\documentclass[a4paper,12pt,twoside,openright]{book}
\usepackage{lipsum} 

\usepackage[T1]{fontenc}
\usepackage{textcomp}

\usepackage[inner=3.2cm,outer=2.2cm,bottom=3.5cm]{geometry}

\usepackage{graphicx} 

\usepackage{booktabs} 
\usepackage{enumitem}
\setlist[enumerate]{label=(\arabic*), leftmargin=*, topsep=0.3em, itemsep=0.2em}

\usepackage{setspace} 
\usepackage{makecell}
\usepackage{listings} 

\usepackage[numbers]{natbib}

\usepackage[table]{xcolor}
\definecolor{codegreen}{rgb}{0,0.6,0}
\definecolor{codegray}{rgb}{0.5,0.5,0.5}
\definecolor{codepurple}{rgb}{0.58,0,0.82}
\definecolor{backcolour}{rgb}{0.98,0.98,0.97}
\lstdefinestyle{custom_style}{
    backgroundcolor=\color{backcolour},
    commentstyle=\color{codegreen},
    keywordstyle=\color{magenta},
    numberstyle=\tiny\color{codegray},
    stringstyle=\color{codepurple},
    basicstyle=\ttfamily\footnotesize,
    breakatwhitespace=false,
    breaklines=true,
    captionpos=b,
    keepspaces=true,
    numbers=left,
    numbersep=5pt,
    showspaces=false,
    showstringspaces=false,
    showtabs=false,
    tabsize=2
}
\lstdefinestyle{source_paper_python}{
    language=Python,
    basicstyle=\ttfamily\small\linespread{1}\selectfont,
    keywordstyle=\color{blue},
    commentstyle=\color{gray},
    stringstyle=\color{red},
    breaklines=true,
    frame=tb
}
\usepackage[
    pagebackref=false,
    breaklinks=true,
    colorlinks,
    bookmarks=true,
    bookmarksopen=true,
    bookmarksnumbered=true,
    citecolor=green
]{hyperref}
\usepackage{bookmark}
\newcommand{\clearemptydoublepage}{\newpage{\pagestyle{empty}\cleardoublepage}}
\usepackage{fancyhdr} 
\fancypagestyle{myfancy}{%
    \fancyhead[LE]{\slshape \nouppercase\rightmark}
    \fancyhead[RE,LO]{}
    \fancyhead[RO]{\slshape \nouppercase\leftmark}
}

\usepackage[nottoc,notlot,notlof]{tocbibind}

\usepackage{amsmath}
\allowdisplaybreaks[3]
\usepackage{amsfonts}
\usepackage{amssymb}
\usepackage{nicefrac}
\usepackage{amsthm}
\usepackage[ruled,vlined]{algorithm2e}
\usepackage{multirow}
\usepackage{wrapfig}
\usepackage{diagbox}
\usepackage{floatrow}
\newfloatcommand{capbtabbox}{table}[][\FBwidth]
\usepackage{tabularx}

\usepackage{mathtools}
\usepackage{mathrsfs}
\usepackage{tasks}
\usepackage{pifont}
\usepackage{multicol}
\usepackage{tikz-cd}
\usepackage{mdframed}
\usepackage{chngcntr}
\usepackage{etoolbox}
\usepackage{xspace}
\usepackage{needspace}
\usepackage{placeins}
\usepackage{subcaption}
\usepackage[capitalise]{cleveref}
\usepackage{crossreftools}
\newcommand{\pdfbookmarkcref}[1]{\crtcrefcounter{#1} \crtrefnumber{#1}}
\usepackage[acronym,toc,hyperfirst=true]{glossaries-extra}

\graphicspath{ {./image/} }

\makeatletter
\@ifundefined{keybox}{%
  \newenvironment{keybox}[1][thesisParisBrown]
    {\begin{parismath}[#1]}
    {\end{parismath}}%
}{}
\makeatother
\makeatletter
\@ifundefined{mymath}{%
}{}
\makeatother

\makeatletter
\@ifundefined{theorem}{}{}
\@ifundefined{thm}{}{}
\@ifundefined{proposition}{}{}
\@ifundefined{prop}{}{}
\@ifundefined{duplicate}{}{}
\@ifundefined{lemma}{}{}
\@ifundefined{dup}{}{}
\@ifundefined{corollary}{}{}
\@ifundefined{definition}{%
  \theoremstyle{definition}
}{}
\@ifundefined{assumption}{%
  \theoremstyle{definition}
}{}
\@ifundefined{example}{%
  \theoremstyle{definition}
}{}
\@ifundefined{remark}{%
  \theoremstyle{remark}
}{}
\theoremstyle{plain}
\makeatother

\makeatletter
\@ifundefined{Hy@raisedlink@left}{%
  \def\Hy@raisedlink@left#1{%
    \ifvmode
      #1%
    \else
      \Hy@SaveSpaceFactor
      \llap{\smash{%
      \begingroup
        \let\HyperRaiseLinkLength\@tempdima
        \setlength\HyperRaiseLinkLength\HyperRaiseLinkDefault
        \HyperRaiseLinkHook
      \expandafter\endgroup
      \expandafter\raise\the\HyperRaiseLinkLength\hbox{%
        \Hy@RestoreSpaceFactor
        #1%
        \Hy@SaveSpaceFactor
      }%
      }}%
      \Hy@RestoreSpaceFactor
      \penalty\@M\hskip\z@
    \fi
  }%
}{}
\makeatother
\robustify{\hyperlink}

\makeatletter
\@ifundefined{newtarget}{%
  \newcommand\newtarget[2]{\Hy@raisedlink@left{\hypertarget{#1}{}}#2}%
}{}
\makeatother
\providecommand\linktoproof[1]{\leavevmode{\normalfont[{\hyperlink{proof:#1}{$\downarrow$}}]}}
\providecommand\linkofproof[1]{\newtarget{proof:#1}}

\usepackage[most]{tcolorbox}

\definecolor{thesisParisBlue}{HTML}{3D91C6}
\definecolor{thesisParisRed}{HTML}{E47B72}
\definecolor{thesisParisGreen}{HTML}{16805D}
\definecolor{thesisParisGold}{HTML}{A88A45}

\definecolor{thesisParisBrown}{RGB}{190,148,110} 

\definecolor{thesisParisBackground}{RGB}{247,249,251} 

\newtcolorbox{parismath}[1][thesisParisBlue]{
  enhanced jigsaw,
  breakable,
  colback=thesisParisBackground,
  colframe=black!18,
  boxrule=0.35pt,
  sharp corners,
  left=8pt,
  right=8pt,
  top=6pt,
  bottom=6pt,
  before skip=6pt,
  after skip=6pt,
  drop fuzzy shadow southeast={black!22},
  overlay unbroken and first={%
    \draw[color=#1,line width=2.2pt]
      (frame.north west) -- ([xshift=2.4cm]frame.north west);%
  }
}

\newcommand{\thesisParisBeginTheorem}[3]{%
  \begin{parismath}[#1]%
  \IfNoValueTF{#3}{\begin{#2}}{\begin{#2}[#3]}%
}

\NewDocumentEnvironment{parisdefinition}{o}
  {\thesisParisBeginTheorem{thesisParisGreen}{definition}{#1}}
  {\end{definition}\end{parismath}}

\NewDocumentEnvironment{parisremark}{o}
  {\thesisParisBeginTheorem{thesisParisRed}{remark}{#1}}
  {\end{remark}\end{parismath}}

\NewDocumentEnvironment{paristheorem}{o}
  {\thesisParisBeginTheorem{thesisParisBlue}{theorem}{#1}}
  {\end{theorem}\end{parismath}}

\NewDocumentEnvironment{paristhm}{o}
  {\thesisParisBeginTheorem{thesisParisBlue}{thm}{#1}}
  {\end{thm}\end{parismath}}

\NewDocumentEnvironment{parisproposition}{o}
  {\thesisParisBeginTheorem{thesisParisBlue}{proposition}{#1}}
  {\end{proposition}\end{parismath}}

\NewDocumentEnvironment{parisprop}{o}
  {\thesisParisBeginTheorem{thesisParisBlue}{prop}{#1}}
  {\end{prop}\end{parismath}}

\NewDocumentEnvironment{parisduplicate}{o}
  {\thesisParisBeginTheorem{thesisParisBlue}{duplicate}{#1}}
  {\end{duplicate}\end{parismath}}

\NewDocumentEnvironment{parislemma}{o}
  {\thesisParisBeginTheorem{thesisParisBlue}{lemma}{#1}}
  {\end{lemma}\end{parismath}}

\NewDocumentEnvironment{parisdup}{o}
  {\thesisParisBeginTheorem{thesisParisBlue}{dup}{#1}}
  {\end{dup}\end{parismath}}

\NewDocumentEnvironment{pariscorollary}{o}
  {\thesisParisBeginTheorem{thesisParisBlue}{corollary}{#1}}
  {\end{corollary}\end{parismath}}

\NewDocumentEnvironment{parisassumption}{o}
  {\thesisParisBeginTheorem{thesisParisBlue}{assumption}{#1}}
  {\end{assumption}\end{parismath}}

\NewDocumentEnvironment{parisexample}{o}
  {\thesisParisBeginTheorem{thesisParisBrown}{example}{#1}}
  {\end{example}\end{parismath}}

\makeatletter
\@ifundefined{gls@notation@type}{\newglossary*{notation}{List of Symbols}}{}
\makeatother
\setglossarystyle{long}
\setabbreviationstyle[acronym]{long-short}
\providecommand{\thesisnewacronym}[4][]{%
  \ifglsentryexists{#2}{}{\newacronym[#1]{#2}{#3}{#4}}%
}

\thesisnewacronym[sort=nn]{CNNs}{CNNs}{Convolutional Neural Networks}
\thesisnewacronym[sort=nn]{RNNs}{RNNs}{Recurrent Neural Networks}
\thesisnewacronym[sort=nn]{DNNs}{DNNs}{Deep Neural Networks}
\thesisnewacronym[sort=nn]{GCN}{GCN}{Graph Convolutional Network}
\thesisnewacronym[sort=nn]{FC}{FC}{Fully Connected}
\thesisnewacronym[sort=nn]{MLR}{MLR}{Multinomial Logistic Regression}
\thesisnewacronym[sort=nn]{RMLR}{RMLR}{Riemannian Multinomial Logistic Regression}
\thesisnewacronym[sort=nn]{RMLRs}{RMLRs}{Riemannian Multinomial Logistic Regressions}
\thesisnewacronym[sort=nn]{BMLR}{BMLR}{Busemann Multinomial Logistic Regression}
\thesisnewacronym[sort=nn]{BMLRs}{BMLRs}{Busemann Multinomial Logistic Regressions}
\thesisnewacronym[sort=nn]{BFC}{BFC}{Busemann Fully Connected}
\thesisnewacronym[sort=nn]{SPDNet}{SPDNet}{SPD Neural Network}
\thesisnewacronym[sort=nn]{GrNet}{GrNet}{Grassmann network}
\thesisnewacronym[sort=nn]{GrConv}{GrConv}{Grassmannian Convolution}
\thesisnewacronym[sort=nn]{HNN}{HNN}{Hyperbolic Neural Network}
\thesisnewacronym[sort=nn]{HNNs}{HNNs}{Hyperbolic Neural Networks}
\thesisnewacronym[sort=nn]{HNN++}{HNN++}{Hyperbolic Neural Network++}
\thesisnewacronym[sort=nn]{HBNN}{HBNN}{Hyperbolic Busemann Neural Network}
\thesisnewacronym[sort=nn]{HBNNs}{HBNNs}{Hyperbolic Busemann Neural Networks}
\thesisnewacronym[sort=nn]{CorNet}{CorNet}{Correlation Network}
\thesisnewacronym[sort=nn]{CorNets}{CorNets}{Correlation Networks}

\thesisnewacronym[sort=bn]{BN}{BN}{Batch Normalization}
\thesisnewacronym[sort=bn]{RBN}{RBN}{Riemannian Batch Normalization}
\thesisnewacronym[sort=bn]{LieBN}{LieBN}{Lie Group Batch Normalization}
\thesisnewacronym[sort=bn]{SPDBN}{SPDBN}{SPD Batch Normalization}
\thesisnewacronym[sort=bn]{GyroBN}{GyroBN}{Gyrogroup Batch Normalization}

\thesisnewacronym[sort=spd]{SPD}{SPD}{Symmetric Positive Definite}
\thesisnewacronym[sort=spd]{AIM}{AIM}{Affine-Invariant Metric}
\thesisnewacronym[sort=spd]{LCM}{LCM}{Log-Cholesky Metric}
\thesisnewacronym[sort=spd]{LEM}{LEM}{Log-Euclidean Metric}
\thesisnewacronym[sort=spd]{PEM}{PEM}{Power-Euclidean Metric}
\thesisnewacronym[sort=spd]{BWM}{BWM}{Bures--Wasserstein Metric}
\thesisnewacronym[sort=spd]{GBWM}{GBWM}{Generalized Bures--Wasserstein Metric}
\thesisnewacronym[sort=spd]{ALEM}{ALEM}{Adaptive Log-Euclidean Metric}
\thesisnewacronym[sort=spd]{ALEMs}{ALEMs}{Adaptive Log-Euclidean Metrics}
\thesisnewacronym[sort=spd]{BWCM}{BWCM}{Bures--Wasserstein--Cholesky Metric}
\thesisnewacronym[sort=spd]{PCM}{PCM}{Power-Cholesky Metric}
\thesisnewacronym[sort=gr]{ONB}{ONB}{Orthonormal Basis}
\thesisnewacronym[sort=gr]{PP}{PP}{Projector Perspective}
\thesisnewacronym[sort=cor]{ECM}{ECM}{Euclidean--Cholesky Metric}
\thesisnewacronym[sort=cor]{LECM}{LECM}{Log-Euclidean--Cholesky Metric}
\thesisnewacronym[sort=cor]{LSM}{LSM}{Log-Scaled Metric}
\thesisnewacronym[sort=cor]{OLM}{OLM}{Off-Log Metric}
\thesisnewacronym[sort=cor]{PHCM}{PHCM}{Poly-Hyperbolic-Cholesky Metric}

\thesisnewacronym[sort=h]{PV}{PV}{Proper Velocity}
\thesisnewacronym[sort=h]{PVNN}{PVNN}{Proper Velocity Neural Network}
\thesisnewacronym[sort=h]{PVNNs}{PVNNs}{Proper Velocity Neural Networks}
\thesisnewacronym[sort=h]{PVCNN}{PVCNN}{Proper Velocity Convolutional Network}
\thesisnewacronym[sort=nn]{TFC}{TFC}{Tangent Fully Connected}
\thesisnewacronym[sort=bn]{TBN}{TBN}{Tangent Batch Normalization}
\thesisnewacronym[sort=score]{MCC}{MCC}{Matthews Correlation Coefficient}
\thesisnewacronym[sort=data]{TEB}{TEB}{Transposable Elements Benchmark}

\ifglsentryexists{manifold}{}{%
  \newglossaryentry{manifold}{%
    type=notation,
    name={$\mathcal{M}$},
    text={$\mathcal{M}$},
    description={Riemannian manifold}%
  }%
}
\ifglsentryexists{metric}{}{%
  \newglossaryentry{metric}{%
    type=notation,
    name={$\mathbf{g}$},
    text={$\mathbf{g}$},
    description={Metric on $\mathcal{M}$}%
  }%
}
\ifglsentryexists{reals}{}{%
  \newglossaryentry{reals}{%
    type=notation,
    name={$\mathbb{R}^n$},
    text={$\mathbb{R}^n$},
    description={A $n$-dimensional Euclidean space}%
  }%
}

\providecommand{\calB}{\mathcal{B}}
\providecommand{\calF}{\mathcal{F}}

\providecommand{\calH}{\mathcal{H}}
\providecommand{\calI}{\mathcal{I}}
\providecommand{\calL}{\mathcal{L}}
\providecommand{\calM}{\mathcal{M}}
\providecommand{\calN}{\mathcal{N}}
\providecommand{\calO}{\mathcal{O}}
\providecommand{\calX}{\mathcal{X}}

\providecommand{\wcalM}{\widetilde{\mathcal{M}}}

\providecommand{\bbA}{\mathbb{A}}

\providecommand{\bbD}{\mathbb{D}}
\providecommand{\bbJ}{\mathbb{J}}
\providecommand{\bbK}{\mathbb{K}}
\providecommand{\bbL}{\mathbb{L}}

\providecommand{\bbM}{\mathbb{M}}

\providecommand{\bbP}{\mathbb{P}}
\providecommand{\bbR}[1]{\mathbb{R}^{#1}}
\providecommand{\bbRplus}{\mathbb{R}_{+}}
\providecommand{\bbRplusscalar}{\mathbb{R}_{++}}
\providecommand{\bbRscalar}{\mathbb{R}}
\providecommand{\bbV}{\mathbb{V}}

\providecommand{\bbX}{\mathbb{X}}
\providecommand{\bbY}{\mathbb{Y}}
\providecommand{\bbZ}{\mathbb{Z}}

\providecommand{\dist}{\operatorname{d}}
\providecommand{\geodesic}[2]{\gamma_{({#1},{#2})}}
\providecommand{\rieexp}{\operatorname{Exp}}
\providecommand{\rielog}{\operatorname{Log}}
\providecommand{\rieExp}[1]{\operatorname{Exp}_{#1}}

\providecommand{\pt}[2]{\operatorname{PT}_{#1 \rightarrow #2}}
\providecommand{\vt}[2]{\mathcal{T}_{#1 \rightarrow #2}}
\providecommand{\diff}{\operatorname{d}}
\providecommand{\diffphi}[1]{\phi_{*,#1}}
\providecommand{\diffphiinv}[1]{\phiinv_{*,#1}}

\providecommand{\fm}{\operatorname{FM}}
\providecommand{\wfm}{\operatorname{WFM}}
\providecommand{\barcenter}{\operatorname{Bar}}

\providecommand{\argmin}{\operatorname{argmin}}
\providecommand{\argmax}{\operatorname{argmax}}
\providecommand{\arccosh}{\operatorname{arccosh}}

\providecommand{\cinf}{C^{\infty}}
\providecommand{\concat}{\mathrm{concat}}

\providecommand{\diag}{\operatorname{diag}}

\providecommand{\dlog}{\operatorname{Dlog}}
\providecommand{\mexp}{\operatorname{exp}}
\providecommand{\mlog}{\operatorname{log}}
\providecommand{\fnorm}[1]{\|{#1}\|_\mathrm{F}}

\providecommand{\inv}{\operatorname{inv}}
\providecommand{\ln}{\operatorname{ln}}
\providecommand{\mle}{\mathrm{MLE}}
\providecommand{\pow}{\operatorname{P}}
\providecommand{\rank}{\operatorname{rank}}

\providecommand{\sech}{\operatorname{sech}}
\providecommand{\sgn}{\operatorname{sgn}}
\providecommand{\sign}{\operatorname{sign}}
\providecommand{\softmax}{\mathrm{softmax}}

\providecommand{\symmetrize}[1]{\left(#1 \right)_{\mathrm{sym}}}
\providecommand{\symmetrizeSum}[1]{\left(#1 \right)_{\mathrm{sym}+}}
\providecommand{\tr}{\operatorname{tr}}

\providecommand{\bfst}{\mathbf{ST}}
\providecommand{\bbzero}{\mathbf{0}}
\providecommand{\id}{\operatorname{id}}

\providecommand{\mrL}{\mathrm{L}}

\providecommand{\na}{\textcolor{gray}{N/A}}
\providecommand{\rmE}{\mathrm{E}}
\providecommand{\rmF}{\mathrm{F}}

\providecommand{\rmpt}{\mathrm{PT}}

\providecommand{\st}{\mathrm{s.t.}}
\providecommand{\vecone}{\boldsymbol{1}}

\providecommand{\phiinv}{\phi^{-1}}
\providecommand{\phiMul}{\odot_{\phi}}
\providecommand{\phiMulScalar}{\circledast_{\phi}}
\providecommand{\gphi}{g^{\phi}}

\providecommand{\clog}{\psi_{\mathrm{LC}}}

\providecommand{\clnchart}{\varphi_{ln}}

\providecommand{\GL}[1]{\mathrm{GL}({#1})}
\providecommand{\SL}[1]{\operatorname{SL}({#1})}
\providecommand{\liebn}{\operatorname{LieBN}}
\providecommand{\gleft}{g^{\mathrm{L}}}
\providecommand{\gright}{g^{\mathrm{R}}}
\providecommand{\ltrans}{\operatorname{L}}
\providecommand{\rtrans}{\operatorname{R}}
\providecommand{\orth}[1]{\mathrm{O}({#1})}

\providecommand{\so}[1]{\mathrm{SO}(#1)}
\providecommand{\soLieAlgebra}[1]{\mathfrak{so}(#1)}
\providecommand{\soprod}[2]{\mathrm{SO}^{#1}(#2)}
\providecommand{\spdtrans}[2]{\Gamma_{#1 \rightarrow #2}}
\providecommand{\stiefel}[1]{\mathrm{St}(#1)}

\providecommand{\CAT}{\mathrm{CAT}}

\providecommand{\inner}[2]{\left\langle #1,#2 \right\rangle}

\providecommand{\norm}[1]{\left\| #1 \right\|}
\providecommand{\zerovec}{\mathbf{0}}
\providecommand{\rzero}[1]{\mathrm{Row}_0({#1})}
\providecommand{\rone}[1]{\mathrm{Row}^+_1({#1})}
\providecommand{\Rone}{\mathbf{1}}

\providecommand{\calMK}[1]{\mathcal{M}_{K}^{#1}}

\providecommand{\Kinner}[2]{\left\langle #1, #2 \right\rangle_{K}}
\providecommand{\Knorm}[1]{\norm{#1}_{K}}
\providecommand{\MKzero}{\overline{\mathbf{0}}}

\providecommand{\MKoplus}{\oplus^\mathcal{M}_K}
\providecommand{\MKominus}{\ominus^\mathcal{M}_K}
\providecommand{\MKodot}{\odot^\mathcal{M}_K}
\providecommand{\stereo}[1]{\mathfrak{st}_{K}^{#1}}
\providecommand{\stoplus}{\oplus_{K}}
\providecommand{\stominus}{\ominus_{K}}
\providecommand{\stodot}{\odot_{K}}
\providecommand{\tank}{\tan_{K}}
\providecommand{\sink}{\sin_{K}}
\providecommand{\cosk}{\cos_{K}}
\providecommand{\isoSTMK}[1]{\pi_{\stereo{#1} \to \calMK{#1}}}
\providecommand{\isoMKST}[1]{\pi_{\calMK{#1} \to \stereo{#1}}}

\providecommand{\gyr}{\operatorname{gyr}}
\providecommand{\gyrinner}[2]{\left\langle #1, #2 \right\rangle_{\mathrm{gyr}}}
\providecommand{\gyrnorm}[1]{\left\| #1 \right\|_{\mathrm{gyr}}}
\providecommand{\gyrdist}{\mathrm{d}_{\mathrm{gyr}}}
\providecommand{\gyrw}{\widetilde{\operatorname{gyr}}}

\providecommand{\gyrnormw}[1]{\left\| #1 \right\|_{\widetilde{\mathrm{gyr}}}}
\providecommand{\gyrdistw}{\mathrm{d}_{\widetilde{\mathrm{gyr}}}}

\providecommand{\hyperspace}[1]{\mathcal{H}^{#1}_{K}}
\providecommand{\Hoplus}{\oplus_\mathcal{H}}

\providecommand{\Hotimes}{\otimes_\mathcal{H}}

\providecommand{\PB}{\mathbb{P}}
\providecommand{\pball}[1]{\mathbb{P}^{#1}_{K}}

\providecommand{\Moplus}{\oplus_\mathrm{M}}
\providecommand{\Mominus}{\ominus_\mathrm{M}}
\providecommand{\Modot}{\odot_\mathrm{M}}

\providecommand{\Mgyr}{\gyr_{\mathrm{M}}}
\providecommand{\unitpball}[1]{\mathbb{P}^{#1}}

\providecommand{\klein}[1]{\mathbb{K}^{#1}_K}

\providecommand{\Eoplus}{\oplus_\mathrm{E}}
\providecommand{\Eominus}{\ominus_\mathrm{E}}
\providecommand{\Eodot}{\odot_\mathrm{E}}
\providecommand{\unitklein}[1]{\mathbb{K}^{#1}}

\providecommand{\lorentz}[1]{\mathbb{L}^{#1}_K}
\providecommand{\unitlorentz}[1]{\mathbb{L}^{#1}}
\providecommand{\Linner}[2]{\left\langle #1, #2 \right\rangle_{\mathcal{L}}}
\providecommand{\Lnorm}[1]{\left\| #1 \right\|_{\mathcal{L}}}
\providecommand{\Lzero}{\overline{\mathbf{0}}}
\providecommand{\Loplus}{\oplus_{\mathbb{L}}}
\providecommand{\Lominus}{\ominus_{\mathbb{L}}}
\providecommand{\Lodot}{\odot_{\mathbb{L}}}

\providecommand{\PV}{\mathbb{PV}}
\providecommand{\PVspace}[1]{\mathbb{PV}^{#1}_{K}}
\providecommand{\PVoplus}{\oplus_{\mathrm{U}}}
\providecommand{\PVominus}{\ominus_{\mathrm{U}}}
\providecommand{\PVotimes}{\otimes_{\mathrm{U}}}
\providecommand{\PVtoPB}{\pi_{\PVspace{n} \to \pball{n}}}
\providecommand{\PBtoPV}{\pi_{\pball{n} \to \PVspace{n}}}
\providecommand{\PVtoL}{\pi_{\PVspace{n} \to \lorentz{n}}}
\providecommand{\LtoPV}{\pi_{\lorentz{n} \to \PVspace{n}}}
\providecommand{\PballtoL}{\pi_{\pball{n} \to \lorentz{n}}}
\providecommand{\LtoPball}{\pi_{\lorentz{n} \to \pball{n}}}

\providecommand{\hs}[1]{\mathrm{H}\mathbb{S}^{#1}}
\providecommand{\projhs}[1]{\mathbb{D}^{#1}_{K}}
\providecommand{\sphere}[1]{\mathbb{S}_{K}^{#1}}
\providecommand{\unitsphere}[1]{\mathbb{S}^{#1}}
\providecommand{\unitprojhs}[1]{\mathbb{D}^{#1}}
\providecommand{\bbPHS}[1]{\mathbb{PH}\mathrm{S}^{#1}}

\providecommand{\bbPPB}[1]{\mathbb{PP}^{#1}}

\providecommand{\sym}[1]{\mathcal{S}^{#1}}
\providecommand{\spd}[1]{\mathcal{S}^{#1}_{++}}
\providecommand{\spsd}[1]{\mathcal{S}_{#1}^{+}}
\providecommand{\chospace}[1]{\mathcal{L}_{++}^{#1}}
\providecommand{\chol}{\operatorname{Chol}}
\providecommand{\lyp}{\mathcal{L}}
\providecommand{\tildept}[2]{\widetilde{\mathrm{PT}}_{{#1} \rightarrow {#2}}}

\providecommand{\bbDspace}[1]{\mathrm{Diag}({#1})}
\providecommand{\bbDplus}[1]{\mathrm{Diag}^{+}({#1})}
\providecommand{\dplus}{\mathcal{D}}
\providecommand{\dstar}{\mathcal{D}^\star}
\providecommand{\diagvec}{\mathrm{Dv}}

\providecommand{\trilAk}{\lfloor A_k \rfloor}
\providecommand{\trilJ}{\lfloor J \rfloor}
\providecommand{\trilK}{\lfloor K \rfloor}
\providecommand{\trilL}{\lfloor L \rfloor}
\providecommand{\trilLi}{\lfloor L_i \rfloor}
\providecommand{\trilLk}{\lfloor L_k \rfloor}

\providecommand{\trilX}{\lfloor X \rfloor}
\providecommand{\trilY}{\lfloor Y \rfloor}

\providecommand{\AI}{\mathrm{AI}}
\providecommand{\BW}{\mathrm{BW}}

\providecommand{\LC}{\mathrm{LC}}
\providecommand{\LE}{\mathrm{LE}}
\providecommand{\PE}{\mathrm{PE}}
\providecommand{\lem}{\mathrm{LEM}}
\providecommand{\lcm}{\mathrm{LCM}}
\providecommand{\BWM}{\text{BWM}}
\providecommand{\LCM}{\text{LCM}}
\providecommand{\PEM}{\theta\text{-EM}}
\providecommand{\GBWM}{M\text{-BWM}}
\providecommand{\alphabeta}{(\alpha,\beta)}

\providecommand{\biparam}{(a,b)}
\providecommand{\biparamEM}{(\alpha,\beta)\text{-EM}}
\providecommand{\biparamAIM}{(\alpha,\beta)\text{-AIM}}
\providecommand{\biparamLEM}{(\alpha,\beta)\text{-LEM}}
\providecommand{\biparamALEM}{(a,b)\text{-ALEM}}
\providecommand{\oplusale}{\oplus^{\mathrm{ALE}}}
\providecommand{\odotale}{\odot^{\mathrm{ALE}}}

\providecommand{\triparamEM}{(\theta,\alpha,\beta)\text{-EM}}
\providecommand{\triparamAIM}{(\theta,\alpha,\beta)\text{-AIM}}
\providecommand{\triparamLEM}{(\theta,\alpha,\beta)\text{-LEM}}
\providecommand{\paramBWM}{2\theta\text{-BWM}}
\providecommand{\paramLCM}{\theta\text{-LCM}}

\providecommand{\defDPM}{\theta\text{-DPM}}
\providecommand{\defDBWM}{\theta\text{-DBWM}}
\providecommand{\defDGBWM}{(\theta,\mathbb{M})\text{-DBWM}}
\providecommand{\defCDEM}{\theta\text{-PCM}}
\providecommand{\defCDBWM}{\theta\text{-BWCM}}
\providecommand{\defCDGBWM}{(\theta,\mathbb{M})\text{-BWCM}}

\providecommand{\DGBWM}{\mathbb{M}\text{-DBWM}}

\providecommand{\gAI}{g^{\mathrm{AI}}}
\providecommand{\gBW}{g^{\mathrm{BW}}}

\providecommand{\gDGBW}{g^{\bbM\text{-DBW}}}

\providecommand{\gDL}{g^{\text{DL}}}

\providecommand{\gE}{g^\mathrm{E}}

\providecommand{\gGBWscalar}{g^{m\text{-BW}}}
\providecommand{\gLC}{g^{\text{LC}}}
\providecommand{\gLE}{g^{\mathrm{LE}}}
\providecommand{\gPE}{g^{\theta\text{-E}}}
\providecommand{\gRplus}{g^{\mathbb{R}_{++}}}
\providecommand{\galem}{g^{\mathrm{ALE}}}

\providecommand{\gbiparamai}{g^{(\alpha,\beta)\text{-AI}}}

\providecommand{\gbiparamLE}{g^{(\alpha,\beta)\text{-LE}}}

\providecommand{\gbiparamalem}{g^{(a,b)\text{-ALE}}}
\providecommand{\gbiparamaE}{g^{(a,b)\text{-E}}}

\providecommand{\gcm}{g^{\mathrm{C}}}
\providecommand{\gcri}{g^{\mathrm{CRI}}}

\providecommand{\gdefDBW}{g^{\theta\text{-DBW}}}
\providecommand{\gdefDE}{g^{\theta\text{-DE}}}
\providecommand{\gdefcri}{g^{\theta\text{-CRI}}}
\providecommand{\gdefDGBW}{g^{(\theta,\bbM)\text{-DBW}}}

\providecommand{\geuc}{g^{\mathrm{E}}}

\providecommand{\glcm}{g^{\mathrm{LC}}}

\providecommand{\gparamBWM}{g^{2\theta\text{-BWM}}}

\providecommand{\gtriparamAI}{g^{(\theta,\alpha,\beta)\text{-AI}}}

\providecommand{\gtriparamEM}{g^{(\theta,\alpha,\beta)\text{-EM}}}
\providecommand{\gtriparamLE}{g^{(\theta,\alpha,\beta)\text{-LE}}}

\providecommand{\dalem}{d^{\mathrm{ALE}}}

\providecommand{\dphi}{d^{\phi}}
\providecommand{\dpow}{\operatorname{DPow}}

\providecommand{\odotai}{\odot^{\mathrm{AI}}}
\providecommand{\odotle}{\odot^{\mathrm{LE}}}
\providecommand{\odotlc}{\odot^{\mathrm{LC}}}

\providecommand{\oplusLieAI}{\oplus^{\mathrm{LieAI}}}
\providecommand{\oplusLieLE}{\oplus^{\mathrm{LE}}}
\providecommand{\oplusLieLC}{\oplus^{\mathrm{LC}}}
\providecommand{\oplusLiePAI}{\oplus^{\theta\text{-AI}}}
\providecommand{\oplusLiePLC}{\oplus^{\theta\text{-LC}}}
\providecommand{\ominusLieAI}{\ominus^{\mathrm{LieAI}}}
\providecommand{\oplusGyrAI}{\oplus^{\mathrm{AI}}}
\providecommand{\goplus}{\oplus_{g}}
\providecommand{\gominus}{\ominus_{g}}
\providecommand{\spsdoplus}{\oplus_{psd,g}}
\providecommand{\spsdominus}{\ominus_{psd,g}}

\providecommand{\oplusChol}{\oplus^\mathcal{C}}
\providecommand{\ominusChol}{\ominus^\mathcal{C}}
\providecommand{\odotChol}{\odot^\mathcal{C}}

\providecommand{\frakU}{\mathfrak{U}}
\providecommand{\grasonb}[1]{\mathrm{Gr}(#1)}
\providecommand{\graspp}[1]{\widetilde{\mathrm{Gr}}(#1)}
\providecommand{\grassoplus}{\oplus_{gr}}
\providecommand{\grassominus}{\ominus_{gr}}

\providecommand{\idonb}{I_{p,n}}
\providecommand{\idpp}{\widetilde{I}_{p,n}}

\providecommand{\oplusGyrONB}{\oplus^{\mathrm{Gr}}}
\providecommand{\oplusGyrPP}{\widetilde{\oplus}^{\mathrm{Gr}}}
\providecommand{\ominusGyrONB}{\ominus^{\mathrm{Gr}}}
\providecommand{\ominusGyrPP}{\widetilde{\ominus}^{\mathrm{Gr}}}
\providecommand{\odotGyrONB}{\odot^{\mathrm{Gr}}}
\providecommand{\odotGyrPP}{\widetilde{\odot}^{\mathrm{Gr}}}
\providecommand{\rp}[1]{\mathbb{RP}^{#1}}

\providecommand{\cor}[1]{\mathrm{Cor}^{+}({#1})}
\providecommand{\coropt}{\operatorname{Cor}}

\providecommand{\chocor}[1]{\mathcal{L}^{#1}}
\providecommand{\LTone}[1]{\mathrm{LT}^1(#1)}
\providecommand{\LTzero}[1]{\mathrm{LT}^0(#1)}
\providecommand{\trilspace}[1]{\mathrm{LT}^{#1}}
\providecommand{\EC}{\mathrm{EC}}
\providecommand{\LEC}{\mathrm{LEC}}
\providecommand{\OL}{\mathrm{OL}}
\providecommand{\LS}{\mathrm{LS}}

\providecommand{\isoecm}{\phi^{\mathrm{EC}}}

\providecommand{\singperm}[1]{\mathfrak{S}^{\pm}(#1)}
\providecommand{\perm}[1]{\mathfrak{S}^{#1}}
\providecommand{\hol}[1]{\mathrm{Hol}({#1})}

\providecommand{\off}{\mathrm{off}}
\providecommand{\holinner}[2]{\left\langle #1, #2 \right\rangle^{(\alpha,\beta,\gamma)}}
\providecommand{\holnorm}[1]{\left\| #1 \right\|^{(\alpha,\beta,\gamma)}}
\providecommand{\offlog}{\operatorname{Log}^{\circ}}
\providecommand{\offexp}{\operatorname{Exp}^{\circ}}
\providecommand{\Sum}{\operatorname{Sum}}

\providecommand{\rzeroinner}[2]{\left\langle #1, #2 \right\rangle^{(\alpha,\delta,\zeta)}}
\providecommand{\rzeronorm}[1]{\left\| #1 \right\|^{(\alpha,\delta,\zeta)}}
\providecommand{\logscaled}{\operatorname{Log}^{\star}}
\providecommand{\expscaled}{\operatorname{Exp}^{\star}}

\providecommand{\red}[1]{\textcolor{red}{#1}}
\providecommand{\firstresults}[1]{{\boldmath\textbf{#1}}}
\providecommand{\redbf}[1]{\textbf{\textcolor{red}{#1}}}
\providecommand{\bluebf}[1]{\textbf{\textcolor{blue}{#1}}}
\providecommand{\cyanbf}[1]{\textbf{\textcolor{cyan}{#1}}}
\providecommand{\greenbf}[1]{\textbf{\textcolor{green}{#1}}}

\providecommand{\mypara}[1]{\textbf{#1}}

\providecommand{\ie}{\emph{i.e.},\xspace}
\providecommand{\eg}{\emph{e.g.},\xspace}

\providecommand{\cmark}{\textcolor{green}{\text{\ding{51}}}}
\providecommand{\xmark}{\textcolor{red}{\text{\ding{55}}}}
\providecolor{HilightColor}{RGB}{220,255,240}
\providecolor{RowGray}{gray}{0.925}

\makeatletter
\@ifundefined{supressingmaincontents}{%
}{}
\makeatother

\crefname{equation}{Eq.}{Eqs.}
\Crefname{equation}{Eq.}{Eqs.}
\crefname{figure}{Fig.}{Figs.}
\Crefname{figure}{Fig.}{Figs.}
\crefname{table}{Tab.}{Tabs.}
\Crefname{table}{Tab.}{Tabs.}
\crefname{algocf}{Alg.}{Algs.}
\Crefname{algocf}{Alg.}{Algs.}
\crefname{algorithm}{Alg.}{Algs.}
\Crefname{algorithm}{Alg.}{Algs.}
\crefname{section}{Sec.}{Secs.}
\Crefname{section}{Sec.}{Secs.}
\crefname{appendix}{App.}{Apps.}
\Crefname{appendix}{App.}{Apps.}
\crefname{subsubsubappendix}{App.}{Apps.}
\Crefname{subsubsubappendix}{App.}{Apps.}
\crefname{theorem}{Thm.}{Thms.}
\Crefname{theorem}{Thm.}{Thms.}
\crefname{thm}{Thm.}{Thms.}
\Crefname{thm}{Thm.}{Thms.}
\crefname{lemma}{Lem.}{Lems.}
\Crefname{lemma}{Lem.}{Lems.}
\crefname{dup}{Lem.}{Lems.}
\Crefname{dup}{Lem.}{Lems.}
\crefname{definition}{Def.}{Defs.}
\Crefname{definition}{Def.}{Defs.}
\crefname{corollary}{Cor.}{Cors.}
\Crefname{corollary}{Cor.}{Cors.}
\crefname{assumption}{Assump.}{Assumps.}
\Crefname{assumption}{Assump.}{Assumps.}
\crefname{example}{Ex.}{Exs.}
\Crefname{example}{Ex.}{Exs.}
\crefname{remark}{Rmk.}{Rmks.}
\Crefname{remark}{Rmk.}{Rmks.}
\crefname{proposition}{Prop.}{Props.}
\Crefname{proposition}{Prop.}{Props.}
\crefname{prop}{Prop.}{Props.}
\Crefname{prop}{Prop.}{Props.}
\crefname{duplicate}{Prop.}{Props.}
\Crefname{duplicate}{Prop.}{Props.}
\crefname{proof}{Pr.}{Prs.}
\Crefname{proof}{Pr.}{Prs.}
\crefname{enumi}{Case}{Cases}
\Crefname{enumi}{Case}{Cases}

\begin{document}
\thispagestyle{empty}

\begin{center}
    {
    \setlength\intextsep{0pt}
    \begin{figure}[h!]
        \centering
        \includegraphics[width=0.6\textwidth,trim={0cm 0cm 0cm 0cm}]{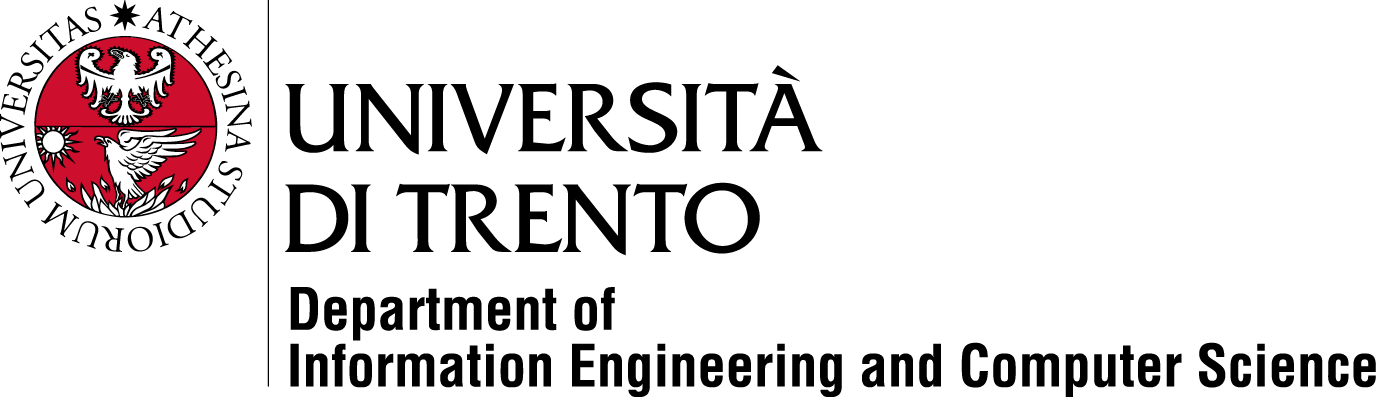}
        \label{fig:university-logo}
    \end{figure}
    }

    \par\noindent\rule{\textwidth}{0.1pt}

    \vspace{0.5 cm}  

    \large\textbf{Doctoral School in} \\
    \large\textbf{Information and Communication Technology}

    \vspace{2 cm}
    
    \Huge\textsc{Riemannian Deep Learning:\\Modules, Networks, and Geometries\\}

    \vspace{2 cm}

    \huge{Ziheng Chen}

\end{center}

\vspace{1.3 cm}

\begin{tabular}{ll}
\multicolumn{2}{l}{\large Advisor}\\
 & \large Prof. Nicu Sebe\\
 & \large Universit\`a di Trento\\
\end{tabular}

\vspace{0.5cm}

\begin{tabular}{ll}
\multicolumn{2}{l}{\large Co-advisor}\\
 & \large Prof. Xiao-Jun Wu\\
 & \large Jiangnan University\\
\end{tabular}

\vspace{0.5cm}

\begin{tabular}{ll}
\multicolumn{2}{l}{\large ELLIS Co-advisor}\\
 & \large Prof. Bernhard Sch\"olkopf\\
 & \large Max Planck Institute for Intelligent Systems\\
\end{tabular}

\begin{center}
    \vspace{1cm}
    \hrule
    \vspace{4pt}
    September 2026
\end{center}

\clearemptydoublepage

\frontmatter
\singlespacing
    \newpage
\section*{Publications}

\begin{small}
\textit{($\dagger$ corresponding author, $\ddagger$ equal contribution, $\clubsuit$ equal supervision)}
\end{small}

\begingroup
\emergencystretch=2em

The dissertation is based on the following publications:

\begin{itemize}
    \item \cref{chapter:normalization}:\\
    {[1]} \textbf{Ziheng Chen}, Yue Song, Yunmei Liu, and Nicu Sebe. ``\href{https://openreview.net/forum?id=okYdj8Ysru}{A Lie Group Approach to Riemannian Batch Normalization}.'' ICLR 2024.\\
    {[2]} \textbf{Ziheng Chen}, Yue Song, Xiao-Jun Wu, and Nicu Sebe. ``\href{https://openreview.net/forum?id=d1NWq4PjJW}{Gyrogroup Batch Normalization}.'' ICLR 2025.
    \item \cref{chapter:rmlr}:\\
    {[3]} \textbf{Ziheng Chen}, Yue Song, Gaowen Liu, Ramana Rao Kompella, Xiao-Jun Wu, and Nicu Sebe. ``\href{https://arxiv.org/abs/2305.11288}{Riemannian Multinomial Logistics Regression for SPD Neural Networks}.'' CVPR 2024.\\
    {[4]} \textbf{Ziheng Chen}, Yue Song, Rui Wang, Xiao-Jun Wu, and Nicu Sebe. ``\href{https://arxiv.org/abs/2409.19433}{RMLR: Extending Multinomial Logistic Regression into General Geometries}.'' NeurIPS 2024.
    \item \cref{chapter:riemannian-neural-networks}:\\
    {[5]} \textbf{Ziheng Chen}$^\ddagger$, Zihan Su$^\ddagger$, Bernhard Sch\"olkopf, and Nicu Sebe. ``\href{https://openreview.net/forum?id=UDIYU1X3vC}{Proper Velocity Neural Networks}.'' ICLR 2026.\\
    {[6]} \textbf{Ziheng Chen}, Bernhard Sch\"olkopf, and Nicu Sebe. ``\href{https://arxiv.org/abs/2602.18858}{Hyperbolic Busemann Neural Networks}.'' CVPR 2026.\\
    {[7]} \textbf{Ziheng Chen}, Xiao-Jun Wu, Bernhard Sch\"olkopf, and Nicu Sebe. ``\href{https://arxiv.org/abs/2605.19073}{Riemannian Networks over Full-Rank Correlation Matrices}.'' ICML 2026.
    \item \cref{chapter:spd-geometries}:\\
    {[8]} \textbf{Ziheng Chen}, Yue Song, Tianyang Xu, Zhiwu Huang, Xiao-Jun Wu, and Nicu Sebe. ``\href{https://arxiv.org/abs/2303.15477}{Adaptive Log-Euclidean Metrics for SPD Matrix Learning}.'' IEEE TIP 2024.\\
    {[9]} \textbf{Ziheng Chen}, Yue Song, Xiao-Jun Wu, and Nicu Sebe. ``\href{https://openreview.net/forum?id=5S8ruWKe8l}{Fast and Stable Riemannian Metrics on SPD Manifolds via Cholesky Product Geometry}.'' ICLR 2026.
\end{itemize}

The following papers are published but are not included in this dissertation:
\begin{enumerate}[start=10]
    \item \textbf{Ziheng Chen}, Yue Song, Xiao-Jun Wu, Gaowen Liu, and Nicu Sebe. ``\href{https://openreview.net/forum?id=q1t0Lmvhty}{Understanding Matrix Function Normalizations in Covariance Pooling through the Lens of Riemannian Geometry}.'' ICLR 2025.
    \item Shaocheng Jin, Tao Zhou, Rui Wang, \textbf{Ziheng Chen}, Xiaoqing Luo, Xiao-Jun Wu, and Josef Kittler. ``Towards Robust EEG Decoding Based on Riemannian Self-Attention.'' KDD 2026.
    \item Rui Wang, Zihao Bi, Chen Hu, Xiaoning Song, Xiao-Jun Wu, Nicu Sebe, and \textbf{Ziheng Chen}$^\dagger$. ``Riemannian Graph Convolutional Network for Skeleton-Based Two-Person Interaction Recognition.'' IJCAI 2026.
    \item Xianglong Shi$^\ddagger$, \textbf{Ziheng Chen}$^{\dagger,\ddagger}$, Yunhan Jiang, and Nicu Sebe. ``\href{https://openreview.net/forum?id=NNnkLi1ALt}{Intrinsic Lorentz Neural Network}.'' ICLR 2026.
    \item Shanglin Li, Shiwen Chu, Okan Ko\c{c}, Yi Ding, Qibin Zhao, Motoaki Kawanabe, and \textbf{Ziheng Chen}$^\dagger$. ``\href{https://openreview.net/forum?id=CNDNRjpVIL}{HEEGNet: Hyperbolic Embeddings for EEG}.'' ICLR 2026.
    \item Chen Hu$^\ddagger$, \textbf{Ziheng Chen}$^\ddagger$, Rui Wang, Yefeng Zheng, and Nicu Sebe. ``\href{https://openreview.net/forum?id=66h1sCMm7F}{Riemannian High-Order Pooling for Brain Foundation Models}.'' ICLR 2026.
    \item Rui Wang, Yuting Jiang, Xiaoqing Luo, Xiao-Jun Wu, Nicu Sebe, and \textbf{Ziheng Chen}$^\dagger$. ``\href{https://www.arxiv.org/abs/2512.09402}{Wasserstein-Aligned Hyperbolic Multi-View Clustering}.'' AAAI 2026 (\textbf{Oral}).
    \item Rui Wang, Chen Hu, Xiaoning Song, Xiao-Jun Wu, Nicu Sebe, and \textbf{Ziheng Chen}$^\dagger$. ``\href{https://openreview.net/forum?id=lovTDtbsdZ}{Towards a General Attention Framework on Gyrovector Spaces for Matrix Manifolds}.'' NeurIPS 2025.
    \item Rui Wang, Shaocheng Jin, Zhenyu Cai, \textbf{Ziheng Chen}$^\dagger$, Xiao-Jun Wu$^\dagger$, and Josef Kittler. ``\href{https://ieeexplore.ieee.org/abstract/document/11128878/}{Learning a Better SPD Network for Signal Classification: A Riemannian Batch Normalization Method}.'' IEEE TNNLS 2025.
    \item Chen Hu, Rui Wang$^\clubsuit$, Xiaoning Song, Tao Zhou, Xiao-Jun Wu, Nicu Sebe, and \textbf{Ziheng Chen}$^\clubsuit$. ``\href{https://ijcai-preprints.s3.us-west-1.amazonaws.com/2025/2120.pdf}{A Correlation Manifold Self-Attention Network for EEG Decoding}.'' IJCAI 2025.
    \item Rui Wang, Shaocheng Jin, \textbf{Ziheng Chen}$^\dagger$, Xiaoqing Luo, and Xiao-Jun Wu. ``\href{https://arxiv.org/abs/2504.00660}{Learning to Normalize on the SPD Manifold under Bures-Wasserstein Geometry}.'' CVPR 2025.
    \item Rui Wang, Jiayao Jin, \textbf{Ziheng Chen}$^\dagger$, Cong Wu$^\dagger$, Xiao-Jun Wu, and Nicu Sebe. ``\href{https://ieeexplore.ieee.org/document/10994294}{Structural Topology Refinement Network for Skeleton-Based Action Recognition}.'' IEEE TIM 2025.
    \item Rui Wang, Chen Hu, \textbf{Ziheng Chen}$^\dagger$, Xiao-Jun Wu$^\dagger$, and Xiaoning Song. ``\href{https://www.ijcai.org/proceedings/2024/564}{A Grassmannian Manifold Self-Attention Network for Signal Classification}.'' IJCAI 2024.
    \item Rui Wang, Xiao-Jun Wu, \textbf{Ziheng Chen}, Cong Hu, and Josef Kittler. ``\href{https://ieeexplore.ieee.org/stamp/stamp.jsp?arnumber=10467142}{SPD Manifold Deep Metric Learning for Image Set Classification}.'' IEEE TNNLS 2024.
\end{enumerate}
\endgroup

    \chapter*{Abstract}

\begin{itshape}
Recently, deep neural networks operating on manifold-valued representations have garnered significant attention across various machine learning applications. However, many basic neural components remain tied to particular manifolds or rely on Euclidean approximations, while the underlying geometry can make repeated computations costly or numerically unstable. This dissertation addresses these limitations by developing reusable modules, exploiting manifold-specific structures when general constructions are insufficient, and introducing fast and stable geometries. We first generalize batch normalization and classification beyond individual manifolds. For normalization, we develop a framework on Lie groups, with theoretical control over Riemannian sample means and variances. To extend this principle beyond Lie groups, we introduce pseudo-reductive gyrogroups, which generalize classical gyrogroups and Lie groups, and build a normalization framework on this structure for a wider range of manifolds. For classification, we first extend Euclidean Multinomial Logistic Regression (MLR), which consists of a fully connected layer followed by softmax, to Symmetric Positive Definite (SPD) manifolds with flat metrics. We then use Riemannian trigonometry to extend MLR to general Riemannian manifolds. When a general formulation cannot exploit useful structure, we design networks for particular representations. For stable hyperbolic deep learning, we use Proper Velocity, an unconstrained representation of hyperbolic space, and develop its geometry and core neural layers. We also use Busemann functions to build intrinsic and efficient hyperbolic classification and fully connected layers. For full-rank correlation matrices, a normalized alternative to SPD matrices, we construct networks that operate directly on the manifold and derive accurate gradients for end-to-end training under two correlation geometries. Finally, we study how the geometry itself can improve learning on SPD manifolds. To move beyond fixed metrics, we make the metric learnable through parameterized matrix logarithms, allowing it to adapt to data and network dynamics with little additional computation. To improve efficiency and numerical stability, we exploit the product structure of Cholesky factors to construct SPD metrics with fast and stable closed-form operators. The proposed methods are supported by theoretical analysis and validated through numerical experiments and empirical applications in vision, signal processing, graph learning, and genomics.
\end{itshape}

\section*{Keywords}
Riemannian deep learning, geometric deep learning, Riemannian manifolds, matrix manifolds, Lie groups, constant-curvature manifolds

    \newpage
\section*{Acknowledgements}

My Ph.D. journey would not have been possible without the guidance, collaboration, and encouragement of many people. First and foremost, I would like to express my deepest gratitude to my advisor, Prof.~Nicu Sebe. He gave me the freedom to pursue the questions that genuinely interested me and created an open research environment in which I could explore new directions with confidence. At the same time, he was always generous with his time and consistently offered encouragement and guidance in our discussions. His support extended well beyond research. He advised me on career decisions, helped me engage with the research community, and supported my scholarship applications. This balance between intellectual freedom and dependable support has profoundly shaped both the work presented in this dissertation and the researcher I have become.

I am also deeply grateful to Prof.~Bernhard Sch\"olkopf. During my research stay with him, his generosity and intellectual openness allowed me to explore distributional geometry in depth and to develop new perspectives on its role in machine learning. I value his insight, trust, and support. I have greatly enjoyed discussing research questions with him and feel fortunate that our collaboration will continue during my postdoctoral research.

I would also like to thank Prof.~Xiaojun Wu, who was my master's advisor. We continued to collaborate throughout my Ph.D., and I deeply appreciate the generous help, guidance, and encouragement he provided along the way.

My sincere thanks also go to my collaborators Yue Song, Rui Wang, and Shanglin Li, as well as to the students I mentored: Zihan Su, Xianglong Shi, Youxing Li, Ying Zhang, Chen Hu, Shaocheng Jin, Zihao Bi, and Yuting Jiang. Our discussions and joint efforts have been an essential part of my Ph.D. experience. I have learned a great deal from each of them, and I am grateful for the ideas, dedication, and enthusiasm they brought to our work together.

Finally, I would like to thank my family. My deepest and most personal thanks go to my girlfriend, Prof.~Yunmei Liu. Throughout this journey, she has supported me with patience and warmth and has always believed in me. She has made the difficult moments easier and the joyful moments more meaningful. She has also brought more good fortune into my life than I could ever have imagined. I am profoundly grateful to have her by my side.
    \tableofcontents
    \listoftables
    \listoffigures
    \chapter*{Notation and Conventions}
\phantomsection
\addcontentsline{toc}{chapter}{Notation and Conventions}
\markboth{Notation and Conventions}{Notation and Conventions}

This chapter collects the notation used throughout the dissertation.

\section*{General Sets and Linear Algebra}

\begin{description}[
    leftmargin=0.32\textwidth,
    labelwidth=0.28\textwidth,
    labelsep=0.04\textwidth,
    style=multiline,
    font=\normalfont
]
\item[$\bbRscalar$, $\bbR{n}$]
The real numbers and the $n$-dimensional Euclidean space.
\item[$\bbRplus$, $\bbRplusscalar$]
The non-negative and positive real numbers.
\item[$I_n$, $\zerovec_n$, $\Rone_n$, $\vecone$, $\bbzero_{n\times n}$]
The $n \times n$ identity matrix, the $n$-dimensional zero vector, the $n$-dimensional all-one vector (also written $\vecone$ when its dimension is clear), and the $n \times n$ zero matrix. Dimension subscripts may be omitted when they are clear from context.
\item[$\inner{x}{y}$, $\norm{x}$]
The Euclidean inner product and the induced norm.
\item[$\norm{A}$, $\tr(A)$, $\rank(A)$]
The Frobenius norm, trace, and rank of a matrix.
\item[$\diag(x)$, $\operatorname{diag}(A)$]
The diagonal matrix induced by a vector and the diagonal part of a matrix, with
the intended meaning specified by context.
\item[$\argmin$, $\argmax$]
The minimizer and maximizer operators.
\end{description}

\section*{Manifold Geometry}

\begin{description}[
    leftmargin=0.32\textwidth,
    labelwidth=0.28\textwidth,
    labelsep=0.04\textwidth,
    style=multiline,
    font=\normalfont
]
\item[$X,Y$]
Generic sets or topological spaces.
\item[$\calX$]
A generic metric space.
\item[$\calM,\calN$]
Smooth or Riemannian manifolds.
\item[$C^{\infty}(\calM)$]
The space of smooth real-valued functions on $\calM$.
\item[$T_p\calM$]
The tangent space of $\calM$ at $p$.
\item[$g_p$]
The Riemannian metric at $p$, viewed as an inner product on $T_p\calM$.
\item[$\gleft$, $\gright$]
Left- and right-invariant Riemannian metrics on a Lie group.
\item[$\inner{v}{w}_p$, $\norm{v}_p$]
The Riemannian inner product and norm in $T_p\calM$.
\item[$\dist_{\calM}(x,y)$]
The geodesic distance between $x,y \in \calM$.
\item[$\alpha$, $L(\alpha)$]
A general curve on a space and its length.
\item[$\gamma$]
A geodesic or geodesic ray.
\item[$\eta$]
The momentum parameter used in LieBN and GyroBN running-statistics updates.
\item[$M_K^n$, $d_K$, $D_K$]
The $n$-dimensional model space of constant curvature $K$, its distance, and the diameter of $M_K^2$.
\item[$\CAT(K)$]
A metric space whose geodesic triangles satisfy the $\CAT(K)$ comparison inequality.
\item[$\pi_C$]
The orthogonal projection onto a convex subset $C$ of a $\CAT(0)$ space.
\item[$\partial\calX$]
The boundary at infinity of a metric space, defined by equivalence classes of asymptotic geodesic rays when available.
\item[$B^{\gamma}(x)$, $HB_{\tau}^{\gamma}$, $H_{\tau}^{\gamma}$]
The Busemann function associated with a geodesic ray $\gamma$, its horoball, and its horosphere at level $\tau$.
\item[$\rieexp_p$, $\rielog_p$]
The Riemannian exponential and logarithmic maps at $p$.
\item[$\pt{p}{q}$]
Parallel transport from $T_p\calM$ to $T_q\calM$.
\item[$\vt{p}{q}$]
A vector transport from $T_p\calM$ to $T_q\calM$.
\item[$H_{a,p}$, $\tilde{H}_{\tilde{A},P}$]
Euclidean and Riemannian hyperplanes used by multinomial logistic regression.
\item[$P_k$, $A_k$, $\tilde{A}_k$]
Class-wise Riemannian prototype, fixed tangent-space parameter, and transported tangent parameter in RMLR.
\item[$z_k$, $r_k$, $v_k(x)$]
The Euclidean direction parameter, scalar offset parameter, and class score used by the PV MLR and PV FC layers.
\item[$M$, $B$, $v^2$, $s$]
The Riemannian mean, biasing parameter, variance, and scaling factor used by Riemannian normalization layers.
\item[$f_{*,p}$, $d_p f(v)$]
The differential of a smooth map $f$ at $p$, written in Chapter 2 as a linear map and, in later computations, equivalently as applied to a tangent vector $v$.
\item[$\nabla_w f$, $\operatorname{grad}_w f$]
The Euclidean gradient of a scalar objective at a Euclidean parameter $w$ and the Riemannian gradient at $w\in\calM$, respectively.
\item[$\mathfrak{X}(\calM)$, $\mathfrak{X}(\alpha)$]
Smooth vector fields on $\calM$ and smooth vector fields along a curve $\alpha$.
\item[{$[X,Y]$}]
The Lie bracket of smooth vector fields $X,Y \in \mathfrak{X}(\calM)$.
\item[$D$, $\frac{D}{dt}$, $V'$]
The Levi-Civita connection and the induced derivative
$V'=\frac{D}{dt}(V)$ along a curve.
\item[$\fm(\{x_i\})$, $\wfm(\{w_i\},\{x_i\})$]
The Fréchet mean and weighted Fréchet mean.
\end{description}

\section*{Algebraic and Gyrovector Operations}

\begin{description}[
    leftmargin=0.32\textwidth,
    labelwidth=0.28\textwidth,
    labelsep=0.04\textwidth,
    style=multiline,
    font=\normalfont
]
\item[$G$, $e$, $E$]
A group-like algebraic structure and its identity element.
\item[$\oplus$]
A binary operation for groups, Lie groups, and gyrogroups.
\item[$\ltrans_x$, $\rtrans_x$]
Left and right translations by $x$ on a Lie group.
\item[$L_x$]
Left gyrotranslation by $x$, defined by $L_x(y)=x\oplus y$.
\item[$\ominus x$]
The inverse of $x$ under an addition-like operation. In gyrogroups, this is called the gyro-inverse.
\item[$\odot$]
Scalar multiplication in gyrovector spaces.
\item[{$\gyr[x,y]$}]
The gyration generated by $x$ and $y$.
\item[$\gyrinner{x}{y}$]
The gyro inner product.
\item[$\gyrnorm{x}$]
The gyronorm.
\item[$\gyrdist(x,y)$]
The gyrodistance.
\item[$\barcenter_\eta$]
The binary barycenter operator with weight $\eta$, used for running mean
updates on gyrogroups.
\end{description}

\section*{Matrix Manifolds}

\begin{description}[
    leftmargin=0.32\textwidth,
    labelwidth=0.28\textwidth,
    labelsep=0.04\textwidth,
    style=multiline,
    font=\normalfont
]
\item[$\sym{n}$]
The vector space of $n \times n$ symmetric matrices.
\item[$\spd{n}$]
The manifold of $n \times n$ symmetric positive definite matrices.
\item[$\spsd{n}$]
The set of $n \times n$ symmetric positive semidefinite matrices.
\item[$\chol(P)$]
The Cholesky factor of an SPD matrix $P$.
\item[$\chospace{n}$]
The space of $n \times n$ lower triangular matrices with positive diagonal entries.
\item[$\log(P)$, $\exp(A)$, $\log_\alpha(P)$, $\log_\alpha^{-1}(A)$]
The matrix logarithm and exponentiation, followed by the general matrix logarithm used by ALEM and its inverse.
\item[$\oplusale$, $\odotale$]
The ALEM-induced element addition and scalar multiplication on $\spd{n}$.
\item[$f(S)$, $f_{*,S}$, $L_f$]
A symmetric matrix function induced by a scalar function $f$, its differential at $S$, and the corresponding Loewner matrix.
\item[$A\circledast B$]
The Hadamard product of matrices $A$ and $B$ with the same size.
\item[$\log_{*,P}$, $\chol_{*,P}$, $\chol^{-1}_{*,L}$]
The differentials of the matrix logarithm, the Cholesky decomposition, and the inverse Cholesky map at the indicated points.
\item[$\pow_\theta(P)$, $(\pow_\theta)_{*,P}$]
The matrix power map $P\mapsto P^\theta$ on SPD matrices and its differential at $P$.
\item[$\phi:\spd{n}\to\sym{n}$, $\phiMul$, $\gphi$, $\dphi$]
An isometry from SPD matrices to a Euclidean space and its induced abelian group operation, pullback Euclidean metric, and geodesic distance.
\item[$\dlog(D)$, $\clog(P)$]
The diagonal element-wise logarithm and the Log-Cholesky map $\clog(P)=\lfloor L\rfloor+\dlog(\bbD(L))$, where $L=\chol(P)$.
\item[{$\calL_P[V]$}]
The Lyapunov operator defined by $\calL_P[V]P+P\calL_P[V]=V$.
\item[$\inner{A}{B}^{\alphabeta}$, $\norm{A}^{\alphabeta}$]
The two-parameter Euclidean inner product on $\sym{n}$ and its induced norm.
\item[$\bfst$]
The admissible parameter set $\{\alphabeta \in \bbRscalar^2 \mid \min(\alpha,\alpha+n\beta)>0\}$.
\item[$\lfloor A \rfloor$, $\bbD(A)$]
The strictly lower triangular part and the diagonal matrix formed from the diagonal of $A$.
\item[$A_{1/2}$]
The Cholesky half-diagonal operator $A_{1/2}=\lfloor A \rfloor+\frac{1}{2}\bbD(A)$.
\item[$\AI$]
Affine-invariant geometry on SPD matrices.
\item[$\LE$]
Log-Euclidean geometry on SPD matrices.
\item[$\LC$]
Log-Cholesky geometry on SPD matrices.
\item[$\PE$]
Power-Euclidean geometry on SPD matrices.
\item[$\biparamAIM$]
The two-parameter Affine-Invariant Metric (AIM) on SPD matrices.
\item[$\biparamLEM$]
The two-parameter Log-Euclidean Metric (LEM) on SPD matrices.
\item[$\triparamAIM$]
The three-parameter Affine-Invariant Metric (AIM) on SPD matrices.
\item[$\paramLCM$]
The parameterized Log-Cholesky Metric (LCM) on SPD matrices.
\item[$\triparamEM$]
The three-parameter Euclidean Metric (EM) on SPD matrices.
\item[$\biparamALEM$]
The Adaptive Log-Euclidean Metric on SPD matrices.
\item[$\defCDEM$]
The deformed Power-Cholesky Metric on SPD matrices.
\item[$\defCDGBWM$]
The deformed Bures--Wasserstein--Cholesky Metric on SPD matrices.
\item[$\oplusGyrAI$, $\odotai$]
AIM-induced SPD gyroaddition and scalar gyromultiplication.
\item[$\oplusLieAI$, $\oplusLieLE$, $\oplusLieLC$, $\oplusLiePAI$, $\oplusLiePLC$, $\ominusLieAI$]
SPD Lie group operations under AIM, LEM, LCM, and their power-deformed AIM and
LCM variants, with $\ominusLieAI$ denoting the inverse under $\oplusLieAI$ when operation-specific disambiguation is needed.
\item[$\oplusLieLE$, $\odotle$, $\oplusLieLC$, $\odotlc$]
LEM- and LCM-induced SPD gyrovector operations, which coincide with linear
operations in the corresponding global charts.
\item[$\BW$]
Bures--Wasserstein geometry on SPD matrices.
\item[$\gAI$, $\gLE$, $\gLC$, $\gPE$, $\gBW$, $\gcri$, $\gdefcri$]
Metric tensors associated with the corresponding SPD geometries.
\item[$\GL{n}$, $\SL{n}$]
The general linear group and the special linear group.
\item[$\orth{n}$]
The orthogonal group.
\item[$\stiefel{p,n}$]
The Stiefel manifold represented by orthonormal frames.
\item[$\grasonb{p,n}$, $\graspp{p,n}$]
The Grassmannian manifold represented by orthonormal bases and projection
matrices.
\item[$\idonb$, $\idpp$]
Identity points of the Grassmannian under the Orthonormal Basis (ONB) and
Projector Perspective (PP) representations.
\item[$\oplusGyrONB$, $\odotGyrONB$, $\ominusGyrONB$]
Grassmannian gyroaddition, scalar multiplication, and inverse under the ONB
representation.
\item[$\oplusGyrPP$, $\odotGyrPP$, $\ominusGyrPP$]
Grassmannian gyroaddition, scalar multiplication, and inverse under the PP
representation.
\item[$\pi:\grasonb{p,n}\to\graspp{p,n}$]
The Grassmannian isometry $\pi(U)=UU^\top$ from ONB to PP.
\item[Matrix commutator]
The operation $[A,B]=AB-BA$.
\item[$\cor{n}$]
The manifold of full-rank correlation matrices.
\item[$\coropt$, $\calI$, $\inv$]
The correlation normalization map, the cor-inversion operator on correlation
matrices, and the matrix inversion operator on SPD matrices.
\item[$\trilspace{n}$]
The Euclidean space of $n \times n$ lower triangular matrices.
\item[$\LTzero{n}$]
The Euclidean space of $n \times n$ strictly lower triangular matrices.
\item[$\LTone{n}$]
The affine space of $n \times n$ lower triangular matrices with unit diagonal.
\item[$\chocor{n}$]
The manifold of Cholesky factors of full-rank correlation matrices, consisting of lower triangular matrices with positive diagonal entries and unit row norm.
\item[$\hol{n}$, $\rzero{n}$]
The spaces of symmetric hollow matrices and symmetric matrices with null row
sum, respectively.
\item[$\Theta$, $\isoecm$, $\offlog$, $\offexp$, $\logscaled$, $\expscaled$]
Diffeomorphisms and charts used by ECM, OLM, and LSM on correlation matrices.
\item[$\dplus(H)$, $\dstar(C)$]
The diagonal correction operators used in OLM and LSM, respectively.
\item[$\bbDspace{n}$, $\bbDplus{n}$]
The vector space of diagonal matrices and the positive diagonal matrix
manifold.
\item[$\perm{n}$, $\singperm{n}$]
The permutation group and the signed permutation group acting on correlation
matrices.
\item[$\symmetrize{X}$, $\symmetrizeSum{X}$]
The averaged symmetrization $(X+X^\top)/2$ and the unnormalized
symmetrization $X+X^\top$, respectively.
\item[$\hs{i}$, $\bbPHS{n-1}$, $\bbPPB{n-1}$]
The open hemisphere, product hemisphere space, and product of unit Poincaré
balls used by PHCM and GyroBN on correlation manifolds.
\item[$\Phi:\cor{n}\to\bbPPB{n-1}$]
The row-wise Cholesky identification from full-rank correlation matrices to the
product of unit Poincaré balls.
\item[$\so{n}$, $\soLieAlgebra{n}$]
The special orthogonal group and its Lie algebra.
\end{description}

\section*{Constant-Curvature Manifolds}

\begin{description}[
    leftmargin=0.32\textwidth,
    labelwidth=0.28\textwidth,
    labelsep=0.04\textwidth,
    style=multiline,
    font=\normalfont
]
\item[$\calMK{n}$]
The $K$-radius model with curvature parameter $K$.
\item[$\stereo{n}$, $\projhs{n}$]
The $K$-stereographic model and its positive-curvature projected hypersphere
case.
\item[$\sphere{n}$]
The spherical model space with curvature parameter $K$.
\item[$\unitsphere{n}$, $\unitprojhs{n}$]
The unit sphere and unit projected hypersphere. Unit-space notation suppresses
the fixed curvature subscript.
\item[$\pball{n}$]
The Poincar\'e ball model of hyperbolic space with curvature $K<0$.
\item[$\unitpball{n}$, $\unitklein{n}$, $\unitlorentz{n}$]
The unit Poincar\'e ball, unit Beltrami--Klein ball, and unit Lorentz space.
Unit-space notation suppresses the fixed curvature subscript.
\item[$\stoplus$, $\stominus$, $\stodot$]
Gyroaddition, gyro-inverse, and scalar gyromultiplication on the
$K$-stereographic model.
\item[$\lambda_x^K$]
The conformal factor of the $K$-stereographic model at $x$.
\item[$\Moplus$, $\Mominus$, $\Modot$]
Möbius gyroaddition, gyro-inverse, and scalar multiplication on the Poincaré
ball.
\item[$\MKoplus$, $\MKominus$, $\MKodot$]
Gyroaddition, gyro-inverse, and scalar gyromultiplication on the $K$-radius
model.
\item[$\lorentz{n}$]
The Lorentz model of hyperbolic space with curvature $K<0$.
\item[$\Loplus$, $\Lominus$, $\Lodot$]
Lorentz gyroaddition, gyro-inverse, and scalar gyromultiplication on the Lorentz model.
\item[$\Linner{x}{y}$, $\Lnorm{x}$]
The Lorentzian inner product and its associated quantity.
\item[$\Kinner{x}{y}$, $\Knorm{x}$]
The curvature-dependent ambient bilinear form on the $K$-radius model and its
induced tangent-space norm. On the $K<0$ branch, the ambient form is
Lorentzian and becomes positive definite only after restriction to a tangent
space.
\item[$\klein{n}$]
The Beltrami--Klein model of hyperbolic space with curvature $K<0$.
\item[$\Eoplus$, $\Eominus$, $\Eodot$]
Einstein gyroaddition, gyro-inverse, and scalar gyromultiplication on the
Beltrami--Klein model.
\item[$\gamma_x$]
The Einstein gamma factor on the Beltrami--Klein model.
\item[$\tank$, $\sink$, $\cosk$]
Curvature-aware trigonometric functions used for constant-curvature models.
\item[$\PVspace{n}$]
The Proper Velocity model of hyperbolic space with curvature $K<0$.
\item[$\PVoplus$, $\PVominus$, $\PVotimes$]
PV gyroaddition, gyro-inverse, and scalar gyromultiplication.
\item[$\beta_x$, $\gamma_y$]
The relativistic beta factor on PV space and the gamma factor used by the PV--Poincar\'e isometry.
\item[$\PVtoPB$, $\PBtoPV$]
The mutually inverse isometries between the PV model and the Poincar\'e ball.
\item[$\PVtoL$, $\LtoPV$]
The mutually inverse isometries between the PV model and the Lorentz model.
\end{description}

\mainmatter
\onehalfspacing
    \chapter{Introduction}
\label{chapter:1}

\section{Riemannian Deep Learning}
\label{sec:ch1-riemannian-deep-learning}

Over the past decade or so, \emph{Deep Neural Networks (DNNs)} have achieved significant progress in machine learning \citep{hochreiter1997long,krizhevsky2012imagenet,he2016deep,vaswani2017attention}. Traditionally, DNNs have been developed under the assumption that the latent geometry of the input data is Euclidean. However, many applications involve non-Euclidean structures, such as manifolds \citep{bronstein2017geometric,guigui2023introduction}. Therefore, deep learning over Riemannian spaces, referred to as \emph{Riemannian deep learning}, has shown great success in diverse applications, such as computer vision \citep{huang2017deep,huang2017riemannian,huang2018building,chen2023riemannian,khrulkov2020hyperbolic}, natural language processing \citep{ganea2018hyperbolic,shimizu2021hyperbolic,lopez2021vector}, graph and knowledge-graph learning \citep{nickel2017poincare,chami2019hyperbolic,chen2023distribution,li2024hygnet}, multimodal learning \citep{desai2023hyperbolic,pal2025compositional}, recommendation systems \citep{yanghg2025former}, signal processing \citep{brooks2019riemannian,kobler2022spd}, human neuroimaging \citep{pan2022matt,li2025spdim,li2026heegnet,zhou2026eegmoce}, medical imaging \citep{chakraborty2020manifoldnet}, astronomy \citep{chen2025galaxy}, and genome sequence modeling \citep{khan2025hyperbolic}. Commonly encountered manifolds include special orthogonal groups \citep{oneill1983semi}, Symmetric Positive Definite (SPD) manifolds \citep{arsigny2005fast}, Grassmannian manifolds \citep{edelman1998geometry,bendokat2024grassmann}, spherical manifolds \citep{oneill1983semi}, and hyperbolic manifolds \citep{ratcliffe2006foundations,cannon1997hyperbolic}. Many of these manifolds admit computationally tractable Riemannian operators, including geodesics, exponential and logarithmic maps, and parallel transport.

Building on these geometric tools, several fundamental Euclidean neural network components have been generalized to manifold-valued data, including normalization \citep{brooks2019riemannian,chakraborty2020manifoldnorm,lou2020differentiating,kobler2022spd}, attention \citep{pan2022matt}, residual blocks \citep{van2023poincare,katsman2024riemannian}, classification \citep{ganea2018hyperbolic,nguyen2023building,nguyen2024matrix}, and Fully Connected (FC) and convolutional layers \citep{huang2017riemannian,huang2017deep,huang2018building,ganea2018hyperbolic,shimizu2021hyperbolic,chen2022fully,nguyen2024matrix}. However, most existing constructions either remain tied to particular geometries or carry out neural computations in intermediate flat spaces that approximate the intrinsic geometry. Existing normalization methods are either confined to particular SPD metrics \citep{brooks2019riemannian,kobler2022spd}, restricted to matrix Lie groups with a specific distance \citep[Sec.~3.2]{chakraborty2020manifoldnorm}, or applicable more generally but lack theoretical guarantees for controlling sample statistics, as in ManifoldNorm \citep[Algs.~1--2]{chakraborty2020manifoldnorm} and RBN \citep[Alg.~2]{lou2020differentiating}. Classification layers frequently map manifold-valued features to tangent Euclidean spaces \citep{huang2017riemannian,brooks2019riemannian}, ambient Euclidean spaces \citep{huang2017deep}, or coordinate Euclidean spaces \citep{chakraborty2018statistical}, while intrinsic alternatives may require the generalized law of sines \citep{ganea2018hyperbolic} or gyrovector structures \citep{nguyen2023building,nguyen2024matrix}. FC and convolutional layers exhibit similar limitations. Early constructions are tailored to SPD, rotation, and Grassmannian manifolds \citep{huang2017riemannian,huang2017deep,huang2018building}, while hyperbolic counterparts rely on tangent-space mappings \citep{ganea2018hyperbolic,mao2024klein}, Lorentz spacetime \citep{chen2022fully}, or Poincar\'e geometry \citep{shimizu2021hyperbolic}. The weighted-Fr\'echet-mean convolution \citep{chakraborty2020manifoldnet} applies more broadly, but constrains the output manifold dimension to match the input dimension. These limitations motivate unified principles that formulate a module once at an appropriate geometric level and then instantiate it across manifolds carrying the required common structure.

Beyond individual modules, deep architectures have been developed for matrix-valued manifolds, including SPD manifolds \citep{huang2017riemannian,chakraborty2020manifoldnet}, Grassmannian manifolds \citep{huang2018building}, and rotation manifolds \citep{huang2017deep}, as well as vector-valued manifolds, including spherical manifolds \citep{bachmann2020constant,skopek2020mixed} and hyperbolic manifolds \citep{ganea2018hyperbolic,bachmann2020constant,skopek2020mixed,chen2022fully,bdeir2024fully}. Nevertheless, existing networks remain concentrated on a limited set of geometric representations and construction tools. Hyperbolic networks, for example, predominantly use the Poincar\'e ball \citep{ganea2018hyperbolic,shimizu2021hyperbolic} and the Lorentz model (also known as the hyperboloid model) \citep{chen2022fully,bdeir2024fully}, while alternative models remain less explored \citep{cannon1997hyperbolic}. Beyond standard Riemannian and gyrovector operators, Busemann functions and horospheres \citep[Ch.~II.8]{bridson2013metric} have been incorporated into hyperbolic SVMs \citep{fan2023horospherical}, hyperbolic PCA \citep{chami2021horopca}, Sliced-Wasserstein distances \citep{bonet2025sliced}, and prototype-learning methods \citep{ghadimi2021hyperbolic}. Despite these developments, their use in neural network design remains limited. Likewise, correlation matrices have received far less attention than SPD covariance representations despite being statistically compact alternatives to covariance matrices \citep{archakov2024canonical}. Their recently developed tractable Riemannian geometries \citep{thanwerdas2022theoretically,thanwerdas2024permutation} indicate a broader design space that remains underexplored.

Finally, both the module and network designs described above ultimately depend on the underlying Riemannian geometry. A Riemannian metric is not merely a means of measuring distance. It provides the theoretical foundation and concrete computational primitives of Riemannian learning algorithms. By assigning inner products to tangent spaces, the metric determines geodesic distances, exponential and logarithmic maps, parallel transport, and Fr\'echet statistics \citep{oneill1983semi}. These operators enter directly into the design of deep-network components. For example, hyperbolic geometry defines Riemannian classifiers \citep{ganea2018hyperbolic,shimizu2021hyperbolic}, weighted Fr\'echet means support convolutional layers \citep{chakraborty2020manifoldnet} and normalization \citep{brooks2019riemannian,chakraborty2020manifoldnorm,kobler2022spd}, and exponential and logarithmic maps underpin attention \citep{pan2022matt}, residual blocks \citep{katsman2024riemannian}, and pooling layers \citep{wang2020deep}. Consequently, changing the Riemannian metric changes not only the geometry, but also the formulas, parameterizations, computational cost, numerical stability, and ultimately the practical behavior of the associated deep networks. This role is especially evident on the SPD manifold, where a broad range of metrics has been developed \citep{pennec2006riemannian,arsigny2005fast,dryden2010power,lin2019riemannian,bhatia2019bures,han2023learning}. However, most existing metric tensors are fixed, which may limit the expressivity of the induced geometry and its ability to adapt to data or the dynamics of deep networks. Moreover, numerical stability is particularly important in deep-network training, where metric-induced operators are repeatedly evaluated. Developing Riemannian metrics that balance flexibility, tractability, efficiency, and stability is therefore a significant problem for Riemannian deep learning.

\section{Contributions and Outlines}
\label{sec:ch1-contributions-outlines}

The contributions of this dissertation are organized around three connected perspectives: unified Riemannian module design across manifolds, manifold-specific Riemannian network design, and the design of the underlying Riemannian geometries. The first perspective formulates principled network modules which can be applied to different geometries, including normalization and classification layers. The second perspective addresses cases in which a fully general construction is either intractable or unable to exploit useful manifold-specific structure. It therefore uses the additional structures of particular manifolds to develop manifold-specific modules and network architectures. The third perspective moves from network design under prescribed geometries to the design of the geometry itself, developing flexible, efficient, and numerically stable metrics on SPD manifolds. The contributions and organization of the remaining chapters are summarized below.

\cref{chapter:2} establishes the mathematical foundations used throughout the dissertation. The chapter is organized into two parts. The first part develops the general theory required by the subsequent methods, including topology, differential and Riemannian geometry, metric geometry, algebraic structures on manifolds, Riemannian optimization, and matrix functions, and relates these constructions to their Euclidean counterparts. The second part turns to the concrete manifolds studied in later chapters: SPD, full-rank correlation, Grassmannian, and constant-curvature manifolds, together with special orthogonal groups.

\cref{chapter:normalization} develops unified frameworks for Batch Normalization (BN) across structured classes of manifolds, with the goal of controlling both the Riemannian mean and variance beyond a single manifold or metric. \cref{sec:ch3-lie-group-approach} first introduces Lie Group Batch Normalization (LieBN) on Lie groups under invariant metrics. LieBN uses group translations for centering and biasing and tangent-space scaling at the identity for variance control, thereby extending Euclidean BN while preserving the manifold structure. Concrete manifestations are developed on SPD, rotation, and full-rank correlation manifolds, together with efficient implementations and experimental validation. \cref{sec:ch3-gyrogroup-approach} first introduces pseudo-reductive gyrogroups, a new algebraic structure that generalizes classical gyrogroups and Lie groups, thereby providing a more general foundation for principled normalization. Building on this structure, it develops Gyrogroup Batch Normalization (GyroBN), which recovers LieBN as a special case and extends the same normalization principle to manifolds that need not possess a Lie group structure. Finally, GyroBN is instantiated on the Grassmannian, constant-curvature, and full-rank correlation manifolds, and experiments evaluate the resulting layers across matrix-manifold, constant-curvature, and graph learning tasks.

\cref{chapter:rmlr} develops a unified framework for intrinsic classification across Riemannian manifolds. \cref{sec:ch4-spd-classifiers} first extends Euclidean Multinomial Logistic Regression (MLR) to SPD manifolds with flat pullback metrics. By formulating classification through the geodesic margin between an SPD input and a decision hyperplane, it derives closed-form classifiers and provides an intrinsic explanation for the widely used LogEig MLR. \cref{sec:ch4-rmlr-general-geometries} then replaces the potentially intractable point-to-hyperplane infimum with a Riemannian-trigonometric formulation, extending the classifier beyond flat SPD geometries. The resulting Riemannian Multinomial Logistic Regression (RMLR) requires only an explicit Riemannian logarithmic map, incorporates several existing manifold classifiers as special cases, and yields new instantiations on SPD manifolds under multiple metric families and on the special orthogonal group. Finally, \cref{rmlr:sec:experiments} evaluates these classifiers within feedforward, residual, graph, and Lie group networks, as well as in direct manifold-valued classification.

\cref{chapter:riemannian-neural-networks} turns to manifold-specific Riemannian network design, exploiting the additional structures of particular hyperbolic models and correlation manifolds. \cref{sec:ch5-pvnn} first introduces the Proper Velocity (PV) model, an unconstrained model of hyperbolic space, and establishes its Riemannian geometry. Based on this geometry, Proper Velocity Neural Networks (PVNNs) develop MLR, FC, convolutional, activation, and normalization layers for stable hyperbolic deep learning. \cref{sec:ch5-hbnn} uses Busemann functions and horospheres to develop intrinsic and efficient Busemann Multinomial Logistic Regression (BMLR) and Busemann Fully Connected (BFC) layers on both the Poincar\'e and Lorentz models. BMLR interprets its logits through point-to-horosphere distances, while BFC uses the same Busemann logits to construct feature transformations. \cref{sec:ch5-cornet} then develops Correlation Networks (CorNets) for full-rank correlation matrices. It constructs MLR, FC, and convolutional layers under five correlation geometries and derives accurate Riemannian backpropagation. Experiments across vision, graph, and genome learning tasks validate these manifold-specific designs.

\cref{chapter:spd-geometries} develops flexible, fast, and numerically stable Riemannian metrics on SPD manifolds. Whereas the preceding chapters design modules and networks under prescribed geometries, this chapter designs the underlying metrics. \cref{sec:alem} first develops Adaptive Log-Euclidean Metrics (ALEMs). By parameterizing general matrix logarithms within a pullback framework, ALEMs adapt the geometry to data while retaining closed-form Riemannian operators and a compatible abelian Lie group structure; the learned metrics are instantiated in SPD networks and other Riemannian building blocks. \cref{sec:pcm-bwcm} then develops product Cholesky geometries, including the Power-Cholesky Metric (PCM) and Bures--Wasserstein--Cholesky Metric (BWCM). These metrics avoid the scalar logarithms and exponentials used by the Log-Cholesky Metric (LCM) and admit fast and stable closed-form Riemannian and algebraic operators. Their effectiveness is evaluated in SPD classification, residual learning, and tensor interpolation, while separate experiments assess computational efficiency, scalability, and numerical stability.

\cref{chapter:conclusion} summarizes the dissertation and discusses future directions.

\section{Summary of Papers Excluded from the Dissertation}
\label{sec:ch1-excluded-papers}

The technical chapters focus on the core theoretical and methodological contributions. Fourteen additional publications on Riemannian deep learning belong to the same research theme and are summarized below.

\begin{itemize}
    \item \mypara{Riemannian attention.} We first developed self-attention on three specific manifolds: the Grassmannian \citep{wang2024grassatt}, the full-rank correlation manifold \citep{hu2025coratt}, and the SPD manifold under the Bures--Wasserstein geometry \citep{jin2026redrsa}. Then, we generalized geometry-specific constructions into a unified attention framework for general matrix manifolds \citep{wang2025gyroatt}.

    \item \mypara{Geometry-aware electroencephalography (EEG) representation learning.} In~\citet{li2026heegnet}, we used hyperbolic embeddings to represent the hierarchical structure of EEG signals and improve cross-domain generalization. In~\citet{hu2026riemannian}, we introduced Riemannian high-order pooling on SPD manifolds to capture second-order correlations in EEG foundation models.

    \item \mypara{Geometry-specific Riemannian batch normalization.} We developed SPD BN methods under specific Riemannian metrics \citep{wang2025cbn,wang2025gbwmbn} for stable training.

    \item \mypara{Skeleton-based action recognition.} We developed two SPD-based approaches to skeleton action recognition: a graph convolutional network that uses Gaussian embeddings of high-order skeletal statistics to model inter-subject interactions and global correlations for two-person interaction recognition \citep{wang2026rgcn}, and an approach that partitions the skeleton into semantic body regions and models their long-range dependencies with SPD representations \citep{wang2025strn}.

    \item \mypara{Hyperbolic neural networks.} In~\citet{shi2026ilnn}, we constructed a fully intrinsic hyperbolic Lorentz neural network whose FC, BN, concatenation, activation, and dropout modules operate within the Lorentz geometry.

    \item \mypara{Hyperbolic multi-view clustering.} In~\citet{wang2026wasserstein}, we developed hyperbolic multi-view clustering that aligns view-specific distributions through a hyperbolic sliced-Wasserstein distance while preserving hierarchical semantics in the Lorentz manifold.

    \item \mypara{Riemannian interpretation of global covariance pooling.} In~\citet{chen2025understanding}, we provided a unified Riemannian interpretation of matrix functions in global covariance pooling, showing that their effectiveness is explained by the Riemannian classifiers they implicitly respect.

    \item \mypara{SPD deep metric learning.} In~\citet{wangspdmetric2024}, we combined an SPD network encoder with a Riemannian decoder, local covariance regularization, and deep metric learning to improve image-set classification.

\end{itemize}

These papers are excluded from this dissertation because they extend or apply the core works developed in this dissertation. For example, CorAtt \citep{hu2025coratt} applies OLM/LSM correlation geometry to EEG attention, while \cref{sec:ch5-cornet} systematizes how to build neural networks and differentiation over the correlation matrices. GyroAtt \citep{wang2025gyroatt} adopts the gyro paradigm, following \cref{sec:ch3-gyrogroup-approach}. HEEGNet \citep{li2026heegnet} applies the Lorentz GyroBN developed in \cref{sec:ch3-gyrogroup-approach} to domain-specific normalization. CBN \citep{wang2025cbn} and GBWBN \citep{wang2025gbwmbn} specialize \cref{sec:ch3-lie-group-approach}'s template to specific metrics. ILNN \citep{shi2026ilnn} accelerates the Lorentz GyroBN developed in \cref{sec:ch3-gyrogroup-approach} and follows \cref{sec:ch5-pvnn}'s point-to-hyperplane design. Finally, RiemGCP \citep{chen2025understanding} applies the RMLR framework developed in \cref{sec:ch4-rmlr-general-geometries} to interpret matrix-function normalization as implicit SPD Riemannian classifiers.

    \chapter{Mathematical Background}
\label{chapter:2}

\section{Introduction}
\label{sec:ch2-introduction}

This chapter presents the mathematical foundations of the dissertation in two parts. The first part develops the general theory and computational tools used throughout the subsequent chapters, beginning with topology and differential geometry \citep{loring2011introduction}, proceeding to Riemannian geometry \citep{oneill1983semi}, metric geometry \citep{bridson2013metric}, and algebraic structures on manifolds \citep{lang2012algebra,ungar2022analytic}, and then reviewing Riemannian optimization \citep{absil2009optimization,boumal2023introduction}, trivialization \citep{lezcano2019trivializations}, and matrix functions and their differentials \citep{bhatia2013matrix}. Although these concepts can be abstract, many generalize familiar Euclidean constructions, so Euclidean prototypes are provided whenever appropriate. The second part reviews the concrete spaces used later: Symmetric Positive Definite (SPD) and full-rank correlation manifolds, Grassmannian manifolds, special orthogonal groups, and constant-curvature manifolds. Their Riemannian and algebraic operators expose both the common primitives that support unified module design across manifolds and the additional structures later exploited by manifold-specific methods.

\section{Topology}

Topology provides the language for continuity and locality before coordinates are introduced \citep[App.~A]{loring2011introduction}. This is essential because a manifold is usually not defined by a single global coordinate system as in Euclidean space, but by a family of local coordinate systems.

Recall classical analysis on $\bbR{n}$, where many basic notions are formulated in terms of open sets. In the standard Euclidean setting $\bbR{n}$, openness is defined through open balls: a subset $U \subset \bbR{n}$ is open if, for every $x \in U$, there exists $r>0$ such that the open ball $B_r(x)=\{y \in \bbR{n} \mid \norm{y-x}<r\}$ is contained in $U$. However, a general abstract space need not carry a norm or a distance, so it may not have a prior notion of open balls. Topology abstracts the essential closure properties of Euclidean open sets, leading to the following definition.

\begin{parisdefinition}[Topological space {\citep[Def.~A.1]{loring2011introduction}}]
\label{def:ch2-topological-space}
A \emph{topology} on a set $X$ is a collection $\mathcal{T}$ of subsets of $X$ such that
\par\leavevmode\vspace{-\baselineskip}
\begin{enumerate}
    \item $\emptyset \in \mathcal{T}$ and $X \in \mathcal{T}$,
    \item every union of elements of $\mathcal{T}$ belongs to $\mathcal{T}$, \ie for any $\{U_{\alpha}\}_{\alpha \in A} \subset \mathcal{T}$,
    \begin{equation}
    \bigcup_{\alpha \in A} U_{\alpha} \in \mathcal{T}.
    \end{equation}
    \item every finite intersection of elements of $\mathcal{T}$ belongs to $\mathcal{T}$, \ie for any integer $m \geq 1$ and any open sets $U_1,\ldots,U_m \in \mathcal{T}$,
    \begin{equation}
    \bigcap_{i=1}^{m} U_i \in \mathcal{T}.
    \end{equation}
\end{enumerate}
The pair $(X,\mathcal{T})$ is called a \emph{topological space}. Elements of $\mathcal{T}$ are called \emph{open sets}. A subset $F \subset X$ is called \emph{closed} if $X \setminus F$ is open, where $X \setminus F=\{x \in X \mid x \notin F\}$.
\end{parisdefinition}

A basis records enough open sets to reconstruct the full topology, and second countability controls the size of this local description.

\begin{parisdefinition}[Basis and second countability {\citep[Defs.~A.6 and A.12]{loring2011introduction}}]
\label{def:ch2-basis}
Let $(X,\mathcal{T})$ be a topological space. A collection $\mathcal{B}\subset \mathcal{T}$ is a \emph{basis} for $\mathcal{T}$ if every open set $U \in \mathcal{T}$ is a union of elements of $\mathcal{B}$. Equivalently, for every $x \in U$ with $U \in \mathcal{T}$, there exists $B \in \mathcal{B}$ such that $x \in B \subset U$. A topological space is \emph{second countable} if its topology has a countable basis.
\end{parisdefinition}

The next proposition gives the corresponding construction criterion when one starts from a candidate family of subsets rather than from an already specified topology.

\begin{parisproposition}[Criterion {\citep[Prop.~A.8]{loring2011introduction}}]
\label{prop:ch2-basis-criterion}
Let $X$ be a set and let $\mathcal{B}$ be a collection of subsets of $X$. Then $\mathcal{B}$ is a basis for some topology $\mathcal{T}$ on $X$ if and only if
\par\leavevmode\vspace{-\baselineskip}
\begin{enumerate}
    \item $X$ is the union of all sets in $\mathcal{B}$,
    \item whenever $x \in B_1 \cap B_2$ with $B_1,B_2 \in \mathcal{B}$, there exists $B_3 \in \mathcal{B}$ such that $x \in B_3 \subset B_1 \cap B_2$.
\end{enumerate}
\end{parisproposition}

In Euclidean space $\bbR{n}$, the family of open balls generates the standard topology. Moreover, $\bbR{n}$ is second countable because the open balls with centers in $\mathbb{Q}^{n}$ and positive rational radii form a countable basis.

Subspace topology is used whenever a geometric object is described as a subset of an ambient space.

\begin{parisdefinition}[Subspace topology {\citep[Sec.~A.2]{loring2011introduction}}]
\label{def:ch2-subspace-topology}
Let $(X,\mathcal{T})$ be a topological space and let $Y \subset X$. The \emph{subspace topology} on $Y$ is
\begin{equation}
\mathcal{T}_Y
=
\left\{Y \cap U \mid U \in \mathcal{T}\right\}.
\end{equation}
With this topology, $Y$ is called a \emph{subspace} of $X$.
\end{parisdefinition}

For Euclidean space, the unit sphere $S^{n-1}=\{x \in \bbR{n} \mid \norm{x}=1\}$ carries the subspace topology inherited from $\bbR{n}$. Its open sets are exactly the sets $S^{n-1}\cap U$, where $U$ is open in $\bbR{n}$.

\begin{parisdefinition}[Continuous map and homeomorphism {\citep[Sec.~A.7]{loring2011introduction}}]
\label{def:ch2-continuity-homeomorphism}
Let $(X,\mathcal{T}_X)$ and $(Y,\mathcal{T}_Y)$ be topological spaces. A map $f:X \to Y$ is \emph{continuous} if $f^{-1}(V) \in \mathcal{T}_X$ for every $V \in \mathcal{T}_Y$. A map $f:X \to Y$ is a \emph{homeomorphism} if it satisfies:
\par\leavevmode\vspace{-\baselineskip}
\begin{enumerate}
    \item $f$ is bijective,
    \item $f$ and $f^{-1}$ are continuous.
\end{enumerate}
\end{parisdefinition}

In the Euclidean special case, this definition recovers the familiar notion of continuity from calculus. For a map $f:\bbR{n}\to \bbR{m}$ with the standard topologies, $f$ is continuous in the sense of \cref{def:ch2-continuity-homeomorphism} if and only if, for every $x \in \bbR{n}$ and every $\varepsilon>0$, there exists $\delta>0$ such that $\norm{f(y)-f(x)}<\varepsilon$ whenever $\norm{y-x}<\delta$. Homeomorphic spaces are topologically identical. A simple homeomorphism is the translation $x \mapsto x+a$ on $\bbR{n}$, whose inverse is $x \mapsto x-a$.

For manifolds, one also needs a separation condition that prevents distinct points from being topologically indistinguishable.

\begin{parisdefinition}[Hausdorff space {\citep[Def.~A.16]{loring2011introduction}}]
\label{def:ch2-hausdorff}
A topological space $X$ is \emph{Hausdorff} if, for any two distinct points $x,y \in X$, there exist disjoint open sets $U,V \subset X$ such that $x \in U$ and $y \in V$.
\end{parisdefinition}

The Hausdorff condition ensures that points can be separated by neighborhoods. For Euclidean space, if $x\neq y$, then the open balls $B_r(x)$ and $B_r(y)$ are disjoint whenever $0<r<\frac{1}{2}\norm{x-y}$. Thus Euclidean space is Hausdorff.

\cref{tab:ch2-topology-euclidean-examples} summarizes the Euclidean special cases of the above topological concepts.

\begin{table}[t]
\centering
\caption{Euclidean prototypes for topological concepts.}
\label{tab:ch2-topology-euclidean-examples}
\begin{tabularx}{0.92\textwidth}{p{0.34\textwidth}X}
\toprule
\textbf{Topological concept} & \textbf{Euclidean special case} \\
\midrule
Open and closed sets &
$B_r(x)=\{y \in \bbR{n} \mid \norm{y-x}<r\}$, $\overline{B}_r(x)=\{y \in \bbR{n} \mid \norm{y-x}\leq r\}$. \\
Basis and second countability &
$\mathcal{B}_{\mathbb{Q}}=\{B_r(q) \mid q \in \mathbb{Q}^{n}, r \in \mathbb{Q}_{>0}\}$. \\
Subspace topology &
$\mathcal{T}_{S^{n-1}}=\{S^{n-1}\cap U \mid U \subset \bbR{n}\text{ open}\}$. \\
Continuous map & Continuous map in calculus. \\
Hausdorff space &
$x\neq y \Rightarrow B_r(x)\cap B_r(y)=\emptyset$ for $0<r<\frac{1}{2}\norm{x-y}$. \\
\bottomrule
\end{tabularx}
\end{table}

\section{Differential Geometry}

In $\bbR{n}$, a basis provides a global coordinate system: every point is represented by a unique coordinate tuple. By contrast, a manifold usually does not admit a single coordinate map covering the entire space. Instead, each point has a neighborhood homeomorphic to an open subset of $\bbR{n}$, and this homeomorphism provides Euclidean coordinates on that neighborhood. This local Euclidean structure permits derivatives, tangent vectors, and smooth maps to be defined intrinsically.

\begin{parisdefinition}[Topological manifold {\citep[Sec.~5]{loring2011introduction}}]
\label{def:ch2-topological-manifold}
An $n$-dimensional \emph{topological manifold} is a topological space $\calM$ such that
\par\leavevmode\vspace{-\baselineskip}
\begin{enumerate}
    \item $\calM$ is Hausdorff,
    \item $\calM$ is second countable,
    \item every point $p \in \calM$ has a neighborhood $U$ homeomorphic to an open subset of $\bbR{n}$.
\end{enumerate}
The integer $n$ is called the \emph{dimension} of $\calM$.
\end{parisdefinition}

The Hausdorff and second-countability assumptions exclude pathological spaces and ensure that local coordinates behave like ordinary Euclidean neighborhoods.

Charts are local coordinate systems, and an atlas is a collection of such coordinate systems covering the whole manifold.

\begin{parisdefinition}[Chart and atlas {\citep[Sec.~5]{loring2011introduction}}]
\label{def:ch2-chart-atlas}
Let $\calM$ be an $n$-dimensional topological manifold. A \emph{chart} on $\calM$ is a pair $(U,\varphi)$ satisfying:
\par\leavevmode\vspace{-\baselineskip}
\begin{enumerate}
    \item $U \subset \calM$ is open,
    \item $\varphi:U \to \varphi(U) \subset \bbR{n}$ is a homeomorphism onto an open subset of $\bbR{n}$.
\end{enumerate}
Writing $\varphi=(x^1,\ldots,x^n)$, the functions $x^i:U\to\bbRscalar$ are called the \emph{coordinate functions} of the chart. Equivalently, if $r^i:\bbR{n}\to\bbRscalar$ denotes the $i$-th standard coordinate projection, then $x^i=r^i\circ\varphi$.
An \emph{atlas} is a collection of charts $\mathcal{A}=\{(U_{\alpha},\varphi_{\alpha})\}_{\alpha \in A}$ satisfying
\begin{equation}
\calM=\bigcup_{\alpha \in A} U_{\alpha}.
\end{equation}
\end{parisdefinition}

To do calculus consistently across charts, changes of coordinates must preserve smoothness.

\begin{parisdefinition}[Smooth atlas and smooth manifold {\citep[Sec.~5]{loring2011introduction}}]
\label{def:ch2-smooth-manifold}
Two charts $(U,\varphi)$ and $(V,\psi)$ on $\calM$ are \emph{smoothly compatible} if the following transition maps are smooth maps between open subsets of Euclidean spaces:
\par\leavevmode\vspace{-\baselineskip}
\begin{enumerate}
    \item $\psi \circ \varphi^{-1}:\varphi(U \cap V) \to \psi(U \cap V)$,
    \item $\varphi \circ \psi^{-1}:\psi(U \cap V) \to \varphi(U \cap V)$.
\end{enumerate}
A \emph{smooth atlas} is an atlas whose charts are pairwise smoothly compatible. A \emph{smooth manifold} is a topological manifold equipped with a maximal smooth atlas.
\end{parisdefinition}

Smooth compatibility means that changing coordinates does not destroy differentiability. This is the formal mechanism that allows a derivative computed in one coordinate chart to represent an intrinsic geometric object.

Smooth maps between manifolds are defined by checking their coordinate representations.

\begin{parisdefinition}[Smooth map and diffeomorphism {\citep[Sec.~6]{loring2011introduction}}]
\label{def:ch2-smooth-map}
Let $\calM$ and $\calN$ be smooth manifolds. A map $f:\calM \to \calN$ is \emph{smooth} if, for every $p \in \calM$, every chart $(U,\varphi)$ around $p$, and every chart $(V,\psi)$ around $f(p)$ with $f(U) \subset V$, the coordinate representation
\begin{equation}
\psi \circ f \circ \varphi^{-1}:\varphi(U) \to \psi(V)
\end{equation}
is smooth. A smooth map $f:\calM \to \calN$ is a \emph{diffeomorphism} if it satisfies:
\par\leavevmode\vspace{-\baselineskip}
\begin{enumerate}
    \item $f$ is bijective,
    \item $f^{-1}:\calN\to\calM$ is smooth.
\end{enumerate}
\end{parisdefinition}

Diffeomorphisms refine homeomorphisms by preserving smooth structure, so diffeomorphic manifolds are identical from the viewpoint of smooth geometry.

Before defining tangent spaces on manifolds, it is useful to recall what a tangent vector does in Euclidean calculus. Fix $x \in \bbR{n}$ and a direction $v \in \bbR{n}$. The directional derivative at $x$ in the direction $v$ can be viewed as an operator on smooth functions,
\begin{equation}
D_{x,v}:C^{\infty}(\bbR{n}) \to \bbRscalar,
\qquad
D_{x,v}(f)=\left.\frac{\diff}{\diff t}\right|_{t=0} f(x+tv).
\end{equation}
This operator is linear in $f$ and, by the ordinary product rule, satisfies
\begin{equation}
D_{x,v}(fh)=f(x)D_{x,v}(h)+h(x)D_{x,v}(f),
\qquad f,h \in C^{\infty}(\bbR{n}).
\end{equation}
Thus a Euclidean tangent vector can be recognized not only as an arrow $v$, but also as a first-order operator that differentiates smooth functions at $x$. The derivation viewpoint keeps precisely this algebraic behavior and extends it to manifolds, where no global vector structure is available.

\begin{parisdefinition}[Tangent space {\citep[Sec.~8]{loring2011introduction}}]
\label{def:ch2-tangent-space}
Let $\calM$ be a smooth manifold and let $p \in \calM$. A \emph{derivation at $p$} is a map $v:C^{\infty}(\calM) \to \bbRscalar$ satisfying:\footnote{Strictly speaking, the domain should be the algebra of \emph{germs} of smooth functions at $p$, namely equivalence classes of smooth functions that agree on some neighborhood of $p$. For simplicity, we write $C^{\infty}(\calM)$.}
\par\leavevmode\vspace{-\baselineskip}
\begin{enumerate}
    \item linearity,
    \begin{equation}
    v(af+bh)=av(f)+bv(h),
    \qquad a,b \in \bbRscalar,\quad f,h \in C^{\infty}(\calM),
    \end{equation}
    \item the Leibniz rule,
\begin{equation}
v(fh)=f(p)v(h)+h(p)v(f),
\qquad f,h \in C^{\infty}(\calM).
\end{equation}
\end{enumerate}
The set of all derivations at $p$ is a vector space, called the \emph{tangent space} of $\calM$ at $p$ and denoted by $T_p\calM$.
\end{parisdefinition}

The tangent space $T_p\calM$ is a vector space whose operations are pointwise addition and scalar multiplication of derivations. Let $(U,\varphi)=(U,x^1,\ldots,x^n)$ be a chart containing $p$. The chart induces the \emph{coordinate tangent vector} $\left.\frac{\partial}{\partial x^i}\right|_p$, which is the derivation at $p$ defined by
\begin{equation}
\left.\frac{\partial}{\partial x^i}\right|_p(f)
=
\frac{\partial\left(f\circ\varphi^{-1}\right)}{\partial r^i}\left(\varphi(p)\right),
\qquad f\in C^\infty(\calM),\quad i=1,\ldots,n.
\end{equation}
Thus $\left.\frac{\partial}{\partial x^i}\right|_p$ differentiates $f$ in the $i$-th Euclidean coordinate direction after $f$ is expressed in the chart. The coordinate tangent vectors $\left.\frac{\partial}{\partial x^1}\right|_p,\ldots,\left.\frac{\partial}{\partial x^n}\right|_p$ form a basis of $T_p\calM$ \citep[Prop.~8.9]{loring2011introduction}. Consequently, every $v \in T_p\calM$ has a unique coordinate representation
\begin{equation}
v=\sum_{i=1}^n v^i\left.\frac{\partial}{\partial x^i}\right|_p,
\qquad v^i \in \bbRscalar,
\end{equation}
so this coordinate basis identifies $T_p\calM$ with $\bbR{n}$.

The differential of a smooth map generalizes the Jacobian matrix.

\begin{parisdefinition}[Differential {\citep[Sec.~8]{loring2011introduction}}]
\label{def:ch2-differential}
Let $f:\calM \to \calN$ be a smooth map and let $p \in \calM$. The \emph{differential} of $f$ at $p$ is the linear map $f_{*,p}:T_p\calM \to T_{f(p)}\calN$ defined by
\begin{equation}
\left(f_{*,p}(v)\right)(h)=v(h \circ f),
\qquad v \in T_p\calM,\quad h \in C^{\infty}(\calN).
\end{equation}
Throughout this chapter, we write the differential as $f_{*,p}$ and its action on $v$ as $f_{*,p}(v)$. Later computations also use the notation $d_p f(v)$.
\end{parisdefinition}

The intrinsic differential recovers the ordinary Jacobian in local coordinates. Let $(U,\varphi)=(U,x^1,\ldots,x^n)$ be a chart around $p\in\calM$ and let $(V,\psi)=(V,y^1,\ldots,y^m)$ be a chart around $f(p)\in\calN$. The differential $f_{*,p}$ is
\begin{align}
&
f_{*,p}\left(
\begin{bmatrix}
\left.\dfrac{\partial}{\partial x^1}\right|_p & \cdots & \left.\dfrac{\partial}{\partial x^n}\right|_p
\end{bmatrix}
\right)
\notag\\
&=
\begin{bmatrix}
\left.\dfrac{\partial}{\partial y^1}\right|_{f(p)} & \cdots & \left.\dfrac{\partial}{\partial y^m}\right|_{f(p)}
\end{bmatrix}
\begin{bmatrix}
\dfrac{\partial\left(y^1\circ f\right)}{\partial x^1}(p) & \cdots & \dfrac{\partial\left(y^1\circ f\right)}{\partial x^n}(p) \\
\vdots & \ddots & \vdots \\
\dfrac{\partial\left(y^m\circ f\right)}{\partial x^1}(p) & \cdots & \dfrac{\partial\left(y^m\circ f\right)}{\partial x^n}(p)
\end{bmatrix}.
\end{align}
The $(i,j)$-th entry of this matrix is $\partial\left(y^i\circ f\right)/\partial x^j(p)$ for $i=1,\ldots,m$ and $j=1,\ldots,n$. Hence, $f_{*,p}$ is represented by the Jacobian matrix of the local coordinate representation $\psi\circ f\circ\varphi^{-1}$ evaluated at $\varphi(p)$ \citep[Prop.~8.11]{loring2011introduction}.

The differential gives an intrinsic definition of the velocity of a curve.

\begin{parisdefinition}[Smooth curve and velocity {\citep[Sec.~8.6]{loring2011introduction}}]
\label{def:ch2-curve-velocity}
Let $I\subset\bbRscalar$ be an open interval and let $c:I\to\calM$ be a smooth curve. The \emph{velocity vector} of $c$ at $t_0\in I$ is
\begin{equation}
c'(t_0)
:=
c_{*,t_0}\left(\left.\frac{\diff}{\diff t}\right|_{t=t_0}\right)
\in T_{c(t_0)}\calM.
\end{equation}
Equivalently, for every $h\in C^\infty(\calM)$,
\begin{equation}
c'(t_0)(h)
=
\left.\frac{\diff}{\diff t}\right|_{t=t_0}h(c(t)).
\end{equation}
\end{parisdefinition}

\begin{parisproposition}[Velocity in local coordinates {\citep[Prop.~8.15]{loring2011introduction}}]
\label{prop:ch2-velocity-local-coordinates}
Let $c:I\to\calM$ be a smooth curve and let $(U,\varphi)=(U,x^1,\ldots,x^n)$ be a chart containing $c(t)$. Then the velocity is obtained by differentiating the coordinate functions:
\begin{equation}
c'(t)
=
\sum_{i=1}^n
\frac{\diff (x^i\circ c)}{\diff t}(t)
\left.\frac{\partial}{\partial x^i}\right|_{c(t)}.
\end{equation}
Thus the coefficients of $c'(t)$ in the chart-induced basis are precisely the ordinary derivatives of the coordinate representation $\varphi\circ c$.
\end{parisproposition}

Every smooth curve through $p$ yields a tangent vector in $T_p\calM$, and conversely every tangent vector is the velocity of some smooth curve through $p$ \citep[Prop.~8.16]{loring2011introduction}. Therefore,
\begin{equation}
T_p\calM
=
\left\{
c'(0)
\mid
\varepsilon>0,\quad
c:(-\varepsilon,\varepsilon)\to\calM\text{ is smooth},\quad
c(0)=p
\right\}.
\end{equation}
For any curve $c$ representing $v\in T_p\calM$ in this way, the derivation $v$ acts as the directional derivative $v(h)=\left.\frac{\diff}{\diff t}\right|_{t=0}h(c(t))$ \citep[Prop.~8.17]{loring2011introduction}.

Curves also provide a classical method for computing differentials.

\begin{parisproposition}[Differential via curves {\citep[Prop.~8.18]{loring2011introduction}}]
\label{prop:ch2-differential-via-curves}
Let $F:\calM\to\calN$ be a smooth map, let $p\in\calM$, and let $v\in T_p\calM$. If $c:(-\varepsilon,\varepsilon)\to\calM$ is any smooth curve satisfying $c(0)=p$ and $c'(0)=v$, then the differential maps the velocity of $c$ to the velocity of its image curve:
\begin{equation}
\label{eq:ch2-differential-via-curves}
F_{*,p}(v)
=
(F\circ c)'(0)
=
\left.\frac{\diff}{\diff t}\right|_{t=0}F(c(t)).
\end{equation}
In particular, this expression is independent of the chosen representative curve.
\end{parisproposition}
The following examples illustrate the above proposition in the vector and matrix cases.
\begin{parisexample}[Sphere]
\label{ex:ch2-sphere-normalization-differential}
Let $\unitsphere{n}=\{y\in\bbR{n+1}\mid\norm{y}=1\}$ and consider the normalization map $\nu:\bbR{n+1}\setminus\{0\}\to\unitsphere{n}$ defined by $\nu(x)=x/\norm{x}$. For $x\neq0$ and $v\in\bbR{n+1}$, the curve $c(t)=x+tv$ has initial velocity $v$. Applying \cref{eq:ch2-differential-via-curves} gives
\begin{equation}
\nu_{*,x}(v)
=
\left.\frac{\diff}{\diff t}\right|_{t=0}
\frac{x+tv}{\norm{x+tv}}
=
\frac{1}{\norm{x}}
\left(
v-
\frac{\inner{x}{v}}{\norm{x}^2}x
\right)
\in T_{\nu(x)}\unitsphere{n}.
\end{equation}
In particular, when $x\in\unitsphere{n}$, the differential is $\nu_{*,x}(v)=v-\inner{x}{v}x$, the orthogonal projection of $v$ onto $T_x\unitsphere{n}$.
\end{parisexample}

\begin{parisexample}[General linear group]
\label{ex:ch2-general-linear-left-multiplication}
Let $\GL{n}=\left\{A\in\bbR{n\times n}\mid\det(A)\neq0\right\}$ denote the manifold of invertible real $n\times n$ matrices. For $G\in\GL{n}$, define $\ltrans_G:\GL{n}\to\GL{n}$ by $\ltrans_G(B)=GB$. Since $\GL{n}$ is an open submanifold of $\bbR{n\times n}$, its tangent spaces are identified with $\bbR{n\times n}$. For $X\in T_{I_n}\GL{n}\cong\bbR{n\times n}$, the curve $c(t)=I_n+tX$ remains in $\GL{n}$ for sufficiently small $t$ and satisfies $c'(0)=X$. Hence,
\begin{equation}
(\ltrans_G)_{*,I_n}(X)
=
\left.\frac{\diff}{\diff t}\right|_{t=0}
G(I_n+tX)
=
GX,
\end{equation}
so the differential of left multiplication is again left multiplication \citep[Ex.~8.19]{loring2011introduction}.
\end{parisexample}

A vector field assigns a tangent vector to each point in a smooth way.

\begin{parisdefinition}[Vector field {\citep[Def.~12.7]{loring2011introduction}}]
\label{def:ch2-vector-field}
A \emph{vector field} on a smooth manifold $\calM$ is a map $X:\calM \to T\calM$ such that $X(p) \in T_p\calM$ for every $p \in \calM$. It is \emph{smooth} if, for every smooth function $f \in C^{\infty}(\calM)$, the function $p \mapsto X(p)f$ is smooth.
\end{parisdefinition}

\cref{tab:ch2-differential-geometry-examples} summarizes the Euclidean special cases of the above differential-geometric concepts.

\begin{table}[t]
\centering
\caption{Euclidean prototypes for differential-geometric concepts.}
\label{tab:ch2-differential-geometry-examples}
\begin{tabular}{ll}
\toprule
\textbf{Differential-geometric concept} & \textbf{Euclidean special case} \\
\midrule
Smooth manifolds &
$\bbR{n}$. \\
Charts &
$\varphi=\id_{\bbR{n}}$. \\
Smooth map & Smooth map in calculus. \\
Tangent space &
$T_x\bbR{n}\cong\bbR{n}$. \\
\bottomrule
\end{tabular}
\end{table}

\section{Riemannian Geometry}

Riemannian geometry equips each tangent space with an inner product. This turns local tangent vectors into measurable directions and induces global geometric objects such as lengths and distances.

\begin{parisdefinition}[Riemannian manifold {\citep[Defs.~3.1 and 3.2]{oneill1983semi}}]
\label{def:ch2-riemannian-manifold}
Let $\calM$ be a smooth manifold. A \emph{Riemannian metric} on $\calM$ is a smooth assignment
\begin{equation}
p \mapsto g_p:T_p\calM \times T_p\calM \to \bbRscalar
\end{equation}
such that, for every $p \in \calM$, $g_p$ satisfies:
\par\leavevmode\vspace{-\baselineskip}
\begin{enumerate}
    \item $g_p$ is bilinear,
    \item $g_p(v,w)=g_p(w,v)$ for all $v,w \in T_p\calM$,
    \item $g_p(v,v)>0$ for all nonzero $v \in T_p\calM$.
\end{enumerate}
The pair $(\calM,g)$ is called a \emph{Riemannian manifold}. We write the value of the metric as $\inner{v}{w}_p=g_p(v,w)$ for $v,w \in T_p\calM$. The induced norm is
\begin{equation}
\norm{v}_p=\sqrt{\inner{v}{v}_p},
\qquad v \in T_p\calM.
\end{equation}
\end{parisdefinition}

The Riemannian metric extends the Euclidean inner product to curved spaces. Unless explicitly stated otherwise, a \emph{manifold} means a Riemannian manifold, and the pair $(\calM,g)$ is abbreviated as $\calM$. We also use $\inner{v}{w}_p$ for the Riemannian metric.

Once each tangent vector has a norm, one can measure the length of curves and define the induced shortest-path distance.

\begin{parisdefinition}[Length and geodesic distance {\citep[Defs.~5.11 and 5.15]{oneill1983semi}}]
\label{def:ch2-length-distance}
Let $\calM$ be a manifold and let $\alpha:[a,b]\to\calM$ be a piecewise smooth curve. The \emph{length} of $\alpha$ is
\begin{equation}
L(\alpha)=\int_a^b \norm{\dot{\alpha}(t)}_{\alpha(t)} \diff t.
\end{equation}
The \emph{geodesic distance} between $x,y \in \calM$ is
\begin{equation}
\dist_{\calM}(x,y)
=\inf_{\alpha} L(\alpha),
\end{equation}
where the infimum is taken over all piecewise smooth curves $\alpha:[a,b]\to\calM$ satisfying $\alpha(a)=x$ and $\alpha(b)=y$.
\end{parisdefinition}

The geodesic distance in $\bbR{n}$ is exactly the straight-line distance. On a connected manifold $\calM$, the distance function $\dist_{\calM}:\calM\times\calM\to[0,\infty)$ is a metric, which makes $\left(\calM,\dist_{\calM}\right)$ a metric space \citep[Prop.~5.18]{oneill1983semi}.

The notion of a connection generalizes the directional derivative of tangent vectors to curved spaces. It is used to define geodesics, parallel transport, and curvature.
\begin{parisdefinition}[Connection and covariant derivative {\citep[Def.~3.9]{oneill1983semi}}]
\label{def:ch2-connection}
Let $\mathfrak{X}(\calM)$ denote the space of smooth vector fields on $\calM$. A \emph{connection} on $\calM$ is a map
\begin{equation}
D:\mathfrak{X}(\calM) \times \mathfrak{X}(\calM) \to \mathfrak{X}(\calM),
\qquad (X,Y) \mapsto D_X Y,
\end{equation}
that satisfies, for all $X,Y,Z \in \mathfrak{X}(\calM)$, $f_1,f_2,f \in C^{\infty}(\calM)$, and $a,b \in \bbRscalar$:
\par\leavevmode\vspace{-\baselineskip}
\begin{enumerate}
    \item $C^{\infty}(\calM)$-linearity in the first argument,
    \begin{equation}
    D_{f_1X+f_2Y}Z=f_1D_XZ+f_2D_YZ,
    \end{equation}
    \item $\bbRscalar$-linearity in the second argument,
    \begin{equation}
    D_X(aY+bZ)=aD_XY+bD_XZ,
    \end{equation}
    \item the Leibniz rule,
    \begin{equation}
    D_X(fY)=X(f)Y+fD_X Y.
    \end{equation}
\end{enumerate}
The vector field $D_X Y$ is called the \emph{covariant derivative} of $Y$ in the direction $X$.
\end{parisdefinition}

A Riemannian metric determines a canonical connection by requiring zero torsion and compatibility with the metric.

\begin{paristheorem}[Levi-Civita connection {\citep[Thm.~3.11]{oneill1983semi}}]
\label{thm:ch2-levi-civita}
Every Riemannian manifold $(\calM,g)$ admits a unique connection $D$ satisfying, for all smooth vector fields $X,Y,Z \in \mathfrak{X}(\calM)$:
\par\leavevmode\vspace{-\baselineskip}
\begin{enumerate}
    \item torsion-freeness,
    \begin{equation}
    D_X Y-D_Y X=[X,Y],
    \end{equation}
    \item compatibility with the metric,
    \begin{equation}
    X\left(\inner{Y}{Z}\right)=\inner{D_X Y}{Z}+\inner{Y}{D_X Z}.
    \end{equation}
\end{enumerate}
This connection is called the \emph{Levi-Civita connection}.
\end{paristheorem}

\begin{parisexample}[Euclidean Levi-Civita connection {\citep[Def.~3.8 and Lem.~3.14]{oneill1983semi}}]
\label{ex:ch2-euclidean-levi-civita}
Let $x^1,\ldots,x^n$ be the standard coordinates on $\bbR{n}$, and let
\begin{equation}
X=\sum_{i=1}^n X^i\frac{\partial}{\partial x^i},
\qquad
Y=\sum_{j=1}^n Y^j\frac{\partial}{\partial x^j}
\end{equation}
be smooth vector fields. The Levi-Civita connection of the Euclidean metric is
\begin{equation}
D_XY
=\sum_{j=1}^n X(Y^j)\frac{\partial}{\partial x^j}
=\sum_{i,j=1}^n X^i\frac{\partial Y^j}{\partial x^i}\frac{\partial}{\partial x^j}.
\end{equation}
Thus, $D_XY$ is the ordinary directional derivative of the component functions of $Y$ along $X$. In particular, the standard coordinate vector fields are parallel:
\begin{equation}
D_{\frac{\partial}{\partial x^i}}\frac{\partial}{\partial x^j}=0,
\end{equation}
for all $i,j$.
\end{parisexample}

The Levi-Civita connection is the canonical connection of Riemannian geometry. It determines geodesics, parallel transport, and curvature. It also induces a covariant derivative for vector fields along a curve.

\begin{parisproposition}[Induced covariant derivative {\citep[Prop.~3.18]{oneill1983semi}}]
\label{prop:ch2-induced-connection-curve}
Let $\calM$ be a manifold with Levi-Civita connection $D$, let $\alpha:I\to\calM$ be a smooth curve, and let $\mathfrak{X}(\alpha)$ denote the space of smooth vector fields along $\alpha$. There is a unique map
\begin{equation}
\mathfrak{X}(\alpha)\to\mathfrak{X}(\alpha),
\qquad
V \mapsto V'=\frac{D}{dt}(V),
\end{equation}
called the \emph{induced covariant derivative} along $\alpha$, satisfying:
\par\leavevmode\vspace{-\baselineskip}
\begin{enumerate}
    \item $(aV+bW)'=aV'+bW'$ for $V,W \in \mathfrak{X}(\alpha)$ and $a,b \in \bbRscalar$.
    \item $(fV)'=\frac{\diff f}{\diff t}V+fV'$ for $V \in \mathfrak{X}(\alpha)$ and $f \in C^{\infty}(I)$.
    \item Let $U \subseteq \calM$ be an open neighborhood of $\alpha(I)$, and let $Y \in \mathfrak{X}(U)$ be a smooth vector field on $U$. If $V \in \mathfrak{X}(\alpha)$ is the vector field along $\alpha$ obtained by restricting $Y$ to the curve, namely $V(t)=Y_{\alpha(t)} \in T_{\alpha(t)}\calM$, then
    \begin{equation}
    V'(t)=D_{\dot{\alpha}(t)}Y,
    \end{equation}
    where the right-hand side denotes the covariant derivative of the field $Y$.
    \item For $V,W \in \mathfrak{X}(\alpha)$,
    \begin{equation}
    \frac{\diff}{\diff t}\inner{V}{W}_{\alpha(t)}
    =
    \inner{V'}{W}_{\alpha(t)}+\inner{V}{W'}_{\alpha(t)}.
    \end{equation}
\end{enumerate}
\end{parisproposition}

Geodesics are the Riemannian counterparts of straight lines. 

\begin{parisdefinition}[Geodesic {\citep[Ch.~3]{oneill1983semi}}]
\label{def:ch2-geodesic}
Let $\calM$ be a manifold with Levi-Civita connection $D$. A smooth curve $\gamma:I\to\calM$ is a \emph{geodesic} if
\begin{equation}
\frac{D\dot{\gamma}}{dt}=D_{\dot{\gamma}}\dot{\gamma}=0.
\end{equation}
Consequently, every geodesic has \emph{constant speed}: $\norm{\dot{\gamma}(t)}_{\gamma(t)}$ is constant on $I$.
\end{parisdefinition}

Unlike straight lines in Euclidean space, geodesics on a general manifold need not be globally length-minimizing. They are locally length-minimizing curves \citep[Lem.~5.14 and Prop.~5.16]{oneill1983semi}.

Geodesics also turn tangent vectors into manifold points through the exponential map, and locally turn nearby manifold points back into tangent vectors through the logarithmic map.

\begin{parisdefinition}[Exponential and logarithmic maps {\citep[Def.~3.29 and Prop.~3.30]{oneill1983semi}}]
\label{def:ch2-exp-log}
Let $\calM$ be a manifold. For $x \in \calM$ and $v \in T_x\calM$, let $\gamma_{x,v}$ be the geodesic satisfying $\gamma_{x,v}(0)=x$ and $\dot{\gamma}_{x,v}(0)=v$. The \emph{Riemannian exponential map} at $x$ is
\begin{equation}
\rieexp_x(v)=\gamma_{x,v}(1),
\end{equation}
where it is defined. As $\rieexp_x$ is locally invertible around $0 \in T_x\calM$, its local inverse is called the \emph{Riemannian logarithmic map} at $x$ and is denoted by $\rielog_x$.
\end{parisdefinition}

This pair is used repeatedly in Riemannian neural networks to move between nonlinear manifold-valued data and linear tangent-space computations.

To compare or aggregate tangent vectors based at different manifold points, tangent vectors must be transported along curves.

\begin{parisdefinition}[Parallel transport {\citep[Prop.~3.19]{oneill1983semi}}]
\label{def:ch2-parallel-transport}
Let $\alpha:[a,b]\to\calM$ be a smooth curve and let $V(t)$ be a vector field along $\alpha$. The field $V$ is \emph{parallel along $\alpha$} if
\begin{equation}
\frac{DV}{dt}=0
\end{equation}
for all $t \in [a,b]$. Given $v \in T_{\alpha(a)}\calM$, the endpoint $V(b) \in T_{\alpha(b)}\calM$ of the unique parallel vector field satisfying $V(a)=v$ is called the \emph{parallel transport} of $v$ along $\alpha$.
\end{parisdefinition}

When the connecting curve is the relevant geodesic from $x$ to $y$, we denote parallel transport by $\pt{x}{y}:T_x\calM \to T_y\calM$.

\begin{parisproposition}[Parallel transport is an isometry {\citep[Lem.~3.20]{oneill1983semi}}]
\label{prop:ch2-parallel-transport-isometry}
Let $V(t)$ and $W(t)$ be parallel vector fields along a smooth curve $\alpha:[a,b]\to\calM$. Then
\begin{equation}
\inner{V(t)}{W(t)}_{\alpha(t)}
\end{equation}
is constant in $t$. Consequently, parallel transport along $\alpha$ defines a linear isometry between tangent spaces.
\end{parisproposition}

The connection also measures the failure of second covariant derivatives to commute, which is encoded by the curvature tensor.

\begin{parisdefinition}[Curvature {\citep[Lem.~3.35]{oneill1983semi}}]
\label{def:ch2-curvature-tensor}
Let $\calM$ be a manifold with Levi-Civita connection $D$. The \emph{Riemannian curvature tensor} is the map
\begin{equation}
R:\mathfrak{X}(\calM) \times \mathfrak{X}(\calM) \times \mathfrak{X}(\calM) \to \mathfrak{X}(\calM)
\end{equation}
defined by
\begin{equation}
R(X,Y)Z=D_X D_Y Z-D_Y D_X Z-D_{[X,Y]} Z,
\end{equation}
where $[X,Y]$ denotes the Lie bracket of vector fields:
\begin{equation}
[X,Y](f)=X(Y(f))-Y(X(f)),
\qquad f \in C^{\infty}(\calM).
\end{equation}
\end{parisdefinition}

Sectional curvature extracts from the curvature tensor the curvature of each two-dimensional tangent plane.

\begin{parisdefinition}[Sectional curvature {\citep[Lem.~3.39]{oneill1983semi}}]
\label{def:ch2-sectional-curvature}
Let $\calM$ be a manifold with curvature tensor $R$. For a two-dimensional subspace $\sigma \subset T_x\calM$ spanned by linearly independent vectors $u,v \in T_x\calM$, the \emph{sectional curvature} of $\sigma$ is
\begin{equation}
K_x(\sigma)
=
\frac{\inner{R(u,v)v}{u}_x}
{\inner{u}{u}_x\inner{v}{v}_x-\inner{u}{v}_x^2}.
\end{equation}
It is independent of the choice of basis $\{u,v\}$ for $\sigma$.
\end{parisdefinition}

Positive, zero, and negative curvature correspond to spherical, Euclidean, and hyperbolic behavior, which will be discussed later.

A central consequence of non-positive curvature is that the exponential map can become globally invertible under suitable assumptions.

\begin{paristheorem}[Cartan--Hadamard theorem {\citep[Thm.~10.22]{oneill1983semi}}]
\label{thm:ch2-cartan-hadamard}
Let $\calM$ be a complete, connected, and simply connected manifold whose sectional curvature is everywhere non-positive. Then, for every $x \in \calM$, the exponential map $\rieexp_x:T_x\calM \to \calM$ is a global diffeomorphism.
\end{paristheorem}

This theorem explains why several geometries in machine learning are algorithmically convenient, as their exponential maps allow tangent-space computations to be performed globally on the manifold.

The distance also supports averaging manifold-valued data through Fréchet means.

\begin{parisdefinition}[Fréchet mean and variance {\citep[Sec.~2]{pennec2006riemannian}}]
\label{def:ch2-frechet-mean}
Let $(\calX,\dist)$ be a metric space and let $x_1,\ldots,x_N \in \calX$ with weights $w_i>0$ satisfying $\sum_{i=1}^N w_i=1$. A \emph{weighted Fréchet mean} is any minimizer
\begin{equation}
\wfm(\{w_i\},\{x_i\})
\in
\argmin_{z \in \calX}
\sum_{i=1}^{N} w_i \dist(x_i,z)^2.
\end{equation}
When $w_i=\frac{1}{N}$ for all $i$, it is called the \emph{Fréchet mean}, and we write $\fm(\{x_i\})$ for the corresponding minimizer. The infimum of the objective is called the \emph{Fréchet variance}.
\end{parisdefinition}

The objective above is defined on any metric space. On a manifold, the distance is usually the geodesic distance. If the data lie in a sufficiently small geodesic ball, the weighted Fréchet mean exists and is unique \citep[Thm.~2.1]{afsari2011riemannian}.

Pullback metrics allow a complex manifold to inherit a computationally convenient geometry from a simpler prototype space.

\begin{parisdefinition}[Pullback and Riemannian isometry {\citep[Defs.~2.8 and 3.6]{oneill1983semi}}]
\label{def:ch2-riemannian-isometry}
Let $\calM$ and $\calN$ be smooth manifolds, let $g$ be a Riemannian metric on $\calN$, and let $f:\calM \to \calN$ be smooth. The \emph{pullback} of $g$ by $f$ is the symmetric tensor field $f^{*}g$ on $\calM$ defined by
\begin{equation}
(f^{*}g)_p(v,w)
=g_{f(p)}\left(f_{*,p}(v),f_{*,p}(w)\right),
\qquad p \in \calM,\quad v,w \in T_p\calM.
\end{equation}
If $f^{*}g$ is positive definite at every point, it is a Riemannian metric on $\calM$. In particular, when $f$ is a diffeomorphism and $\calM$ is equipped with the pullback metric $f^{*}g$, the map $f:(\calM,f^{*}g)\to(\calN,g)$ is a \emph{Riemannian isometry}.
\end{parisdefinition}

A Riemannian isometry is the Riemannian version of a diffeomorphism: it preserves the metric, lengths, geodesic distances, geodesics, exponential maps, logarithmic maps, parallel transport, curvature, and distance-based quantities such as Fréchet means and variances \citep[Ch.~3]{oneill1983semi}.

\cref{tab:ch2-riemannian-examples} summarizes the Euclidean special cases of the above geometric concepts.

\begin{table}[t]
\centering
\caption{Euclidean prototypes for Riemannian-geometric concepts.}
\label{tab:ch2-riemannian-examples}
\begin{tabular}{ll}
\toprule
\textbf{Geometric concept} & \textbf{Euclidean special case} \\
\midrule
Riemannian metric &
$\inner{u}{v}_x=\inner{u}{v}$. \\
Geodesic distance &
$\dist(x,y)=\norm{x-y}$. \\
Geodesic &
$\gamma(t)=(1-t)x+ty$. \\
Exponential map &
$\rieexp_x(v)=x+v$. \\
Logarithmic map &
$\rielog_x(y)=y-x$. \\
Parallel transport &
$\pt{x}{y}=\id_{\bbR{n}}$. \\
Sectional curvature &
$K\equiv0$ on $\bbR{n}$. \\
Weighted Fréchet mean &
$\wfm(\{w_i\},\{x_i\})=\sum_{i=1}^{N}w_i x_i$. \\
Fréchet variance &
$\operatorname{Var}=\min_{z}\sum_{i=1}^{N}w_i\norm{x_i-z}^2$. \\
\bottomrule
\end{tabular}
\end{table}

\section{Metric Geometry}
\label{sec:ch2-metric-geometry}

Metric geometry extends Riemannian geometry to the more general setting of metric spaces, where no differentiable structure is assumed.\footnote{Here, the missing differentiable structure refers specifically to a smooth manifold structure. Metric analogues of geodesics and curvature can still be defined without it.} The theory develops geodesics and curvature without smooth structure, providing the metric tools used later for hyperbolic neural layers. We begin by recalling basic notions in metric spaces.

\begin{parisdefinition}[Metric space {\citep[Def.~I.1.1]{bridson2013metric}}]
\label{def:ch2-metric-space}
A \emph{metric space} is a pair $(\calX,d)$ where $\calX$ is a nonempty set and $d:\calX \times \calX \to \bbRscalar$ satisfies, for all $x,y,z \in \calX$:
\par\leavevmode\vspace{-\baselineskip}
\begin{enumerate}
    \item $d(x,y)\geq 0$ and $d(x,y)=0$ if and only if $x=y$,
    \item $d(x,y)=d(y,x)$,
    \item $d(x,z)\leq d(x,y)+d(y,z)$.
\end{enumerate}
\end{parisdefinition}

A connected Riemannian manifold becomes a metric space when equipped with its geodesic distance \citep[Prop.~5.18]{oneill1983semi}. The metric-space perspective abstracts away coordinates and focuses on distances, geodesics, and comparison geometry.

Geodesics, rays, and lines generalize unit-speed minimizing geodesics to metric spaces.
\begin{parisdefinition}[Geodesic, geodesic ray, and geodesic line {\citep[Ch.~I.1]{bridson2013metric}}]
\label{def:ch2-geodesic-ray-line}
Let $(\calX,d)$ be a metric space. A \emph{geodesic} joining $x$ to $y$ is a continuous map $\gamma:[0,l]\to\calX$ with $\gamma(0)=x$ and $\gamma(l)=y$ such that
\begin{equation}
d\left(\gamma(t),\gamma(t')\right)=|t-t'|,
\qquad t,t' \in [0,l].
\end{equation}
A \emph{geodesic ray} is a continuous map $\gamma:[0,\infty)\to\calX$ such that $d\left(\gamma(t),\gamma(t')\right)=|t-t'|$ for all $t,t'\geq 0$. A \emph{geodesic line} is a continuous map $\gamma:\bbRscalar\to\calX$ such that $d\left(\gamma(t),\gamma(t')\right)=|t-t'|$ for all $t,t'\in\bbRscalar$.
\end{parisdefinition}

Geodesic metric spaces abstract the requirement that every pair of points be joined by a distance-realizing geodesic.
\begin{parisdefinition}[Geodesic metric space {\citep[Ch.~I.1]{bridson2013metric}}]
\label{def:ch2-geodesic-metric-space}
The metric space $(\calX,d)$ is a \emph{geodesic metric space}, or more briefly a \emph{geodesic space}, if every pair of points in $\calX$ is joined by a geodesic. It is \emph{uniquely geodesic} if there is exactly one geodesic joining $x$ to $y$ for all $x,y\in\calX$.
\end{parisdefinition}

Convexity is defined through geodesics, generalizing linear convexity in Euclidean space and geodesic convexity on manifolds.
\begin{parisdefinition}[Convex subset {\citep[Ch.~I.1]{bridson2013metric}}]
\label{def:ch2-convex-subset-metric}
Let $(\calX,d)$ be a metric space. A subset $C\subseteq\calX$ is \emph{convex} if every pair $x,y\in C$ can be joined by a geodesic in $\calX$ and the image of every such geodesic is contained in $C$.
\end{parisdefinition}

We next review concepts that extend curvature from manifolds to metric spaces. The reference spaces are the model spaces of constant curvature.
\begin{parisdefinition}[Model space {\citep[Ch.~I.2]{bridson2013metric}}]
\label{def:ch2-model-space}
For $K\in\bbRscalar$, the \emph{model space} $(M_K^n,d_K)$ is given by
\begin{equation}
(M_K^n,d_K)
=
\begin{cases}
\left(\unitsphere{n}, \frac{1}{\sqrt{K}}d\right), & K>0,\\
\left(\bbR{n}, d\right), & K=0,\\
\left(\unitlorentz{n}, \frac{1}{\sqrt{-K}}d\right), & K<0,
\end{cases}
\end{equation}
where $d$ is the geodesic distance in the corresponding manifold. The unit sphere $\unitsphere{n}$ and unit Lorentz manifold $\unitlorentz{n}$ are reviewed in \cref{sec:ch2-constant-curvature-manifolds}. The diameter of $M_K^2$ is denoted by
\begin{equation}
D_K
=
\begin{cases}
\pi/\sqrt{K}, & K>0,\\
\infty, & K\le 0.
\end{cases}
\end{equation}
\end{parisdefinition}

\begin{parisdefinition}[Comparison triangle {\citep[Lem.~I.2.14 and Sec.~II.1]{bridson2013metric}}]
\label{def:ch2-comparison-triangle}
Let $(\calX,d)$ be a geodesic metric space and let $\triangle(x,y,z)$ be a geodesic triangle in $\calX$ with side lengths $a=d(y,z)$, $b=d(x,z)$, and $c=d(x,y)$. A \emph{comparison triangle} for $\triangle(x,y,z)$ in $M_K^2$ is a triangle $\triangle(\bar{x},\bar{y},\bar{z})$ such that $d_K(\bar{y},\bar{z})=a$, $d_K(\bar{x},\bar{z})=b$, and $d_K(\bar{x},\bar{y})=c$. When $a+b+c<2D_K$, the comparison triangle exists. If $p$ lies on the side from $x$ to $y$, then a \emph{comparison point} for $p$ is the point $\bar{p}$ on the side from $\bar{x}$ to $\bar{y}$ such that $d(x,p)=d_K(\bar{x},\bar{p})$ and $d(y,p)=d_K(\bar{y},\bar{p})$. Comparison points on the other two sides are defined analogously.
\end{parisdefinition}

$\CAT(K)$ spaces encode curvature through triangle comparison with the model plane $M_K^2$. Intuitively, a $\CAT(K)$ space is a metric space whose triangles are thinner than the corresponding comparison triangles in $M_K^2$.
\begin{parisdefinition}[$\CAT(K)$ space {\citep[Def.~II.1.1]{bridson2013metric}}]
\label{def:ch2-catk}
Let $(\calX,d)$ be a metric space and let $K\in\bbRscalar$. Let $\Delta$ be a geodesic triangle in $\calX$ with perimeter less than $2D_K$, and let $\bar{\Delta}\subset M_K^2$ be a comparison triangle for $\Delta$. The triangle $\Delta$ satisfies the \emph{$\CAT(K)$ inequality} if, for all $p,q\in\Delta$ and all corresponding comparison points $\bar{p},\bar{q}\in\bar{\Delta}$,
\begin{equation}
d(p,q)\leq d_K(\bar{p},\bar{q}).
\end{equation}
Then $\calX$ is called a \emph{$\CAT(K)$ space} as follows:
\par\leavevmode\vspace{-\baselineskip}
\begin{enumerate}
    \item If $K\leq0$, then $\calX$ is a $\CAT(K)$ space if $\calX$ is a geodesic space all of whose geodesic triangles satisfy the $\CAT(K)$ inequality.
    \item If $K>0$, then $\calX$ is a $\CAT(K)$ space if $\calX$ is $D_K$-geodesic and all geodesic triangles in $\calX$ of perimeter less than $2D_K$ satisfy the $\CAT(K)$ inequality.
\end{enumerate}
Here, $D_K$-geodesic means that for every pair of points $x,y\in\calX$ with $d(x,y)<D_K$, there is a geodesic joining $x$ to $y$.
\end{parisdefinition}

\begin{parisdefinition}[Hadamard space {\citep[p.~159]{bridson2013metric}}]
\label{def:ch2-hadamard-space}
A \emph{Hadamard space} is a complete $\CAT(0)$ space.
\end{parisdefinition}

In particular, a Hadamard manifold, a complete, simply connected Riemannian manifold with non-positive sectional curvature, is a Hadamard space \citep[Thm.~II.4.1]{bridson2013metric}.

\begin{parisproposition}[Orthogonal projection {\citep[Prop.~II.2.4]{bridson2013metric}}]
\label{prop:ch2-orthogonal-projection}
Let $(\calX,d)$ be a $\CAT(0)$ space and let $C\subseteq\calX$ be a convex subset that is complete in the induced metric. For every $x\in\calX$, there exists a unique point $\pi_C(x)\in C$ such that
\begin{equation}
d\left(x,\pi_C(x)\right)
=
\inf_{y\in C} d(x,y)
=
d(x,C).
\end{equation}
If $x'$ belongs to the geodesic segment $\left[x,\pi_C(x)\right]$, then
\begin{equation}
\pi_C(x')=\pi_C(x).
\end{equation}
The map $\pi_C:\calX\to C$ is called the \emph{orthogonal projection}, or simply the \emph{projection}.
\end{parisproposition}

With geodesics, one can define asymptotic rays, boundary points, Busemann functions, horoballs, and horospheres in metric spaces.
\begin{parisdefinition}[Asymptotic rays and boundary points {\citep[Def.~II.8.1]{bridson2013metric}}]
\label{def:ch2-asymptotic-rays}
Let $(\calX,d)$ be a metric space. Two geodesic rays $\gamma,\eta:[0,\infty)\to\calX$ are \emph{asymptotic} if
\begin{equation}
\text{there exists } C\geq0 \text{ such that } d\left(\gamma(t),\eta(t)\right)\leq C \text{ for all } t\geq0.
\end{equation}
The set $\partial\calX$ of \emph{boundary points}, also called \emph{points at infinity} or \emph{ideal points}, is the set of equivalence classes of geodesic rays, where two geodesic rays are equivalent if and only if they are asymptotic.
\end{parisdefinition}

Asymptotic rays generalize parallel lines. In hyperbolic geometry, they point toward the same ideal boundary point, which becomes a direction for defining Busemann functions.
\begin{parisdefinition}[Busemann function, horoball, and horosphere {\citep[Def.~II.8.17]{bridson2013metric}}]
\label{def:ch2-busemann-horosphere}
Let $(\calX,d)$ be a metric space and let $\gamma:[0,\infty)\to\calX$ be a geodesic ray. If the limit exists, the \emph{Busemann function} associated with $\gamma$ is
\begin{equation}
B^{\gamma}(x)
=
\lim_{t\to\infty}
\left(d\left(x,\gamma(t)\right)-t\right),
\qquad x\in\calX.
\end{equation}
For $\tau\in\bbRscalar$, the sublevel set
\begin{equation}
HB_{\tau}^{\gamma}
=
\left\{x\in\calX \mid B^{\gamma}(x)\leq\tau\right\}
\end{equation}
is a \emph{horoball}, and the level set
\begin{equation}
H_{\tau}^{\gamma}
=
\left\{x\in\calX \mid B^{\gamma}(x)=\tau\right\}
\end{equation}
is a \emph{horosphere}.
\end{parisdefinition}

In Hadamard spaces, the Busemann limit exists \citep[Lem.~II.8.18]{bridson2013metric}. Moreover, Busemann functions associated with asymptotic rays agree up to an additive constant.

\begin{pariscorollary}[Busemann functions of asymptotic rays {\citep[Cor.~II.8.20]{bridson2013metric}}]
\label{cor:ch2-busemann-asymptotic-rays}
If $\calX$ is a Hadamard space, then the Busemann functions associated with asymptotic rays in $\calX$ are equal up to addition of a constant.
\end{pariscorollary}

In Euclidean space, if $\gamma(t)=tv$ with $\norm{v}=1$, then $B^{\gamma}(x)=-\inner{x}{v}$. Thus the Busemann function provides an intrinsic generalization of the Euclidean inner product, up to sign. Therefore, horospheres generalize Euclidean hyperplanes. \cref{tab:ch2-metric-geometry-examples} summarizes the Euclidean special cases of the metric-geometric concepts.

\begin{table}[t]
\centering
\caption{Euclidean prototypes for metric-geometric concepts.}
\label{tab:ch2-metric-geometry-examples}
\begin{tabular}{ll}
\toprule
\textbf{Metric-geometric concept} & \textbf{Euclidean special case} \\
\midrule
Geodesic, geodesic ray, and geodesic line &
Straight segment, ray, and line. \\
Asymptotic geodesic rays &
Parallel rays. \\
Busemann function $-B^{\gamma}(x)$ &
Inner product $\inner{x}{v}$. \\
Horosphere $H_{\tau}^{\gamma}$ &
Hyperplane. \\
Horospheres associated with asymptotic rays &
Parallel hyperplanes. \\
\bottomrule
\end{tabular}
\end{table}

\section{Algebraic Structures on Manifolds}
\label{sec:ch2-algebraic-structures}

Euclidean neural networks rely on vector addition and scalar multiplication to combine, translate, and rescale features. Manifold-valued representations generally lack a global linear structure, which motivates algebraic structures on manifolds that play analogous roles. This section reviews groups, Lie groups, gyrogroups and gyrovector spaces, which generalize addition, subtraction, and scalar multiplication to curved spaces.

\begin{parisdefinition}[Group {\citep[Ch.~I.2]{lang2012algebra}}]
\label{def:ch2-group}
A \emph{group} is a nonempty set $G$ equipped with a binary operation\footnote{The group operation is usually written multiplicatively and called multiplication, as it is usually noncommutative. For a commutative group, it is often written additively and called addition. In this dissertation, we use $\oplus$ for simplicity.} $\oplus:G \times G \to G$ satisfying, for all $x,y,z\in G$:
\par\leavevmode\vspace{-\baselineskip}
\begin{enumerate}
    \item associativity, $x\oplus(y\oplus z)=(x\oplus y)\oplus z$,
    \item identity, there exists $e\in G$, called the \emph{identity element} or \emph{neutral element}, such that $e\oplus x=x\oplus e=x$,
    \item inverse, for every $x\in G$, there exists $\ominus x \in G$ such that $(\ominus x)\oplus x=x\oplus(\ominus x)=e$.
\end{enumerate}
\end{parisdefinition}

\begin{parisdefinition}[Abelian group {\citep[Ch.~I.2]{lang2012algebra}}]
\label{def:ch2-abelian-group}
A group $(G,\oplus)$ is an \emph{abelian group}, or \emph{commutative group}, if $x\oplus y=y\oplus x$ for all $x,y\in G$.
\end{parisdefinition}

Lie groups are simultaneously algebraic and geometric objects. They add smooth structure to groups and require the group operations to be smooth.

\begin{parisdefinition}[Lie group {\citep[Def.~6.20]{loring2011introduction}}]
\label{def:ch2-lie-group}
A \emph{Lie group} is a smooth manifold $G$ equipped with a binary operation $\oplus:G \times G \to G$ satisfying:
\par\leavevmode\vspace{-\baselineskip}
\begin{enumerate}
    \item $(G,\oplus)$ is a group,
    \item the multiplication map $(x,y) \mapsto x \oplus y$ is smooth,
    \item the inversion map $x \mapsto \ominus x$ is smooth.
\end{enumerate}
\end{parisdefinition}

\begin{parisdefinition}[Abelian Lie group]
\label{def:ch2-abelian-lie-group}
A Lie group $(G,\oplus)$ is an \emph{abelian Lie group}, or \emph{commutative Lie group}, if its underlying group is abelian, namely $x\oplus y=y\oplus x$ for all $x,y\in G$.
\end{parisdefinition}

\begin{parisexample}[Matrix Lie groups]
\label{ex:ch2-matrix-lie-groups}
The general linear group $\GL{n}$ introduced in \cref{ex:ch2-general-linear-left-multiplication} is a Lie group under matrix multiplication. Standard matrix Lie subgroups of $\GL{n}$ include the \emph{special linear group}, the \emph{orthogonal group}, and the \emph{special orthogonal group}:
\begin{equation}
\begin{aligned}
\SL{n}
&=
\left\{
A \in \GL{n}
\mid
\det(A)=1
\right\},\\
\orth{n}
&=
\left\{
A \in \GL{n}
\mid
A^{\top}A=I_n
\right\},\\
\so{n}
&=
\orth{n}\cap\SL{n}.
\end{aligned}
\end{equation}
\end{parisexample}

Gyrogroups relax the associativity axiom of groups. The deviation from associativity is controlled by gyrations through the gyroassociative law.

\begin{parisdefinition}[Gyrogroup {\citep[Def.~2.7]{ungar2022analytic}}]
\label{def:ch2-gyrogroup}
Given a nonempty set $G$ with a binary operation $\oplus:G \times G \rightarrow G$, $(G,\oplus)$ forms a gyrogroup if its binary operation satisfies the following axioms for any $x,y,z \in G$:
\par\leavevmode\vspace{-\baselineskip}
\begin{enumerate}
    \item Left identity: there exists at least one element $e \in G$, called a left identity or neutral element, such that $e \oplus x=x$.
    \item Left inverse: there exists an element $\ominus x \in G$, called a left inverse of $x$, such that $\ominus x \oplus x=e$.
    \item Left gyroassociative law: there exists an automorphism $\gyr[x,y]:G \rightarrow G$ for each $x,y \in G$ such that
    \begin{equation}
    x \oplus (y \oplus z) = (x \oplus y) \oplus \gyr[x,y]z.
    \end{equation}
    The automorphism $\gyr[x,y]$ is called the gyroautomorphism, or the gyration of $G$ generated by $x,y$.
    \item Left reduction law:
    \begin{equation}
    \gyr[x,y]=\gyr[x \oplus y,y].
    \end{equation}
\end{enumerate}
\end{parisdefinition}

If every gyration is the identity map, the gyroassociative law reduces to ordinary associativity. Hence, gyrogroups naturally generalize groups. 

\begin{parisdefinition}[Gyrocommutative gyrogroup {\citep[Def.~2.8]{ungar2022analytic}}]
\label{def:ch2-gyrocommutative}
A gyrogroup $(G,\oplus)$ is gyrocommutative if it satisfies
\begin{equation}
x \oplus y = \gyr[x,y](y \oplus x)
\quad \text{(gyrocommutative law).}
\end{equation}
\end{parisdefinition}

Gyrocommutativity replaces commutativity. It says that exchanging two operands is possible after applying the appropriate gyration. Similarly, a gyrovector space generalizes a vector space.

\begin{parisdefinition}[Gyrovector space {\citep{chen2025gyrobnextension}}]
\label{def:ch2-gyrovector-space}
A gyrocommutative gyrogroup $(G,\oplus)$ equipped with a scalar gyromultiplication $\odot:\bbRscalar \times G \rightarrow G$ is called a \emph{gyrovector space} if it satisfies the following axioms for $s,t \in \bbRscalar$ and $x,y,z \in G$:
\par\leavevmode\vspace{-\baselineskip}
\begin{enumerate}
    \item Identity scalar multiplication:
    \begin{equation}
    1 \odot x=x.
    \end{equation}
    \item Scalar distributive law:
    \begin{equation}
    (s+t)\odot x=s\odot x \oplus t\odot x.
    \end{equation}
    \item Scalar associative law:
    \begin{equation}
    (s t)\odot x=s\odot(t\odot x).
    \end{equation}
    \item Gyroautomorphism:
    \begin{equation}
    \gyr[x,y](t\odot z)=t\odot\gyr[x,y]z.
    \end{equation}
    \item Identity gyroautomorphism:
    \begin{equation}
\gyr[s\odot x,t\odot x]=\id,
    \end{equation}
where $\id$ is the identity map.
\end{enumerate}
\end{parisdefinition}

\begin{parisremark}
\label{rem:ch2-gyrovector-space-definition}
\citet[Def.~2.3]{nguyen2022gyro} presented a similar definition, except that the identity scalar multiplication axiom also includes $0\odot x=t\odot e=e$ and $(-1)\odot x=\ominus x$. As implied by \citet[Thm.~6.4]{ungar2022analytic}, these conditions are redundant.
\end{parisremark}

Just as an inner product space augments a vector space with a compatible inner product, a real inner product gyrovector space augments a gyrovector space with an ambient inner product and corresponding compatibility axioms.

\begin{parisdefinition}[Real inner product gyrovector space {\citep[Def.~6.2]{ungar2022analytic}}]
\label{def:ch2-real-inner-product-gyrovector-space}
Let $(G,\oplus,\odot)$ be a gyrovector space and let $\inner{\cdot}{\cdot}$ denote the Euclidean inner product on $\bbR{n}$ with associated norm $\norm{\cdot}$. We call $(G,\oplus,\odot,\inner{\cdot}{\cdot})$ a \emph{real inner product gyrovector space} if the following conditions hold.
\par\leavevmode\vspace{-\baselineskip}
\begin{enumerate}
    \item $G\subseteq\bbR{n}$ and inherits the inner product $\inner{\cdot}{\cdot}$ and norm $\norm{\cdot}$.
    \item Inner product gyroinvariance:
    \begin{equation}
    \inner{\gyr[x,y]u}{\gyr[x,y]v}=\inner{u}{v},
    \quad \forall x,y,u,v\in G.
    \end{equation}
    \item Scaling property:
    \begin{equation}
    \frac{|s|\odot x}{\norm{s\odot x}}=\frac{x}{\norm{x}},
    \quad \forall x\in G\setminus\{\zerovec\},
    \quad \forall s\in\bbRscalar\setminus\{0\}.
    \end{equation}
    \item Let $\norm{G}=\{\pm\norm{x}\mid x\in G\}\subset\bbRscalar$. The set $\norm{G}$ forms a one-dimensional real vector space with respect to the vector addition and scalar multiplication induced by $\oplus$ and $\odot$ on $G$.
    \item Homogeneity property:
    \begin{equation}
    \norm{s\odot x}=|s|\odot\norm{x},
    \quad \forall x\in G,
    \quad \forall s\in\bbRscalar.
    \end{equation}
    \item Gyrotriangle inequality:
    \begin{equation}
    \norm{x\oplus y}\leq\norm{x}\oplus\norm{y},
    \quad \forall x,y\in G.
    \end{equation}
\end{enumerate}
\end{parisdefinition}

\begin{parisdefinition}[Gyrovector space isomorphisms {\citep[Def.~6.89]{ungar2022analytic}}]
\label{def:ch2-gyrovector-space-isomorphisms}
Let
\begin{equation}
(G_1, \oplus_1, \odot_1)
\quad\text{and}\quad
(G_2, \oplus_2, \odot_2)
\end{equation}
be real inner product gyrovector spaces. A map $\phi: G_1 \rightarrow G_2$ is a \emph{gyrovector space isomorphism} if it is bijective and satisfies
\begin{equation}
\phi(x \oplus_1 y) = \phi(x) \oplus_2 \phi(y), \quad \forall x, y \in G_1,
\end{equation}
\begin{equation}
\phi(t \odot_1 x) = t \odot_2 \phi(x), \quad \forall x \in G_1, \forall t \in \bbRscalar,
\end{equation}
and preserves the inner product of unit gyrovectors,
\begin{equation}
\frac{\inner{\phi(x)}{\phi(y)}}{\norm{\phi(x)} \norm{\phi(y)}} = \frac{\inner{x}{y}}{\norm{x} \norm{y}}, \quad \forall x, y \in G_1 \text{ with } x \neq \zerovec, y \neq \zerovec.
\end{equation}
\end{parisdefinition}

A useful property is that gyrovector space isomorphisms preserve the gyration, inverse, and identity.
\begin{parisproposition}
\label{prop:ch2-isomorphism-preserves-properties}\linktoproof{prop:ch2-isomorphism-preserves-properties}
Let $(G_1, \oplus_1, \odot_1)$ and $(G_2, \oplus_2, \odot_2)$ be real inner product gyrovector spaces with gyrations $\gyr_1$ and $\gyr_2$, respectively. If $\phi: G_1 \to G_2$ is a gyrovector space isomorphism, then for all $x, y, z \in G_1$,
\begin{align}
\phi\left(\gyr_1[x, y] z\right) &= \gyr_2[\phi(x), \phi(y)] \phi(z), \\
\phi(e_1) &= e_2, \\
\phi(\ominus_1 x) &= \ominus_2 \phi(x),
\end{align}
where $e_1$ and $e_2$ are the gyro identities in $G_1$ and $G_2$, respectively.
\end{parisproposition}

Intuitively, a gyrovector space generalizes a vector space to curved spaces. The gyro-operations associated with a manifold $\calM$ can be defined as follows. Given a predefined origin $e \in \calM$, we assume that the relevant exponential maps, logarithmic maps, and parallel transports are well-defined. Following \citet[Eqs.~(1)--(3)]{nguyen2023building}, for $x,y,z \in \calM$ and $t \in \bbRscalar$, these operations are defined as
\begin{align}
x \oplus y
&=
\rieexp_x\left(\pt{e}{x}\left(\rielog_e(y)\right)\right),
\label{eq:ch2-riem-gyro-addition}\\
t \odot x
&=
\rieexp_e\left(t\rielog_e(x)\right),
\label{eq:ch2-riem-gyro-scalar}\\
\ominus x
&=
\rieexp_e\left(-\rielog_e(x)\right),
\label{eq:ch2-riem-gyro-inverse}\\
\gyr[x,y]z
&=
(\ominus(x\oplus y))\oplus(x\oplus(y\oplus z)).
\label{eq:ch2-riem-gyration}
\end{align}
The corresponding \emph{gyro inner product}, \emph{gyronorm}, and \emph{gyrodistance} are
\begin{align}
\gyrinner{x}{y}
&=
\inner{\rielog_e(x)}{\rielog_e(y)}_e,
\label{eq:ch2-gyro-inner}\\
\gyrnorm{x}
&=
\sqrt{\gyrinner{x}{x}},
\label{eq:ch2-gyro-norm}\\
\gyrdist(x,y)
&=
\gyrnorm{\ominus x \oplus y}.
\label{eq:ch2-gyro-distance}
\end{align}
If the above operations satisfy the axioms in \cref{def:ch2-gyrovector-space}, then $(\calM,\oplus,\odot)$ is a gyrovector space. In Euclidean space, the above definitions reduce to the ordinary vector-space structure.

\section{Riemannian Optimization}
\label{sec:ch2-riemannian-optimization}

This section briefly reviews Riemannian optimization, which addresses the following manifold-constrained optimization problems:
\begin{equation}
\label{eq:ch2-riemannian-optimization-objective}
\min_{w\in\calM} f(w).
\end{equation}
Here, $\calM$ is a Riemannian manifold and $f:\calM\to\bbRscalar$ is a smooth objective function.

\mypara{Riemannian gradient {\citep[Def.~3.47]{oneill1983semi}}.} The \emph{Riemannian gradient} of $f$ at $w\in\calM$, denoted by $\operatorname{grad}_w f\in T_w\calM$, is the unique tangent vector satisfying
\begin{equation}
\label{eq:ch2-riemannian-gradient}
\inner{\operatorname{grad}_w f}{v}_w
=f_{*,w}(v),
\qquad \forall v\in T_w\calM.
\end{equation}
\cref{eq:ch2-riemannian-gradient} characterizes the Riemannian gradient through the directional derivative $f_{*,w}(v)$. This is the same characterization as in Euclidean space, where the Euclidean gradient satisfies $f_{*,w}(v)=\inner{\nabla_w f}{v}$ for $w,v\in\bbR{n}$. As in $\bbR{n}$, the Cauchy--Schwarz inequality shows that, among all unit tangent directions, the directional derivative is minimized by $-\operatorname{grad}_w f/\norm{\operatorname{grad}_w f}_w$ whenever $\operatorname{grad}_w f\neq0$. Therefore, $-\operatorname{grad}_w f$ is the direction of steepest descent.

\mypara{Riemannian optimizers.} To solve the objective in \cref{eq:ch2-riemannian-optimization-objective}, we take \emph{Stochastic Gradient Descent (SGD)} \citep{robbins1951stochastic} as an example to show how to generalize a Euclidean optimizer to the Riemannian setting \citep{absil2009optimization}. At iteration $t$, let $f_t$ denote a stochastic or mini-batch objective associated with $f$. When $\calM=\bbR{m}$, Euclidean SGD updates an iterate $w^{(t)}\in\bbR{m}$ as
\begin{equation}
\label{eq:ch2-euclidean-sgd}
w^{(t+1)}
=w^{(t)}-\alpha_t\nabla_{w^{(t)}}f_t,
\end{equation}
where $\alpha_t>0$ is the learning rate and $\nabla_{w^{(t)}}f_t$ is the Euclidean gradient of $f_t$. On a general manifold, the descent direction must belong to the tangent space at the current iterate, and ordinary vector addition cannot return that direction to the manifold. The corresponding \emph{Riemannian Stochastic Gradient Descent (RSGD)} \citep[Sec.~2.3]{bonnabel2013stochastic} update is
\begin{equation}
\label{eq:ch2-rsgd}
w^{(t+1)}
=\rieexp_{w^{(t)}}\left(
-\alpha_t\operatorname{grad}_{w^{(t)}}f_t
\right),
\end{equation}
whenever the exponential map is defined for this step. In practical algorithms, a retraction can replace the exact exponential map by a first-order approximation \citep[Def.~3.47 and Sec.~4.3]{boumal2023introduction}, but a systematic treatment of retractions is beyond the scope of this chapter. A widely used package is the \texttt{Geoopt} \citep{kochurov2020geoopt}, which integrates manifold-valued parameters and Riemannian optimizers into PyTorch, supporting RSGD and adaptive optimizers such as Riemannian Adam \citep{becigneul2019riemannian}. In-depth discussions of Riemannian optimization can be found in \citet{absil2009optimization,boumal2023introduction}.

\mypara{Trivialization.} Apart from solving \cref{eq:ch2-riemannian-optimization-objective} directly on the manifold, another approach is to parameterize manifold-valued variables with Euclidean parameters and thereby use Euclidean optimization, which is called \emph{trivialization} \citep{lezcano2019trivializations}. Let $\phi:\bbR{m}\to\calM$ be a smooth surjective map, let $z\in\bbR{m}$ be a trainable Euclidean parameter, and set $w=\phi(z)\in\calM$. The constrained objective can then be written as
\begin{equation}
\label{eq:ch2-trivialization-objective}
\min_{z\in\bbR{m}} f\left(\phi(z)\right).
\end{equation}
In practice, the map $\phi$ could be the exponential map or retraction.

\section{Matrix Functions}
\label{sec:ch2-matrix-functions}
This section reviews some matrix functions widely used on matrix manifolds.

\subsection{Matrix Functions and Differentials}
\label{sec:ch2-matrix-functions-differentials}
We denote the Euclidean space of $n \times n$ real symmetric matrices by $\sym{n}$ and the SPD manifold of $n \times n$ SPD matrices by $\spd{n}$. Let $\mathring{I}$ be an open interval of $\bbRscalar$ and let $f:\mathring{I}\rightarrow\bbRscalar$ be a smooth function. For any symmetric matrix $S$ whose eigenvalues lie in $\mathring{I}$, the associated symmetric matrix function is defined by
\begin{equation}
    \label{eq:ch2-symmetric-matrix-function}
    f: S \longmapsto U f(\Sigma) U^\top \in \sym{n}, \text{ with } S=U \Sigma U^\top \text{ as the eigendecomposition.}
\end{equation}
Its differential is known as the Daleckii--Krein formula:
\begin{align}
    \label{eq:ch2-symmetric-matrix-function-differential}
    f_{*,S} (V) &=U\left(L_f \circledast \left(U^{\top} V U\right)\right) U^{\top}, \quad \forall V \in \sym{n}, \\
    \label{eq:ch2-loewner-matrix}
    [L_f]_{i,j}&=
    \begin{cases}
    \frac{f(\sigma_i)-f(\sigma_j)}{\sigma_i-\sigma_j}, & \text { if } \sigma_i \neq \sigma_j \\
    f^{\prime}(\sigma_i), & \text { otherwise }
    \end{cases}
\end{align}
where $L_f$ is called the Loewner matrix, its $(i,j)$-th entry is defined in \cref{eq:ch2-loewner-matrix}, and $\circledast$ denotes the Hadamard product. Three special cases are the matrix logarithm $\log:\spd{n}\rightarrow\sym{n}$, the matrix exponential $\exp:\sym{n}\rightarrow\spd{n}$, which is its inverse, and the matrix power map $\pow_\theta:\spd{n}\rightarrow\spd{n}$ defined by $\pow_\theta(P)=P^\theta$. See \citet[Eqs.~(2.38)--(2.40)]{bhatia2009positive} or \citet[Thm.~V.3.3]{bhatia2013matrix} for more details.

Let $\chospace{n}$ be the Cholesky space of lower triangular matrices with positive diagonal entries. The Cholesky map is $\chol:\spd{n}\to\chospace{n}$ with inverse $\chol^{-1}(L)=LL^\top$. Let $P\in\spd{n}$, $L=\chol(P)$, $V\in T_P\spd{n}$, and $X\in T_L\chospace{n}$. For any square matrix $A$, set $A_{1/2}=\lfloor A \rfloor+\frac{1}{2}\bbD(A)$, where $\lfloor A \rfloor$ is the strictly lower triangular part and $\bbD(A)$ is the diagonal matrix formed from the diagonal of $A$. As shown by \citet[Prop.~4]{lin2019riemannian}, the differentials of $\chol$ and $\chol^{-1}$ are
\begin{equation}
\label{eq:ch2-cholesky-differentials}
\chol_{*,P}(V)=L\left(L^{-1}VL^{-\top}\right)_{1/2},
\qquad
\chol^{-1}_{*,L}(X)=XL^\top+LX^\top.
\end{equation}

\subsection{Backpropagation Through Matrix Functions}
\label{sec:ch2-matrix-functions-backpropagation}

\mypara{Symmetric matrix functions.} The above symmetric matrix functions can be backpropagated using the Daleckii--Krein formula. Although PyTorch supports automatic differentiation through eigendecomposition \citep{paszke2019pytorch}, its backward pass requires computing $(\sigma_i-\sigma_j)^{-1}$ \citep[Prop.~1]{ionescu2015matrix}, which may trigger numerical instability when two eigenvalues approach each other. Following \citet[Eq.~(13)]{brooks2019riemannian}, we instead use the Daleckii--Krein expression in \cref{eq:ch2-symmetric-matrix-function-differential,eq:ch2-loewner-matrix} for the backward pass \citep[Thm.~V.3.3]{bhatia2013matrix}. As shown in \cref{eq:ch2-loewner-matrix}, the divided difference converges to the derivative $f'(\sigma_i)$ when two eigenvalues approach each other, making the expression numerically more stable.

\mypara{Cholesky decomposition.} The backpropagation of the Cholesky decomposition has been studied by \citet{murray2016differentiation}. In the experiments, we use \texttt{torch.linalg.cholesky} and its automatic differentiation.

\section{Example Manifolds}

This section collects the concrete model spaces that recur in later chapters. Each manifold is first described by its underlying set and representation, and then summarized through the Riemannian and algebraic operators.

\subsection{Symmetric Positive Definite Manifolds}
\label{sec:ch2-spd-manifolds}

The SPD manifold has shown great success in diverse applications \citep{huang2017riemannian,chakraborty2020manifoldnet,wang2020deep,lopez2021vector,chen2021hybrid,kobler2022spd}. We denote the set of $n \times n$ SPD matrices by $\spd{n}$ and the vector space of $n \times n$ real symmetric matrices by $\sym{n}$. As shown by \citet{arsigny2005fast}, $\spd{n}$ forms an open submanifold of the Euclidean space $\sym{n}$. We review five common Riemannian metrics on $\spd{n}$: the \emph{affine-invariant metric (AIM)} \citep{pennec2006riemannian}, the \emph{log-Euclidean metric (LEM)} \citep{arsigny2005fast}, the \emph{power-Euclidean metric (PEM)} \citep{dryden2010power}, the \emph{log-Cholesky metric (LCM)} \citep{lin2019riemannian}, and the \emph{Bures--Wasserstein metric (BWM)} \citep{bhatia2019bures}. LEM, AIM, and PEM are represented by the parameterized families $\biparamLEM$, $\biparamAIM$, and $\triparamEM$, respectively. Their common $(\alpha,\beta)$ parameters refer to the following $\orth{n}$-invariantinner product on $\sym{n}$ \citep{thanwerdas2023n}:
\begin{equation}
\label{eq:ch2-spd-alpha-beta-inner}
\inner{A}{B}^{\alphabeta}
=
\alpha\inner{A}{B}+\beta\tr(A)\tr(B),
\end{equation}
where $A,B \in \sym{n}$ and $\alphabeta \in \bfst=\{\alphabeta \in \bbRscalar^2 \mid \min(\alpha,\alpha+n\beta)>0\}$. 
The standard LEM and AIM are recovered from $\biparamLEM$ and $\biparamAIM$, respectively, at $(\alpha,\beta)=(1,0)$. Likewise, the standard $\theta$-PEM is recovered from $\triparamEM$ at $(\alpha,\beta)=(1,0)$. The $\triparamEM$ formulas below assume $\theta\neq0$; their limit as $\theta\to0$ is $\biparamLEM$.

\cref{tab:ch2-spd-lie-operators,tab:ch2-spd-bwm-pem-operators} summarize the associated operators under these five geometries with the following notation.
\begin{enumerate}
    \item \mypara{General notation.} Let $P,Q \in \spd{n}$ be SPD matrices, let $V,W \in T_P\spd{n}$ be tangent vectors, and let $\{P_i\}_{i=1}^{N}\subset\spd{n}$ be a dataset. The norms induced by $\inner{\cdot}{\cdot}^{\alphabeta}$ and the standard Frobenius inner product $\inner{\cdot}{\cdot}$ are denoted by $\norm{\cdot}^{\alphabeta}$ and $\norm{\cdot}$, respectively.\footnote{We use $\norm{\cdot}$ for both the Frobenius norm of matrices and the $\ell_2$ norm of vectors, since both norms are induced by the standard Euclidean inner product on the ambient Euclidean space.}
    \item \mypara{Symmetric matrix functions.} Symmetric matrix functions and their differentials are reviewed in \cref{sec:ch2-matrix-functions-differentials}. In addition to the matrix logarithm and exponential recorded there, the SPD operator tables use the matrix power map $\pow_\theta(P)=P^\theta$, whose differential is obtained by specializing \cref{eq:ch2-symmetric-matrix-function-differential}.
    \item \mypara{LCM.} The Cholesky map, its inverse, the half-diagonal operator, and their differentials are reviewed in \cref{sec:ch2-matrix-functions-differentials}. Let $L=\chol(P)$ and $K=\chol(Q)$.
    In the LCM column, the corresponding tangent vectors are $X=\chol_{*,P}(V)$ and $Y=\chol_{*,P}(W)$. We define the Log-Cholesky map by $\clog(P)=\lfloor L\rfloor+\dlog(\bbD(L))$, where $\dlog(\cdot)$ is the diagonal element-wise logarithm. The diagonal matrices $\bbK$, $\bbL$, $\bbX$, and $\bbY$ contain the diagonal entries of $K$, $L$, $X$, and $Y$, respectively.
    \item \mypara{Gyrovector.} The three geometries in \cref{tab:ch2-spd-lie-operators} also admit gyrovector spaces. As shown by \citet[Sec.~3.1]{nguyen2022gyro} and \citet[Sec.~2.4]{nguyen2023building}, AIM, LEM, and LCM each induce a gyro-structure via \crefrange{eq:ch2-riem-gyro-addition}{eq:ch2-gyro-distance}. For LEM and LCM, the gyrovector spaces reduce to vector spaces:
    \begin{equation}
    \begin{aligned}
    P\oplusLieLE Q&=\exp\left(\log(P)+\log(Q)\right),
    &
    t\odotle P&=\exp\left(t\log(P)\right),\\
    P\oplusLieLC Q&=\clog^{-1}(\clog(P)+\clog(Q)),
    &
    t\odotlc P&=\clog^{-1}(t\clog(P)).
    \end{aligned}
    \end{equation}
    The above two vector additions are the Lie group operations in \cref{tab:ch2-spd-lie-operators}.
    For AIM, the gyroaddition and scalar gyromultiplication are
    \begin{equation}
    \label{eq:ch2-aim-gyro-operations}
    P\oplusGyrAI Q=P^{1/2}QP^{1/2},
    \qquad
    t\odotai P=P^t.
    \end{equation}
    In particular, the AIM gyroaddition above is not the AIM Lie group operation; the latter is listed in \cref{tab:ch2-spd-lie-operators}.
    \item \mypara{BWM.} The Lyapunov operator $\calL_P[V]$ is defined by
    \begin{equation}
    \calL_P[V]P+P\calL_P[V]=V.
    \end{equation}
    For BWM parallel transport, we only present the case where $P$ and $Q$ are commuting matrices, namely $P=U\Sigma U^\top$ and $Q=U\Delta U^\top$, with $\Sigma=\diag(\sigma_i)$ and $\Delta=\diag(\delta_i)$.

    \item \mypara{Completeness.} BWM and $\triparamEM$ with $\theta\neq0$ are geodesically incomplete, meaning that their exponential maps are not defined globally. For BWM, the exponential map is locally defined only for tangent vectors satisfying $\calL_P[V]+I_n\in\spd{n}$. For $\triparamEM$, the exponential map is locally defined only for tangent vectors satisfying $P^\theta+(\pow_\theta)_{*,P}(V)\in\spd{n}$.

    \item \mypara{Fréchet mean.} Karcher flow \citep{karcher1977riemannian} refers to the standard iterative solver for the Fréchet mean objective in \cref{def:ch2-frechet-mean}.
\end{enumerate}

\begin{table}[t]
\centering
\caption{Lie group structures and associated Riemannian operators on $\spd{n}$.}
\label{tab:ch2-spd-lie-operators}
\resizebox{\textwidth}{!}{%
\begin{tabular}{c|ccc}
\toprule
\textbf{Metric} & $\biparamLEM$ & $\biparamAIM$ & \textbf{LCM} \\
\midrule
$Q \oplus P$ &
$\exp\left(\log(P)+\log(Q)\right)$ &
$KPK^{\top}$ &
$\chol^{-1}(\lfloor L+K \rfloor+\bbK\bbL)$ \\
\midrule
$g_P(V,W)$ &
$\inner{\log_{*,P}(V)}{\log_{*,P}(W)}^{\alphabeta}$ &
$\inner{P^{-1}V}{WP^{-1}}^{\alphabeta}$ &
$\inner{\lfloor X \rfloor}{\lfloor Y \rfloor}+\inner{\bbX\bbL^{-1}}{\bbY\bbL^{-1}}$ \\
\midrule
$\dist(P,Q)$ &
$\norm{\log(P)-\log(Q)}^{\alphabeta}$ &
$\norm{\log(Q^{-1/2}PQ^{-1/2})}^{\alphabeta}$ &
$\norm{\clog(P)-\clog(Q)}$ \\
\midrule
$\fm(\{P_i\})$ &
$\exp\left(\frac{1}{N}\sum_{i=1}^{N}\log(P_i)\right)$ &
Karcher flow &
$\clog^{-1}\left(\frac{1}{N}\sum_{i=1}^{N}\clog(P_i)\right)$ \\
\midrule
$\rielog_P Q$ &
$(\log_{*,P})^{-1}\left[\log(Q)-\log(P)\right]$ &
$P^{1/2}\log(P^{-1/2}QP^{-1/2})P^{1/2}$ &
$\chol^{-1}_{*,L}\left[\lfloor K \rfloor-\lfloor L \rfloor+\bbL\dlog(\bbL^{-1}\bbK)\right]$ \\
\midrule
$\rieexp_P V$ &
$\exp\left(\log(P)+\log_{*,P}(V)\right)$ &
$P^{1/2}\exp\left(P^{-1/2}VP^{-1/2}\right)P^{1/2}$ &
$\chol^{-1}\left(\lfloor L \rfloor+\lfloor X \rfloor+\bbL\exp\left(\bbX\bbL^{-1}\right)\right)$ \\
\midrule
$\gamma(t;P,Q)$ &
$\exp\left[\log(P)+t(\log(Q)-\log(P))\right]$ &
$P^{1/2}(P^{-1/2}QP^{-1/2})^tP^{1/2}$ &
$\chol^{-1}\left\{\lfloor L \rfloor+t(\lfloor K \rfloor-\lfloor L \rfloor)+\frac{\bbK^t}{\bbL^{t-1}}\right\}$ \\
\midrule
References &
\citep{arsigny2005fast,thanwerdas2023n}; \cref{alem:thm:rethk_lem_lcm} &
\citep{pennec2006riemannian,thanwerdas2023n} &
\citep{lin2019riemannian}; \cref{alem:thm:rethk_lem_lcm} \\
\bottomrule
\end{tabular}}
\end{table}

\begin{table}[t]
\centering
\caption{Riemannian operators of $\triparamEM$ and BWM on $\spd{n}$.}
\label{tab:ch2-spd-bwm-pem-operators}
\resizebox{0.95\textwidth}{!}{%
\begin{tabular}{ccc}
\toprule
\textbf{Metric} & $\triparamEM$ & \textbf{BWM} \\
\midrule
$g_P(V,W)$ &
$\frac{1}{\theta^2}\inner{(\pow_\theta)_{*,P}(V)}{(\pow_\theta)_{*,P}(W)}^{\alphabeta}$ &
$\frac{1}{2}\inner{\calL_P[V]}{W}$ \\
\midrule
$\dist(P,Q)$ &
$\frac{1}{\left|\theta\right|}\norm{Q^\theta-P^\theta}^{\alphabeta}$ &
$\left[\tr(P)+\tr(Q)-2\tr\left((PQ)^{1/2}\right)\right]^{1/2}$ \\
\midrule
$\rielog_P Q$ &
$\left[(\pow_\theta)_{*,P}\right]^{-1}\left(Q^\theta-P^\theta\right)$ &
$(PQ)^{1/2}+(QP)^{1/2}-2P$ \\
\midrule
$\pt{P}{Q}(V)$ &
$\left[(\pow_\theta)_{*,Q}\right]^{-1}\circ(\pow_\theta)_{*,P}(V)$ &
$U\left[\sqrt{\frac{\delta_i+\delta_j}{\sigma_i+\sigma_j}}\left[U^\top VU\right]_{ij}\right]U^\top$ \\
\midrule
$\rieexp_P V$ &
$\left(P^\theta+(\pow_\theta)_{*,P}(V)\right)^{1/\theta}$ &
$P+V+\calL_P[V]P\calL_P[V]$ \\
\midrule
$\gamma(t;P,Q)$ &
$\left((1-t)P^\theta+tQ^\theta\right)^{1/\theta}$ &
\makecell[c]{$A_tA_t^\top$\\
$A_t=(1-t)P^{1/2}+tP^{-1/2}\left(P^{1/2}QP^{1/2}\right)^{1/2}$} \\
\midrule
References &
\citep{dryden2010power,thanwerdas2023n} &
\citep{bhatia2019bures,thanwerdas2023n} \\
\bottomrule
\end{tabular}}
\end{table}

\subsection{Full-Rank Correlation Manifolds}
\label{sec:ch2-full-rank-correlation-manifolds}

Given a covariance matrix $\Sigma$, its correlation matrix is defined as
\begin{equation}
C=\coropt(\Sigma)=\bbD(\Sigma)^{-\nicefrac{1}{2}}\Sigma\bbD(\Sigma)^{-\nicefrac{1}{2}},
\end{equation}
where $\bbD(\cdot)$ extracts the diagonal part of $\Sigma$ as a diagonal matrix. This diagonal normalization yields a scale-invariant representation: for any positive diagonal matrix $D$, $\coropt(D\Sigma D)=\coropt(\Sigma)$. Hence, correlation matrices remove marginal scales and emphasize pairwise dependencies rather than raw variances.

\begin{figure}[t]
\centering
\includegraphics[width=0.58\textwidth,trim={0cm 0cm 0cm 0cm}]{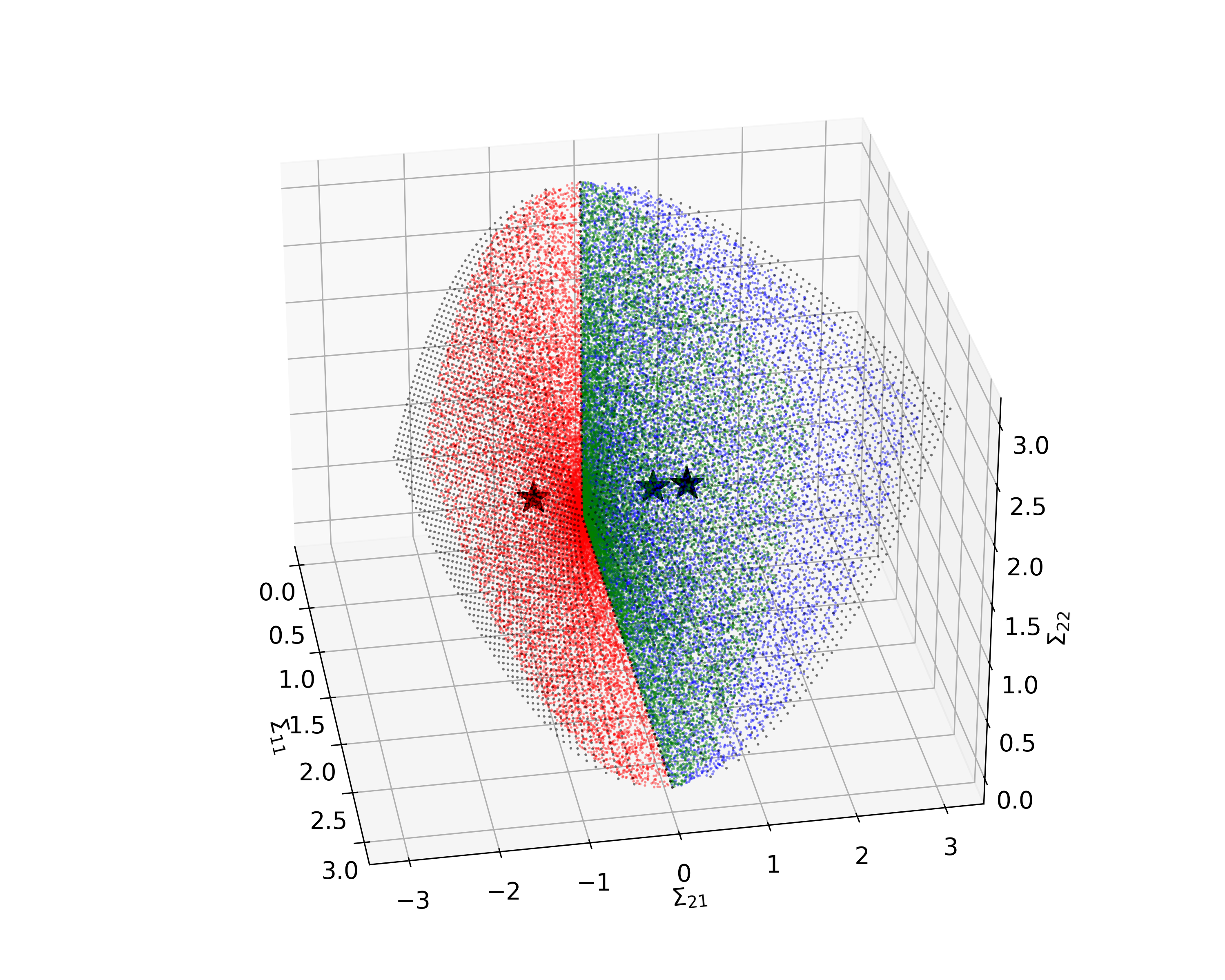}
\caption{The black stars denote $2 \times 2$ correlation matrices, while the red, green, and blue dots denote corresponding SPD matrices. The black dots denote the boundary of the SPD cone.}
\label{fig:ch2-cor-in-spd}
\end{figure}

Only recently have Riemannian structures been developed for correlation matrices.
The space of $n \times n$ full-rank correlation matrices, denoted by $\cor{n}$, forms a Riemannian manifold and can be identified as a quotient manifold of the SPD manifold \citep[Thm.~1]{david2019riemannian}. As illustrated in \cref{fig:ch2-cor-in-spd}, each correlation matrix corresponds to a surface in the SPD manifold. However, this quotient geometry does not guarantee uniqueness or closed forms of the Riemannian logarithm and Fréchet mean \citep[Sec.~1.1]{thanwerdas2022theoretically}. To address this limitation, recent advances introduced five convenient Riemannian metrics on $\cor{n}$: the \emph{Euclidean--Cholesky metric (ECM)} \citep{thanwerdas2022theoretically}, \emph{log-Euclidean--Cholesky metric (LECM)} \citep{thanwerdas2022theoretically}, \emph{poly-hyperbolic--Cholesky metric (PHCM)} \citep{thanwerdas2022theoretically}, \emph{off-log metric (OLM)} \citep{thanwerdas2024permutation}, and \emph{log-scaled metric (LSM)} \citep{thanwerdas2024permutation}. These metrics are pullback metrics from simpler prototype spaces: ECM, LECM, OLM, and LSM are induced from Euclidean spaces, while PHCM is induced from a product of hyperbolic open hemispheres.

We first review the associated prototype spaces.
\begin{enumerate}
    \item $\LTone{n}$ is the affine space of $n \times n$ lower triangular matrices with unit diagonal.
    \item $\LTzero{n}$ is the Euclidean space of $n \times n$ lower triangular matrices with null diagonal.
    \item $\chocor{n}$ is the manifold of $n \times n$ lower triangular matrices with positive diagonals and unit row $\ell_2$-norm.
    \item $\hol{n}$ is the Euclidean space of $n \times n$ symmetric matrices with null diagonals. The tangent space $T_C\cor{n}$ at $C \in \cor{n}$ can be identified with $\hol{n}$.
    \item $\rzero{n}$ is the Euclidean space of $n \times n$ symmetric matrices with null row sum.
\end{enumerate}

\mypara{ECM.} It is derived from $\LTone{n}$ by
\begin{equation}
\cor{n}\xrightleftharpoons[\Theta^{-1}=\coropt\circ\chol^{-1}]{\Theta=\bbD(\chol(\cdot))^{-1}\chol(\cdot)}\LTone{n},
\end{equation}
where $\Theta(C)=\bbD(\chol(C))^{-1}\chol(C)$ for any $C\in\cor{n}$. Here, $\chol(C)$ is the Cholesky decomposition $C=\chol(C)\chol(C)^\top$ and $\bbD(\cdot)$ returns a diagonal matrix consisting of the input diagonals. As $\LTone{n}=I_n+\LTzero{n}$, ECM is essentially induced from the Euclidean space of $\LTzero{n}$.

\begin{parisproposition}[ECM]
\label{prop:ch2-correlation-ecm}
Let $\isoecm(C)=\lfloor\Theta(C)\rfloor$, where $\lfloor\cdot\rfloor$ returns a strictly lower triangular matrix. ECM over $\cor{n}$ is the pullback metric from the Euclidean space $\LTzero{n}$ by $\isoecm$.
\end{parisproposition}

\mypara{LECM.} It is defined by further pulling back ECM:
\begin{equation}
\cor{n}\xrightleftharpoons[(\log\circ\Theta)^{-1}=\coropt\circ\chol^{-1}\circ\exp]{\log\circ\Theta}\LTzero{n},
\end{equation}
where $\log(\cdot):\LTone{n}\longrightarrow\LTzero{n}$ is the matrix logarithm with the matrix exponential $\exp(\cdot)$ as its inverse.

\mypara{OLM.} It is derived from a permutation-invariant inner product over $\hol{n}$ by
\begin{equation}
\cor{n}\xrightleftharpoons[\offexp]{\offlog=\off\circ\log}\hol{n}.
\end{equation}
For any symmetric hollow matrix $H\in\hol{n}$, the operator $\dplus(H)$ returns a unique diagonal matrix such that $\offexp(\cdot):\hol{n}\ni H\longmapsto\exp\left(\dplus(H)+H\right)\in\cor{n}$ is a diffeomorphism. As shown by \citet[Cor.~1 and Sec.~5]{archakov2021new}, $\dplus(H)$ can be computed by the following exponentially converging algorithm: $D_{k+1}=D_k-\log\left(\bbD\left(\exp\left(D_k+H\right)\right)\right)$, with $D_0=\bbzero_{n\times n}$ as the zero matrix.

\mypara{LSM.} It is derived from a permutation-invariant inner product over $\rzero{n}$ by
\begin{equation}
\cor{n}\xrightleftharpoons[\expscaled=\coropt\circ\exp]{\logscaled}\rzero{n}.
\end{equation}
For any correlation matrix $C\in\cor{n}$, there exists a unique positive diagonal matrix $\dstar(C)$ such that $\logscaled(\cdot):\cor{n}\ni C\longmapsto\log(\dstar(C)C\dstar(C))\in\rzero{n}$ is a diffeomorphism. As shown by \citet[Sec.~3.5]{thanwerdas2024permutation}, $\dstar(C)$ corresponds to the unique zero of $f:x\in\bbRplusscalar^n\longmapsto Cx-\frac{1}{x}$, where $\bbRplusscalar^n$ denotes the set of $n$-dimensional positive vectors and $\frac{1}{x}=\left(\frac{1}{x_1},\ldots,\frac{1}{x_n}\right)$. This equation can be solved by damped Newton's method.

\cref{tab:ch2-correlation-maps} summarizes the associated prototype spaces and diffeomorphisms.
\cref{tab:ch2-correlation-ecm-lecm,tab:ch2-correlation-olm-lsm} summarize the vector operations and Riemannian operators for ECM, LECM, OLM, and LSM. We use the following notation and record the following remarks.

\begin{enumerate}
    \item \mypara{General notation.} Let $C,C' \in \cor{n}$ be correlation matrices, let $\{C_i\}_{i=1}^{N}\subset\cor{n}$, and let $V,W \in T_C\cor{n}\cong\hol{n}$ be tangent vectors. Let $L=\chol(C)$.
    \item \mypara{ECM and LECM.} For any $K \in \LTone{n}$ and $X,\xi\in\LTzero{n}$, the maps and differentials involved in ECM and LECM are
    \begin{align}
    \Theta(C)&=\bbD(L)^{-1}L,\\
    \Theta^{-1}(K)&=\bbD(KK^\top)^{-\frac{1}{2}}KK^\top\bbD(KK^\top)^{-\frac{1}{2}},\\
    \log(K)&=\sum_{k=1}^{n-1}\frac{(-1)^{k-1}}{k}\left(K-I_n\right)^k,\\
    \exp(\xi)&=\sum_{k=0}^{n-1}\frac{1}{k!}\xi^k,\\
    \Theta_{*,C}(V)&=\Theta(C)\left(L^{-1}VL^{-\top}\right)_{\frac{1}{2}}-\frac{1}{2}\bbD\left(L^{-1}VL^{-\top}\right)\Theta(C),\\
    \left(\Theta_{*,C}\right)^{-1}(\xi)&=\left(L\xi^\top-C\bbD\left(L\xi^\top\right)\right)\bbD(L)+\bbD(L)\left(\xi L^\top-\bbD\left(L\xi^\top\right)C\right),\\
    \log_{*,K}(\xi)&=\sum_{k=1}^{n-1}\frac{(-1)^{k-1}}{k}\left[\left(K-I_n\right)^{k-1}\xi+\cdots+\xi\left(K-I_n\right)^{k-1}\right],\\
    \exp_{*,X}(\xi)&=\sum_{k=1}^{n-1}\frac{1}{k!}\left(X^{k-1}\xi+X^{k-2}\xi X+\cdots+\xi X^{k-1}\right),\\
    (\log\circ\Theta)_{*,C}(V)&=\log_{*,\Theta(C)}\left(\Theta_{*,C}(V)\right).
    \end{align}
    Due to the nilpotency of $\LTzero{n}$, the matrix logarithm over $\LTone{n}$ and exponentiation over $\LTzero{n}$ are free from eigendecomposition. Although the Euclidean inner product in the ECM and LECM columns can be any inner product, we use the canonical one in this dissertation.

    \item \mypara{OLM and LSM.} Let $H,W\in\hol{n}$, $S=H+\dplus(H)$, $\Sigma=\dstar(C)C\dstar(C)$, $X=\logscaled(C)=\log(\Sigma)\in\rzero{n}$, and $Y\in\rzero{n}$. The involved maps and differentials are \citep[Thms.~2.4 and 4.1]{thanwerdas2024permutation}:
    \begin{align}
    \offlog_{*,C}(V)&=\off\left(\log_{*,C}(V)\right),\\
    \offexp_{*,H}(W)&=\exp_{*,S}\left(W+\dplus_{*,H}(W)\right),\\
    \dplus_{*,H}(W)&=-\diag\left(\left(H^0\right)^{-1}\bbD\left(\exp_{*,S}(W)\right)\vecone\right),\\
    H^0&=[H^0_{il}]\in\spd{n},
    \qquad H^0_{il}=\sum_{j,k}U_{ij}U_{ik}U_{lj}U_{lk}[L_{\exp}]_{j,k},\\
    \logscaled_{*,C}(V)&=\log_{*,\Sigma}\left(\Delta V\Delta+\frac{1}{2}\left(V^0\Sigma+\Sigma V^0\right)\right),\\
    \expscaled_{*,X}(Y)&=\Delta^{-1}\left[\exp_{*,X}(Y)-\frac{1}{2}\left(\Delta^{-2}\bbD\left(\exp_{*,X}(Y)\right)\Sigma\right.\right.\\
    &\left.\left.+\Sigma\bbD\left(\exp_{*,X}(Y)\right)\Delta^{-2}\right)\right]\Delta^{-1},
    \end{align}
    where $S=U\diag\left(\lambda_1,\ldots,\lambda_n\right)U^\top$, $L_{\exp}$ is the Loewner matrix of $\exp_{*,S}$, and $\vecone$ is the vector of all ones. Here, $\log_*$ and $\exp_*$ are given by the Daleckii--Krein formula in \cref{eq:ch2-symmetric-matrix-function-differential,eq:ch2-loewner-matrix}, while $\diag(\cdot):\bbR{n}\to\bbDspace{n}$ returns a diagonal matrix from an input vector. The remaining auxiliary quantities are
    \begin{equation}
    \Delta=\bbD(\Sigma)^{\frac{1}{2}},
    \qquad
    V^0=-2\diag\left((I_n+\Sigma)^{-1}\Delta V\Delta\vecone\right).
    \end{equation}
    \item \mypara{Permutation.} Let $\perm{n}$ be the group of permutation matrices $P_\sigma=\left[\delta_{i,\sigma(j)}\right]_{1\leq i,j\leq n}$ associated with permutations $\sigma$, and let $\mathcal{D}^{\pm}(n)=\left\{\diag\left(\varepsilon_1,\ldots,\varepsilon_n\right)\mid\varepsilon\in\{-1,1\}^n\right\}$ be the group of diagonal matrices with entries in $\{-1,1\}$. \citet[Thm.~1.1]{thanwerdas2024permutation} showed that the largest congruence action on full-rank correlation matrices is the action of signed permutation matrices:
    \begin{equation}
    \begin{aligned}
    \star:(A,C)&\in\singperm{n}\times\cor{n}\longmapsto ACA^\top\in\cor{n},\\
    \singperm{n}&=\mathcal{D}^{\pm}(n)\perm{n}.
    \end{aligned}
    \end{equation}
    As both $\logscaled_{*}$ and $\offlog_{*}$ are permutation-equivariant \citep[Thms.~2.2(1) and 3.6(1)]{thanwerdas2024permutation}, permutation-invariant metrics over the correlation manifold can be induced by permutation-invariant inner products over $\hol{n}$ and $\rzero{n}$, respectively.
    \item \mypara{Invariant inner products on $\hol{n}$.} For $n\geq4$, permutation-invariant inner products on $\hol{n}$ are \citep[Thm.~8.7]{thanwerdas2022riemannian}
    \begin{equation}
    \begin{aligned}
    \holinner{X_1}{X_2}
    &=
    \alpha\tr(X_1X_2)+\beta\Sum\left(X_1X_2\right)\\
    &\quad+\gamma\Sum(X_1)\Sum(X_2),\quad \forall X_1,X_2\in\hol{n},
    \end{aligned}
    \end{equation}
    with $\alpha>0$, $2\alpha+(n-2)\beta>0$, and $\alpha+(n-1)(\beta+n\gamma)>0$. For $n=3$, permutation-invariant inner products have the same form with $\alpha=0$:
    \begin{equation}
    \begin{aligned}
    \holinner{X_1}{X_2}
    &=
    \beta\Sum(X_1X_2)+\gamma\Sum(X_1)\Sum(X_2),\\
    &\quad \text{with } \beta>0 \text{ and } \beta+3\gamma>0.
    \end{aligned}
    \end{equation}
    For $n=2$, they have the same form with $\alpha=\beta=0$:
    \begin{equation}
    \holinner{X_1}{X_2}=\gamma\Sum(X_1)\Sum(X_2),\quad \text{with } \gamma>0.
    \end{equation}
    \item \mypara{Invariant inner products on $\rzero{n}$.} For $n\geq4$, permutation-invariant inner products on $\rzero{n}$ are \citep[Thm.~4.2]{thanwerdas2024permutation}
    \begin{equation}
    \begin{aligned}
    \rzeroinner{Y_1}{Y_2}
    &=
    \alpha\tr(Y_1Y_2)+\delta\tr(\bbD(Y_1)\bbD(Y_2))\\
    &\quad+\zeta\tr(Y_1)\tr(Y_2),\quad \forall Y_1,Y_2\in\rzero{n},
    \end{aligned}
    \end{equation}
    with $\alpha>0$, $n\alpha+(n-2)\delta>0$, and $n\alpha+(n-1)(\delta+n\zeta)>0$. For $n=3$, the permutation-invariant inner products have the same form with $\alpha=0$. For $n=2$, they have the same form with $\alpha=\delta=0$.
    \item \mypara{OLM and LSM invariance.} Combining the permutation-equivariant diffeomorphisms with the above invariant inner products gives permutation-invariant OLM and LSM. As shown by \citet[Thm.~2.7]{thanwerdas2024permutation}, OLM is further invariant under signed permutations when $\beta=\gamma=0$, in which case the associated $\inner{\cdot}{\cdot}^{(\alpha,0,0)}$ reduces to the scaled canonical Euclidean inner product:
    \begin{equation}
    \inner{V}{W}^{(\alpha,0,0)}=\alpha\inner{V}{W},\quad \forall V,W\in\hol{n}.
    \end{equation}
    \Needspace{4\baselineskip}
    In this dissertation, we assume that $\holinner{\cdot}{\cdot}$ and $\rzeroinner{\cdot}{\cdot}$ are the canonical Euclidean inner products. For $n\leq3$, these inner products remain permutation invariant; the dimension-specific forms above account for redundancies among the displayed trace and sum terms.
    \item \mypara{Inverse consistency.} We briefly review inverse-consistency, a property exclusive to LSM. The cor-inversion is defined as $\calI:\cor{n}\ni C\longmapsto\coropt\left(C^{-1}\right)\in\cor{n}$ \citep[Def.~1.4]{thanwerdas2024permutation}. It corresponds to the matrix inversion $\inv:\spd{n}\ni \Sigma\longmapsto\Sigma^{-1}\in\spd{n}$, as represented by the following commuting diagram:
    \begin{equation}
    \begin{tikzcd}
    \spd{n} \arrow[r, "\inv"] \arrow[d, "\coropt"'] & \spd{n} \arrow[d, "\coropt"] \\
    \cor{n} \arrow[r, "\calI"] & \cor{n}
    \end{tikzcd}
    \end{equation}
    As shown by \citet[Thm.~1.7]{thanwerdas2024permutation}, LSM enjoys inverse-consistency:
    \begin{equation}
    \logscaled(\calI(C))=-\logscaled(C),\quad \forall C\in\cor{n}.
    \end{equation}
    \item \mypara{Vector structure.} ECM, LECM, OLM, and LSM are all pulled back from Euclidean vector spaces. It is therefore natural to inherit vector addition and scalar multiplication from their prototype spaces. For each corresponding isometry $\phi$, these operations take the form
    \begin{equation}
    C \oplus C' = \phi^{-1}\left(\phi(C)+\phi(C')\right),
    \qquad
    t\odot C = \phi^{-1}\left(t\phi(C)\right),
    \end{equation}
    with $t\in\bbRscalar$.
\end{enumerate}

\FloatBarrier
\begin{table}[t]
\centering
\caption{Isometric prototype spaces and diffeomorphisms on the correlation manifold.}
\label{tab:ch2-correlation-maps}
\resizebox{\textwidth}{!}{%
\begin{tabular}{cccc}
\toprule
\textbf{Metric} & \textbf{Prototype space} & \textbf{Diffeomorphisms} & \textbf{Properties} \\
\midrule
\makecell{ECM \\ \citep{thanwerdas2022theoretically}} &
$\LTone{n}=\LTzero{n}+I_n$ &
\makecell{$\Theta:C\in\cor{n}\longmapsto\bbD(\chol(C))^{-1}\chol(C)\in\LTone{n}$ \\ $\Theta^{-1}=\coropt\circ\chol^{-1}:\LTone{n}\longrightarrow\cor{n}$} &
Null curvature \\
\midrule
\makecell{LECM \\ \citep{thanwerdas2022theoretically}} &
$\LTzero{n}$ &
\makecell{$\log\circ\Theta:\cor{n}\longrightarrow\LTzero{n}$ \\ $(\log\circ\Theta)^{-1}=\coropt\circ\chol^{-1}\circ\exp:\LTzero{n}\longrightarrow\cor{n}$} &
Null curvature \\
\midrule
\makecell{OLM \\ \citep{thanwerdas2024permutation}} &
$\hol{n}$ &
\makecell{$\offlog:C\in\cor{n}\longmapsto(\off\circ\log)(C)\in\hol{n}$ \\ $(\offlog)^{-1}=\offexp:H\in\hol{n}\longmapsto\exp\left(\dplus(H)+H\right)\in\cor{n}$} &
\makecell{Permutation-invariance \\ Null curvature} \\
\midrule
\makecell{LSM \\ \citep{thanwerdas2024permutation}} &
$\rzero{n}$ &
\makecell{$\logscaled:C\in\cor{n}\longmapsto\log(\dstar(C)C\dstar(C))\in\rzero{n}$ \\ $(\logscaled)^{-1}=\expscaled:R\in\rzero{n}\longmapsto\coropt(\exp(R))\in\cor{n}$} &
\makecell{Permutation-invariance \\ Inverse-consistency \\ Null curvature} \\
\midrule
\makecell{PHCM \\ \citep{thanwerdas2022theoretically}} &
$\bbPHS{n-1}$ &
\makecell{$\chol:\cor{n}\longrightarrow\chocor{n}\cong\bbPHS{n-1}$ \\ $\chol^{-1}:\chocor{n}\cong\bbPHS{n-1}\longrightarrow\cor{n}$} &
\makecell{Nonpositive \\ sectional curvature} \\
\bottomrule
\end{tabular}}
\end{table}

\begin{table}[t]
\centering
\caption{Vector operations and Riemannian operators under ECM and LECM.}
\label{tab:ch2-correlation-ecm-lecm}
\resizebox{\textwidth}{!}{%
\begin{tabular}{ccc}
\toprule
\textbf{Operation} & \textbf{ECM} & \textbf{LECM} \\
\midrule
$C\oplus C'$ & $(\isoecm)^{-1}\left(\isoecm(C)+\isoecm(C')\right)$ & $(\log\circ\Theta)^{-1}\left(\log\circ\Theta(C)+\log\circ\Theta(C')\right)$ \\
$t\odot C$ & $(\isoecm)^{-1}\left(t\isoecm(C)\right)$ & $(\log\circ\Theta)^{-1}\left(t\log\circ\Theta(C)\right)$ \\
$g_C(V,W)$ & $\inner{\Theta_{*,C}(V)}{\Theta_{*,C}(W)}$ & $\inner{(\log\circ\Theta)_{*,C}(V)}{(\log\circ\Theta)_{*,C}(W)}$ \\
$\rieexp_C(V)$ & $\Theta^{-1}\left(\Theta(C)+\Theta_{*,C}(V)\right)$ & $(\log\circ\Theta)^{-1}\left(\log\circ\Theta(C)+(\log\circ\Theta)_{*,C}(V)\right)$ \\
$\rielog_C(C')$ & $\Theta^{-1}_{*,\Theta(C)}\left(\Theta(C')-\Theta(C)\right)$ & $(\log\circ\Theta)^{-1}_{*,\log\circ\Theta(C)}\left(\log\circ\Theta(C')-\log\circ\Theta(C)\right)$ \\
$\gamma(t;C,C')$ & $\Theta^{-1}\left((1-t)\Theta(C)+t\Theta(C')\right)$ & $(\log\circ\Theta)^{-1}\left((1-t)\log\circ\Theta(C)+t\log\circ\Theta(C')\right)$ \\
$\dist(C,C')$ & $\norm{\Theta(C)-\Theta(C')}$ & $\norm{\log\circ\Theta(C)-\log\circ\Theta(C')}$ \\
Fréchet mean & $\Theta^{-1}\left(\frac{1}{N}\sum_{i=1}^{N}\Theta(C_i)\right)$ & $(\log\circ\Theta)^{-1}\left(\frac{1}{N}\sum_{i=1}^{N}(\log\circ\Theta)(C_i)\right)$ \\
Curvature & $0$ & $0$ \\
$\pt{C}{C'}(V)$ & $(\Theta_{*,C'})^{-1}\left(\Theta_{*,C}(V)\right)$ & $((\log\circ\Theta)_{*,C'})^{-1}\left((\log\circ\Theta)_{*,C}(V)\right)$ \\
\bottomrule
\end{tabular}}
\end{table}

\begin{table}[t]
\centering
\caption{Vector operations and Riemannian operators under OLM and LSM.}
\label{tab:ch2-correlation-olm-lsm}
\resizebox{\textwidth}{!}{%
\begin{tabular}{ccc}
\toprule
\textbf{Operation} & \textbf{OLM} & \textbf{LSM} \\
\midrule
$C\oplus C'$ & $\offexp\left(\offlog(C)+\offlog(C')\right)$ & $\expscaled\left(\logscaled(C)+\logscaled(C')\right)$ \\
$t\odot C$ & $\offexp\left(t\offlog(C)\right)$ & $\expscaled\left(t\logscaled(C)\right)$ \\
$g_C(V,W)$ & $\holinner{\offlog_{*,C}(V)}{\offlog_{*,C}(W)}$ & $\rzeroinner{\logscaled_{*,C}(V)}{\logscaled_{*,C}(W)}$ \\
$\rieexp_C(V)$ & $\offexp\left(\offlog(C)+\offlog_{*,C}(V)\right)$ & $\expscaled\left(\logscaled(C)+\logscaled_{*,C}(V)\right)$ \\
$\rielog_C(C')$ & $\offexp_{*,\offlog(C)}\left(\offlog(C')-\offlog(C)\right)$ & $\expscaled_{*,\logscaled(C)}\left(\logscaled(C')-\logscaled(C)\right)$ \\
$\gamma(t;C,C')$ & $\offexp\left((1-t)\offlog(C)+t\offlog(C')\right)$ & $\expscaled\left((1-t)\logscaled(C)+t\logscaled(C')\right)$ \\
$\dist(C,C')$ & $\holnorm{\offlog(C)-\offlog(C')}$ & $\rzeronorm{\logscaled(C)-\logscaled(C')}$ \\
Fréchet mean & $\offexp\left(\frac{1}{N}\sum_{i=1}^{N}\offlog(C_i)\right)$ & $\expscaled\left(\frac{1}{N}\sum_{i=1}^{N}\logscaled(C_i)\right)$ \\
Curvature & $0$ & $0$ \\
$\pt{C}{C'}(V)$ & $(\offlog_{*,C'})^{-1}\left(\offlog_{*,C}(V)\right)$ & $(\logscaled_{*,C'})^{-1}\left(\logscaled_{*,C}(V)\right)$ \\
\bottomrule
\end{tabular}}
\end{table}

\FloatBarrier
\mypara{PHCM \citep[Def.~4.3 and Thm.~4.4]{thanwerdas2022theoretically}.} It is defined through the Cholesky decomposition and a product of open-hemisphere models of hyperbolic space. For a correlation matrix $C\in\cor{n}$, let $L=\chol(C)$. For $k=2,\ldots,n$, define the nonzero part of the $k$-th row of $L$ by
\begin{equation}
\ell_k=(L_{k1},\ldots,L_{kk})\in\hs{k-1},
\qquad
\hs{k-1}=\left\{x\in\bbR{k}\mid\norm{x}=1,\ x_k>0\right\}.
\end{equation}
Thus, $\chocor{n}$ is identified with the product of $n-1$ open hemispheres, denoted by $\bbPHS{n-1}=\prod_{i=1}^{n-1}\hs{i}$. The first row, for which $L_{11}=1$ and $\hs{0}=\{1\}$, is trivial and omitted from the product. PHCM is the pullback by the Cholesky decomposition of the product metric on $\prod_{i=1}^{n-1}(\hs{i},\alpha_i g^{\hs{i}})$, where each $\alpha_i$ is a positive weight and $g^{\hs{i}}$ denotes the metric tensor on $\hs{i}$. In particular, PHCM with all weights equal to 1 is called the canonical PHCM, on which we focus below. Given $C\in\cor{n}$ and $L=\chol(C)\in\chocor{n}$, define
\begin{equation}
\Psi=\psi^1\times\cdots\times\psi^{n-1}:
\chocor{n}\to\prod_{i=1}^{n-1}\hs{i},
\end{equation}
where
\begin{equation}
\psi^i(L)=\left(L_{i+1,1},\ldots,L_{i+1,i+1}\right)\in\hs{i}.
\end{equation}
For $Z\in T_L\chocor{n}$, its differential is the corresponding row extraction,
\begin{equation}
\psi^i_{*,L}(Z)=\left(Z_{i+1,1},\ldots,Z_{i+1,i+1}\right)
\in T_{\psi^i(L)}\hs{i},
\qquad
\Psi_{*,L}=\psi^1_{*,L}\times\cdots\times\psi^{n-1}_{*,L}.
\end{equation}
The Riemannian operators under PHCM are obtained from the product geometry and the geometry of each $\hs{i}$. Let $C'\in\cor{n}$, $L'=\chol(C')$, and $V,W\in T_C\cor{n}$. Write $a_i=\psi^i(L)$, $a_i'=\psi^i(L')$, and $\xi_i(V)=\psi^i_{*,L}(\chol_{*,C}(V))$. Then
\begin{align}
g_C(V,W)
&=
\left\langle
\bbD(L)^{-1}L\left(L^{-1}VL^{-\top}\right)_{\frac{1}{2}},
\right.\notag\\
&\qquad\left.
\bbD(L)^{-1}L\left(L^{-1}WL^{-\top}\right)_{\frac{1}{2}}
\right\rangle, \\
\rieexp_C(V)
&=
\chol^{-1}\left(
\Psi^{-1}\left(
\begin{gathered}
\rieexp_{a_1}^{\hs{1}}\left(\xi_1(V)\right),\ldots,\\
\rieexp_{a_{n-1}}^{\hs{n-1}}\left(\xi_{n-1}(V)\right)
\end{gathered}
\right)\right), \\
\rielog_C(C')
&=
\chol^{-1}_{*,L}\left(
(\Psi_{*,L})^{-1}\left(
\begin{gathered}
\rielog_{a_1}^{\hs{1}}\left(a_1'\right),\ldots,\\
\rielog_{a_{n-1}}^{\hs{n-1}}\left(a_{n-1}'\right)
\end{gathered}
\right)\right), \\
\gamma(t;C,C')
&=
\chol^{-1}\left(
\Psi^{-1}\left(
\begin{gathered}
\gamma^{\hs{1}}\left(t;a_1,a_1'\right),\ldots,\\
\gamma^{\hs{n-1}}\left(t;a_{n-1},a_{n-1}'\right)
\end{gathered}
\right)\right), \\
\dist(C,C')^2
&=
\sum_{i=1}^{n-1}
\arccosh\left(
1+\frac{1-\inner{a_i}{a_i'}}{L_{i+1,i+1}L'_{i+1,i+1}}
\right)^2.
\end{align}
Here, $\rielog^{\hs{i}}$, $\rieexp^{\hs{i}}$, and $\gamma^{\hs{i}}$ are the corresponding operators on $\hs{i}$; the distance and these closed forms follow from the Lorentz--hemisphere isometry \citep[Thm.~4.2]{thanwerdas2022theoretically}.

\subsection{Grassmannian Manifolds}

The Grassmannian has been widely applied in machine learning, ranging from action recognition \citep{huang2018building} to question answering \citep{nguyen2023building}, shape generation \citep{yataka2023grassmann}, image classification \citep{wang2023get}, and signal analysis \citep{wang2024grassatt}.

The \emph{Grassmannian manifold} is the set of $p$-dimensional subspaces of $\bbR{n}$ \citep{bendokat2024grassmann}. It has two matrix representations: the \emph{projector perspective (PP)} and the \emph{orthonormal-basis perspective (ONB)}:
\begin{equation}
\begin{aligned}
\graspp{p,n}
&=
\left\{P \in \sym{n} \mid P^2=P,\quad \rank(P)=p\right\}, \\
\grasonb{p,n}
&=
\left\{[U] \mid [U]:=\left\{\widetilde{U} \in \stiefel{p,n} \mid \widetilde{U}=UR,\quad R \in \orth{p}\right\}\right\},
\end{aligned}
\end{equation}
where $\sym{n}$ is the Euclidean space of symmetric matrices, $\stiefel{p,n}$ is the Stiefel manifold, and $\orth{p}$ is the orthogonal group. By abuse of notation, we use $[U]$ and $U$ interchangeably for elements of $\grasonb{p,n}$, where the $n \times p$ column-wise orthonormal matrix $U$ is taken as a representative of an equivalence class. \citet{helmke2012optimization} show that the ONB perspective is diffeomorphic to the PP representation by
\begin{equation}
\label{eq:iso_grass}
\pi:\grasonb{p,n}\ni U\mapsto UU^\top\in\graspp{p,n}.
\end{equation}
As shown by \citet[Sec.~3.2]{nguyen2022gyro} and \citet[Sec.~2.3.1]{nguyen2023building}, the Grassmannian admits gyro-structures defined by \crefrange{eq:ch2-riem-gyro-addition}{eq:ch2-gyro-distance}.

Under the ONB perspective, tangent vectors are represented by horizontal lifts. Given an orthogonal complement $U_{\perp}\in\stiefel{n-p,n}$ of $U$, every tangent vector $\Delta\in T_U\grasonb{p,n}$ can be written as
\begin{equation}
\Delta=U_{\perp}B,
\qquad
B\in\bbR{(n-p)\times p}.
\end{equation}
Under the PP perspective, every point can be written as $P=O\idpp O^\top$ with $O\in\orth{n}$, and every tangent vector $A\in T_P\graspp{p,n}$ can be written as
\begin{equation}
A
=
O
\begin{bmatrix}
\bbzero & B^\top \\
B & \bbzero
\end{bmatrix}
O^\top,
\qquad
B\in\bbR{(n-p)\times p}.
\end{equation}
\cref{tab:ch2-grassmann-riemannian-operators} and \cref{tab:ch2-grassmann-gyro-operators} summarize the Riemannian and gyro operators, respectively; their formulas use the \emph{singular value decomposition (SVD)}. In these tables, $U,V\in\grasonb{p,n}$ and $\Delta,\Xi\in T_U\grasonb{p,n}$ refer to the ONB perspective, while $P,Q\in\graspp{p,n}$ and $A,A_1,A_2\in T_P\graspp{p,n}$ refer to the PP perspective. To distinguish the two perspectives, PP Riemannian operators are marked by a tilde, such as $\widetilde{\rielog}$ and $\widetilde{\rieexp}$. The differential of $\pi$ is $\pi_{*,U}(\Delta)=U\Delta^\top+\Delta U^\top$. For gyro operators, define $\overline{P}=\widetilde{\rielog}_{\idpp}(P)$. Let $\Omega_U=[\overline{UU^\top},\idpp]$ and $\Omega_P=[\overline{P},\idpp]$. The identities are $\idonb=\left[I_p,\bbzero\right]^\top$ and $\idpp=\idonb\idonb^\top$.

\begin{table}[t]
\centering
\caption{Riemannian operators on the Grassmannian under ONB and PP.}
\label{tab:ch2-grassmann-riemannian-operators}
\resizebox{\textwidth}{!}{%
\begin{tabular}{ccc}
\toprule
\textbf{Operator} & \textbf{ONB: $\grasonb{p,n}$} & \textbf{PP: $\graspp{p,n}$} \\
\midrule
\makecell{$g_U(\Delta,\Xi)$ \\ or $\widetilde{g}_P(A_1,A_2)$} &
$g_U(\Delta,\Xi)=\inner{\Delta}{\Xi}$ &
$\widetilde{g}_P(A_1,A_2)=\frac{1}{2}\inner{A_1}{A_2}$ \\
\midrule
\makecell{$\dist(U,V)$ \\ or $\widetilde{\dist}(P,Q)$} &
\makecell{$\norm{\arccos(\Sigma)}$ \\ $U^\top V\stackrel{\mathrm{SVD}}{:=}O\Sigma R^\top$} &
$\frac{1}{2\sqrt{2}}\norm{\log\left(\left(I_n-2Q\right)\left(I_n-2P\right)\right)}$ \\
\midrule
\makecell{$\rielog_U V$ \\ or $\widetilde{\rielog}_P Q$} &
\makecell{$O\arctan(\Sigma)R^\top$ \\ $\left(I_n-UU^\top\right)V\left(U^\top V\right)^{-1}\stackrel{\mathrm{SVD}}{:=}O\Sigma R^\top$} &
$\frac{1}{2}\left[\log\left(\left(I_n-2Q\right)\left(I_n-2P\right)\right),P\right]$ \\
\midrule
\makecell{$\rieexp_U \Delta$ \\ or $\widetilde{\rieexp}_P A$} &
\makecell{$UR\cos(\Sigma)R^\top+O\sin(\Sigma)R^\top$ \\ $\Delta\stackrel{\mathrm{SVD}}{:=}O\Sigma R^\top$} &
$\exp\left([A,P]\right)P\exp\left(-[A,P]\right)$ \\
\midrule
\makecell{$\gamma(t;U,V)$ \\ or $\widetilde{\gamma}(t;P,Q)$} &
\makecell{$UR\cos(t\Sigma)R^\top+O\sin(t\Sigma)R^\top$ \\ $\rielog_U V\stackrel{\mathrm{SVD}}{:=}O\Sigma R^\top$} &
$\exp\left(t[\widetilde{\rielog}_P Q,P]\right)P\exp\left(-t[\widetilde{\rielog}_P Q,P]\right)$ \\
\midrule
\makecell{$\pt{U}{V}(\Delta)$ \\ or $\widetilde{\pt{P}{Q}}(A)$} &
\makecell{$\left(\begin{bmatrix}UR & O\end{bmatrix}\begin{bmatrix}-\sin(\Sigma) \\ \cos(\Sigma)\end{bmatrix}O^\top+I_n-OO^\top\right)\Delta$ \\ $\rielog_U V\stackrel{\mathrm{SVD}}{:=}O\Sigma R^\top$} &
$\exp\left([\widetilde{\rielog}_P Q,P]\right)A\exp\left(-[\widetilde{\rielog}_P Q,P]\right)$ \\
\midrule
\makecell{$\fm(\{U_i\})$ \\ or $\widetilde{\fm}(\{P_i\})$} &
Karcher flow &
Karcher flow \\
\midrule
References &
\citep{edelman1998geometry,bendokat2024grassmann} &
\citep{bendokat2024grassmann} \\
\bottomrule
\end{tabular}}
\end{table}

\begin{table}[t]
\centering
\caption{Gyro operators on the Grassmannian under ONB and PP.}
\label{tab:ch2-grassmann-gyro-operators}
\resizebox{0.88\textwidth}{!}{%
\begin{tabular}{ccc}
\toprule
\textbf{Operator} & \textbf{ONB: $\grasonb{p,n}$} & \textbf{PP: $\graspp{p,n}$} \\
\midrule
Gyroaddition &
$U\oplusGyrONB V=\exp(\Omega_U)V$ &
$P\oplusGyrPP Q=\exp(\Omega_P)Q\exp\left(-\Omega_P\right)$ \\
\midrule
Gyro identity &
$\idonb$ &
$\idpp$ \\
\midrule
Scalar gyromultiplication &
$t\odotGyrONB U=\exp\left(t\Omega_U\right)\idonb$ &
$t\odotGyrPP P=\exp\left(t\Omega_P\right)\idpp\exp\left(-t\Omega_P\right)$ \\
\midrule
Gyro-inverse &
$\ominusGyrONB U=\exp\left(-\Omega_U\right)\idonb$ &
$\ominusGyrPP P=\exp\left(-\Omega_P\right)\idpp\exp(\Omega_P)$ \\
\midrule
References &
\citep{nguyen2023building} &
\citep{nguyen2022gyro} \\
\bottomrule
\end{tabular}}
\end{table}

\subsection{Special Orthogonal Groups}
\label{sec:ch2-special-orthogonal-groups}

The set of $n \times n$ rotation matrices forms a Lie group, known as the \emph{special orthogonal group} and denoted by $\so{n}$ \citep{loring2011introduction}:
\begin{equation}
\so{n}
=
\left\{
R \in \bbR{n \times n}
\mid
R^{\top}R=I_n,\quad \det(R)=1
\right\}.
\end{equation}
Its group operation is the matrix product, with the identity matrix as the neutral element. Any tangent vector $A \in T_R\so{n}$ can be represented as $A=RV$, with $V \in \soLieAlgebra{n}$. Here, $\soLieAlgebra{n}$ is the Lie algebra of $\so{n}$, which is the tangent space at the identity matrix, formed by the set of $n \times n$ skew-symmetric matrices:
\begin{equation}
\soLieAlgebra{n}
=
\left\{
\Omega \in \bbR{n \times n}
\mid
\Omega^{\top}=-\Omega
\right\}.
\end{equation}
The Fréchet mean can be obtained by Karcher flow \citep{manton2004globally}. Furthermore, if all rotations lie in a closed ball of radius $r<\nicefrac{\pi}{2}$, then Karcher flow converges to the unique mean \citep[Thm.~5]{manton2004globally}. Given $R,S\in\so{n}$ and tangent vectors $A,A_1,A_2\in T_R\so{n}$, \cref{tab:ch2-so-operators} summarizes all the associated operators on $\so{n}$.

\begin{table}[t]
\centering
\caption{Lie group structures and Riemannian operators on rotation matrices.}
\label{tab:ch2-so-operators}
\resizebox{\textwidth}{!}{%
\begin{tabular}{cccccccc}
\toprule
\textbf{Operator} & $R\oplus S$ & $g_R(A_1,A_2)$ & $\dist(R,S)$ & $\rielog_R S$ & $\rieexp_R(A)$ & $\gamma(t;R,S)$ & \textbf{FM} \\
\midrule
Expression &
$RS$ &
$\inner{A_1}{A_2}$ &
$\norm{\log(R^{\top}S)}$ &
$R\log(R^{\top}S)$ &
$R\exp\left(R^{\top}A\right)$ &
$R\exp\left(t\log(R^{\top}S)\right)$ &
Karcher flow \\
\midrule
References &
\multicolumn{7}{c}{\citep{manton2004globally}} \\
\bottomrule
\end{tabular}}
\end{table}

Across these matrix-manifold examples, the tables expose the Euclidean template behind the spaces used later: flat pullback metrics use ordinary addition and scaling in a chart, quotient metrics use horizontal representatives, product metrics act factor by factor, and Lie-group metrics use group translation.

\subsection{Constant-Curvature Manifolds}
\label{sec:ch2-constant-curvature-manifolds}

\begin{parisdefinition}[Constant-curvature space {\citep[Ch.~8]{oneill1983semi}}]
\label{def:ch2-constant-curvature}
A \emph{constant-curvature space (CCS)} is a complete, simply connected, $n$-dimensional Riemannian manifold of constant curvature $K$.
\end{parisdefinition}

By \citet[Cor.~8.25]{oneill1983semi}, any two CCSs with the same dimension and the same curvature $K$ are isometric. Thus, a CCS is determined up to isometry by $K$: Euclidean space corresponds to $K=0$, spherical space to $K>0$, and hyperbolic space to $K<0$. In the following, we introduce several concrete models of these spaces that are used later: the $K$-stereographic model, the $K$-radius model, and the Beltrami--Klein model.

\mypara{$K$-stereographic model \citep{bachmann2020constant}.} This model has shown success in different applications, including computer vision \citep{van2023poincare}, natural language processing \citep{ganea2018hyperbolic,shimizu2020hyperbolic}, graph learning \citep{bachmann2020constant,grover2025curvgad,grover2025spectro}, and astronomy \citep{chen2025galaxy}. It is defined as a model $\stereo{n}$ with the conformal metric
\begin{equation}
\inner{u}{v}_x^{\mathrm{st}}
=
(\lambda_x^K)^2\inner{u}{v},
\qquad
\lambda_x^K=\frac{2}{1+K\norm{x}^{2}},
\end{equation}
where $K \in \bbRscalar$ is the constant curvature and $\lambda_x^K$ is a conformal factor. In particular, $\stereo{n}$ is the scaled $\bbR{n}$ when $K=0$. It unifies the spherical projected hypersphere $\projhs{n}$, Euclidean space $\bbR{n}$, and the hyperbolic Poincar\'e ball $\pball{n}$:
\begin{equation}
\stereo{n}
=
\begin{cases}
\projhs{n}=\bbR{n}, & \text{For } K>0, \text{ spherical geometry,}\\
\bbR{n}, & \text{For } K=0, \text{ Euclidean geometry,}\\
\pball{n}=\left\{x\in\bbR{n}\mid \norm{x}^2<-\frac{1}{K}\right\}, & \text{For } K<0, \text{ hyperbolic geometry.}
\end{cases}
\end{equation}
Although $\projhs{n}=\bbR{n}$ for $K>0$, its metric is conformal to the Euclidean one. We abbreviate the $K$-stereographic model as the stereographic model. \citet[Eqs.~2--3]{bachmann2020constant} show that this model admits a gyro-structure.

\mypara{$K$-radius model \citep{skopek2020mixed}.} This model has been effective in various applications \citep{chami2019hyperbolic,chen2022fully,bdeir2024fully,pal2025compositional,he2025lorentzian,khan2025hyperbolic}. It provides an extrinsic representation of the space with constant curvature $K \in \bbRscalar$, encompassing the sphere $\sphere{n}$, Euclidean space $\bbR{n}$, and Lorentz space $\lorentz{n}$:
\begin{equation*}
\calMK{n}
=
\begin{cases}
\sphere{n}=\left\{x\in\bbR{n+1}\mid \norm{x}^2=\frac{1}{K}\right\}, & \text{For } K>0, \text{ spherical geometry,}\\
\bbR{n}, & \text{For } K=0, \text{ Euclidean geometry,}\\
\lorentz{n}=\left\{x\in\bbR{n+1}\mid \Linner{x}{x}=\frac{1}{K}, x_t>0\right\}, & \text{For } K<0, \text{ hyperbolic geometry,}
\end{cases}
\end{equation*}
where $\Linner{x}{x}=\norm{x_s}^2-x_t^2$ is the Lorentzian quadratic form. Following the conventions of Lorentz geometry, we write $x=(x_t,x_s^\top)^\top$, where $x_t\in\bbRscalar$ is the time component and $x_s\in\bbR{n}$ is the spatial component \citep{ratcliffe2006foundations}. When $K\neq0$, the model can be written compactly as
\begin{equation}
\calMK{n}
=
\left\{
x\in\bbR{n+1}
\mid
\Kinner{x}{x}=\frac{1}{K}
\right\},
\qquad
\Kinner{\cdot}{\cdot}
=
\begin{cases}
\inner{\cdot}{\cdot}, & K>0,\\
\Linner{\cdot}{\cdot}, & K<0,
\end{cases}
\end{equation}
with the hyperbolic branch restricted to $x_t>0$. We abbreviate the $K$-radius model as the radius model. On its negative-curvature branch, we identify $\calMK{n}=\lorentz{n}$; in hyperbolic-only contexts, the specializations of $\MKoplus$, $\MKominus$, and $\MKodot$ are denoted by $\Loplus$, $\Lominus$, and $\Lodot$, respectively. Its gyro-structure is introduced later in \cref{sec:ch3-gyrogroup-approach}.

\mypara{Beltrami--Klein model.} There are five models of hyperbolic space \citep{cannon1997hyperbolic}. Apart from the above Poincar\'e ball and Lorentz models, we further study the Beltrami--Klein model:
\begin{equation*}
\klein{n}
=
\left\{
x\in\bbR{n}
\mid
\norm{x}^{2}<-\frac{1}{K}
\right\},
\qquad
\text{with }
g_{x}^{\mathbb{K}}(v,w)
=
\frac{\inner{v}{w}}{1+K\norm{x}^{2}}
-
\frac{K\inner{x}{v}\inner{x}{w}}{\left(1+K\norm{x}^{2}\right)^2},
\end{equation*}
where $K<0$ is the constant curvature and $g^{\mathbb{K}}$ is its Riemannian metric.
Although the Poincar\'e ball and Beltrami--Klein models share the same underlying set, their Riemannian metrics differ. This model admits an Einstein gyrovector space \citep[Sec.~6.18]{ungar2022analytic}.

The gyro-structures for the stereographic and Beltrami--Klein models are summarized in \cref{tab:ch2-constant-curvature-gyro-operators}. In the table, $x,y,z\in\stereo{n}$ for the stereographic model, $x,y,z\in\klein{n}$ for the Beltrami--Klein model, $t\in\bbRscalar$, and $\gamma_x^K=\left(1+K\norm{x}^2\right)^{-1/2}$ is the Einstein gamma factor. The stereographic convention is $\tank=\tanh$ for $K<0$ and $\tank=\tan$ for $K>0$.

\begin{table}[t]
\centering
\caption{Gyro operators on stereographic and Beltrami--Klein models.}
\label{tab:ch2-constant-curvature-gyro-operators}
\resizebox{\textwidth}{!}{%
\begin{tabular}{ccc}
\toprule
\textbf{Operator} & \textbf{Stereographic model $\stereo{n}$} & \textbf{Beltrami--Klein model $\klein{n}$} \\
\midrule
$x\oplus y$ &
$\displaystyle \frac{\left(1-2K\inner{x}{y}-K\norm{y}^{2}\right)x+\left(1+K\norm{x}^{2}\right)y}{1-2K\inner{x}{y}+K^2\norm{x}^{2}\norm{y}^{2}}$ &
$\displaystyle \frac{1}{1-K\inner{x}{y}}\left(x+\frac{1}{\gamma_x^K}y-K\frac{\gamma_x^K}{1+\gamma_x^K}\inner{x}{y}x\right)$ \\
$\zerovec$ &
$\zerovec$ &
$\zerovec$ \\
$t\odot x$ &
$\displaystyle \frac{\tank\left(t\tank^{-1}\left(\sqrt{|K|}\norm{x}\right)\right)}{\sqrt{|K|}}\frac{x}{\norm{x}}$ &
$\displaystyle \frac{\tanh\left(t\tanh^{-1}\left(\sqrt{-K}\norm{x}\right)\right)}{\sqrt{-K}}\frac{x}{\norm{x}}$ \\
$\ominus x$ &
$-x$ &
$-x$ \\
$\gyr[x,y]z$ &
$\displaystyle z+\frac{2\left(A_{\mathrm{st}}x+B_{\mathrm{st}}y\right)}{D_{\mathrm{st}}}$ &
$\displaystyle z+\frac{A_{\mathrm{E}}x+B_{\mathrm{E}}y}{D_{\mathrm{E}}}$ \\
\midrule
References &
\citep{ungar2022analytic} &
\citep{ungar2022analytic} \\
\bottomrule
\end{tabular}}
\end{table}

The stereographic gyration in \cref{tab:ch2-constant-curvature-gyro-operators}, following \citet[App.~C.2.6]{bachmann2020constant}, is
\begin{equation}
\label{eq:stereographic_gyration}
\gyr[x,y]z
=
z+2\frac{A_{\mathrm{st}}x+B_{\mathrm{st}}y}{D_{\mathrm{st}}},
\end{equation}
with
\begin{equation}
\begin{aligned}
A_{\mathrm{st}}
&=
-K^2\inner{x}{z}\norm{y}^{2}-K\inner{y}{z}+2K^2\inner{x}{y}\inner{y}{z},\\
B_{\mathrm{st}}
&=
-K^2\inner{y}{z}\norm{x}^{2}+K\inner{x}{z},\\
D_{\mathrm{st}}
&=
1-2K\inner{x}{y}+K^2\norm{x}^{2}\norm{y}^{2} \\
&=
\left(1-K\inner{x}{y}\right)^2
+K^2\left(\norm{x}^{2}\norm{y}^{2}-\inner{x}{y}^{2}\right)
\geq
0.
\end{aligned}
\end{equation}
The Cauchy--Schwarz inequality gives the last relation. The gyration formula applies whenever $D_{\mathrm{st}}>0$; singular positive-curvature configurations with $D_{\mathrm{st}}=0$ are excluded.
The Einstein gyration coefficients for the Beltrami--Klein model are
\begin{equation}
\begin{aligned}
A_{\mathrm{E}}
&=
K\frac{\left(\gamma_x^K\right)^2}{\gamma_x^K+1}\left(\gamma_y^K-1\right)\inner{x}{z}
-K\gamma_x^K\gamma_y^K\inner{y}{z}\\
&\quad
+2K^2\frac{\left(\gamma_x^K\right)^2\left(\gamma_y^K\right)^2}{\left(\gamma_x^K+1\right)\left(\gamma_y^K+1\right)}\inner{x}{y}\inner{y}{z},\\
B_{\mathrm{E}}
&=
K\frac{\gamma_y^K}{\gamma_y^K+1}\left(\gamma_x^K\left(\gamma_y^K+1\right)\inner{x}{z}+\left(\gamma_x^K-1\right)\gamma_y^K\inner{y}{z}\right),\\
D_{\mathrm{E}}
&=
1+\gamma_x^K\gamma_y^K\left(1-K\inner{x}{y}\right)
=1+\gamma_{x\Eoplus y}^K.
\end{aligned}
\end{equation}

\cref{tab:ch2-constant-curvature-operators} summarizes the associated Riemannian operators for the stereographic and radius models when $K\neq0$; at $K=0$, they reduce to the usual Euclidean operators. On the sphere, logarithmic maps and parallel transports are restricted away from antipodal pairs. The gyro operators of the radius model and the closed-form Riemannian operators of the Beltrami--Klein model are introduced later in \cref{sec:ch3-gyrogroup-approach}.

In \cref{tab:ch2-constant-curvature-operators}, for $K \neq 0$, the curvature-aware trigonometric functions are
\begin{equation*}
\begin{gathered}
\tank(\cdot)
=
\begin{cases}
\tan(\cdot), & K>0, \\
\tanh(\cdot), & K<0,
\end{cases}
\quad
\sink(\cdot)
=
\begin{cases}
\sin(\cdot), & K>0, \\
\sinh(\cdot), & K<0,
\end{cases}
\\
\cosk(\cdot)
=
\begin{cases}
\cos(\cdot), & K>0, \\
\cosh(\cdot), & K<0,
\end{cases}
\end{gathered}
\end{equation*}

\begin{table}[t]
\centering
\caption{Riemannian operator templates for stereographic and radius constant-curvature coordinates.}
\label{tab:ch2-constant-curvature-operators}
\resizebox{\textwidth}{!}{%
\begin{tabular}{ccc}
\toprule
\textbf{Operator} & \textbf{Stereographic: $\pball{n}$, $\projhs{n}$} & \textbf{Radius: $\lorentz{n}$, $\sphere{n}$} \\
\midrule
$\inner{u}{v}_x$ &
$(\lambda_x^K)^2\inner{u}{v}$ &
$\Kinner{u}{v}$ \\
$\dist(x,y)$ &
$\frac{2}{\sqrt{|K|}}\tank^{-1}\left(\sqrt{|K|}\norm{-x\stoplus y}\right)$ &
$\frac{1}{\sqrt{|K|}}\cosk^{-1}(K\Kinner{x}{y})$ \\
$\rieexp_x(v)$ &
$x\stoplus\left(\tank\left(\sqrt{|K|}\frac{\lambda_x^K\norm{v}}{2}\right)\frac{v}{\sqrt{|K|}\norm{v}}\right)$ &
$\cosk\left(\sqrt{|K|}\Knorm{v}\right)x+\frac{\sink\left(\sqrt{|K|}\Knorm{v}\right)}{\sqrt{|K|}\Knorm{v}}v$ \\
$\rielog_x(y)$ &
$\frac{2\tank^{-1}\left(\sqrt{|K|}\norm{-x\stoplus y}\right)}{\sqrt{|K|}\lambda_x^K}\frac{-x\stoplus y}{\norm{-x\stoplus y}}$ &
$\frac{\cosk^{-1}(\beta)}{\sqrt{\sign(K)(1-\beta^2)}}(y-\beta x)$, $\beta=K\Kinner{x}{y}$ \\
$\pt{x}{y}(v)$ &
$\frac{\lambda_x^K}{\lambda_y^K}\gyr[y,-x]v$ &
$v-\frac{K\Kinner{y}{v}}{1+K\Kinner{x}{y}}(x+y)$ \\
Fréchet mean &
\makecell{$K<0$: \citep[Alg.~1]{lou2020differentiating} \\ $K>0$: Karcher flow} &
\makecell{$K<0$: \citep[Alg.~3]{lou2020differentiating} \\ $K>0$: Karcher flow} \\
\midrule
References &
\citep{skopek2020mixed} &
\citep{skopek2020mixed} \\
\bottomrule
\end{tabular}}
\end{table}

All ratios in these tables are understood by continuous extension at removable zero denominators: in particular, $t\stodot\zerovec=\zerovec$, $\rieexp_x(0)=x$, and $\rielog_x(x)=0$. On the positive-curvature branch, logarithmic maps and parallel transports are restricted away from antipodal pairs and other stated singular configurations. For the radius model, if $x\in\calMK{n}$ and $v\in T_x\calMK{n}$, then $\Knorm{v}=\sqrt{\Kinner{v}{v}}$; for $K<0$, the Lorentzian form is positive definite only after restriction to this tangent space. These formulas show that constant-curvature neural layers can be implemented by choosing a model, applying the matching exponential, logarithmic, and transport maps, and reducing to Euclidean vector operations when $K=0$.

    \chapter{Riemannian Batch Normalization}
\label{chapter:normalization}

\section{Introduction}
\label{sec:ch3-introduction}

Motivated by the great success of normalization techniques \citep{ioffe2015batch,ba2016layer,ulyanov2016instance,wu2018group}, researchers have sought to devise normalization layers tailored for manifold-valued data. \citet{brooks2019riemannian} introduced \emph{Riemannian Batch Normalization (RBN)} designed specifically for the SPD manifold, with the ability to normalize the Riemannian mean.
\citet{kobler2022spd} extended this approach to further control the Riemannian variance. \textit{However, the above methods are constrained to AIM on the SPD manifold, limiting their applicability.} On the other hand, \citet{chakraborty2020manifoldnorm} proposed two distinct Riemannian normalization frameworks: one for Riemannian homogeneous spaces \citep[Algs.~1--2]{chakraborty2020manifoldnorm} and another for matrix Lie groups \citep[Algs.~3--4]{chakraborty2020manifoldnorm}. \textit{Nonetheless, the normalization designed for Riemannian homogeneous spaces can normalize neither the mean nor the variance, while the one for matrix Lie groups is confined to a specific type of distance \citep[Sec.~3.2]{chakraborty2020manifoldnorm}.} Meanwhile, \citet[Alg.~2]{lou2020differentiating} proposed an RBN layer for general geometries. However, similar to \citet[Algs.~1--2]{chakraborty2020manifoldnorm}, it lacks theoretical guarantees for normalizing sample statistics. Therefore, a principled Riemannian normalization framework capable of controlling both Riemannian mean and variance remains unexplored.

Given that \emph{Batch Normalization (BN)}~\citep{ioffe2015batch} serves as the foundational prototype for various types of normalization, this chapter focuses on RBN, with the potential to be extended to other normalization variants. We first present a general RBN framework for Lie groups, referred to as \emph{Lie Group Batch Normalization (LieBN)}, which can normalize both the Riemannian mean and variance under invariant metrics. To extend this normalization principle beyond Lie groups, we next introduce pseudo-reductive gyrogroups, a relaxation of classical gyrogroups that also encompasses Lie groups as special cases. Building on this algebraic structure, we develop \emph{Gyrogroup Batch Normalization (GyroBN)}, which generalizes LieBN beyond group structures. We then instantiate LieBN and GyroBN on SPD manifolds, rotation matrices, correlation matrices, the Grassmannian, and different CCSs. Extensive experiments across multiple tasks demonstrate the effectiveness of this unified design.

\section{Lie Group Batch Normalization}
\label{sec:ch3-lie-group-approach}

\subsection{Introduction}
\label{sec:intro}

\begin{figure}[t]
\centering
\includegraphics[width=\linewidth,trim={0cm 0cm 0cm 0cm}]{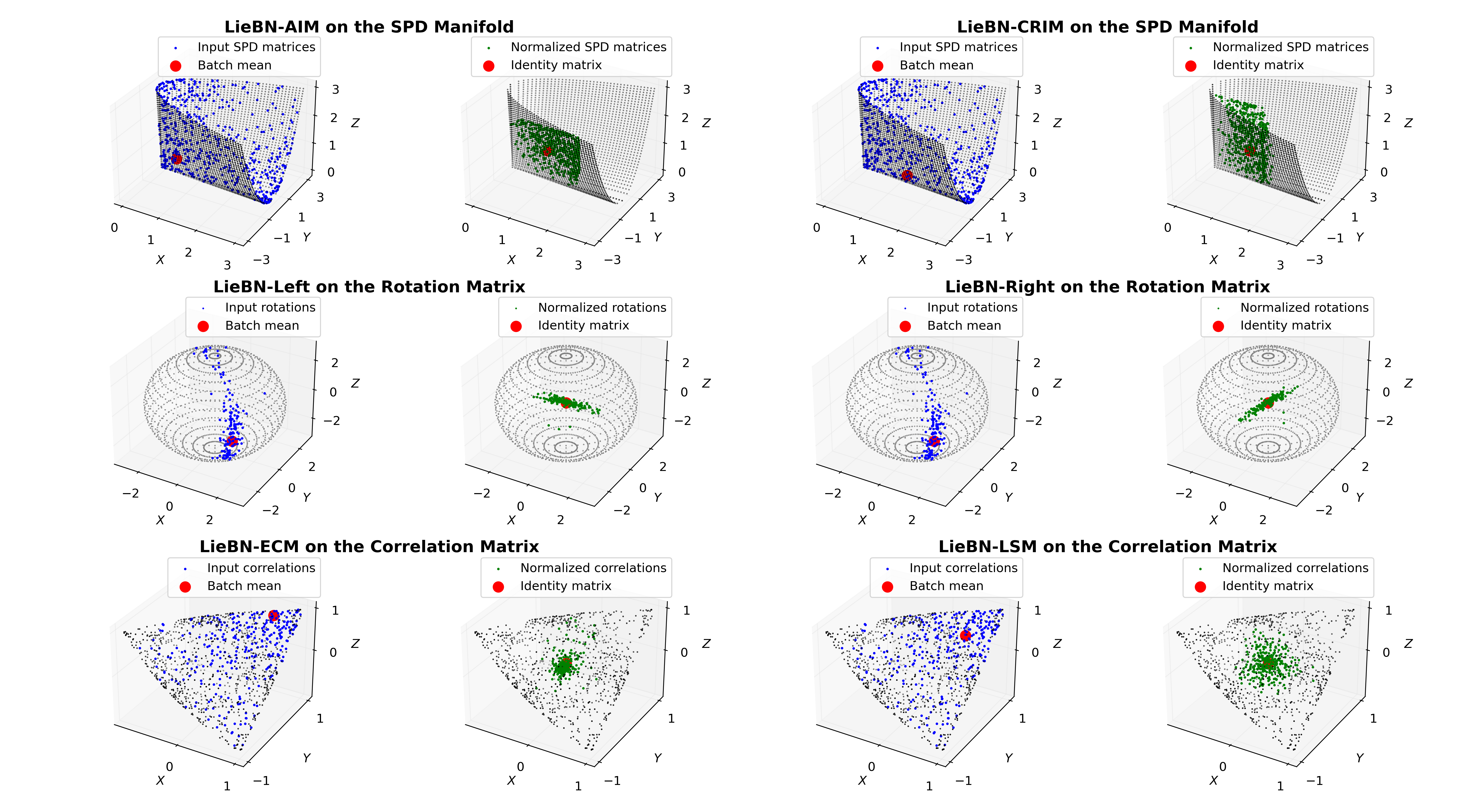}
\caption[Illustration of LieBN on the SPD, rotation, and correlation Lie groups.]{Illustration of LieBN on the SPD, rotation, and correlation Lie groups. The $2 \times 2$ SPD, $3 \times 3$ rotation, and $3 \times 3$ correlation manifolds can be embedded into $\bbR{3}$ as an open cone \citep{yair2019parallel}, a closed ball with antipodal points identified \citep{hartley2013rotation}, and an open elliptope \citep{thanwerdas2022theoretically}, respectively. LieBN is illustrated by (1) the left-invariant AIM and the proposed right-invariant CRIM geometry on the SPD manifold, (2) left or right translation under a bi-invariant metric on the rotation manifold, and (3) the bi-invariant ECM and LSM geometry on the correlation manifold. On the SPD and correlation manifolds, the batch mean and variance of the same input samples differ under different geometries. In all sub-figures, the black, blue, green, and red dots denote the boundary of the space, the input Lie group samples, the normalized samples, and the batch mean, respectively. As illustrated, our LieBN effectively normalizes the Lie group distribution.
}
\vspace{-5mm}
\label{fig:illustration}
\end{figure}

Since several manifold-valued measurements form Lie groups, such as SPD manifolds \citep{arsigny2005fast,lin2019riemannian,thanwerdas2022theoretically}, special orthogonal groups $\so{n}$ \citep{boumal2011discrete}, and full-rank correlation matrices \citep{thanwerdas2022theoretically,thanwerdas2024permutation}, we direct our attention to Lie groups. As each Lie group naturally admits left- and right-invariant metrics \citep[Ch.~1.2]{do1992riemannian}, we propose a principled framework for RBN over Lie groups under invariant metrics, referred to as LieBN. Compared to previous work, our framework provides a theoretical guarantee for normalizing the Riemannian sample mean and variance.

Empirically, we focus on the SPD, special orthogonal, and full-rank correlation manifolds. On SPD manifolds, we generalize three existing Lie group structures into parameterized ones by matrix power deformation. Additionally, we propose a novel right-invariant metric, which, to the best of our knowledge, is the \textbf{first} non-trivial right-invariant SPD metric\footnote{Although some metrics are bi-invariant, the associated group structures are commutative \citep{arsigny2005fast,lin2019riemannian}. Therefore, their bi-invariance is reduced to left-invariance.}, referred to as the \emph{Cholesky Right Invariant Metric (CRIM)}. We then instantiate our LieBN framework on SPD manifolds under these four Lie group structures. For rotation matrices, we adopt the popular bi-invariant metric \citep{boumal2011discrete}, which will induce two types of LieBN: one w.r.t. left-invariance and another w.r.t. right-invariance. On the correlation manifold, we manifest our LieBN under four recently developed correlation geometries \citep{thanwerdas2022theoretically,thanwerdas2024permutation}. To facilitate usage, we provide a LieBN toolbox compatible with PyTorch, which can be used as a drop-in module. \cref{fig:illustration} illustrates our LieBN on different geometries, while \cref{fig:liebn_examples} illustrates a minimal demo. Extensive experiments on SPD, rotation, and correlation manifolds involving radar recognition, human action recognition, and electroencephalography (EEG) classification demonstrate the effectiveness of our methods.

\begin{figure}[t]
\begin{lstlisting}[language=Python]
from LieBN import LieBNSPD, LieBNRot, LieBNCor
from LieBN.Geometry.SPD import SPDMatrices
from LieBN.Geometry.Rotations import RotMatrices
from LieBN.Geometry.Correlation import Correlation

# ==== SPD matrices ====
P_spd = SPDMatrices(n=5).random(4, 2, 5, 5)
# Implemented metrics: LEM,ALEM,LCM,AIM,CRIM
liebn_spd = LieBNSPD([2, 5, 5], metric="LEM", batchdim=[0])
output_spd = liebn_spd(P_spd)

# ==== SO(3) matrices ====
P_so3 = RotMatrices().random(4, 2, 3, 3, 3)
# LieBN-Left if is_left else -Right
liebn_so3 = LieBNRot([3, 3, 3], batchdim=[0, 1], is_left=False)
output_so3 = liebn_so3(P_so3)

# ==== Correlation matrices ====
P_cor = Correlation(n=5).random(4, 2, 5, 5)
# Implemented metrics: ECM,LECM,OLM,LSM
liebn_cor = LieBNCor([2, 5, 5], metric="ECM", batchdim=[0])
output_cor = liebn_cor(P_cor)
\end{lstlisting}
\caption{Minimal examples of applying LieBN.}
\label{fig:liebn_examples}
\end{figure}

We emphasize that our work is fundamentally distinct from \citet{brooks2019riemannian,kobler2022spd,lou2020differentiating} in theory and more general than \citet{chakraborty2020manifoldnorm}. Previous RBN methods are either designed for specific geometries \citep{brooks2019riemannian,kobler2022spd,chakraborty2020manifoldnorm} or fail to control both the mean and variance \citep{lou2020differentiating}. In contrast, our LieBN ensures the normalization of both the mean and variance across general Lie groups. In summary, our main contributions are:
\begin{itemize}
    \item
    A general LieBN framework with controllable first- and second-order moments;
    \item
    A novel right-invariant metric on the SPD manifold, which is the first non-trivial right-invariant SPD metric;
    \item
    Concrete instantiations of our LieBN framework on different geometries: four on SPD manifolds, one on rotation matrices, and four on correlation manifolds;
    \item
    Validation of the effectiveness of our LieBN framework by extensive experiments on different geometries.\footnote{The code is available at \url{https://github.com/GitZH-Chen/LieBN.git}.}
\end{itemize}

\mypara{Outline.} \cref{sec:preliminary} recalls the invariant metrics and Lie structures used by LieBN. \cref{sec:revisit_normalization} revisits Euclidean BN and RBN. \cref{sec:liebn} develops LieBN on Lie groups under left- and right-invariant metrics and establishes its statistical control. \cref{sec:manifestations} instantiates LieBN on SPD, rotation, and full-rank correlation manifolds. \cref{sec:experiments} reports experiments that validate LieBN across these geometries. Proofs are deferred to \cref{app:liebn-proofs}.

\subsection{Preliminaries}
\label{sec:preliminary}

An invariant metric can be understood as the Lie-group analogue of the Euclidean inner product being unaffected by translations. In Euclidean space, adding the same vector on the left or on the right preserves inner products, while on a Lie group the corresponding requirement is that left or right group translations preserve the Riemannian metric.

\begin{parisdefinition}[Invariance {\citep{do1992riemannian}}]
\label{def:left_invariance}
A Riemannian metric $\gleft$ over a Lie group $\{\calM, \oplus\}$ is left-invariant if, for any $x,y \in \calM$ and $V_1,V_2 \in T_y\calM$, it satisfies $\gleft _y(V_1,V_2) = \gleft_{\ltrans _x(y)} \left(\ltrans _{x*,y}(V_1), \ltrans _{x*,y}(V_2) \right)$, with $\ltrans _x(y) = x \oplus y$ as the left translation by $x$, and $\ltrans _{x*,y}$ as the differential map of $\ltrans _x$ at $y$. Similarly, a right-invariant metric $\gright$ satisfies $\gright _y(V_1,V_2) = \gright_{\rtrans _x(y)} \left(\rtrans _{x*,y}(V_1), \rtrans _{x*,y}(V_2) \right)$, with $\rtrans _x(y) = y \oplus x$ as the right translation by $x$, and $\rtrans _{x*,y}$ as the differential map of $\rtrans_x$ at $y$.
\end{parisdefinition}

Many popular matrix manifolds used in machine learning form Lie groups, including the SPD manifold, the full-rank correlation manifold, and the rotation group. Although the corresponding Riemannian geometries have been reviewed in \cref{sec:ch2-spd-manifolds,sec:ch2-full-rank-correlation-manifolds,sec:ch2-special-orthogonal-groups}, \cref{tab:riem_lie_spd,tab:riem_lie_so,tab:riem_lie_cor} provide a focused recap of their Lie structures. The notation follows \cref{tab:ch2-spd-lie-operators,tab:ch2-correlation-maps,tab:ch2-correlation-ecm-lecm,tab:ch2-correlation-olm-lsm,tab:ch2-so-operators}.

\begin{table}[t]
    \centering
    \caption[Review of SPD Lie groups and invariant metrics.]{Review of SPD Lie groups and invariant metrics.}
    \label{tab:riem_lie_spd}
    \resizebox{0.99\linewidth}{!}{%
    \begin{tabular}{c|ccc}
        \toprule
        \textbf{Operator} & \textbf{\boldmath$\biparamAIM$} & \textbf{\boldmath$\biparamLEM$} & \textbf{LCM} \\
        \midrule
        $Q \oplus P$ & $KPK^\top$ & $\mexp\left(\mlog(P)+\mlog(Q)\right)$ & $\chol^{-1}(\lfloor L + K \rfloor + \bbK \bbL)$ \\
        $\ominus P$ & $\chol^{-1}(L^{-1})$ & $\mexp\left(-\mlog(P)\right)$ & $\chol^{-1}(-\lfloor L\rfloor+\bbL^{-1})$ \\
        Identity & $I_n$ & $I_n$ & $I_n$ \\
        WFM & Karcher Flow & $\mexp\left(\sum_i w_i\mlog(P_i)\right)$ & $\clog^{-1}\left(\sum_i w_i\clog(P_i)\right)$ \\
        Invariance & Left-invariance & Bi-invariance & Bi-invariance \\
        \bottomrule
    \end{tabular}}
\end{table}

\begin{table}[t]
    \centering
    \caption[Review of the rotation Lie group and invariant metric.]{Review of the rotation Lie group and invariant metric.}
    \label{tab:riem_lie_so}
    \begin{tabular}{c|ccccc}
        \toprule
        \textbf{Group} & \textbf{$Q \oplus P$} & \textbf{$\ominus P$} & \textbf{Identity} & \textbf{WFM} & \textbf{Invariance} \\
        \midrule
        $\so{n}$ & $QP$ & $P^{-1}=P^\top$ & $I_n$ & Karcher Flow & Bi-invariance \\
        \bottomrule
    \end{tabular}
\end{table}

\begin{table}[t]
    \centering
    \caption[Review of full-rank correlation Lie groups and invariant metrics.]{Review of full-rank correlation Lie groups and invariant metrics. Here $\phi$ denotes $\Theta$, $\log \circ \Theta$, $\offlog$, and $\logscaled$ for ECM, LECM, OLM, and LSM, respectively.}
    \label{tab:riem_lie_cor}
    \begin{tabular}{c|cccc}
        \toprule
        \textbf{Operator} & \textbf{ECM} & \textbf{LECM} & \textbf{OLM} & \textbf{LSM} \\
        \midrule
        $C \oplus C'$ & \multicolumn{4}{c}{$\phi^{-1}\left(\phi(C)+\phi(C')\right)$} \\
        $\ominus C$ & \multicolumn{4}{c}{$\phi^{-1}\left(-\phi(C)\right)$} \\
        Identity & \multicolumn{4}{c}{$\phi^{-1}(\bbzero_{n\times n})$} \\
        WFM & \multicolumn{4}{c}{$\phi^{-1}\left(\sum_{i=1}^N w_i \phi(C_i)\right)$} \\
        Invariance & \multicolumn{4}{c}{Bi-invariance} \\
        \bottomrule
    \end{tabular}
\end{table}

\subsection{Revisiting Normalization}
\label{sec:revisit_normalization}

\begin{table}[t]
    \centering
    \caption{Summary of some representative RBN methods.}
    \label{tab:sum_rbn}
    \resizebox{\linewidth}{!}{%
    \begin{tabular}{ccccc}
         \toprule
         \textbf{Methods} & \makecell{\textbf{Involved} \\ \textbf{Statistics}} & \makecell{\textbf{Controllable} \\ \textbf{Mean}} & \makecell{\textbf{Controllable} \\ \textbf{Variance}}  & \textbf{Geometries}\\
         \midrule
         SPDBN \citep[Alg.~1]{brooks2019riemannian} & Mean & \cmark & \na & SPD manifolds under AIM \\
         SPDBN \citep[Alg.~1]{kobler2022controlling} & Mean+Variance & \cmark & \cmark & SPD manifolds under AIM \\
         SPDDSMBN \citep{kobler2022spd} & Mean+Variance & \cmark & \cmark & SPD manifolds under AIM \\
         ManifoldNorm \citep[Algs.~1--2]{chakraborty2020manifoldnorm} & Mean+Variance & \xmark & \xmark & Riemannian homogeneous spaces \\
        ManifoldNorm \citep[Algs.~3--4]{chakraborty2020manifoldnorm} & Mean+Variance & \cmark & \cmark & A specific Lie group structure and distance  \\
         RBN \citep[Alg.~2]{lou2020differentiating} & Mean+Variance & \xmark & \xmark & Geodesically complete manifolds \\
         \midrule
         \rowcolor{HilightColor} LieBN (Ours) & Mean+Variance & \cmark & \cmark & Lie groups\\
         \bottomrule
    \end{tabular}%
    }
\end{table}

\subsubsection{Revisiting Euclidean Normalization}
In Euclidean DNNs, normalization is a significant technique for accelerating network training by mitigating the issue of internal covariate shift \citep{ioffe2015batch}. While various normalization methods have been introduced \citep{ioffe2015batch, ba2016layer, ulyanov2016instance, wu2018group}, they all share a common purpose: the normalization of the first and second moments. We focus on BN, the prototype of other normalization variants.

Given a batch of activations $\{x_i\}_{i=1}^N$, the core operations in the standard Euclidean BN can be expressed as:
\begin{equation} \label{eq:ebn}
    \forall i \leq N, x_i \gets \gamma \frac{x_i-\mu_{b}}{\sqrt{v^2_{b}+\epsilon}} + \beta
\end{equation}
where $\mu_{b}$ is the batch mean, $v^2_b$ is the batch variance, $\gamma$ is the scaling parameter, $\beta$ is the biasing parameter, and $\epsilon$ is a small scalar for stability.

\subsubsection{Revisiting RBN}
\label{subsec:revisit_rbn}

Although endeavors have been made to develop Riemannian normalization approaches tailored for manifolds, none of the existing methods effectively handle the first and second moments in a principled manner.

\citet{brooks2019riemannian} introduced RBN over SPD manifolds under AIM. The core operations are defined as follows:
\begin{align}
    \label{eq:spdnetbn_centering}
    \text{Centering from mean } M \in \spd{n}: \bar{P}_i \gets M^{-\frac{1}{2}} P_i M^{-\frac{1}{2}}, \\
    \label{eq:spdnetbn_biasing}
    \text{Biasing towards parameter } B \in \spd{n}: \hat{P}_i \gets B^{\frac{1}{2}} \bar{P}_i B^{\frac{1}{2}},
\end{align}
where $\{P_i\}_{i=1}^N$ are SPD matrices, and $M$ is their Fréchet mean under AIM. Let
\begin{equation}
    \spdtrans{P}{Q}(S)=\rieexp_{Q} \left[\pt{P}{Q} \left(\rielog_P (S)\right)\right],
\end{equation}
where $P, Q, S \in \spd{n}$. Under AIM, \cref{eq:spdnetbn_centering,eq:spdnetbn_biasing} can be more generally expressed as
\begin{equation} \label{eq:brooksbn_prototype}
    \spdtrans{I}{B} [\spdtrans{M}{I}(P_i)].
\end{equation}
However, \cref{eq:spdnetbn_centering,eq:spdnetbn_biasing} only consider the Riemannian mean\footnote{Although not discussed in \citet{brooks2019riemannian}, the congruent actions in \cref{eq:spdnetbn_centering,eq:spdnetbn_biasing} can transfer the batch mean to a desired value under AIM.} and do not consider the Riemannian variance.
To remedy this limitation, \citet{kobler2022spd} further extended the RBN to involve the second-order statistics. The key operation is formulated as
\begin{equation} \label{eq:kobler_rbn}
    \forall i \leq N, \bar{P}_i \gets \spdtrans{I}{B}[(\spdtrans{M}{I}(P_i))^{\frac{s}{v}}],
\end{equation}
where $v^2$ is the Fréchet variance, and $s \in \bbRscalar$ is a scaling factor.
However, this method is still limited to SPD manifolds under AIM.
In parallel, \citet[Algs.~1--2]{chakraborty2020manifoldnorm} proposed a general framework for Riemannian homogeneous spaces based on \cref{eq:brooksbn_prototype}, which involves both first and second moments.
However, \cref{eq:brooksbn_prototype} does not generally guarantee control over the Riemannian mean, resulting in agnostic Riemannian statistics \citep[Sec.~3.1]{chakraborty2020manifoldnorm}. To mitigate this limitation, \citet[Algs.~3--4]{chakraborty2020manifoldnorm} further proposed normalization over matrix Lie groups. However, the discussion is limited to a certain distance, limiting the applicability of their method. On the other hand, \citet[Alg.~2]{lou2020differentiating} proposed an RBN based on a variant of \cref{eq:brooksbn_prototype}. Similarly, their approach suffers from the same problem of agnostic Riemannian statistics on general manifolds.

In summary, prevailing Riemannian normalization approaches lack a principled guarantee for controlling the first- and second-order statistics.
In contrast, our method can normalize first- and second-order statistics over general Lie groups. We summarize the above RBN methods in \cref{tab:sum_rbn}.

\subsection{LieBN}
\label{sec:liebn}
Since every Lie group naturally admits invariant metrics, we propose BN over Lie groups based on invariant metrics, referred to as LieBN. We first introduce the core operations under left-invariant metrics and then extend them to right-invariant metrics. Finally, we present the theoretical LieBN framework. In the following, we denote the neutral element in the Lie group $\calM$ as $E$\footnote{The neutral element $E$ is not necessarily the identity matrix.}.

\subsubsection{Ingredients under Left-invariant Metrics}

In this subsection, we always assume that the Lie group $\calM$ admits a left-invariant metric $\gleft$. Recalling the standard Euclidean BN \citep{ioffe2015batch} in \cref{eq:ebn}, two key points are noteworthy:
(a) the Euclidean BN implicitly assumes a Gaussian distribution and can effectively normalize the latent Gaussian distribution;
(b) the centering and biasing operations control the mean, while the scaling controls the variance. Therefore, extending BN to Lie groups requires Lie-group counterparts of the Gaussian distribution, centering, biasing, and scaling.

There are several notions of Gaussian distribution over manifolds
\citep{pennec2004probabilities,wang2006error,chakraborty2019statistics,barbaresco2021gaussian}. We adopt the intrinsic definition from \citet{chakraborty2019statistics}, which characterizes a Gaussian distribution on the Lie group $\calM$ with a mean parameter $M \in \calM$ and variance $\sigma^2$. This distribution is denoted as $\calN(M,\sigma^2)$, and its probability density function (PDF) is
\begin{equation} \label{eq:lie_gaussian}
    p\left(X \mid M, \sigma^2\right)=k(\sigma) \exp \left(-\frac{\dist(X, M)^2}{2 \sigma^2}\right),
\end{equation}
where $k(\sigma)$ is the normalizing constant and $\dist(\cdot,\cdot)$ is the geodesic distance. When $\calM$ is $\bbRscalar$ with the standard Euclidean metric, \cref{eq:lie_gaussian} reduces to the Euclidean Gaussian.

On Lie groups, the natural counterparts of addition and subtraction in \cref{eq:ebn} are group operations. Therefore, centering and biasing on Lie groups can be defined by the left translation. Additionally, we define scaling via the tangent space. Specifically, for a batch of activations $\{P_i\}_{i=1}^N \subset \calM$, we define the key operations of LieBN as follows:
\begin{align}
    \label{eq:liebn_centering}
    \text{Centering from mean } M \in \calM: \bar{P}_i \gets \ltrans_{\ominus M}(P_i), \\
    \label{eq:liebn_scaling}
    \text{Scaling: } \hat{P}_i \gets \rieexp_{E} \left [ \frac{s}{\sqrt{v^2+\epsilon}} \rielog_{E}(\bar{P}_i) \right], \\
    \label{eq:liebn_biasing}
    \text{Biasing towards parameter } B \in \calM: \tilde{P}_i \leftarrow \ltrans_B\left(\hat{P}_i\right),
\end{align}
where $M$ is the Fréchet mean, $v^2$ is the Fréchet variance, $\ominus M \in \calM$ is the group inverse of $M$, $\ltrans_{\ominus M}$ and $\ltrans_{B}$ are left translations ($\ltrans_{B}(P_i)= B \oplus P_i$), and $s \in \bbRscalar \setminus \{0\}$ is a scaling parameter. The following two propositions demonstrate the above operations in normalizing mean and variance: one related to population statistics and the other related to sample statistics.

\begin{parisproposition}[Population]
    \label{props:population_gaussian}\linktoproof{population_gaussian}
    Given a random point $X$ over $\{ \calM, \oplus, \gleft \}$, and the Gaussian distribution $\calN(M,v^2)$ defined in \cref{eq:lie_gaussian}, we have the following for the population statistics:
    \par\leavevmode\vspace{-\baselineskip}
    \begin{enumerate}
        \item \label{pro:mle_m}
        (MLE of $M$) Given $\{P_i\}_{i=1}^N \subset \calM$ i.i.d. sampled from $\calN(M,v^2)$, the maximum likelihood estimator (MLE) of $M$ is the sample Fréchet mean.
        \item (Gaussian homogeneity) \label{pro:hom}
        Given $X \sim \calN(M,v^2)$ and $B \in \calM$, we have
        \begin{equation}
            \ltrans _{B}(X) \sim \calN(\ltrans_B(M),v^2).
        \end{equation}
    \end{enumerate}
\end{parisproposition}

\begin{parisproposition} [Sample]
    \label{props:samples}\linktoproof{samples}
    Given $N$ samples $\{P_i\}_{i=1}^N$ over the Lie group $\{ \calM, \oplus, \gleft \}$, define
    \begin{equation}
        \phi_{s}(P_i)=\rieexp_{E} \left [ s \rielog_{E}(P_i) \right].
    \end{equation}
    We then have the following for the sample statistics.
    \begin{itemize}
        \item Sample mean homogeneity:
        \begin{equation}
            \label{eq:hom_fm_lie_group}
            \fm\{\ltrans _{B} (P_i) \} = \ltrans _{B} (\fm\{ P_i \}), \forall B \in \calM.
        \end{equation}

        \item Controllable dispersion from $E$:
        \begin{equation}
            \label{eq:variance_lie_group}
            \sum\nolimits_{i=1}^N  w_i \dist^2(\phi_{s}(P_i), E) = s^2 \sum\nolimits_{i=1}^N w_i \dist^2(P_i, E),
        \end{equation}
    \end{itemize}
    where $\{w_i\}_{i=1}^N$ are weights satisfying a convexity constraint, \ie $\forall i, w_i>0$ and $\sum_i w_i=1$.
\end{parisproposition}

\cref{props:population_gaussian} and \cref{eq:hom_fm_lie_group} imply that our centering and biasing in \cref{eq:liebn_centering,eq:liebn_biasing} can transfer the sample and population mean. As the post-centering mean is $E$, \cref{eq:variance_lie_group} implies that \cref{eq:liebn_scaling} can control the sample variance. More interestingly, the latent Gaussian distribution can be transferred under some geometries, such as SPD manifolds under LEM and LCM.

\begin{parisremark}
The MLE of the mean of the Gaussian distribution has been examined in several previous works \citep{said2017riemannian,chakraborty2019statistics,chakraborty2020manifoldnorm}. However, these studies primarily focus on particular manifolds or specific metrics. In contrast, our contribution lies in presenting a general result for Lie groups.
\end{parisremark}
\begin{parisremark}
    While \cref{eq:lie_gaussian} appeared in \citet{kobler2022controlling}, the authors only focus on SPD manifolds under AIM. The transformation of the population under their proposed RBN remains unexplored as well. Besides, while \citet{chakraborty2020manifoldnorm} analyzed the population properties for their RBN over matrix Lie groups, their results were confined within a specific distance. In contrast, our work provides a more extensive examination, encompassing both population and sample properties of our LieBN in a general manner.
\end{parisremark}

\subsubsection{Ingredients under Right-invariant Metrics}
The key insight beneath \cref{eq:liebn_centering,eq:liebn_biasing,props:population_gaussian,props:samples} is that left translation is an isometry under left-invariant metrics. Similarly, right translation is an isometry under right-invariant metrics. Therefore, it can be used for centering and biasing under right-invariant metrics. Following the previous notations, we define the centering and biasing under a right-invariant metric $\gright$ as
\begin{align}
    & \text{centering to } E \text{: }
    \bar{P}_i \gets \rtrans _{\ominus M}(P_i),\\
    & \text{ biasing towards $B$: }
    \tilde{P}_i \gets \rtrans _{B}(\hat{P}_i).
\end{align}

Similar to the case under left-invariant metrics, \cref{props:population_gaussian,props:samples} can be easily extended to right-invariant metrics. Notably, the proofs for the MLE of $M$ in \cref{props:population_gaussian} and controllable dispersion in \cref{props:samples} can be directly applied to the right-invariant metric. Therefore, we only show the homogeneity in the following proposition.

\begin{parisproposition}
    \label{props:core_operation_right}\linktoproof{core_operation_right}
    Given a random point $X \sim \calN(M,v^2)$ over $\{ \calM, \oplus, \gright \}$, $B \in \calM$, and $N$ samples $\{P_i\}_{i=1}^N$ over $\calM$, we have:
    \par\leavevmode\vspace{-\baselineskip}
    \begin{enumerate}
        \item Gaussian homogeneity: $\rtrans_{B}(X) \sim \calN(\rtrans_B(M),v^2)$;\label{pro:hom_gauss_right}
        \item
        Sample homogeneity: $\fm\{\rtrans_{B} (P_i) \} = \rtrans_{B} (\fm\{ P_i \})$.
    \end{enumerate}
\end{parisproposition}

\subsubsection{LieBN under Invariant Metrics}

With the above ingredients, \cref{alg:liebn} presents our theoretical LieBN framework. Similar to \citet{ioffe2015batch}, we use the moving average to update the running statistics. For a bi-invariant metric, LieBN can be implemented using either left or right translation. If the Lie group is commutative, LieBN under left and right translations are equivalent. \cref{tab:liebn_types} summarizes the LieBN types under different conditions.

The centering and biasing in Euclidean BN correspond to the group action of $\bbRscalar$. From a geometric perspective, the standard Euclidean metric is invariant under this group operation. Consequently, it is not surprising that our LieBN algorithm naturally generalizes the standard Euclidean BN.

\begin{parisproposition} \label{prop:liebn_natural_extension_ebn}\linktoproof{liebn_natural_extension_ebn}
    The LieBN algorithm presented in \cref{alg:liebn} is equivalent to the standard Euclidean BN when $\calM= \bbR{n}$, both during the training and testing phases.
\end{parisproposition}

\begin{table}[t]
    \centering
    \caption{Summary of LieBN types.}
    \label{tab:liebn_types}
    \resizebox{0.8\linewidth}{!}{
    \begin{tabular}{c|ccc|c}
        \toprule
        \textbf{Commutativity} & \multicolumn{3}{c|}{Non-commutative} & Commutative \\
        \midrule
        \textbf{Invariance} & Left & Right & Bi & Left = Right = Bi  \\
        \midrule
        \textbf{LieBN Types} & Left & Right & Left \& Right & Left = Right \\
        \bottomrule
    \end{tabular}
    }
\end{table}%

\begin{algorithm}[t] \SetKwInOut{Input}{Input}\SetKwInOut{Output}{Output}\SetKwInOut{Parameters}{Parameters}
\caption{Lie Group Batch Normalization (LieBN)}
\label{alg:liebn}
\Input{
A batch of activations $\{P_i\}_{i=1}^N$ over Lie groups $\{ \calM, \oplus, g\}$, a small positive constant $\epsilon$, and momentum $\eta \in [0,1]$, running mean $M_r=E$, running variance $v^2_r=1$, biasing parameter $B \in \calM$, and scaling parameter $s \in \bbRscalar \setminus \{0\}$.
}

\Output{Normalized activations $\{\tilde{P}_i\}_{i=1}^N$.}
\BlankLine
\uIf{training}{
    Compute batch mean $M_b$ and variance $v_b^2$\\
    Update running statistics:
    $M_r \gets \wfm(\{1-\eta,\eta\},\{M_r,M_b\})$
    $v^2_r \gets (1-\eta)v^2_r + \eta v^2_b$

    Use the batch statistics, $M \gets M_b, v^2 \gets v^2_b$
}
\Else{Use the running statistics, $M \gets M_r, v^2 \gets v^2_r$}

\For{$i \gets 1$ \KwTo $N$}{
    Centering to the neutral element $E$: \\
    \Indp \uIf{$g$ is left-invariant}{$\bar{P}_i \gets \ltrans _{\ominus M}(P_i)$}
    \Else{$\bar{P}_i \gets \rtrans _{\ominus M}(P_i)$} \Indm

    Scaling the variance: \\
    \hspace{1.5em} $\hat{P}_i \gets \rieexp_{E} \left[ \frac{s}{\sqrt{v^2+\epsilon}} \rielog_{E}(\bar{P}_i) \right]$

    Biasing towards parameter $B$: \\
    \Indp \uIf{$g$ is left-invariant}{$\tilde{P}_i \gets \ltrans _{B}(\hat{P}_i)$}
    \Else{$\tilde{P}_i \gets \rtrans _{B}(\hat{P}_i)$} \Indm
}
\end{algorithm}

\subsection{Manifestations}
\label{sec:manifestations}
This section instantiates our LieBN in \cref{alg:liebn} on nine different Lie groups, including four on the SPD manifold, one on rotation matrices, and four on the correlation manifold.

\subsubsection{LieBN on SPD Manifolds}
\label{subsubsec:spd_param_lie_groups}
We first extend the current Lie groups on SPD manifolds by the matrix power deformation, resulting in three families of parameterized Lie groups. Then, we propose a novel right-invariant metric on the SPD manifold, the first non-trivial right-invariant metric on this manifold. Finally, we construct LieBN layers based on these Lie structures.

\mypara{Deformed Lie structures on SPD manifolds.} As shown in \cref{tab:riem_lie_spd}, there are three Lie groups on SPD manifolds, each with a left-invariant metric. These metrics include $\biparamAIM$, $\biparamLEM$, and LCM. For clarity, we denote the group operations w.r.t. $\biparamAIM$, $\biparamLEM$ and LCM as $\oplusLieAI$, $\oplusLieLE$ and $\oplusLieLC$, respectively.

Recently, \citet{thanwerdas2019exploration} further extended $(\alpha,\beta)$-AIM into a three-parameter family of metrics via the pullback of the matrix power function $\mathrm{P}_\theta(\cdot)$, scaled by $\frac{1}{\theta^2}$ and denoted by $\triparamAIM$.
The matrix power serves as a deformation, wherein $\triparamAIM$ encompasses $(\alpha,\beta)$-AIM with $\theta = 1$, and becomes $(\alpha,\beta)$-LEM as $\theta$ approaches 0 \citep{thanwerdas2019affine}.
Inspired by the deforming utility of the power function, we define the power-deformed metrics of $(\alpha,\beta)$-LEM and LCM as the pullback metrics by $\pow_\theta$ and scaled by $\frac{1}{\theta^2}$.
We denote these two metrics as $\triparamLEM$ and $\paramLCM$, respectively.
We have the following results with respect to the deformation.

\begin{parisproposition} [Deformation]\label{prop:spd_param_lem_lcm_deformation}\linktoproof{spd_param_lem_lcm_deformation}
    $\triparamLEM$ is equal to $\biparamLEM$.
    $\theta$-LCM interpolates between $\tilde{g}$-LEM (as $\theta \to 0$) and LCM ($\theta=1$). Here, given any $P \in \spd{n}$ and tangent vectors $V,W \in T_P\spd{n}$, $\tilde{g}$-LEM is defined as
    \begin{equation}
        \langle V,W \rangle_P = \tilde{g}(\mlog_{*,P}(V),\mlog_{*,P}(W)),
    \end{equation}
    where $\tilde{g}(V_1,V_2)=\frac{1}{2} \langle V_1, V_2 \rangle -\frac{1}{4} \langle \bbD(V_1), \bbD(V_2) \rangle$, $\bbD(V_i)$ is a diagonal matrix consisting of the diagonal elements of $V_i$, and $\mlog_{*,P}$ is the differential map at $P$.
\end{parisproposition}

As $\triparamLEM$ is equal to $\biparamLEM$, we focus on $\biparamLEM$, $\triparamAIM$, and $\paramLCM$ in the following. As a diffeomorphism, $\pow_{\theta}$ can also pull back the group operations $\oplusLieAI$ and $\oplusLieLC$, denoted by $\oplusLiePAI$ and $\oplusLiePLC$, respectively.
We have the following proposition on the invariance.

\begin{parisproposition} [Invariance]\label{prop:spd_param_invariance}\linktoproof{spd_param_invariance}
    $\triparamAIM$ is left-invariant w.r.t. $\oplusLiePAI$, while $\paramLCM$ is bi-invariant w.r.t. $\oplusLiePLC$.
\end{parisproposition}

\mypara{SPD right-invariant metrics.} AIM is left-invariant w.r.t. $\oplusLieAI$. We can also define a right-invariant metric w.r.t. $\oplusLieAI$ by definition \citep[Ch.~1.2]{do1992riemannian}:
\begin{equation}
    \gcri_{P} (V,W) = \left\langle (\rtrans_{\ominusLieAI P})_{*,P} (V) , (\rtrans_{\ominusLieAI P})_{*,P} (W)\right\rangle_{I}
\end{equation}
where $\rtrans_{(\cdot)}$ denotes Lie group right translation, $\ominusLieAI P$ is the inverse of $P$ under $\oplusLieAI$, and $\left\langle \cdot , \cdot \right\rangle_{I}$ denotes an arbitrary inner product on $T_I\spd{n}$. We set $\left\langle \cdot, \cdot \right\rangle_{I}$ to be the same as the AIM at $I$, \ie $\left\langle \cdot, \cdot \right\rangle^{\alphabeta}$. We call this metric CRIM, as the group operation is defined by the matrix product of Cholesky factors \citep[Sec.~3.2]{thanwerdas2022theoretically}.

\begin{paristheorem} \label{thm:crim}\linktoproof{crim}
Given any SPD matrices $P,Q$ and tangent vector $V \in T_P\spd{n}$, the Riemannian operators on $\{\spd{n},\gcri\}$ are
\begin{align}
    \gcri_{P}(V,V)
    &= \left( \left \| \symmetrizeSum{ L(L^{-1} V L^{-\top})_{\frac{1}{2}} L^{-1}} \right \|^{\alphabeta} \right)^2 \\
    \label{eq:dist_crim}
    \dist(P,Q)
    &= \left \| \mlog\left( \widetilde{Q}^{-\frac{1}{2}} \widetilde{P} \widetilde{Q}^{-\frac{1}{2}}\right) \right \| ^{\alphabeta}, \\
    \label{eq:exp_crim}
    \rieexp_{P}(V)
    &= \ominusLieAI\left( \rieexp_{\widetilde{P}}^{\mathrm{AI}} \left(-\bar{V}  \right)\right),\\
    \label{eq:log_crim}
    \rielog_{P}(Q)
    &=-\symmetrizeSum{L L^{\top} \left( L \widetilde{V}L^{\top} \right)_{\frac{1}{2}}^{\top}},
\end{align}
where $L$ is the Cholesky factor of $P=LL^\top$, $\ominusLieAI(\cdot)$ is the group inverse, $\widetilde{P}$ and $\widetilde{Q}$ are the group inverses of $P$ and $Q$, $\bar{V}=\symmetrizeSum{ \left(L^{-1} V L^{-\top}\right)_{\frac{1}{2}} L^{-1} L^{-\top}}$, and $\widetilde{V}=\rielog^{\mathrm{AI}} _{\widetilde{P}} \left(\widetilde{Q} \right)$. Here, $\symmetrizeSum{X}=X + X^\top, \forall X \in \bbR{n \times n}$ denotes unnormalized symmetrization, and $(X)_{\frac{1}{2}}=\lfloor X \rfloor + \frac{1}{2}\bbX$.
\end{paristheorem}

\begin{pariscorollary} \label{cor:crim_geodesic}\linktoproof{crim_geodesic}
    CRIM is geodesically complete, and the associated geodesic connecting SPD matrices $P$ and $Q$ is
    \begin{equation}
    \begin{aligned}
        \gamma _{(P,Q)} (t)
        &= \ominusLieAI\left\{ \gamma^{\mathrm{AI}}(t; \widetilde{P},\widetilde{Q}) \right\}\\
        &= \ominusLieAI\left\{\widetilde{P}^{\frac{1}{2}} \left( \widetilde{P}^{-\frac{1}{2}} \widetilde{Q} \widetilde{P}^{-\frac{1}{2}} \right)^t \widetilde{P}^{\frac{1}{2}} \right\},
     \end{aligned}
    \end{equation}
    where $\widetilde{P}=\ominusLieAI P$ and $\widetilde{Q} = \ominusLieAI Q$ are group inverses, with $\gamma^{\mathrm{AI}}$ as the geodesic under AIM.
\end{pariscorollary}

Similar to the discussion in \cref{subsubsec:spd_param_lie_groups}, we define $\theta$-CRIM as the deformed metric of CRIM by the pullback of matrix power function $\mathrm{P}_\theta(\cdot)$ and scaled by $\frac{1}{\theta^2}$. As the pullback of CRIM, $\theta$-CRIM is right-invariant w.r.t. $\oplusLiePAI$ by definition.
\begin{parisproposition}\label{prop:spd_deformedcrim_invariance}
    $\theta$-CRIM is right-invariant w.r.t. $\oplusLiePAI$.
\end{parisproposition}

\begin{table}[t]
  \centering
  \caption[Key operators in calculating LieBN on SPD manifolds.]{Key operators in calculating LieBN on SPD manifolds.}
  \label{tab:ops_liebn_spd}
  \resizebox{0.99\linewidth}{!}{
    \begin{tabular}{c|c|cccc}
        \toprule
        \multicolumn{2}{c|}{\textbf{Metric}} & $\triparamAIM$  & $\biparamLEM$ & $\theta$-LCM & $\theta$-CRIM \\
        \midrule
        \multicolumn{2}{c|}{\textbf{Invariance}} & Left-invariance  & \multicolumn{2}{c}{Bi-invariance} & Right-invariance \\
        \midrule
        \multicolumn{2}{c|}{\textbf{LieBN Type}} & LieBN-Left  & \multicolumn{2}{c}{LieBN-Left = LieBN-Right} & LieBN-Right \\
        \midrule
        \multicolumn{2}{c|}{\textbf{Pullback Map}} &  $\pow_{\theta}$ &    $\mlog$ & $\pow_{\theta} \circ \clog$ & $\pow_{\theta}$ \\
        \midrule
        \multicolumn{2}{c|}{\textbf{Codomain}} &  $\{\spd{n},\oplusLieAI,\frac{1}{\theta^2}\gbiparamai\}$  & $\{\sym{n}, \langle \cdot , \cdot \rangle^{\alphabeta} \}$ & $\{\trilspace{n}, \frac{1}{\theta^2} \langle \cdot , \cdot \rangle\}$ & $\{\spd{n},\oplusLieAI,\frac{1}{\theta^2}\gcri\}$ \\
        \midrule
        \multirow{5}[15]{2.5cm}{\parbox{2.5cm}{\centering \textbf{Riemannian\\ and Lie\\ group\\ operators\\ in the\\ codomain}}} &  $\ltrans_{Q}(P)$ or $\rtrans_{Q}(P)$  &  $ KPK^\top $  &    $P+Q$  &  $P+Q$ & $LQL^\top$ \\
        \cmidrule(l){2-6}
        &   $\ltrans_{\ominus Q}(P)$ or $\rtrans_{\ominus Q}(P)$   &  $K^{-1}PK^{-\top}$  &  $P-Q$  &  $P-Q$ & $L^{-1}QL^{-\top}$  \\
        \cmidrule(l){2-6}
        &  $ \rieexp_{E} \left [ s \rielog_{E}(P) \right]$   &  $P^{s}$      &    $sP$     &    $sP$ & $\ominusLieAI\left( \left( \ominusLieAI P \right)^s \right)$     \\
        \cmidrule(l){2-6}
        &  FM    &    Karcher Flow     &    \makecell{Arithmetic \\ average}  &    \makecell{Arithmetic \\ average}  & Karcher Flow   \\
        \cmidrule(l){2-6}
        &  $\wfm(\{1-\eta,\eta\},\{P_1,P_2\})$  &  $P_1^{\frac{1}{2}}\left(P_1^{-\frac{1}{2}} P_2 P_1^{-\frac{1}{2}}\right)^\eta P_1^{\frac{1}{2}}$ &  \makecell{Arithmetic \\ weighted average} &  \makecell{Arithmetic \\ weighted average} & $\ominusLieAI\left(\widetilde{P}_1^{\frac{1}{2}}\left(\widetilde{P}_1^{-\frac{1}{2}} \widetilde{P}_2 \widetilde{P}_1^{-\frac{1}{2}}\right)^\eta \widetilde{P}_1^{\frac{1}{2}} \right)$\\
        \bottomrule
    \end{tabular}
    }
\end{table}

\begin{table}[t]
    \centering
    \caption[Key operators in calculating LieBN on the rotation matrices.]{Key operators in calculating LieBN on the rotation matrices.}
    \label{tab:liebn_rotations}
    \resizebox{0.99\linewidth}{!}{
    \begin{tabular}{cccccccc}
    \toprule
    \textbf{Invariance} & \textbf{LieBN Type} & $\ominus R$ & $\ltrans _{R} (S)$ & $\rtrans_{R}(S)$ & $ \rieexp_{I} \left [ s \rielog_{I}(R) \right]$ & \textbf{FM} & $\wfm(\{1-\eta,\eta\},\{R,S\})$  \\
    \midrule
    Bi-invariance & LieBN-Left \& LieBN-Right & $R^{-1}$ & $RS$ &  $SR$ & $\mexp \left( s \mlog \left(R\right) \right)$  & \citep[Alg.~1]{manton2004globally} & $R \mexp (\eta \mlog(R^\top S))$ \\
    \bottomrule
    \end{tabular}
    }
\end{table}

\mypara{Manifestations on SPD manifolds.} So far, there are four families of invariant metrics on the SPD Lie groups: (1) left-invariant $\triparamAIM$ w.r.t. $\oplusLieAI$; (2) bi-invariant $\biparamLEM$ w.r.t. $\oplusLieLE$ and $\theta$-LCM w.r.t. $\oplusLieLC$; (3) right-invariant $\theta$-CRIM w.r.t. $\oplusLieAI$. Since all the above metrics are pullback metrics, the LieBN based on these metrics can be simplified and calculated in the codomain. We first show a general result on LieBN under the pullback metric. We denote \cref{alg:liebn} on the Lie group $\calM$ as
\begin{equation}
    \liebn(P_i;B,s,\epsilon,\eta), \qquad P_i \in \{P_j\}_{j=1}^N \subset \calM.
\end{equation}
Then we can obtain the following theorem.

\begin{paristheorem} \label{thm:liebn_pullback}\linktoproof{liebn_pullback}
    Given a Lie group $\calM_1$, a Lie group $\calM_2$ with an invariant metric $g^2$, and a map $f:\calM_1 \rightarrow \calM_2$ that is both a diffeomorphism and a Lie-group isomorphism, the map $f$ induces an invariant metric $g^1$ on $\calM_1$, denoted as $g^1=f^*g^2$.
    For a batch of activations $\{P_i\}_{i=1}^N$ in $\calM_1$, $\liebn^1(P_i;B,s,\epsilon,\eta)$ in $\calM_1$ can be calculated in $\calM_2$ by the following process:
    \begin{align}
        &\text{Mapping data into } \calM_2: \bar{P_i} =f(P_i), \bar{B}=f(B),\\
        \label{eq:liebn_pm_codomain}
        &\text{Performing LieBN in } \calM_2 : \hat{P_i} = \liebn^2(\bar{P_i};\bar{B},s,\epsilon,\eta),\\
        &\text{Mapping the resulting data back to } \calM_1: \tilde{P_i} =f^{-1}(\hat{P}_i),
    \end{align}
    where $\liebn^2$ is the LieBN on $\calM_2$.
\end{paristheorem}

Given a metric $g$ on $\spd{n}$, the power-deformed metric $\tilde{g}=\frac{1}{\theta^2}\pow_{\theta}^*g$ is equal to $\pow_{\theta}^*(\frac{1}{\theta^2} g)$.
\cref{thm:liebn_pullback} indicates that the LieBN under $\tilde{g}$ can be calculated by the LieBN under $\frac{1}{\theta^2} g$. Besides, as the Christoffel symbols remain the same under constant scaling, the LieBNs under $\frac{1}{\theta^2} g$ and $g$ only differ in the variance. We denote $\gbiparamai$ and $\gtriparamAI$ as the metric tensors of $\biparamAIM$ and $\triparamAIM$, respectively. Based on the above discussions, the computations of the LieBN under $\gtriparamAI$ are reduced to the LieBN under $\frac{1}{\theta^2}\gbiparamai$. Similarly, denoting $\gcri$ and $\gdefcri$ as the metric tensors of CRIM and $\theta$-CRIM, then the LieBN under $\theta$-CRIM can be calculated by the one under $\frac{1}{\theta^2}\gcri$. Furthermore, as shown in \cref{spdmlr:sec:deformed_metrics}, $\biparamLEM$ is a pullback metric from the Euclidean space $\sym{n}$ of symmetric matrices, while $\theta$-LCM is a pullback metric from the Euclidean space $\trilspace{n}$ of lower triangular matrices. As shown in \cref{prop:liebn_natural_extension_ebn}, the LieBN in the Euclidean space $\sym{n}$ or $\trilspace{n}$ is simplified to the standard Euclidean BN. Therefore, the LieBNs under $\biparamLEM$ and $\theta$-LCM can be calculated by the Euclidean BN over $\sym{n}$ and $\trilspace{n}$, respectively.

We denote the LieBN under left and right translations as LieBN-Left and LieBN-Right, respectively. Then, the LieBNs under $\triparamAIM$ and $\theta$-CRIM correspond to LieBN-Left and LieBN-Right, respectively. As $\oplusLieLC$ and $\oplusLieLE$ are commutative, the LieBN-Left and LieBN-Right under $\biparamLEM$ and $\theta$-LCM are equivalent. We denote $P, Q, P_1$ and $P_2$ as points in the codomain, \ie $\spd{n}$ with scaled CRIM for $\theta$-CRIM, $\spd{n}$ with scaled $\biparamAIM$ for $\triparamAIM$, $\sym{n}$ for $\biparamLEM$, and $\trilspace{n}$ for $\theta$-LCM, respectively. For CRIM, we denote $\widetilde{P}_i= \ominusLieAI P_i$ for $i=1,2$. We summarize all the necessary ingredients in \cref{tab:ops_liebn_spd} for calculating SPD LieBN. Note that for $\triparamAIM$, our scaling operation defined in \cref{eq:liebn_scaling} encompasses the scaling operation in \citet[Eq.~(9)]{kobler2022controlling} as a special case, when $(\theta,\alpha,\beta)=(1,1,0)$.

\subsubsection{LieBN on Rotation Matrices}
\label{subsec:liebn_son}

As the Riemannian metric on the rotation matrices is bi-invariant, there are two instantiations of LieBN on this manifold, \ie LieBN-Left based on the left translation and LieBN-Right based on the right translation. In particular, the scaling can be further simplified: $\rieexp_{I} \left(  s \rielog_{I} \left( R \right) \right)
    = \mexp \left( s \mlog \left(R\right) \right)$.
For the specific $\so{3}$, the matrix exp and log can be efficiently calculated without matrix decomposition \citep[Sec.~3.2]{hartley2013rotation}. \cref{tab:liebn_rotations} presents the expressions of the required operators in \cref{alg:liebn}.

\subsubsection{LieBN on Full-Rank Correlation Matrices}
\label{subsec:liebn_cor}

\begin{table}[t]
  \centering
  \caption[Summary of LieBN on the correlation.]{Summary of LieBN on the correlation. $\holinner{\cdot}{\cdot}$ and $\rzeroinner{\cdot}{\cdot}$ are permutation-invariant inner products \citep{thanwerdas2024permutation}.}
  \label{tab:ops_liebn_cor}
  \resizebox{0.99\linewidth}{!}{
    \begin{tabular}{c|cccc}
        \toprule
        \textbf{Metric} & \textbf{ECM}  & \textbf{LECM} & \textbf{OLM} & \textbf{LSM} \\
        \midrule
        \textbf{Invariance} & \multicolumn{4}{c}{Bi-invariance} \\
        \midrule
        \textbf{LieBN Type} & \multicolumn{4}{c}{LieBN-Left = LieBN-Right} \\
        \midrule
        \textbf{Pullback Map} & $ \Theta$ & $\log \circ \Theta$ & $\offlog$ & $\logscaled$ \\
        \midrule
        \textbf{Codomain} & $\{\LTone{n}, \inner{\cdot}{\cdot} \}$ & $\{\LTzero{n}, \inner{\cdot}{\cdot} \}$ & $\{\hol{n}, \holinner{\cdot}{\cdot} \}$ & $\{\rzero{n}, \rzeroinner{\cdot}{\cdot} \}$\\
        \bottomrule
    \end{tabular}
    }
\end{table}

As summarized in \cref{tab:riem_lie_cor}, all four correlation metrics are bi-invariant, and their associated Lie groups are commutative. Consequently, LieBN-Left is identical to LieBN-Right. Moreover, all four correlation metrics are pullback metrics from simpler Euclidean spaces. Therefore, LieBN over the correlation can be implemented according to \cref{thm:liebn_pullback}: (1) map the correlation into the prototype Euclidean space, (2) apply Euclidean BN, and (3) map back to the correlation.

\mypara{Optimization.} Finally, we discuss the optimization of the correlation-valued biasing parameter $B \in \cor{n}$. As reviewed in \cref{sec:ch2-full-rank-correlation-manifolds}, the correlation matrix can be identified by the product of hyperbolic spaces via the Cholesky decomposition. Given $C \in \cor{n}$, the $k$-th row of the Cholesky factor $L=\chol(C)$ is $\left(L_{k 1}, \ldots, L_{k, k-1}, L_{k k}, 0, \ldots, 0\right)$ with $L_{k k}>0$, which belongs to the hyperbolic space of an open hemisphere
\begin{equation}
    \hs{k-1}=\left\{x \in \bbR{k} \mid \norm{x}=1, x_k>0 \right\}.
\end{equation}
Besides, the open hemisphere $\hs{n}$ is isometric to the Poincaré ball
\begin{equation}
    \unitpball{n}=\left\{x \in \bbR{n} \mid \norm{x} < 1 \right\}
\end{equation}
by
\begin{equation}
    \pi _{\hs{n} \rightarrow \unitpball{n}} ((x^\top, x_{n+1})^\top) = \frac{x}{1+x_{n+1}}.
\end{equation}
Therefore, each correlation can be parameterized with $n-1$ Poincaré vectors. Each Poincaré vector can be optimized directly on its manifold using the Riemannian optimization strategy reviewed in \cref{sec:ch2-riemannian-optimization}. The above process can be expressed as
\begin{equation}
    C {\mapsto}
    \begin{pmatrix}
    1 & 0 & \cdots & 0 \\
    L_{21} & L_{22} & \cdots & 0 \\
    \vdots & \vdots & \ddots & \vdots \\
    L_{n1} & L_{n2} & \cdots & L_{nn}
    \end{pmatrix}
    {\mapsto}
    \begin{pmatrix}
    x_1 \in \unitpball{1} \\
    \vdots \\
    x_{n-1} \in \unitpball{n-1}
    \end{pmatrix}.
\end{equation}

\subsection{Experiments}
\label{sec:experiments}
This section validates our LieBN on nine invariant metrics across the SPD, rotation, and correlation matrices.
More details on data sets and experimental settings are provided in \cref{app:datasets,app:liebn-experimental-details}.

\subsubsection{Experiments of LieBN on the SPD Manifold}
\label{subsec:exp_spd}
Note that our LieBN layers are architecture-agnostic and can be applied to any existing SPD neural network. Following the previous work \citep{huang2017riemannian,brooks2019riemannian,kobler2022spd}, we focus on two network architectures:
(1) SPDNet \citep{huang2017riemannian} for drone recognition on the Radar data set \citep{brooks2019riemannian}, and human action recognition on the HDM05 \citep{muller2007documentation} and FPHA \citep{garcia2018first} data sets;
(2) TSMNet \citep{kobler2022spd} for EEG classification on the Hinss2021 data set \citep{hinss2021eegdata}.
In the EEG application, TSMNet is endowed with \emph{SPD Domain-Specific Momentum Batch Normalization (SPDDSMBN)} (denoted TSMNet+SPDDSMBN) \citep{kobler2022spd}, which is a domain adaptation version of \citet{kobler2022controlling}. For a fair comparison, we also implement a domain-specific momentum LieBN, referred to as DSMLieBN. The backbone network architectures are represented as $\{d_0, d_1, \ldots, d_L\}$, where the dimension of the parameter in the $i$-th BiMap layer is $d_i \times d_{i-1}$. As $(\alpha,\beta)$ only affects variance calculation throughout LieBN, we simply set $(\alpha,\beta)=(1,0)$ and only tune the deformation factor $\theta$. For each family of LieBN or DSMLieBN, we report two representatives: the standard one induced from the standard metric ($\theta=1$), and the one induced from the deformed metric with proper $\theta$. \textit{If the standard one is already saturated, we only report the result of that standard variant.}

\begin{table}[t]
    \caption[10-fold average results of SPDNet with and without SPDBN or LieBN.]{10-fold average results of SPDNet with and without SPDBN or LieBN on the Radar, HDM05, and FPHA data sets. If the LieBN under the standard metric ($\theta=1$) is not saturated, the rightmost columns report the deformed LieBN. The best results are \firstresults{bold}.}
    \label{tab:results_spdnet}
    \begin{subtable}[h]{\textwidth}
        \centering
        \caption{Radar data set.}
        \label{tab:results_radar}
        \resizebox{\linewidth}{!}{%
        \begin{tabular}{c|cc|*{4}{>{\columncolor{HilightColor}}c}|>{\columncolor{HilightColor}}c}
            \toprule
            \multirow{3}[6]{*}{\textbf{Acc}} & \multirow{3}[6]{*}{\textbf{SPDNet}} & \multirow{3}[6]{*}{\textbf{SPDNetBN}} & \multicolumn{5}{>{\columncolor{HilightColor}}c}{\textbf{SPDNetLieBN}} \\
            \cmidrule{4-8}          &       &       & \multicolumn{4}{>{\columncolor{HilightColor}}c|}{$\theta=1$} & \multicolumn{1}{>{\columncolor{HilightColor}}c}{\textbf{Best $\theta$}} \\
            \cmidrule{4-8}          &       &       & \textbf{AIM-(1)} & \textbf{LEM-(1)} & \textbf{LCM-(1)} & \textbf{CRIM-(1)} & \textbf{LCM-(-0.5)} \\
            \midrule
            \multicolumn{1}{c|}{Fit time (s)} & 0.98  & 1.56  & 1.62  & 1.28  & 1.11  & 1.82  & 1.43 \\
            Mean±STD & 93.25±1.10 & 94.85±0.99 & \firstresults{95.47±0.90} & 94.89±1.04 & 93.52±1.07 & 94.35±0.68 & 94.80±0.71 \\
            Max   & 94.4  & 96.13 & 96.27 & \firstresults{96.8} & 95.2  & 95.6  & 95.73 \\
            \bottomrule
        \end{tabular}%
        }
    \end{subtable}

    \begin{subtable}[h]{\linewidth}
        \centering
        \caption{HDM05 data set.}
        \label{tab:results_HDM05}
        \resizebox{\linewidth}{!}{%
        \begin{tabular}{c|c|c|*{4}{>{\columncolor{HilightColor}}c}|*{3}{>{\columncolor{HilightColor}}c}}
            \toprule
            \multicolumn{1}{c|}{\multirow{3}[6]{*}{\textbf{Acc}}} & \multirow{3}[6]{*}{\textbf{SPDNet}} & \multirow{3}[6]{*}{\textbf{SPDNetBN}} & \multicolumn{7}{>{\columncolor{HilightColor}}c}{\textbf{SPDNetLieBN}} \\
            \cmidrule{4-10}    \multicolumn{1}{c|}{} &       &       & \multicolumn{4}{>{\columncolor{HilightColor}}c|}{$\theta=1$} & \multicolumn{3}{>{\columncolor{HilightColor}}c}{\textbf{Best $\theta$}} \\
            \cmidrule{4-10}    \multicolumn{1}{c|}{} &       &       & \textbf{AIM-(1)} & \textbf{LEM-(1)} & \textbf{LCM-(1)} & \textbf{CRIM-(1)} & \textbf{AIM-(1.5)} & \textbf{LCM-(0.5)} & \textbf{CRIM-(0.5)} \\
            \midrule
            Fit time (s) & 0.57  & 0.97  & 1.14  & 0.87  & 0.66  & 1.37  & 1.46  & 1.01  & 1.74 \\
            Mean±STD & 59.13±0.67 & 66.72±0.52 & 67.79±0.65 & 65.05±0.63 &  66.68±0.71 & 63.25±0.88 & 68.16±0.68 & \firstresults{70.84±0.92} & 65.76±0.54 \\
            Max   & 60.34 & 67.66 & 68.75 & 66.05 & 68.52 & 64.94 & 69.25 & \firstresults{72.27} & 66.96 \\
            \bottomrule
        \end{tabular}%
        }
     \end{subtable}

    \begin{subtable}[h]{\linewidth}
        \centering
        \caption{FPHA data set.}
        \label{tab:results_FPHA}
        \resizebox{\linewidth}{!}{%
        \begin{tabular}{c|c|c|*{4}{>{\columncolor{HilightColor}}c}|*{3}{>{\columncolor{HilightColor}}c}}
            \toprule
            \multirow{3}[6]{*}{\textbf{Acc}} & \multirow{3}[6]{*}{\textbf{SPDNet}} & \multirow{3}[6]{*}{\textbf{SPDNetBN}} & \multicolumn{7}{>{\columncolor{HilightColor}}c}{\textbf{SPDNetLieBN}} \\
            \cmidrule{4-10}          &       &       & \multicolumn{4}{>{\columncolor{HilightColor}}c|}{$\theta=1$} & \multicolumn{3}{>{\columncolor{HilightColor}}c}{\textbf{Best $\theta$}} \\
            \cmidrule{4-10}          &       &       & \textbf{AIM-(1)} & \textbf{LEM-(1)} & \textbf{LCM-(1)} & \textbf{CRIM-(1)} & \textbf{AIM-(1.5)} & \textbf{LCM-(0.5)} & \textbf{CRIM-(-0.5)} \\
            \midrule
            Fit time (s) & 0.32  & 0.62  & 0.8   & 0.55  & 0.39  & 0.92  & 1.03  & 0.65  & 1.21 \\
            Mean±STD & 85.59±0.72 & 89.33±0.49 & 89.70±0.51 & 86.56±0.79 & 77.64±1.00 & 84.65±1.20 & \firstresults{90.39±0.66} & 86.33±0.43 & 86.40±0.57 \\
            Max   & 86    & 90.17 & 90.5  & 87.83 & 79    & 86.67 & \firstresults{92.17} & 87    & 87.17 \\
            \bottomrule
        \end{tabular}%
        }
     \end{subtable}
\end{table}

\mypara{Application to SPDNet.} As SPDNet is a classical SPD network, we apply our LieBN to SPDNet on the Radar, HDM05, and FPHA data sets. Additionally, we compare our method with SPDNetBN, which applies the SPDBN in \cref{eq:spdnetbn_centering,eq:spdnetbn_biasing} to SPDNet. Following \citet{brooks2019riemannian}, we use the architectures of $\{20,16,8\}$, $\{93,30\}$, and $\{63,33\}$ for the Radar, HDM05, and FPHA data sets, respectively. The 10-fold average results, including the average training time (s/epoch), are summarized in \cref{tab:results_spdnet}. We have three key observations regarding the choice of metrics, deformation, and training efficiency.
\begin{itemize}
    \item
    \mypara{The choice of metrics.} The metric that yields the most effective LieBN layer differs for each data set. Specifically, the optimal LieBN layers on these three data sets are the ones induced by AIM-(1), LCM-(0.5), and AIM-(1.5), respectively, \textbf{which improve the performance of SPDNet by 2.22, 11.71, and 4.80 percentage points}. Additionally, although the LCM-based LieBN performs worse than other LieBN variants on the Radar and FPHA data sets, it exhibits the best performance on the HDM05 data set. These observations demonstrate the value of a unified normalization framework that can be instantiated under multiple admissible SPD geometries.
    \item
    \mypara{The effect of deformation.} Deformation patterns also vary across data sets. Firstly, the standard AIM and CRIM are already saturated on the Radar data set. Secondly, the appropriate deformation $\theta$ can further enhance the performance of LieBN. Notably, even though the LieBNs induced by LCM-(1) and CRIM-(1) impede the learning of SPDNet on the FPHA data set, they can improve the performance under an appropriate deformation $\theta$. These findings highlight the efficacy of the deforming geometry on the SPD manifold.
    \item
    \mypara{Efficiency.} Although our LieBN involves additional computations on variance compared with SPDNetBN, our LieBN achieves comparable or even better efficiency than SPDNetBN. In particular, the LieBN induced by standard LEM or LCM exhibits better efficiency than SPDNetBN. Even with deformation, the LCM-based LieBN is still comparable with SPDNetBN in terms of efficiency. This phenomenon could be attributed to the fast and simple computation of LCM and LEM.
\end{itemize}

\begin{table}[t]
    \caption[Cross-validation results of TSMNet with SPDDSMBN and DSMLieBN.]{Cross-validation results of TSMNet with SPDDSMBN and DSMLieBN on the Hinss data set. If the DSMLieBN under the standard metric ($\theta=1$) is not saturated, the bottom rows report deformed DSMLieBN.}
    \label{tab:results_TSMNet}
    \begin{subtable}[t]{0.48\linewidth}
        \centering
        \caption{Inter-session classification}
        \label{tab:results_inter_session}
        \resizebox{\linewidth}{!}{
        \begin{tabular}{c|c|cc}
            \toprule
            \multicolumn{2}{c|}{\textbf{Method}} & \textbf{Fit Time} & \textbf{Mean±STD} \\
            \midrule
            \multicolumn{2}{c|}{SPDDSMBN} & 0.16  & 54.12±9.87 \\
            \midrule
            \multirow{5}[4]{*}{DSMLieBN} & \cellcolor{HilightColor} AIM-(1) & \cellcolor{HilightColor} 0.16  & \cellcolor{HilightColor} \firstresults{55.10±7.61} \\
            & \cellcolor{HilightColor} LEM-(1) & \cellcolor{HilightColor} 0.13  & \cellcolor{HilightColor} 54.95±10.09 \\
            & \cellcolor{HilightColor} LCM-(1) & \cellcolor{HilightColor} 0.10  & \cellcolor{HilightColor} 51.54±6.88 \\
            & \cellcolor{HilightColor} CRIM-(1) & \cellcolor{HilightColor} 0.29  & \cellcolor{HilightColor} 51.86±9.21 \\
            \cmidrule{2-4}          & \cellcolor{HilightColor} LCM-(0.5) & \cellcolor{HilightColor} 0.15  & \cellcolor{HilightColor} 53.11±5.65 \\
            \bottomrule
        \end{tabular}%
        }
    \end{subtable}
    \hfill
    \begin{subtable}[t]{0.45\linewidth}
        \centering
        \caption{Inter-subject classification}
        \label{tab:results_inter_subject}
        \resizebox{\linewidth}{!}{
        \begin{tabular}{c|c|cc}
            \toprule
            \multicolumn{2}{c|}{\textbf{Method}} & \textbf{Fit Time} & \textbf{Mean±STD} \\
            \midrule
            \multicolumn{2}{c|}{SPDDSMBN} & 7.74  & 50.10±8.08 \\
            \midrule
            \multirow{6}[4]{*}{DSMLieBN} & \cellcolor{HilightColor} AIM-(1) & \cellcolor{HilightColor} 6.94  & \cellcolor{HilightColor} 50.04±8.01 \\
            & \cellcolor{HilightColor} LEM-(1) & \cellcolor{HilightColor} 4.71  & \cellcolor{HilightColor} 50.95±6.40 \\
            & \cellcolor{HilightColor} LCM-(1) & \cellcolor{HilightColor} 3.59  & \cellcolor{HilightColor} 51.86±4.53 \\
            & \cellcolor{HilightColor} CRIM-(1) & \cellcolor{HilightColor} 16.35 & \cellcolor{HilightColor} 50.71±8.1 \\
            \cmidrule{2-4}          & \cellcolor{HilightColor} CRIM-(1.5) & \cellcolor{HilightColor} 19.51 &  \cellcolor{HilightColor} 51.34±5.82 \\
            & \cellcolor{HilightColor} AIM-(-0.5) & \cellcolor{HilightColor} 8.71  & \cellcolor{HilightColor} \firstresults{53.97±8.78} \\
            \bottomrule
        \end{tabular}%
        }
     \end{subtable}
\end{table}

\mypara{Application to EEG classification.} We apply our method to TSMNet under two scenarios, inter-session and inter-subject. Following \citet{kobler2022spd}, we adopt the architecture of $\{40,20\}$. Compared to SPDDSMBN, DSMLieBN-AIM obtains the highest average scores of 55.10\% and 53.97\% in these two scenarios, \textbf{outperforming SPDDSMBN by 0.98 and 3.87 percentage points, respectively}. In the inter-subject scenario, the efficiency advantage of our LieBN over SPDDSMBN is more evident.
Specifically, both the LEM- and LCM-based DSMLieBN achieve similar or better performance compared to SPDDSMBN, while requiring considerably less training time.
For example, DSMLieBN-LCM-(1) achieves better results with only half the training time of SPDDSMBN on inter-subject tasks. Interestingly, under the standard AIM, the sole difference between SPDDSMBN and our DSMLieBN is the way of centering and biasing. SPDDSMBN applies the matrix inverse square root and matrix square root to fulfill centering and biasing, while AIM-induced LieBN uses more efficient Cholesky decomposition. As such, the DSMLieBN induced by the standard AIM is more efficient than SPDDSMBN, particularly on the inter-subject task. On the other hand, the CRIM-based LieBN shows less efficiency, due to the relatively complex Riemannian computation of this metric.

\begin{figure}[t]
\centering
\includegraphics[width=\linewidth,trim={0cm 0cm 0cm 0cm}]{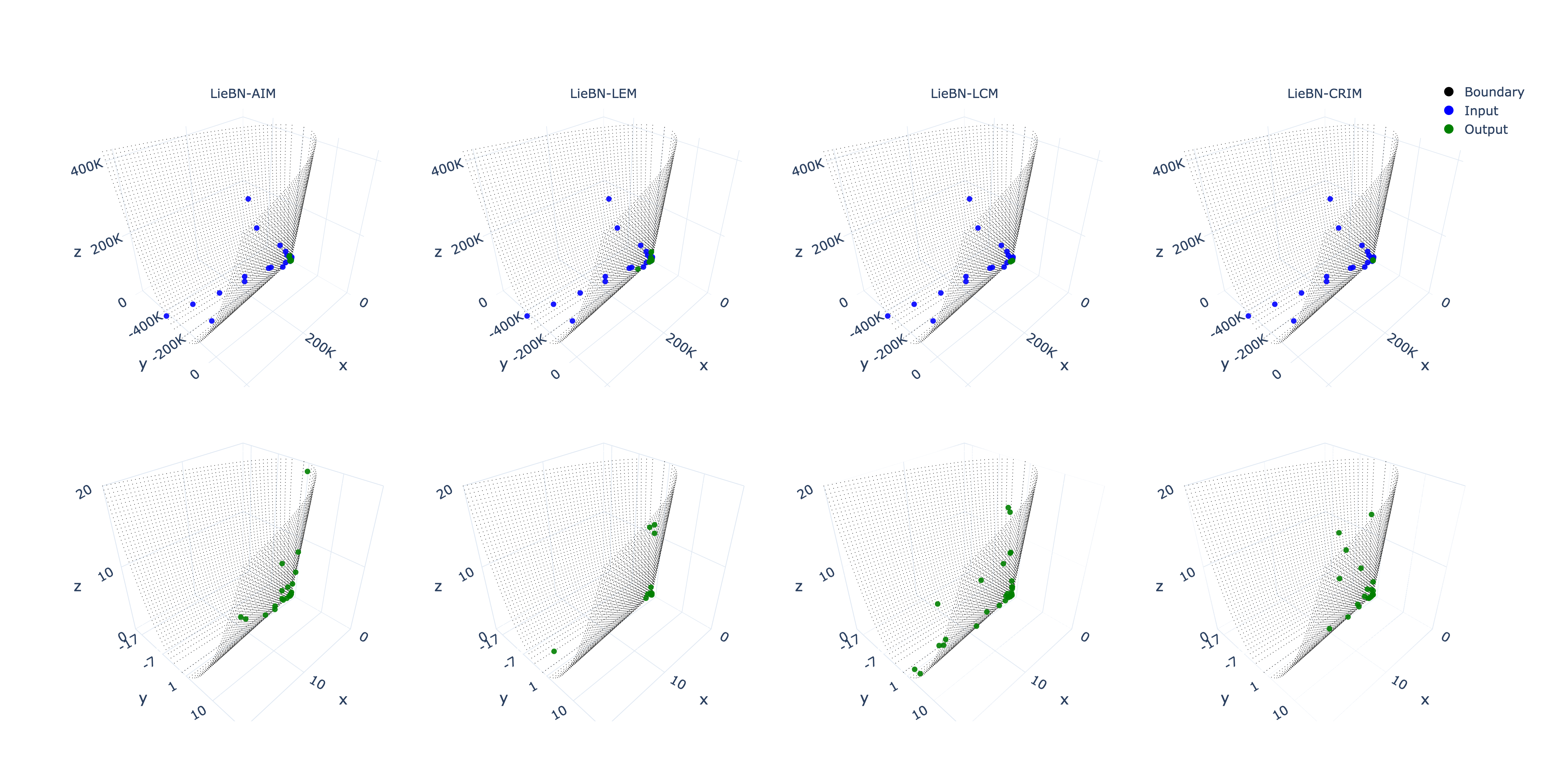}
\caption[Visualization of input and output SPD matrices in LieBN.]{Visualization of input and output $30 \times 30$ SPD matrices in LieBN using $2 \times 2$ Riemannian t-SNE embeddings. The first row shows the input and output under different metrics. Due to the significant difference in magnitude between the t-SNE embeddings of LieBN’s input and output, the second row separately visualizes the LieBN output (\textbf{at a smaller scale}).}
\label{fig:visualization_hdm05}
\end{figure}

\mypara{Visualization.} We randomly select 50 samples and visualize the input and output of LieBN on the HDM05 data set. Using Riemannian t-SNE \citep{surrel2025geometryaware}, we map the $30 \times 30$ SPD matrices to $2 \times 2$ low-dimensional representations. As shown in \cref{fig:visualization_hdm05}, LieBN effectively normalizes the data distribution. Specifically, the input t-SNE embeddings are largely scattered and their elements can reach up to 400K, whereas those of the output embeddings are mostly constrained within 20.

\subsubsection{Experiments of LieBN on Rotation Matrices}
This subsection implements our LieBN on the special orthogonal groups, \ie $\so{n}$, also known as rotation matrices. As the Riemannian metric on $\so{n}$ is bi-invariant, there are two instantiations of our LieBN on this group: LieBN-Left based on the left translation and LieBN-Right based on the right translation. We apply our LieBN to the classic LieNet backbone \citep{huang2017deep}, where the latent space is the special orthogonal group. Following \citet{huang2017deep}, we use three action recognition data sets, the G3D \citep{bloom2012g3d}, HDM05 \citep{muller2007documentation}, and NTU60 \citep{shahroudy2016ntu} data sets. We denote the LieNet models with our LieBN-Left and LieBN-Right as LieNetLieBN-Left and LieNetLieBN-Right, respectively.

\begin{figure}[t]
\centering
\includegraphics[width=0.99\linewidth,trim={0cm 0cm 0cm 0cm}]{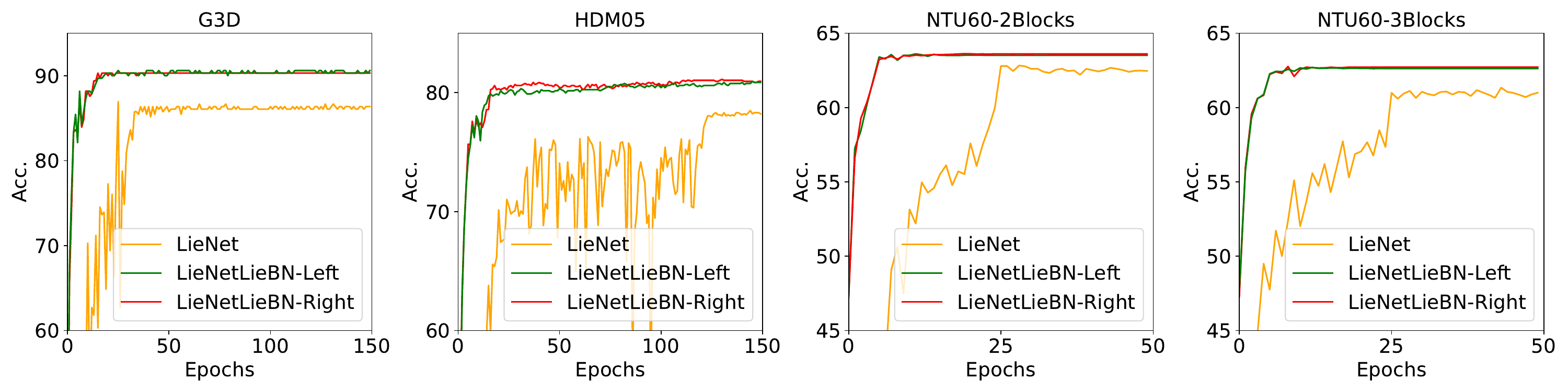}
\caption[Test accuracy curves of LieNet with rotation LieBN.]{Test accuracy curves corresponding to \cref{tab:results_liebn_so3}.}
\label{fig:acc_liebn_so3}
\end{figure}

\begin{table}[t]
    \centering
    \caption{Results of LieNet with or without rotation LieBN.}
    \label{tab:results_liebn_so3}%
    \resizebox{0.99\linewidth}{!}{
    \begin{tabular}{c|cc|cc|cc}
        \toprule
        \multirow{2}[4]{*}{\textbf{Method}} & \multicolumn{2}{c|}{\textbf{G3D}} & \multicolumn{2}{c|}{\textbf{HDM05}} & \multicolumn{2}{c}{\textbf{NTU60}} \\
        \cmidrule{2-7}          & \textbf{Mean±STD} & \textbf{Max}   & \textbf{Mean±STD} & \textbf{Max}   & \textbf{2Blocks} & \textbf{3Blocks} \\
        \midrule
        LieNet & 87.91±0.90 & 89.73 & 76.92±1.27 & 79.11 & 62.4  & 60.91 \\
        \midrule
        \rowcolor{HilightColor} LieNetLieBN-Left &  \firstresults{88.88±1.62} &  \firstresults{90.67} &  78.89±1.07 &  \firstresults{80.88} & 63.51 & 62.62 \\
        \rowcolor{HilightColor} LieNetLieBN-Right & 88.12±1.12 & 90.3  & \firstresults{79.39±1.13} & 80.67 & \firstresults{63.6} & \firstresults{62.72} \\
        \bottomrule
    \end{tabular}
    }
\end{table}%

\mypara{Results.} We conduct 10-fold experiments on the G3D and HDM05 data sets under the suggested
3Blocks\footnote{Each block consists of a RotMap layer followed by a RotPooling layer. For more details, please refer to \citet{huang2017deep}.} and 2Blocks architectures, respectively. On the NTU60 data set, we validate LieBN under the 2Blocks and 3Blocks settings. The results are presented in \cref{tab:results_liebn_so3}. Due to differences in software, our reimplemented LieNet (in PyTorch) performs slightly differently from the results reported by \citet{huang2017deep} (in MATLAB). However, we still observe a clear improvement when applying our LieBN to the vanilla LieNet backbone. Additionally, LieBN-Right performs slightly better than LieBN-Left. Although the effects of left and right translations on the sample statistics under the bi-invariant metric are identical, their transformations on each sample differ, as illustrated in \cref{fig:illustration}. This difference could slightly affect the network performance. The specific optimal choice of left or right translations depends on the data set's characteristics.

\mypara{Training dynamics.} \cref{fig:acc_liebn_so3} presents the test accuracy curves. We have the following additional observations, which can be attributed to the mitigated covariate shift by our LieBN, as our LieBN can effectively normalize the sample statistics.
\begin{itemize}
    \item \mypara{Accelerated convergence.} LieBN significantly accelerates the convergence of LieNet. Specifically, on the NTU60 data set, the largest data set involved, LieNet with LieBN converges by the 5th epoch, whereas the vanilla LieNet does not converge until the 25th epoch. A similar phenomenon can also be observed on the HDM05 data set.
    \item \mypara{More stable performance.} LieBN enhances the stability of network training. In particular, on the HDM05 and G3D data sets, the initial training fluctuations are greatly mitigated by our LieBN.
\end{itemize}

\subsubsection{Experiments of LieBN on Correlation Matrices}
\label{subsec:exp_liebn_cor}

We apply our correlation LieBN (LieBN-Cor) to SPD networks. Our experiments focus on the SPDNet backbone using the FPHA and HDM05 data sets. LieBN-Cor is applied before the final classification layer. Specifically, SPD features are first activated by the power function, then mapped into correlation matrices via $\coropt(\cdot)$, and finally processed by LieBN-Cor.

\begin{table}[t]
  \centering
  \caption{Results of SPDNet with or without correlation LieBN under different invariant metrics.}
  \label{tab:results_liebn_cor}%
  \resizebox{0.99\linewidth}{!}{
    \begin{tabular}{c|c|*{4}{>{\columncolor{HilightColor}}c}}
    \toprule
    \multirow{2}[4]{*}{\textbf{Data Set}} & \multirow{2}[4]{*}{\textbf{SPDNet}} & \multicolumn{4}{>{\columncolor{HilightColor}}c}{\textbf{SPDNetLieBN-Cor}} \\
\cmidrule{3-6}          &       & \textbf{ECM}   & \textbf{LECM}  & \textbf{OLM}   & \textbf{LSM} \\
    \midrule
    HDM05 & 59.13±0.67 & \firstresults{65.37 ± 1.07} & 61.35 ± 0.34 & 60.33 ± 0.12 & 60.00 ± 0.27 \\
    FPHA  & 85.59±0.72 & \firstresults{87.20 ± 0.12} & 87.03 ± 0.32 & 86.80 ± 0.12 & 86.77 ± 0.29 \\
    \bottomrule
    \end{tabular}%
    }
\end{table}%

\mypara{Results.} The 5-fold average results are presented in \cref{tab:results_liebn_cor}. Although LieBN-Cor is not specifically designed for SPD networks, it still improves SPDNet's performance, demonstrating its effectiveness. Among the four invariant metrics, ECM achieves the best performance. LieBN-SPD outperforms LieBN-Cor when applied to SPDNet, as expected because SPDNet is tailored to SPD matrices. However, this does not undermine the validity of LieBN-Cor. The consistent improvement over vanilla SPDNet highlights the potential of applying LieBN-Cor to correlation manifolds.

\section{Gyrogroup Batch Normalization}
\label{sec:ch3-gyrogroup-approach}

\subsection{Introduction}
\label{sec:ch3-gyrobn-introduction}
Although LieBN can normalize sample statistics, many important geometries in machine learning do not admit a Lie group structure. As a result, existing methods still lack a principled solution for Riemannian normalization. Recently, gyro-structures have emerged as effective tools for building Riemannian networks across various geometries, including SPD~\citep{nguyen2022gyro}, Grassmannian~\citep{nguyen2022gyro}, hyperbolic~\citep{ganea2018hyperbolic}, and spherical manifolds~\citep{skopek2020mixed}. They naturally extend Euclidean vector structures while encompassing Lie groups and non-group geometries. For instance, the Grassmannian, hyperbolic, and spherical manifolds do not form Lie groups but instead form gyrogroups.

\begin{figure}[t]
\centering
\includegraphics[width=0.9\linewidth,trim={0cm 0cm 0cm 0cm}]{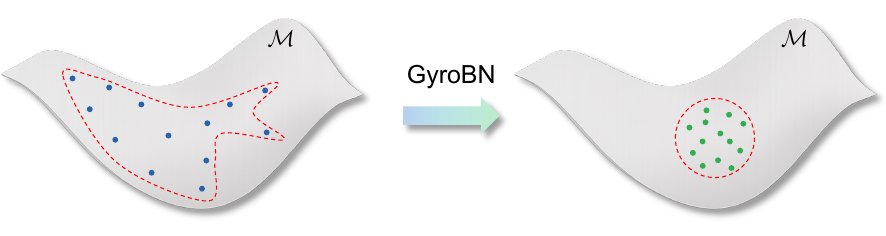}
\caption{Illustration of GyroBN on manifold-valued data. Blue points, green points, and the red dashed curves indicate the input samples, normalized outputs, and data distributions, respectively.}
\label{fig:illustration_gyrobn}
\end{figure}

\begin{table}[t]
    \centering
    \resizebox{0.99\linewidth}{!}{
    \begin{tabular}{cccc}
        \toprule
         \textbf{Method} & \makecell{\textbf{Controllable}\\ \textbf{Statistics}} & \textbf{Applied Geometries} & \textbf{Incorporated by GyroBN}\\
         \midrule
         SPDBN \citep{brooks2019riemannian} &  M & SPD manifolds under AIM & \cmark\\
         SPDBN \citep{kobler2022controlling} & M+V & SPD manifolds under AIM & \cmark\\
         SPDDSMBN \citep{kobler2022spd} & M+V & SPD manifolds under AIM & \cmark\\
         ManifoldNorm \citep[Algs.~1--2]{chakraborty2020manifoldnorm}  & \redbf{N/A} & Riemannian homogeneous spaces & \xmark\\
         ManifoldNorm \citep[Algs.~3--4]{chakraborty2020manifoldnorm} & M+V & \makecell{Matrix Lie groups under the distance \\
         $d(X, Y)=\left\|\mlog \left(X^{-1} Y\right)\right\|$}  & \cmark\\
         RBN \citep[Alg.~2]{lou2020differentiating} & \redbf{N/A} & Geodesically complete manifolds & \xmark\\
         LieBN (\cref{sec:ch3-lie-group-approach}) & M+V & \makecell{Lie groups under\\ invariant metrics} & \cmark\\
         \midrule
         \rowcolor{HilightColor} GyroBN & M+V & \shortstack{Pseudo-reductive gyrogroups \\ with gyroisometric gyrations} & \na \\
         \bottomrule
    \end{tabular}
    }
     \caption{Comparison of previous RBN methods with GyroBN, where M and V denote the sample mean and variance.}
    \label{tab:rbn_summary}
\end{table}

Based on the analysis above, this part of the dissertation first introduces the \emph{pseudo-reductive gyrogroup}, a relaxation of the classical gyrogroup that provides a broader algebraic foundation for Riemannian normalization. Building on this structure, we develop GyroBN, a general RBN framework on pseudo-reductive gyrogroups, as illustrated in \cref{fig:illustration_gyrobn}. We employ gyrosubtraction, gyroaddition, and scalar gyromultiplication to generalize the centering (vector subtraction), biasing (vector addition), and scaling (scalar multiplication) in Euclidean BN to curved manifolds in a principled manner. We clarify why centering and biasing in GyroBN rely on \emph{left} gyroaddition, rather than other candidates, such as right gyroaddition or gyrocoaddition~\citep[Def.~2.9]{ungar2022analytic}. We show that when gyrations are \emph{gyroisometries}, GyroBN enjoys theoretical control over sample statistics. These conditions are satisfied by all known gyrogroups in machine learning, providing a principled and unified normalization mechanism. Moreover, several existing RBN methods arise as special cases of GyroBN, including LieBN and various SPD-based variants, as summarized in \cref{tab:rbn_summary}.

Beyond the LieBN instantiations in \cref{sec:manifestations}, we instantiate GyroBN on seven representative geometries: the Grassmannian~\citep{bendokat2024grassmann}, five CCSs~\citep{ganea2018hyperbolic,lee2018introduction,bachmann2020constant}, and the full-rank correlation manifold~\citep{thanwerdas2022theoretically}. For the Grassmannian, we propose an efficient implementation. For CCSs, we cover five models: Poincaré ball, Lorentz, Beltrami--Klein, sphere, and projected hypersphere. To enable these instantiations, we refine the projected hypersphere structure~\citep{bachmann2020constant}, derive closed-form gyro-structures for Lorentz and spherical geometries, and develop the Riemannian structure of the Beltrami--Klein model. For the correlation manifold, we demonstrate that its gyro-structure can be defined row-wise on its Cholesky factor. We also provide a PyTorch-compatible toolbox~\citep{paszke2019pytorch} with drop-in GyroBN layers, illustrated in \cref{fig:gyrobn_minimal_examples}. Experiments on networks over these seven geometries validate the effectiveness of our framework.

\begin{figure}[t]
\begin{lstlisting}[language=Python]
from GyroBN import *
from GyroBN.Geometry import *

# ==== Grassmannian ====
manifold = GrassmannianGyro(n=50, p=10)
X_gr = manifold.random_normal(30, 50, 10)
gybn_gr = GyroBNGr(shape=[50, 10])
out_gr = gybn_gr(X_gr)

# ==== Five CCSs ====
models = [
    ("Poincare",    Stereographic(K=-1.0)),
    ("Hyperboloid", Hyperboloid(K=-1.0)),
    ("Klein",       Klein(K=-1.0)),
    ("Sphere",      Sphere(K= 1.0)),
    ("ProjSphere",  Stereographic(K= 1.0)),
]
for name, manifold in models:
    X_ccs = manifold.random_normal(30, 16)
    gybn_ccs = GyroBNCCS(shape=[16], model=name, K=manifold.K)
    out_ccs = gybn_ccs(X_ccs)

# ==== Full-Rank Correlation ====
manifold = CorPolyHyperbolicCholeskyMetric(n=10)
X_cor = manifold.random(30, 10, 10)
gybn_cor = GyroBNCor(shape=[10, 10])
out_cor = gybn_cor(X_cor)
\end{lstlisting}
\caption{Minimal examples of applying GyroBN.}
\label{fig:gyrobn_minimal_examples}
\end{figure}

In summary, the main contributions of this part are
\begin{itemize}
    \item Theoretical foundation: pseudo-reductive gyrogroups as a relaxation of classical gyrogroups;
    \item General framework: GyroBN as a plug-and-play normalization mechanism, with pseudo-reduction and gyroisometric gyrations ensuring theoretical control of batch statistics;
    \item Geometric insights: refined projected hypersphere gyro-structure, closed-form gyro-structures for Lorentz and spherical geometries, Riemannian structure of the Beltrami--Klein model, and row-wise correlation manifold gyro-structure;
    \item Practical instantiations: implementations on the Grassmannian, five CCSs, and the correlation manifold with extensive experiments.\footnote{The code is available at \url{https://github.com/GitZH-Chen/GyroBN.git}.}
\end{itemize}

\mypara{Outline.} \cref{sec:ch3-pseudo-reductive-gyrogroups} introduces pseudo-reductive gyrogroups and analyzes their theoretical properties, while \cref{sec:ch3-gyrobn-general} develops the GyroBN framework. \cref{sec:ch3-gyrobn-instantiations} shows that prior RBN methods are special cases and instantiates GyroBN on seven representative geometries. \cref{sec:ch3-gyrobn-experiments} reports experiments that validate GyroBN across these geometries. Proofs are deferred to \cref{app:gyrobn-proofs}.

\subsection{Pseudo-Reductive Gyrogroups}
\label{sec:ch3-pseudo-reductive-gyrogroups}

\begin{figure}[t]
\centering
\includegraphics[width=0.8\linewidth,trim={0cm 0cm 0cm 0cm}]{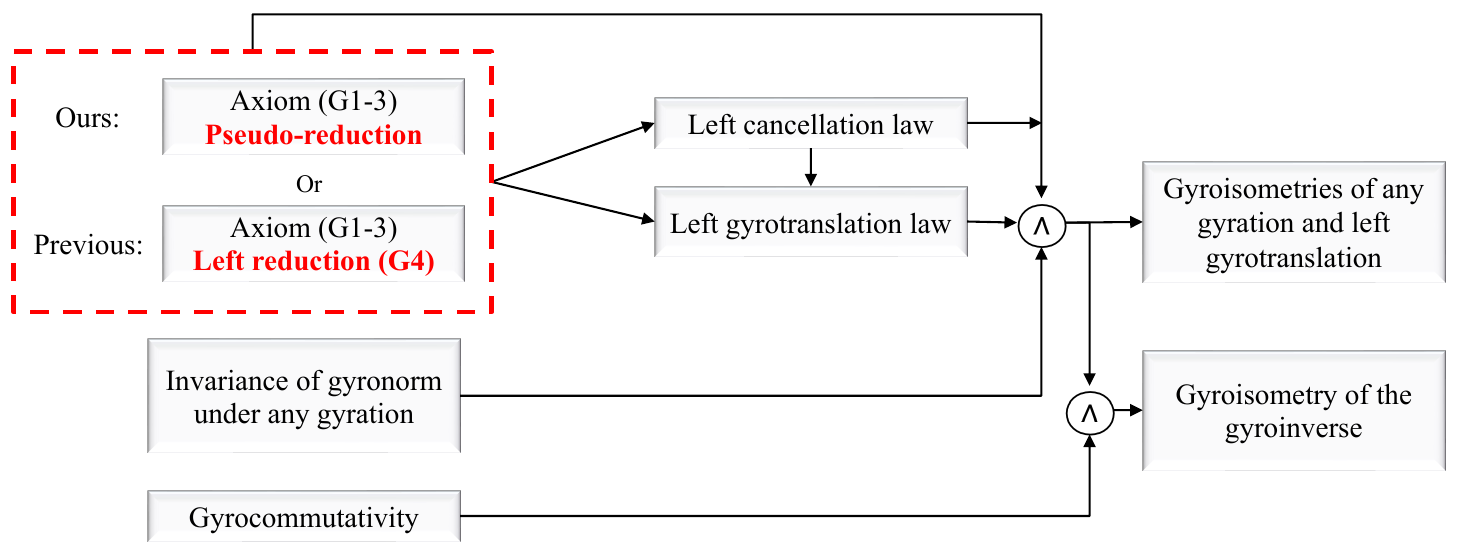}
\caption[Comparison of derivation logic for gyrotranslation isometries.]{A conceptual comparison of the derivation logic in our work with that in previous work \citep{nguyen2023building}, where the left gyrotranslation law is presented in \cref{app:lem:left_gyrotranslation}.
The previous work proves the results on the SPD and Grassmannian manifolds in a case-by-case manner. In contrast, we relax the left reduction into pseudo-reduction and give a general analysis. Our framework also corrects the proof for the Grassmannian cases.
}
\label{fig:logics_isometry}
\end{figure}

Given a gyrogroup $(G, \oplus)$, the left gyrotranslation by $x \in G$ is defined as
\begin{equation}
     L_x : G \rightarrow G, \quad L_{x}(y) = x \oplus y, \quad \forall y \in G.
\end{equation}
If any gyrotranslation is a gyroisometry, we can use gyrotranslation to center manifold-valued samples for the normalization layer. \citet{nguyen2023building} shows that any left gyrotranslation on the SPD and Grassmannian manifolds is a gyroisometry. However, the proof relies on the left cancellation law of gyrogroups, which does not hold for \emph{non-reductive gyrogroups}, such as the Grassmannian. Here, non-reductive gyrogroups refer to gyro-like groupoids that satisfy the first three gyrogroup axioms but fail the last axiom in \cref{def:ch2-gyrogroup}, namely the left reduction law (G4), $\gyr[x,y]=\gyr[x \oplus y,y]$. Therefore, the proof is not generally valid for the Grassmannian. We propose an intermediate structure, referred to as \emph{pseudo-reductive gyrogroups}, which supports the left cancellation law and, therefore, the gyroisometry of gyrotranslation. This structure forms the algebraic foundation for building the normalization layer. As illustrated in \cref{fig:logics_isometry}, our derivation extends the case-by-case approach by \citet{nguyen2023building} to general pseudo-reductive gyrogroups.

\subsubsection{From Gyrogroups to Pseudo-Reductive Gyrogroups}

\begin{parisdefinition}[Pseudo-reductive gyrogroups]
\label{def:pseudo_reductive_gyrogroup}
    A groupoid $(G, \oplus)$ is a pseudo-reductive gyrogroup if it satisfies the axioms (G1), (G2), (G3) and the following pseudo-reductive law:
    \begin{equation} \label{eq:pseudo_reduction}
        \gyr[a, x] = \id, \text{ for any left inverse } a \text{ of } x \text{ in } G,
    \end{equation}
    where $\id$ is the identity map.
\end{parisdefinition}

\cref{eq:pseudo_reduction} can be intuitively viewed as an intermediate between reduction and non-reduction. For gyrogroups, \cref{eq:pseudo_reduction} can be directly obtained from left gyroassociativity (G3) and reduction (G4) \citep[Thm.~2.10, item~3]{ungar2022analytic}. However, there is no theoretical guarantee that \cref{eq:pseudo_reduction} holds for non-reductive gyrogroups. Therefore, we name \cref{eq:pseudo_reduction} pseudo-reduction. Nevertheless, for the specific non-reductive Grassmannian, it is indeed pseudo-reductive.

\begin{parisproposition}\label{prop:grassmannian_pseudo_reductive_gyrogroups}
    \linktoproof{prop:grassmannian_pseudo_reductive_gyrogroups}
    $\grasonb{p,n}$ and $\graspp{p,n}$ are pseudo-reductive gyrocommutative gyrogroups.
\end{parisproposition}

Our pseudo-reductive gyrogroup naturally generalizes the vanilla gyrogroup, as it shares most of the basic properties of gyrogroups \citep[Thms.~2.10--2.11]{ungar2022analytic}.

\begin{paristheorem}[First pseudo-reductive gyrogroup properties]
    \label{thm:pseudo_reductive_gyrogroups_properties}
    \linktoproof{thm:pseudo_reductive_gyrogroups_properties}
    Let $(G, \oplus)$ be a pseudo-reductive gyrogroup. For any elements $x, y, z, a \in G$, we have:
    \par\leavevmode\vspace{-\baselineskip}
    \begin{enumerate}
        \item \label{enu:prgp_1}
        If $x \oplus y = x \oplus z$, then $y = z$ (General Left Cancellation law; see \cref{enu:prgp_8} below).
        \item \label{enu:prgp_2}
        $\gyr[e, x] = \id$ for any left identity $e$ in $G$.
        \item \label{enu:prgp_3}
        $\gyr[a, x] = \id$ for any left inverse $a$ of $x$ in $G$.
        \item \label{enu:prgp_4}
        There is a left identity that is a right identity.
        \item \label{enu:prgp_5}
        There is only one left identity.
        \item \label{enu:prgp_6}
        Every left inverse is a right inverse.
        \item \label{enu:prgp_7}
        There is only one left inverse, $\ominus x$, of $x$, and $\ominus(\ominus x) = x$.
        \item \label{enu:prgp_8}
        The left cancellation law: $\ominus x \oplus (x \oplus y) = y$.
        \item \label{enu:prgp_9}
        The gyrator identity: $\gyr[x, y] a = \ominus(x \oplus y) \oplus \{x \oplus (y \oplus a)\}$.
        \item \label{enu:prgp_10}
        $\gyr[x, y] e = e$.
        \item \label{enu:prgp_11}
        $\gyr[x, y](\ominus a) = \ominus \gyr[x, y] a$.
        \item \label{enu:prgp_12}
        $\gyr[x, e] = \id$.
        \item \label{enu:prgp_13}
        The gyrosum inversion law: $\ominus(x \oplus y) = \gyr[x, y](\ominus y \oplus \ominus x)$.
    \end{enumerate}
\end{paristheorem}
\begin{parisremark}
In non-reductive gyrogroups, the identities in \cref{enu:prgp_2,enu:prgp_3} are not guaranteed to hold. Consequently, any property relying on them, such as \cref{enu:prgp_4} and those from \cref{enu:prgp_6} to \cref{enu:prgp_10}, is also not guaranteed to hold. The absence of these basic properties undermines the rationality of non-reductive gyrogroups. In contrast, our pseudo-reductive gyrogroups preserve most of the fundamental properties of gyrogroups.
\end{parisremark}

\subsubsection{Isometries over Pseudo-Reductive Gyrogroups}
The gyro-structure in the following is assumed to be defined as \cref{eq:ch2-riem-gyro-addition}--\cref{eq:ch2-gyro-distance}. We first clarify that the Riemannian distance agrees with the gyrodistance, and the Riemannian isometry agrees with the gyroisometry. These justify gyrodistance and gyroisometry for gyrospaces over manifolds.
\begin{parislemma}[Distances]
\label{lem:gyro_geodesic_dist}
\linktoproof{lem:gyro_geodesic_dist}
    Given a pseudo-reductive gyrogroup $(\calM, \oplus)$, we have
    \begin{equation}
            \dist(x,y) = \norm{\rielog_x (y)}_x = \gyrnorm{ \ominus x \oplus y} = \gyrdist(x,y), \quad \forall x,y \in \calM,
    \end{equation}
    where $\dist$ denotes the geodesic distance.\footnotemark
\end{parislemma}
\footnotetext{On Cartan--Hadamard manifolds, the statement holds for all $x,y \in \calM$. More generally, the equality requires $x,y$ to lie within a geodesic ball of convexity radius to ensure the well-definedness of the minimizing geodesic and logarithm. In this section, we implicitly assume these conditions are satisfied.}
\begin{parislemma}[Isometries]
\label{lem:isometry_for_gyro}
\linktoproof{lem:isometry_for_gyro}
Let $(\calM, \oplus)$ and $(\widetilde{\calM}, \widetilde{\oplus})$ be two pseudo-reductive gyrogroups. Their gyro identity elements are $e \in \calM$ and $\widetilde{e} \in \widetilde{\calM}$, respectively. If $\phi: \calM \to \widetilde{\calM}$ is a Riemannian isometry with $\widetilde{e}=\phi(e)$, then the following hold.
\par\leavevmode\vspace{-\baselineskip}
\begin{enumerate}
    \item
    The Riemannian isometry is a gyroisometry:
    \begin{equation}
        \gyrdist(x, y) = \widetilde{\gyrdist}(\phi(x), \phi(y)),
    \end{equation}
    where $\gyrdist$ and $\widetilde{\gyrdist}$ are the gyrodistances over $\calM$ and $\widetilde{\calM}$, respectively.
    \item
    If the gyroinverse, gyration, or left gyrotranslation over $\calM$ is a gyroisometry, its counterpart over $\widetilde{\calM}$ is also a gyroisometry.
\end{enumerate}
\end{parislemma}
\cref{lem:gyro_geodesic_dist} implies that, as long as the gyro-structure is defined by \cref{eq:ch2-riem-gyro-addition}--\cref{eq:ch2-gyro-distance}, the gyrodistance coincides with the geodesic distance. Unless otherwise specified, we shall not distinguish between the two and uniformly denote them by $\dist(\cdot,\cdot)$. Besides, the second result in \cref{lem:isometry_for_gyro} is particularly useful, as several geometries are isometric, such as the ONB and PP Grassmannian, as well as different models in hyperbolic geometry.

Now, we analyze gyroisometries over pseudo-reductive gyrogroups. The most related property in \cref{thm:pseudo_reductive_gyrogroups_properties} is the left cancellation law, one of the key prerequisites for a gyrotranslation to be a gyroisometry. Note that the left cancellation comes from left gyroassociativity and \cref{eq:pseudo_reduction} \citep[Thm.~2.10, item~9]{ungar2022analytic}. Therefore, left cancellation does not generally hold for non-reductive gyrogroups but exists in pseudo-reductive gyrogroups. We first present an if-and-only-if statement about gyroisometry, which will be useful in the following.

\begin{paristheorem}
    \label{thm:iff_gyroauto_gyroisometries}
    \linktoproof{thm:iff_gyroauto_gyroisometries}
    Given a pseudo-reductive gyrogroup $(G, \oplus)$, $\gyr[x, y]$ preserves the gyronorm for any $x,y \in G$ if and only if $\gyr[x, y]$ is a gyroisometry for any $x,y \in G$.
\end{paristheorem}

The gyroisometry of any gyration is a prerequisite for other operators to be gyroisometries.
\begin{paristheorem}[Gyroisometries]
    \label{thm:gyroisometries}
    \linktoproof{thm:gyroisometries}
    Given a pseudo-reductive gyrogroup $(G, \oplus)$ with any $\gyr[\cdot, \cdot]$ as a gyroisometry, we have the following.
    \par\leavevmode\vspace{-\baselineskip}
    \begin{enumerate}
        \item
        The left gyrotranslation is a gyroisometry.
        \item
        If $(G, \oplus)$ is gyrocommutative, then any gyroinverse is a gyroisometry.
    \end{enumerate}
\end{paristheorem}
Now, we discuss the gyroisometries for the gyro-structures reviewed in \cref{tab:ch2-spd-lie-operators,tab:ch2-grassmann-gyro-operators,tab:ch2-constant-curvature-gyro-operators}.
\begin{paristheorem}
    \label{thm:gyroinvariance}
    \linktoproof{thm:gyroinvariance}
    For the pseudo-reductive gyrogroups corresponding to the SPD manifold (under AIM, LEM, and LCM), the ONB and PP Grassmannian, and the stereographic model with $K \leq 0$ (the Poincaré ball for $K<0$ and Euclidean space for $K=0$), the gyrodistance coincides with the geodesic distance. Moreover, the gyroinverse, any gyration, and any left gyrotranslation are gyroisometries.
\end{paristheorem}
\begin{proof}[Credit and sketch of the proof]
As \cref{lem:gyro_geodesic_dist} already shows that the gyrodistance coincides with the geodesic distance, it remains to establish the isometries. For the Grassmannian and SPD manifolds, these were proved by \citet[Thms.~2.12--2.14 and~2.16--2.18]{nguyen2023building}, although their arguments implicitly treated the non-reductive Grassmannian as a gyrogroup by using left cancellation. Our \cref{prop:grassmannian_pseudo_reductive_gyrogroups,thm:pseudo_reductive_gyrogroups_properties} confirms that the Grassmannian is pseudo-reductive and does satisfy left cancellation, thereby validating their results. However, we can directly establish these properties from \cref{thm:iff_gyroauto_gyroisometries,thm:gyroisometries}. The complete proof is given in \cref{app:subsec:proof_gyroinvariance}.
\end{proof}
\begin{parisremark}
The remaining constant-curvature manifolds reviewed in \cref{sec:ch2-constant-curvature-manifolds}, including the stereographic model with $K>0$, radius model, and Beltrami--Klein model, also satisfy these properties. Their verifications will be presented in \cref{sec:ch3-gyrobn-instantiations}.
\end{parisremark}

\subsection{GyroBN on Pseudo-Reductive Gyrogroups}
\label{sec:ch3-gyrobn-general}
Building on \cref{thm:gyroinvariance}, which establishes that several geometries admit isometric gyrotranslations, we develop RBN in a principled way for general pseudo-reductive gyrogroups, referred to as GyroBN. Throughout, we assume $(\calM, \oplus)$ is a pseudo-reductive gyrogroup with the gyro-structure defined by \cref{eq:ch2-riem-gyro-addition}--\cref{eq:ch2-gyro-distance}.\footnote{In GyroBN, $\odot$ is not required to satisfy the axioms of a gyrovector space (\cref{def:ch2-gyrovector-space}).} As \cref{lem:gyro_geodesic_dist} establishes the equivalence between gyrodistance and geodesic distance, we use the terms ``gyromean'' and ``gyrovariance'' interchangeably with their Riemannian counterparts.

\subsubsection{GyroBN}
To generalize the Euclidean BN in \cref{eq:ebn} to gyrogroups, we first define sample mean, sample variance, centering, biasing, and scaling over gyrogroups. Then, we introduce the GyroBN framework with a theoretical analysis of the ability to normalize sample statistics.

We define the gyromean as the Fréchet mean \citep{frechet1948elements} under gyrodistance:
\begin{equation} \label{eq:gyromean}
    \mu = \fm(\{ x_i \in \calM \}_{i=1}^N) = \underset{y \in \calM}{\argmin} \frac{1}{N} \sum\nolimits_{i=1}^N  \dist^2\left(x_i, y \right).
\end{equation}
The gyrovariance is the corresponding Fréchet variance. By \cref{lem:gyro_geodesic_dist}, the gyromean and gyrovariance coincide with the Riemannian mean and variance, \ie the Fréchet mean and variance under geodesic distance. The existence and local uniqueness of the Fr\'echet mean are reviewed in \cref{def:ch2-frechet-mean}.

Easy computation shows that the Euclidean BN operations in \cref{eq:ebn} have direct gyrogroup counterparts in $\bbR{n}$. Centering corresponds to gyrosubtraction (\cref{eq:ch2-riem-gyro-addition,eq:ch2-riem-gyro-inverse}), biasing to gyroaddition (\cref{eq:ch2-riem-gyro-addition}), and scaling to scalar gyromultiplication (\cref{eq:ch2-riem-gyro-scalar}). Motivated by this, we define a normalization layer over gyrogroups via gyro operations. Given a batch of activations $\{x_i\}_{i=1}^N \subset \calM$, the core operations of GyroBN are
\begin{equation}
    \label{eq:gyrobn_centering}
    \forall i \leq N, \quad \tilde{x}_i \gets  \overbrace{ \beta \oplus}^{\text{Biasing}} \left( \overbrace{\frac{s}{\sqrt{v^2+\epsilon}} \odot }^{\text{Scaling}} \left( \overbrace{\ominus \mu  \oplus x_i}^{\text{Centering}} \right) \right),
\end{equation}
where $\mu \in \calM$ and $v^2$ are gyromean and gyrovariance, $\beta \in \calM$ is the bias parameter, $s \in \bbRscalar$ is the scaling parameter, and $\epsilon$ is a small value for numerical stability. The following theorem shows that \cref{eq:gyrobn_centering} can normalize manifold-valued data.

\begin{paristheorem}[Homogeneity]
    \label{thm:gyrobn} \linktoproof{thm:gyrobn}
    Let $(\calM, \oplus)$ be a pseudo-reductive gyrogroup in which every gyration $\gyr[\cdot,\cdot]$ is a gyroisometry. For $N$ samples $\{x_i\}_{i=1}^N \subset \calM$ and any $t \in \bbRscalar$, we have
    \begin{align}
        & \text{Homogeneity of gyromean: }
        \fm( \{\beta \oplus x_i \}_{i=1}^N ) = \beta \oplus \fm( \{ x_i \}_{i=1}^N ), \quad \forall \beta \in \calM,\\
        & \text{Homogeneity of dispersion from $e$: }
        \frac{1}{N} \sum\nolimits_{i=1}^N \dist^2(t \odot x_i, e) = \frac{t^2}{N} \sum\nolimits_{i=1}^N \dist^2(x_i, e),
    \end{align}
\end{paristheorem}

The most important property of the Euclidean BN~\citep{ioffe2015batch} lies in its ability to normalize the sample mean and variance. \cref{thm:gyrobn} shows that the formulation in \cref{eq:gyrobn_centering} enjoys the same property: homogeneity of the gyromean guarantees that centering and biasing shift the gyromean, while homogeneity of the dispersion from $e$ ensures that scaling controls the sample variance. As a result, GyroBN provides a theoretical guarantee of normalization on any pseudo-reductive gyrogroup with isometric gyrations. Moreover, since the gyromean and gyrovariance coincide with their Riemannian counterparts, GyroBN also normalizes Riemannian statistics.

\begin{algorithm}[t]
\SetKwInOut{Input}{Require}
\SetKwInOut{Output}{Return}
\SetKwInOut{Parameters}{Parameters}
\caption{Gyrogroup Batch Normalization (GyroBN)}
\label{alg:gyrobn}
\Input{
batch of activations $\{x_i\}_{i=1}^N \subset \calM$, small positive constant $\epsilon$, and momentum $\eta \in [0,1]$, running mean $\mu_r$, running variance $v^2_r$, bias parameter $\beta \in \calM$, scaling parameter $s \in \bbRscalar $.
}
\Output{normalized batch $\{\tilde{x}_i\}_{i=1}^N \subset \calM$}
\BlankLine
\If{training}{
    Compute batch mean $\mu_b$ and variance $v_b^2$ of $\{x_i\}_{i=1}^N$;\\
    Update running statistics
    $\mu_r = \barcenter_\eta (\mu_b,\mu_r)$, and $v^2_r = \eta v^2_b + (1-\eta)v^2_r$;
}

$(\mu,v^2) = (\mu_b, v^2_b )$ \textbf{if} \textit{training} \textbf{else} $(\mu_r, v_r^2)$

$\forall i \leq N, \ \tilde{x}_i = \beta \oplus \left( \frac{s}{\sqrt{v^2+\epsilon}} \odot \left( \ominus \mu \oplus x_i \right) \right)$
\end{algorithm}

To finalize GyroBN, we define the running mean updates over gyrogroups as the binary barycenter based on gyrodistance:
\begin{equation}
        \barcenter_\eta (x_1, x_2) = {\argmin}_{y \in \calM}  \left( \eta \dist^2\left(x_1, y \right) + (1-\eta) \dist^2\left(x_2, y \right)  \right), \quad \eta \in [0,1],
\end{equation}
which can be calculated by the geodesic. With these ingredients, the general framework for GyroBN is presented in \cref{alg:gyrobn}. In particular, it recovers the classic Euclidean BN~\citep{ioffe2015batch} when $\calM=\bbR{n}$.

\begin{parisremark}
We make the following two remarks with respect to left gyrotranslation.
\begin{itemize}
    \item
    \mypara{Other candidates.} There are three alternatives to left gyrotranslation. However, they are not necessarily
    gyroisometries and therefore cannot support the general GyroBN construction without additional isometry assumptions.
    Specifically, analogous to the left gyrotranslation, the right gyrotranslation is
    \begin{equation}
         R_x : G \rightarrow G, \quad R_{x}(y) = y \oplus x, \quad \forall y \in G.
    \end{equation}
    Along with the gyroaddition, there is the gyrogroup coaddition \citep[Def.~2.9]{ungar2022analytic}:
    \begin{equation}
        x \boxplus y = x \oplus \gyr[x, \ominus y] y, \quad \forall x, y \in G.
    \end{equation}
    Coaddition is symmetric to gyroaddition in many ways; for instance, when gyroaddition is
    gyrocommutative, coaddition is commutative \citep[Thm.~3.3]{ungar2022analytic}.
    However, the right gyrotranslation, as well as the left and right translations by coaddition,
    are not guaranteed to be gyroisometries. Numerical experiments confirm that these three
    translations on the hyperbolic Poincaré ball fail to preserve gyrodistance (see
    \texttt{gyrocoadd.py}). A theoretical reason is that they all lack a counterpart of the left
    gyrotranslation law, which is important for the translation to be a gyroisometry
    (see~\cref{fig:logics_isometry}).

    \item
    \mypara{Special cases.} For Lie groups with a right-invariant Riemannian metric, right translations are isometries, and GyroBN can then be formulated using them, with \cref{thm:gyrobn} extending directly. In general, however, only left gyrotranslations are guaranteed to be isometries.
\end{itemize}
\end{parisremark}

\subsection{Instantiations}
\label{sec:ch3-gyrobn-instantiations}
As indicated by \cref{thm:gyroinvariance,thm:gyrobn}, GyroBN can be applied to different geometries with guaranteed normalization of the sample statistics. Once the required operators are specified, \cref{alg:gyrobn} can be used in a plug-and-play manner. We first show that existing RBN methods with control over sample statistics, such as LieBN on Lie groups and AIM-based SPDBNs, are special cases of the framework. We then instantiate GyroBN on seven representative geometries: the Grassmannian, five CCS models (Poincaré ball, projected hypersphere, Lorentz, sphere, and Beltrami--Klein), and the full-rank correlation manifold. To support these instantiations, we simplify the Grassmannian operators for efficient computation, refine the gyro-structures on the Poincaré ball and projected hypersphere, establish new gyro-structures for Lorentz and spherical geometries, characterize the Beltrami--Klein geometry, and formulate a row-wise realization for the correlation manifold.

\subsubsection{LieBN as a Special Case}
\label{subsec:liebn-as-gyrobn}
\citet[Algs.~3--4]{chakraborty2020manifoldnorm} introduced the Riemannian normalization on matrix Lie groups under a specific distance. The extension to general Lie groups, yielding LieBN with theoretical normalization over the Riemannian mean and variance, is presented in \cref{sec:ch3-lie-group-approach}. This subsection shows that LieBN is a special case of GyroBN.

LieBN is formulated with an invariant metric on a Lie group. Centering and biasing are performed through group translations, while scaling is defined in the tangent space at the identity element. Since every Lie group is automatically a gyrogroup, gyrotranslation reduces exactly to the group translation. Consequently, the centering, biasing, and scaling in LieBN are identical to those in GyroBN. Moreover, as shown in \cref{lem:gyro_geodesic_dist}, the mean, variance, and running mean update defined via the geodesic distance in LieBN are equivalent to their counterparts based on gyrodistance in GyroBN. Therefore, LieBN is a special case of GyroBN. The LieBN instantiations presented in \cref{sec:manifestations} are therefore incorporated by GyroBN.

\subsubsection{AIM-Based SPDBNs as Special Cases}
\label{subsec:spdbn-as-gyrobn}
As reviewed in \cref{subsec:revisit_rbn}, several RBNs on the SPD manifold were developed based on AIM \citep{brooks2019riemannian,kobler2022controlling,kobler2022spd}. These SPD normalization methods can be expressed in the following unified form:
\begin{equation} \label{eq:spdbn_aim}
    \text{Normalization: }
    \forall i \leq N, \quad \widetilde{P}_i \gets B^{\frac{1}{2}} \left( M^{-\frac{1}{2}} P_i M^{-\frac{1}{2}} \right)^{\frac{s}{\sqrt{v^2+\epsilon}}} B^{\frac{1}{2}},
\end{equation}
where $M$ and $v^2$ are the Riemannian mean and variance. The running mean is updated by the binary barycenter under the geodesic distance.

As gyrodistance is identical to the geodesic distance, the gyromean, gyrovariance, and running mean updates are identical to the Riemannian ones. Using the AIM gyro-operations in \cref{eq:ch2-aim-gyro-operations}, \cref{eq:spdbn_aim} is exactly the specific implementation of \cref{eq:gyrobn_centering} under the AIM-based gyrogroup on the SPD manifold. Therefore, the SPDBNs developed by \citet{brooks2019riemannian,kobler2022controlling,kobler2022spd} are also special cases of GyroBN.
\begin{parisremark}
\citet{brooks2019riemannian} only considers centering and biasing.
\citet{kobler2022controlling} uses a running mean for centering during training. \citet{kobler2022spd} uses different momentum parameters to update running statistics for training and testing, along with multi-channel mechanisms for domain adaptation. Nevertheless, all of them are based on \cref{eq:spdbn_aim}.
Therefore, tricks such as multi-channel and separate momentum can also be applied to GyroBN. This is what we mean by claiming that GyroBN incorporates their approaches.
\end{parisremark}

\subsubsection{Grassmannian Manifold}
\label{subsec:grassmannian_gyrobn}

We focus on the ONB perspective. Let $U,V \in \grasonb{p,n}$, $t \in \bbRscalar$, and $\Delta \in T _U \grasonb{p,n}$. Using the Grassmannian operators reviewed in \cref{tab:ch2-grassmann-riemannian-operators,tab:ch2-grassmann-gyro-operators}, we instantiate GyroBN on the ONB Grassmannian.

\mypara{Instantiation.} Building on these operators, we now implement the ONB Grassmannian GyroBN. Given a batch of activations $\{U_{1 \cdots N}\}$, the three core steps of GyroBN are
\begin{align}
    \text{Centering to the identity $\idonb$: }&
    U^1_i = \mexp \left( -[\overline{M M^\top}, \idpp]\right) U_i,\\
    \text{Scaling the dispersion from $\idonb$: }&
     U^2_i = \mexp \left(  \frac{s}{\sqrt{v^2+\epsilon}} [\overline{U^1_i(U^1_i)^\top}, \idpp] \right) \idonb,\\
     \label{eq:gyrobn_grass_biasing}
     \text{Biasing towards $B \in \calM$: }&
     U^3_i = \mexp \left([\overline{B}, \idpp] \right)U^2_i .
\end{align}
Here $\overline{(\cdot)} = \widetilde{\rielog}_{\idpp}(\cdot)$ is the Riemannian logarithm under the PP Grassmannian, $M$ (resp. $v^2$) is the Riemannian batch mean (resp. variance), and $\idpp=\idonb \idonb^\top$ is the PP identity. The mean $M$ can be obtained by the Karcher flow~\citep{karcher1977riemannian}, with the Riemannian logarithm computed by \citet[Alg.~5.3]{bendokat2024grassmann}.

\mypara{Efficient computation.} The commutators $[\overline{M M^\top}, \idpp]$ and $[\overline{U^1_i(U^1_i)^\top},\idpp]$ can be efficiently computed by the following result.

\begin{parisproposition}
    \label{prop:fast_bracket_grassmannian}
    \linktoproof{prop:fast_bracket_grassmannian}
    Given $U=(U_1^\top,U_2^\top)^\top \in \grasonb{p,n}$ with $U_1 \in \bbR{p \times p}$ and $U_2 \in \bbR{(n-p) \times p}$, then
    \begin{equation}
        [\overline{U U^\top},\idpp] =
        \begin{pmatrix}
             \bbzero_{p\times p} & -\widetilde{U}_2^\top \\
             \widetilde{U}_2 & \bbzero_{(n-p)\times(n-p)}
        \end{pmatrix},
    \end{equation}
    where $\widetilde{U}_2 = U_2 Q \frac{\arcsin(\hat{S})}{\hat{S}}R^\top$ and $U_1^\top \stackrel{\mathrm{SVD}}{:=} QSR^\top$.
    Here $S$ is in ascending order, $Q$ and $R$ are flipped column-wise, and $\hat{S} = \sqrt{I_p-S^2}$.
\end{parisproposition}

\begin{parisremark}
Two technical issues are worth noting.
\begin{itemize}
    \item \mypara{Cut locus.} The logarithm $\rielog_U(V)$ exists only when $U$ and $V$ are not in each other's cut locus~\citep{bendokat2024grassmann}. Similarly, gyroaddition and gyromultiplication are not globally defined due to the cut locus~\citep[Sec.~3.2]{nguyen2022gyro}. However, \citet[Alg.~5.3]{bendokat2024grassmann} provides a numerical remedy.
    \item \mypara{PP Grassmannian.} Although our derivation is based on the ONB Grassmannian, GyroBN under the PP Grassmannian can be obtained by mapping data via $\pi^{-1}: \graspp{p,n} \to \grasonb{p,n}$, normalizing, and mapping back via $\pi$. This follows from the isometry $\pi: \grasonb{p,n} \to \graspp{p,n}$.
\end{itemize}
\end{parisremark}

\subsubsection{Stereographic Model}
\label{subsec:stereo_model}

We begin by analyzing its gyro-structure and then instantiate GyroBN.

\mypara{Stereographic gyrovector space.} As reviewed in \cref{sec:ch2-constant-curvature-manifolds}, the stereographic model $\stereo{n}$ unifies CCS geometries: the hyperbolic Poincaré ball $\pball{n}$ for $K<0$, Euclidean space $\bbR{n}$ for $K=0$, and the spherical projected hypersphere $\projhs{n}$ for $K>0$. For $x,y,z \in \stereo{n}$, $t \in \bbRscalar$, and $v \in T_x\stereo{n}$, its Riemannian and gyro operators are reviewed in \cref{tab:ch2-constant-curvature-operators,tab:ch2-constant-curvature-gyro-operators}.

For $K<0$, the Poincaré ball $(\pball{n},\stoplus,\stodot)$ forms a Möbius gyrovector space, as reviewed in \cref{tab:ch2-constant-curvature-gyro-operators}. For $K>0$, however, the situation is subtler \citep{bachmann2020constant}:
\begin{itemize}
    \item gyroaddition is well-defined except when $x = \frac{y}{K\norm{y}^2}$;
    \item gyromultiplication is well-defined except when $r\tan^{-1}(\sqrt{K}\norm{x}) = \nicefrac{\pi}{2} + k\pi$ for some $k \in \bbZ$.
\end{itemize}
Even if assumed well-defined, it remains unclear whether $(\stereo{n},\stoplus,\stodot)$ with $K>0$ satisfies the axioms of a gyrovector space, in contrast to the Poincaré ball. Closing this gap is part of our contribution: under the assumption of well-definedness, we show that the stereographic gyro operations coincide with those in \cref{eq:ch2-riem-gyro-addition,eq:ch2-riem-gyro-scalar} and prove that the stereographic model satisfies all axioms of a gyrovector space.

\begin{parisproposition} \label{prop:stereographic_gyro_from_riemannian}
    \linktoproof{prop:stereographic_gyro_from_riemannian}
    The stereographic gyroaddition and gyromultiplication coincide with the Riemannian definitions:
    \begin{align}
        x \stoplus y &= \rieexp_{x}\left(\pt{\zerovec}{x}(\rielog_{\zerovec}(y))\right), \quad \forall x,y \in \stereo{n}, \\
        t \stodot x &= \rieexp_{\zerovec}\left(t \rielog_{\zerovec}(x)\right), \quad \forall t \in \bbRscalar, x \in \stereo{n}.
    \end{align}
\end{parisproposition}

\begin{paristheorem} \label{thm:stereographic_gyro}
    \linktoproof{thm:stereographic_gyro}
    For any $K\in\bbRscalar$, the stereographic model $(\stereo{n},\stoplus)$ satisfies all axioms of a gyrocommutative gyrogroup. When further endowed with gyromultiplication $\stodot$, it satisfies all axioms of a gyrovector space.
\end{paristheorem}

\mypara{Stereographic GyroBN.} Next, we extend \cref{thm:gyroinvariance} to the stereographic model with arbitrary curvature.

\begin{paristheorem}
    \label{thm:gyroinvariance_stereo}
    \linktoproof{thm:gyroinvariance_stereo}
    For the stereographic model, the gyrodistance coincides with the geodesic distance. Moreover, the gyroinverse, any gyration, and any left gyrotranslation are gyroisometries.
\end{paristheorem}

Combining \cref{thm:gyrobn} and \cref{thm:gyroinvariance_stereo}, GyroBN in the stereographic model is theoretically guaranteed to normalize sample statistics. Practically, implementation only requires substituting the stereographic operators reviewed in \cref{tab:ch2-constant-curvature-operators,tab:ch2-constant-curvature-gyro-operators} into \cref{alg:gyrobn}. For efficient computation, the Poincaré Fréchet mean can be obtained using the algorithm of \citet[Alg.~1]{lou2020differentiating}, while the mean on the sphere is computed via the Karcher flow \citep{karcher1977riemannian}.

\subsubsection{Radius Model}
\label{subsec:radius-model}
As reviewed in \cref{sec:ch2-constant-curvature-manifolds}, the radius model $\calMK{n}$ is another representation of constant-curvature spaces, unifying Lorentz space $\lorentz{n}$ for $K<0$, the sphere $\sphere{n}$ for $K>0$, and the Euclidean space $\bbR{n}$ for $K=0$. The case $K=0$ reduces trivially to Euclidean space, so the following analysis of the radius-model gyro-structure considers $K \neq 0$. Although this model has been effective in various applications \citep{chami2019hyperbolic,chen2022fully,bdeir2024fully,pal2025compositional,he2025lorentzian,khan2025hyperbolic}, its gyro-structure has not been formalized. We first analyze the gyro-structure on $\calMK{n}$ and then instantiate GyroBN.

\mypara{Radius gyrovector space.} The radius model $\calMK{n}$ is isometric to the stereographic model $\stereo{n}$ via stereographic projection fixing the south pole \citep{skopek2020mixed}:
\begin{align}
    \label{eq:iso_calMK_to_stereo}
    \isoMKST{n}
    &: \calMK{n} \ni
    \begin{bmatrix}
    \xi \in \bbRscalar \\
    x \in \bbR{n}
    \end{bmatrix} \longmapsto \frac{x}{1+\sqrt{|K|} \xi} \in \stereo{n}, \\
    \label{eq:iso_stereo_to_calMK}
    \isoSTMK{n}
    &: \stereo{n} \ni y \longmapsto
    \begin{bmatrix}
    \frac{1}{\sqrt{|K|}} \frac{1-K\| y \|^2}{1+K\| y \|^2} \\
    \frac{2 y}{1+ K\| y \|^2}
    \end{bmatrix} \in \calMK{n}.
\end{align}
The origin in $\calMK{n}$ is defined as $\MKzero=[\sqrt{1/|K|}, 0, \ldots, 0]^\top$, corresponding to $\zerovec \in \stereo{n}$. For $K>0$, we have $\calMK{n}=\sphere{n}$ and $\stereo{n}=\projhs{n}$. Since \cref{eq:iso_calMK_to_stereo} is undefined at the south pole $-\MKzero$ when $K>0$, we use the one-point compactification $\projhs{n}\cup\{\infty\}$ with the identification $\pi_{\calMK{n}\to\stereo{n}}(-\MKzero)=\infty$ \citep[Rmk.~A.9]{skopek2020mixed}. For simplicity, we use $\projhs{n}$ and $\projhs{n}\cup\{\infty\}$ interchangeably.

\cref{tab:ch2-constant-curvature-operators} reviews the Riemannian operators. We adopt the following curvature-aware functions:
\begin{equation}
    \begin{aligned}
        \sin _K &=
        \begin{cases}
            \sin & \text{if } K>0, \\
            \sinh & \text{if } K<0,
        \end{cases}
        &
        \cos _K &=
        \begin{cases}
            \cos & \text{if } K>0, \\
            \cosh & \text{if } K<0,
        \end{cases} \\
        \Kinner{\cdot}{\cdot} &=
        \begin{cases}
            \inner{\cdot}{\cdot} & \text{if } K>0, \\
            \Linner{\cdot}{\cdot} & \text{if } K<0.
        \end{cases}
    \end{aligned}
\end{equation}
For $v\in T_x\calMK{n}$, $\Knorm{v}=\sqrt{\Kinner{v}{v}}$ is the induced Riemannian norm. On the $K<0$ branch, the ambient Lorentzian form is positive definite only after this tangent-space restriction. Moreover, $(\cdot)_s$ denotes the space vector, and $(\cdot)_t$ denotes the time scalar.

Using the radius-model Riemannian operators reviewed in \cref{tab:ch2-constant-curvature-operators}, we define the gyroaddition and gyromultiplication as \cref{eq:ch2-riem-gyro-addition,eq:ch2-riem-gyro-scalar}:
\begin{align}
    \label{eq:calmk_gyroadd_def}
    x \MKoplus y
    &= \rieexp_{x}\left(\pt{\MKzero}{x} \left(\rielog _{\MKzero}(y)\right)\right), \quad \forall x,y \in \calMK{n}, \\
    \label{eq:calmk_gyro_prod_def}
    t \MKodot x
    &= \rieexp _{\MKzero} \left( t \rielog _{\MKzero} (x) \right), \quad \forall t \in \bbRscalar, \forall x \in \calMK{n}.
\end{align}

We give the following clarifications regarding the above two gyro operations.
\begin{itemize}
    \item
    For hyperbolic geometry ($K < 0$), \cref{eq:calmk_gyroadd_def} has been employed in prior work~\citep{chami2019hyperbolic,he2025lorentzian}. However, there exists no closed-form expression, which could be more efficient than the composition of Riemannian operators. Besides, the underlying gyro-structure has not been formally discussed.
    \item
    For spherical geometry ($K>0$), the geodesic between antipodal points ($x$ and $-x$) is not unique, making the logarithm and parallel transport along such a geodesic ill-defined. For the sphere, \cref{eq:calmk_gyroadd_def} assumes that $x \neq -\MKzero$ and $y \neq -\MKzero$, while \cref{eq:calmk_gyro_prod_def} assumes that $x \neq -\MKzero$. Like before, we always make these assumptions implicitly.
\end{itemize}

In the following, we first give the closed-form expressions of \cref{eq:calmk_gyroadd_def,eq:calmk_gyro_prod_def}. Then, we show that \cref{eq:calmk_gyroadd_def,eq:calmk_gyro_prod_def} conform to all the axioms of a gyrovector space.

\begin{parisproposition}[Gyromultiplication and gyroinverse]
    \label{prop:calmk_scalar_prod_inv}
    \linktoproof{prop:calmk_scalar_prod_inv}
    Let $x=[x_t,x_s^\top]^\top$ be a point in $\calMK{n}$, where $x_t \in \bbRscalar$ is the time scalar, and $x_s \in \bbR{n}$ is the spatial part. The gyromultiplication and inverse have closed-form expressions:
    \begin{align}
    \label{eq:gyro_prod_calmk}
    t \MKodot x
    &=
    \begin{cases}
    \MKzero, & t = 0 \lor x=\MKzero, \\
    \frac{1}{\sqrt{|K|}}
    \begin{bmatrix}
    \cosk \left( t \cosk^{-1}(\sqrt{|K|}  x_t) \right) \\
    \dfrac{\sink \left( t \cosk^{-1}(\sqrt{|K|}  x_t) \right)}{\norm{x_s}}  x_s
    \end{bmatrix}, & t \neq 0,
    \end{cases} \\
    \MKominus x &
    = -1 \MKodot x
    =
    \begin{bmatrix}
    x_t \\
    -x_s
    \end{bmatrix}.
    \end{align}
    In particular, the gyro identity is $\MKzero$. Besides, the following shows that the sphere gyromultiplication is still valid for the singular cases in the projected hypersphere. Assume $K>0$ and $x \neq \pm \MKzero$, and let $\stereo{n} \ni u=\isoMKST{n}(x)$ be its stereographic image. For $t\in \bbRscalar$, set $\theta = \cos^{-1} \left( \sqrt{K} x_t \right) \in (0,\pi)$. Then the following are equivalent:
    \begin{equation}
    \textup{(i) } t \tan^{-1}\left(\sqrt{K} \norm{u} \right) = \tfrac{\pi}{2}+k\pi
      \Longleftrightarrow
    \textup{(ii) } t \theta = (2k+1)\pi, \quad k \in \bbZ.
    \end{equation}
    In these singular cases, \cref{eq:gyro_prod_calmk} is still valid, while the stereographic gyromultiplication returns infinity:
    \begin{equation} \label{eq:singular-case-radius}
    t \MKodot x = -\MKzero \in \calMK{n},
    \quad
    t \stodot u = \infty \in \stereo{n},
    \end{equation}
    where $\infty$ denotes the added point in the one-point compactification $\projhs{n}\cup\{\infty\}$ ($\stereo{n}=\projhs{n}$ for $K>0$) and $\isoMKST{n}(-\MKzero)=\infty$.
\end{parisproposition}

\begin{parisproposition}[Gyroaddition]
\label{prop:calmk_gyroaddition}
\linktoproof{prop:calmk_gyroaddition}
Let $x=[x_t,x_s^\top]^\top$ and $y=[y_t,y_s ^\top]^\top$ be points in $\calMK{n}$, where $x_t,y_t\in\bbRscalar$ are the time scalars, and $x_s,y_s\in\bbR{n}$ are the spatial parts. Then, the gyroaddition on $\calMK{n}$ admits the closed form:
\begin{equation} \label{eq:gyroadd_calmk}
    x\MKoplus y =
    \begin{cases}
    x, & y=\MKzero, \\
    y, & x=\MKzero, \\
    \begin{bmatrix}
    \frac{1}{\sqrt{|K|}} \frac{D - K N}{D + K N} \\
    \frac{2 \left( A_s x_s + A_y y_s \right)}{ D + K N }
    \end{bmatrix},  & \text{otherwise}.
    \end{cases}
\end{equation}
Here, $A_s = ab^2 - 2K b s_{xy} - K a n_y$ and $A_y = b(a^2 + K n_x)$, where the following quantities are defined by
\begin{equation}
a=1+\sqrt{|K|}x_t,
b=1+\sqrt{|K|} y_t,
n_x=\norm{x_s}^2,
n_y=\norm{y_s}^2,
s_{xy}=\langle x_s,y_s\rangle.
\end{equation}
\begin{equation}
D = a^2b^2 - 2K ab s_{xy} + K^2 n_x n_y,
N = a^2 n_y + 2ab s_{xy} + b^2 n_x .
\end{equation}
Besides, the following shows that the sphere gyroaddition is still valid under the singular cases in the projected hypersphere. Assume $K>0$ and $x,y \neq \pm \MKzero$, and let $u=\isoMKST{n}(x)$ and $v=\isoMKST{n}(y)$ be the stereographic images. The following statements are equivalent:
\par\leavevmode\vspace{-\baselineskip}
\begin{enumerate}
\item $u=\dfrac{v}{K\norm{v}^2}$ ($v\neq \zerovec$);
\item $x_s=y_s$ and $x_t=-y_t$ (same meridian, mirrored across the equator);
\item $D=0$.
\end{enumerate}
In such singular cases, we have $N>0$ and \cref{eq:gyroadd_calmk} is still valid, while the stereographic gyroaddition returns infinity:
\begin{equation}
    x \MKoplus y= -\MKzero,
    \quad
    u \stoplus v=\infty,
\end{equation}
where $\infty$ is the point added in the one-point compactification of $\projhs{n}$.
\end{parisproposition}

The above two propositions immediately imply that $\isoMKST{n}$ preserves the gyro operations.
\begin{pariscorollary}[Isomorphism]
    \label{cor:isomorphism_calmk_stereo}
    \linktoproof{cor:isomorphism_calmk_stereo}
    For the hyperbolic geometry ($K<0$), the isometry $\LtoPball: \lorentz{n} \to \pball{n}$ preserves the gyro operations:
    \begin{equation}
        \begin{aligned}
            x \Loplus y &= \PballtoL \left( \LtoPball(x) \Moplus \LtoPball(y) \right), \quad \forall x, y \in \lorentz{n},\\
            r \Lodot x &= \PballtoL \left( r \Modot \LtoPball(x) \right), \quad \forall r \in \bbRscalar, \forall x \in \lorentz{n}.
        \end{aligned}
    \end{equation}
    For the spherical geometry ($K>0$), the isometry $\isoMKST{n}: \sphere{n} \to \projhs{n} \cup \{\infty\}$ preserves the gyro operations:
    \begin{equation}
        \begin{aligned}
            x \MKoplus y &= \isoSTMK{n} \left( \isoMKST{n}(x) \stoplus \isoMKST{n}(y) \right), \quad \forall x, y \in \calMK{n} / \{ -\MKzero \},\\
            r \MKodot x &= \isoSTMK{n} \left( r \stodot \isoMKST{n}(x) \right), \quad \forall r \in \bbRscalar, \forall x \in \calMK{n} / \{ -\MKzero \}.
        \end{aligned}
    \end{equation}
\end{pariscorollary}

\begin{parisremark}
    We provide two clarifications regarding \cref{cor:isomorphism_calmk_stereo}.
    \begin{itemize}
        \item
        \mypara{Compactified projected hypersphere.} Although stereographic operations for $K>0$ may be undefined in certain cases, they become well-defined on the one-point compactification $\projhs{n}\cup\{\infty\}$, where undefined cases correspond to $\infty$.

        \item
        \mypara{Sphere vs.\ projected hypersphere.} \cref{cor:isomorphism_calmk_stereo} suggests a numerical advantage of the sphere: gyro-operations are well-defined on $\sphere{n}$ at all points except the single south pole $-\MKzero$, including cases corresponding to singularities on the projected hypersphere. This broader domain can make computations on $\sphere{n}$ more stable.
    \end{itemize}
\end{parisremark}

From the above corollary, it is expected that the operations $\MKoplus$ and $\MKodot$ also satisfy the axioms of a gyrovector space for both negative and positive curvature $K$.
\begin{paristheorem}[Radius gyrovector spaces]
\label{thm:calmk_gyrovector}
\linktoproof{thm:calmk_gyrovector}
$(\calMK{n},\MKoplus)$ forms a gyrocommutative gyrogroup, and $(\calMK{n},\MKoplus,\MKodot)$ forms a gyrovector space.\footnote{For $K>0$, we implicitly assume the addition and multiplication are well-defined; whenever they are, all corresponding axioms hold.}
\end{paristheorem}

Due to the isometry, \cref{lem:isometry_for_gyro} implies gyroisometries over $\calMK{n}$.
\begin{paristheorem} \label{thm:radius_iso}
On the radius model, the gyrodistance is identical to the geodesic distance, whereas the gyroinverse, gyration, and left gyrotranslation are gyroisometries.
\end{paristheorem}

\mypara{Radius GyroBN.} \cref{thm:radius_iso} guarantees that GyroBN on the radius model normalizes the sample mean and variance. Substituting the radius Riemannian operators from \cref{tab:ch2-constant-curvature-operators} and the gyro operators from \cref{prop:calmk_scalar_prod_inv,prop:calmk_gyroaddition} into \cref{alg:gyrobn} can directly yield the radius GyroBN. For $K<0$ (Lorentz), the Fréchet mean can be computed efficiently by \citet[Alg.~3]{lou2020differentiating}; for $K>0$ (sphere), we compute the Fréchet mean using the Karcher flow \citep{karcher1977riemannian}.

\subsubsection{Hyperbolic Beltrami--Klein}
\label{subsec:klein_einstein}
The Beltrami--Klein model has recently emerged as a promising alternative to the Poincaré ball for representing hyperbolic geometry \citep{mao2024klein}. As reviewed in \cref{sec:ch2-constant-curvature-manifolds}, the Poincaré ball admits the Möbius gyrovector space and the Beltrami--Klein model admits the Einstein gyrovector space. While prior studies mainly focused on the case $K=-1$ \citep{mao2024klein}, we develop the Beltrami--Klein Riemannian structure under arbitrary negative curvature and relate it to the Einstein gyrospace. This allows us to establish the equivalence between the gyro and Riemannian formulations and, ultimately, to instantiate GyroBN on this model.

\mypara{Beltrami--Klein Riemannian structure.} The standard Einstein gyro operations on the Beltrami--Klein model are reviewed in \cref{tab:ch2-constant-curvature-gyro-operators}. For $K=-1$, prior work related the Einstein operations to the Riemannian-form gyro operations in \cref{eq:ch2-riem-gyro-addition,eq:ch2-riem-gyro-scalar} \citep[Sec.~4.2]{mao2024klein}. We extend this equivalence to arbitrary $K<0$ and derive the corresponding closed-form Beltrami--Klein Riemannian operators. To this end, we first establish the isometry between the Beltrami--Klein and Poincaré models.

\begin{parisproposition}[Beltrami--Klein isometries]
    \label{props:klein_poincare_isometry_differentials}
    \linktoproof{props:klein_poincare_isometry_differentials}
    The following maps are Riemannian isometries between the Beltrami--Klein and Poincaré ball models:
    \begin{align}
        \pi_{\klein{n} \to \pball{n}}
        &: \klein{n} \ni x  \longmapsto
        \frac{1}{ 1 +  \sqrt{1 + K \norm{x}^2} } x \in \pball{n},\\
        \pi_{\pball{n} \to \klein{n}}
        &: \pball{n} \ni x \longmapsto \frac{2}{1 - K \norm{x}^2} x \in \klein{n}.
    \end{align}
    In particular, $\pi_{\pball{n} \to \klein{n}}(\zerovec)=\zerovec$. Given $x$ in the hyperbolic model $\calH \in \{ \klein{n}, \pball{n}\}$ and tangent vector $v \in T _x \calH$, the differential maps of $\pi_{\klein{n} \to \pball{n}}$ and $\pi_{\pball{n} \to \klein{n}}$ are
    \begin{align*}
    (\pi_{\klein{n} \to \pball{n}})_{*,x} (v)
    &= \frac{1}{1 + \sqrt{1+K\norm{x}^2}} v
    - \frac{K \inner{x}{v}}{\left(1 + \sqrt{1+K\norm{x}^2}\right)^2 \sqrt{1+K\norm{x}^2}} x, \\
    (\pi_{\pball{n} \to \klein{n}})_{*,x} (v)
    &= \frac{2}{\left(1-K\norm{x}^2\right)}v
    + \frac{4 K \inner{x}{v} }{\left(1-K\norm{x}^2\right)^2 }x.
    \end{align*}
    In particular, the differential maps at the zero vector are
    \begin{align}
        \label{app:eq:k2p_diff_0}
        (\pi_{\klein{n} \to \pball{n}})_{*,{\zerovec}} (v)
        &= \frac{1}{2} v, \\
        \label{app:eq:p2k_diff_0}
        (\pi_{\pball{n} \to \klein{n}})_{*,{\zerovec}} (v)
        &= 2 v.
    \end{align}
    Moreover, these isometries preserve gyroaddition and gyromultiplication:
    \begin{equation}
        \label{eq:iso_addition}
        \begin{aligned}
        \pi_{\pball{n} \to \klein{n}}(x \Moplus y) &= \pi_{\pball{n} \to \klein{n}}(x) \Eoplus \pi_{\pball{n} \to \klein{n}}(y), \quad \forall x,y \in \pball{n}, \\
        \pi_{\pball{n} \to \klein{n}} (t \Modot x) &= t \Eodot \pi_{\pball{n} \to \klein{n}}(x), \quad \forall t \in \bbRscalar, \forall x \in \pball{n},
    \end{aligned}
    \end{equation}
    where $\Moplus$ and $\Modot$ are Möbius operations, while $\Eoplus$ and $\Eodot$ are the Einstein counterparts.
\end{parisproposition}

\begin{paristheorem}[Einstein by Beltrami--Klein]
    \label{thm:einstein_klein}
    \linktoproof{thm:einstein_klein}
    The Einstein gyro operations can be rewritten as \cref{eq:ch2-riem-gyro-addition,eq:ch2-riem-gyro-scalar}:
    \begin{align}
        x \Eoplus y &= \rieexp _x \left(\pt{{\zerovec}}{x} (\rielog _{\zerovec} (y)) \right), \quad \forall x,y \in \klein{n}, \\
        t \Eodot x &= \rieexp _{\zerovec} \left( t \rielog _{\zerovec} (x)\right), \quad \forall x \in \klein{n}, \forall t \in \bbRscalar.
    \end{align}
\end{paristheorem}

\cref{thm:einstein_klein} demonstrates that the Einstein gyro-structure can be expressed by the Beltrami--Klein geometry. Conversely, the Beltrami--Klein geometry can also be formulated by the Einstein gyro-structure.

\begin{paristheorem} [Beltrami--Klein by Einstein]
    \label{thm:riem_klein}
    \linktoproof{thm:riem_klein}
    Given $x, y \in \klein{n}$ and $v \in T _x \klein{n}$, the distance, exponential, and logarithmic operators under the Beltrami--Klein geometry are
    \begin{align*}
        \dist (x, y)
        & =\frac{2}{\sqrt{|K|}} \tanh ^{-1}\left(\sqrt{|K|} \frac{ \norm{-x \Eoplus y}}{1+\sqrt{1+K\left\|-x \Eoplus y\right\|^2}} \right), \\
        \rieexp _{x}(v)
        &= x \Eoplus \rieexp _{\zerovec}\left( \frac{1}{\sqrt{1+K\norm{x}^2}} v
            - \frac{K \inner{x}{v}}{\left(1 + \sqrt{1+K\norm{x}^2}\right) (1+K\norm{x}^2)} x  \right), \\
        \rielog _x (y)
        &= \frac{1}{\lambda_{\widetilde{x}}^K} (\pi_{\pball{n} \to \klein{n}}) _{*, \widetilde{x}}\left(\rielog _{\zerovec}\left( -x \Eoplus  y\right)\right),
    \end{align*}
    where $\widetilde{x} = \pi_{\klein{n} \to \pball{n}}(x)$. In particular, the exponential and logarithmic maps at the zero vector $\zerovec$ are identical across the Beltrami--Klein and Poincaré ball models:
    \begin{align}
        \label{eq:exp_0_klein}
        \rieexp_{\zerovec}(v) &= \tanh (\sqrt{|K|}\norm{v}) \frac{v}{\sqrt{|K|}\norm{v}}, \quad \forall v \in T_\zerovec \calH,\\
        \label{eq:log_0_klein}
        \rielog_{\zerovec}(x) &= \tanh ^{-1}(\sqrt{|K|}\norm{x}) \frac{x}{\sqrt{|K|}\norm{x}}, \quad \forall x \in \calH,
    \end{align}
    with $\calH \in \{ \klein{n}, \pball{n} \}$.
\end{paristheorem}
\begin{parisremark}
    Since the Beltrami--Klein and Poincaré ball models share the same $\rieexp_{\zerovec}$ and $\rielog_{\zerovec}$, it naturally follows that the Einstein and Möbius gyromultiplication coincide.
\end{parisremark}

As the Beltrami--Klein model is isometric to the Poincaré ball, \cref{lem:isometry_for_gyro} implies gyroisometries.
\begin{paristheorem} \label{thm:klein_iso}
On the Beltrami--Klein model, the gyrodistance is identical to the geodesic distance, whereas the gyroinverse, gyration, and left gyrotranslation are gyroisometries.
\end{paristheorem}

\mypara{Beltrami--Klein GyroBN.} \cref{thm:klein_iso} ensures that GyroBN on the Beltrami--Klein model normalizes the sample mean and variance. Owing to the isometry between the Beltrami--Klein and Poincaré models, the Fréchet mean can be computed via the Poincaré ball: map the data to the Poincaré model using $\pi_{\klein{n} \to \pball{n}}$, compute the Poincaré Fréchet mean \citep[Alg.~1]{lou2020differentiating}, and map the result back using $\pi_{\pball{n} \to \klein{n}}$. Together with the gyro and Riemannian operators in \cref{subsec:klein_einstein}, we have all the ingredients to implement \cref{alg:gyrobn}.

\subsubsection{Correlation Manifolds}
\label{subsec:cor_manifolds}

As reviewed in \cref{sec:ch2-full-rank-correlation-manifolds}, ECM, LECM, OLM, and LSM are correlation metrics induced from Euclidean, zero-curvature prototype spaces, and these four metrics have been instantiated for LieBN in \cref{subsec:liebn_cor}. Here, we further handle the nonzero-curvature correlation metric, namely PHCM. Under PHCM, any correlation matrix can be identified with a product of hyperbolic spaces via its Cholesky decomposition. Given $C \in \cor{n}$, let $L=\chol(C)$ be its Cholesky factor. The $k$-th row of $L$ has the form $(L_{k1},\ldots,L_{k,k-1},L_{kk},0,\ldots,0)$ with $L_{kk}>0$, which belongs to the hyperbolic open hemisphere\footnote{Also known as the Jemisphere model, where the ``J'' is pronounced as in Spanish \citep[Sec.~7]{cannon1997hyperbolic}.}
\begin{equation}
    \hs{k-1}=\left\{x \in \bbR{k} \mid \norm{x}=1, x_k>0 \right\}.
\end{equation}
As detailed in \cref{sec:ch5-cornet}, $\hs{n}$ is isometric to the unit Poincaré ball
\begin{equation}
    \unitpball{n}=\left\{x \in \bbR{n} \mid \norm{x} < 1 \right\}
\end{equation}
by
\begin{equation}
    \pi _{\hs{n} \rightarrow \unitpball{n}} \left(
    \begin{bmatrix}
        x \\
        x_{n+1}
    \end{bmatrix} \right)
    = \frac{x}{1+x_{n+1}}.
\end{equation}
Therefore, each correlation matrix can be identified with $n-1$ Poincaré vectors:
\begin{equation} \label{eq:iso_cor_phc}
    \cor{n} \ni C {\mapsto}
    \begin{bmatrix}
    1 & 0 & \cdots & 0 \\
    L_{21} & L_{22} & \cdots & 0 \\
    \vdots & \vdots & \ddots & \vdots \\
    L_{n1} & L_{n2} & \cdots & L_{nn}
    \end{bmatrix}
    {\mapsto}
    \begin{bmatrix}
    x_1 \in \unitpball{1} \\
    \vdots \\
    x_{n-1} \in \unitpball{n-1}
    \end{bmatrix}.
\end{equation}
Here, $x_{i} = \pi _{\hs{i} \rightarrow \unitpball{i}} \left(L_{(i+1, 1)}, \cdots, L_{(i+1, i+1)} \right)^\top$ corresponds to the $(i+1)$-th row of the Cholesky factor. Let $\bbPPB{n-1} = \prod_{i=1}^{n-1}\unitpball{i}$ denote the product of unit Poincaré balls. We denote the identification in \cref{eq:iso_cor_phc} by $\Phi: \cor{n} \to \bbPPB{n-1}$. GyroBN on the correlation manifold can then be realized via the Poincaré GyroBN applied row-wise: first map $C$ to $\bbPPB{n-1}$ via $\Phi$, apply $\text{GyroBN}_i$ independently on each $\unitpball{i}$, and finally map back with $\Phi^{-1}$. For a batch of activations $\{C^i\}_{i=1}^N \subset \cor{n}$, the process can be expressed as
\begin{equation}
    \forall i \leq N, \quad C^i \overset{\Phi}{\longmapsto}
    \begin{bmatrix}
    x^i_1 \in \unitpball{1} \\
    \vdots \\
    x^i_{n-1} \in \unitpball{n-1}
    \end{bmatrix}
    \begin{array}{c}
         \overset{\text{GyroBN}_1}{\longmapsto}  \\
         \vdots \\
         \overset{\text{GyroBN}_{n-1}}{\longmapsto}  \\
    \end{array}
    \begin{bmatrix}
    \widetilde{x}^i_1 \in \unitpball{1} \\
    \vdots \\
    \widetilde{x}^i_{n-1} \in \unitpball{n-1}
    \end{bmatrix}
    \overset{\Phi^{-1}}{\longmapsto}
    \widetilde{C}^i.
\end{equation}

\subsubsection{Summary}

\begin{table}[t]
\centering
\renewcommand{\arraystretch}{1.2}
\begin{subtable}[t]{\linewidth}
    \centering
    \begin{tabular}{cccc}
    \toprule
    $U \oplusGyrONB V$ & $\ominusGyrONB U$ & $t \odotGyrONB U$ & \textbf{Fr\'echet mean} \\
    \midrule
     $\mexp(\Omega_U) V$
    & $\mexp\left(-\Omega_U\right) \idonb$
    & $ \mexp\left(t\Omega_U\right) \idonb$
    & Karcher flow \\
    \bottomrule
    \end{tabular}
    \caption{The ONB Grassmannian}
    \end{subtable}

    \begin{subtable}[t]{\linewidth}
    \centering
    \resizebox{\linewidth}{!}{
    \begin{tabular}{cccc}
    \toprule
    \textbf{Operator} & $\stereo{n}$ & $\calMK{n}$ & $\klein{n}$ \\
    \midrule
    $x \oplus y$
    & $\displaystyle \frac{(1-2K\langle x,y\rangle - K\|y\|^{2})x + (1+K\|x\|^{2})y}{1 - 2K\langle x,y\rangle + K^{2}\|x\|^{2}\|y\|^{2}}$
    & \cref{eq:gyroadd_calmk}
    & \cref{tab:ch2-constant-curvature-gyro-operators} \\
    $\ominus x$ & $-x$
    & $\displaystyle \begin{bmatrix}
    x_t \\
    -x_s
    \end{bmatrix}$ & $-x$ \\
    $t \odot x$
    & $\displaystyle \frac{\tank \left(t \tank ^{-1}\left( \sqrt{|K|} \|x\| \right)\right)}{\sqrt{|K|}}  \frac{x}{\|x\|}$
    & \cref{eq:gyro_prod_calmk}
    & \cref{tab:ch2-constant-curvature-gyro-operators}\\
    \textbf{Fréchet mean}
    & \makecell{$K<0$: \citep[Alg.~1]{lou2020differentiating} \\ $K>0$: Karcher flow}
    & \makecell{$K<0$: \citep[Alg.~3]{lou2020differentiating} \\ $K>0$: Karcher flow}
    & \makecell{Via Poincaré ball \\ (see \cref{subsec:klein_einstein})} \\
    \bottomrule
    \end{tabular}
    }
    \caption{Constant curvature spaces}
    \end{subtable}
    \caption{Summary of operators for GyroBN across representative manifolds.}
\label{tab:gyrobn-summary}
\end{table}

To conclude this section, \cref{tab:gyrobn-summary} summarizes the key gyro operators needed to implement GyroBN on representative manifolds. The correlation manifold is excluded, since its GyroBN is realized row-wise through the Poincaré ball.

\subsection{Experiments}
\label{sec:ch3-gyrobn-experiments}
GyroBN layers are backbone-agnostic and can be integrated into different networks whenever the underlying manifold admits the required gyro operations. This subsection evaluates GyroBN on the Grassmannian, five CCSs, and the correlation manifold. The main findings are summarized as follows.
More details on data sets and experimental settings are provided in \cref{app:datasets,app:gyrobn-experimental-details}.
\begin{keybox}
\begin{itemize}
    \item
    \mypara{Numerical experiments (\cref{subsec:ch3-gyrobn-experiments:numerical}).} The closed-form expression derived for radius gyroaddition in \cref{eq:gyroadd_calmk} significantly accelerates the computation compared with its definition-based counterpart in \cref{eq:calmk_gyroadd_def}, achieving $2{\times}$--$3{\times}$ speedups. Furthermore, visualizations demonstrate that GyroBN effectively normalizes sample distributions across diverse geometries.
    \item
    \mypara{Performance (\cref{subsec:ch3-gyrobn-experiments:grass_nn,subsec:ch3-gyrobn-experiments:ccs_nn,subsec:ch3-gyrobn-experiments:cor_nn}).} On networks over the Grassmannian, CCSs, and the correlation manifold, GyroBN generally improves backbone networks, whereas existing RBN methods often degrade performance. Compared to these methods, GyroBN is generally faster or comparable in runtime, requires fewer or equal parameters, and improves robustness.
\end{itemize}
\end{keybox}

\subsubsection{Numerical Experiments}
\label{subsec:ch3-gyrobn-experiments:numerical}

\begin{table}[t]
  \centering
  \resizebox{0.7\linewidth}{!}{
    \begin{tabular}{c|cc|cc}
    \toprule
    \textbf{Geometry} & \multicolumn{2}{c|}{\textbf{Lorentz}} & \multicolumn{2}{c}{\textbf{Sphere}} \\
    \midrule
    \textbf{Dim} & \textbf{Riemannian} & \cellcolor{HilightColor}\textbf{Closed form} & \textbf{Riemannian} & \cellcolor{HilightColor}\textbf{Closed form} \\
    \midrule
    16    & 361.22 & \cellcolor{HilightColor}\firstresults{121.68 (33.69\%)} & 323.58 & \cellcolor{HilightColor}\firstresults{122.02 (37.71\%)} \\
    32    & 363.41 & \cellcolor{HilightColor}\firstresults{123.08 (33.87\%)} & 327.75 & \cellcolor{HilightColor}\firstresults{123.71 (37.74\%)} \\
    64    & 387.94 & \cellcolor{HilightColor}\firstresults{181.50 (46.78\%)} & 453.42 & \cellcolor{HilightColor}\firstresults{181.05 (39.93\%)} \\
    128   & 574.37 & \cellcolor{HilightColor}\firstresults{272.10 (47.37\%)} & 644.18 & \cellcolor{HilightColor}\firstresults{270.21 (41.95\%)} \\
    256   & 1149.58 & \cellcolor{HilightColor}\firstresults{534.09 (46.46\%)} & 1137.62 & \cellcolor{HilightColor}\firstresults{538.58 (47.34\%)} \\
    1024  & 3364.23 & \cellcolor{HilightColor}\firstresults{1414.67 (42.05\%)} & 3551.69 & \cellcolor{HilightColor}\firstresults{1421.67 (40.03\%)} \\
    2048  & 6479.95 & \cellcolor{HilightColor}\firstresults{2497.47 (38.54\%)} & 6930.69 & \cellcolor{HilightColor}\firstresults{2449.56 (35.34\%)} \\
    \bottomrule
    \end{tabular}%
    }
    \caption{Efficiency (in~$\mu$s) of gyroaddition on the radius manifold: closed form versus Riemannian definition. Values in parentheses indicate the runtime of the closed-form implementation as a percentage of the corresponding Riemannian implementation. The best results are \firstresults{bold}.}
  \label{tab:res:efficiency_gyroadd}%
\end{table}%

\mypara{Efficiency of the closed-form radius gyroaddition.} Recalling \cref{subsec:radius-model}, we derive closed-form expressions for the radius gyrooperations to improve computational efficiency. To assess this, we compare two variants of radius gyroaddition: (i) the definition-based operator \cref{eq:calmk_gyroadd_def}, implemented via a composition of the Riemannian logarithm, parallel transport, and exponential map; and (ii) the closed-form operator \cref{eq:gyroadd_calmk}. We report the mean wall-clock time (in~$\mu$s), averaged over 100 runs with a batch size of 10,000, across varying dimensions. As shown in \cref{tab:res:efficiency_gyroadd}, the closed-form implementation consistently outperforms its definition-based counterpart, achieving speedups of roughly $2{\times}$--$3{\times}$ across Lorentz and spherical geometries.

\begin{figure}[t]
\centering
\includegraphics[width=0.8\linewidth,trim={0cm 0cm 0cm 1cm}]{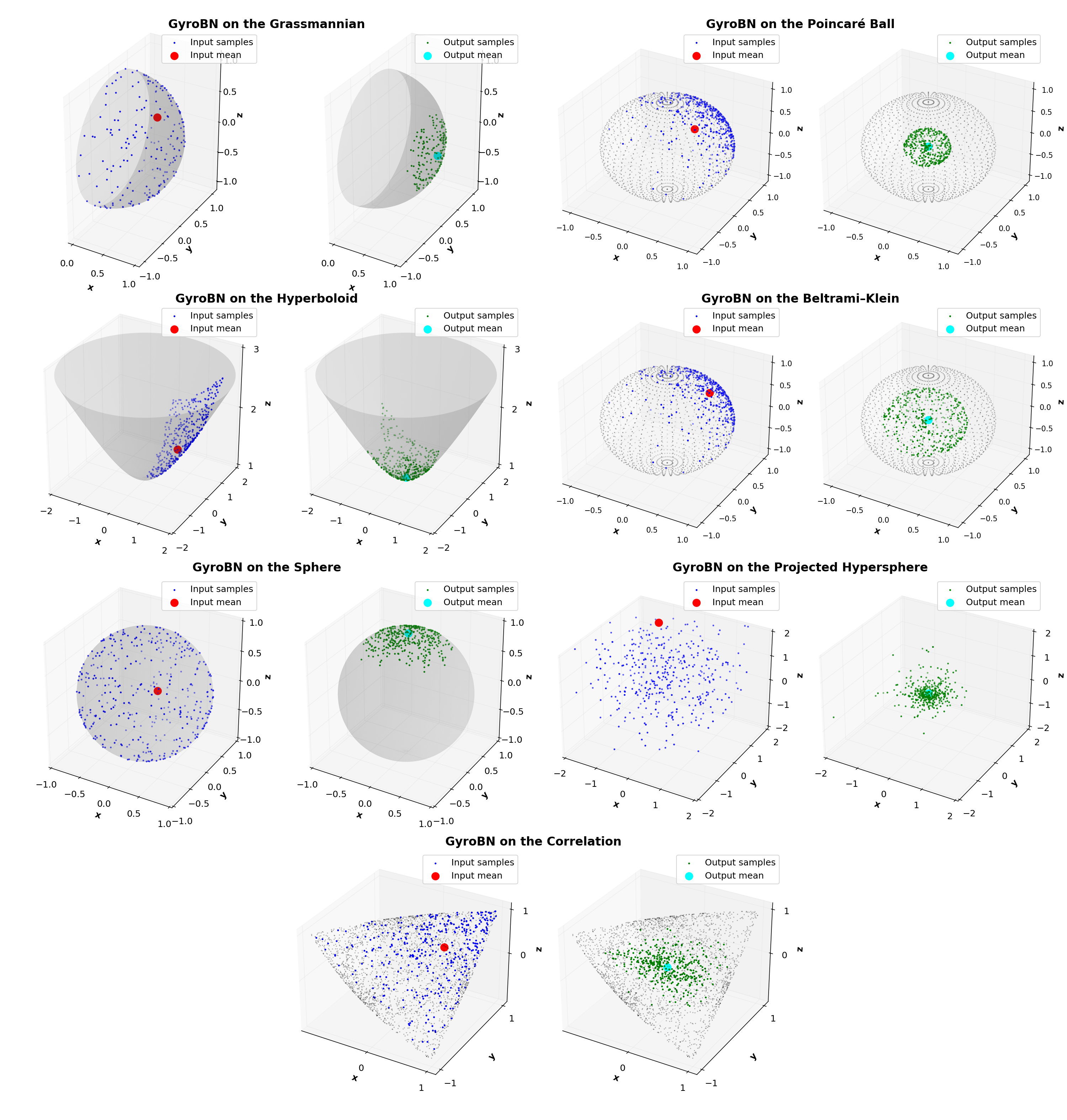}
\caption{Visualization of GyroBN across different geometries. Blue and green points represent input and normalized data, respectively. Red and cyan points denote the input and output batch means. Black points mark the manifold boundary, and the gray surface depicts the manifold.}
\label{fig:illustration_gyrobn_grass_ccs_cor}
\end{figure}

\mypara{Visualization of GyroBN on different geometries.} To intuitively illustrate the effect of GyroBN, we visualize its behavior on the ONB Grassmannian $\grasonb{1,3}$, five CCSs with $|K|=1$, and the correlation manifold $\cor{3}$. For each geometry, we randomly generate a batch of points, compute their batch mean, apply GyroBN, and then plot the normalized batch along with the resulting mean. The visualizations are constructed using the following embeddings:
\begin{itemize}
    \item
    \mypara{Grassmannian.} Since $\grasonb{1,3}$ is homeomorphic to the real projective space $\rp{2}$, it is depicted as the unit hemisphere with antipodal points identified.
    \item
    \mypara{CCSs.} The unit Poincaré ball $\unitpball{3}$ and unit Beltrami--Klein ball $\unitklein{3}$ are shown as the interior of the unit ball in $\bbR{3}$. The unit Lorentz space $\unitlorentz{2}$ is visualized as the upper sheet of its defining two-sheeted quadric in $\bbR{3}$. The unit sphere $\unitsphere{2}$ is embedded as a $2$-sphere in $\bbR{3}$, while the projected hypersphere $\unitprojhs{3}$ coincides with $\bbR{3}$ itself.
    \item
    \mypara{Correlation manifold.} $\cor{3}$ is embedded in $\bbR{3}$ as an open elliptope using its strictly lower triangular part.
\end{itemize}

For better visualization, we fix the bias parameter to the gyro identity element and set the scaling parameter to $0.4$ for the Grassmannian, correlation manifold, and sphere; $0.7$ for the Poincaré, Beltrami--Klein, and Lorentz models; and $1$ for the projected hypersphere. As shown in \cref{fig:illustration_gyrobn_grass_ccs_cor}, GyroBN consistently normalizes data distributions across these geometries. Notably, although the Poincaré and Beltrami--Klein inputs are identical, their GyroBN behavior and resulting sample distributions differ due to their distinct Riemannian metrics, underscoring that GyroBN faithfully respects the underlying geometry.

\subsubsection{Experiments on Grassmannian Neural Networks}
\label{subsec:ch3-gyrobn-experiments:grass_nn}

\mypara{Data sets and preprocessing.} In line with previous work \citep{huang2018building,nguyen2023building}, we evaluate GyroBN on three skeleton-based action recognition tasks, including HDM05 \citep{muller2007documentation}, NTU60 \citep{shahroudy2016ntu}, and NTU120 \citep{liu2019ntu} data sets, focusing on mutual actions for NTU60 and NTU120. Each sequence is represented as a Grassmannian matrix of size $93 \times 10$, $150 \times 10$, and $150 \times 10$ for HDM05, NTU60, and NTU120, respectively.

\mypara{Comparative methods.} Since no Grassmannian-specific BN methods exist, we adapt two previous approaches to the Grassmannian: ManifoldNorm~\citep[Algs.~1--2]{chakraborty2020manifoldnorm} and the RBN method of \citet[Alg.~2]{lou2020differentiating}, which we denote LRBN for clarity. Although neither was originally designed for the Grassmannian, they can be adapted by employing Riemannian operators such as geodesics, exponential/logarithmic maps, and parallel transport. \textit{The key difference is that GyroBN can normalize data distributions across different geometries, whereas the other two methods cannot.}

\mypara{Backbone networks.} We adopt the recently proposed GyroGr architecture \citep{nguyen2023building} as the backbone, which is briefly reviewed in \cref{app:backbone-grassmannian}. GyroGr replaces the non-intrinsic FRMap + ReOrth block in GrNet \citep{huang2018building} with Grassmannian left gyrotranslation, thereby improving numerical stability and performance. It consists of three basic components: left gyrotranslation, pooling \citep{huang2018building}, and the Projection Map (ProjMap) \citep{huang2018building}, where ProjMap maps Grassmannian points to symmetric matrices for classification. We consider both the 1-block and $L$-block variants. The 1-block version is structured as: gyrotranslation $\rightarrow$ pooling $\rightarrow$ ProjMap $\rightarrow$ classification, where the classification is implemented as an FC layer with softmax. The $L$-block version stacks $L$ blocks of gyrotranslation and pooling, followed by a final ProjMap and classification layer. Since each pooling step approximately halves the dimensionality, we omit the pooling operation in the last block when $L > 1$. Following \citet{huang2018building}, the number of channels is fixed to $8$.

\mypara{Implementation details.} Following \citet{nguyen2023building}, we use the Cayley map to approximate the matrix exponential of skew-symmetric matrices and apply the trivialization strategy reviewed in \cref{sec:ch2-riemannian-optimization} to parameterize the Grassmannian variables in both the gyrotranslation and GyroBN layers. Specifically, each trainable Grassmannian parameter is represented by Euclidean coordinates through the exponential map at the identity. This allows direct use of PyTorch optimizers \citep{paszke2019pytorch} and avoids direct Riemannian updates of these parameters. In contrast, we find that the Grassmannian LRBN benefits from Riemannian optimization. Thus, we employ \texttt{Geoopt} \citep{kochurov2020geoopt} to optimize its Grassmannian bias parameter. Similarly, we use \texttt{Geoopt} to update the orthogonal bias parameter in ManifoldNorm. For all models, the BN layer is inserted after the first pooling layer with a momentum of $0.1$. Training uses SGD with a learning rate of $5e^{-2}$, batch size $30$, and $400$, $200$, and $200$ epochs for HDM05, NTU60, and NTU120, respectively. All models are optimized with a standard cross-entropy loss. Following previous normalization methods on matrix manifolds \citep{kobler2022spd,wang2025gbwmbn} and the LieBN implementation in \cref{app:liebn-experimental-details}, we adopt a single Fréchet mean iteration.

\begin{table}[t]
  \centering
  \resizebox{0.99\linewidth}{!}{
    \begin{tabular}{c|ccc|ccc|ccc}
    \toprule
    \multirow{2}[4]{*}{\textbf{Method}} & \multicolumn{3}{c|}{\makecell{\textbf{HDM05} \\ (47 \ensuremath{\times} 10)}} & \multicolumn{3}{c|}{\makecell{\textbf{NTU60} \\ (75 \ensuremath{\times} 10)}} & \multicolumn{3}{c}{\makecell{\textbf{NTU120} \\ (75 \ensuremath{\times} 10)}} \\
    \cmidrule{2-10}          & \textbf{Acc}   & \textbf{Fit Time} &  \textbf{\#Params (M)} & \textbf{Acc}   & \textbf{Fit Time} &  \textbf{\#Params (M)} & \textbf{Acc}   & \textbf{Fit Time} &  \textbf{\#Params (M)} \\
    \midrule
    GyroGr & 48.97\ensuremath{\pm}0.24 & 2.09  & 2.0744 & 70.13\ensuremath{\pm}0.16 & 28.16  & 0.5062 & 53.76\ensuremath{\pm}0.18 & 49.62  & 1.1812 \\
    GyroGr-ManifoldNorm & 49.67\ensuremath{\pm}0.76 & 32.90  & \red{2.0921} & 68.56\ensuremath{\pm}0.43 & 232.60  & \red{0.5512} & 51.41\ensuremath{\pm}0.38 & 399.78  & \red{1.2262} \\
    GyroGr-LRBN & 48.64\ensuremath{\pm}0.77 & 33.31  & 2.0781 & 67.77\ensuremath{\pm}0.52 & 238.53  & 0.5122 & 50.56\ensuremath{\pm}0.22 & 403.40  & 1.1872 \\
    \midrule
    \rowcolor{HilightColor} GyroGr-GyroBN & \firstresults{51.89\ensuremath{\pm}0.37} & 3.04  & 2.0773 & \firstresults{72.60\ensuremath{\pm}0.04} & 35.85  & 0.5114 & \firstresults{55.47\ensuremath{\pm}0.10} & 67.37  & 1.1864 \\
    \bottomrule
    \end{tabular}%
    }
    \caption{Comparison of GyroBN against other Grassmannian BN methods under the GyroGr backbone. Here, accuracy is reported as a percentage, fit time denotes the average training time per epoch (s/epoch), and \#Params is reported in millions. Values in parentheses specify the dimension of the Grassmannian input to the BN layer. The largest number of parameters is marked in \red{red}.}
  \label{tab:grass_results}
\end{table}

\begin{figure}[t]
\centering
\includegraphics[width=0.9\linewidth,trim={0cm 0cm 0cm 0cm}]{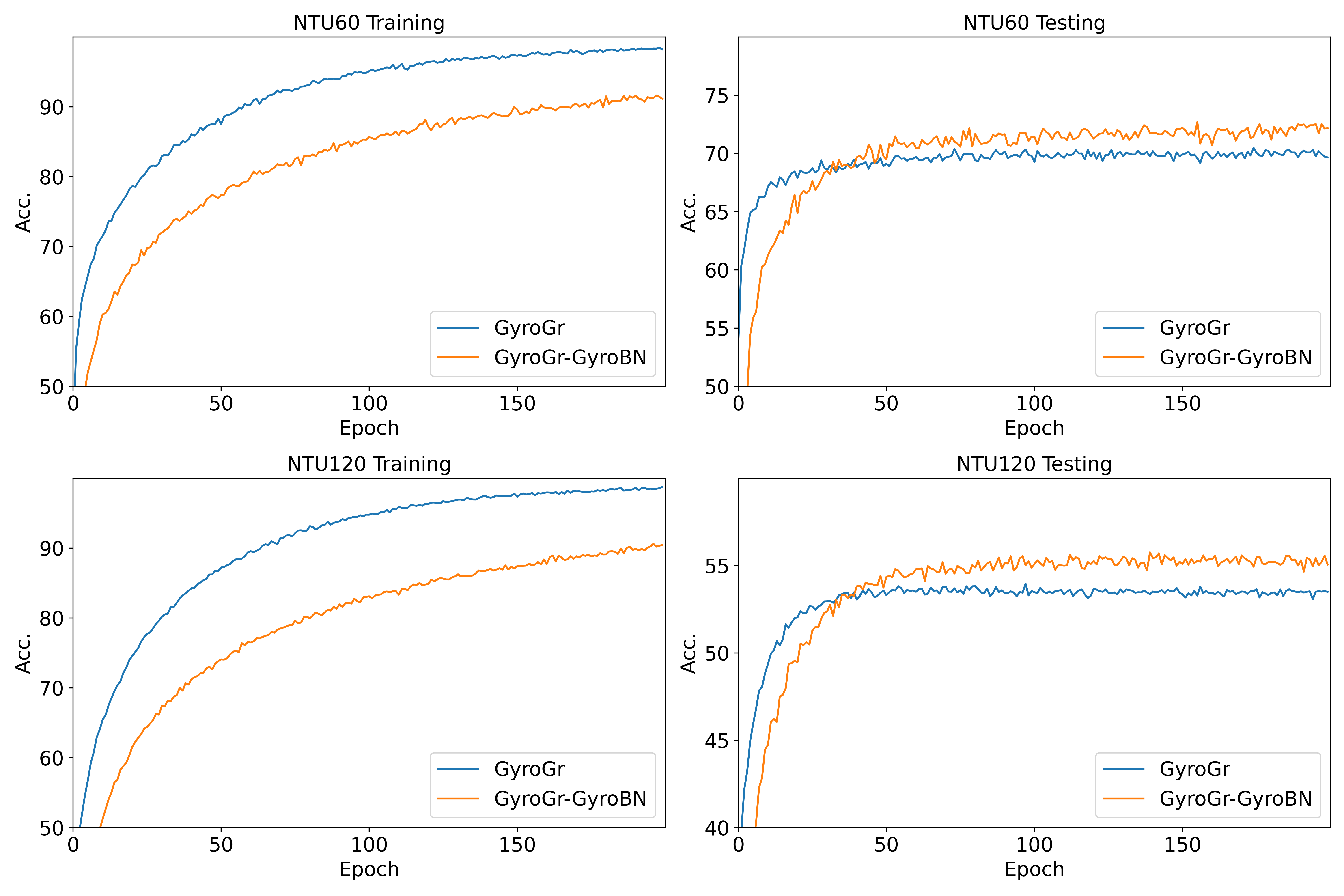}
\caption{Training and testing curves of 1-block GyroGr on two NTU data sets.
}
\label{fig:gyrobn_generalization}
\end{figure}

\mypara{Main results.} We compare GyroBN with ManifoldNorm and LRBN under the 1-block GyroGr backbone. The 5-fold results are presented in \cref{tab:grass_results}. We have the following four findings, which highlight the effectiveness of GyroBN in facilitating network training.
\begin{itemize}
    \item
    \mypara{Improved accuracy.} Across all three data sets, GyroBN consistently improves performance, enhancing the accuracy of the vanilla GyroGr by 2.92, 2.47, and 1.71 percentage points on HDM05, NTU60, and NTU120, respectively. In contrast, both ManifoldNorm and LRBN often degrade performance, particularly on NTU60 and NTU120. This advantage comes from the theoretical guarantee of GyroBN for normalizing sample statistics, which is absent in the other two methods (see \cref{tab:rbn_summary}).

    \item
    \mypara{Enhanced efficiency.} GyroBN is substantially more efficient than ManifoldNorm and LRBN.
    The efficiency gain is mainly attributed to: (i) replacing computationally expensive Riemannian operators (e.g., parallel transport, exponential/logarithmic maps) with simpler gyro operations,
    (ii) reducing matrix multiplications from $n \times p$ to $(n-p)\times p$ or $p \times p$ (see \cref{prop:fast_bracket_grassmannian}),
    and (iii) applying trivialization to avoid costly Riemannian optimization.

    \item
    \mypara{Improved parameter economy.} GyroBN also requires fewer parameters than LRBN and ManifoldNorm. The key difference lies in the bias parameter. Due to the trivialization, GyroBN only needs an $(n-p)\times p$ Euclidean matrix, whereas LRBN requires an $n\times p$ Grassmannian matrix
    and ManifoldNorm requires an $n\times n$ orthogonal matrix.

    \item
    \mypara{Stronger generalization.} As shown in \cref{fig:gyrobn_generalization}, we observe that GyroBN can narrow the gap between training and testing accuracy, indicating a stronger generalization ability.
\end{itemize}

\subsubsection{Experiments on Networks over CCSs}
\label{subsec:ch3-gyrobn-experiments:ccs_nn}

\mypara{Data sets.} Following \citet{lou2020differentiating}, we focus on the link prediction task on four graph data sets: Cora \citep{sen2008collective}, Disease \citep{anderson1991infectious}, Airport \citep{zhang2018link}, and PubMed \citep{namata2012query}.

\mypara{Comparative methods.} We compare GyroBN with LRBN \citep[Alg.~2]{lou2020differentiating}. As the original LRBN is only implemented in the Poincaré ball, we extend it to the other four CCSs. In particular, the core difference between LRBN and GyroBN lies in normalization: GyroBN can normalize sample statistics across different geometries, while LRBN lacks this guarantee.

\mypara{Backbone networks.} We use a \emph{Hyperbolic Neural Network (HNN)} \citep{ganea2018hyperbolic} for the Poincaré ball and Klein HNN (KNN) \citep{mao2024klein} for the Beltrami--Klein. For the sphere and projected hypersphere, we mimic the transformation and activation in HNN \citep[Sec.~3.2]{ganea2018hyperbolic} to build the corresponding layers. The layers on the four spaces above can be expressed as
\begin{align*}
    \text{Transformation: } x^k &= \rieexp _e \left(M^k \rielog _e(x^{k-1})\right) \oplus b^k, \text{ with } b^k \in \calN, \text{ and } M^k \in \bbR{m \times n}, \\
    \text{Activation: } x^k &= \rieexp _e \left( \phi\left(\rielog _e(x^{k-1})\right)\right), \text{ with } \phi \text{ as an activation},
\end{align*}
where $e$ is the origin, $\oplus$ is the gyroaddition, and $\calN \in \{\pball{n}, \klein{n}, \sphere{n}, \projhs{n}\}$. For Lorentz geometry, we use the Lorentz fully-connected layer \citep[Eq.~3]{chen2022fully} and Lorentz activation layer \citep[Eq.~13]{bdeir2024fully}, which are briefly reviewed in \cref{app:backbone-hyperbolic-spaces}. The above backbone network is referred to as HNN, KNN, SNN, PHNN, and LNN, respectively. We collectively call them \emph{Constant Curvature Neural Networks (CCNNs)}.

\mypara{Implementation details on CCNNs.} We follow the official implementations of HGCN\footnote{\url{https://github.com/HazyResearch/hgcn}} \citep{chami2019hyperbolic}, LRBN\footnote{\url{https://github.com/CUAI/Differentiable-Frechet-Mean}} \citep{lou2020differentiating}, and HCNN\footnote{\url{https://github.com/kschwethelm/HyperbolicCV}} \citep{bdeir2024fully} to conduct experiments, where we adopt the same training settings as \citet[Sec.~H.1]{lou2020differentiating}. Specifically, the baseline encoder is a CCNN with two transformation layers: the first maps the input feature dimension to 128, and the second maps 128 to 128. After each transformation layer, we use a ReLU activation, except for the Cora data set where activation is omitted. A BN layer, GyroBN or LRBN, is inserted after each transformation layer. The curvature is set as $|K|=1$. For the manifold-valued bias parameter in GyroBN and LRBN, we apply the exponential map $\rieexp_e(v)$ to trivialize it via a Euclidean parameter $v$. The optimization is performed with Adam \citep{kingma2014adam}, using a learning rate of $1e^{-2}$ and a weight decay of $1e^{-3}$, except for the Cora data set, where weight decay is set to 0. The Fréchet mean iterations are performed until convergence.

\begin{table}[t]
  \centering
  \begin{subtable}[t]{0.99\linewidth}
    \centering
    \resizebox{\linewidth}{!}{
    \begin{tabular}{c|c|ccc|ccc|ccc}
    \toprule
    \multicolumn{2}{c|}{\textbf{Space}} & \multicolumn{3}{c|}{\textbf{Poincaré Ball} $\pball{n}$} & \multicolumn{3}{c|}{\textbf{Lorentz} $\lorentz{n}$} & \multicolumn{3}{c}{\textbf{Beltrami--Klein} $\klein{n}$} \\
    \midrule
    \multicolumn{2}{c|}{\textbf{Method}} & \textbf{HNN}   & \textbf{HNN-LRBN} & \cellcolor{HilightColor} \textbf{HNN-GyroBN} & \textbf{LNN}   & \textbf{LNN-LRBN} & \cellcolor{HilightColor} \textbf{LNN-GyroBN} & \textbf{KNN}   & \textbf{KNN-LRBN} & \cellcolor{HilightColor} \textbf{KNN-GyroBN} \\
    \midrule
    \multirow{3}[2]{*}{Disease} & ROC   & 79.21 \ensuremath{\pm} 2.14 & \red{76.58 \ensuremath{\pm} 2.15} & \cellcolor{HilightColor} \firstresults{81.18 \ensuremath{\pm} 0.93} & 87.71 \ensuremath{\pm} 1.42 & \red{85.31 \ensuremath{\pm} 0.95} & \cellcolor{HilightColor} \firstresults{88.87 \ensuremath{\pm} 0.33} & 81.31 \ensuremath{\pm} 1.37 & \red{78.98 \ensuremath{\pm} 1.51} & \cellcolor{HilightColor} \firstresults{81.56 \ensuremath{\pm} 0.70} \\
    & Fit Time & 0.027  & 0.088  & \cellcolor{HilightColor} 0.084  & 0.012  & 0.057  & \cellcolor{HilightColor} 0.057  & 0.020  & 0.089  & \cellcolor{HilightColor} 0.077  \\
    &  \#Params (M) & 0.0180  & 0.0183  & \cellcolor{HilightColor} 0.0183  & 0.0184  & 0.0187  & \cellcolor{HilightColor} 0.0187  & 0.0180  & 0.0183  & \cellcolor{HilightColor} 0.0183  \\
    \midrule
    \multirow{3}[2]{*}{Airport} & ROC   & 94.63 \ensuremath{\pm} 0.19 & \red{94.17 \ensuremath{\pm} 0.40} & \cellcolor{HilightColor} \firstresults{ 95.40 \ensuremath{\pm} 0.17} &  93.86 \ensuremath{\pm} 0.21 & \red{ 93.05 \ensuremath{\pm} 1.00} & \cellcolor{HilightColor} \firstresults{ 95.06 \ensuremath{\pm} 0.12} & 95.00 \ensuremath{\pm} 0.05 & \red{94.47 \ensuremath{\pm} 0.33} & \cellcolor{HilightColor} \firstresults{96.14 \ensuremath{\pm} 0.05} \\
    & Fit Time & 0.054  & 0.122  & \cellcolor{HilightColor} 0.119  & 0.047  & 0.089  & \cellcolor{HilightColor} 0.092  & 0.056  & 0.136  & \cellcolor{HilightColor} 0.126  \\
    &  \#Params (M) & 0.0182  & 0.0184  & \cellcolor{HilightColor} 0.0184  & 0.0186  & 0.0188  & \cellcolor{HilightColor} 0.0188  & 0.0182  & 0.0184  & \cellcolor{HilightColor} 0.0184  \\
    \midrule
    \multirow{3}[2]{*}{PubMed} & ROC   & 95.02\ensuremath{\pm} 0.42 & \red{93.40 \ensuremath{\pm} 0.20} & \cellcolor{HilightColor} \firstresults{95.83 \ensuremath{\pm} 0.11} & 95.36 \ensuremath{\pm} 0.10 & 95.88 \ensuremath{\pm} 0.09 & \cellcolor{HilightColor} \firstresults{95.89 \ensuremath{\pm} 0.11} & 95.87 \ensuremath{\pm} 0.11 & \red{89.83 \ensuremath{\pm} 0.27} & \cellcolor{HilightColor} \firstresults{96.23 \ensuremath{\pm} 0.12} \\
    & Fit Time & 0.125  & 0.342  & \cellcolor{HilightColor} 0.335  & 0.111  & 0.242  & \cellcolor{HilightColor} 0.249  & 0.125  & 0.353  & \cellcolor{HilightColor} 0.337  \\
    &  \#Params (M) & 0.0806  & 0.0809  & \cellcolor{HilightColor} 0.0809  & 0.0815  & 0.0818  & \cellcolor{HilightColor} 0.0818  & 0.0806  & 0.0809  & \cellcolor{HilightColor} 0.0809  \\
    \midrule
    \multirow{3}[2]{*}{Cora} & ROC   & 89.96 \ensuremath{\pm} 0.44 & 93.47 \ensuremath{\pm} 0.49 & \cellcolor{HilightColor} \firstresults{94.32 \ensuremath{\pm} 0.22} & 91.84 \ensuremath{\pm} 1.01 & 92.62 \ensuremath{\pm} 0.17 & \cellcolor{HilightColor} \firstresults{93.66 \ensuremath{\pm} 0.30} & 90.03 \ensuremath{\pm} 0.32 & 93.38 \ensuremath{\pm} 0.12 & \cellcolor{HilightColor} \firstresults{93.48 \ensuremath{\pm} 0.25} \\
    & Fit Time & 0.032  & 0.091  & \cellcolor{HilightColor} 0.076  & 0.114  & 0.244  & \cellcolor{HilightColor} 0.242  & 0.038  & 0.091  & \cellcolor{HilightColor} 0.076  \\
    &  \#Params (M) & 0.2001 & 0.2003 & \cellcolor{HilightColor} 0.2003 & 0.2019 & 0.2021 & \cellcolor{HilightColor} 0.2021 & 0.2001 & 0.2003 & \cellcolor{HilightColor} 0.2003 \\
    \bottomrule
    \end{tabular}%
    }
    \caption{Results on three hyperbolic spaces.}
    \label{tab:res:hyperbolic}
  \end{subtable}

  \begin{subtable}[t]{0.8\linewidth}
    \centering
    \resizebox{\linewidth}{!}{
    \begin{tabular}{c|c|ccc|ccc}
    \toprule
    \multicolumn{2}{c|}{\textbf{Space}} & \multicolumn{3}{c|}{\textbf{Projected Hypersphere} $\projhs{n}$} & \multicolumn{3}{c}{\textbf{Sphere} $\sphere{n}$} \\
    \midrule
    \multicolumn{2}{c|}{\textbf{Method}} & \textbf{PHNN}  & \textbf{PHNN-LRBN} & \cellcolor{HilightColor} \textbf{PHNN-GyroBN} & \textbf{SNN}   & \textbf{SNN-LRBN} & \cellcolor{HilightColor} \textbf{SNN-GyroBN} \\
    \midrule
    \multirow{3}[2]{*}{Disease} & ROC   & 69.70 \ensuremath{\pm} 2.01 & \red{60.25 \ensuremath{\pm} 1.25} & \cellcolor{HilightColor} \firstresults{72.26 \ensuremath{\pm} 0.61} & 54.19 \ensuremath{\pm} 2.21 & \red{53.38 \ensuremath{\pm} 4.07} & \cellcolor{HilightColor} \firstresults{71.84 \ensuremath{\pm} 0.89} \\
    & Fit Time & 0.022  & 0.066  & \cellcolor{HilightColor} 0.063  & 0.029  & 0.042  & \cellcolor{HilightColor} 0.043  \\
    &  \#Params (M) & 0.0180  & 0.0183  & \cellcolor{HilightColor} 0.0183  & 0.0181  & 0.0183  & \cellcolor{HilightColor} 0.0183  \\
    \midrule
    \multirow{3}[2]{*}{Airport} & ROC   & 89.60 \ensuremath{\pm} 0.99 & \red{87.06 \ensuremath{\pm} 0.46} & \cellcolor{HilightColor} \firstresults{90.44 \ensuremath{\pm} 0.93} & 83.63 \ensuremath{\pm} 0.77 & 86.14 \ensuremath{\pm} 0.79 & \cellcolor{HilightColor} \firstresults{91.12 \ensuremath{\pm} 1.57} \\
    & Fit Time & 0.056  & 0.102  & \cellcolor{HilightColor} 0.095  & 0.053  & 0.070  & \cellcolor{HilightColor} 0.071  \\
    &  \#Params (M) & 0.0182  & 0.0184  & \cellcolor{HilightColor} 0.0184  & 0.0182  & 0.0184  & \cellcolor{HilightColor} 0.0184  \\
    \midrule
    \multirow{3}[2]{*}{PubMed} & ROC   & 89.86 \ensuremath{\pm} 0.39 & 90.06 \ensuremath{\pm} 0.23 & \cellcolor{HilightColor} \firstresults{92.06 \ensuremath{\pm} 0.64} & 79.94 \ensuremath{\pm} 1.76 & 90.10 \ensuremath{\pm} 0.21 & \cellcolor{HilightColor} \firstresults{93.31 \ensuremath{\pm} 0.08} \\
    & Fit Time & 0.121  & 0.175  & \cellcolor{HilightColor} 0.174  & 0.120  & 0.140  & \cellcolor{HilightColor} 0.156  \\
    &  \#Params (M) & 0.0806  & 0.0809  & \cellcolor{HilightColor} 0.0809  & 0.0806  & 0.0809  & \cellcolor{HilightColor} 0.0809  \\
    \midrule
    \multirow{3}[2]{*}{Cora} & ROC   & 92.88 \ensuremath{\pm} 0.26 & \red{92.03 \ensuremath{\pm} 0.42} & \cellcolor{HilightColor} \firstresults{93.26 \ensuremath{\pm} 0.42} & 92.10 \ensuremath{\pm} 0.40 & \red{82.01 \ensuremath{\pm} 0.71} & \cellcolor{HilightColor} \firstresults{93.16 \ensuremath{\pm} 0.32} \\
    & Fit Time & 0.026  & 0.067  & \cellcolor{HilightColor} 0.063  & 0.025  & 0.044  & \cellcolor{HilightColor} 0.045  \\
    &  \#Params (M) & 0.2001 & 0.2003 & \cellcolor{HilightColor} 0.2003 & 0.2001 & 0.2003 & \cellcolor{HilightColor} 0.2003 \\
    \bottomrule
    \end{tabular}%
    }
    \caption{Results on two spherical spaces.}
    \label{tab:res:sphere}
  \end{subtable}
    \caption{Comparison of GyroBN against LRBN across five CCSs. ROC is the testing AUC reported as a percentage, fit time is measured in s/epoch, and \#Params is reported in millions. When LRBN degenerates the backbone network, the results are highlighted with \red{red}.}
  \label{tab:res:ccs}
\end{table}%

\mypara{Main results.} We compare GyroBN with LRBN across five CCSs under the CCNN backbone. \cref{tab:res:ccs} reports the 5-fold average testing AUC on four data sets. We highlight the following findings.
\begin{itemize}
    \item
    \mypara{Improved performance.} GyroBN consistently improves performance over the vanilla CCNNs across all data sets and geometries, whereas LRBN degrades performance in several cases (highlighted in \red{red}). The gains are especially pronounced on the sphere (SNN), where GyroBN achieves improvements of 17.65 (Disease), 7.49 (Airport), and 13.37 (PubMed) percentage points. This contrast underscores the advantage of GyroBN’s theoretical guarantee of normalizing sample statistics.
    \item
    \mypara{Efficiency.} As shown in the ``Fit Time'' rows of \cref{tab:res:ccs}, GyroBN is more efficient than LRBN on the Poincaré ball, Beltrami--Klein and projected hypersphere, due to the simplicity of gyro operations. Under Lorentz and spherical geometries, GyroBN and LRBN exhibit comparable efficiency.
    \item
    \mypara{Parameter equivalence.} GyroBN and LRBN require the same number of parameters, only marginally more than those of the vanilla backbone. Thus, the performance gains of GyroBN cannot be attributed to parameter size, but rather to its principled normalization mechanism.
    \end{itemize}

\subsubsection{Experiments on Correlation Neural Networks}
\label{subsec:ch3-gyrobn-experiments:cor_nn}

\mypara{Data sets and preprocessing.} Following the CorNet protocol in \cref{sec:ch5-cornet}, we use the Radar, HDM05, and FPHA data sets and model each input sequence as multichannel correlation matrices.

\mypara{Comparative methods.} Similar to \cref{subsec:ch3-gyrobn-experiments:ccs_nn}, we compare GyroBN against LRBN.

\mypara{Backbone networks.} We adopt the PHCM-based CorNet detailed in \cref{sec:ch5-cornet} as the backbone network. CorNet identifies each correlation matrix $C \in \cor{n}$ with a collection of Poincaré vectors and applies Poincaré layers. Specifically, $C$ is mapped to the poly-Poincaré space $\bbPPB{n-1}$ via \cref{eq:iso_cor_phc}. The resulting multi-channel Poincaré vectors are then merged into a single Poincaré representation through $\beta$-concatenation \citep[Sec.~3.3]{shimizu2021hyperbolic}. Then, a Poincar\'e FC layer followed by a Poincaré MLR layer constructs the network.

\mypara{Implementation details.} Since CorNet operates in the Poincaré geometry, both GyroBN and LRBN are instantiated in the Poincaré model and applied after the Poincaré FC layer. To stabilize training, we scale the learning rate of the scaling parameter $s$ by factors of $0.1$ and $0.5$ for Radar and HDM05, respectively. On FPHA, we further regularize the scaling by clamping: $\min \left(\tfrac{s}{\sqrt{v^2+\epsilon}},4\right)$. For better efficiency, the number of Fréchet mean iterations is set to 2.

\begin{table}[t]
  \centering
  \resizebox{0.99\linewidth}{!}{
    \begin{tabular}{c|ccc|ccc|ccc}
    \toprule
    \multirow{2}[4]{*}{\textbf{Methods}} & \multicolumn{3}{c|}{\textbf{Radar}} & \multicolumn{3}{c|}{\textbf{HDM05}} & \multicolumn{3}{c}{\textbf{FPHA}} \\
\cmidrule{2-10}          & \textbf{Acc}   & \textbf{Fit Time} &  \textbf{\#Params (M)} & \textbf{Acc}   & \textbf{Fit Time} &  \textbf{\#Params (M)} & \textbf{Acc}   & \textbf{Fit Time} &  \textbf{\#Params (M)} \\
    \midrule
    CorNet & 96.56 \ensuremath{\pm} 0.86 & 2.12  & 0.0438  & \firstresults{82.26 \ensuremath{\pm} 0.92} & 0.74  & 0.4091  & 90.03 \ensuremath{\pm} 0.63 & 0.70  & 1.1210  \\
    \midrule
    CorNet-LRBN & \red{92.85 \ensuremath{\pm} 2.46} & 2.33  & 0.0439  & \red{N/A} & 1.64  & 0.4094  & \red{81.53 \ensuremath{\pm} 0.72} & 1.12 & 1.1213  \\
    \midrule
    \rowcolor{HilightColor} CorNet-GyroBN & \firstresults{97.67 \ensuremath{\pm} 0.36} & 2.19  & 0.0439  & \red{80.82 \ensuremath{\pm} 0.86} & 1.33  & 0.4094  & \firstresults{92.88 \ensuremath{\pm} 0.20} & 1.07  & 1.1213  \\
    \bottomrule
    \end{tabular}%
    }
    \caption{Comparison of CorNet with or without RBN layers. Accuracy is reported as a percentage, fit time is measured in s/epoch, and \#Params is reported in millions.}
    \label{tab:res:cornet_gyrobn}%
\end{table}

\mypara{Main results.} We summarize the comparison of CorNet with or without normalization layers in \cref{tab:res:cornet_gyrobn}. Overall, GyroBN demonstrates clear benefits on Radar and FPHA with negligible parameter cost and modest efficiency trade-offs. On HDM05, however, neither GyroBN nor LRBN improves the baseline, with LRBN even diverging and failing to converge.

\section{Conclusion}
\label{sec:ch3-conclusion}

This chapter developed a unified approach to RBN across broad families of manifolds. We first presented LieBN, which operates under left-, right-, and bi-invariant metrics. LieBN uses the group translations for centering and biasing and performs scaling in the tangent space at the identity element. These operations provide theoretical control over Riemannian sample and population statistics. Their instantiations on SPD, rotation, and full-rank correlation manifolds further demonstrate how a choice of Lie group structure and invariant metric yields a concrete normalization layer, including the power-deformed SPD structures and CRIM developed in this chapter.

The chapter then introduced pseudo-reductive gyrogroups as a relaxation of classical gyrogroups, providing a broader algebraic foundation for Riemannian normalization. Building on this structure, we developed GyroBN and showed that gyroisometric gyrations enable theoretical control over sample statistics. Since every Lie group is a gyrogroup with identity gyrations, GyroBN recovers LieBN as a special case. GyroBN replaces group subtraction, addition, and tangent-space scaling with gyrosubtraction, gyroaddition, and scalar gyromultiplication, thereby extending the normalization principle to broader geometries. Finally, we instantiated the framework on the Grassmannian, five CCS models, and the full-rank correlation manifold. Experiments across all considered geometries support the effectiveness of this extension.

Together, LieBN and GyroBN extend normalization with theoretical control of intrinsic statistics from Lie groups to pseudo-reductive gyrogroups. Gyrogroups substantially broaden the scope beyond Lie groups but do not encompass all manifolds. A natural direction for future work is therefore to extend this normalization principle to manifolds that do not admit suitable gyro-structures. The next chapter considers intrinsic classification.

    \chapter{Riemannian Multinomial Logistic Regression}
\label{chapter:rmlr}

\section{Introduction}
\label{sec:ch4-introduction}

Although Riemannian networks demonstrated success in many applications, most approaches still rely on Euclidean spaces for classification, such as tangent spaces \citep{huang2017riemannian, huang2017deep, brooks2019riemannian,wang2021symnet,wang2022learning,nguyen2022gyro,kobler2022spd,wang2022dreamnet,chen2023riemannian}, ambient Euclidean spaces \citep{wang2020deep, song2021approximate, song2022eigenvalues}, or coordinate systems \citep{chakraborty2018statistical}.\footnote{Notice that there are also works designing classifiers for grid-based manifold-valued data \citep{chakraborty2020manifoldnet}, but we focus on non-gridded data in line with many previous SPD networks.}
However, these strategies distort the intrinsic geometry of the manifold, undermining the effectiveness of Riemannian networks.
Researchers have recently started directly developing \emph{Riemannian Multinomial Logistic Regression (RMLR)} on manifolds.
Inspired by the idea of hyperplane margin \citep{lebanon2004hyperplane}, \citet{ganea2018hyperbolic} developed a hyperbolic MLR in the Poincar\'e ball for HNNs.
Motivated by HNNs, \citet{nguyen2023building} developed three kinds of gyro SPD MLRs based on three distinct gyro-structures of the SPD manifold.
\citet{nguyen2024matrix} proposed gyro MLRs for the \emph{Symmetric Positive Semidefinite (SPSD) manifold} based on the product of gyro spaces.
However, these classifiers often rely on manifold-specific geometric structures, limiting their generalizability to other geometries. For instance, the hyperbolic MLR \citep{ganea2018hyperbolic} relies on the generalized law of sines, while the gyro MLRs \citep{nguyen2023building,nguyen2024matrix} rely on gyro-structures.

This chapter proceeds in two stages. We first study SPD manifolds endowed with pullback Euclidean metrics, whose flat geometry reduces the point-to-hyperplane infimum to a Euclidean problem and yields closed-form intrinsic MLRs. We then extend the classification principle to general Riemannian manifolds. Rather than evaluating the potentially intractable point-to-hyperplane infimum on each manifold, the general RMLR adopts a Riemannian-trigonometric formulation that requires only a well-defined Riemannian logarithm. Since the SPD classifiers in \cref{sec:ch4-spd-classifiers} are special cases of the general RMLR framework in \cref{sec:ch4-rmlr-general-geometries}, and the two parts use the same experimental settings, all experiments\footnote{The code is available at \url{https://github.com/GitZH-Chen/RMLR}.} are presented together in \cref{rmlr:sec:experiments}.

Throughout this chapter and its accompanying appendix material, all parameters are assumed to satisfy the standing admissibility conditions.\footnote{All normal vectors are nonzero. Whenever a displayed formula contains $1/\theta$ or $1/\theta^2$, we assume $\theta\neq0$, except when an explicit limit as $\theta\to0$ is considered. The inner-product parameters satisfy $\alphabeta\in\bfst$, as defined in \cref{eq:ch2-spd-alpha-beta-inner}. These conditions are not repeated below.}

\section{Multinomial Logistic Regression on SPD Manifolds}
\label{sec:ch4-spd-classifiers}

\subsection{Introduction}
\label{spdmlr:sec:intro}

SPD matrices are commonly encountered in a diverse range of scientific fields, such as medical imaging \citep{chakraborty2018statistical,chakraborty2020manifoldnet}, signal processing \citep{arnaudon2013riemannian,hua2017matrix,brooks2019exploring,brooks2019riemannian}, elasticity \citep{moakher2006averaging,guilleminot2012generalized}, question answering \citep{lopez2021vector,nguyen2022gyro}, graph classification \citep{chen2023distribution}, and computer vision \citep{huang2017riemannian,harandi2018dimensionality,zhen2019dilated,chakraborty2020manifoldnorm,chen2020covariance,zhang2020deep,chakraborty2020manifoldnet,chen2021hybrid,song2021approximate,song2022fast}.
Despite their ubiquitous presence, traditional learning algorithms are ineffective in handling the non-Euclidean geometry of SPD matrices.
To address this limitation, several Riemannian metrics \citep{pennec2006riemannian,arsigny2005fast,lin2019riemannian} have been proposed.
With these Riemannian metrics, various machine learning techniques can be generalized to SPD manifolds.

Inspired by the great success of deep learning \citep{hochreiter1997long, krizhevsky2012imagenet,he2016deep}, several deep networks have been developed on SPD manifolds.
Despite their promising performance, many approaches still rely on Euclidean spaces for classification, such as tangent spaces \citep{huang2017riemannian, brooks2019riemannian,wang2021symnet,nguyen2022gyro,kobler2022spd,wang2022dreamnet,chen2023riemannian}, ambient Euclidean spaces \citep{wang2020deep, song2021approximate, song2022eigenvalues}, and coordinate systems \citep{chakraborty2018statistical}.
However, these strategies distort the intrinsic geometry of the SPD manifold, undermining the effectiveness of SPD neural networks. 
Notably, there are also some similarity-based classifiers originally designed for shallow learning methods \citep{gao2019robust,harandi2018dimensionality,chen2021hybrid}.
Although these classifiers can be extended to deep SPD neural networks \citep{wang2022learning,wangspdmetric2024}, the calculation of pairwise distances might undermine training efficiency.
Recently, motivated by HNNs \citep{ganea2018hyperbolic}, three kinds of SPD \emph{Multinomial Logistic Regression (MLR)} classifiers based on the gyro-structures induced by LEM, LCM, and AIM were developed by \citet{nguyen2023building}.
However, the proposed SPD MLRs rely on the gyro-structures, limiting their generality.
Besides, \citet{chakraborty2020manifoldnet} also introduced an invariant layer for manifold-valued data mimicking the invariant FC layer in \emph{Convolutional Neural Networks (CNNs)}. However, it is designed for gridded manifold-valued data, which is not the primary data type encountered in many other SPD networks. Following the convention of most SPD networks, we only focus on non-gridded cases.

\textit{In fact, SPD MLR can be directly derived under LEM and LCM without the assistance of gyro-structures.} More generally, LEM and LCM are pullback Euclidean metrics, which are metrics pulled back from the Euclidean space.
This section develops a unified construction of SPD MLR across pullback Euclidean metrics.
On the empirical side, we focus on the parameterized LEM and LCM defined in \cref{subsubsec:spd_param_lie_groups}, which generalize the standard LEM and LCM by the pullback of matrix power. Their deformation behavior is established in \cref{prop:spd_param_lem_lcm_deformation}.
We showcase our SPD MLRs under these parameterized metrics.
Besides, our framework encompasses the gyro SPD MLRs induced by the standard LEM and LCM in \citet{nguyen2023building}.
More importantly, our framework also provides an intrinsic explanation for the commonly used LogEig classifier on SPD manifolds, which consists of successive matrix logarithm, FC, and softmax layers. 
The main \textbf{contributions} are summarized as follows:
\begin{enumerate}
    \item 
    We develop a unified construction of SPD MLR under pullback Euclidean metrics and instantiate it under two parameterized metric families.
    \item
    Our framework offers an intrinsic explanation of the most popular LogEig classifier, which stacks matrix logarithm, the FC layer, and softmax.
\end{enumerate}

\mypara{Outline.} \cref{spdmlr:sec:spd_mlr} develops SPD MLRs under pullback Euclidean metrics. The deformed-metric instantiation follows in \cref{spdmlr:sec:deformed_metrics}. \cref{spdmlr:sec:logeig} reinterprets the existing LogEig classifier intrinsically. The corresponding experiments are presented together with the general RMLR experiments in \cref{rmlr:sec:experiments}. Proofs are deferred to \cref{app:spdmlr-proofs}.

\subsection{SPD Multinomial Logistic Regression}
\label{spdmlr:sec:spd_mlr}

This section first reformulates the Euclidean MLR. Then, we deal with SPD MLR under an arbitrary pullback Euclidean metric on SPD manifolds. We use the pullback metric in \cref{def:ch2-riemannian-isometry} with a Euclidean codomain. The induced abelian Lie-group structure, bi-invariant metric, distance, and Riemannian operators required below are established later in \cref{alem:lem:g_spd}.

\subsubsection{Reformulation of Euclidean MLR}
\label{spdmlr:subsec:reform_emlr}

The Euclidean MLR was first reformulated by \citet{lebanon2004hyperplane} from the perspective of distances to margin hyperplanes. Hyperbolic MLR was designed based on this reformulation \citep{ganea2018hyperbolic}.
\citet{nguyen2023building} further proposed three gyro SPD MLRs based on the gyro-structures induced by AIM, LEM, and LCM.
We now briefly review the reformulation of Euclidean MLR.

Given $C$ classes, MLR in $\bbR{n}$ computes the following softmax probabilities:
\begin{equation}
    \label{spdmlr:eq:EMLR_reform_start}
    \forall k \in \{1,\ldots,C\}, \quad p(y=k \mid x) \propto \exp\left(\left\langle a_k,x\right\rangle-b_k\right),
\end{equation}
where $b_k\in\bbRscalar$ and $x,a_k\in\bbR{n}$.
As shown in \citet[Sec.~5]{lebanon2004hyperplane} and \citet[Sec.~3.1]{ganea2018hyperbolic}, \cref{spdmlr:eq:EMLR_reform_start} can be reformulated as
\begin{equation}
    \label{spdmlr:eq:EMLR_reform_end}
    \begin{aligned}
    p(y=k \mid x) \propto
    \exp\left(\operatorname{sign}\left(\left\langle a_k,x-p_k\right\rangle\right)\left\|a_k\right\|d\left(x,H_{a_k,p_k}\right)\right),
    \end{aligned}
\end{equation}
where $\langle a_k,p_k\rangle=b_k$, and $H_{a_k,p_k}$ is referred to as a hyperplane, defined as
\begin{equation}
    \label{spdmlr:eq:euc_hyperplane}
    H_{a_k,p_k}=\left\{x\in\bbR{n}\mid\left\langle a_k,x-p_k\right\rangle=0\right\}.
\end{equation}

As reviewed in \cref{tab:ch2-riemannian-examples}, $\rielog_p x$ is the natural generalization of the directional vector $\overrightarrow{px}=x-p$ starting at $p$ and ending at $x$, while the Riemannian metric at $p$ corresponds to the inner product.
Therefore, the MLR in \cref{spdmlr:eq:EMLR_reform_end} and hyperplane in \cref{spdmlr:eq:euc_hyperplane} can be readily generalized to the SPD manifold $(\spd{n},g)$.

\begin{parisdefinition}[SPD hyperplanes]
\label{spdmlr:def:spd_hyperplanes}
Given $P\in\spd{n}$ and $A\in T_P\spd{n}\backslash\{\bbzero\}$, we define the \emph{SPD hyperplane} as
\begin{equation}
    \label{spdmlr:eq:spd_hyperplane}
    \tilde{H}_{A,P}=\left\{S\in\spd{n}\mid g_P\left(\rielog_P S,A\right)=\left\langle\rielog_P S,A\right\rangle_P=0\right\},
\end{equation}
where $P$ and $A$ are referred to as shift and normal matrices, respectively.
\end{parisdefinition}

\begin{parisdefinition}[SPD MLR]
\label{spdmlr:def:spd_mlr}
The \emph{SPD MLR} is defined as
\begin{equation}
    \label{spdmlr:eq:spd_mlr}
    \begin{aligned}
    p(y=k \mid S)
    &\propto \exp\left(\operatorname{sign}\left(\left\langle A_k,\rielog_{P_k}(S)\right\rangle_{P_k}\right)\left\|A_k\right\|_{P_k} d\left(S,\tilde{H}_{A_k,P_k}\right)\right),
    \end{aligned}
\end{equation}
where $P_k\in\spd{n}$, $A_k\in T_{P_k}\spd{n}\backslash\{\bbzero\}$, $\langle\cdot,\cdot\rangle_{P_k}=g_{P_k}$, $\|\cdot\|_{P_k}$ is the norm on $T_{P_k}\spd{n}$ induced by $g$ at $P_k$, and $\tilde{H}_{A_k,P_k}$ is a margin hyperplane in $\spd{n}$ as defined in \cref{spdmlr:eq:spd_hyperplane}.
$d\left(S,\tilde{H}_{A_k,P_k}\right)$ denotes the margin distance between $S$ and the SPD hyperplane $\tilde{H}_{A_k,P_k}$, which is formulated as
\begin{equation}
    \label{spdmlr:eq:spd_pem_dist_hyperplane}
    d\left(S,\tilde{H}_{A_k,P_k}\right)=\inf_{Q\in\tilde{H}_{A_k,P_k}}d(S,Q),
\end{equation}
where $d(S,Q)$ is the geodesic distance induced by $g$.
\end{parisdefinition}

In geometry, the hyperplane in \cref{spdmlr:eq:euc_hyperplane} is actually a regular submanifold of the trivial manifold $\bbR{n}$.
As for our definition of SPD hyperplanes, we have a similar result.

\begin{parisproposition}[Submanifolds]
\label{spdmlr:prop:hyperplanes_as_submanifolds}
\linktoproof{spdmlr:prop:hyperplanes_as_submanifolds}
The SPD hyperplane, as defined in \cref{spdmlr:eq:spd_hyperplane}, is a regular submanifold of the SPD manifold if $\rielog_P:\spd{n}\rightarrow T_P\spd{n}$ is a global diffeomorphism.
\end{parisproposition}

\cref{spdmlr:prop:hyperplanes_as_submanifolds} rationalizes \cref{spdmlr:def:spd_hyperplanes}, as both the SPD hyperplane and Euclidean hyperplane are submanifolds.
Nevertheless, we still follow the nomenclature of \citet{ganea2018hyperbolic,lebanon2004hyperplane} and call $\tilde{H}_{A,P}$ an SPD hyperplane.

\subsubsection{SPD MLRs under Pullback Euclidean Metrics}
\label{spdmlr:subsec:general_spd_mlr}

For our SPD MLR in \cref{spdmlr:def:spd_mlr}, under most Riemannian metrics on SPD manifolds, all the operators involved in \cref{spdmlr:eq:spd_mlr} have closed-form expressions except the margin distance in \cref{spdmlr:eq:spd_pem_dist_hyperplane}.
Therefore, the only difficulty lies in the calculation of the margin distance. This subsection proposes a general expression for SPD MLRs under pullback Euclidean metrics, which are defined by a pullback from Euclidean metric, as detailed in \cref{alem:lem:g_spd}. We choose the identity matrix $I$ as a fixed anchor.

We choose pullback Euclidean metrics as our starting metrics mainly because of their extensive inclusion and easy computation. Several Riemannian metrics, including LEM, LCM, and their variants \citep{thanwerdas2023n,thanwerdas2022geometry}, are pullback Euclidean metrics. Besides, due to their fast and simple calculation, the margin distance under a pullback Euclidean metric has a closed-form expression, while obtaining distances to hyperplanes under other metrics, such as AIM, would be complicated.

We start by calculating the margin distance in \cref{spdmlr:eq:spd_pem_dist_hyperplane} under a given pullback Euclidean metric.

\begin{parislemma}
\label{spdmlr:lem:dist_to_hyperplane_pems}
\linktoproof{spdmlr:lem:dist_to_hyperplane_pems}
Given a pullback Euclidean metric $g$, the margin distance defined in \cref{spdmlr:eq:spd_pem_dist_hyperplane} has a closed-form solution:
\begin{align}
    d\left(S,\tilde{H}_{A_k,P_k}\right)
    &=d\left(\phi(S),H_{\phi_{*,P_k}(A_k),\phi(P_k)}\right)\\
    \label{spdmlr:eq:dist_s_to_hyperplane}
    &=\frac{\left|\left\langle\phi(S)-\phi(P_k),\phi_{*,P_k}(A_k)\right\rangle\right|}{\left\|A_k\right\|_{P_k}},
\end{align}
where $|\cdot|$ is the absolute value.
\end{parislemma}

Putting \cref{spdmlr:eq:dist_s_to_hyperplane} into \cref{spdmlr:eq:spd_mlr}, we obtain our SPD MLR under a given pullback Euclidean metric:
\begin{align}
    p(y=k \mid S)
    &\propto\exp\left(\left\langle A_k,\rielog_{P_k}(S)\right\rangle_{P_k}\right)\\
    \label{spdmlr:eq:spd_dist_s_to_h}
    &=\exp\left(\left\langle\phi(S)-\phi(P_k),\phi_{*,P_k}(A_k)\right\rangle\right),
\end{align}
where $S,P_k\in\spd{n}$ and $A_k\in T_{P_k}\spd{n}\backslash\{\bbzero\}$.
When $P_k$ is fixed, $A_k\in T_{P_k}\spd{n}$ indeed lies in a Euclidean space.
However, $P_k$ would vary during training, making $A_k$ non-Euclidean.
To remedy this issue, $A_k$ can be generated from a Euclidean parameter in a fixed tangent space through one of three mechanisms: Riemannian parallel transport \citep{do1992riemannian}, vector transport\footnote{Vector transport is often used in optimization as a substitute for parallel transport because its expressions are typically simpler and cheaper to compute \citep[Sec.~10.5]{boumal2023introduction}.} \citep{absil2009optimization}, or the differential of a group translation, including Lie group translation \citep[Sec.~20]{loring2011introduction} and gyrogroup translation \citep{ungar2005analytic}.
We focus on Riemannian parallel transport and the differential of a Lie group translation, for which we establish an equivalence under pullback Euclidean metrics.
Under parallel transport, we write $A_k=\pt{Q}{P_k}(\tilde{A}_k)$ with $\tilde{A}_k\in T_Q\spd{n}$ as a Euclidean parameter.
This is the solution also adopted by HNNs \citep{ganea2018hyperbolic}, where the tangent point is the zero vector.
Since the Lie groups associated with pullback Euclidean metrics are abelian, we only consider the left translation.
We have the following two lemmas to show the relation between parallel transport and the differential of left translation.

\begin{parislemma}
\label{spdmlr:lem:equ_pt_and_lt}
\linktoproof{spdmlr:lem:equ_pt_and_lt}
Given a pullback Euclidean metric, any parallel transport is equivalent to the differential map of a left translation and vice versa.
\end{parislemma}

\begin{parislemma}
\label{spdmlr:lem:pt_anchor_invariance}
\linktoproof{spdmlr:lem:pt_anchor_invariance}
Given two fixed SPD matrices $Q_1,Q_2\in\spd{n}$, we have the following equivalence for parallel transports under a pullback Euclidean metric:
\begin{equation}
    \begin{aligned}
    &\forall \tilde{A}_{1,k}\in T_{Q_1}\spd{n},\ \exists!\ \tilde{A}_{2,k}\in T_{Q_2}\spd{n},\\
    &\st \pt{Q_1}{P_k}\left(\tilde{A}_{1,k}\right)=\pt{Q_2}{P_k}\left(\tilde{A}_{2,k}\right).
    \end{aligned}
\end{equation}
\end{parislemma}

\cref{spdmlr:lem:equ_pt_and_lt} indicates that under pullback Euclidean metrics, the above two solutions are equivalent, while \cref{spdmlr:lem:pt_anchor_invariance} implies that anchor points can be arbitrarily chosen.
Therefore, without loss of generality, we generate $A_k$ from the tangent space at the identity matrix $I$ by parallel transport, \ie $A_k=\pt{I}{P_k}(\tilde{A}_k)$ with $\tilde{A}_k\in T_I\spd{n}\cong\sym{n}$.
Together with \cref{alem:eq:gene_pt_spd}, \cref{spdmlr:eq:spd_dist_s_to_h} can be further simplified.

\begin{paristheorem}[SPD MLR under a pullback Euclidean metric]
\label{spdmlr:thm:general_mlr_pems}
\linktoproof{spdmlr:thm:general_mlr_pems}
Under any pullback Euclidean metric, SPD MLR and SPD hyperplane are
\begin{align}
    \label{spdmlr:eq:spd_dist_s_to_h_final}
    p(y=k \mid S)&\propto\exp\left(\left\langle\phi(S)-\phi(P_k),\phi_{*,I}(\tilde{A}_k)\right\rangle\right),\\
    \label{spdmlr:eq:spd_hyperplane_final}
    \tilde{H}_{\tilde{A}_k,P_k}
    &=\left\{S\in\spd{n}\mid\left\langle\phi(S)-\phi(P_k),\phi_{*,I}(\tilde{A}_k)\right\rangle=0\right\},
\end{align}
where $\tilde{A}_k\in T_I\spd{n}\backslash\{\bbzero\}\cong\sym{n}\backslash\{\bbzero\}$ is a symmetric matrix, and $P_k\in\spd{n}$ is an SPD matrix.
\end{paristheorem}

\subsection{SPD MLRs under Deformed LEM and LCM}
\label{spdmlr:sec:deformed_metrics}

In this section, we first review the deformed LEM and LCM, and then showcase our SPD MLR in \cref{spdmlr:thm:general_mlr_pems} under these deformed metrics.

As reviewed in \cref{subsubsec:spd_param_lie_groups}, inspired by the deforming utility of the matrix power function \citep{thanwerdas2019affine,thanwerdas2022geometry}, we define $\triparamLEM$ and $\paramLCM$ as the pullback metrics of $\biparamLEM$ and LCM by the matrix power function $(\cdot)^\theta$ and scaled by $\frac{1}{\theta^2}$ for $\theta\neq0$.
As shown in \cref{prop:spd_param_lem_lcm_deformation}, $\triparamLEM$ is equal to $\biparamLEM$ and $\paramLCM$ interpolates between the standard LCM for $\theta=1$ and an LEM-like metric as $\theta\rightarrow0$.

Besides, both $\biparamLEM$ and $\paramLCM$ are pullback Euclidean metrics, as summarized in \cref{tab:ops_liebn_spd}.
Therefore, the SPD MLRs under these two families of metrics can be directly obtained by \cref{spdmlr:thm:general_mlr_pems}.

\begin{pariscorollary}[SPD MLRs under the deformed LEM and LCM]
\label{spdmlr:cor:spd_mlr_param_lem_lcm}
\linktoproof{spdmlr:cor:spd_mlr_param_lem_lcm}
The SPD MLR under $\biparamLEM$ is
\begin{equation}
    \label{spdmlr:eq:spd_mlr_param_lem}
    \begin{aligned}
    p(y=k \mid S)\propto
    \exp\left[\left\langle\mlog(S)-\mlog(P_k),\tilde{A}_k\right\rangle^{\alphabeta}\right],
    \end{aligned}
\end{equation}
where $\tilde{A}_k\in T_I\spd{n}\cong\sym{n}$ and $P_k\in\spd{n}$. The SPD MLR under $\paramLCM$ is
\begin{equation}
    \label{spdmlr:eq:spd_mlr_param_lcm}
    p(y=k \mid S)\propto\exp\left[\frac{1}{\theta}\left\langle X,Y\right\rangle\right],
\end{equation}
with $X$ and $Y$ defined as
\begin{align}
    X&=\lfloor\tilde{K}\rfloor-\lfloor\tilde{L}_k\rfloor+\left[\dlog\left(\bbD(\tilde{K})\right)-\dlog\left(\bbD(\tilde{L}_k)\right)\right],\\
    Y&=\lfloor\tilde{A}_k\rfloor+\frac{1}{2}\bbD(\tilde{A}_k),
\end{align}
where $\tilde{K}=\chol(S^\theta)$, $\tilde{L}_k=\chol(P_k^\theta)$, and $\bbD(\tilde{A}_k)$ denotes a diagonal matrix with the diagonal elements of $\tilde{A}_k$.
\end{pariscorollary}

\begin{figure}[t]
\centering
\includegraphics[width=\textwidth,trim={0cm 0cm 0cm 0cm}]{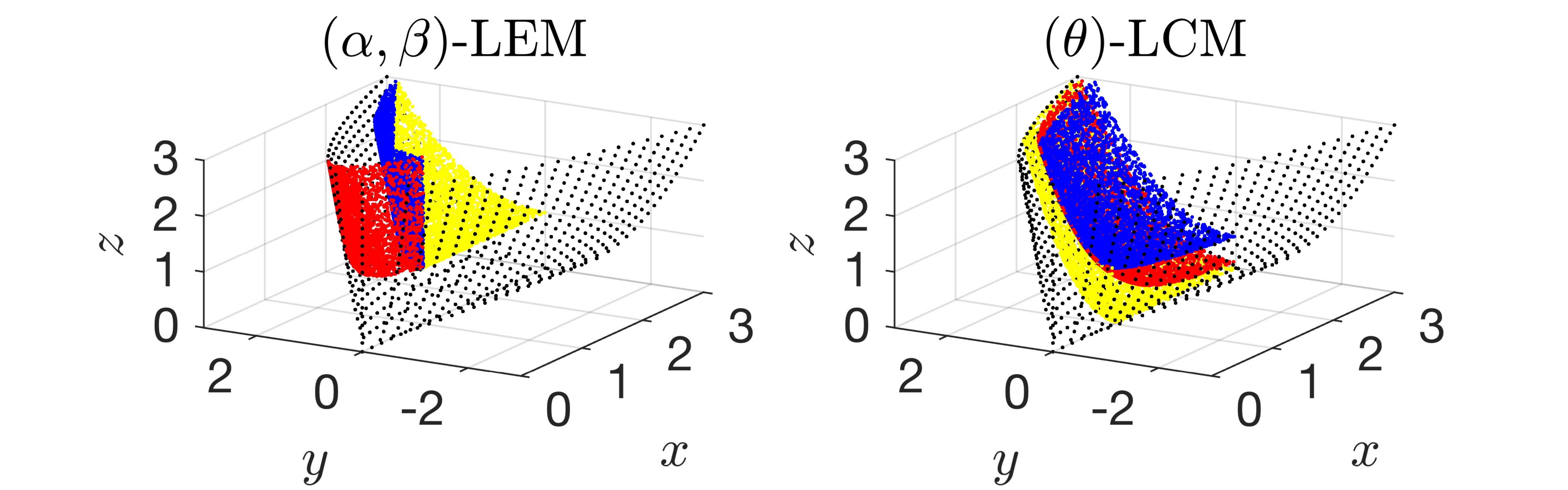}
\caption{Conceptual illustration of SPD hyperplanes induced by $\biparamLEM$ and $\paramLCM$. In each subfigure, the black dots are SPSD matrices, denoting the boundary of $\spd{2}$, while the blue, red, and yellow dots denote three SPD hyperplanes.}
\label{spdmlr:fig:hyperplanes}
\vspace{-5mm}
\end{figure}

$\spd{2}$ can be visualized as an open cone in $\bbR{3}$ by the condition that $P=\left(\begin{array}{ll}x&y\\y&z\end{array}\right)\in\sym{2}$ is positive definite if and only if $x,z>0\land xz>y^2$.
\cref{spdmlr:fig:hyperplanes} illustrates SPD hyperplanes induced by $\biparamLEM$ and $\paramLCM$.

\begin{parisremark}
This construction incorporates the results with respect to LEM and LCM presented by \citet{nguyen2023building}.
For $\biparamLEM$, when $\alphabeta=(1,0)$, $\biparamLEM$ becomes the standard LEM.
Our margin distance to the hyperplane in \cref{spdmlr:lem:dist_to_hyperplane_pems} becomes the pseudo-gyrodistance under LEM \citep[Thm.~2.23]{nguyen2023building}.
For $\paramLCM$, when $\theta=1$, $\paramLCM$ becomes the standard LCM.
\cref{spdmlr:lem:dist_to_hyperplane_pems} then becomes the pseudo-gyrodistance induced by LCM \citep[Thm.~2.24]{nguyen2023building}.
However, our framework does not require gyro-structures and directly obtains the margin distance and SPD MLR based on the Riemannian metric.
\end{parisremark}

\subsection{Rethinking the Existing LogEig Classifier}
\label{spdmlr:sec:logeig}

Many existing SPD neural networks \citep{huang2017riemannian,brooks2019riemannian,wang2021symnet,wang2022dreamnet,chen2023riemannian} rely on a Euclidean MLR in the codomain of matrix logarithm, \ie a matrix logarithm followed by an FC layer and a softmax layer. For simplicity, we call this classifier LogEig MLR. The existing explanation of LogEig MLR is that it approximates manifolds by a tangent space.
However, our framework can offer a novel intrinsic explanation for this widely used MLR.

When $\alphabeta=(1,0)$ for $\biparamLEM$, the SPD MLR in \cref{spdmlr:eq:spd_mlr_param_lem} is very similar to the LogEig MLR.
However, due to the nonlinearity of $\mlog(\cdot)$ and the non-Euclideanness of the SPD parameter $P_k$, SPD MLR cannot be hastily viewed as equivalent to LogEig MLR.
Nevertheless, under special circumstances, \cref{spdmlr:eq:spd_mlr_param_lem} is indeed equivalent to a LogEig MLR.

\begin{parisproposition}
\label{spdmlr:prop:equivalence_lem_mlr}
\linktoproof{spdmlr:prop:equivalence_lem_mlr}
The LEM-based SPD MLR is equivalent to a LogEig MLR with parameters in the FC layer optimized by Euclidean SGD when SPD manifolds are endowed with the standard LEM, the SPD parameter $P_k$ in \cref{spdmlr:eq:spd_mlr_param_lem} is optimized by LEM-based RSGD, and the Euclidean parameter $\tilde{A}_k$ is optimized by Euclidean SGD.
\end{parisproposition}

\cref{spdmlr:prop:equivalence_lem_mlr} implies that, when optimized by LEM-based RSGD, the LEM-based SPD MLR is equivalent to the Euclidean MLR in the codomain of matrix logarithm.
Nevertheless, a substantial body of prior work underscores the theoretical and empirical superiority of AIM-based optimization over its LEM-based counterpart \citep{sra2015conic,han2021riemannian}.
Therefore, we adopt the AIM-based optimizer to update the involved SPD parameters.

\section{Extension to General Riemannian Manifolds}
\label{sec:ch4-rmlr-general-geometries}

\subsection{Introduction}
\label{rmlr:sec:intro}
The SPD MLR framework developed in \cref{sec:ch4-spd-classifiers} constructs an intrinsic classifier from the geodesic distance between an SPD matrix and a margin hyperplane. Its central quantity is the point-to-hyperplane distance in \cref{spdmlr:eq:spd_pem_dist_hyperplane}, defined as the infimum of the geodesic distance over all points on the hyperplane. For the pullback Euclidean metrics considered in \cref{spdmlr:subsec:general_spd_mlr}, this optimization admits a closed-form solution. On a general Riemannian manifold, however, directly evaluating this infimum can require solving a difficult, potentially non-convex optimization problem, which prevents an extension of the preceding framework.

We circumvent this obstacle by reinterpreting the Euclidean margin distance from a trigonometric perspective rather than directly solving the infimum. This alternative characterization expresses the margin distance through the angle between geodesics and the geodesic distance from the input to the hyperplane anchor. Lifting this characterization to Riemannian manifolds yields a closed-form RMLR. Accordingly, the resulting framework only requires an explicit expression of the Riemannian logarithm, which is the minimal geometric requirement for extending Euclidean MLR to manifolds. Since this requirement is satisfied by many manifolds commonly encountered in machine learning, the framework applies broadly across different geometries. This reliance on an operator shared by many manifolds, rather than on additional manifold-specific structure, makes RMLR a unified classification module across geometries.

We instantiate the framework on SPD manifolds and rotation matrices. On the SPD manifold, we systematically develop SPD MLRs under five families of power-deformed metrics and provide a complete theoretical discussion of their geometric properties. On the Lie group $\so{n}$, we construct a Lie MLR under the widely used bi-invariant metric, providing the first extension of Euclidean MLR to Lie groups. Moreover, the framework incorporates several existing Riemannian MLRs, including gyro SPD MLRs \citep{nguyen2023building}, the SPD MLRs developed in \cref{sec:ch4-spd-classifiers}, and gyro SPSD MLRs \citep{nguyen2024matrix}.

Our SPD MLRs are validated on four SPD backbone networks, including SPDNet \citep{huang2017riemannian} on the radar and human action recognition tasks and TSMNet \citep{kobler2022spd} on the EEG classification tasks for the Riemannian feedforward network, RResNet \citep{katsman2023riemannian} on the human action recognition task for the Riemannian residual network, and SPDGCN \citep{zhao2023modeling} on the node-classification task for the Riemannian graph neural network.
Our Lie MLR is validated on the classic LieNet \citep{huang2017deep} backbone for the human action recognition task.
Compared with previous non-intrinsic classifiers, our MLRs achieve consistent performance gains.
In particular, our SPD MLRs outperform the previous classifiers by \textbf{14.23 percentage points} on SPDNet and \textbf{13.72 percentage points} on RResNet for human action recognition, and \textbf{4.46 percentage points} on TSMNet for EEG inter-subject classification.
Furthermore, our Lie MLR can improve both the training stability and performance.
In summary, our \textbf{main theoretical contributions} are the following:
\begin{enumerate}
    \item
    We develop a unified RMLR framework for general Riemannian manifolds that requires only the Riemannian logarithm and incorporates several existing manifold-specific RMLRs as special cases.
    \item
    We systematically propose five families of SPD MLRs based on different geometries of the SPD manifold.
    \item
    We propose a novel Lie MLR for deep neural networks on $\so{n}$.
\end{enumerate}

\mypara{Outline.} \cref{rmlr:sec:rmlr} revisits the existing Riemannian MLRs and proposes the general RMLR framework. \cref{rmlr:sec:spd_mlrs} instantiates the framework on SPD manifolds under five families of power-deformed metrics. \cref{rmlr:sec:lie_mlr} presents the Lie MLR on $\so{n}$. \cref{rmlr:sec:experiments} reports experiments on Riemannian feedforward, residual, and graph networks, direct SPD classification, and LieNet. Proofs are deferred to \cref{app:rmlr-proofs}.

\subsection{Riemannian Multinomial Logistic Regression}
\label{rmlr:sec:rmlr}

Inspired by \citet{lebanon2004hyperplane}, prior work extended the Euclidean MLR to hyperbolic, SPD, and SPSD manifolds \citep{ganea2018hyperbolic,nguyen2023building,nguyen2024matrix}. \cref{sec:ch4-spd-classifiers} further develops SPD MLRs under pullback Euclidean metrics.
However, these classifiers rely on specific Riemannian properties, such as the generalized law of sines, gyro-structures, and flat metrics, which limit their generality.
In this section, we first revisit several existing MLRs and then propose our Riemannian classifiers with minimal geometric requirements. 

\subsubsection{Revisiting Existing Multinomial Logistic Regressions}
\label{rmlr:subsec:re_exist_MLR}

The Euclidean MLR and its reformulation through margin distances to hyperplanes have already been reviewed in \cref{spdmlr:subsec:reform_emlr}. By abstracting the expressions for the SPD MLR and SPD hyperplanes in \cref{spdmlr:eq:spd_mlr,spdmlr:eq:spd_hyperplane}, we can readily extend this construction from the SPD manifold to a general Riemannian manifold $\calM$:
\begin{align}
    \label{rmlr:eq:rmlr_v1}
    p(y=k \mid S) &\propto \exp \left(\operatorname{sign}(\langle \tilde{A}_k, \rielog_{P_k}(S) \rangle_{P_k})\|\tilde{A}_k\|_{P_k} \tilde{d} (S, \tilde{H}_{\tilde{A}_k, P_k}) \right),\\
    \label{rmlr:eq:r_hyperplane}
    \tilde{H}_{\tilde{A}_k, P_k} &= \left\{S \in \calM\mid g_{P_k}( \rielog_{P_k} S, \tilde{A}_k) =0\right\},
\end{align}
where $P_k \in \calM, \tilde{A}_k \in T_{P_k}\calM \backslash \{ \zerovec \}$, $g_{P_k}$ is the Riemannian metric at ${P_k}$, and $\rielog_{P_k}$ is the Riemannian logarithm at ${P_k}$.
The margin distance is defined as an infimum:
\begin{equation} \label{rmlr:eq:dist_hyperplane}
    \tilde{d} (S, \tilde{H}_{\tilde{A}_k, P_k}) =\inf _{Q \in \tilde{H}_{\tilde{A}_k, P_k}} d(S, Q).
\end{equation}

The MLRs in \citet{lebanon2004hyperplane,ganea2018hyperbolic,nguyen2023building} and \cref{sec:ch4-spd-classifiers} can be viewed as different implementations of \cref{rmlr:eq:rmlr_v1,rmlr:eq:r_hyperplane,rmlr:eq:dist_hyperplane}. To calculate the MLR in \cref{rmlr:eq:rmlr_v1}, one has to compute the associated Riemannian metrics, logarithmic maps, and margin distance.
The associated Riemannian metrics and logarithmic maps often have closed-form expressions on the frequently encountered manifolds in machine learning. 
However, the computation of the margin distance can be challenging. 
On the Poincar\'e ball of hyperbolic manifolds, the generalized law of sines simplifies the calculation of \cref{rmlr:eq:dist_hyperplane} \citep{ganea2018hyperbolic}.
However, the generalized law of sines is not universally guaranteed on other manifolds.
For pullback Euclidean metrics on SPD manifolds, \cref{spdmlr:lem:dist_to_hyperplane_pems} provides a closed-form solution for the margin distance. For curved manifolds, solving \cref{rmlr:eq:dist_hyperplane} would become a non-convex optimization problem.
To address this challenge, \citet{nguyen2023building} defined gyro-structures on the SPD manifold and proposed a pseudo-gyrodistance to calculate the margin distance.
Similarly, \citet{nguyen2024matrix} proposed a pseudo-gyrodistance on the SPSD manifold based on the gyro product space.
However, gyro-structures do not necessarily exist in general geometries.
\textit{In summary, the aforementioned methods often rely on specific properties of their associated Riemannian metrics, which usually do not generalize to general geometries.}

\subsubsection{Riemannian Multinomial Logistic Regression} \label{rmlr:subsec:general_RMLR}
Recalling \cref{rmlr:eq:rmlr_v1,rmlr:eq:r_hyperplane}, the minimum requirement for extending Euclidean MLR to manifolds is the well-definedness of $\rielog_{P_k}(S)$ for each $k$. In this subsection, we will develop Riemannian MLR, which depends solely on the Riemannian logarithm, without additional requirements, such as gyro-structures and the generalized law of sines.
In the following, we always assume the well-definedness of the Riemannian logarithm.
We start by reformulating the Euclidean margin distance to the hyperplane from a trigonometric perspective and then present our Riemannian MLR.

As we discussed before, obtaining the margin distance of \cref{rmlr:eq:dist_hyperplane} could be challenging.
Inspired by \citet{nguyen2023building}, we resort to the perspective of trigonometry to reinterpret Euclidean margin distance.
In Euclidean space, the margin distance is equivalent to
\begin{equation} \label{rmlr:eq:reform_e_dits_v2}
    d\left(x,H_{a,p}\right)=\sin\left(\angle xpy^*\right)d(x,p), \quad \text{with } y^*=\underset{y\in H_{a,p}\backslash\{p\}}{\argmax}\cos\angle xpy.
\end{equation}
We extend \cref{rmlr:eq:reform_e_dits_v2} to manifolds by the Riemannian trigonometry and geodesic distance, the counterparts of Euclidean trigonometry and distance.

\begin{parisdefinition}[Riemannian margin distance] \label{rmlr:def:riem_margin_dist}
   Let $\tilde{H}_{\tilde{A}, P}$ be a Riemannian hyperplane defined in \cref{rmlr:eq:r_hyperplane}, and $S \in \calM$. 
    The Riemannian margin distance from $S$ to $\tilde{H}_{\tilde{A}, P}$ is defined as
    \begin{equation} \label{rmlr:eq:margin_dist_reform_v2}
        d\left(S,\tilde{H}_{\tilde{A},P}\right)=\sin\left(\angle SPY^*\right)d(S,P),
    \end{equation}
    where $d(S, P)$ is the geodesic distance, and
    \begin{equation}
        Y^*=\underset{Y\in\tilde{H}_{\tilde{A},P}\backslash\{P\}}{\argmax}\cos\angle SPY.
    \end{equation}
    The initial velocities of geodesics define $\cos \angle SPY$:
    \begin{equation} \label{rmlr:eq:riem_trigo}
        \cos \angle SPY=\frac{\langle \rielog_P Y, \rielog_P S \rangle_P}{\|\rielog_P Y\|_P\|\rielog_P S\|_P},
    \end{equation}
    where  $\langle \cdot,\cdot \rangle_P$ is the Riemannian metric at $P$, and $\| \cdot \|_P$ is the associated norm.
\end{parisdefinition}
The Riemannian margin distance in \cref{rmlr:def:riem_margin_dist} has a closed-form expression.

\begin{paristheorem}
    \label{rmlr:thm:rie_margin_dist} 
    \linktoproof{rmlr:thm:rie_margin_dist}
    The Riemannian margin distance defined in \cref{rmlr:def:riem_margin_dist} is given by
    \begin{equation} \label{rmlr:eq:rie_margin_dist}
         d(S,\tilde{H}_{\tilde{A}, P}) 
        = \frac{|\langle  \rielog_P S, \tilde{A} \rangle _P|}{\| \tilde{A} \|_P}.
    \end{equation}
\end{paristheorem}
Putting \cref{rmlr:eq:rie_margin_dist} into \cref{rmlr:eq:rmlr_v1}, we can obtain a closed-form expression for Riemannian MLR.

\begin{paristheorem} [RMLR]
    \label{rmlr:thm:rmlr}
    \linktoproof{rmlr:thm:rmlr}
    Given a Riemannian manifold $(\calM,g)$, the Riemannian MLR induced by $g$ is
    \begin{equation} \label{rmlr:eq:rmlr_final}
        p(y=k \mid S \in \calM) \propto \exp \left( \langle \rielog_{P_k} S,  \tilde{A}_k \rangle_{P_k} \right),
    \end{equation}
    where $P_k \in \calM$, $\tilde{A}_k \in T_{P_k}\calM \backslash \{ \zerovec \}$,
    and $\rielog$ is the Riemannian logarithm.
\end{paristheorem}

As discussed for SPD MLRs in \cref{spdmlr:subsec:general_spd_mlr}, a tangent vector attached to a trainable base point can be generated from a Euclidean parameter in a fixed tangent space through Riemannian parallel transport, vector transport, or the differential of a group translation. Following the hyperbolic and gyro MLRs \citep{ganea2018hyperbolic,nguyen2023building}, we focus on parallel transport and Lie group translation:
\begin{align}
    \label{rmlr:eq:A_by_pt} 
    \tilde{A}_k &=\Gamma_{Q \rightarrow P_k} A_k,\\
    \label{rmlr:eq:A_by_lt} 
    \tilde{A}_k &= \left(L_{P_k \odot Q_{\odot}^{-1}}\right)_{*,Q} A_k,
\end{align}
where $Q \in \calM$ is a fixed point, $A_k \in T_Q\calM \backslash \{\zerovec\}$, $\Gamma$ is the parallel transport along the geodesic connecting $Q$ and $P_k$, and $\left(L_{P_k \odot Q_{\odot}^{-1}}\right)_{*,Q}$ denotes the differential map at $Q$ of left translation $L_{P_k \odot Q_{\odot}^{-1}}$ with $P_k \odot Q_{\odot}^{-1}$ denoting the Lie group product and inverse.
In this way, $A_k$ lies in a fixed tangent space and, therefore, can be optimized by a Euclidean optimizer.

\begin{parisremark}
    We make the following remarks regarding our Riemannian MLR.
    
    (a)
    The reformulations of \cref{rmlr:eq:reform_e_dits_v2} in gyro MLR \citep{nguyen2023building,nguyen2024matrix} and in our work are different.
    Gyro MLR adopts gyro trigonometry and gyro distance to reformulate \cref{rmlr:eq:reform_e_dits_v2}, while our method directly uses Riemannian trigonometry and geodesic distance.   
    
    (b)
    Compared with the hyperbolic, gyro SPD, and gyro SPSD MLRs \citep{ganea2018hyperbolic,nguyen2023building,nguyen2024matrix} and the SPD MLRs in \cref{sec:ch4-spd-classifiers}, our framework enjoys broader applicability, as our framework only requires the Riemannian logarithm. This property is commonly satisfied by most manifolds encountered in machine learning, such as the five metrics on SPD manifolds mentioned in \cref{sec:ch2-spd-manifolds}, the invariant metric on $\so{n}$ \citep{boumal2011discrete}, and the hyperbolic and spherical models in \cref{sec:ch2-constant-curvature-manifolds}.
    Besides, several existing MLRs on different geometries are special cases of our Riemannian MLR, which are detailed in \cref{rmlr:tab:mlr_as_ours_cases}.

    (c)
    The well-definedness of the Riemannian logarithm is a much weaker requirement compared to the existence of the gyro-structure. 
    The gyro-structure not only requires the Riemannian logarithm but also implicitly requires geodesic completeness \citep[Eqs.~(1)--(2)]{nguyen2023building}.
    For instance, on SPD manifolds, EM and BWM \citep{thanwerdas2023n} are incomplete, undermining the well-definedness of gyro operations. 
\end{parisremark}

\begin{table}[t]
\centering
\caption{Several MLRs on different geometries are special cases of our MLR.}
\label{rmlr:tab:mlr_as_ours_cases}
\resizebox{\linewidth}{!}{
\begin{tabular}{cccc}
\toprule
\textbf{MLR} & \textbf{Geometries} & \textbf{Requirements} & \makecell{\textbf{Incorporated} \\ \textbf{by Our MLR}}\\
\midrule
Euclidean MLR (\cref{spdmlr:eq:EMLR_reform_start}) & Euclidean geometry & \na & \cmark (\cref{rmlr:app:sec:rmlr_gen_emlr}) \\
\midrule
Gyro SPD MLRs \citep{nguyen2023building} & AIM, LEM \& LCM on $\spd{n}$   & Gyro-structures & \cmark (\cref{rmlr:rmk:extending_spd_mlr})  \\
\midrule
Gyro SPSD MLRs \citep{nguyen2024matrix} & SPSD product gyro spaces & \makecell{Gyro-structures} & \cmark (\cref{rmlr:app:sec:gyro_spsd_mlr_as_rmlr})   \\
\midrule
Flat SPD MLRs (\cref{sec:ch4-spd-classifiers}) & $(\alpha,\beta)$-LEM \& $(\theta)$-LCM on $\spd{n}$ & \makecell{Pullback metrics from \\ the Euclidean space} & \cmark (\cref{rmlr:rmk:extending_spd_mlr})  \\
\midrule
Ours & General geometries & Riemannian logarithm & \na \\
\bottomrule
\end{tabular}
}
\end{table}

\subsection{SPD Multinomial Logistic Regressions}
\label{rmlr:sec:spd_mlrs}

\begin{figure}[t]
\centering
\includegraphics[width=\linewidth,trim={0cm 5mm 0cm 0cm}]{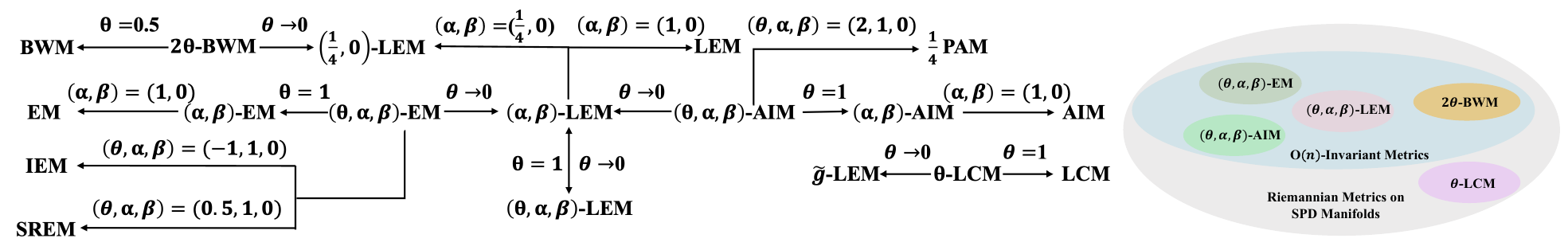}
\caption{Illustration of the deformation (\textbf{left}) and Venn diagram (\textbf{right}) of metrics on SPD manifolds, where IEM, SREM, and $\frac{1}{4}$~PAM denote Inverse Euclidean Metric, Square Root Euclidean Metric, and Polar Affine Metric scaled by $\nicefrac{1}{4}$, respectively.
}
\label{rmlr:fig:illustration_metrics}
\end{figure}

This section showcases our RMLR framework on the SPD manifold.
We first systematically discuss the power-deformed geometries of SPD manifolds.
Based on these metrics, we will develop five families of deformed SPD MLRs.

\subsubsection{Deformed Geometries of SPD Manifolds} \label{rmlr:subsec:geom_spd}

\begin{table}[t]
  \centering
  \caption{Properties of deformed metrics on SPD manifolds ($\theta \neq 0$ and $\min (\alpha, \alpha+n \beta)>0$).}
  \label{rmlr:tab:properties_rie_metrics}
  \resizebox{0.8\linewidth}{!}{
    \begin{tabular}{cc}
    \toprule
    \textbf{Name} & \textbf{Properties} \\
    \midrule
    $(\theta,\alpha,\beta)$-LEM &
    Bi-invariance, $\orth{n}$-invariance, Geodesic Completeness \\  
    \midrule
    $(\theta,\alpha,\beta)$-AIM &
    Lie Group Left-Invariance, $\orth{n}$-invariance, Geodesic Completeness \\
    \midrule
    $(\theta,\alpha,\beta)$-EM &
    $\orth{n}$-Invariance \\
    \midrule
    $\theta$-LCM &
    Lie Group Bi-Invariance, Geodesic Completeness\\
    \midrule
    $2\theta$-BWM &
    $\orth{n}$-Invariance \\
    \bottomrule    
    \end{tabular}
    }
\end{table}

As discussed in \cref{sec:ch2-spd-manifolds}, there are five popular Riemannian metrics on SPD manifolds. These metrics can all be extended to power-deformed metrics.
For a metric $g$ on $\spd{n}$, the power-deformed metric is defined as
\begin{equation} \label{rmlr:eq:pow_deform_meric}
    \tilde{g}_P \left(V,W \right) = \frac{1}{\theta^2} g_{P^\theta} \left( (\phi_\theta)_{*,P} (V), (\phi_\theta)_{*,P} (W)\right), \forall P \in \spd{n}, V,W \in T_P\spd{n},
\end{equation}
where $\phi_\theta(P)=P^\theta$ is the matrix power, and $(\phi_\theta)_{*,P}$ is the differential map.
The deformed metric $\tilde{g}$ can interpolate between a LEM-like metric ($\theta \rightarrow 0$) and $g$ ($\theta=1$) \citep{thanwerdas2022geometry}.
Previous work power-deformed $\biparamAIM$ and BWM into $\triparamAIM$ \citep{thanwerdas2019affine} and $\paramBWM$ \citep{thanwerdas2022geometry}, respectively. Earlier, we introduced $\triparamLEM$ and $\paramLCM$ in \cref{subsubsec:spd_param_lie_groups}. They are the power deformations of $\biparamLEM$ and LCM, respectively. By \cref{prop:spd_param_lem_lcm_deformation}, $\triparamLEM$ is equal to $\biparamLEM$. We therefore use $\biparamLEM$ for this family. Here, we further define the power deformation of $\biparamEM$ through \cref{rmlr:eq:pow_deform_meric}, denoted by $\triparamEM$.
We have the following result for $\triparamEM$.

\begin{parisproposition}
    \label{rmlr:prop:spd_parametrized_metrics}
    \linktoproof{rmlr:prop:spd_parametrized_metrics}
   $(\theta,\alpha,\beta)$-EM interpolates between $(\alpha,\beta)$-LEM ($\theta \rightarrow 0$) and $\biparamEM$ ($\theta=1$).
\end{parisproposition}

So far, all five popular Riemannian metrics on SPD manifolds have been generalized to power-deformed families of metrics.
We summarize their associated properties in \cref{rmlr:tab:properties_rie_metrics} and present their theoretical relation in \cref{rmlr:fig:illustration_metrics}. 
We leave technical details in \cref{rmlr:app:sec:deformed_metrics_theory}.

\subsubsection{Five Families of SPD Multinomial Logistic Regressions}

\begin{figure}[t]
\centering
\includegraphics[width=\linewidth,trim={0cm 2cm 0cm 0cm}]{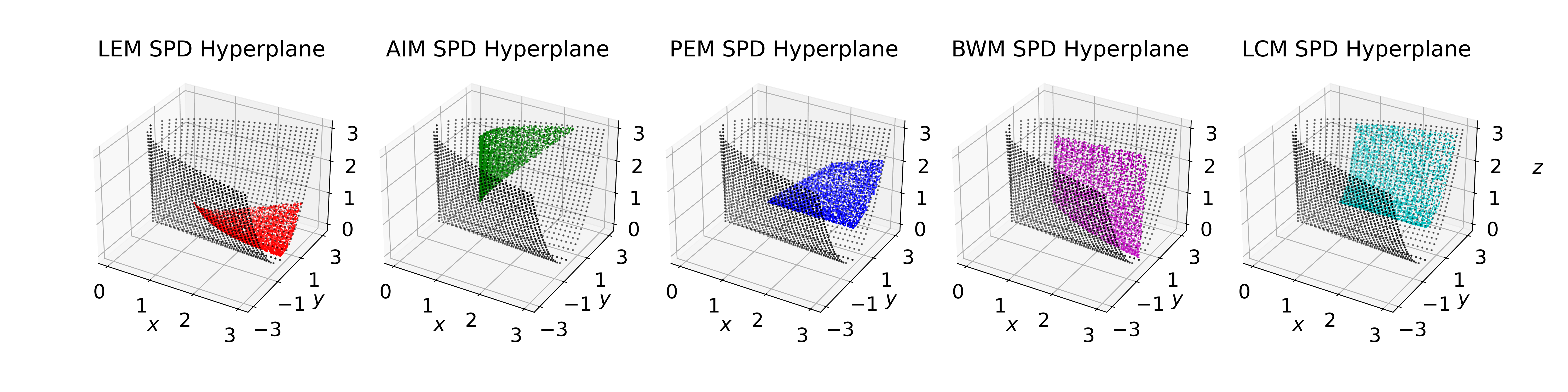}
\caption{Conceptual illustration of SPD hyperplanes induced by five families of Riemannian metrics.
The black dots denote the boundary of $\spd{2}$.
}
\label{rmlr:fig:hyperplanes}
\end{figure}

This subsection presents five families of specific SPD MLRs using our general framework in \cref{rmlr:thm:rmlr} and the metrics discussed in \cref{rmlr:subsec:geom_spd}.
We focus on generating $\tilde{A}_k$ by parallel transport from the identity matrix, except for $2\theta$-BWM.
Since the parallel transport under $2\theta$-BWM would undermine numerical stability (please refer to \cref{rmlr:app:sec:detail_bwm_spdmlr} for more details), we resort to a newly developed Lie group operation \citep{thanwerdas2022theoretically}:
\begin{equation}
    S_1 \odot S_2 = L_1 S_2 L_1^\top, \qquad \forall S_1, S_2 \in \spd{n},
\end{equation}
where $L_1=\chol(S_1)$ is the Cholesky factor.

\begin{paristheorem}[SPD MLRs]
    \label{rmlr:thm:spdmlrs} 
    \linktoproof{rmlr:thm:spdmlrs}
    By abuse of notation, we omit the subscripts $k$ of $A_k$ and $P_k$.
    Given an SPD feature $S$, the SPD MLRs, $p(y=k \mid S \in \spd{n})$, are proportional to
    \begin{align}
        \biparamLEM: 
        & \exp \left [ \langle \log(S)-\log(P), A \rangle^{\alphabeta} \right ], \\
        \triparamAIM:
        &\exp \left[ \frac{1}{\theta} \left\langle \log\left(P^{-\frac{\theta}{2}} S^\theta P^{-\frac{\theta}{2}}\right), A \right\rangle^{\alphabeta} \right], \\
        \triparamEM:
         &\exp \left[ \frac{1}{\theta} \langle S^\theta-P^\theta, A \rangle^{\alphabeta} \right], \\
        \paramLCM: 
        &\exp\left[\frac{1}{\theta}\left\langle
        \begin{aligned}
        &\lfloor\tilde{K}\rfloor-\lfloor\tilde{L}\rfloor+\left[\dlog(\bbD(\tilde{K}))-\dlog(\bbD(\tilde{L}))\right],\\
        &\lfloor A\rfloor+\frac{1}{2}\bbD(A)
        \end{aligned}
        \right\rangle\right], \\
        \label{rmlr:eq:bwm_spdmlr}
        \paramBWM:
        &\exp\left[\frac{1}{4\theta}\left\langle
        \begin{aligned}
        &(P^{2\theta}S^{2\theta})^{\frac{1}{2}}+(S^{2\theta}P^{2\theta})^{\frac{1}{2}}-2P^{2\theta},\\
        &\calL_{P^{2\theta}}\left(\bar{L}A\bar{L}^\top\right)
        \end{aligned}
        \right\rangle\right],
    \end{align}
    where $A \in T_I\spd{n} \backslash \{\bbzero\}$ is a symmetric matrix, $\log(\cdot)$ is the matrix logarithm, $\calL_P[V]$ is the solution to the matrix linear system $\calL_P[V] P+ P \calL_P[V]=V$, known as the Lyapunov operator,
    $\dlog(\cdot)$ is the diagonal element-wise logarithm, 
    $\lfloor \cdot \rfloor$ is the strictly lower part of a square matrix, and $\bbD(\cdot)$ is a diagonal matrix with diagonal elements of a square matrix.
    Besides, $\log_{*,P}$ is the differential map at $P$, $\tilde{K}= \chol(S^\theta)$, $\tilde{L}= \chol(P^\theta)$, and $\bar{L}=\chol(P^{2\theta})$.
\end{paristheorem}

The Lyapunov operator in \cref{rmlr:eq:bwm_spdmlr} requires the eigendecomposition.
However, the backpropagation of eigendecomposition involves $\nicefrac{1}{(\sigma_i-\sigma_j)}$ \citep{ionescu2015training}, undermining numerical stability.
Therefore, we propose a numerically stable backpropagation for the Lyapunov operator, detailed in \cref{rmlr:app:sec:detail_bwm_spdmlr}.

As $2 \times 2$ SPD matrices can be embedded into $\bbR{3}$ as an open cone \citep{yair2019parallel}, we illustrate SPD hyperplanes induced by five families of metrics in \cref{rmlr:fig:hyperplanes}.

\begin{parisremark} \label{rmlr:rmk:extending_spd_mlr}
    Our SPD MLRs extend the gyro SPD MLRs of \citet{nguyen2023building} and the flat SPD MLRs in \cref{sec:ch4-spd-classifiers}.
    The pseudo-gyrodistance to an SPD hyperplane in \citet[Thms.~2.23--2.25]{nguyen2023building} is incorporated by our \cref{rmlr:thm:rie_margin_dist}, while the flat SPD MLRs under $\biparamLEM$ and $\paramLCM$ in \cref{spdmlr:cor:spd_mlr_param_lem_lcm} are special cases of our \cref{rmlr:thm:spdmlrs}.
    Furthermore, our approach extends the scope of prior work because neither the framework in \cref{sec:ch4-spd-classifiers} nor the gyro SPD MLRs of \citet{nguyen2023building} cover SPD MLRs based on $\triparamEM$ and $\paramBWM$.
    The gyro operations in \citet[Eq.~(1)]{nguyen2023building} implicitly require geodesic completeness, whereas $\triparamEM$ and $\paramBWM$ are incomplete.
    As neither $\triparamEM$ nor $\paramBWM$ belong to pullback Euclidean metrics, the framework in \cref{spdmlr:thm:general_mlr_pems} cannot be applied to these metrics.
    To the best of our knowledge, our work is the \textbf{first} to apply PEM and BWM to establish Riemannian neural networks, opening up new possibilities for utilizing these metrics in machine learning applications. Besides, neither the gyro SPD MLRs of \citet{nguyen2023building} nor the flat SPD MLRs in \cref{sec:ch4-spd-classifiers} cover the deformed metrics for building SPD MLRs.
\end{parisremark}

\subsection{Lie Multinomial Logistic Regression}
\label{rmlr:sec:lie_mlr}

This section introduces our Lie MLR on $\so{n}$ based on the general RMLR framework in \cref{rmlr:thm:rmlr}.
The Riemannian metric on $\so{n}$ is assumed to be the invariant metric in \cref{tab:ch2-so-operators}.

We employ the vector transport on $\so{n}$ given by \citet[Tab.~1]{boumal2011discrete}, which coincides with the differential of left translation in \cref{rmlr:eq:A_by_lt}.

\begin{parislemma}
    \label{rmlr:lem:equi_lie_mlr}
    \linktoproof{rmlr:lem:equi_lie_mlr}
    \begin{equation}
        \vt{Q}{P}(H)
        = \left(L_{PQ^{-1}}\right)_{*,Q}(H)
        = PQ^{\top}H,
        \qquad \forall P,Q \in \so{n},\quad H\in T_Q\so{n}.
    \end{equation}
\end{parislemma}

Similar to SPD MLRs, we set $Q=I$. 
The Lie MLR on $\so{n}$ is presented in the following.

\begin{paristheorem}
    \label{rmlr:thm:lie_mlr}
    \linktoproof{rmlr:thm:lie_mlr}
    The Lie MLR on $\so{n}$ is given by
    \begin{equation}
        p(y=k \mid R \in \so{n}) \propto \exp\left(\left\langle\log\left(P_k^\top R\right),A_k\right\rangle\right),
    \end{equation}
    where $P_k \in \so{n}$ and $A_k \in \soLieAlgebra{n}$.
\end{paristheorem}

\begin{figure}[t]
    \centering
    \includegraphics[width=0.8\textwidth,trim={0cm 0cm 0cm 0cm}]{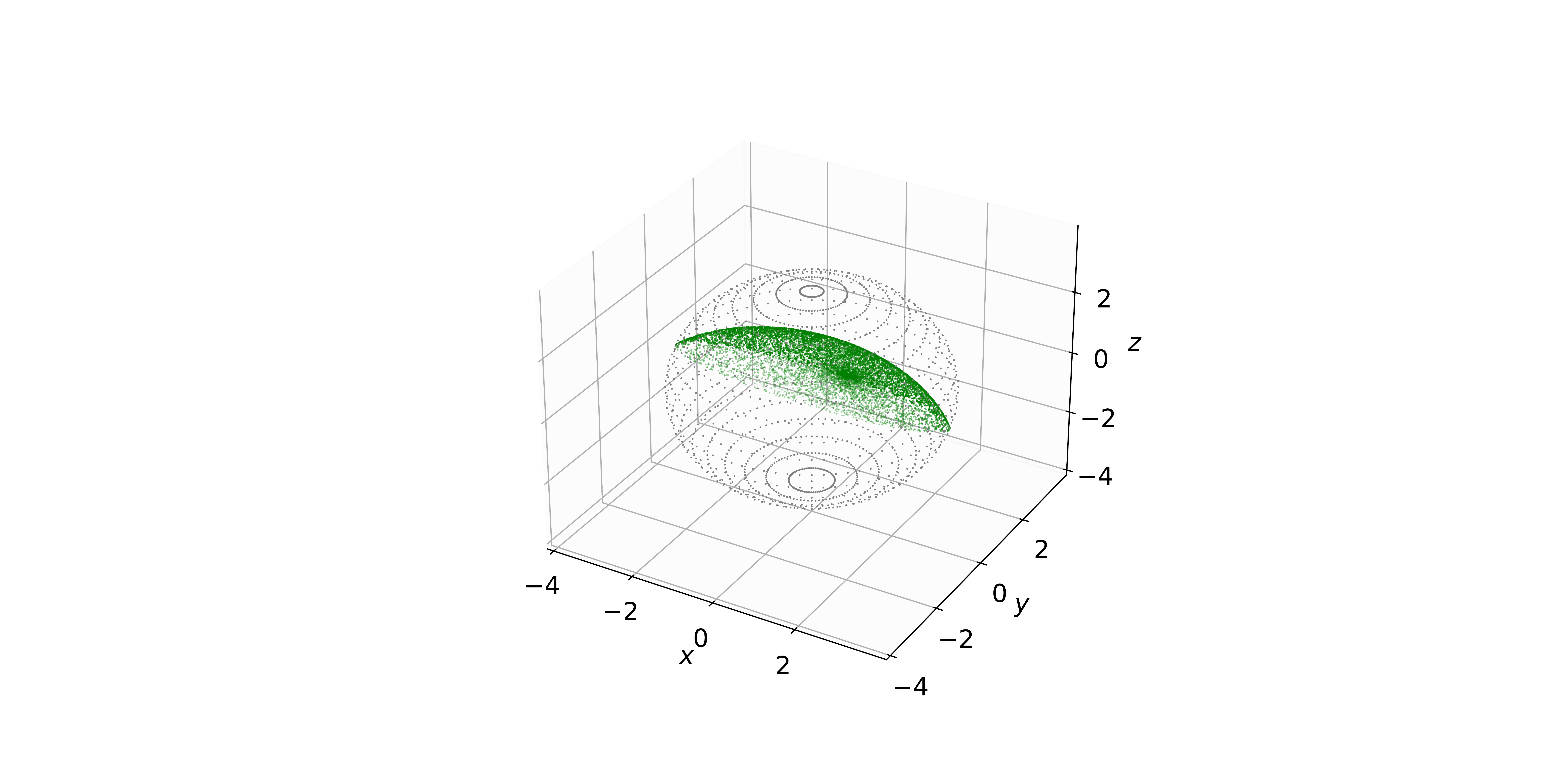}
    \caption{Conceptual illustration of a Lie hyperplane. Each pair of antipodal black dots corresponds to a rotation matrix with an Euler angle of $\pi$, while the green dots denote a Lie hyperplane.
    }
    \label{rmlr:fig:rot_hyperplane}
\end{figure}

We refer to the Riemannian hyperplanes (\cref{rmlr:eq:r_hyperplane}) on $\so{n}$ as Lie hyperplanes.
As $\so{3}$ is homeomorphic to 3-dimensional real projective space $\mathbb{RP}^3$ \citep{hartley2013rotation}, \cref{rmlr:fig:rot_hyperplane} illustrates Lie hyperplanes in the closed ball in $\bbR{3}$ of radius $\pi$.

\section{Experiments}
\label{rmlr:sec:experiments}

\begin{table}[H]
    \centering
    \caption{Comparison of SPDNet with LogEig against SPD MLRs on the Radar data set. The best results are \firstresults{bold}.}
    \label{rmlr:tb:results_radar}
    \resizebox{0.99\linewidth}{!}{
    \begin{tabular}{c|c|>{\columncolor{HilightColor}}c|*{2}{>{\columncolor{HilightColor}}c}|*{2}{>{\columncolor{HilightColor}}c}|*{2}{>{\columncolor{HilightColor}}c}|*{2}{>{\columncolor{HilightColor}}c}}
    \toprule
        \multirow{2}[4]{*}{\textbf{Architectures}} & \multirow{2}[4]{*}{\textbf{LogEig MLR}} & \textbf{$(\theta,\alpha,\beta)$-AIM} & \multicolumn{2}{>{\columncolor{HilightColor}}c|}{\textbf{$(\theta,\alpha,\beta)$-EM}} & \multicolumn{2}{>{\columncolor{HilightColor}}c|}{\textbf{$(\alpha,\beta)$-LEM}} & \multicolumn{2}{>{\columncolor{HilightColor}}c|}{\textbf{$2\theta$-BWM}} & \multicolumn{2}{>{\columncolor{HilightColor}}c}{\textbf{$\theta$-LCM}} \\
    \cmidrule{3-11}          &       & (1,1,0) & (1,1,0) & (1,1,$\nicefrac{1}{8}$) & (1,1,0) & (1,1,1) & (0.5) & (0.25) & (1)   & (0.5) \\
    \midrule
    2-Block & 92.88$\pm$1.05 & \firstresults{94.53$\pm$0.95} & 94.24$\pm$0.55 & \firstresults{94.93$\pm$0.60} & 93.55$\pm$1.21 & \firstresults{95.64$\pm$0.83} & 92.22$\pm$0.83 & \firstresults{94.99$\pm$0.47} & 93.49$\pm$1.25 & \firstresults{94.59$\pm$0.82} \\
    5-Block & 93.47$\pm$0.45 & \firstresults{94.32$\pm$0.94} & \firstresults{95.11$\pm$0.82} & 95.01$\pm$0.84 & 94.60$\pm$0.70 & \firstresults{95.87$\pm$0.58} & 93.69$\pm$0.66 & \firstresults{94.84$\pm$0.68} & 93.93$\pm$0.98 & \firstresults{95.16$\pm$0.67} \\
    \bottomrule
    \end{tabular}%
    }
\end{table}%

\begin{table}[H]
    \centering
    \caption{Comparison of SPDNet with LogEig against SPD MLRs on the HDM05 data set.}
    \label{rmlr:tb:results_hdm05}
    \resizebox{0.99\linewidth}{!}{
    \begin{tabular}{c|c|>{\columncolor{HilightColor}}c|*{2}{>{\columncolor{HilightColor}}c}|>{\columncolor{HilightColor}}c|>{\columncolor{HilightColor}}c|*{2}{>{\columncolor{HilightColor}}c}}
    \toprule
    \multirow{2}[4]{*}{\textbf{Architectures}} & \multirow{2}[4]{*}{\textbf{LogEig MLR}} & \textbf{$(\theta,\alpha,\beta)$-AIM} & \multicolumn{2}{>{\columncolor{HilightColor}}c|}{\textbf{$(\theta,\alpha,\beta)$-EM}} & \textbf{$(\alpha,\beta)$-LEM} & \textbf{$2\theta$-BWM} & \multicolumn{2}{>{\columncolor{HilightColor}}c}{\textbf{$\theta$-LCM}} \\
    \cmidrule{3-9}          &       & {(1,1,0)} & (1,1,0) & (0.5,1.0,$\nicefrac{1}{30}$) & (1,1,0) & (0.5) & (1) & (0.5) \\
    \midrule
    1-Block & 57.42$\pm$1.31 & 58.07$\pm$0.64  & 66.32$\pm$0.63 & \firstresults{71.65$\pm$0.88} &  56.97$\pm$0.61 & \firstresults{70.24$\pm$0.92} &  63.84$\pm$1.31 &	\firstresults{65.66$\pm$0.73} \\
    2-Block & 60.69$\pm$0.66 &  60.72$\pm$0.62 & 66.40$\pm$0.87 & \firstresults{70.56$\pm$0.39} &  60.69$\pm$1.02 & \firstresults{70.46$\pm$0.71} & 62.61$\pm$1.46 &	\firstresults{65.79$\pm$0.63} \\
    3-Block & 60.76$\pm$0.80 & 61.14$\pm$0.94 & 66.70$\pm$1.26 & \firstresults{70.22$\pm$0.81} &  60.28$\pm$0.91 & \firstresults{70.20$\pm$0.91} &  62.33$\pm$2.15 & \firstresults{65.71$\pm$0.75} \\
    \bottomrule
    \end{tabular}%
    }
\end{table}

\begin{table}[H]
    \centering
    \caption[Inter-session TSMNet results on Hinss2021.]{Inter-session experiments of TSMNet with different MLRs on the Hinss2021 data set.}
    \label{rmlr:tb:results_hinss_inter_session}
    \resizebox{0.99\linewidth}{!}{
    \begin{tabular}{c|c|*{2}{>{\columncolor{HilightColor}}c}|>{\columncolor{HilightColor}}c|>{\columncolor{HilightColor}}c|>{\columncolor{HilightColor}}c|*{2}{>{\columncolor{HilightColor}}c}}
    \toprule
    \multirow{2}[4]{*}{\textbf{Classifiers}} & \multirow{2}[4]{*}{\textbf{LogEig MLR}} & \multicolumn{2}{>{\columncolor{HilightColor}}c|}{\textbf{$(\theta,\alpha,\beta)$-AIM}} & \textbf{$(\theta,\alpha,\beta)$-EM} & \textbf{$(\alpha,\beta)$-LEM} & \textbf{$2\theta$-BWM} & \multicolumn{2}{>{\columncolor{HilightColor}}c}{\textbf{$\theta$-LCM}} \\
    \cmidrule{3-9}          &       & (1,1,0) & (0.5,1,0.05) & (1,1,0) & (1,1,0) & (0.5) & (1) & (1.5) \\
    \midrule
    Balanced Acc. & 53.83$\pm$9.77 &  53.36$\pm$9.92 & \firstresults{55.27$\pm$8.68} & \firstresults{54.48$\pm$9.21} &  53.51$\pm$10.02 & \firstresults{55.54$\pm$7.45} &  55.71$\pm$8.57 & \firstresults{56.43$\pm$8.79} \\
    \bottomrule
    \end{tabular}
    }
\end{table}%

\begin{table}[H]
    \centering
    \caption[Inter-subject TSMNet results on Hinss2021.]{Inter-subject experiments of TSMNet with different MLRs on the Hinss2021 data set.}
    \label{rmlr:tb:results_hinss_inter_subject}
    \resizebox{0.99\linewidth}{!}{
    \begin{tabular}{c|c|*{2}{>{\columncolor{HilightColor}}c}|*{2}{>{\columncolor{HilightColor}}c}|>{\columncolor{HilightColor}}c|*{2}{>{\columncolor{HilightColor}}c}|*{2}{>{\columncolor{HilightColor}}c}}
    \toprule
    \multirow{2}[4]{*}{\textbf{Classifiers}} & \multirow{2}[4]{*}{\textbf{LogEig MLR}} & \multicolumn{2}{>{\columncolor{HilightColor}}c|}{\textbf{$(\theta,\alpha,\beta)$-AIM}} & \multicolumn{2}{>{\columncolor{HilightColor}}c|}{\textbf{$(\theta,\alpha,\beta)$-EM}} & \textbf{$(\alpha,\beta)$-LEM} & \multicolumn{2}{>{\columncolor{HilightColor}}c|}{\textbf{$2\theta$-BWM}} & \multicolumn{2}{>{\columncolor{HilightColor}}c}{\textbf{$\theta$-LCM}} \\
    \cmidrule{3-11}          &       & (1,1,0) & (1.5,1,0) & (1,1,0) & (1.5,1,$\nicefrac{1}{20}$) & (1,1,0) & (0.5) & (0.75) & (1)   & (0.5) \\
    \midrule
    Balanced Acc. & 49.68$\pm$7.88 &  50.65$\pm$8.13 & \firstresults{51.15$\pm$7.83} &  50.02$\pm$5.81 & \firstresults{51.38$\pm$5.77} & \firstresults{51.41$\pm$7.98} &  50.26$\pm$7.23 & \firstresults{51.67$\pm$8.73} &  52.93$\pm$7.76 & \firstresults{54.14$\pm$8.36} \\
    \bottomrule
    \end{tabular}%
    }
\end{table}%

We first instantiate our SPD MLRs in four SPD neural networks: SPDNet \citep{huang2017riemannian} and TSMNet \citep{kobler2022spd} for Riemannian feedforward networks, RResNet \citep{katsman2023riemannian} for Riemannian residual networks, and SPDGCN \citep{zhao2023modeling} for Riemannian graph neural networks.
Then, we proceed with experiments of our Lie MLR under the classic LieNet architecture \citep{huang2017deep}.
The classifier in all the above networks is the LogEig MLR (matrix logarithm + FC + softmax), a Euclidean MLR on the tangent space at the identity matrix.
We substitute the original non-intrinsic LogEig MLR in each baseline model with our RMLRs.
Notably, the gyro SPD MLRs \citep{nguyen2023building} are special cases of our SPD MLRs under the standard AIM, LEM, and LCM ($(\theta,\alpha,\beta)=(1,1,0)$), while the flat SPD MLRs in \cref{sec:ch4-spd-classifiers} are incorporated by our SPD MLRs under $\biparamLEM$ and $\paramLCM$.
More details on data sets and experimental settings are provided in \cref{app:datasets,rmlr:app:experimental-details}.

\subsection{Experiments on the Proposed SPD MLRs}
In the following, we abbreviate \emph{SPD MLR-metric} as \emph{metric}.
For instance, $\triparamAIM$ denotes the baseline endowed with the SPD MLR induced by $\triparamAIM$, with $(1,1,0)$ as the value of $(\theta,\alpha,\beta)$.

\mypara{Experiments on the Riemannian feedforward network.} We evaluate our SPD MLRs for Riemannian feedforward networks under the SPDNet and TSMNet backbones.
Following \citet{huang2017riemannian,brooks2019riemannian}, on SPDNet, we use the Radar data set \citep{brooks2019riemannian} for radar recognition and the HDM05 data set \citep{muller2007documentation} for human action recognition.
TSMNet \citep{kobler2022spd} is one of the state-of-the-art methods for the EEG classification task.
Following \citet{kobler2022spd}, we use the Hinss2021 \citep{hinss2021eegdata} data set.
For each family of SPD MLRs, we report the SPD MLR induced by the standard metric ($\theta=1,\alpha=1,\beta=0$) and the one induced by the deformed metric with the best $(\theta,\alpha,\beta)$.
Besides, if the standard SPD MLR is already saturated, we only report the results of the standard one.
Under each metric, we highlight the results of our SPD MLR under the best hyperparameters in bold.

\begin{enumerate}
    \item \mypara{Radar.} In line with \citet{brooks2019riemannian}, we evaluate our classifiers under two network architectures: 2-Block and 5-Block configurations.
    The 10-fold results (mean$\pm$std) are presented in \cref{rmlr:tb:results_radar}.
    Note that the SPD MLR induced by standard AIM is saturated.
    Generally speaking, our SPD MLRs achieve superior performance against the vanilla LogEig MLR.
    Moreover, for most families of metrics, the associated SPD MLRs with proper $(\theta,\alpha,\beta)$ outperform the standard SPD MLR, demonstrating the effectiveness of our parameterization.
    Besides, among all SPD MLRs, the ones induced by $\biparamLEM$ achieve the best performance.

    \item \mypara{HDM05.} Following \citet{huang2017riemannian}, three architectures are adopted: 1-Block, 2-Block and 3-Block configurations.
    The 10-fold results (mean$\pm$std) are presented in \cref{rmlr:tb:results_hdm05}.
    Note that the standard SPD MLRs under AIM, LEM, and BWM are already saturated on this data set.
    As on the Radar data set, similar observations can be made on this data set.
    Our SPD MLRs can bring consistent performance gains for SPDNet, and properly selected hyperparameters can bring further improvement.
    Particularly, among all the SPD MLRs, the ones based on the $\paramBWM$ and $\triparamEM$ achieve the best performance.
    Compared to the vanilla LogEig MLR, \textbf{the highest performance improvement is 14.23 percentage points}, highlighting our approach's effectiveness.
    Notably, since $\paramBWM$ and $\triparamEM$ are geodesically incomplete and not pulled back from a Euclidean space, the SPD MLR under these two metrics cannot be derived by the framework of gyro or flat MLR.
    This contrast confirms the applicability of our theoretical framework to a broader range of geometries.

    \item \mypara{Hinss2021.} The results (mean$\pm$std) of leave-5\%-out cross-validation are reported in \cref{rmlr:tb:results_hinss_inter_session,rmlr:tb:results_hinss_inter_subject}.
    Once again, our intrinsic classifiers demonstrate improved performance compared to the LogEig MLR in both inter-session and inter-subject scenarios.
    Besides, the SPD MLRs based on $\theta$-LCM achieve the best performance, \textbf{outperforming the vanilla classifier by 2.60 percentage points for inter-session and by 4.46 percentage points for inter-subject}.
    This finding highlights the versatility of our framework.
\end{enumerate}

\begin{table}[t]
    \centering
    \caption{Comparison of LogEig against SPD MLRs under the RResNet architecture.}
    \label{rmlr:tab:results_rresnet}
    \resizebox{0.99\linewidth}{!}{
        \begin{tabular}{cc*{5}{>{\columncolor{HilightColor}}c}}
        \toprule
         \textbf{Data Sets} & \textbf{LogEig MLR} & \textbf{$\triparamAIM$} & \textbf{$\triparamEM$} & \textbf{$\biparamLEM$} & \textbf{$\paramBWM$} & \textbf{$\paramLCM$} \\
        \midrule
        HDM05 & 58.17 $\pm$ 2.07 & 60.23 $\pm$ 1.26 & \firstresults{71.89 $\pm$ 0.60 ($\uparrow$ 13.72)} & 59.44 $\pm$ 0.87 & 69.85 $\pm$ 0.23 & 65.76 $\pm$ 0.96 \\
        NTU60 & 45.22 $\pm$ 1.23 & 48.94 $\pm$ 0.68 & 52.24 $\pm$ 1.25 & 46.99 $\pm$ 0.41 & 50.56 $\pm$ 0.59 & \firstresults{53.63 $\pm$ 0.95 ($\uparrow$ 8.41)} \\
        \bottomrule
        \end{tabular}
    }
\end{table}

\mypara{Experiments on the Riemannian residual network.} Following \citet{katsman2023riemannian}, we use the HDM05 and NTU60 \citep{shahroudy2016ntu} data sets on the RResNet backbone.
For the hyperparameter $(\theta,\alpha,\beta)$ in our SPD MLRs, we borrow the best ones from \cref{rmlr:tb:results_hdm05}.
\cref{rmlr:tab:results_rresnet} reports the 10-fold and 5-fold results on the HDM05 and NTU60 data sets, respectively.
The SPD MLRs still consistently outperform the vanilla LogEig MLR. Besides, similar to the SPD MLRs under the SPDNet backbone for action recognition (\cref{rmlr:tb:results_hdm05}), the SPD MLR based on $\paramLCM$, $\paramBWM$, or $\triparamEM$ outperforms the vanilla LogEig MLR by a large margin.
In particular, \textbf{the highest performance improvements are 13.72 and 8.41 percentage points} on these two data sets.

\begin{table}[t]
\centering
\caption{Comparison of LogEig against SPD MLRs under the SPDGCN architecture.}
\label{rmlr:tab:results_spdgcn}
\resizebox{0.8\linewidth}{!}{
\begin{tabular}{ccccccc}
    \toprule
    \multirow{2}[0]{*}{\textbf{Classifiers}} & \multicolumn{2}{c}{\textbf{Disease}} & \multicolumn{2}{c}{\textbf{Cora}} & \multicolumn{2}{c}{\textbf{Pubmed}} \\
    \cmidrule{2-7} 
          & \textbf{Mean$\pm$STD} & \textbf{Max} & \textbf{Mean$\pm$STD} & \textbf{Max} & \textbf{Mean$\pm$STD} & \textbf{Max} \\
    \midrule
    LogEig MLR & 90.55 $\pm$ 4.83 & 96.85 & 78.04 $\pm$ 1.27 & 79.6 & 70.99 $\pm$ 5.12 & 77.6 \\
    \midrule
    \rowcolor{HilightColor} $\triparamAIM$ & 94.84 $\pm$ 2.27 & 98.43 & 79.79 $\pm$ 1.44 & 81.6 & 77.83 $\pm$ 1.08 & \firstresults{80} \\
    \rowcolor{HilightColor} $\triparamEM$ & 90.87 $\pm$ 5.14 & 98.03 & 79.05 $\pm$ 1.23 & 81 & 78.16 $\pm$ 2.41 & 79.5 \\
    \rowcolor{HilightColor} $\biparamLEM$ & \firstresults{96.33 $\pm$ 2.19} & \firstresults{98.82} & \firstresults{79.89 $\pm$ 0.99} & \firstresults{81.8} & \firstresults{78.16 $\pm$ 2.41} & 79.5 \\
    \rowcolor{HilightColor} $\paramBWM$ & 91.93 $\pm$ 3.64 & 96.85 & 73.46 $\pm$ 2.18 & 77.7 & 73.22 $\pm$ 4.06 & 78.1 \\
    \rowcolor{HilightColor} $\paramLCM$ & 93.01 $\pm$ 2.14 & 98.43 & 77.59 $\pm$ 1.20 & 80.1 & 74.46 $\pm$ 5.81 & 78.9 \\
    \bottomrule
\end{tabular}
}
\end{table}

\mypara{Experiments on the Riemannian graph network.} We use SPDGCN \citep{zhao2023modeling} as the backbone network for the Riemannian graph network.
Following \citet{zhao2023modeling}, we use the Disease \citep{anderson1991infectious}, Cora \citep{sen2008collective}, and Pubmed \citep{namata2012query} data sets for node classification.
The 10-fold average and maximum results of the vanilla LogEig MLR against our SPD MLR with the best $(\theta,\alpha,\beta)$ are reported in \cref{rmlr:tab:results_spdgcn}.
Similar to the previous results, our SPD MLRs generally outperform the LogEig MLR. Besides, the SPD MLR based on $\biparamLEM$ generally achieves the best performance for SPDGCN.

\begin{table}[t]
\centering
\caption{Comparison of LogEig against SPD MLRs for direct classification.}
\label{rmlr:tab:results_direct_classification}
\resizebox{0.99\linewidth}{!}{
\begin{tabular}{ccccc}
    \toprule
    \multirow{2}[0]{*}{\textbf{Classifiers}} & \multirow{2}[0]{*}{\textbf{Radar}} & \multirow{2}[0]{*}{\textbf{HDM05}} & \multicolumn{2}{c}{\textbf{Hinss2021}} \\
          &       &       & \textbf{Inter-session} & \textbf{Inter-subject} \\
    \midrule
    LogEig MLR & 91.93 $\pm$ 1.30 & 48.43 $\pm$ 1.25 & 39.76 $\pm$ 7.60 & 44.66 $\pm$ 7.17 \\
    \midrule
    \rowcolor{HilightColor} $\triparamAIM$ & 95.21 $\pm$ 0.81 & 49.17 $\pm$ 1.08 & 41.14 $\pm$ 7.26 & 45.89 $\pm$ 6.52 \\
    \rowcolor{HilightColor} $\triparamEM$ & 92.25 $\pm$ 1.20 & 61.60 $\pm$ 0.69 & \firstresults{45.78 $\pm$ 8.51 ($\uparrow$ 6.02)} & 45.84 $\pm$ 4.75 \\
    \rowcolor{HilightColor} $\biparamLEM$ & 95.09 $\pm$ 0.57 & 49.05 $\pm$ 0.91 & 40.88 $\pm$ 7.46 & \firstresults{46.02 $\pm$ 5.96 ($\uparrow$ 1.36)} \\
    \rowcolor{HilightColor} $\paramBWM$ & 94.89 $\pm$ 0.41 & \firstresults{66.77 $\pm$ 1.34 ($\uparrow$ 18.34)} & 44.84 $\pm$ 8.00 & 45.21 $\pm$ 7.44 \\
    \rowcolor{HilightColor} $\paramLCM$ & \firstresults{95.67 $\pm$ 0.61 ($\uparrow$ 3.74)} & 58.66 $\pm$ 0.51 & 43.17 $\pm$ 6.21 & 45.10 $\pm$ 6.20 \\
    \bottomrule
\end{tabular}
}
\end{table}
\mypara{Ablations of SPD MLRs on direct classification.} For a more straightforward comparison, we compare LogEig against our SPD MLRs for direct classification.
We adopt the Radar, HDM05, and Hinss2021 data sets.
We follow the preprocessing of SPDNet and TSMNet to model features into the SPD manifold and directly use LogEig or our SPD MLRs for classification.
The average results are presented in \cref{rmlr:tab:results_direct_classification}.
The hyperparameters $(\theta,\alpha,\beta)$ are borrowed from \cref{rmlr:tb:results_radar,rmlr:tb:results_hdm05,rmlr:tb:results_hinss_inter_session,rmlr:tb:results_hinss_inter_subject}.
Our SPD MLRs consistently outperform the vanilla LogEig MLR.
In particular, on the HDM05 data set, \textbf{the highest performance improvement by our SPD MLRs is 18.34 percentage points}, surpassing the non-intrinsic LogEig MLR by a large margin.

\subsection{Experiments on the Proposed Lie MLR}

\begin{table}[t]
    \centering
    \caption{Results of LogEig MLR against Lie MLR under the LieNet architecture.}
    \label{rmlr:tab:results_lie_mlr}%
    \resizebox{0.75\linewidth}{!}{
    \begin{tabular}{ccccc}
    \toprule
    \multirow{2}[0]{*}{\textbf{Classifiers}} & \multicolumn{2}{c}{\textbf{G3D}} & \multicolumn{2}{c}{\textbf{HDM05}} \\
    \cmidrule{2-5}          & \textbf{Mean$\pm$STD} & \textbf{Max} & \textbf{Mean$\pm$STD} & \textbf{Max} \\
    \midrule
    LogEig MLR & 87.91$\pm$0.90 & 89.73 & 76.92$\pm$1.27 & 79.11 \\
    \midrule
    \rowcolor{HilightColor} Lie MLR & \firstresults{89.13$\pm$1.7} & \firstresults{92.12} & \firstresults{78.24$\pm$1.03} & \firstresults{80.25} \\
    \bottomrule
    \end{tabular}
    }
\end{table}%

We apply our Lie MLR to the classic $\so{n}$ network, \ie LieNet \citep{huang2017deep}, where features are on the Lie group of $\so{3} \times \cdots \times \so{3}$.
More precisely, this feature space is a product manifold of $\so{3}$ factors, and \cref{rmlr:thm:lie_mlr} extends naturally under the product metric, with each class logit obtained by summing the factorwise inner products.
Following LieNet \citep{huang2017deep}, we use G3D \citep{bloom2012g3d} and HDM05 \citep{muller2007documentation} data sets.
We also extend the Riemannian optimization package \texttt{Geoopt} \citep{kochurov2020geoopt} to $\so{3}$, allowing for the direct Riemannian optimization reviewed in \cref{sec:ch2-riemannian-optimization}.
We find that RSGD performs best for LieNet.
\cref{rmlr:tab:results_lie_mlr} presents the 10-fold average results of LieNet with or without Lie MLR.
Note that on the HDM05 data set, LieNet might fail to converge, with the validation accuracy fluctuating between 70\% and 75\%.
Therefore, we select the 10 best-performing folds out of 20 experimental folds.
It can be observed that our Lie MLR can improve the performance of LieNet.
Besides, our Lie MLR can also improve the training stability.
On the HDM05 data set, LieNet fails to converge in 8 out of 20 folds.
However, when endowed with our Lie MLR, LieNet+LieMLR only encounters convergence failures in 2 folds.

\clearpage
\section{Conclusion}
\label{sec:ch4-conclusion}
This chapter developed a unified approach to intrinsic classification in two stages, progressing from a structured family of flat SPD geometries to general Riemannian manifolds. The first part considered SPD manifolds endowed with pullback Euclidean metrics. Their flat geometry reduces the infimum defining the geodesic distance from an SPD point to a margin hyperplane to a Euclidean point-to-hyperplane problem. This yields a closed-form margin distance and, consequently, a unified construction of SPD MLR. We instantiated this construction under deformed LEM and LCM and showed that, under the corresponding optimization scheme, its LEM instance recovers the widely used LogEig classifier, thereby providing an intrinsic interpretation of the existing pipeline.

The second part addressed the central obstacle to extending this construction beyond flat geometries. On a general Riemannian manifold, evaluating the point-to-hyperplane distance through its infimum can require solving a difficult, potentially non-convex optimization problem and may not admit a closed-form solution. Instead of solving this minimization problem, we replaced the infimum-based margin formulation with a Riemannian-trigonometric one that combines the geodesic distance from an input to the hyperplane anchor with the angle between the corresponding geodesics. This reformulation yields a closed-form RMLR that requires only a well-defined Riemannian logarithm, extending the classification principle from flat SPD geometries to a broad range of Riemannian manifolds.

We instantiated the general framework as five families of SPD MLRs under power-deformed metrics and as a Lie MLR on $\so{n}$. Experiments across Riemannian feedforward, residual, graph, and Lie-group networks demonstrated the broad applicability of the framework.

    \chapter{Riemannian Neural Networks}
\label{chapter:riemannian-neural-networks}

\section{Introduction}
\label{sec:ch5-introduction}

The preceding chapters developed two fundamental network modules through unified geometric formulations that can be instantiated across different manifolds. Such unified constructions make essential modules reusable across manifold families, but not every neural component admits a sufficiently tractable or effective formulation based only on broadly shared Riemannian properties. The general RMLR in \cref{rmlr:subsec:general_RMLR}, for example, achieves broad applicability by replacing the potentially intractable point-to-hyperplane infimum with a Riemannian-trigonometric formulation that requires only a well-defined Riemannian logarithm. In contrast, hyperbolic and flat correlation geometries permit exact evaluation of the corresponding point-to-hyperplane infima, while Busemann functions and horospheres provide an alternative hyperbolic decision principle. These examples illustrate how additional geometric or algebraic structure of a particular manifold can support more direct and better-tailored modules and architectures.

This chapter therefore studies manifold-specific Riemannian network design through three complementary routes. In \cref{sec:ch5-pvnn}, we introduce the unconstrained Proper Velocity (PV) model, establish its Riemannian toolkit, and construct MLR, fully connected, convolutional, activation, and normalization layers. In \cref{sec:ch5-hbnn}, we exploit Busemann functions and horospheres to derive intrinsic and batch-efficient BMLR and BFC layers for the Poincar\'e and Lorentz models. Finally, \cref{sec:ch5-cornet} exploits the specific geometries of full-rank correlation manifolds to construct MLR, fully connected, and convolutional layers together with Riemannian backpropagation.

\section{Proper Velocity Neural Networks}
\label{sec:ch5-pvnn}

\subsection{Introduction}
\label{pvnn:sec:intro}
Hyperbolic representations have recently delivered strong performance across different applications because the exponential volume growth of negatively curved manifolds enables low-distortion embeddings of tree-like and hierarchical structure \citep{nickel2017poincare}. These advantages have been validated in computer vision \citep{gao2021curvature,khrulkov2020hyperbolic,ermolov2022hyperbolic,van2023poincare,gao2023exploring,bdeir2024fully,he2025lorentzian,bdeir2025robust,sur2025hyperbolic,liu2025hyperbolic,wang2026wasserstein}, graph learning \citep{chami2019hyperbolic,bachmann2020constant,fu2024hyperbolic,sun2024geometry}, multimodal learning \citep{desai2023hyperbolic,pal2025compositional}, recommendation systems \citep{yanghg2025former}, astronomy \citep{chen2025galaxy}, genome sequence learning \citep{khan2025hyperbolic}, natural language processing \citep{nickel2017poincare,ganea2018hyperbolic,nickel2018learning,gulcehre2019hyperbolic,he2025helm,yang2025hyperbolic}, and brain signal decoding \citep{li2026heegnet}. Recently, the focus has shifted from hyperbolic embeddings to building HNNs that operate entirely within hyperbolic space. As reviewed in \cref{sec:ch2-constant-curvature-manifolds}, hyperbolic geometry admits multiple models, so the choice of representation is central to the design of hyperbolic networks. Most recent works rely on the Poincar\'e ball and Lorentz models, which provide convenient Riemannian or gyrovector structures, thereby facilitating neural network construction. However, both models are constrained spaces, which can lead to numerical instabilities. In particular, as embeddings in the Poincar\'e ball approach the boundary, numerical computations become unstable and might cause gradients to vanish~\citep{guo2022clipped}.

On the other hand, the \emph{Proper Velocity (PV) model} originates from Einstein's special relativity, where proper velocity provides a natural parameterization for relativistic velocity addition~\citep[Ch.~10]{ungar2022analytic}. Algebraically, PV admits a gyrovector space~\citep[Ch.~6]{ungar2022analytic}, analogous to the M\"obius gyrovector space of the Poincar\'e ball. Unlike the constrained Poincar\'e ball and Lorentz models, PV offers an unconstrained representation that alleviates numerical instabilities. These properties have made the PV model successful in relativistic physics and motivate its exploration as a stable alternative geometry for HNNs. However, its Riemannian operators, including exponential and logarithmic maps and parallel transport, remain largely unexplored, despite being fundamental for constructing neural networks.

Inspired by the above discussions, we propose \emph{Proper Velocity Neural Networks (PVNNs)}. To this end, we first establish the complete Riemannian geometry of PV by deriving closed-form expressions for the exponential map, logarithmic map, geodesic distance, and parallel transport. Building on this foundation, we extend several fundamental neural layers into PV space, including MLR classification, FC, convolutional, activation, and BN layers. Together, these layers form a complete PVNN framework from which different network architectures can be constructed. We validate the framework through four sets of experiments, including numerical stability, image classification, graph learning, and genomic sequence learning, demonstrating both the stability of PV embeddings and effectiveness of PVNNs. To our knowledge, the PV model has remained largely unexplored in machine learning, and our work provides the first systematic study of its use for representation learning. In summary, our \textbf{contributions} are threefold:
\begin{enumerate}
    \item We establish the complete Riemannian geometric toolkit of the PV manifold, deriving closed-form operators that enable its use as a new alternative to classical hyperbolic models.  
    \item We develop fundamental building blocks in PV space, including MLR, FC, convolutional, activation, and BN layers.
    \item We validate the stability and effectiveness of PVNNs through experiments on four tasks: numerical stability, image classification, graph node classification, and genomic sequence learning.\footnote{The code is available at \url{https://github.com/NickyoyoSu/PVNN}.}
\end{enumerate}

\mypara{Outline.} In \cref{pvnn:sec:preliminaries}, we introduce the PV model and its gyrovector operations. In \cref{pvnn:sec:pv-geometry}, we develop the Riemannian geometry and closed-form operators of PV space. In \cref{pvnn:sec:pv-layers}, we construct the core layers of PVNNs. In \cref{pvnn:sec:pv-and-lorentz}, we connect PV constructions to Lorentz neural layers, and in \cref{pvnn:sec:experiments}, we evaluate their numerical stability and effectiveness. Proofs are deferred to \cref{app:pvnn-proofs}.

\subsection{Preliminaries}
\label{pvnn:sec:preliminaries}
\mypara{PV Space \citep{ungar2022analytic}.} As shown in \cref{sec:ch2-constant-curvature-manifolds}, hyperbolic space is a space with constant negative curvature $K<0$ and admits several models one can work with. Popular choices include Poincaré and Lorentz. The PV model $\PVspace{n} = \bbR{n}$ is an alternative representation of hyperbolic geometry, which was initially named the Ungar gyrovector space and is used to describe algebraic structures of relativistic proper velocities \citep{ungar2022analytic}. Unlike the constrained Poincaré and Lorentz models, the PV model is an unconstrained space, offering better numerical stability. Its Riemannian metric is given by \cref{pvnn:app:pv-metrics}:
\begin{equation}
\label{pvnn:eq:pv-metric}
g_x(u,v) 
= \inner{u}{v} + K \beta_x^2 \inner{x}{u} \inner{x}{v}, \quad \forall x \in \PVspace{n}, \forall u,v \in T_x \PVspace{n}.
\end{equation}
Here, $\beta_x = \frac{1}{\sqrt{1-K\norm{x}^2}}$ is the relativistic beta factor. In Ungar's notation, the curvature is parametrized by a positive constant $s$ with $s^2 = -1/K$, where $s$ plays the role of the vacuum speed of light in special relativity \citep[Sec.~3.8]{ungar2022analytic}.

\mypara{PV Gyrovector \citep{ungar2022analytic}.} From an algebraic point of view, the PV space forms a gyrovector space~\citep[Def.~6.2]{ungar2022analytic}, which extends the Euclidean vector space to manifolds. Given $x,y,z \in \PVspace{n}$ and $t \in \bbRscalar$, PV gyroaddition $\PVoplus$ and scalar gyromultiplication $\PVotimes$~\citep[Chs.~3.11 and 6.20]{ungar2022analytic} are defined as\footnote{The subscript U refers to the initial letter of Ungar.}
\begin{align}
x \PVoplus y 
&= x + y 
  + \left\{
     \frac{1-\beta_y}{\beta_y} -K \frac{\beta_x}{1+\beta_x} \inner{x}{y}
   \right\} x, \\
t \PVotimes y 
&= 
   \sinh\left(t \sinh^{-1}\left(\sqrt{-K}\norm{y}\right)\right)
   \frac{y}{\sqrt{-K}\norm{y}},  \quad \left(t \PVotimes \zerovec=\zerovec\right).
\end{align}
In particular, the PV inverse is $\PVominus x = -x$, and the PV identity is the zero vector: $\zerovec \PVoplus x = x \PVoplus \zerovec = x$. 

\mypara{PV Gyration.} As shown by \citet[Eqs.~3.220 and 3.221]{ungar2022analytic}, the PV gyration for any $x,y,z \in \PVspace{n}$ is given by
\begin{equation}
\label{pvnn:eq:pv-gyration-closed-form}
\gyr[x,y]z = z + \frac{A x + B y}{D},
\end{equation}
where the coefficients are
\begin{align}
A &= (1-\beta_y^2)K \inner{x}{z} - (1+\beta_x)(1+\beta_y)\beta_x\beta_y K \inner{y}{z} + 2\beta_x^2\beta_y^2 K^2 \inner{x}{y} \inner{y}{z}, \\
B &= (1-\beta_x^2)\beta_y^2 K \inner{y}{z} + (1+\beta_x)(1+\beta_y)\beta_x\beta_y K \inner{x}{z}, \\
D &= (1+\beta_x)(1+\beta_y)\left(1-\beta_x\beta_y K \inner{x}{y} + \beta_x\beta_y\right).
\end{align}
Here, $\beta_x = \frac{1}{\sqrt{1-K\norm{x}^2}}$ is the relativistic beta factor.

\subsection{Proper Velocity Geometry}
\label{pvnn:sec:pv-geometry}

\subsubsection{From Gyro Isomorphism to Riemannian Isometry}
The Poincaré ball also admits a gyrovector space, named the Möbius gyrovector space, as reviewed in \cref{sec:ch2-constant-curvature-manifolds}. Algebraically, the PV and Möbius gyrovector spaces are isomorphic. We further show that PV and the Poincaré ball are geometrically isometric.

The following bijections define the gyrovector space isomorphism \citep[Tab.~6.1]{ungar2022analytic}:
\begin{equation}
\label{pvnn:eq:pv-poincare-isos}
\PVtoPB: \PVspace{n} \ni x \mapsto \frac{\beta_x}{1+\beta_x}x \in \pball{n},
\quad
\PBtoPV: \pball{n} \ni y \mapsto 2 \gamma_y^2 y \in \PVspace{n},
\end{equation}

where $\gamma_y = \frac{1}{\sqrt{1+K\norm{y}^2}}$ is the gamma factor. The isomorphism preserves the gyro operations:
\begin{align}
    \PVtoPB\left(x \PVoplus y\right)
    &= \PVtoPB(x) \Moplus \PVtoPB(y), \quad \forall x, y \in \PVspace{n},\\
    \PVtoPB\left(r \PVotimes x\right)
    &= r \Modot \PVtoPB(x), \quad \forall x \in \PVspace{n}, \forall r \in \bbRscalar,
\end{align}
where $\Modot$ and $\Moplus$ are the Möbius gyro operations reviewed in \cref{sec:ch2-constant-curvature-manifolds}.

\begin{parislemma}[Differentials]
\label{pvnn:lem:pv-poincare-diffs}
\linktoproof{pvnn:lem:pv-poincare-diffs}
The differentials of $\PVtoPB$ and $\PBtoPV$ are
\begin{align*}
d_x\PVtoPB(v) &= K \frac{\beta_x^{3}}{(1+\beta_x)^2}\inner{x}{v}x
 + \frac{\beta_x}{1+\beta_x}v, \quad \forall x \in \PVspace{n}, \forall v \in T_x \PVspace{n}, \\
d_y\PBtoPV(w) &= -4K\gamma_y^4\inner{y}{w}y + 2\gamma_y^2 w, \quad \forall y \in \pball{n}, \forall w \in T_y\pball{n}.
\end{align*}
Let $\id$ be the identity map. The differentials at the origin $\zerovec$ are
\begin{equation}
d_{\zerovec}\PVtoPB = \tfrac{1}{2} \id, \quad d_{\zerovec}\PBtoPV = 2 \id.
\end{equation}
\end{parislemma}
Based on \cref{pvnn:lem:pv-poincare-diffs}, we can prove that the above isomorphisms are isometries.

\begin{paristheorem}[Isometries]
\label{pvnn:thm:pv-poincare-isometry}
\linktoproof{pvnn:thm:pv-poincare-isometry}
The mappings in \cref{pvnn:eq:pv-poincare-isos} are Riemannian isometries.
\end{paristheorem}

\subsubsection{Proper Velocity Riemannian Operators}

The Poincaré ball admits the closed-form Riemannian operators reviewed in \cref{sec:ch2-constant-curvature-manifolds}. By \cref{pvnn:thm:pv-poincare-isometry}, we can readily obtain the counterparts on PV space via the properties of Riemannian isometries reviewed in \cref{def:ch2-riemannian-isometry}.

\Needspace{12\baselineskip}
\begin{paristheorem}[PV Riemannian operators]
\label{pvnn:thm:pv-exp-log-pt-at-x}
\linktoproof{pvnn:thm:pv-exp-log-pt-at-x}
Let $\pi=\PVtoPB$. Given $x,y \in \PVspace{n}$ and $v \in T_x\PVspace{n}$, the Riemannian operators on the PV space are
\begin{align}
\rieexp_{x}(v)
&= x \PVoplus
\left(
\frac{1}{\sqrt{-K}}
\sinh\left(
    \frac{\sqrt{-K} (1+\beta_x)}{\beta_x} \norm{d_x\pi(v)}
    \right)
\frac{d_x\pi(v)}{\norm{d_x\pi(v)}}
\right), \\
\rielog_{x}(y) 
&= \sigma(x,y)z + \tau(x,y)\inner{x}{z}x, \\
\pt{x}{y}(v)
&=
\frac{1+\beta_x}{\beta_x} \tilde{v}
- K \frac{(1+\beta_x)\beta_y}{(1+\beta_y)\beta_x} \inner{y}{\tilde{v}}y,\\
\dist(x,y)
&= \frac{2}{\sqrt{-K}}
    \tanh^{-1}\left(
    \sqrt{-K}\norm{\pi(-x\PVoplus y)}
    \right),
\end{align}
with $z = (-x)\PVoplus y$. For the parallel transport, $\tilde{v}=\Mgyr[\bar{y}, -\bar{x}]\left(d_x\pi(v)\right)$ with $\Mgyr$ as the Möbius gyration in \cref{sec:ch2-constant-curvature-manifolds}, $\bar{x}=\frac{\beta_x}{1+\beta_x}x$ and $\bar{y}=\frac{\beta_y}{1+\beta_y}y$. Here, the scalar coefficients in the logarithm are
\begin{equation}
\begin{aligned}
\sigma(x,y)
&= \frac{2}{\sqrt{-K}}
\frac{\tanh^{-1}\left(\sqrt{-K}\norm{\pi(z)}\right)}{\norm{z}},
\\
\tau(x,y)
&= \frac{2\beta_x}{1+\beta_x}\frac{\sqrt{-K}\tanh^{-1}\left(\sqrt{-K}\norm{\pi(z)}\right)}
{\norm{z}}.
\end{aligned}
\end{equation}
At the identity $\zerovec$, the above operators can be further simplified:
\begin{align}
\rieexp_{\zerovec}(v)
&= \frac{1}{\sqrt{-K}}\sinh\left(\sqrt{-K}\norm{v}\right)\frac{v}{\norm{v}}, \\
\rielog_{\zerovec}(y)
&= \frac{1}{\sqrt{-K}}\sinh^{-1}\left(\sqrt{-K}\norm{y}\right)\frac{y}{\norm{y}}, \\
\pt{\zerovec}{y}(v)
&= v - K \frac{\beta_y}{1+\beta_y}\inner{y}{v}y, \\
\pt{x}{\zerovec}(v)
&= v + K \frac{\beta_x^{2}}{1+\beta_x}\inner{x}{v}x, \\
\dist(\zerovec,y)
&= \frac{1}{\sqrt{-K}}\sinh^{-1}\left(\sqrt{-K}\norm{y}\right).
\end{align}
The expressions containing normalized vectors or $\norm{z}^{-1}$ are understood by continuous extension in the zero cases. Thus, $\rieexp_x(\zerovec)=x$ and $\rielog_x(x)=\zerovec$. In particular, $\rieexp_\zerovec(\zerovec)=\rielog_\zerovec(\zerovec)=\zerovec$.
\end{paristheorem}
This implies that PV gyro operations can be expressed via Riemannian operations.
\Needspace{8\baselineskip}
\begin{paristheorem}[Gyro by Riemannian]
    \label{pvnn:thm:gyro-riem-pv}
    \linktoproof{pvnn:thm:gyro-riem-pv}
    The PV gyro operations can be rewritten as
    \begin{equation}
        \begin{aligned}
        x \PVoplus y
        &= \rieexp _x \left(\pt{{\zerovec}}{x} (\rielog _{\zerovec} (y)) \right),
        && \forall x,y \in \PVspace{n}, \\
        t \PVotimes x
        &= \rieexp _{\zerovec} \left( t \rielog _{\zerovec} (x)\right),
        && \forall x \in \PVspace{n}, \forall t \in \bbRscalar.
        \end{aligned}
    \end{equation}
\end{paristheorem}

\subsection{Proper Velocity Neural Networks}
\label{pvnn:sec:pv-layers}
Building on the above gyrovector and Riemannian tools, we introduce fundamental building blocks for PV neural networks, including MLR, FC, convolutional, activation, and BN layers, thereby enabling the construction of concrete deep architectures in this space.

\subsubsection{Proper Velocity Multinomial Logistic Regression}
\label{pvnn:subsec:pv-mlr}
Following the point-to-hyperplane formulation in \cref{spdmlr:subsec:reform_emlr}, we define the PV margin hyperplane and solve the corresponding point-to-hyperplane infimum under the PV geometry. We define the PV hyperplane as
\begin{equation}
H_{a,p} = \left\{x\in\PVspace{n} \mid \inner{\rielog_{p}(x)}{a}_{p}=0\right\}, 
\qquad p\in\PVspace{n}, a\in T_{p}\PVspace{n},
\end{equation}
where $p\in\PVspace{n}$ and $a\in T_{p}\PVspace{n}$ are the hyperplane parameters. As the Poincaré hyperplane can be expressed by the Möbius gyro operations \citep[Eq.~(22)]{ganea2018hyperbolic}, the PV hyperplane can also be expressed by the PV gyro operations. In addition, building PV MLR requires the PV point-to-hyperplane distance. The following theorem provides these results.

\begin{paristheorem}
\label{pvnn:thm:pv-hyperplane-distance}
\linktoproof{pvnn:thm:pv-hyperplane-distance}
Let $\pi=\PVtoPB$. Given $x, p \in \PVspace{n}$ and $a\in T_{p}\PVspace{n}$, we have
\begin{align*}
&H_{a,p} 
= \left\{x\in\PVspace{n} \mid \inner{\rielog_{p}(x)}{a}_{p}=0\right\} 
= \left\{x\in\PVspace{n} \mid \inner{-p \PVoplus x}{d_p\pi(a)} =0\right\}, \\
&\dist(y, H_{a,p})
= \inf _{w \in H_{a, p}} \dist(y, w)
=\frac{1}{\sqrt{-K}}
\sinh^{-1}\left(
\frac{\sqrt{-K} \left|\inner{-p \PVoplus y}{d_p\pi(a)}\right|}{\|d_p\pi(a)\|}
\right).
\end{align*}
\end{paristheorem}

By \cref{pvnn:thm:pv-hyperplane-distance}, we define the $C$-class PV MLR as
\begin{equation}
    \label{pvnn:eq:pv-mlr-start}
    \begin{aligned}
    p(y=k\mid x) &\propto \exp\left(v_k(x)\right), \\
    v_k(x) &= \sign\left(\inner{-p_k \PVoplus x}{d_{p_k}\pi(a_k)}\right)
    \|a_k\|_{p_k}\dist\left(x,H_{a_k,p_k}\right),
    \end{aligned}
\end{equation}
where $p_k \in \PVspace{n}$ and $a_k \in T_{p_k}\PVspace{n}$ are the PV MLR parameters for class $k$. However, the above expression has three drawbacks: (i) the parameter $p_k$ is over-parameterized, as it corresponds to the scalar bias parameter in the Euclidean MLR; (ii) the gyroaddition in $\inner{-p_k \PVoplus x}{d_{p_k}\pi(a_k)}$ complicates the computation; and (iii) the parameters $(p_k, a_k)$ are constrained, making optimization costly. To address these drawbacks, we follow \citet{shimizu2021hyperbolic} and adopt the parameterization $p_k = \rieexp_\zerovec \left(r_k z_k / \|z_k\|\right)$, $a_k = \pt{\zerovec}{p_k}(z_k)$ with $z_k \in T_\zerovec \PVspace{n} \cong \bbR{n}$ and $r_k \in \bbRscalar$. This parameterization avoids Riemannian optimization in PV MLR and further simplifies the formulation.

\begin{paristheorem}[PV MLR]
\label{pvnn:thm:pv-mlr}
\linktoproof{pvnn:thm:pv-mlr}
For $x \in \PVspace{n}$, the score $v_k(x)$ in \cref{pvnn:eq:pv-mlr-start} for each class $k$ is
{\footnotesize
\begin{equation}
\label{pvnn:eq:pv-mlr-undirectional} 
v_k(x) = \frac{\|z_k\|}{\sqrt{-K}} \sinh^{-1}\left(\cosh(\sqrt{-K}r_k)\frac{\sqrt{-K}}{\|z_k\|}\inner{x}{z_k} - \sinh(\sqrt{-K}r_k)\sqrt{1-K\|x\|^2}\right),
\end{equation}
}
where $z_k \in \bbR{n}$ and $r_k \in \bbRscalar$ are parameters for class $k$. In particular, as $K \to 0^{-}$ we have $v_k(x)\to \inner{x}{z_k} + b_k$ with $b_k = - r_k \|z_k\|$, which recovers the Euclidean MLR reviewed in \cref{spdmlr:subsec:reform_emlr}.
\end{paristheorem}

The parameterization $(z_k, r_k)$ is essential for efficiency. In the original form \cref{pvnn:eq:pv-mlr-start}, computing $v_k(x)$ for a batch $x \in \bbR{b \times n}$ and $C$ classes requires explicit gyroaddition $-p_k \PVoplus x$ for each class, producing an intermediate tensor of size $b \times C \times n$ that could cause out-of-memory errors in high dimensions. One could instead loop over classes, but this is computationally inefficient. In contrast, \cref{pvnn:eq:pv-mlr-undirectional} depends on inner products $\inner{x}{z_k}$, which can be implemented as a matrix multiplication.

\subsubsection{Proper Velocity Fully Connected Layer}
\label{pvnn:subsec:pv-fc-hnnpp}

The Euclidean \emph{Fully Connected (FC)} layer is defined as $y = Ax + b$ with $A \in \bbR{m \times n}$ and $b \in \bbR{m}$. It can be expressed element-wise as $y_k = \inner{a_k}{x} - b_k= \inner{a_k}{x - p_k}$ with $a_k, p_k \in \bbR{n}$ and $\inner{p_k}{a_k} = b_k$. As shown by \citet[Sec.~3.2]{shimizu2021hyperbolic} and \citet[Sec.~3.1]{chen2025riefc}, the LHS $y _k$ is the signed distance from $y$ to the hyperplane passing through the origin and orthogonal to the $k$-th axis of the output space, which can be formulated as
\begin{equation} \label{pvnn:eq:euc-fc}
    \sign\left(\inner{e_k}{y - \zerovec}\right) \dist (y, H_{e_k, \zerovec}) = \inner{a_k}{x - p_k}, \quad \forall 1 \leq k \leq m,
\end{equation}
where $e_k$ denotes the vector whose $k$-th element is 1 and all others are 0.

For the PV model, the LHS of \cref{pvnn:eq:euc-fc} can be formulated by the signed point-to-hyperplane distance, while the RHS can be formulated by the $v_k$ in PV MLR. Specifically, the PV FC layer $\calF: \PVspace{n} \to \PVspace{m}$ from the $n$-dimensional to the $m$-dimensional PV spaces for the input $x \in \PVspace{n}$ returns the output $y \in \PVspace{m}$ by solving the $m$ equations:
\begin{equation} \label{pvnn:eq:pv-fc}
    \sign\left(\inner{d_{\zerovec}\pi(e_k)}{-\zerovec \PVoplus y}\right)
     \dist (y, H_{e_k, \zerovec})
     = v_k(x), \quad \forall 1 \leq k \leq m,
\end{equation}
where $H_{e_k, \zerovec}$ and $v_k(x)$ are given by \cref{pvnn:thm:pv-hyperplane-distance,pvnn:eq:pv-mlr-undirectional}, respectively. This definition has an explicit solution.

\Needspace{12\baselineskip}
\begin{paristheorem}[PV FC layer]
\label{pvnn:thm:pv-fc}
\linktoproof{pvnn:thm:pv-fc}
The output $y = \calF(x) \in \PVspace{m}$ has the closed form
\begin{equation}
\label{pvnn:eq:pv-fc-final}
y_k 
= \frac{1}{\sqrt{-K}}\sinh(\sqrt{-K} v_k(x)),
\quad 1 \le k \le m,
\end{equation}
where $v_k(x)$ is defined in \cref{pvnn:eq:pv-mlr-undirectional} with $z_k \in \bbR{n}$ and $r_k \in \bbRscalar$ as the FC parameters. In particular, as $K \to 0^{-}$ we have $y_k \to \inner{x}{z_k} + b_k$ with $b_k = - r_k \|z_k\|$, which recovers the Euclidean FC layer.
\end{paristheorem}

\mypara{Generalization.} We can jointly express the Euclidean FC layer and activation $\sigma$, which yields the RHS of \cref{pvnn:eq:euc-fc} with $\sigma\left(\inner{a_k}{x-p_k}\right)$. Accordingly, we extend the PV FC by applying the activation to $v_k(x)$ in \cref{pvnn:eq:pv-fc}. Then, \cref{pvnn:eq:pv-fc-final} becomes
\begin{equation}
\label{pvnn:eq:pv-fc-activation}
y_k = \frac{1}{\sqrt{-K}}\sinh(\sqrt{-K} \sigma(v_k(x))), 
\quad 1 \le k \le m.
\end{equation}

\subsubsection{Proper Velocity Convolution and Activation}
\label{pvnn:subsec:pv-conv-activation}
\mypara{Convolution.} As shown by \citet{shimizu2021hyperbolic,bdeir2024fully,chen2025riefc}, Euclidean convolution consists of linear maps between kernel weights and concatenated values in each receptive field. To define convolution on PV space, it therefore suffices to define PV concatenation, since we already have the PV FC layer. Because PV space is unconstrained, we define PV concatenation to coincide with Euclidean concatenation. For simplicity, we consider the 1D case. For PV inputs $\{x_{i} \in \PVspace{n} \}_{i=1}^{k}$ in a 1D receptive field (where $k$ is the kernel size), the PV convolution output $y \in \PVspace{m}$ for this receptive field is
$y = \calF\left(\operatorname{Concat}\left(x_{1}, \dots, x_{k}\right)\right)$,
where $\operatorname{Concat}(\cdot)$ is standard Euclidean concatenation and $\calF$ is the PV FC layer.

\mypara{Activation.} A natural choice is to apply a Euclidean activation $\sigma$ in the tangent space at the origin via the mapping
$x \mapsto \rieexp_{\zerovec}\left(\sigma\left(\rielog_{\zerovec}(x)\right)\right)$,
which has been shown to be effective in Poincaré networks \citep{ganea2018hyperbolic}. Alternatively, since PV space is unconstrained, we can apply the activation directly in PV space as $x \mapsto \sigma(x)$. This direct PV-space activation avoids exponential and logarithmic maps and is therefore more efficient.

\subsubsection{Proper Velocity Normalization}
\label{pvnn:subsec:normalization}
We instantiate the GyroBN framework in \cref{sec:ch3-gyrobn-general} on the PV space and show that PV GyroBN can normalize sample statistics.

Given activations $\{ x_i \in \PVspace{n}\}_{i=1}^N$, the core operations of PV GyroBN are
\begin{equation}
    \label{pvnn:eq:pvgyrobn}
    \forall i \leq N, \quad 
    \tilde{x}_i \gets  
            \overbrace{B \PVoplus}^{\text{Biasing}}
    \left(
        \overbrace{\frac{s}{\sqrt{v^2+\epsilon}} \PVotimes}^{\text{Scaling}}
        \left(
            \overbrace{- M \PVoplus x_i}^{\text{Centering}}
        \right)
    \right),
\end{equation}
where $M$ and $v^2$ denote the Fr\'echet mean and variance, and $B \in \PVspace{n}$ and $s \in \bbRscalar$ are parameters. Owing to the isometry between the PV space and Poincaré ball, the PV Fréchet mean can be computed via the Poincaré ball: map the data to the Poincaré ball, compute the Poincaré mean \citep[Alg.~1]{lou2020differentiating}, and map the result back.

The following theorem guarantees that PV GyroBN can normalize sample statistics.
\begin{paristheorem}[Homogeneity]
    \label{pvnn:thm:homogeneity}
    \linktoproof{pvnn:thm:homogeneity}
    For $N$ samples $\{x_i\}_{i=1}^N \subset \PVspace{n}$, we have
    \begin{align*}
        & \text{Homogeneity of mean: }
        \fm\left( \{B \PVoplus x_i \}_{i=1}^N \right) 
        = B \PVoplus \fm\left( \{ x_i \}_{i=1}^N \right), 
        \quad \forall B \in \PVspace{n},\\
        & \text{Homogeneity of dispersion from $\zerovec$: } 
        \frac{1}{N} \sum\nolimits_{i=1}^N 
        \dist^2(t \PVotimes x_i, \zerovec) 
        = t^2 \cdot \frac{1}{N} \sum\nolimits_{i=1}^N 
        \dist^2(x_i, \zerovec).
    \end{align*}
\end{paristheorem}
\cref{pvnn:thm:homogeneity} directly explains the PV GyroBN in \cref{pvnn:eq:pvgyrobn}. After the centering, the batch mean is shifted to the identity $\zerovec$. After the scaling, the variance becomes $s^2$. After the biasing, the batch mean is translated to $B$.

\subsection{Connections to Lorentz Geometry}
\label{pvnn:sec:pv-and-lorentz}
This subsection discusses the connections between the PV and Lorentz models. We first show the isometry between the two models. Then, we show that several current Lorentz network layers can be rewritten as PV layers.

\Needspace{18\baselineskip}
\begin{parisproposition}[PV--Lorentz isometries]
\label{pvnn:prop:pv-lorentz-isometry}
\linktoproof{pvnn:prop:pv-lorentz-isometry}
The following maps are Riemannian isometries between $\lorentz{n}$ and $\PVspace{n}$:
\begin{align}
\LtoPV: 
&\lorentz{n} \ni
\begin{bmatrix}
x_t \\
x_s
\end{bmatrix} \mapsto x_s \in \PVspace{n},\\
\PVtoL: 
&\PVspace{n} \ni x \mapsto \begin{bmatrix}
\displaystyle \sqrt{\norm{x}^2 - \tfrac{1}{K}} \\
\displaystyle x
\end{bmatrix} \in \lorentz{n}.
\end{align}
\end{parisproposition}

The PV--Lorentz isometries in \cref{pvnn:prop:pv-lorentz-isometry} imply that several standard layers in Lorentz networks can be rewritten as PV layers composed with $\LtoPV$ and $\PVtoL$.

The Lorentz activation \citep[Eq.~(13)]{bdeir2024fully}, Lorentz FC layer \citep[Sec.~3.1]{chen2022fully} and Lorentz concatenation \citep[Eq.~(32)]{bdeir2024fully} are
\begin{align}
\mathrm{LAct}\left(
\begin{bmatrix}
x_t \\
x_s
\end{bmatrix}
\right)
&=
\begin{bmatrix}
\sqrt{\norm{\sigma(x_s)}^2 - \tfrac{1}{K}} \\
\sigma(x_s)
\end{bmatrix}, \\
\mathrm{LFC}\left(
\begin{bmatrix}
x_t \\
x_s
\end{bmatrix}
\right)
&=
\begin{bmatrix}
\sqrt{\norm{W x_s + b}^2 - \tfrac{1}{K}} \\
W x_s + b
\end{bmatrix}, \\
\mathrm{LCat}(\{x_i\}_{i=1}^N)
&=
\begin{bmatrix}
\sqrt{\sum_{i=1}^{N} x_{i,t}^2 + \tfrac{N-1}{K}} \\
x_{1,s} \\
\vdots \\
x_{N,s}
\end{bmatrix} \in \lorentz{nN},
\end{align}
where $x = [x_t, x_s^\top]^\top \in \lorentz{n}$ and $x_i = [x_{i,t}, x_{i,s}^\top]^\top \in \lorentz{n}$ for $1 \leq i \leq N$. Then \cref{pvnn:prop:pv-lorentz-isometry} implies that the above Lorentz layers can be rewritten in terms of PV layers as follows:
\begin{align}
\mathrm{LAct}(x)
&= \PVtoL(\sigma(\LtoPV(x))), \\
\mathrm{LFC}(x)
&=
\PVtoL(W \LtoPV(x) + b), \\
\mathrm{LCat}(\{x_i\}_{i=1}^N)
&=
\PVtoL \left(\operatorname{Concat}(\LtoPV(x_1), \ldots, \LtoPV(x_N)) \right).
\end{align}
These identities show that many Lorentz constructions effectively operate by mapping to PV space, applying Euclidean building blocks there, and mapping back through $\PVtoL$. 
This perspective naturally motivates designing networks directly in PV space, instead of repeatedly switching between equivalent models. Moreover, even if one follows the pattern $\lorentz{n} \to \PVspace{n} \to \PVspace{m} \to \lorentz{m}$ to construct layers, the intermediate map should be the PV layers, such as \cref{pvnn:thm:pv-fc}, rather than Euclidean layers, since PV is a non-linear Riemannian manifold.

\subsection{Experiments}
\label{pvnn:sec:experiments}
We evaluate PV embeddings and PVNNs on four representative tasks:
\begin{itemize}
  \item \cref{pvnn:subsec:stability} evaluates the numerical advantage of the PV model against Poincaré and Lorentz.
  \item \cref{pvnn:sec:exp2} compares PV, Poincaré, and Lorentz MLRs on image classification.
  \item \cref{pvnn:sec:exp1} evaluates our PV MLR, FC, and GyroBN layers on graph learning.
  \item \cref{pvnn:sec:exp3} compares fully PV convolutional networks with fully Lorentz convolutional networks on genomic sequence learning.
\end{itemize}
More details on data sets and experimental settings are provided in \cref{app:datasets,pvnn:app:experimental-details}.

\subsubsection{Numerical Stability}
\label{pvnn:subsec:stability}

\begin{wraptable}{r}{0.5\textwidth}
    \centering
    \caption{Failure and violation rates (\%) of $r \Hotimes x$ in FP32.}
    \label{pvnn:tab:fail_prob_precision}
\resizebox{\linewidth}{!}{
\begin{tabular}{c|>{\columncolor{HilightColor}}c cc|>{\columncolor{HilightColor}}c cc}
    \toprule
    \multirow{2}{*}{$r$} & \multicolumn{3}{c|}{\textbf{Failure rate}} & \multicolumn{3}{c}{\textbf{Violation rate}} \\
    \cmidrule(lr){2-4} \cmidrule(l){5-7}
     & $\PVspace{n}$ & $\pball{n}$ & $\lorentz{n}$ & $\PVspace{n}$ & $\pball{n}$ & $\lorentz{n}$ \\
    \midrule
    1     & 0 & 0 & 0      & \na & 0 & 32.50 \\
    5     & 0 & 0 & 0      & \na & 0 & 92.36 \\
    10    & 0 & 0 & 0      & \na & 0 & 99.76 \\
    20    & 0 & 0 & 4.23   & \na & 0 & 100 \\
    50    & 0 & 0 & 64.42  & \na & 0 & 100 \\
    75    & 0 & 0 & 79.63  & \na & 0 & 100 \\
    100   & 0 & 0 & 88.26  & \na & 0 & 100 \\
    150   & 0 & 0 & 96.43  & \na & 0 & 100 \\
    200   & 0 & 0 & 100    & \na & 0 & 100 \\
    1000  & 0 & 0 & 100    & \na & 0 & 100 \\
    \bottomrule
\end{tabular}
}
\end{wraptable}
We study three aspects: gyro operator, Riemannian operator, and gradient behavior. All experiments use curvature $K=-1$, dimension $n=16$, and batch size $4096$.

\mypara{Gyro Operator.} We use scalar gyromultiplication $r \Hotimes x$ as a probe of numerical stability across hyperbolic models. Given random batches $x$ and radii $r$, we evaluate two metrics. The \emph{failure rate} is the fraction of outputs that contain NaN/Inf. The \emph{violation rate} is defined only for models with manifold constraints: Poincaré ball requires $\norm{x}^2 < -1/K$, and Lorentz requires $\Linner{x}{x}=\norm{x_s}^2-x_t^2=\frac{1}{K}$ for $x = [x_t,x_s^\top]^\top$. The tolerance is set to $10^{-8}$. As PV is unconstrained, its violation rate is reported as \na. As shown in \cref{pvnn:tab:fail_prob_precision}, PV maintains zero failures up to $r=1000$ in FP32. The Poincaré ball has zero failure and violation rates, whereas Lorentz starts to fail around $r=20$ and quickly accumulates both NaN/Inf outputs and off-manifold points under large scalar multipliers, revealing pronounced numerical instability.

\begin{wraptable}{r}{0.4\textwidth}
    \centering
    \caption{$\norm{\rielog_\zerovec(\rieexp_\zerovec(v)) - v}$.}
    \label{pvnn:tab:invertibility}
    \resizebox{\linewidth}{!}{
    \begin{tabular}{ccc}
        \toprule
        \textbf{Model} & \textbf{FP32} & \textbf{FP64} \\
        \midrule
        $\pball{n}$ & $2.1 \times 10^{-4}$ & $4.3 \times 10^{-11}$ \\
        $\lorentz{n}$ & $1.0 \times 10^{0}$ & $1.0 \times 10^{0}$ \\
        \rowcolor{HilightColor} $\PVspace{n}$ & $2.1 \times 10^{-7}$ & $6.7 \times 10^{-16}$ \\
        \bottomrule
    \end{tabular}
    }
\end{wraptable}
\mypara{Riemannian Operator.} We evaluate the exponential and logarithmic maps by measuring the round-trip error $\norm{\rielog_\zerovec(\rieexp_\zerovec(v)) - v}$ for tangent vectors $v$ with large norm $\norm{v} = 10$. Since this quantity is theoretically zero, any non-zero value reflects numerical instability. We sample a batch of such vectors and report the average error in \cref{pvnn:tab:invertibility}. PV achieves stable behavior in both FP32 and FP64, whereas the Poincaré ball already exhibits noticeable errors in FP32 and Lorentz remains unstable in both precisions.

\begin{wraptable}{r}{0.6\textwidth}
    \centering
    \caption{Gradient magnitude $\norm{\nabla_x f_r(x)}$ across varying radii.}
    \label{pvnn:tab:gradients}
    \resizebox{\linewidth}{!}{
    \begin{tabular}{ccc}
        \toprule
        \textbf{Model} & {\boldmath\textbf{$\norm{\nabla_x f_r(x)}$ Range}} & \textbf{Gradient behavior} \\
        \midrule
        $\pball{n}$ & $[7.6 \times 10^{-13}, 1.1 \times10^{-11}]$ & Vanishing gradients \\
        $\lorentz{n}$ & $[0, \text{NaN}]$ & Exploding gradients \\
        \rowcolor{HilightColor} $\PVspace{n}$ & $[2.1 \times 10^{-6}, 1.1 \times10^{-4}]$ & Stable gradients \\
        \bottomrule
    \end{tabular}
    }
\end{wraptable}
\mypara{Gradient.} To compare gradient behavior, we study the gradient of $f_r(x) = \norm{r \Hotimes x - x}$ with respect to $x$. Specifically, we sample 24 logarithmically spaced radii $r \in [1, 1000]$ and, for each radius, measure the $\norm{\nabla_x f_r(x)}$ on a random batch. The range of $\norm{\nabla_x f_r(x)}$ is summarized in \cref{pvnn:tab:gradients}. The Poincaré ball exhibits severe gradient vanishing near the boundary. In contrast, Lorentz yields gradients that vary from $0$ to NaN, reflecting gradient explosion. PV maintains gradients in a safer band.

\subsubsection{Image Classification}
\label{pvnn:sec:exp2}

We compare our PV MLR against previous Poincar\'e MLRs \citep{ganea2018hyperbolic,shimizu2021hyperbolic} and Lorentz MLR \citep{bdeir2024fully}. Following \citet{bdeir2024fully}, we train a ResNet-18 backbone \citep{he2016deep} on CIFAR-10 and CIFAR-100 \citep{krizhevsky2009learning}, replacing the final Euclidean MLR with a hyperbolic MLR. The backbone output is lifted to the target geometry via the exponential map at the identity. Since PV space is unconstrained, we also consider a direct variant that skips $\rieexp_{\zerovec}$ and treats the backbone output as PV coordinates. We denote these two PV heads as PV MLR (with $\rieexp_{\zerovec}$) and PV MLR (without $\rieexp_{\zerovec}$). \cref{pvnn:tab:vision-results} reports the 5-fold results. PV MLR matches or outperforms prior hyperbolic baselines, with the largest gains on CIFAR-100 where the decision boundaries are more complex. Both PV variants, with and without $\rieexp_{\zerovec}$, achieve similar accuracies.

\begin{table}[t]
\caption{Top-1 image classification accuracy (\%) of hyperbolic MLRs on ResNet-18. The best results are \firstresults{bold}. $\delta$ represents the $\delta$-hyperbolicity (lower is more hyperbolic), which comes from \citet[Tab.~1]{bdeir2024fully}.}
\label{pvnn:tab:vision-results}
\centering
\resizebox{0.8\linewidth}{!}{
\begin{tabular}{llcc}
\toprule
\textbf{Model} & \textbf{Method} & \makecell{\textbf{CIFAR-10} \\ ($\delta=0.26$)} & \makecell{\textbf{CIFAR-100} \\ ($\delta=0.23$)} \\
\midrule
\multirow{2}{*}{$\pball{n}$}
 & Poincar\'e MLR~\citep[Eq.~(25)]{ganea2018hyperbolic}
 & $95.09 \pm 1.51$
 & $76.78 \pm 0.67$ \\
 & Unidirectional MLR~\citep[Eq.~(6)]{shimizu2021hyperbolic}
 & $95.12 \pm 0.20$
 & $77.19 \pm 0.10$ \\
\midrule
$\lorentz{n}$
 & Lorentz MLR~\citep[Thm.~2]{bdeir2024fully}
 & $95.02 \pm 0.12$
 & $77.96 \pm 0.09$ \\
\midrule
\multirow{2}{*}{$\PVspace{n}$}
 & \cellcolor{HilightColor} PV MLR (with $\rieexp_{\zerovec}$)
 & \cellcolor{HilightColor} $95.27 \pm 0.12$
 & \cellcolor{HilightColor} $78.19 \pm 0.59$ \\
 & \cellcolor{HilightColor} PV MLR (without $\rieexp_{\zerovec}$)
 & \cellcolor{HilightColor} \firstresults{$95.30 \pm 0.18$}
 & \cellcolor{HilightColor} \firstresults{$78.20 \pm 0.37$} \\
\bottomrule
\end{tabular}
}
\end{table}

\subsubsection{Graph Learning}
\label{pvnn:sec:exp1}

\mypara{Data and Setup.} We study node classification on four standard graph data sets: Disease~\citep{anderson1991infectious}, Airport~\citep{zhang2018link}, Cora~\citep{sen2008collective}, and PubMed~\citep{namata2012query}. All models share the same architecture consisting of two FC layers with nonlinear activations followed by an MLR classifier; they differ only in the underlying hyperbolic model. Baselines include KNN~\citep{mao2024klein} for the Klein ball, HNN/HNN++~\citep{chami2019hyperbolic,shimizu2021hyperbolic} for the Poincaré ball, and LNN~\citep{bdeir2024fully} for Lorentz. Our PVNN is built from PV FC, activation, and MLR layers.

\begin{table}[t]
    \centering
    \caption{Accuracies of hyperbolic networks on graph learning. The best results are \firstresults{bold}. $\delta$ represents the $\delta$-hyperbolicity (lower is more hyperbolic).}
    \label{pvnn:tab:pvnn_results}
    \resizebox{\linewidth}{!}{
    \begin{tabular}{llcccc}
    \toprule
    \textbf{Model} & \textbf{Method} & \makecell{\textbf{Disease} \\ ($\delta=0$)} & \makecell{\textbf{Airport} \\ ($\delta=1$)} & \makecell{\textbf{PubMed} \\ ($\delta=3.5$)} & \makecell{\textbf{Cora} \\ ($\delta=11$)} \\
    \midrule
    $\klein{n}$ & KNN~\citep{mao2024klein} & $79.41 \pm 0.55$ & $92.10 \pm 0.97$ & $69.36 \pm 0.76$ & $52.26 \pm 1.99$ \\
    \midrule
    \multirow{2}{*}{$\pball{n}$} 
          & HNN~\citep{ganea2018hyperbolic} & $79.90 \pm 0.01$ & $82.16 \pm 2.95$ & $69.28 \pm 0.85$ & $49.68 \pm 1.25$ \\
          & HNN++~\citep{shimizu2021hyperbolic} & $80.57 \pm 0.23$ & $88.40 \pm 0.17$ & $73.68 \pm 0.39$ & $52.06 \pm 0.90$ \\
    \midrule
    $\lorentz{n}$ & LNN~\citep{bdeir2024fully} & $79.90 \pm 0.01$ & $75.20 \pm 1.08$ & $68.82 \pm 0.88$ & \firstresults{$53.34 \pm 1.65$} \\
    \midrule
    \rowcolor{HilightColor} $\PVspace{n}$ & PVNN & \firstresults{$81.15 \pm 0.23$} & \firstresults{$97.96 \pm 0.42$} & \firstresults{$74.33 \pm 0.22$} & $51.42 \pm 1.33$ \\
    \bottomrule
    \end{tabular}
    }
\end{table}

\mypara{Main Results.} For a fair comparison, we use a tangent activation in each model and set $\sigma=\id$ for the PV FC layer in \cref{pvnn:eq:pv-fc-activation}. \cref{pvnn:tab:pvnn_results} summarizes the 5-fold results. On the three more hyperbolic data sets (Disease, Airport, and PubMed), PVNN consistently achieves the best performance, with especially large gains on Airport where it improves over the strongest baseline by $5.86\%$. On the weakly hyperbolic Cora data set, PVNN remains comparable to Poincaré- and Klein-based networks, and worse than the Lorentz-based one. Overall, these results suggest that PV geometry is more effective on strongly hyperbolic graphs.

\begin{table}[t]
    \centering
    \caption{Results of Tangent FC (TFC) vs PV FC, and Tangent BN (TBN) vs GyroBN.}
    \label{pvnn:tab:ablations-euclidean}
    \resizebox{0.8\linewidth}{!}{
    \begin{tabular}{lcccc}
    \toprule
    \textbf{Method} & \textbf{Disease} & \textbf{Airport} & \textbf{PubMed} & \textbf{Cora} \\
    \midrule
    PVNN+TFC & $80.86 \pm 0.30$ & $86.99 \pm 0.61$ & \firstresults{$74.40 \pm 0.43$} & \firstresults{$53.58 \pm 0.81$} \\
    \rowcolor{HilightColor} PVNN & \firstresults{$81.24 \pm 0.36$} & \firstresults{$97.93 \pm 0.29$} & $74.16 \pm 0.32$ & $52.26\pm 1.32$ \\
    \midrule
    PVNN+TBN & $80.67 \pm 0.38$ & $98.71 \pm 0.36$ & $73.52 \pm  0.12$ & $45.36 \pm 2.44$ \\
    \rowcolor{HilightColor} PVNN+GyroBN & \firstresults{$81.24 \pm 0.19$} & \firstresults{$99.03 \pm 0.18$} & \firstresults{$74.34 \pm 0.31$} & \firstresults{$46.64 \pm 5.45$} \\
    \bottomrule
    \end{tabular}
    }
\end{table}

\begin{table}[t]
\caption{Comparison of methods in calculating mean and variance in PV GyroBN. Time is measured in milliseconds per training epoch.}
\label{pvnn:tab:pv-gyrobn}
\centering
\resizebox{\linewidth}{!}{
\begin{tabular}{c|cc|cc|cc|cc}
\toprule
\multirow{2.5}{*}{\textbf{Method}}
& \multicolumn{2}{c}{\textbf{Disease}}
& \multicolumn{2}{c}{\textbf{Airport}}
& \multicolumn{2}{c}{\textbf{PubMed}}
& \multicolumn{2}{c}{\textbf{Cora}} \\
\cmidrule{2-9}
& \textbf{Acc} & \textbf{Fit Time} & \textbf{Acc} & \textbf{Fit Time} & \textbf{Acc} & \textbf{Fit Time} & \textbf{Acc} & \textbf{Fit Time} \\
\midrule
  Tangent
    & $81.15 \pm 0.23$ & 26.08
    & $98.56 \pm 0.36$ & 55.48
    & $61.50 \pm 5.75$ & 3.10
    & $33.10 \pm 1.58$ & 7.12 \\
  Euclidean
    & $81.15 \pm 0.23$ & 25.80
    & $98.75 \pm 0.31$ & 55.19
    & $69.82 \pm 3.58$ & 2.99
    & $32.62 \pm 0.65$ & 7.29 \\
  \midrule
  Fr\'echet 1 iter
    & $81.05 \pm 0.23$ & 29.79
    & $88.93 \pm 1.17$ & 65.19
    & $62.52 \pm 8.44$ & 3.38
    & $42.84 \pm 6.15$ & 7.67 \\
  Fr\'echet 2 iters
    & $81.05 \pm 0.23$ & 30.12
    & $94.11 \pm 0.46$ & 67.37
    & $73.78 \pm 0.20$ & 3.49
    & $45.68 \pm 4.36$ & 8.21 \\
  Fr\'echet 5 iters
    & $81.24 \pm 0.36$ & 30.90
    & $98.50 \pm 0.16$ & 82.28
    & $73.92 \pm 0.44$ & 4.02
    & \firstresults{$49.50 \pm 1.83$} & 9.15 \\
  \rowcolor{HilightColor} Fr\'echet 10 iters
    & \firstresults{$81.24 \pm 0.19$} & 30.49
    & \firstresults{$99.03 \pm 0.18$} & 105.79
    & \firstresults{$74.34 \pm 0.31$} & 3.96
    & $46.64 \pm 5.45$ & 9.77 \\
  Fr\'echet $\infty$
    & $80.86 \pm 0.00$ & 31.29
    & $98.46 \pm 0.15$ & 122.37
    & $71.16 \pm 3.93$ & 4.46
    & $47.32 \pm 4.73$ & 9.27 \\
\bottomrule
\end{tabular}}
\end{table}

\begin{table}[t]
    \centering
    \caption{Ablations on PVNN with or without exponential map for the input PV feature.}
    \label{pvnn:tab:ablations-no-expmap}
    \resizebox{0.8\linewidth}{!}{%
    \begin{tabular}{lcccc}
    \toprule
    $\rieexp_\zerovec$ & \textbf{Disease} & \textbf{Airport} & \textbf{PubMed} & \textbf{Cora} \\
    \midrule
    \xmark  & $81.05 \pm 0.23$ & $97.71 \pm 0.34$ & \firstresults{$74.22 \pm 0.26$} & $51.92 \pm 2.01$ \\
    \midrule
    \rowcolor{HilightColor} \cmark & \firstresults{$81.24 \pm 0.36$} & \firstresults{$97.93 \pm 0.29$} & $74.16 \pm 0.32$ & \firstresults{$52.26\pm 1.32$} \\
    \bottomrule
    \end{tabular}
    }
\end{table}

\begin{table}[t]
    \centering
    \caption{Ablations on PV activations.}
    \label{pvnn:tab:ablations-architecture}
    \resizebox{\linewidth}{!}{%
    \begin{tabular}{lcccc}
    \toprule
    \textbf{Method} & \textbf{Disease} & \textbf{Airport} & \textbf{PubMed} & \textbf{Cora} \\
    \midrule
    Tangent Act. & $81.24 \pm 0.36$ & $97.93 \pm 0.29$ & $74.16 \pm 0.32$ & \firstresults{$52.26\pm 1.32$}  \\
    FC $\sigma$     & $81.34 \pm 0.43$ & \firstresults{$99.40 \pm 0.15$} & $74.02 \pm 0.17$ & $51.34 \pm 0.46$ \\
    FC $\sigma$ + Tangent Act.  & $80.96 \pm 0.19$ & $99.15 \pm 0.38$ & $73.96 \pm 0.22$ & $51.30 \pm 1.65$ \\
    \midrule
    Euc. Act.   & \firstresults{$81.34 \pm 0.43$} & $98.87 \pm 0.35$ & \firstresults{$74.56 \pm 0.59$} & $38.10 \pm 3.30$ \\
    \bottomrule
    \end{tabular}
    }
\end{table}

\mypara{Tangent vs. Riemannian.} A natural construction of hyperbolic layers is to work in the tangent space. To validate the benefits of our Riemannian PV layers, we compare our PV FC with TFC of the form $\rieexp_\zerovec\left(A\rielog_\zerovec(x)+b\right)$, and our GyroBN with TBN given by $\rieexp_\zerovec(\mathrm{BN}(\rielog_\zerovec(x)))$~\citep{ioffe2015batch}. We denote these variants by PVNN+TFC and PVNN+TBN, respectively. As shown in \cref{pvnn:tab:ablations-euclidean}, PVNN consistently outperforms PVNN+TFC on the more hyperbolic Disease and Airport data sets, while performance on the other two data sets is comparable and TFC can be slightly better. For normalization, PVNN+GyroBN improves over PVNN+TBN on all data sets. Overall, these ablations validate the effectiveness of our Riemannian PV constructions, especially in strongly hyperbolic settings.

\mypara{Ablations on Batch Statistics.} PV GyroBN in \cref{pvnn:eq:pvgyrobn} uses Fréchet mean and variance, which require iterative solvers. We also consider two efficient variants. A tangent variant computes batch statistics in the tangent space at the identity via
\begin{equation}
\begin{aligned}
M &= \rieexp_{\zerovec}\left(\frac{1}{N}\sum_{i=1}^N \rielog_{\zerovec}(x_i)\right), \\
v^2 &= \frac{1}{N}\sum_{i=1}^N \norm{\rielog_{\zerovec}(x_i) - \rielog_{\zerovec}(M)}^2,
\end{aligned}
\end{equation}
and a Euclidean variant computes standard Euclidean mean and variance directly in the unconstrained PV space. \cref{pvnn:tab:pv-gyrobn} shows that Tangent and Euclidean are up to $2\times$ faster while achieving similar accuracies on Disease and Airport. Although Fréchet-based GyroBN attains the best accuracies, it is more computationally expensive.

\mypara{Ablations on PV Embedding.} In the main experiments, the input features are first lifted to PV via $\rieexp_{\zerovec}$ and then processed by PVNN. Since PV space is unconstrained, we also consider a variant that feeds the Euclidean features directly as PV coordinates. \cref{pvnn:tab:ablations-no-expmap} compares these two settings. The two variants perform similarly, while using $\rieexp_{\zerovec}$ provides small improvements on Disease, Airport, and Cora. This differs from image classification in \cref{pvnn:tab:vision-results}, where the variant without $\rieexp_{\zerovec}$ is marginally better. This slight discrepancy may stem from the different nature of the inputs. In vision, the ResNet encoder can adapt its learned representation to the chosen lifting, whereas in graphs the raw node features benefit slightly from the explicit exponential map.

\mypara{Ablations on Activation.} We ablate three types of activations in PVNN: the internal nonlinearity $\sigma$ in the PV FC layer (fixed to $\tanh$), and explicit activations applied either directly in PV (\textit{Euc. Act.}) or in the tangent space (\textit{Tangent Act.}). \cref{pvnn:tab:ablations-architecture} reports the results. First, when comparing these three choices individually, the differences are small on Disease and PubMed, while FC $\sigma$ performs best on Airport, Tangent Act. performs best on Cora, and Euc. Act. degrades substantially on Cora. Second, when comparing the composite variant FC $\sigma$ + Tangent Act. against Tangent Act., the composite does not yield consistent gains, suggesting redundancy.

\FloatBarrier
\subsubsection{Genomic Sequence Learning}
\label{pvnn:sec:exp3}

\citet{khan2025hyperbolic} recently proposed \emph{Hyperbolic Convolutional Neural Networks (HCNNs)} under Lorentz geometry for genomic sequence learning, demonstrating that HCNNs outperform Euclidean CNNs on this task. Following \citet{khan2025hyperbolic}, we evaluate on the \emph{Transposable Elements Benchmark (TEB)} data set for DNA transposable element prediction. To ensure a fair comparison, all models share the same backbone network architecture, which consists of two convolutional blocks followed by an FC layer and a final MLR classifier~\citep{khan2025hyperbolic}. We use a single curvature shared for all layers. \cref{pvnn:tab:teb_results} reports 5-fold \emph{Matthews Correlation Coefficient (MCC)}. The \emph{PV Convolutional Network (PVCNN)} achieves the best performance on all TEB tasks, with particularly strong gains on SINEs, where it improves over HCNN-S by about $9$ MCC points. These results demonstrate the benefits of PV convolutional networks.

\begin{table}[t]
\centering
\caption{Comparison in MCC of hyperbolic and Euclidean convolutional networks, including PVCNN, on TEB data sets.}
\label{pvnn:tab:teb_results}
\resizebox{\linewidth}{!}{%
\begin{tabular}{lcccc}
\toprule
\textbf{Task} & \textbf{Data Set} & \textbf{Euclidean CNN} & \textbf{HCNN-S} & \cellcolor{HilightColor} \textbf{PVCNN} \\
\midrule
\multirow{2}{*}{Retrotransposons}
& LINEs  & $70.63 \pm 1.24$ & $76.12 \pm 2.16$ & \cellcolor{HilightColor} \firstresults{$81.83 \pm 0.27$} \\
& SINEs  & $85.15 \pm 1.64$ & $85.45 \pm 1.16$ & \cellcolor{HilightColor} \firstresults{$93.78\pm 0.54$} \\
\midrule
\multirow{1}{*}{DNA transposons}
& hAT-Ac    & $87.45 \pm 0.90$ & $89.61 \pm 1.34$ & \cellcolor{HilightColor} \firstresults{$92.08 \pm 0.80$} \\
\midrule
\multirow{2}{*}{Pseudogenes}
& processed   & $60.66 \pm 0.82$ & $68.30 \pm 0.93$ & \cellcolor{HilightColor} \firstresults{$71.27 \pm 0.78$} \\
& unprocessed & $51.94 \pm 2.69$ & $56.10 \pm 0.56$ & \cellcolor{HilightColor} \firstresults{$62.31 \pm 0.78$} \\
\bottomrule
\end{tabular}
}
\end{table}
\FloatBarrier

\section{Hyperbolic Busemann Neural Networks}
\label{sec:ch5-hbnn}

\subsection{Introduction}
\label{hbnn:sec:intro}

The preceding section introduced the unconstrained PV representation and its neural layers, based on geodesic hyperplanes. We next use another powerful geometric tool, the Busemann function, to construct hyperbolic neural networks. For simplicity, we focus on the widely used Poincar\'e ball and Lorentz models.

To support deep learning fully in hyperbolic spaces, several key building blocks in neural networks have recently been generalized to Poincaré or Lorentz spaces, including attention \citep{gulcehre2019hyperbolic,chen2022fully,yang2024hypformer}, BN \citep{lou2020differentiating,bdeir2024fully,chen2025gyrobn,chen2025gyrobnextension}, linear feed-forward layers \citep{ganea2018hyperbolic,shimizu2021hyperbolic,chen2022fully}, activation \citep{ganea2018hyperbolic,bdeir2024fully}, residual blocks \citep{van2023poincare,katsman2024riemannian,he2025lorentzian}, MLR \citep{shimizu2021hyperbolic,bdeir2024fully,nguyen2025neural}, and graph convolution \citep{chami2019hyperbolic,liu2019hyperbolic,bachmann2020constant,dai2021hyperbolic}. Among these components, MLR classification and FC layers play a fundamental role in final decision-making and feature transformation.

Recently, hyperplanes and point-to-hyperplane distances, which have been explored in \cref{chapter:rmlr,sec:ch5-pvnn}, have been adopted to construct hyperbolic MLR in both Poincaré \citep{ganea2018hyperbolic,shimizu2021hyperbolic} and Lorentz \citep{bdeir2024fully} models. \citet[Sec.~3.1]{ganea2018hyperbolic} introduced the first Poincaré MLR based on Poincaré hyperplanes, but the formulation suffers from over-parameterization and lacks batch efficiency. \citet[Sec.~3.1]{shimizu2021hyperbolic} alleviated these issues through a re-parameterization strategy. Building on these ideas, \citet[Sec.~4.3]{bdeir2024fully} proposed a Lorentz MLR. However, its hyperplane is defined by the ambient Minkowski space, which is model-specific and may distort Lorentzian geometry.

For hyperbolic FC layers, three main formulations exist. \citet[Sec.~3.2]{ganea2018hyperbolic} introduced Möbius matrix-vector multiplication through the tangent space on the Poincaré ball. \citet[Sec.~3.2]{shimizu2021hyperbolic} further proposed the Poincaré FC layer, defined intrinsically but restricted to the Poincaré model. On the Lorentz model, \citet[Sec.~2.2]{chen2022fully} constructed a Lorentz FC layer by applying linear transformations in the ambient Minkowski space followed by projection onto the Lorentz model. Thus, Möbius and Lorentz FC rely on flat-space (tangent or ambient) approximations that could distort intrinsic geometry, whereas Poincaré FC is intrinsic but model-specific.

On the other hand, the Busemann function and its level sets, horospheres, have emerged as powerful intrinsic tools for hyperbolic learning. They enjoy convenient metric properties \citep[Ch.~II.8]{bridson2013metric} and admit closed-form expressions on both the Poincaré and Lorentz models \citep[Prop.~9]{bonet2025sliced}. These operators have supported several hyperbolic algorithms, including SVM \citep{fan2023horospherical}, PCA \citep{chami2021horopca}, Sliced Wasserstein distances \citep{bonet2025sliced}, and prototype learning \citep{ghadimi2021hyperbolic}. We also note that \citet[Cor.~4.3]{nguyen2025neural} proposed a Poincaré MLR based on the Busemann function. However, its induced point-to-hyperplane distance is pseudo, coincides with the true distance only in Euclidean geometry, remains over-parameterized, and is not batch efficient.

These observations motivate intrinsic and batch-efficient formulations for MLR and FC layers that can operate on both the Poincaré ball and the Lorentz model. To address this need, we propose \emph{Busemann Multinomial Logistic Regression (BMLR)} and \emph{Busemann Fully Connected (BFC)} layers, two Busemann-based components for hyperbolic networks. Our contributions are summarized as follows:
\begin{itemize}
\item We introduce BMLR, deriving intrinsic logits directly from Busemann functions with a point-to-horosphere distance interpretation. BMLR uses a compact per-class parameterization, eliminates manifold-valued parameters in prior MLRs, remains batch-efficient, and recovers Euclidean MLR as curvature tends to zero.
\item We develop BFC layers by generalizing the FC and activation layers through the Busemann function, providing intrinsic constructions on both the Poincaré and Lorentz models. BFC preserves comparable complexity and parameter counts, and recovers Euclidean FC in the zero curvature limit.
\item We provide empirical validation across image classification, genome sequence learning, node classification, and link prediction. BMLR and BFC generally outperform existing hyperbolic layers. BMLR shows particularly large gains as the number of classes increases, and the Lorentz BMLR is the fastest among all hyperbolic MLRs.\footnote{The code is available at \url{https://github.com/GitZH-Chen/HBNN}.}
\end{itemize}

\mypara{Outline.} In \cref{hbnn:subsec:preliminaries}, we recall Busemann functions and horospheres in hyperbolic space. In \cref{hbnn:subsec:bmlr}, we introduce BMLR and its point-to-horosphere interpretation. In \cref{hbnn:subsec:bfc}, we develop BFC layers, and in \cref{hbnn:subsec:experiments}, we evaluate both components. Proofs are deferred to \cref{app:hbnn-proofs}.

\subsection{Preliminaries}
\label{hbnn:subsec:preliminaries}

The metric-geometric notions of geodesic rays, asymptotic rays, Busemann functions, horoballs, horospheres, and Hadamard spaces have been reviewed in \cref{sec:ch2-metric-geometry}. The Poincaré and Lorentz models and their Riemannian operators have been reviewed in \cref{sec:ch2-constant-curvature-manifolds}. Their gyro operators are presented in \cref{tab:ch2-constant-curvature-gyro-operators,subsec:radius-model}, respectively. Their gyrovector spaces are denoted by $\{\pball{n}, \Moplus, \Modot\}$ and $\{\lorentz{n}, \Loplus, \Lodot\}$, respectively. In the following, we review the Busemann function on the hyperbolic space.

In Euclidean space, the Busemann function associated with the geodesic $\gamma(t)=t v$ that starts at $\zerovec$ with unit direction $v \in \unitsphere{n-1}$ is $B^v(x)=-\inner{x}{v}$, which coincides, up to a sign, with the inner product. Let $\hyperspace{n}\in\{\pball{n},\lorentz{n}\}$ be a hyperbolic space. We write $B^v(x)$ for the Busemann function associated with the ray that emanates from the origin $e \in \hyperspace{n}$ in the direction $v \in \unitsphere{n-1} \subset T_e\hyperspace{n}$. For $K<0$, $v \in \unitsphere{n-1}$, and $x \in \hyperspace{n}$, closed forms of the Poincaré and Lorentz Busemann functions \citep[Prop.~9]{bonet2025sliced} are
\begin{align}
\label{hbnn:eq:poincare-busemann}
\pball{n}:\quad B^{v}(x)
&=\frac{1}{\sqrt{-K}}\log\left(\frac{\norm{v-\sqrt{-K}x}^{2}}{1+K\norm{x}^{2}}\right),\\
\label{hbnn:eq:lorentz-busemann}
\lorentz{n}:\quad B^{v}(x)
&=\frac{1}{\sqrt{-K}}\log\left(\sqrt{-K}\left(x_t-\inner{x_s}{v}\right)\right).
\end{align}

\begin{figure}[t]
\centering
\includegraphics[width=0.8\linewidth,trim={0cm 0cm 0cm 0cm}]{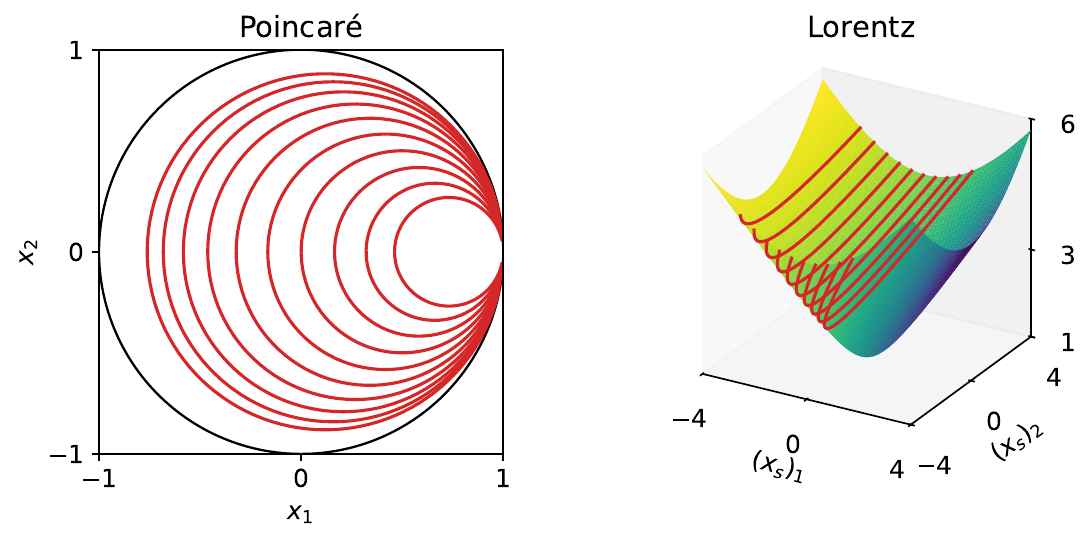}
\caption{Illustration: red curves are different horospheres of $B^{v}$.}
\label{hbnn:fig:horospheres}
\end{figure}

The level sets of a Busemann function are horospheres, the hyperbolic counterpart of Euclidean hyperplanes. In Euclidean space, for a unit direction $v$, the hyperplanes $H^v_\tau=\{x\in\bbR{n}\mid\inner{x}{v}=\tau\}$ with $\tau\in\bbRscalar$ are parallel. Analogously, for fixed $v$, the horospheres $H^v_\tau=\{x\in\hyperspace{n}\mid B^v(x)=\tau\}$ are equidistant, as established later in \cref{hbnn:thm:distance-horospheres}. \cref{hbnn:fig:horospheres} illustrates such horospheres, and \cref{tab:ch2-metric-geometry-examples} summarizes the above correspondence between Euclidean and hyperbolic notions.

\subsection{Busemann Multinomial Logistic Regression}
\label{hbnn:subsec:bmlr}
We begin by reformulating the Euclidean MLR, then lift it to hyperbolic space via the Busemann function, introducing BMLR. We also present a point-to-horosphere interpretation. Finally, we compare BMLR with existing hyperbolic MLRs, highlighting our advantages in geometric fidelity, parameterization, and computational efficiency.

\subsubsection{Formulation}
\label{hbnn:subsubsec:bmlr-formulation}
The Euclidean MLR $\softmax(Ax+b)$ computes the multinomial probability for each class $k\in\{1,\ldots,C\}$ given an input $x\in\bbR{n}$. It admits the inner product form:
\begin{equation}
\label{hbnn:eq:euc-mlr}
\forall k,\quad p(y=k\mid x)=\frac{\exp\left(\inner{a_k}{x}+b_k\right)}{\sum_{j=1}^C\exp\left(\inner{a_j}{x}+b_j\right)},
\end{equation}
where $a_k\in\bbR{n}$ and $b_k\in\bbRscalar$ are the weight and bias for class $k$. We write $p(y=k\mid x)\propto\exp\left(u_k(x)\right)$ with $u_k(x)=\inner{a_k}{x}+b_k$. Decomposing the weight vector into a magnitude $\alpha_k=\norm{a_k}>0$ and a unit direction $v_k=\frac{a_k}{\norm{a_k}}\in\unitsphere{n-1}$, each logit is
\begin{equation}
\label{hbnn:eq:vk-decomp}
u_k(x)=\alpha_k\inner{v_k}{x}+b_k.
\end{equation}

As reviewed in \cref{hbnn:subsec:preliminaries}, the Busemann function naturally generalizes the Euclidean inner product. Analogously to \cref{hbnn:eq:vk-decomp}, we define the hyperbolic logits via the Busemann function, yielding BMLR:
\begin{align}
\forall k,\quad p(y=k\mid x)
&=\frac{\exp\left(u_k(x)\right)}{\sum_{j=1}^C\exp\left(u_j(x)\right)},\\
\label{hbnn:eq:b-logits}
u_k(x)&=-\alpha_kB^{v_k}(x)+b_k,
\end{align}
with $\alpha_k>0$, $v_k\in\unitsphere{n-1}$, and $b_k\in\bbRscalar$ as parameters.

The following result shows that, as $K\to0^{-}$, both the Poincaré and Lorentz BMLRs reduce to the Euclidean MLR.

\begin{paristheorem}[Limits of BMLRs]
\label{hbnn:thm:limits-bmlr}
\linktoproof{hbnn:thm:limits-bmlr}
As $K\to0^{-}$, the hyperbolic Busemann functions converge to the Euclidean inner product:
\begin{align}
\text{Poincaré:}\quad &B^v(x)\xrightarrow{K\to0^{-}}-2\inner{v}{x},\\
\label{hbnn:eq:lorentz-b-limits}
\text{Lorentz:}\quad &B^v(x)\xrightarrow{K\to0^{-}}-\inner{v}{x_s}.
\end{align}
The hyperbolic BMLRs converge to the Euclidean MLR:
\begin{align}
\text{Poincaré:}\quad &u_k(x)\xrightarrow{K\to0^{-}}2\alpha_k\inner{v_k}{x}+b_k,\\
\text{Lorentz:}\quad &u_k(x)\xrightarrow{K\to0^{-}}\alpha_k\inner{v_k}{x_s}+b_k.
\end{align}
\end{paristheorem}

\begin{parisremark}[Intuition]
On the Poincaré ball, letting $K\to0^{-}$ recovers Euclidean geometry \citep[App.~A.4.2]{skopek2020mixed}. For the Lorentz model, as $K\to0^{-}$, the temporal coordinate diverges while the spatial component approaches $\bbR{n}$, making $\lorentz{n}$ converge to a Euclidean space. Consistently, the Poincaré and Lorentz Busemann functions and the associated BMLR logits reduce to their Euclidean counterparts, providing a natural generalization of Euclidean MLR.
\end{parisremark}

\subsubsection{Geometric Interpretation}
\label{hbnn:subsubsec:bmlr-interpretation}
The point-to-hyperplane strategy underlying MLR has been established in \cref{spdmlr:subsec:reform_emlr}. We now show that BMLR admits the corresponding interpretation through point-to-horosphere distances.

\begin{paristheorem}[Hadamard horosphere distance]
\label{hbnn:thm:distance-horospheres}
\linktoproof{hbnn:thm:distance-horospheres}
Let $(\calX,\dist)$ be a geodesically complete Hadamard space, and let $B^\gamma:\calX\to\bbRscalar$ be the Busemann function associated with a geodesic ray $\gamma:[0,\infty)\to\calX$. For any $\tau_1,\tau_2\in\bbRscalar$, define the horospheres by
\begin{equation}
H^\gamma_{\tau_i}=\left\{x\in\calX\mid B^\gamma(x)=\tau_i\right\},\qquad i=1,2.
\end{equation}
The distance between these horospheres is constant:
\begin{equation}
\label{hbnn:eq:equidistance}
\dist\left(H^\gamma_{\tau_1},H^\gamma_{\tau_2}\right)=\dist\left(H^\gamma_{\tau_2},H^\gamma_{\tau_1}\right)=\left|\tau_2-\tau_1\right|.
\end{equation}
In particular, the point-to-horosphere distance is
\begin{equation}
\dist\left(x,H^\gamma_\tau\right)=\left|B^\gamma(x)-\tau\right|,\qquad\forall x\in\calX.
\end{equation}
\end{paristheorem}

\begin{pariscorollary}[Point-to-horosphere distance]
\label{hbnn:cor:poincare-lorentz}
In a hyperbolic space
\begin{equation}
\hyperspace{n}\in\{\pball{n},\lorentz{n}\}
\end{equation}
with curvature $K<0$, the point-to-horosphere distance is
\begin{equation}
\dist\left(x,H^v_\tau\right)=\left|B^v(x)-\tau\right|,
\end{equation}
where $H^v_\tau=\left\{x\mid B^v(x)=\tau\right\}$ denotes the horosphere with respect to direction $v\in\unitsphere{n-1}$.
\end{pariscorollary}

A Euclidean hyperplane can be parameterized by a unit direction, a positive magnitude, and a scalar bias. Similarly, we parameterize a hyperbolic horosphere as
\begin{equation}
\label{hbnn:eq:horosphere-param}
H_{v,\alpha,b}=\left\{x\in\hyperspace{n}\mid-\alpha B^v(x)+b=0\right\},
\end{equation}
with $v\in\unitsphere{n-1}$, $\alpha>0$, and $b\in\bbRscalar$. With this parameterization, the signed point-to-horosphere logit is
\begin{equation}
\label{hbnn:eq:horosphere-mlr}
u_k(x)=\sign_k\alpha_k\dist\left(x,H_{v_k,\alpha_k,b_k}\right),
\end{equation}
where $\sign_k=\sign\left(-\alpha_kB^{v_k}(x)+b_k\right)$. By \cref{hbnn:cor:poincare-lorentz}, the exact point-to-horosphere distance is
\begin{equation}
\label{hbnn:eq:point-to-horosphere-dist-param}
\dist\left(x,H_{v,\alpha,b}\right)=\frac{\left|-\alpha B^v(x)+b\right|}{\alpha}.
\end{equation}
Consequently, \cref{hbnn:eq:horosphere-mlr} equals the BMLR logit in \cref{hbnn:eq:b-logits}.

\begin{parisremark}[Generality]
Since $B^v(x)=-\inner{v}{x}$ in Euclidean geometry, \cref{hbnn:eq:horosphere-param,hbnn:eq:point-to-horosphere-dist-param,hbnn:eq:horosphere-mlr} naturally generalize to their Euclidean counterparts. We also acknowledge \citet[Eq.~(2) and Prop.~3.1]{fan2023horospherical}, who used horospheres and point-to-horosphere distances to construct a hyperbolic SVM. However, they considered only the unit Poincaré ball with curvature $K=-1$, which is a special case of \cref{hbnn:eq:horosphere-param,hbnn:eq:point-to-horosphere-dist-param}.
\end{parisremark}

\begin{table}[t]
    \centering
    \caption{Comparison of $C$-class MLR. In Dist, \greenbf{Real} means the point-to-hyperplane distance is the real distance, obtained by $\inf_{y \in H} \dist(x,y)$, where $H$ is a hyperplane and $\dist$ is the geodesic distance; \redbf{Pseudo} denotes a surrogate that coincides with the real distance only in Euclidean geometry. Compact params indicate whether each logit avoids an additional manifold-valued parameter. Batch efficiency indicates whether the MLR can avoid inefficient per-class loops in implementation (see \cref{hbnn:app:subsubsec:mlr-comparison}). In \#Params, we highlight the heaviest in \redbf{red}. In FLOPs, we mark the slowest in \redbf{red} and the fastest in \greenbf{green}.}
    \label{hbnn:tab:logit-comparison}
    \resizebox{\linewidth}{!}{
    \begin{tabular}{cccccccc}
        \toprule
        \textbf{Method} & \textbf{Logit $u_k(x), \ \forall k \in \{1,\ldots, C\}$} & \textbf{Space} & \textbf{Dist} & \textbf{\#Params} & \makecell{\textbf{Compact} \\ \textbf{params}} & \textbf{FLOPs} & \makecell{\textbf{Batch} \\ \textbf{efficiency}}\\
        \midrule
        Euclidean MLR & $\inner{a_k}{x} + b_k$, with $a_k \in \bbR{n}, b_k \in \bbRscalar$ & $\bbR{n}$ & \greenbf{Real} & $C(n+1)$ & \cmark & $C(2n)$ & \cmark \\
        \midrule
        \makecell{Poincaré MLR \\ \citep[Eq.~(25)]{ganea2018hyperbolic}} & \makecell{$\displaystyle \frac{\lambda_{p_k}^{K} \norm{a_k}}{\sqrt{-K}} \sinh^{-1}\left(\frac{2\sqrt{-K} \inner{-p_k \Moplus x}{a_k}}{\left(1 + K \norm{-p_k \Moplus x}^{2}\right) \norm{a_k}}\right)$, \\ with $p_k \in \pball{n}, a_k \in T_{p_k} \pball{n}$} & $\pball{n}$ & \greenbf{Real} & $C(2n)$ & \xmark & $C(19n+29)$ & \xmark \\
        \midrule
        \makecell{Poincaré MLR \\ \citep[Eq.~(6)]{shimizu2021hyperbolic}} & \makecell{$\displaystyle \frac{2}{\sqrt{-K}} \alpha_k \sinh^{-1}\left( \alpha - \beta \right)$, \\ $\alpha = \lambda_{x}^{K} \sqrt{-K} \inner{x}{v_k} \cosh\left(2\sqrt{-K} b_k\right)$, \\ $\beta = \left(\lambda_{x}^{K} - 1\right) \sinh\left(2\sqrt{-K} b_k\right)$, \\ with $\alpha_k > 0, v_k \in \unitsphere{n-1}, b_k \in \bbRscalar$} & $\pball{n}$ & \greenbf{Real} & $C(n+2)$ & \cmark & $C(4n+52)$ & \cmark \\
        \midrule
        \makecell{Pseudo-Busemann MLR \\ \citep[Cor.~4.3]{nguyen2025neural}} & \makecell{$\displaystyle - \dist(x, p_k) \frac{B^{v_k}\left(- p_k \Moplus x\right)}{\norm{- p_k \Moplus x}}$, \\ with $p_k \in \pball{n}, v_k \in \unitsphere{n-1}$} & $\pball{n}$ & \redbf{Pseudo} & \redbf{$C(2n)$} & \xmark & \redbf{$C(19n+34)$} & \xmark \\
        \midrule
        \makecell{ Lorentz MLR \\ \citep[Eq.~(12)]{bdeir2024fully}} & 
        \makecell{$\displaystyle \frac{1}{\sqrt{-K}} \sign(\alpha) \beta\left|\sinh ^{-1}\left(\sqrt{-K} \frac{\alpha}{\beta}\right)\right|$, \\
        $\alpha=\cosh\left(\sqrt{-K} b_k\right)\inner{z_k}{x_s}-\sinh\left(\sqrt{-K} b_k\right)$, \\
        $\beta=\sqrt{ \norm{\cosh (\sqrt{-K} b_k) z_k }^2-(\sinh (\sqrt{-K} b_k)\norm{z_k})^2}$, \\ with $z_k \in \bbR{n},\ b_k \in \bbRscalar$}
        & $\lorentz{n}$ & \greenbf{Real} & $C(n+1)$ & \cmark & $C(4n+52)$ & \cmark \\
        \midrule
        \rowcolor{HilightColor} BMLR & \Gape[0pt][2pt]{\makecell{$-\alpha_k B^{v_k}(x) + b_k$, \\ with $\alpha_k > 0, v_k \in \unitsphere{n-1}, b_k \in \bbRscalar$}} & \Gape[0pt][2pt]{\makecell{$\pball{n}$ \\ $\lorentz{n}$}} & \greenbf{Real} & $C(n+2)$ & \cmark & \makecell{ $\pball{n}: C(6n+12)$ \\ \greenbf{$\lorentz{n}: C(2n+12)$} } & \cmark \\
        \bottomrule
    \end{tabular}
    }
\end{table}

\subsubsection{Comparison with Existing Hyperbolic MLRs}
\label{hbnn:subsubsec:bmlr-comparison}
Based on the point-to-hyperplane reformulation, recent work extended MLR to the Poincaré \citep{ganea2018hyperbolic,shimizu2021hyperbolic,nguyen2025neural} and Lorentz \citep{bdeir2024fully} models. \citet[Sec.~3.1]{ganea2018hyperbolic} introduced the first Poincaré MLR by replacing the Euclidean point-to-hyperplane distance with its hyperbolic counterpart, where the hyperplane is defined by geodesics and the resulting distance is the real point-to-hyperplane distance, obtained as an infimum over the hyperplane. However, the formulation is not batch efficient (see \cref{hbnn:app:subsubsec:mlr-comparison}). It also requires per-class parameters $a_k\in T_{p_k}\pball{n}$ and $p_k\in\pball{n}$, which leads to over-parameterization. \citet[Sec.~3.1]{shimizu2021hyperbolic} alleviated such issues via re-parameterization. \citet[Sec.~4.3]{bdeir2024fully} further developed a Lorentz MLR, but its hyperplanes are defined by the ambient Minkowski space, which is tailored to the Lorentz model and does not fully respect intrinsic hyperbolic geometry. Moreover, \citet[Cor.~4.3]{nguyen2025neural} proposed a Poincaré MLR based on the Busemann function. We refer to it as Pseudo-Busemann MLR, as the induced point-to-hyperplane distance is pseudo, coinciding with the real point-to-hyperplane distance only in Euclidean geometry. It also suffers from over-parameterization and is not batch efficient.

As summarized in \cref{hbnn:tab:logit-comparison},\footnote{Relative to \citep[Def.~4.2, Cor.~4.3, and App.~B.1.2]{nguyen2025neural}, the Pseudo-Busemann MLR written here includes an additional sign $-$; this is intentional and matches their official implementation.} BMLR unifies advantages that prior hyperbolic MLRs offer only partially. In particular, BMLR respects the real point-to-horosphere distance, uses compact parameters without an additional manifold-valued point, attains the lowest FLOPs on $\lorentz{n}$ and a competitive cost on $\pball{n}$, and supports batch-efficient computation. On $\lorentz{n}$, its FLOPs are even close to those of the Euclidean MLR.

\subsection{Busemann Fully Connected Layer}
\label{hbnn:subsec:bfc}
We first reformulate the Euclidean FC layer by the Busemann function, then present the manifestations in the Poincaré and Lorentz models.

\subsubsection{Formulation}
\label{hbnn:subsubsec:bfc-formulation}
As discussed in \cref{pvnn:subsec:pv-fc-hnnpp}, the Euclidean FC layer can be written as
\begin{equation}
\label{hbnn:eq:euc-fc-p2h}
\bar{\dist}\left(y,H_{e_k,\zerovec}\right)=\inner{a_k}{x}+b_k,\qquad\forall k\in\{1,\ldots,m\},
\end{equation}
where $\bar{\dist}\left(y,H_{e_k,\zerovec}\right)=\sign\left(\inner{e_k}{y - \zerovec}\right) \dist (y, H_{e_k, \zerovec})$ is the signed distance. To extend \cref{hbnn:eq:euc-fc-p2h} into hyperbolic space, the right-hand side can be replaced by \cref{hbnn:eq:b-logits}, as it generalizes $\inner{a_k}{x}+b_k$. For the left-hand side, a natural idea is to use the signed point-to-horosphere distance. However, as detailed in \cref{hbnn:app:subsec:busemann-fc-p2h-distance}, this may fail to admit a solution for $y$. We therefore follow the point-to-hyperplane distance in \citep[Thm.~5]{ganea2018hyperbolic} for the Poincaré model and the one in \citep[Eq.~(44)]{bdeir2024fully} for the Lorentz model. Given $x\in\hyperspace{n}$, the hyperbolic BFC layer $\calF:\hyperspace{n}\ni x\mapsto y\in\hyperspace{m}$ is given by solving $y$ via the following $m$ equations:
\begin{equation}
\label{hbnn:eq:hyp-fc-p2h}
\bar{\dist}\left(y,H_{e_k,e}\right)=u_k(x),\qquad\forall k\in\{1,\ldots,m\},
\end{equation}
where $u_k(x)=-\alpha_kB^{v_k}(x)+b_k$ with $\{\alpha_k>0,v_k\in\unitsphere{n-1},b_k\in\bbRscalar\}$ as parameters. Here, $\bar{\dist}\left(y,H_{e_k,e}\right)$ is the hyperbolic signed distance from $y$ to the hyperplane passing through the output-space origin $e\in\hyperspace{m}$. Next, we show that the above implicit definition has an explicit solution for the output $y$.

\begin{paristheorem}[Poincaré BFC]
\label{hbnn:thm:bfc-poincare}
\linktoproof{hbnn:thm:bfc-poincare}
Given an input $x\in\pball{n}$, the Poincaré BFC layer $\calF:\pball{n}\to\pball{m}$ is given by
\begin{equation}
y=\frac{\omega}{1+\sqrt{1-K\norm{\omega}^{2}}},\qquad
\omega=\left[\frac{\sinh\left(\sqrt{-K}u_k(x)\right)}{\sqrt{-K}}\right]_{k=1}^m,
\end{equation}
where $u_k(x)=-\alpha_kB^{v_k}(x)+b_k$ with $\{\alpha_k>0,v_k\in\unitsphere{n-1},b_k\in\bbRscalar\}$ as parameters for $k=1,\ldots,m$.
\end{paristheorem}

\begin{paristheorem}[Lorentz BFC]
\label{hbnn:thm:bfc-lorentz}
\linktoproof{hbnn:thm:bfc-lorentz}
Given an input $x\in\lorentz{n}$, the Lorentz BFC layer $\calF:\lorentz{n}\to\lorentz{m}$ is given by
\begin{equation}
y=\begin{bmatrix}y_t\\y_s\end{bmatrix}
=\begin{bmatrix}
\sqrt{\frac{1}{-K}+\norm{y_s}^{2}}\\[6pt]
\frac{1}{\sqrt{-K}}\sinh\left(\sqrt{-K}u(x)\right)
\end{bmatrix},
\end{equation}
where $u(x)=\left(u_1(x),\ldots,u_m(x)\right)^\top$ with $u_k(x)=-\alpha_kB^{v_k}(x)+b_k$. Here, $\{\alpha_k>0,v_k\in\unitsphere{n-1},b_k\in\bbRscalar\}$ are parameters for $k=1,\ldots,m$.
\end{paristheorem}

Analogously to \cref{hbnn:thm:limits-bmlr}, our BFC layers converge to their Euclidean counterparts as $K\to0^{-}$.
\begin{paristheorem}[Limits of BFC layers]
\label{hbnn:thm:limits-bfc}
\linktoproof{hbnn:thm:limits-bfc}
As $K\to0^{-}$, the hyperbolic BFC layer $\hyperspace{n}\ni x\mapsto y\in\hyperspace{m}$ reduces to a Euclidean FC layer:
\begin{align}
\text{Poincaré:}\quad &y_k\xrightarrow{K\to0^{-}}\alpha_k\inner{v_k}{x}+\frac{1}{2}b_k,\\
\text{Lorentz:}\quad &(y_s)_k\xrightarrow{K\to0^{-}}\alpha_k\inner{v_k}{x_s}+b_k.
\end{align}
\end{paristheorem}

\subsubsection{Generalization}
\label{hbnn:subsubsec:busemann-fc-generalization}
Following \cref{pvnn:subsec:pv-fc-hnnpp}, we extend the hyperbolic BFC by inserting the activation into \cref{hbnn:eq:hyp-fc-p2h}:
\begin{equation}
\bar{\dist}\left(y,H_{e_k,e}\right)=\phi\left(u_k(x)\right),\qquad\forall k\in\{1,\ldots,m\}.
\end{equation}
This is reflected in \cref{hbnn:thm:bfc-poincare,hbnn:thm:bfc-lorentz} by replacing every $u_k(x)$ with $\phi\left(-\alpha_kB^{v_k}(x)+b_k\right)$. Moreover, inspired by the Poincaré Möbius transformation \citep[Sec.~3.2]{ganea2018hyperbolic}, a BFC transformation could be further followed by a gyroaddition $\Hoplus$: $\hyperspace{n}\ni x\mapsto\calF(x)\Hoplus b\in\hyperspace{m}$ with $b\in\hyperspace{m}$ as a gyro bias. For example, the Lorentz BFC layer is generalized as
\begin{equation}
\lorentz{n}\ni x\mapsto y=
\begin{bmatrix}
\sqrt{\frac{1}{-K}+\norm{y_s}^{2}}\\[6pt]
\frac{1}{\sqrt{-K}}\sinh\left(\sqrt{-K}u(x)\right)
\end{bmatrix}\Loplus b\in\lorentz{m},
\end{equation}
where $u_k(x)=\phi\left(-\alpha_kB^{v_k}(x)+b_k\right)$ with parameters $\{\alpha_k>0,v_k\in\unitsphere{n-1},b_k\in\bbRscalar\}_{k=1}^m$ and $b\in\lorentz{m}$.

\begin{table}[t]
\centering
\caption{Comparison of hyperbolic FC layers. For simplicity, BFC layers do not involve the gyroaddition and assume $\phi$ is the identity map, which is in line with the Möbius and Lorentz FC layers.}
\label{hbnn:tab:fc-comparison}
\resizebox{\linewidth}{!}{%
\begin{tabular}{ccccccc}
\toprule
\textbf{Method} & $\calF:\hyperspace{n}\ni x\mapsto y\in\hyperspace{m}$ & \textbf{Space} & \textbf{Methodology} & \textbf{Parameters} & \textbf{\#Params} & \textbf{FLOPs}\\
\midrule
\makecell{Möbius\\\citep[Eq.~(27)]{ganea2018hyperbolic}} & $\displaystyle\frac{1}{\sqrt{-K}}\tanh\left(\frac{\norm{Wx}}{\norm{x}}\tanh^{-1}\left(\sqrt{-K}\norm{x}\right)\right)\frac{Wx}{\norm{Wx}}$ & $\pball{n}$ & Tangent & $W\in\bbR{m\times n}$ & $mn$ & \makecell{$2nm+2n$\\$+2m+24$}\\
\midrule
\makecell{Poincaré FC\\\citep[Eq.~(7)]{shimizu2021hyperbolic}} & \makecell{$\displaystyle y=\frac{\omega}{1+\sqrt{1-K\norm{\omega}^{2}}}$, $\displaystyle\omega_k=\frac{\sinh\left(\sqrt{-K}u_k(x)\right)}{\sqrt{-K}}$,\\with $u_k(x)$ in \cref{hbnn:tab:logit-comparison}} & $\pball{n}$ & \makecell{Poincaré\\geometry} & \makecell{$\alpha_k>0,v_k\in\unitsphere{n-1}$,\\$b_k\in\bbRscalar$, $k=1,\ldots,m$} & $m(n+2)$ & $4nm+71m+4$\\
\midrule
\makecell{Lorentz FC\\\citep[Eq.~(3)]{chen2022fully}} & \makecell{$\displaystyle y=\begin{bmatrix}\sqrt{\norm{\psi(Wx,v)}^2-1/K}\\\psi(Wx,v)\end{bmatrix}$,\\$\displaystyle\psi(Wx,v)=\lambda\sigma\left(v^\top x+b'\right)\frac{W\phi(x)+b}{\norm{W\phi(x)+b}}$} & $\lorentz{n}$ & \makecell{Ambient\\Minkowski} & \makecell{$W\in\bbR{m\times(n+1)}$,\\$v\in\bbR{n+1}$, $b\in\bbR{m}$, $b'\in\bbRscalar$, $\lambda>0$} & \makecell{$m(n+1)+m$\\$+(n+1)+2$} & \makecell{$2nm+8m$\\$+2n+10$}\\
\midrule
\rowcolor{HilightColor} BFC & \makecell{$\displaystyle y=\frac{\omega}{1+\sqrt{1-K\norm{\omega}^{2}}}$, $\displaystyle\omega=\frac{\sinh\left(\sqrt{-K}u(x)\right)}{\sqrt{-K}}$;\\$\displaystyle y_s=\frac{1}{\sqrt{-K}}\sinh\left(\sqrt{-K}u(x)\right)$, $\displaystyle y_t=\sqrt{\frac{1}{-K}+\norm{y_s}^{2}}$,\\with $\displaystyle u_k(x)=\phi\left(-\alpha_kB^{v_k}(x)+b_k\right)$} & \makecell{$\pball{n}$\\$\lorentz{n}$} & Busemann & \makecell{$\alpha_k>0,v_k\in\unitsphere{n-1}$,\\$b_k\in\bbRscalar$, $k=1,\ldots,m$} & $m(n+2)$ & \makecell{$6nm+29m+4$\\$2nm+30m+2$}\\
\bottomrule
\end{tabular}}
\end{table}

\subsubsection{Comparison with Existing Hyperbolic FC Layers}
\label{hbnn:subsubsec:bfc-comparison}
\cref{hbnn:tab:fc-comparison} compares BFC with prior hyperbolic FC layers. BFC faithfully respects hyperbolic geometry, whereas the Möbius and Lorentz FC layers apply Euclidean transformations in the tangent or ambient Minkowski space, which can distort intrinsic geometry. BFC also offers flexibility across models, while Poincaré FC and Lorentz FC are tailored to their respective models. In addition, BFC uses a comparable parameterization and maintains $\calO(nm)$ FLOPs. On $\lorentz{n}$, its FLOPs are $\calO(2mn)$, matching the fastest layers.

\subsection{Experiments}
\label{hbnn:subsec:experiments}

We first compare BMLRs with prior hyperbolic MLRs on three architectures: ResNet-18 (image classification), CNN (genome sequences), and HGCN (node classification). We then compare BFC with prior hyperbolic FC layers on link prediction. All experiments use both the Poincaré and Lorentz models. More details on data sets and experimental settings are provided in \cref{app:datasets,hbnn:app:experimental-details}.

\begin{table}[t]
\centering
  \caption{Top-1 image classification accuracy (\%) of MLR methods on the ResNet-18 backbone. The best results within each hyperbolic model are \firstresults{bold}. The slowest MLR and largest parameter count are shown in \redbf{red}.}
\label{hbnn:tab:exp-image-classification}
\resizebox{\linewidth}{!}{%
\begin{tabular}{c|c|ccc|ccc|ccc|ccc}
\toprule
\multirow{2}[4]{*}{\textbf{Space}} & \multirow{2}[4]{*}{\textbf{Method}} & \multicolumn{3}{c|}{\makecell{\textbf{CIFAR-10}\\(Num. classes: 10)}} & \multicolumn{3}{c|}{\makecell{\textbf{CIFAR-100}\\(Num. classes: 100)}} & \multicolumn{3}{c|}{\makecell{\textbf{Tiny-ImageNet}\\(Num. classes: 200)}} & \multicolumn{3}{c}{\makecell{\textbf{ImageNet-1k}\\(Num. classes: 1000)}}\\
\cmidrule{3-14}
& & \textbf{Acc} & \textbf{Fit Time} & \textbf{\#Params} & \textbf{Acc} & \textbf{Fit Time} & \textbf{\#Params} & \textbf{Acc} & \textbf{Fit Time} & \textbf{\#Params} & \textbf{Acc} & \textbf{Fit Time} & \textbf{\#Params}\\
\midrule
$\bbR{n}$ & MLR & 95.14 ± 0.12 & 10.66 & 5.13K & 77.72 ± 0.15 & 10.60 & 51.30K & 65.19 ± 0.12 & 69.17 & 102.60K & 71.87 & 2263.12 & 513K\\
\midrule
\multirow{3}[2]{*}{$\pball{n}$} & PMLR & 95.04 ± 0.13 & 11.94 & 5.14K & 77.19 ± 0.50 & 12.11 & 51.40K & 64.93 ± 0.38 & 71.90 & 102.80K & 71.77 & 2300.11 & 514K\\
& PBMLR-P & 95.23 ± 0.08 & \redbf{21.92} & \redbf{10.24K} & 77.78 ± 0.15 & \redbf{76.84} & \redbf{102.40K} & 65.43 ± 0.27 & \redbf{336.58} & \redbf{204.80K} & 71.46 & \redbf{3907.12} & \redbf{1024K}\\
& \cellcolor{HilightColor}BMLR-P & \cellcolor{HilightColor}\firstresults{95.32 ± 0.14} & \cellcolor{HilightColor}12.01 & \cellcolor{HilightColor}5.14K & \cellcolor{HilightColor}\firstresults{78.10 ± 0.35} & \cellcolor{HilightColor}12.13 & \cellcolor{HilightColor}51.40K & \cellcolor{HilightColor}\firstresults{66.16 ± 0.19} & \cellcolor{HilightColor}71.98 & \cellcolor{HilightColor}102.80K & \cellcolor{HilightColor}\firstresults{73.36} & \cellcolor{HilightColor}2300.77 & \cellcolor{HilightColor}514K\\
\midrule
\multirow{2}[2]{*}{$\lorentz{n}$} & LMLR & 94.98 ± 0.12 & 11.55 & 5.13K & 78.03 ± 0.21 & 11.72 & 51.30K & 65.63 ± 0.10 & 69.27 & 102.60K & 72.46 & 2277.17 & 513K\\
& \cellcolor{HilightColor}BMLR-L & \cellcolor{HilightColor}\firstresults{95.25 ± 0.02} & \cellcolor{HilightColor}\firstresults{11.08} & \cellcolor{HilightColor}5.14K & \cellcolor{HilightColor}\firstresults{78.07 ± 0.26} & \cellcolor{HilightColor}\firstresults{11.22} & \cellcolor{HilightColor}51.40K & \cellcolor{HilightColor}\firstresults{65.99 ± 0.14} & \cellcolor{HilightColor}\firstresults{69.19} & \cellcolor{HilightColor}102.80K & \cellcolor{HilightColor}\firstresults{73.24} & \cellcolor{HilightColor}\firstresults{2276.53} & \cellcolor{HilightColor}514K\\
\bottomrule
\end{tabular}}
\end{table}

\begin{figure}[t]
\centering
\includegraphics[width=\linewidth,trim={0cm 0cm 0cm 0cm}]{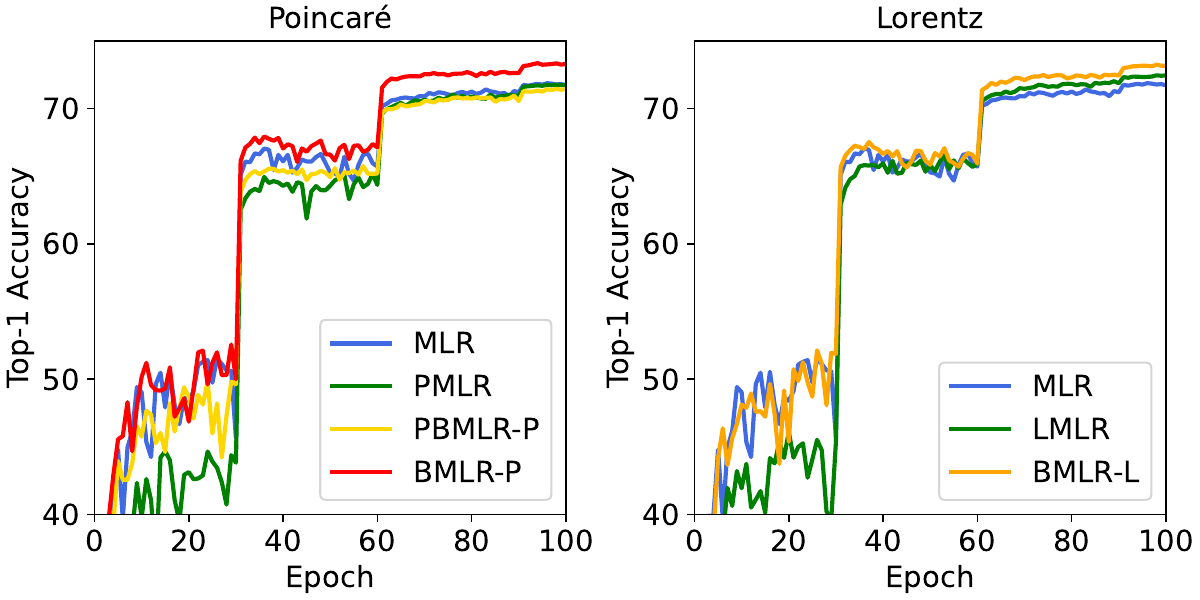}
\caption{Validation accuracy curves on ImageNet-1k.}
\label{hbnn:fig:acc-curves-imagenet}
\end{figure}

\begin{table}[t]
\centering
\caption{Genomic MCC of MLR methods under the CNN backbone. The best results within each hyperbolic model are \firstresults{bold}.}
\label{hbnn:tab:exp-genome-sequence-learning}
\resizebox{\linewidth}{!}{%
\begin{tabular}{ccc|c|ccc|cc}
\toprule
\multirow{2}[3]{*}{\textbf{Benchmark}} & \multirow{2}[3]{*}{\textbf{Task}} & \multirow{2}[3]{*}{\textbf{Data Set}} & \textbf{Num.} & \multicolumn{3}{c|}{$\pball{n}$} & \multicolumn{2}{c}{$\lorentz{n}$}\\
\cmidrule{5-9}
& & & \textbf{classes} & \textbf{PMLR} & \textbf{PBMLR-P} & \cellcolor{HilightColor}\textbf{BMLR-P} & \textbf{LMLR} & \cellcolor{HilightColor}\textbf{BMLR-L}\\
\midrule
\multirow{7}[5]{*}{TEB} & \multirow{3}[1]{*}{Retrotransposons} & LTR Copia & 2 & 75.34 ± 1.02 & 74.37 ± 1.48 & \cellcolor{HilightColor}\firstresults{76.73 ± 1.08} & 73.01 ± 1.07 & \cellcolor{HilightColor}\firstresults{75.86 ± 1.52}\\
& & LINEs & 2 & 85.54 ± 0.61 & 85.92 ± 0.65 & \cellcolor{HilightColor}\firstresults{86.05 ± 1.08} & 83.14 ± 0.80 & \cellcolor{HilightColor}\firstresults{86.72 ± 0.58}\\
& & SINEs & 2 & 95.30 ± 0.85 & 95.34 ± 1.58 & \cellcolor{HilightColor}\firstresults{95.99 ± 0.74} & \firstresults{96.70 ± 0.87} & \cellcolor{HilightColor}96.29 ± 0.59\\
\cmidrule{2-9}
& \multirow{2}[2]{*}{DNA transposons} & CMC-EnSpm & 2 & 83.39 ± 0.56 & 83.62 ± 1.00 & \cellcolor{HilightColor}\firstresults{84.03 ± 0.71} & 81.78 ± 1.05 & \cellcolor{HilightColor}\firstresults{84.15 ± 1.00}\\
& & hAT-Ac & 2 & 89.38 ± 0.90 & \firstresults{89.86 ± 0.54} & \cellcolor{HilightColor}89.62 ± 0.74 & 88.94 ± 0.69 & \cellcolor{HilightColor}\firstresults{90.70 ± 0.51}\\
\cmidrule{2-9}
& \multirow{2}[2]{*}{Pseudogenes} & processed & 2 & 72.45 ± 1.49 & 71.99 ± 2.04 & \cellcolor{HilightColor}\firstresults{73.09 ± 1.66} & \firstresults{73.71 ± 1.76} & \cellcolor{HilightColor}73.32 ± 1.65\\
& & unprocessed & 2 & 75.37 ± 2.27 & 71.99 ± 1.47 & \cellcolor{HilightColor}\firstresults{75.71 ± 1.89} & 74.54 ± 1.98 & \cellcolor{HilightColor}\firstresults{76.15 ± 1.61}\\
\midrule
\multirow{9}[8]{*}{GUE} & \multirow{3}[2]{*}{Core Promoter Detection} & tata & 2 & \firstresults{80.95 ± 1.47} & 79.32 ± 2.44 & \cellcolor{HilightColor}80.29 ± 1.63 & 80.90 ± 1.15 & \cellcolor{HilightColor}\firstresults{81.76 ± 1.16}\\
& & notata & 2 & 70.02 ± 0.52 & \firstresults{70.60 ± 0.75} & \cellcolor{HilightColor}70.48 ± 0.35 & \firstresults{71.26 ± 0.56} & \cellcolor{HilightColor}70.43 ± 0.39\\
& & all & 2 & 67.64 ± 0.77 & 68.02 ± 0.63 & \cellcolor{HilightColor}\firstresults{68.50 ± 0.61} & 67.63 ± 0.56 & \cellcolor{HilightColor}\firstresults{68.36 ± 1.07}\\
\cmidrule{2-9}
& \multirow{3}[2]{*}{Promoter Detection} & tata & 2 & 80.30 ± 1.59 & 80.27 ± 2.71 & \cellcolor{HilightColor}\firstresults{82.83 ± 1.69} & \firstresults{83.27 ± 1.95} & \cellcolor{HilightColor}82.55 ± 1.54\\
& & notata & 2 & 92.63 ± 0.36 & \firstresults{93.05 ± 0.32} & \cellcolor{HilightColor}92.75 ± 0.51 & 91.74 ± 0.57 & \cellcolor{HilightColor}\firstresults{92.60 ± 0.49}\\
& & all & 2 & 90.53 ± 0.50 & \firstresults{90.79 ± 0.77} & \cellcolor{HilightColor}90.20 ± 0.65 & 89.34 ± 0.40 & \cellcolor{HilightColor}\firstresults{89.82 ± 0.45}\\
\cmidrule{2-9}
& Covid Variant Classification & Covid & 9 & \firstresults{74.09 ± 0.25} & 70.84 ± 0.80 & \cellcolor{HilightColor}73.40 ± 0.30 & 64.07 ± 0.51 & \cellcolor{HilightColor}\firstresults{72.45 ± 0.21}\\
\cmidrule{2-9}
& \multirow{2}[2]{*}{Species Classification} & Virus & 20 & 67.24 ± 2.10 & 59.17 ± 3.32 & \cellcolor{HilightColor}\firstresults{77.12 ± 1.23} & 71.34 ± 2.05 & \cellcolor{HilightColor}\firstresults{77.21 ± 1.04}\\
& & Fungi & 25 & 15.06 ± 1.32 & 18.75 ± 1.77 & \cellcolor{HilightColor}\firstresults{30.01 ± 0.76} & 15.07 ± 1.88 & \cellcolor{HilightColor}\firstresults{30.14 ± 2.48}\\
\bottomrule
\end{tabular}}
\end{table}

\subsubsection{Image Classification}
\label{hbnn:subsubsec:exp-image-classification}
\mypara{Setup.} Following \cref{pvnn:sec:exp2}, we use a hybrid architecture with a ResNet-18 \citep{he2016deep} backbone and an MLR head. We compare Euclidean MLR with hyperbolic variants in both models. In Poincaré, we evaluate Poincaré MLR (PMLR) re-parameterized by \citet{shimizu2021hyperbolic}, Pseudo-Busemann MLR (PBMLR-P) \citep{nguyen2025neural}, and our BMLR-P. In Lorentz, we evaluate Lorentz MLR (LMLR) \citep{bdeir2024fully} and our BMLR-L. For hyperbolic MLRs, we map the ResNet-18 features to the target hyperbolic space before classification. We evaluate on CIFAR-10 \citep{krizhevsky2009learning}, CIFAR-100 \citep{krizhevsky2009learning}, Tiny-ImageNet \citep{le2015tiny}, and ImageNet-1k \citep{deng2009imagenet}. On the first three data sets, we conduct five-fold experiments.

\mypara{Results.} \cref{hbnn:tab:exp-image-classification} reports top-1 validation accuracy, fit time per epoch, and classifier-head parameters. \cref{hbnn:fig:acc-curves-imagenet} presents the ImageNet-1k accuracy curves. Overall, BMLR-P and BMLR-L consistently outperform prior hyperbolic MLRs with comparable parameters. Within each hyperbolic model, the accuracy margin over prior hyperbolic MLRs increases with the number of classes, from CIFAR-10 to CIFAR-100 and Tiny-ImageNet, with the largest gains on ImageNet-1k. This demonstrates the advantage of BMLR as task complexity increases. Besides, PBMLR-P uses approximately double the head parameters and is markedly slower due to complex batch-inefficient computation, whereas BMLR-L achieves the fastest fit time among all hyperbolic MLRs.

\subsubsection{Genome Sequence Learning}
\label{hbnn:subsubsec:exp-genome-sequence-learning}

\begin{table}[t]
\centering
\caption{Fit time (s/epoch) on genome sequence learning. The fastest times are \firstresults{bold} and the slowest ones are \redbf{red}.}
\label{hbnn:tab:exp-genome-sequence-learning-efficiency}
\resizebox{0.9\linewidth}{!}{%
\begin{tabular}{c|ccc|cc}
\toprule
\multirow{2}[4]{*}{\textbf{Data Set}} & \multicolumn{3}{c|}{$\pball{n}$} & \multicolumn{2}{c}{$\lorentz{n}$}\\
\cmidrule{2-6}
& \textbf{PMLR} & \textbf{PBMLR-P} & \cellcolor{HilightColor}\textbf{BMLR-P} & \textbf{LMLR} & \cellcolor{HilightColor}\textbf{BMLR-L}\\
\midrule
LTR Copia & 3.96 & \redbf{5.11} & \cellcolor{HilightColor}4.06 & 3.92 & \cellcolor{HilightColor}\firstresults{3.77}\\
LINEs & 5.80 & \redbf{6.90} & \cellcolor{HilightColor}5.73 & 5.50 & \cellcolor{HilightColor}\firstresults{5.32}\\
SINEs & 1.36 & \redbf{1.80} & \cellcolor{HilightColor}1.39 & 1.38 & \cellcolor{HilightColor}\firstresults{1.28}\\
\midrule
CMC-EnSpm & 3.11 & \redbf{4.38} & \cellcolor{HilightColor}3.04 & 2.89 & \cellcolor{HilightColor}\firstresults{2.82}\\
hAT-Ac & 4.37 & \redbf{5.37} & \cellcolor{HilightColor}4.38 & 4.13 & \cellcolor{HilightColor}\firstresults{3.96}\\
\midrule
processed & 5.02 & \redbf{5.94} & \cellcolor{HilightColor}4.89 & 4.68 & \cellcolor{HilightColor}\firstresults{4.58}\\
unprocessed & 3.30 & \redbf{3.90} & \cellcolor{HilightColor}3.29 & 3.05 & \cellcolor{HilightColor}\firstresults{2.95}\\
\midrule
CPD-tata & 0.72 & \redbf{1.35} & \cellcolor{HilightColor}0.70 & 0.67 & \cellcolor{HilightColor}\firstresults{0.62}\\
CPD-notata & 6.13 & \redbf{12.40} & \cellcolor{HilightColor}6.00 & 5.73 & \cellcolor{HilightColor}\firstresults{5.71}\\
CPD-all & 6.85 & \redbf{14.35} & \cellcolor{HilightColor}6.59 & 6.35 & \cellcolor{HilightColor}\firstresults{6.33}\\
\midrule
PD-tata & 0.98 & \redbf{1.20} & \cellcolor{HilightColor}0.97 & 1.04 & \cellcolor{HilightColor}\firstresults{0.94}\\
PD-notata & 8.33 & \redbf{10.53} & \cellcolor{HilightColor}8.29 & 8.11 & \cellcolor{HilightColor}\firstresults{7.84}\\
PD-all & 9.40 & \redbf{12.07} & \cellcolor{HilightColor}9.19 & 9.06 & \cellcolor{HilightColor}\firstresults{8.83}\\
\midrule
Covid & 28.96 & \redbf{45.52} & \cellcolor{HilightColor}27.97 & 27.58 & \cellcolor{HilightColor}\firstresults{26.67}\\
\midrule
Virus & 25.28 & \redbf{29.57} & \cellcolor{HilightColor}25.67 & 25.12 & \cellcolor{HilightColor}\firstresults{24.85}\\
Fungi & 6.41 & \redbf{8.96} & \cellcolor{HilightColor}6.41 & \firstresults{6.25} & \cellcolor{HilightColor}\firstresults{6.25}\\
\bottomrule
\end{tabular}}
\end{table}

\mypara{Setup.} Similar to \cref{pvnn:sec:exp3}, we evaluate hyperbolic MLRs on genome sequence learning. Following \citet{khan2025hyperbolic}, we adopt a CNN backbone, which consists of three convolutional blocks and an MLR head. Similar to \cref{hbnn:subsubsec:exp-image-classification}, we compare our BMLR against previous hyperbolic MLR heads by replacing the final Euclidean MLR with a hyperbolic MLR. We validate on two benchmarks: TEB \citep{khan2025hyperbolic} and \emph{Genome Understanding Evaluation (GUE)} \citep{zhou2024dnabert}, covering a total of 16 data sets.

\mypara{Results.} \cref{hbnn:tab:exp-genome-sequence-learning} summarizes five-fold average MCC across TEB and GUE. Compared with other hyperbolic MLRs, our BMLR-P and BMLR-L achieve higher MCC in most tasks. Similar to \cref{hbnn:subsubsec:exp-image-classification}, the gains are more pronounced on complex data sets with more classes, \eg, Virus (20 classes) and Fungi (25 classes), demonstrating the effectiveness of our approach. \cref{hbnn:tab:exp-genome-sequence-learning-efficiency} reports fit time per epoch, where PBMLR-P is consistently the slowest due to batch inefficiency, and BMLR-L is the fastest.

\begin{table}[t]
\centering
\caption{Comparison of hyperbolic FC layers on link prediction. The best results within each hyperbolic model are \firstresults{bold}.}
\label{hbnn:tab:exp-lp-hnn}
\resizebox{0.85\linewidth}{!}{%
\begin{tabular}{c|cc|cccc}
\toprule
\multirow{2}[2]{*}{\textbf{Space}} & \multirow{2}[2]{*}{\textbf{Method}} & \multirow{2}[2]{*}{\textbf{Methodology}} & \textbf{Disease} & \textbf{Airport} & \textbf{PubMed} & \textbf{Cora}\\
& & & $\delta=0$ & $\delta=1$ & $\delta=3.5$ & $\delta=11$\\
\midrule
\multirow{3}[2]{*}{$\pball{n}$} & Möbius & Tangent & 76.35 ± 1.83 & 93.31 ± 0.41 & \firstresults{94.93 ± 0.06} & 90.80 ± 0.56\\
& Poincaré FC & Poincaré geometry & 79.45 ± 1.01 & 94.31 ± 0.16 & 94.24 ± 0.25 & 88.21 ± 0.72\\
& \cellcolor{HilightColor}BFC-P & \cellcolor{HilightColor}Busemann & \cellcolor{HilightColor}\firstresults{80.45 ± 0.93} & \cellcolor{HilightColor}\firstresults{94.88 ± 0.39} & \cellcolor{HilightColor}94.85 ± 0.07 & \cellcolor{HilightColor}\firstresults{91.94 ± 0.32}\\
\midrule
\multirow{3}[2]{*}{$\lorentz{n}$} & LTFC & Tangent & 71.32 ± 5.36 & 92.68 ± 0.35 & 94.85 ± 0.17 & 89.37 ± 0.64\\
& Lorentz FC & Ambient Minkowski & 72.78 ± 2.04 & 92.99 ± 0.33 & 94.20 ± 0.10 & 92.06 ± 0.62\\
& \cellcolor{HilightColor}BFC-L & \cellcolor{HilightColor}Busemann & \cellcolor{HilightColor}\firstresults{78.36 ± 0.51} & \cellcolor{HilightColor}\firstresults{95.37 ± 0.17} & \cellcolor{HilightColor}\firstresults{94.90 ± 0.04} & \cellcolor{HilightColor}\firstresults{92.28 ± 0.12}\\
\bottomrule
\end{tabular}}
\end{table}

\begin{table}[t]
\centering
\caption{Node classification F1 scores of hyperbolic MLRs on the HGCN backbone, where $\delta$ denotes graph hyperbolicity (lower is more hyperbolic). The best results within each hyperbolic model are \firstresults{bold}.}
\label{hbnn:tab:exp-node-classification}
\resizebox{\linewidth}{!}{%
\begin{tabular}{c|c|cccc}
\toprule
\multirow{2}[3]{*}{\textbf{Space}} & \multirow{2}[3]{*}{\textbf{Method}} & \textbf{Disease} & \textbf{Airport} & \textbf{PubMed} & \textbf{Cora}\\
& & $\delta=0$ & $\delta=1$ & $\delta=3.5$ & $\delta=11$\\
\midrule
\multirow{4}[2]{*}{$\pball{n}$} & HGCN & 86.87 ± 2.58 & 85.34 ± 1.16 & 76.29 ± 0.98 & 76.56 ± 0.81\\
& HGCN-PMLR & 88.98 ± 1.96 & 84.78 ± 1.48 & 76.02 ± 1.09 & 77.47 ± 1.15\\
& HGCN-PBMLR-P & 89.05 ± 0.78 & 85.04 ± 0.97 & 75.89 ± 0.78 & 77.90 ± 1.00\\
& \cellcolor{HilightColor}HGCN-BMLR-P & \cellcolor{HilightColor}\firstresults{92.45 ± 0.96} & \cellcolor{HilightColor}\firstresults{86.02 ± 0.53} & \cellcolor{HilightColor}\firstresults{77.36 ± 0.73} & \cellcolor{HilightColor}\firstresults{78.48 ± 1.52}\\
\midrule
\multirow{3}[2]{*}{$\lorentz{n}$} & HGCN & 87.83 ± 0.77 & 84.94 ± 1.40 & 76.49 ± 0.88 & 77.37 ± 1.72\\
& HGCN-LMLR & 89.72 ± 1.51 & 82.61 ± 1.01 & 75.44 ± 1.17 & 69.91 ± 3.61\\
& \cellcolor{HilightColor}HGCN-BMLR-L & \cellcolor{HilightColor}\firstresults{90.80 ± 1.15} & \cellcolor{HilightColor}\firstresults{85.27 ± 1.17} & \cellcolor{HilightColor}\firstresults{77.30 ± 0.41} & \cellcolor{HilightColor}\firstresults{77.65 ± 2.10}\\
\bottomrule
\end{tabular}}
\end{table}

\subsubsection{Node Classification}
\label{hbnn:subsubsec:exp-node-classification}
\mypara{Setup.} Following \citet{nguyen2025neural}, we adopt the HGCN \citep{chami2019hyperbolic} backbone to evaluate our BMLR on graph data sets, including Disease \citep{anderson1991infectious}, Airport \citep{zhang2018link}, PubMed \citep{namata2012query}, and Cora \citep{sen2008collective}. The HGCN backbone consists of a hyperbolic \emph{Graph Convolutional Network (GCN)} and an MLR as the final classification layer. Both the GCN and the MLR are built on the hyperbolic space. The vanilla HGCN uses a tangent MLR, which maps features into the tangent space via $\rielog_e$ and applies a Euclidean MLR. We replace this with different hyperbolic MLRs.

\mypara{Results.} \cref{hbnn:tab:exp-node-classification} reports average F1 scores. Our BMLRs consistently outperform prior hyperbolic MLRs within each hyperbolic model. As graphs become less hyperbolic, that is, for larger $\delta$, existing hyperbolic heads could underperform the vanilla tangent-based MLR, for example, PBMLR-P on PubMed, and LMLR on Airport, PubMed, and Cora. Especially on Cora, which has the largest $\delta$, LMLR lags the tangent baseline by a large margin (69.91 vs. 77.37). In contrast, BMLR remains the top performer across all $\delta$ values, indicating that Busemann-based decoding robustly strengthens HGCN over a broader range of graph hyperbolicity.

\subsubsection{Link Prediction}
\label{hbnn:subsubsec:exp-link-prediction}
\mypara{Setup.} We compare our BFC layers with prior hyperbolic FC layers, including the Möbius layer \citep{ganea2018hyperbolic} that operates via the tangent space, the Lorentz FC layer \citep{chen2022fully} that operates through the ambient Minkowski space, and the Poincaré FC layer \citep{shimizu2021hyperbolic}. Mimicking the Möbius layer, we also implement a Lorentz tangent FC layer, $\rieexp_{\Lzero}\left(M\rielog_{\Lzero}(x)\right)$, referred to as LTFC. Following \citet{chami2019hyperbolic}, we evaluate on Disease, Airport, PubMed, and Cora. Following the HNN implementation \citep{ganea2018hyperbolic,chami2019hyperbolic}, all methods share the same backbone with two FC layers. For a fair comparison, all hyperbolic FC layers are followed by a gyroaddition biasing. For BFC, we use $\phi=\tanh$ on Airport and Cora and the identity map on the other two data sets.

\mypara{Results.} \cref{hbnn:tab:exp-lp-hnn} reports five-fold test AUC. Our BFC layers generally outperform prior hyperbolic FC layers. The gains are most pronounced on Disease, which is the most hyperbolic ($\delta=0$), where Busemann-based decoding is markedly more effective than tangent or ambient methods, indicating better capture of intrinsic hyperbolic geometry. This observation aligns with geometric intuition, since tangent space or ambient space approximations inherently struggle to represent curved manifolds in highly non-Euclidean cases.

\begin{table}[t]
  \centering
  \caption{Efficiency comparison: fit time (s/epoch) and parameter count. Slowest results and largest parameter counts are in \redbf{red}.}
   \label{hbnn:tab:exp-efficiency}%
    \resizebox{\linewidth}{!}{%
    \begin{tabular}{c|c|cc|cc|cc|cc}
    \toprule
    \multirow{2}[2]{*}{\textbf{Space}} & \multirow{2}[2]{*}{\textbf{Method}} & \multicolumn{2}{c|}{\textbf{Disease}} & \multicolumn{2}{c|}{\textbf{Airport}} & \multicolumn{2}{c|}{\textbf{PubMed}} & \multicolumn{2}{c}{\textbf{Cora}} \\
    \cmidrule{3-10}
     &  & \textbf{Fit Time} &  \textbf{\#Params} & \textbf{Fit Time} &  \textbf{\#Params} & \textbf{Fit Time} &  \textbf{\#Params} & \textbf{Fit Time} &  \textbf{\#Params} \\
     \midrule
    \multirow{3}[2]{*}{$\pball{n}$} & M\"obius & 0.0200  & 464   & 0.0535  & 480   & 0.1120  & 8288  & 0.0229  & 23216  \\
     & Poincaré FC & 0.0198  & 528 & 0.0536  & 544   & 0.1176  & 8352  & 0.0248  & 23280  \\
     & \cellcolor{HilightColor}BFC-P & \cellcolor{HilightColor}0.0201  & \cellcolor{HilightColor}528 & \cellcolor{HilightColor}0.0512  & \cellcolor{HilightColor}544   & \cellcolor{HilightColor}0.1123  & \cellcolor{HilightColor}8352  & \cellcolor{HilightColor}0.0231  & \cellcolor{HilightColor}23280  \\
    \midrule
    \multirow{3}[2]{*}{$\lorentz{n}$} & LTFC & \redbf{{0.0343}} & 464   & \redbf{{0.0818}} & 480   & \redbf{{0.1633}} & 8288  & \redbf{{0.0370}} & 23216  \\
     & Lorentz FC & 0.0232  & \redbf{{563}} & 0.0715  & \redbf{{580}} & 0.1537  & \redbf{{8876}} & 0.0261  & \redbf{{24737}} \\
     & \cellcolor{HilightColor}BFC-L & \cellcolor{HilightColor}0.0244  & \cellcolor{HilightColor}528   & \cellcolor{HilightColor}0.0713  & \cellcolor{HilightColor}544   & \cellcolor{HilightColor}0.1525  & \cellcolor{HilightColor}8352  & \cellcolor{HilightColor}0.0280  & \cellcolor{HilightColor}23280  \\
    \bottomrule
    \end{tabular}%
    }
\end{table}%

\mypara{Training Time and Parameter Count.} \cref{hbnn:tab:exp-efficiency} summarizes fit time per epoch and parameter counts. Our BFC layers achieve training time and model size comparable to existing layers. In particular, LTFC is the slowest due to costly logarithmic and exponential maps, and LFC uses the largest number of parameters among Lorentz variants.

\section{Full-Rank Correlation Networks}
\label{sec:ch5-cornet}

\subsection{Introduction}
\label{cornet:subsec:introduction}
The preceding two sections studied manifold-specific designs for hyperbolic learning. We now turn to neural networks on full-rank correlation manifolds. Covariance matrices in the SPD manifold have achieved success in various applications, with many deep network architectures adapted to leverage their Riemannian geometries \citep{huang2017riemannian,brooks2019riemannian,chakraborty2020manifoldnet,cruceru2021computationally,pan2022matt,kobler2022spd,wang2023towards,chen2023riemannian,katsman2024riemannian,li2025spdim,pouliquen2025schur,wang2025gbwmbn,kang2025smlnet,hu2026riemannian}. In contrast, correlation matrices, despite serving as statistically compact alternatives to covariance matrices \citep{archakov2024canonical}, remain unexplored in deep learning.

As discussed in \cref{sec:ch2-full-rank-correlation-manifolds}, Riemannian structures for correlation matrices have only recently been developed. \citet{david2019riemannian} identified full-rank correlation matrices as a quotient manifold of the SPD manifold, referred to as the correlation manifold. However, this quotient geometry does not guarantee uniqueness or closed forms of the Riemannian logarithm and Fréchet mean \citep[Sec.~1.1]{thanwerdas2022theoretically}. To close this gap, \citet{thanwerdas2022theoretically} proposed three theoretically and computationally convenient geometries: ECM, LECM, and PHCM. \citet{thanwerdas2024permutation} further introduced two efficient permutation-invariant metrics: OLM and LSM. These Riemannian structures provide promising foundations for extending Euclidean deep learning to the correlation manifold.

On the other hand, several fundamental layers in Euclidean deep learning, such as MLR, FC, and convolutional layers, have been extended to different manifolds by leveraging their rich Riemannian or algebraic structures \citep{huang2017riemannian,huang2017deep,huang2018building,ganea2018hyperbolic,chakraborty2020manifoldnet,chen2022fully,shimizu2021hyperbolic,bdeir2024fully,chen2024rmlr,nguyen2024matrix}. For the SPD manifold, these layers have been constructed using bilinear mapping \citep{huang2017riemannian}, weighted Fréchet means \citep{chakraborty2020manifoldnet}, gyrovector spaces \citep{nguyen2023building,nguyen2024matrix}, and Riemannian geometry \citep{chen2024spdrmlr,chen2024rmlr}.

Inspired by these advancements, we develop MLR, FC, and convolutional layers for correlation manifolds in a geometrically intrinsic manner. We begin by systematically introducing four types of correlation-based MLR, FC, and convolutional layers, corresponding to ECM, LECM, OLM, and LSM, respectively. Besides, we discuss backpropagation through Riemannian computations over the correlation manifold, with novel approaches for accurate backpropagation under OLM and LSM. As the above four metrics have zero curvature, our next focus is to build correlation layers under the geometry of non-zero curvature. We target PHCM, induced by the product of multiple hyperbolic spaces \citep[Thm.~4.4]{thanwerdas2022theoretically}. By adapting existing Poincar\'e-based hyperbolic MLR, FC, and convolutional layers designed for a single Poincar\'e ball \citep{ganea2018hyperbolic,shimizu2021hyperbolic}, we construct their counterparts on the correlation manifold. Together with the corresponding backpropagation mechanisms, these layers constitute complete \emph{Correlation Networks (CorNets)} under different geometries. The effectiveness is validated by experiments comparing our approach against existing SPD and Grassmannian baselines.

\cref{cornet:tab:euc-cor-layers-comparison} summarizes the correspondence between Euclidean and our correlation layers. In summary, our \textbf{main contributions} are as follows:
\begin{enumerate}
    \item 
    We systematically extend MLR, FC, and convolutional layers to the correlation manifold under five geometries: four with zero curvature and one with non-zero curvature. The developed layers enable flexible variation of the latent geometry under a consistent network architecture, allowing for straightforward comparisons across different correlation geometries.
    \item 
    We develop accurate backpropagation of Riemannian computations under OLM and LSM.
    \item
    We conduct experiments against existing SPD and Grassmannian networks to demonstrate the effectiveness of correlation embeddings and networks. 
\end{enumerate}

\begin{table}[t]
\centering
\caption{Correspondence between Euclidean and correlation-based layers. For convolution, kernel-based FC refers to applying a convolution kernel to a receptive field, which is an FC transformation.}
\label{cornet:tab:euc-cor-layers-comparison}
\resizebox{\linewidth}{!}{
\begin{tabular}{c|c|c}
\toprule
\textbf{Space} & \textbf{Euclidean $\bbR{n}$} & \textbf{Correlation $\cor{n}$} \\
\midrule
$C$-class MLR &
$f: \bbR{n} \ni x \mapsto p = \softmax(Ax + b) \in \bbR{C}$ &
$f: \cor{n} \ni X \mapsto p \in \bbR{C}$ \\

FC layer &
$\calF: \bbR{n} \ni x \mapsto y = Ax + b \in \bbR{m}$ &
$\calF: \cor{n} \ni X \mapsto Y \in \cor{m}$ \\

Convolution & Kernel-based FC in each receptive field & Kernel-based correlation FC in each receptive field \\

Geometry &
Euclidean &
ECM, LECM, OLM, LSM and PHCM \\
\bottomrule
\end{tabular}
}
\end{table}

\mypara{Outline.} \cref{cornet:subsec:log-euclidean-layers} constructs correlation MLR, FC, and convolutional layers under four flat geometries. \cref{cornet:subsec:phcm-layers} develops their counterparts under a non-zero-curvature geometry and establishes the order-invariance of the associated $\beta$-operations. \cref{cornet:subsec:backpropagation} presents backpropagation over the correlation geometries, and \cref{cornet:subsec:experiments} evaluates CorNets under these five geometries. Proofs are deferred to \cref{app:cornet-proofs}.

\subsection{Log-Euclidean Correlation Layers}
\label{cornet:subsec:log-euclidean-layers}
Since ECM, LECM, OLM, and LSM are derived via diffeomorphisms from Euclidean spaces, they are collectively termed Log-Euclidean metrics \citep{thanwerdas2024permutation}.
This motivates the principled development of MLR, FC, and convolutional layers \citep{thanwerdas2024permutation}.

\subsubsection{Log-Euclidean Correlation MLRs}
\label{cornet:subsubsec:log-euclidean-mlrs}
As discussed in \cref{rmlr:subsec:re_exist_MLR}, the MLR can be rewritten by point-to-hyperplane formulations. We use the same margin-distance infimum, while the Euclidean isometries of ECM, LECM, OLM, and LSM make it possible to solve this infimum exactly under all four geometries. To avoid over-parameterization, we follow \cref{pvnn:subsec:pv-mlr} and set $P_k=\rieexp_E\left(\gamma_k[Z_k]\right)$ and $A_k=\pt{E}{P_k}(Z_k)$, with $[Z_k]=\frac{Z_k}{\norm{Z_k}_E}$ as the unit direction vector of $Z_k$. Here, $E$ is the origin of $\calM$, while $\gamma_k\in\bbRscalar$ and $Z_k\in T_E\calM\cong\bbR{m}$ are the MLR parameters. This is a concrete instance of the trivialization strategy reviewed in \cref{sec:ch2-riemannian-optimization}. Under this trivialization, each hyperplane $H_{A_k,P_k}$ is denoted as $H_{Z_k,\gamma_k}$.

As all Log-Euclidean metrics are isometric to Euclidean spaces, the corresponding MLRs admit principled closed forms.

\begin{paristheorem}
    \label{cornet:thm:flat-mlr}
    \linktoproof{cornet:thm:flat-mlr}
    Let $\left(\calM,g^{\calM} \right)$ be an $m$-dimensional manifold that is isometric to the standard Euclidean space $\bbR{m}$ via the diffeomorphism $\phi: \calM \rightarrow \bbR{m}$. Denoting $E = \phi^{-1}(\zerovec)$ with $\zerovec$ as the zero vector, each $v_{k}(X)$ and margin hyperplane $H_{Z_k,\gamma_k}$ in the $C$-class Riemannian MLR are $v_{k}(X) = \left \langle \phi(X), \phi_{*,E}(Z_k) \right \rangle - \gamma_k \norm{\phi_{*,E}(Z_k)}$ and $H_{Z_k,\gamma_k} = \{X \in \calM \mid v_{k}(X) =0 \}$, respectively. Here, $Z_{k} \in T_E \calM \cong \bbR{m}$ and $\gamma_{k} \in \bbRscalar$ for $1 \leq k \leq C$ are MLR parameters, while $\phi_{*}$ is the differential.
\end{paristheorem}

Simple computations show that
\begin{equation}
    \begin{aligned}
    \text{ECM: } \phi^{\mathrm{EC}}(I) = \bbzero, & \quad \text{LECM: } \log \circ \Theta(I) = \bbzero, \\
    \text{OLM: } \mathrm{Log}^\circ(I) = \bbzero, & \quad \text{LSM: } \mathrm{Log}^\star(I) = \bbzero. \\
    \end{aligned}
\end{equation}
Therefore, we define the origin of the correlation manifold under four Log-Euclidean metrics as the identity matrix. Besides, \cref{cornet:thm:flat-mlr} suggests that Log-Euclidean MLRs can be obtained modulo the calculation of diffeomorphisms and their differentials at the identity matrix $I$.

\begin{parisproposition}[Differentials]
    \label{cornet:prop:diff-at-I}
    \linktoproof{cornet:prop:diff-at-I}
    For any tangent vector $V \in T_I \cor{n} \cong \hol{n}$, the differentials of $\isoecm$, $\log \circ \Theta$, $\offlog$, and $\logscaled$ at the identity matrix $I$ are 
\begin{equation}
    \begin{aligned}
    \phi^{\EC}_{*,I}(V) = \lfloor V\rfloor, & \quad ( \log \circ \Theta )_{*,I}(V) = \lfloor V\rfloor, \\
    \offlog_{*,I}(V) = V, & \quad \logscaled_{*,I}(V) = V- \diag (V \vecone),
    \end{aligned}
\end{equation}
    where $\diag : \bbR{n} \rightarrow \bbDspace{n}$ returns a diagonal matrix, and $\vecone = (1, \cdots, 1)^\top \in \bbR{n}$.
\end{parisproposition}

Putting \cref{cornet:prop:diff-at-I} into \cref{cornet:thm:flat-mlr}, we obtain correlation MLRs under four Log-Euclidean metrics.

\begin{paristheorem}[Log-Euclidean MLRs]
    \label{cornet:thm:cormlr}
    Given $C \in \cor{n}$, the logits $v_{k}(C)$ for the $k$-th class in the correlation MLRs under four Log-Euclidean metrics are
    \begin{equation}
    \label{cornet:eq:cormlr-v-k}
    \begin{aligned}
    & v^{\EC}_k (C) =\inner{\lfloor\Theta(C)\rfloor}{\lfloor Z_k\rfloor} - \gamma_{k} \norm{\lfloor Z_k\rfloor}, \\
    & v^{\LEC}_k (C) =\inner{\log \circ \Theta(C)}{\lfloor Z_k\rfloor} - \gamma_{k} \norm{\lfloor Z_k\rfloor}, \\
    & v^{\OL}_k (C) =\inner{\offlog(C)}{Z_k} - \gamma_{k} \norm{Z_k}, \\
    & v^{\LS}_k (C) =\inner{\logscaled(C)}{\logscaled_{*,I}(Z_k)} - \gamma_{k} \norm{\logscaled_{*,I}(Z_k)},
    \end{aligned}
    \end{equation}
    where $Z_k \in \hol{n}$ and $\gamma_{k} \in \bbRscalar$ are parameters.
\end{paristheorem}

\subsubsection{Log-Euclidean FC and Convolutional Layers}
\label{cornet:subsubsec:lem-cor-fc}
Following \cref{pvnn:subsec:pv-fc-hnnpp}, we now generalize this point-to-hyperplane FC construction to the correlation manifold.

\begin{parisdefinition}[Correlation FC layers] \label{cornet:def:cor-fc}
    Given a metric $g$, the correlation FC layer $\calF: \cor{n} \ni X \mapsto Y \in \cor{m}$ returns the output $Y$ by solving the following $d=\nicefrac{m(m-1)}{2}$ equations: 
    \begin{equation} \label{cornet:eq:riem-fc}
        s_k \dist (Y, H _{O_k, I} ) = v _{k}(X; Z_k,\gamma_k), \quad 1 \leq k \leq d,
    \end{equation}
    where $s_k=\sign \left( \inner{\rielog _{I}(Y)}{O_k}_I \right)$, $I$ is the identity matrix, $d$ is the dimension of $\cor{m}$, $\{ O_k \}_{k=1}^d$ is an orthonormal basis over $T_{I}\cor{m}$, $\dist(\cdot,\cdot)$ is the margin distance to the hyperplane $H _{O_k, I}$, and $v _{k}$ is defined by \cref{rmlr:subsec:re_exist_MLR} for $\cor{n}$. The FC parameters are $\{Z_k \in \hol{n}\}_{k=1}^d$ and $\{\gamma_k \in \bbRscalar \}_{k=1}^d$.
\end{parisdefinition}
\cref{cornet:app:subsec:connection-fc-layers} details how \cref{cornet:def:cor-fc} extends the existing SPD, Poincaré, and Euclidean FC layers. Although \cref{cornet:def:cor-fc} is implicitly defined by $d$ equations, the FC layers under four Log-Euclidean geometries admit explicit expressions in a principled manner. Analogous to \cref{cornet:thm:flat-mlr}, a corresponding result for the FC layer is presented in \cref{cornet:app:lem:flat-fc}, which yields the Log-Euclidean FC layers.

\begin{paristheorem}[Log-Euclidean FC layers]
    \label{cornet:thm:flat-cor-fc}
    \linktoproof{cornet:thm:flat-cor-fc}
    Given an input correlation $C \in \cor{n}$, the correlation FC layers $\calF(\cdot): \cor{n} \rightarrow \cor{m}$ under different Log-Euclidean metrics are
    \begin{align}
        \label{cornet:eq:ecm-fc}
        \text{ECM: } &
        Y = \coropt \circ \chol^{-1} \left( V^\EC + I_{m} \right), \\
        \label{cornet:eq:lecm-fc}
        \text{LECM: } &
        Y = \coropt \circ \chol^{-1} \circ \exp \left( V^\LEC \right), \\
        \label{cornet:eq:olm-fc}
        \text{OLM: } &
        Y = \offexp \left( V^\OL \right), \\
        \label{cornet:eq:lsm-fc}
        \text{LSM: } &
        Y = \coropt \circ \exp \left( V^\LS \right),
    \end{align}
    where the $(i, j)$-th elements in $V^\EC \in \LTzero{m}$, $V^\LEC \in \LTzero{m}$, $V^\OL \in \hol{m}$, and $V^\LS \in \rzero{m}$ are
    \begin{align}
        & V^{\EC}_{ij}=
        \begin{cases}
        v^{\EC}_{ij}(C), & \text { if } i>j \\
        0, & \text{ otherwise }
        \end{cases} \\
        & V^{\LEC}_{ij}=
        \begin{cases}
        v^{\LEC}_{ij}(C), & \text { if } i>j \\
        0, & \text{ otherwise }
        \end{cases} \\
        & V^{\OL}_{ij}=
        \begin{cases}
        \frac{v^{\OL}_{ij}(C)}{\sqrt{2}}, & \text { if } i>j \\
        V^{\OL}_{ji}, & \text { if } i<j \\
        0, & \text{ otherwise }
        \end{cases} \\
        & V^{\LS}_{ij}=
        \begin{cases}
        \nicefrac{v^{\LS}_{ij}(C)}{\sqrt{6}}, & \text { if } m > i > j \geq 1 \\
        \nicefrac{v^{\LS}_{ii}(C)}{\sqrt{3}}, & \text { if } m > i \geq 1 \\
        V^{\LS}_{ji}, & \text { if } i<j \\
        - \sum_{k=1}^{m-1} V^{\LS}_{kj}, & \text { if } i=m, 1 \leq j < m \\
        \sum_{k=1}^{m-1} \sum_{l=1}^{m-1} V^{\LS}_{l k}, & \text { if } i=j=m
        \end{cases}
    \end{align}
    Each $v^{g}_{ij}$ with $g \in \{\EC,\LEC,\OL,\LS\}$ is defined by \cref{cornet:eq:cormlr-v-k} with parameters $Z_{ij} \in \hol{n}$ and $\gamma_{ij} \in \bbRscalar$. For $v^{\EC}_{ij}$, $v^{\LEC}_{ij}$, and $v^{\OL}_{ij}$, the indices satisfy $i,j=1,\ldots,m$ and $i>j$. For $v^{\LS}_{ij}$, they satisfy $i,j=1,\ldots,m-1$ and $i\geq j$.
\end{paristheorem}

\begin{figure}[t]
\centering
\includegraphics[width=\linewidth,trim={0cm 0cm 0cm 0cm}]{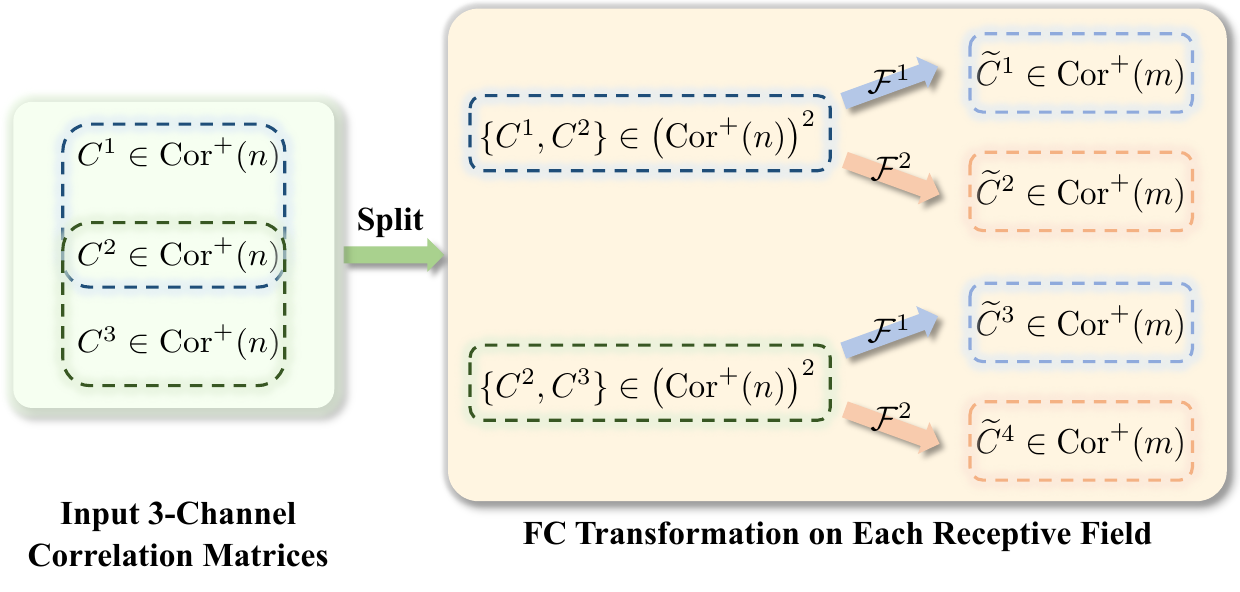}
\caption{Illustration of the Log-Euclidean 1D convolution with two kernels. The $3$-channel input is first split into two receptive fields along the channel dimension. In each receptive field, two kernels are applied to the product space.
}
\label{cornet:fig:conv-layers}
\end{figure}

\mypara{Correlation Convolution.} Following the convolution-as-FC construction in \cref{pvnn:subsec:pv-conv-activation}, we develop the correlation convolution. The ${c}$-channel correlation matrices $\{C_i \in \cor{n} \}_{i=1}^c$ within a receptive field are first concatenated into $\boldsymbol{C} \in (\cor{n})^{c}$. For each convolution kernel, $\boldsymbol{C}$ is then fed into a correlation FC layer.\footnote{\cref{cornet:thm:flat-cor-fc} naturally supports product geometries, which are detailed in \cref{cornet:app:subsec:lem-fc-product}.} \cref{cornet:fig:conv-layers} illustrates the above process.

\subsection{Poly-Hyperbolic-Cholesky Layers}
\label{cornet:subsec:phcm-layers}

As detailed in \cref{sec:ch2-full-rank-correlation-manifolds}, the space $\chocor{n}$, consisting of the Cholesky factors of $\cor{n}$, can be identified with the product of $n-1$ hyperbolic open hemispheres, $\bbPHS{n-1} = \prod_{i=1}^{n-1} \hs{i}$. We focus on the widely used hyperbolic Poincaré ball, whose MLR, FC, and $\beta$-concatenation components are reviewed in \cref{app:backbone-hyperbolic-spaces}. In the following, we focus on the canonical Poincaré ball ($K=-1$), namely the unit Poincaré ball $\unitpball{n}$. We first identify the correlation manifold with the poly-Poincaré space $\bbPPB{n-1} = \prod _{i=1}^{n-1} \unitpball{i}$, the product of $n-1$ unit Poincaré balls. Then, we develop correlation layers from the layers on a single Poincaré space.

\subsubsection{Correlation Geometry via Poincaré Balls}

\begin{parisproposition}[Isometries]
    \label{cornet:prop:isometry-hs-pball}
    \linktoproof{cornet:prop:isometry-hs-pball}
    The open hemisphere $\hs{n}$ is isometric to the unit Poincaré ball $\unitpball{n}$ by
    \begin{equation}
    \begin{aligned}
        \psi _{\hs{n} \rightarrow \unitpball{n}} ((x^\top, x_{n+1})^\top) &= \frac{x}{1+x_{n+1}}, \\
        \psi _{\unitpball{n} \rightarrow \hs{n}} (y) &= \frac{1}{1+\norm{y}^2}
        \left(
        \begin{array}{c}
            2y \\
            1-\norm{y}^2
        \end{array} \right),
    \end{aligned}
    \end{equation}
    with $(x^\top,x_{n+1})^\top \in \hs{n} \subset \bbR{n} \times \bbR{+}$ and $y \in \unitpball{n} \subset \bbR{n}$.
\end{parisproposition}

\cref{cornet:prop:isometry-hs-pball} indicates that $\cor{n}$ can be identified with $\bbPPB{n-1} = \prod_{i=1}^{n-1} \unitpball{i}$ via the diffeomorphism $\Phi$:
\begin{equation} \label{cornet:eq:cor-to-ppb}
    C \overset{\chol}{\longmapsto} 
    \begin{pmatrix}
    1 & 0 & \cdots & 0 \\
    L_{21} & L_{22} & \cdots & 0 \\
    \vdots & \vdots & \ddots & \vdots \\
    L_{n1} & L_{n2} & \cdots & L_{nn}
    \end{pmatrix}
    \overset{\prod_{i=1}^{n-1}\psi_i}{\longmapsto}
    \begin{array}{c}
    \psi _{1} (h_1) \\
    \vdots \\
    \psi _{n-1} (h_{n-1}) \\
    \end{array}
\end{equation}
with $C \in \cor{n}$, $h_i = (L_{i+1,1}, \cdots, L_{i+1,i+1})^\top \in \hs{i}$, and $\psi _{i} = \psi _{\hs{i} \rightarrow \unitpball{i}}$. This identification motivates us to construct the correlation layers using the corresponding layers over Poincaré spaces.

\subsubsection{Revisiting Poincaré Layers} \label{cornet:subsubsec:revisit-poincare-layers}
The Poincaré MLR and FC layers are reviewed in \cref{app:backbone-hyperbolic-spaces} and follow the point-to-hyperplane logic discussed in \cref{chapter:rmlr,sec:ch5-pvnn,sec:ch5-hbnn}. The convolutional construction uses the Poincaré $\beta$-concatenation defined below.

The Poincaré convolutional layer shares a logic similar to the correlation convolution, except it uses $\beta$-concatenation to concatenate the Poincaré vectors in each receptive field \citep[Secs.~3.3--3.4]{shimizu2021hyperbolic}, which can stabilize the norm of the Poincaré vector. The Poincaré $\beta$-concatenation generalizes the Euclidean concatenation via the scaled concatenation in the tangent space. Given inputs $\{x_i \in \unitpball{n_i}\}_{i=1}^N$, it is defined as $\rieexp_{\zerovec} \left( \beta_n \left( \beta_{n_1}^{-1} v_1^\top, \cdots, \beta_{n_N}^{-1} v_{N}^\top \right) \right)^\top \in \unitpball{n}$, where $v_i = \rielog_{\zerovec}(x_i)$ and $n=\sum_{i=1}^N n_i$. Here, $\beta_{n_i}$ and $\beta_n$ are defined by the beta function $\beta_\alpha = \mathrm{B}\left(\nicefrac{\alpha}{2}, \nicefrac{1}{2}\right)$. The inverse is called the Poincaré $\beta$-split. The Poincaré convolution is: (1) $\beta$-concatenating the multi-channel feature in a given receptive field; and (2) performing the Poincaré FC transformation.

\begin{figure}[t]
\centering
\includegraphics[width=\linewidth,trim={0cm 0cm 0cm 0cm}]{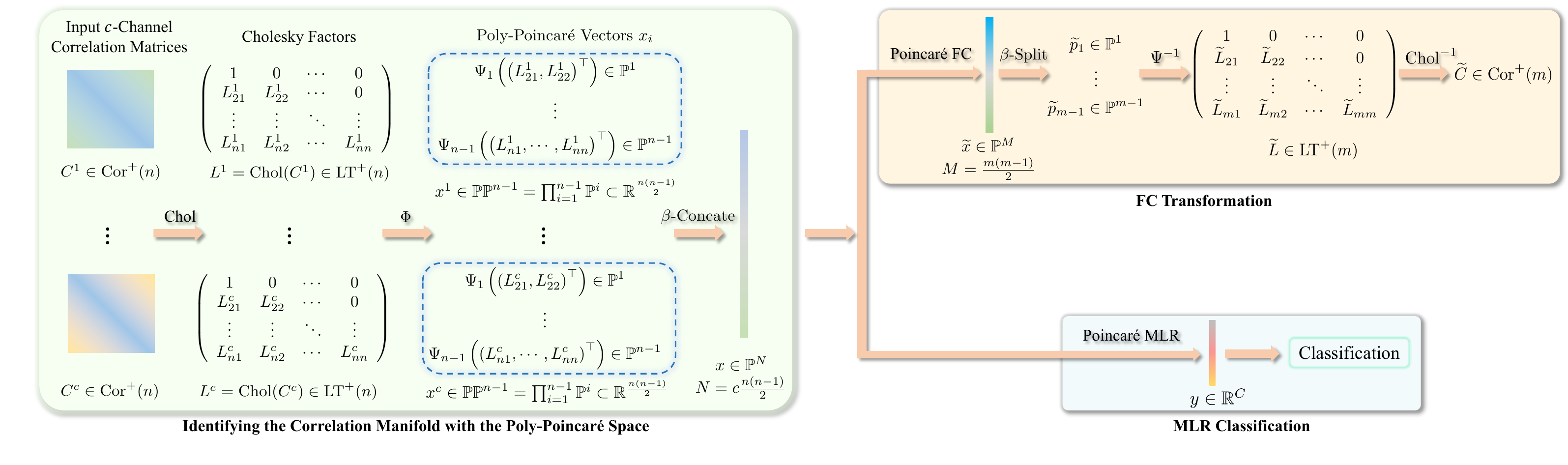}
\caption{Illustration of the PHCM convolution and MLR. The multi-channel input correlation matrices are denoted as $\{C^i\}_{i=1}^c$. 
For the convolutional layer, the illustration focuses on the transformation within a receptive field and assumes a single-channel output.
}
\label{cornet:fig:phcm-layers}
\end{figure}

\subsubsection{Building Poly-Hyperbolic-Cholesky Layers}

\mypara{PHCM MLR.}  The input multi-channel correlation matrices, $\boldsymbol{C} = \{ C^i \in \cor{n} \}_{i=1}^c$, are first mapped into poly-Poincaré spaces as $\boldsymbol{x} = \{x^i = \Phi(C^i) \in \bbPPB{n-1} \}_{i=1}^c$. The resulting Poincaré vectors are then $\beta$-concatenated into a single Poincaré vector $x \in \unitpball{N}$, where $N=c\frac{n(n-1)}{2}$. This concatenated vector is subsequently fed into the Poincaré MLR for classification.

\mypara{PHCM Convolutional and FC Layer.} The convolutional layer follows a logic similar to Log-Euclidean convolution. The multi-channel correlation matrices within a receptive field $\boldsymbol{C} = \{ C^i \in \cor{n} \}_{i=1}^c$ are first mapped to a $\beta$-concatenated Poincaré vector $x \in \unitpball{N}$ as in the PHCM MLR, which is then fed into the Poincaré FC layer for dimensionality transformation. This produces a vector $\widetilde{x} \in \unitpball{M}$, with $M=k\frac{m(m-1)}{2}$, which is then split using $\beta$-split. Subsequently, applying $\Phi^{-1}$ reconstructs new $k \times m \times m$ correlation matrices. When the input is a single correlation matrix, it is reduced to the correlation FC.

\cref{cornet:fig:phcm-layers} illustrates the PHCM layers. However, there is an underlying ambiguity in the above discussion. To clarify, we write each $x^i \in \bbPPB{n-1}$ in $\boldsymbol{x}$ as $x^i = \{p^i_1 \in \unitpball{1}, \cdots, p^i_{n-1} \in \unitpball{n-1}\}$, which gives $\boldsymbol{x}=\{ p^i_j \in \unitpball{j} \}_{i=1,j=1}^{i=c,j=n-1}$. We can either concatenate twice by $i \rightarrow j$ or once along both $i$ and $j$. A similar issue arises with $\beta$-split. The following theorem establishes this invariance.

\begin{paristheorem}[Order-invariance]
    \label{cornet:thm:invariance-beta-concat}
    \linktoproof{cornet:thm:invariance-beta-concat}
    Given multichannel data ${x_{i_1, \dots, i_n} \in \unitpball{n_{i_n}}}$ with $i_j \in \{1, \dots, N_j\}$, applying the $\beta$-concatenation sequentially $n$ times in the order $i_n \rightarrow \dots \rightarrow i_1$ is equivalent to a single $\beta$-concatenation along all indices simultaneously. Similarly, $\beta$-splitting $x \in \unitpball{N}$ into multichannel data ${x_{i_1, \dots, i_n} \in \unitpball{n_{i_n}}}$ with $i_j \in \{1, \dots, N_j\}$ and $N=\left(\prod_{j=1}^{n-1}N_j\right)\sum_{i_n=1}^{N_n}n_{i_n}$ under the sequential order $i_1 \rightarrow \dots \rightarrow i_n$ is identical to the one under a single $\beta$-split to generate all indices simultaneously.
\end{paristheorem}

Therefore, we always conduct the $\beta$-operation simultaneously along both $i$ and $j$.

\subsection{Backpropagation over Correlation Geometries}
\label{cornet:subsec:backpropagation}

Except for $\dplus$ and $\dstar$, all computations involved in the five metrics can be backpropagated using existing techniques or PyTorch's auto-differentiation. The matrix logarithm, matrix exponentiation, and Cholesky decomposition, together with their differentials and backpropagation, are reviewed in \cref{sec:ch2-matrix-functions}. It therefore remains to discuss $\dplus$ and $\dstar$.

\mypara{$\dplus$ and $\dstar$.} Their gradients can be backpropagated either approximately through their iterative algorithms or accurately using the following two propositions.

\begin{parisproposition}[Gradients w.r.t. $\dplus$]
    \label{cornet:prop:gradients-d-plus}
    \linktoproof{cornet:prop:gradients-d-plus}
    Let $l(\cdot)$ be the loss function and define $F:\hol{n}\rightarrow\sym{n}$ by $F(H)=Y=\dplus(H)+H$ for any symmetric hollow matrix $H$, where $\sym{n}$ is the Euclidean space of $n \times n$ symmetric matrices. Let $Y = U \Delta U^\top$ be the eigendecomposition with $(\delta_1, \cdots, \delta_n)$ as eigenvalues. Given the succeeding gradient $\frac{\partial l}{\partial Y}$, the output gradient $\frac{\partial l}{\partial H}$ is
    \begin{equation}
        \frac{\partial l}{\partial H}
        = \off \left(\frac{\partial l}{\partial Y} - \exp_{*,Y} \left( \bbD \left( (H^0)^{-1} \diagvec \left( \frac{\partial l}{\partial Y} \right) \vecone^\top \right) \right) \right),
    \end{equation}
    with $H^0\in\spd{n}$ having entries $H_{i l}^0=\sum_{j, k} U_{i j} U_{i k} U_{l j} U_{l k} [L_{\exp}]_{j,k}$, where $L_{\exp}$ is the Loewner matrix in \cref{eq:ch2-loewner-matrix} specialized to $f=\exp$ and $\sigma_i=\delta_i$. Here, $\bbD(\cdot): \bbR{n \times n} \rightarrow \bbDspace{n}$ extracts the diagonal matrix, while $\diagvec(\cdot): \bbR{n \times n} \rightarrow \bbR{n}$ returns a vector of diagonal elements. Besides, $\off(\cdot)$ subtracts the diagonal matrix from a matrix, and $\exp_{*,Y}$ is the differential of the symmetric matrix exponential given by \cref{eq:ch2-symmetric-matrix-function-differential}.
\end{parisproposition}

\begin{parisproposition}[Gradients w.r.t. $\dstar$]
    \label{cornet:prop:gradients-d-star}
    \linktoproof{cornet:prop:gradients-d-star}
    Following the notation in \cref{cornet:prop:gradients-d-plus}, define $F:\cor{n}\rightarrow\rone{n}$ by $F(C)=\Sigma=\dstar(C) C \dstar(C)$, where $\rone{n}$ is the manifold of $n \times n$ SPD matrices with unit row sum. Given the succeeding gradient $\frac{\partial l}{\partial \Sigma}$, the output gradient $\frac{\partial l}{\partial C}$ is
    \begin{equation}
        \frac{\partial l}{\partial C}
        = \Delta \left( \frac{\partial l}{\partial \Sigma}  -\symmetrize{ (I+\Sigma)^{-1} \widetilde{v} \vecone^{\top}} \right) \Delta,
    \end{equation}
    where $\Delta=\bbD(\Sigma)^{\nicefrac{1}{2}}$, $\widetilde{v}=\diagvec \left(\Sigma \frac{\partial l}{\partial \Sigma} + \frac{\partial l}{\partial \Sigma} \Sigma \right)$, $I$ is the identity matrix, and $\vecone \in \bbR{n}$ is the vector with all entries equal to 1. Here, $\symmetrize{A}=\frac{A+A^\top}{2}$.
\end{parisproposition}

\FloatBarrier

\subsection{Experiments}
\label{cornet:subsec:experiments}

\begin{table}[t]
    \centering
    \caption{Five-fold results and training time per epoch on four data sets. The top 3 results are highlighted with \redbf{red}, \bluebf{blue}, and \cyanbf{cyan}. $^*$ denotes reproduced results due to missing official code.}
    \label{cornet:tab:main-results}%
    \resizebox{\linewidth}{!}{
    \begin{tabular}{c|c|cc|cc|cc|cc}
    \toprule
    \multirow{2}[4]{*}{\textbf{Manifold}} & \multirow{2}[4]{*}{\textbf{Method}} & \multicolumn{2}{c|}{\textbf{Radar}} & \multicolumn{2}{c|}{\textbf{HDM05}} & \multicolumn{2}{c|}{\textbf{FPHA}} & \multicolumn{2}{c}{\textbf{NTU120}} \\
    \cmidrule{3-10}          &       & \textbf{Mean±STD} & \textbf{Time}  & \textbf{Mean±STD} & \textbf{Time}  & \textbf{Mean±STD} & \textbf{Time}  & \textbf{Mean±STD} & \textbf{Time} \\
    \midrule
    \multirow{3}[2]{*}{Grassmann} & GrNet \citep{huang2018building} & 90.48 ± 0.76 & 1.39  & 63.19 ± 0.70 & 1.64  & 85.31 ± 0.90 & 0.70  & 57.59 ± 0.22 & 50.97  \\
      & GyroGr$^*$ \citep{nguyen2023building} & 90.64 ± 0.57 & 1.38  & 58.32 ± 1.23 & 2.48  & 79.62 ± 0.49 & 0.70  & 53.76 ± 0.18 & 136.96  \\
      & GyroGr-Scaling$^*$ \citep{nguyen2023building} & 88.88 ± 1.52 & 1.63  & 39.75 ± 0.93 & 3.52  & 58.62 ± 1.66 & 1.03  & 43.90 ± 0.23 & 338.01  \\
    \midrule
    \multirow{11}[1]{*}{SPD} & SPDNet \citep{huang2017riemannian} & 93.25 ± 1.10 & 0.66  &   64.57 ± 0.61 & 0.50  & 85.59 ± 0.72 & 0.28  & 51.25 ± 0.36 & 12.77  \\
      & SPDNetBN \citep{brooks2019riemannian} & 94.85 ± 0.99 & 1.25  &   71.28 ± 0.79 & 0.94  & 89.33 ± 0.49 & 0.58  & 54.35 ± 0.43 & 19.78  \\
      & SPDResNet-AIM \citep{katsman2024riemannian} & 95.71 ± 0.37 & 0.96  &   64.95 ± 0.82 & 1.23  & 86.63 ± 0.55 & 0.69  & 57.33 ± 0.35 & 23.84  \\
      & SPDResNet-LEM \citep{katsman2024riemannian} & 95.89 ± 0.86 & 0.77  & 70.12 ± 2.45 & 0.55  & 85.07 ± 0.99 & 0.30  & 61.34 ± 2.02 & 13.00  \\
      & SPDNetLieBN-AIM \citep{chen2024liebn} & 95.47 ± 0.90 & 1.21  & 71.83 ± 0.69 & 1.15  & 90.39 ± 0.66 & 0.97  & 58.20 ± 0.46 & 31.10  \\
      & SPDNetLieBN-LCM \citep{chen2024liebn} & 94.80 ± 0.71 & 1.10  & 71.78 ± 0.44 & 1.11  & 86.33 ± 0.43 & 0.59  & 57.96 ± 0.43 & 22.06  \\
      & SPDNetMLR \citep{chen2024rmlr} & 94.59 ± 0.82 & 0.66  & 65.90 ± 0.93 & 5.46  & 85.60 ± 0.43 & 0.88  & 58.59 ± 0.13 & 22.48  \\
      & GyroLE$^*$ \citep{nguyen2023building} & 96.24 ± 0.24 & 0.79  & 73.17 ± 0.37 & 2.86  & 90.73 ± 0.92 & 1.59  & 59.29 ± 0.42 & 22.08  \\
      & GyroLC$^*$ \citep{nguyen2023building} & 93.60 ± 1.31 & 0.66  & 67.53 ± 0.85 & 1.49  & 76.10 ± 0.63 & 0.78  & 59.29 ± 0.42 & 14.14  \\
      & GyroAI$^*$ \citep{nguyen2023building} & 96.29 ± 0.48 & 0.99  & 72.34 ± 1.06 & 22.80  & 89.60 ± 0.37 & 12.62  & 62.21 ± 0.29 & 98.31  \\
      & GyroSPD++$^*$ \citep{nguyen2024matrix}& 95.20 ± 0.88 & 5.09  & 69.82 ± 1.79 & 103.57  & 89.50 ± 0.37 & 66.35  & 61.57 ± 0.30 & 216.46  \\
      \midrule
    \multirow{5}[1]{*}{Correlation} & \cellcolor{HilightColor}CorNet-ECM & \cellcolor{HilightColor}\bluebf{97.71 ± 0.61} & \cellcolor{HilightColor}1.01  & \cellcolor{HilightColor}\cyanbf{81.35 ± 1.27} & \cellcolor{HilightColor}0.60  & \cellcolor{HilightColor}\redbf{92.17 ± 0.49} & \cellcolor{HilightColor}0.50  & \cellcolor{HilightColor}\redbf{65.04 ± 0.14} & \cellcolor{HilightColor}12.06  \\
    & \cellcolor{HilightColor}CorNet-LECM & \cellcolor{HilightColor}\redbf{98.40 ± 0.70} & \cellcolor{HilightColor}1.12  & \cellcolor{HilightColor}78.05 ± 1.14 & \cellcolor{HilightColor}0.64  & \cellcolor{HilightColor}\cyanbf{91.17 ± 0.32} & \cellcolor{HilightColor}0.54  & \cellcolor{HilightColor}\bluebf{65.03 ± 0.10} & \cellcolor{HilightColor}12.68  \\
    & \cellcolor{HilightColor}CorNet-OLM & \cellcolor{HilightColor}\cyanbf{97.57 ± 0.76} & \cellcolor{HilightColor}1.35  & \cellcolor{HilightColor}\bluebf{81.46 ± 0.61} & \cellcolor{HilightColor}0.93  & \cellcolor{HilightColor}\bluebf{91.63 ± 0.12} & \cellcolor{HilightColor}0.79  & \cellcolor{HilightColor}\cyanbf{64.41 ± 0.23} & \cellcolor{HilightColor}16.07  \\
    & \cellcolor{HilightColor}CorNet-LSM & \cellcolor{HilightColor}96.24 ± 1.48 & \cellcolor{HilightColor}1.50  & \cellcolor{HilightColor}74.89 ± 1.07 & \cellcolor{HilightColor}0.98  & \cellcolor{HilightColor}83.43 ± 0.65 & \cellcolor{HilightColor}0.83  & \cellcolor{HilightColor}60.69 ± 0.85 & \cellcolor{HilightColor}16.28  \\
    & \cellcolor{HilightColor}CorNet-PHCM & \cellcolor{HilightColor}96.56 ± 0.86 & \cellcolor{HilightColor}2.37  & \cellcolor{HilightColor}\redbf{82.26 ± 0.92} & \cellcolor{HilightColor}1.10  & \cellcolor{HilightColor}90.03 ± 0.63 & \cellcolor{HilightColor}0.77  & \cellcolor{HilightColor}60.01 ± 0.22 & \cellcolor{HilightColor}16.92  \\
    \bottomrule
    \end{tabular}%
    }
    \end{table}%

We construct Riemannian networks on the correlation manifold, termed CorNets, using the proposed convolutional and MLR layers. Following previous work \citep{huang2017riemannian,brooks2019riemannian,chen2024liebn}, we evaluate our approach on the Radar data set \citep{brooks2019riemannian} for radar signal classification, along with the HDM05 \citep{muller2007documentation}, FPHA \citep{garcia2018first} and NTU120 \citep{liu2019ntu} data sets for human action recognition. More details on data sets and experimental settings are provided in \cref{app:datasets,cornet:app:experimental-details}.

\mypara{Implementation.} We denote CorNet-Metric as the CorNet composed of correlation convolution and MLR layers under a specified metric. In line with \citet{nguyen2024matrix}, each CorNet consists of one correlation convolutional layer followed by a correlation MLR layer, trained with cross-entropy loss. Following \citet{wang2024grassatt,nguyen2024matrix}, each raw feature is modeled as a multi-channel $[c, n, n]$ SPD tensor. Since matrix power effectively activates SPD matrices by deforming their geometry, as detailed in \cref{rmlr:subsec:geom_spd} and prior work \citep{thanwerdas2022geometry,chen2025understanding}, we first apply a matrix power, and then convert the result to correlation matrices as the input of CorNet. Due to trivialization, all trainable manifold-valued parameters are represented by Euclidean parameters and optimized by standard Euclidean optimizers. We compare CorNets against representative Grassmannian and SPD networks, including GrNet \citep{huang2018building}, GyroGr \citep{nguyen2023building}, GyroGr-Scaling \citep{nguyen2023building}, SPDNet \citep{huang2017riemannian}, SPDNetBN \citep{brooks2019riemannian}, RResNet \citep{katsman2024riemannian}, LieBN \citep{chen2024liebn}, SPD MLR \citep{chen2024rmlr}, Gyro \citep{nguyen2023building}, and GyroSPD++\citep{nguyen2024matrix}.

\subsubsection{Main Results}
\cref{cornet:tab:main-results} reports the five-fold results comparing our CorNets against existing SPD and Grassmannian baselines. We summarize the key observations below.
\begin{itemize}
\item \mypara{Effectiveness.} CorNets consistently outperform both SPD and Grassmannian networks. Specifically, CorNets surpass the classic SPDNet by \emph{5.15\%}, \emph{17.69\%}, \emph{6.58\%}, and \emph{13.79\%} on four data sets, respectively, and outperform the best Grassmannian networks by \emph{7.76\%}, \emph{19.07\%}, \emph{6.86\%}, and \emph{7.45\%}. Despite not using BN or residual blocks, CorNets achieve superior performance compared to SPDNetBN, SPDNetLieBN, and RResNet. Notably, although CorNets share the same high-level architecture as GyroSPD++ (one manifold convolutional layer followed by one manifold MLR layer), CorNets exhibit better performance. These results highlight the effectiveness of correlation embedding and our method for constructing correlation networks.
\item \mypara{Optimal Metric.} The optimal metric for CorNets varies across data sets, indicating that the choice of geometry is a critical hyperparameter in Riemannian networks. Our framework enables seamless switching among five correlation geometries in a consistent architecture, demonstrating the adaptability of our approach to different tasks.
\item \mypara{Efficiency.} CorNets achieve efficiency comparable to or better than several baseline methods. The most efficient CorNet variant is based on ECM, owing to the simplest computations of ECM. Although GyroSPD++ uses the same architecture, CorNets achieve significantly greater efficiency, attributed to the heavy computational cost of the AIM-based computations in GyroSPD++ and the lightweight Riemannian computations on the correlation manifold. Particularly, on the largest NTU120 data set, CorNet-ECM and CorNet-LECM are the top two most efficient ones.
\end{itemize}

\subsubsection{Ablations on Mixed Geometries}

\begin{table}[t]
  \centering
  \caption{
    Ablations on mixed geometries. Each row shows the metric used for Convolution (Conv), and each column is the metric for MLR. The \colorbox{HilightColor}{diagonal} entries indicate configurations where both layers use the same metric. The best result in each row is \firstresults{bold}.
    }
  \label{cornet:tab:ablations-metrics}%
  \resizebox{\linewidth}{!}{
    \begin{tabular}{c|ccccc|ccccc}
    \toprule
    \textbf{Data Set} & \multicolumn{5}{c|}{\textbf{HDM05}}            & \multicolumn{5}{c}{\textbf{FPHA}} \\
    \midrule
    \diagbox{\textbf{Conv}}{\textbf{MLR}} & \textbf{ECM}   & \textbf{LECM}  & \textbf{OLM}   & \textbf{LSM}   & \textbf{PHCM}  & \textbf{ECM}   & \textbf{LECM}  & \textbf{OLM}   & \textbf{LSM}   & \textbf{PHCM} \\
    \midrule
    ECM   & \cellcolor{HilightColor} \firstresults{81.35 ± 1.27} & 73.38 ± 0.34 &  80.11 ± 0.77 & 78.54 ± 0.43 & 80.80 ± 0.54 & \cellcolor{HilightColor} \firstresults{92.17 ± 0.49} & 91.50 ± 0.21 & 91.67 ± 0.28 & 87.37 ± 1.14 & 91.97 ± 0.24 \\
    LECM  & 66.49 ± 1.13 & \cellcolor{HilightColor} 78.05 ± 1.14 & \firstresults{79.21 ± 1.23} & 73.61 ± 0.99 & 58.37 ± 2.24 & 87.90 ± 0.57 & \cellcolor{HilightColor} \firstresults{91.17 ± 0.32} & 90.25 ± 0.25 & 89.63 ± 0.31 & 86.09 ± 0.98 \\
    OLM   & 77.82 ± 0.48 & 76.56 ± 0.89 & \cellcolor{HilightColor} \firstresults{81.46 ± 0.61} & 80.77 ± 0.81 & 77.39 ± 1.29 & 92.17 ± 0.58 & \firstresults{92.27 ± 0.78} & \cellcolor{HilightColor} 91.63 ± 0.12 & 89.90 ± 0.67 & 91.83 ± 0.15 \\
    LSM   & 68.83 ± 1.19 & 70.41 ± 1.57 & 67.56 ± 1.52 & \cellcolor{HilightColor} \firstresults{74.89 ± 1.07} & 72.69 ± 3.56 & 78.97 ± 2.80 & 75.10 ± 1.15 & 82.25 ± 3.38 & \cellcolor{HilightColor} \firstresults{83.43 ± 0.65} & 78.97 ± 4.97 \\
    PHCM   & 81.16 ± 0.40 & 80.05 ± 0.45 & 81.96 ± 0.51 & 78.28 ± 0.64 & \cellcolor{HilightColor} \firstresults{82.26 ± 0.92} & 88.30 ± 0.81 & 79.80 ± 0.69 &  87.37 ± 0.72 & 86.63 ± 0.27 & \cellcolor{HilightColor} \firstresults{90.03 ± 0.63} \\
    \bottomrule
    \end{tabular}%
    }
    \end{table}

Our main experiments use the same metric for convolution and MLR. To evaluate mixed geometries, we assign different metrics to the two layers. \cref{cornet:tab:ablations-metrics} reports five-fold results on HDM05 and FPHA. Overall, consistent metrics yield the best accuracy.

\subsubsection{Visualization}

\begin{figure}[H]
\centering
\includegraphics[width=\linewidth,trim={0cm 0cm 0cm 0cm}]{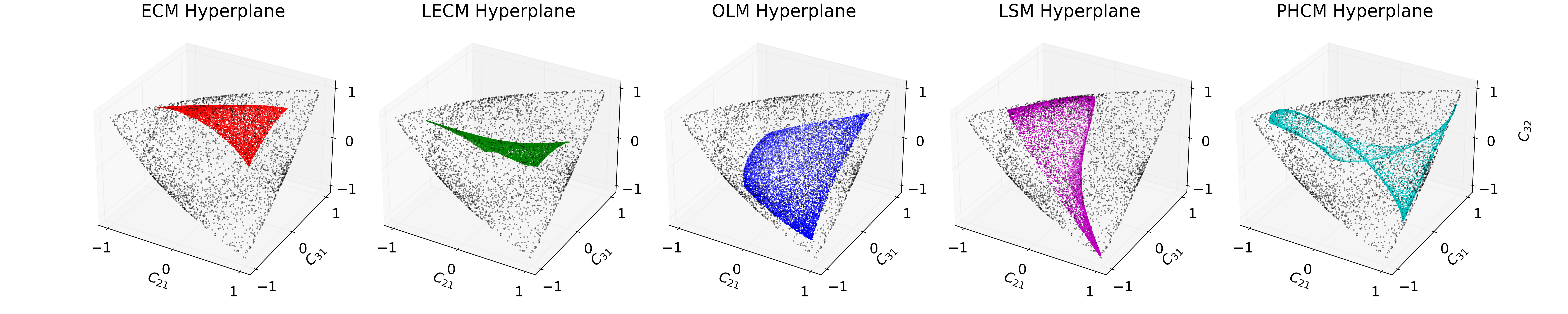}
\caption{Illustration of the decision hyperplanes in the correlation MLRs under five different geometries. The $3 \times 3$ correlation manifold can be embedded as an open elliptope in $\bbR{3}$, by visualizing the strictly lower triangular part of each $C \in \cor{3}$. The black dots denote the boundary. The PHCM hyperplane is defined by the one in the $\beta$-concatenated Poincaré space.
}
\label{cornet:fig:hyperplanes}
\end{figure}

\cref{cornet:fig:hyperplanes} shows that different metrics induce visibly distinct curved hyperplanes.

\Needspace{12\baselineskip}
\subsubsection{Potential and Necessity}

\begin{wraptable}{r}{0.6\linewidth}
    \centering
        \caption{SPDNet: SPD vs. correlation.}
    \label{cornet:tab:ablations-corinput}
    \resizebox{\linewidth}{!}{
    \begin{tabular}{c|ccc}
    \toprule 
    \textbf{Input} & \textbf{Radar} & \textbf{HDM05} & \textbf{FPHA} \\
    \midrule
    SPD   & \firstresults{93.25 ± 1.10} & 64.57 ± 0.61 & \firstresults{85.59 ± 0.72} \\
    \midrule
    \rowcolor{HilightColor} Correlation & 89.49 ± 0.67 & \firstresults{66.81 ± 0.73} & 83.37 ± 0.40 \\
    \bottomrule
    \end{tabular}
    }
    \end{wraptable}
Although correlation matrices are still SPD, naively treating them as SPD inputs and feeding them into existing SPD networks fails to leverage their intrinsic geometric structures. To illustrate this, we use the classic SPDNet \citep{huang2017riemannian} but replace its covariance inputs with correlation matrices. The five-fold average results in \cref{cornet:tab:ablations-corinput} reveal two key insights: (1) on the HDM05 data set, correlation inputs lead to improved performance, suggesting that correlation embeddings can serve as compact and effective alternatives to covariance representations; and (2) on the other two data sets, the performance degrades, indicating that ignoring the specific geometry of correlation matrices can be detrimental. These findings highlight both the promise and the necessity of designing networks respecting the unique geometry of the correlation manifold.

\subsubsection{Ablations on Correlation Embeddings}

\begin{table}[H]
  \centering
  \caption{Comparison of SPDMLR-Trivlz on raw covariances against CorMLR on raw correlations on all three data sets. The input matrix dimensions are $93 \times 93$, $63 \times 63$, and $20 \times 20$, respectively.}
  \label{cornet:tab:ablations-cormlr-spdmlr}%
  \resizebox{\linewidth}{!}{
    \begin{tabular}{c|c|ccc|ccccc}
    \toprule
    \multirow{2}[4]{*}{\textbf{Data Set}} & \multirow{2}[4]{*}{\textbf{Measurement}} & \multicolumn{3}{c|}{\textbf{SPDMLR-Trivlz}} & \multicolumn{5}{c}{\cellcolor{HilightColor} \textbf{CorMLR}} \\
    \cmidrule{3-10}          &       & \textbf{LEM}   & \textbf{LCM}   & \textbf{AIM}   & \cellcolor{HilightColor}\textbf{ECM} & \cellcolor{HilightColor}\textbf{LECM} & \cellcolor{HilightColor}\textbf{OLM} & \cellcolor{HilightColor}\textbf{LSM} & \cellcolor{HilightColor}\textbf{PHCM} \\
    \midrule
    \multirow{2}[2]{*}{Radar} & Acc   & 95.47 ± 0.66 & \firstresults{95.55 ± 0.35} & 94.87 ± 0.87 & \cellcolor{HilightColor}89.47 ± 0.93 & \cellcolor{HilightColor}87.41 ± 0.23 & \cellcolor{HilightColor}85.79 ± 0.83 & \cellcolor{HilightColor}91.63 ± 0.32 & \cellcolor{HilightColor}83.33 ± 1.29 \\
    & Fit Time (s/epoch) & 0.65  & 0.63  & 0.99  & \cellcolor{HilightColor}\firstresults{0.56} & \cellcolor{HilightColor}0.62  & \cellcolor{HilightColor}0.78  & \cellcolor{HilightColor}0.68  & \cellcolor{HilightColor}0.74 \\
    \midrule
    \multirow{2}[2]{*}{HDM05} & Acc   & 54.31 ± 1.65 & 45.12 ± 1.05 & 52.46 ± 2.44 & \cellcolor{HilightColor}\firstresults{65.57 ± 0.62} & \cellcolor{HilightColor}64.44 ± 0.63 & \cellcolor{HilightColor}62.86 ± 0.65 & \cellcolor{HilightColor}64.01 ± 0.92 & \cellcolor{HilightColor}62.78 ± 0.85 \\
    & Fit Time (s/epoch) & 3.24  & 5.38  & 260.67 & \cellcolor{HilightColor}3.18  & \cellcolor{HilightColor}3.87  & \cellcolor{HilightColor}3.39  & \cellcolor{HilightColor}3.57  & \cellcolor{HilightColor}\firstresults{2.73} \\
    \midrule
    \multirow{2}[2]{*}{FPHA} & Acc   & 84.13 ± 1.14 & 76.62 ± 0.43 & 83.25 ± 0.59 & \cellcolor{HilightColor}\firstresults{85.37 ± 0.16} & \cellcolor{HilightColor}85.24 ± 0.22 & \cellcolor{HilightColor}84.67 ± 0.27 & \cellcolor{HilightColor}80.17 ± 0.15 & \cellcolor{HilightColor}73.67 ± 0.32 \\
    & Fit Time (s/epoch) & 0.51  & 0.52  & 18.96 & \cellcolor{HilightColor}0.51  & \cellcolor{HilightColor}0.64  & \cellcolor{HilightColor}0.8   & \cellcolor{HilightColor}0.81  & \cellcolor{HilightColor}\firstresults{0.45} \\
    \bottomrule
    \end{tabular}
    }
    \end{table}%

To further evaluate the effectiveness of correlation embeddings, we compare the performance of directly classifying raw covariance matrices using the SPD MLRs in \cref{rmlr:thm:spdmlrs} with that of classifying corresponding raw correlation matrices using correlation MLR (CorMLR). The original SPDMLR involves an SPD matrix parameter for each class, which causes heavy Riemannian computations. For a fair comparison, we also implement a similar trivialization as \cref{cornet:subsubsec:log-euclidean-mlrs}, denoted as SPDMLR-Trivlz. We implement SPDMLR-Trivlz under LEM, LCM, and AIM, respectively. \cref{cornet:tab:ablations-cormlr-spdmlr} presents the 5-fold average results on all three data sets. CorMLR performs better than SPDMLR-Trivlz on HDM05 and FPHA. Although CorMLR performs worse on Radar, we emphasize that these comparisons are conducted on a single MLR layer, which fails to fully uncover the potential of correlation matrices. Besides, SPDMLR under AIM is much slower than others, especially on HDM05, due to its complex computations. In contrast, CorMLR, especially under ECM and PHCM, offers competitive or superior efficiency.

\raggedbottom
\subsubsection{Analysis of Covariance versus Correlation}
\label{cornet:subsubsec:covariance-vs-correlation}
In this section, we analyze when and why correlation matrices provide stronger representations than covariance matrices. The coefficient of variation of diagonal variances quantifies the variability of diagonal variances via per-sample coefficients of variation, and the ratio of diagonal to off-diagonal entries compares the magnitudes of diagonal and off-diagonal entries via their ratios. These analyses lead to two insights:
(1) large variability and magnitude of diagonal elements can act as nuisance noise for SPD networks by overshadowing informative off-diagonal correlations;
(2) under such cases, correlation representations that normalize variances and emphasize pairwise correlations tend to be more effective, which is especially evident on HDM05.

\mypara{Coefficient of Variation of Diagonal Variances.}

\begin{figure}[H]
    \centering
    \includegraphics[width=0.9\linewidth]{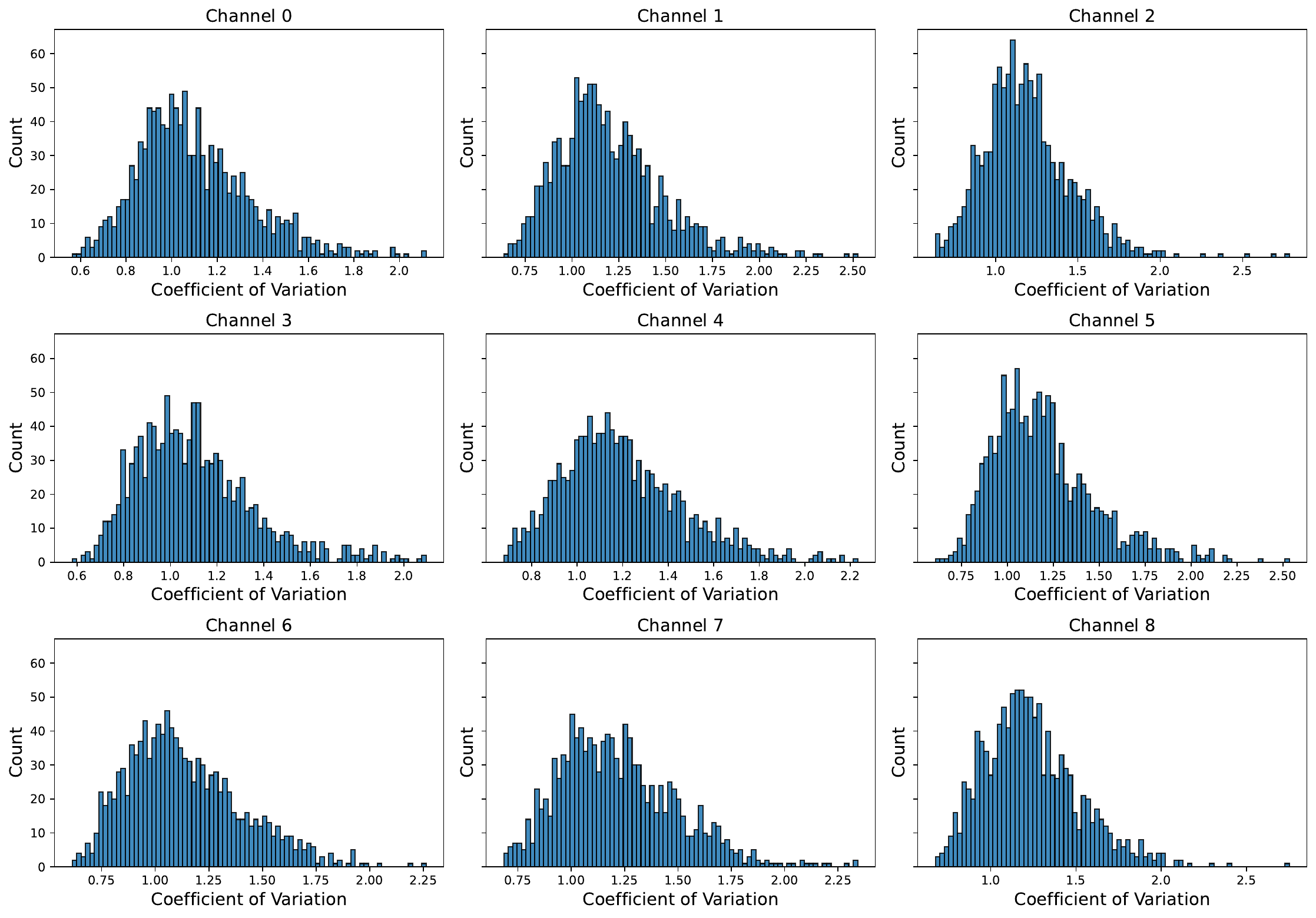}
    \caption{Distribution of per-sample coefficients of variation of diagonal variances on FPHA. Higher values indicate stronger diagonal variability, which could cause nuisance noise.}
    \label{cornet:fig:fpha-cov-cv}
\end{figure}

\begin{figure}[H]
    \centering
    \includegraphics[width=\linewidth]{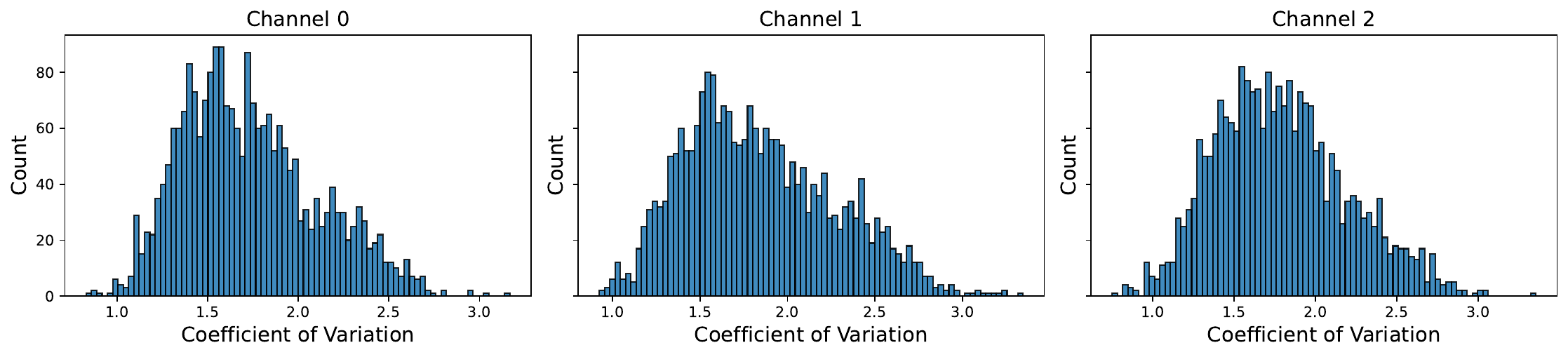}
    \caption{Distribution of per-sample coefficients of variation of diagonal variances on HDM05. Higher values indicate stronger diagonal variability, which could cause nuisance noise.}
    \label{cornet:fig:hdm05-cov-cv}
\end{figure}
This section investigates why CorNets yield substantially larger gains over SPD networks on HDM05 compared to FPHA.

\mypara{Setup.} For each covariance matrix $\Sigma \in \spd{n}$ we extract the diagonal vector
\begin{equation}
v = (\Sigma_{11}, \ldots, \Sigma_{nn}).
\end{equation}
We compute the coefficient of variation of $v$ as
\begin{equation}
\mathrm{CV} = \frac{\mathrm{std}(v)}{\mathrm{mean}(v) + \varepsilon} ,
\end{equation}
where $\varepsilon = 10^{-8}$ ensures numerical stability.
As shown in \cref{cornet:app:subsubsec:input-data}, each sequence is modeled as a $c$-channel tensor of covariance matrices. The above procedure yields one coefficient of variation per channel for each sample. We visualize their empirical distributions per channel.

\mypara{Analysis.} \cref{cornet:fig:fpha-cov-cv,cornet:fig:hdm05-cov-cv} show that the coefficients of variation w.r.t. diagonal variance are large on both data sets. On FPHA, most values fall between $0.8$ and $2.0$. On HDM05, they are even larger, typically between $1.0$ and $3.0$. Such large fluctuations indicate that diagonal variances change substantially and could bring nuisance noise for SPD networks. In contrast, correlation matrices allow CorNets to focus on pairwise relationships. This explains the consistent improvements over SPD networks and the larger gains on HDM05.

\FloatBarrier
\mypara{Ratio of Diagonal to Off-Diagonal Entries in Covariance Features.}

\begin{figure}[H]
    \centering
    \includegraphics[width=\linewidth]{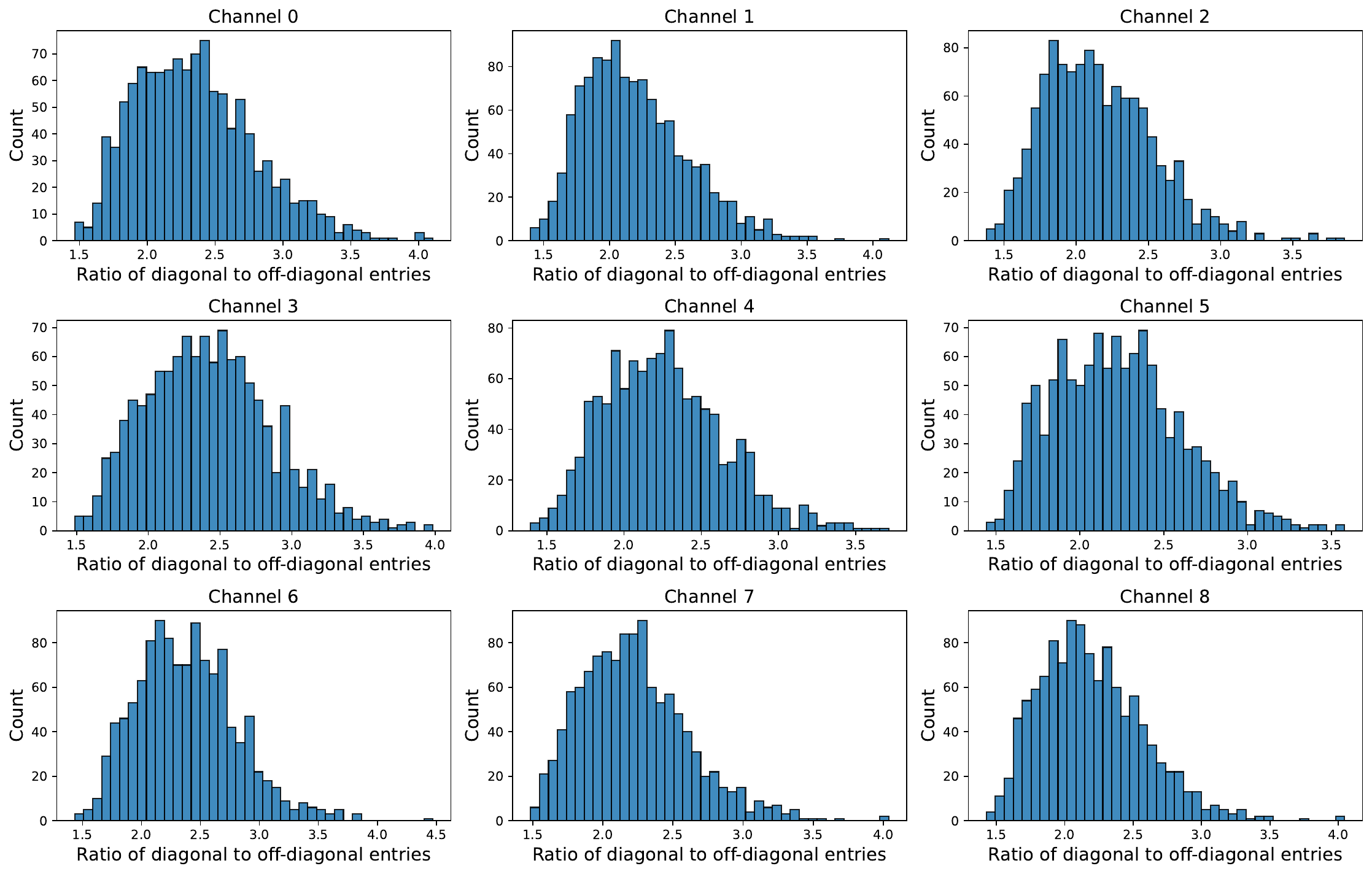}
    \caption{Distribution of ratios of diagonal to off-diagonal entries on FPHA.}
    \label{cornet:fig:fpha-ddr}
\end{figure}

\begin{figure}[H]
    \centering
    \includegraphics[width=\linewidth]{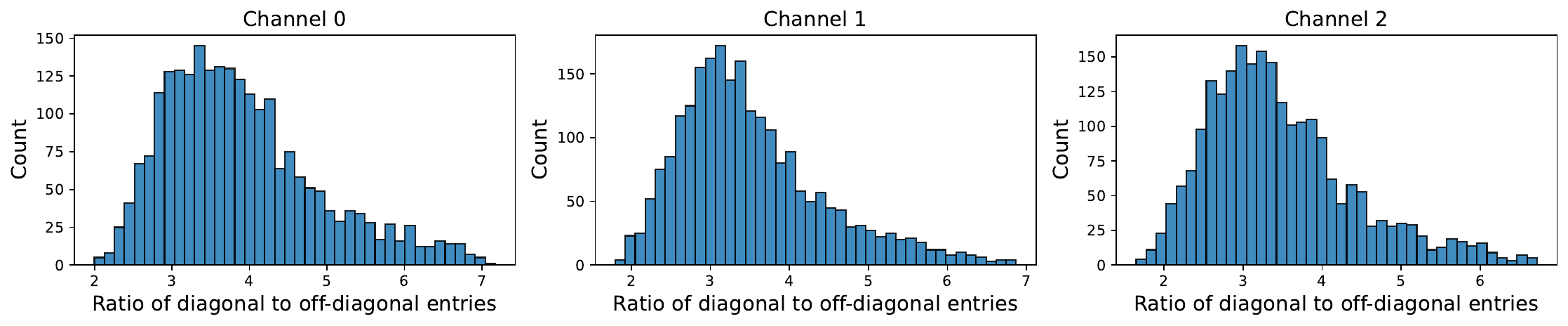}
    \caption{Distribution of ratios of diagonal to off-diagonal entries on HDM05.}
    \label{cornet:fig:hdm-ddr}
\end{figure}
This section further examines why CorNets achieve larger gains over SPD networks on HDM05 than on FPHA. We analyze the ratio of diagonal to off-diagonal entries in covariance matrices on FPHA and HDM05, to quantify how strongly variance terms overshadow pairwise correlations.

\mypara{Setup.} For each covariance matrix $\Sigma \in \spd{n}$ we compute the mean magnitude of diagonal entries
\begin{equation}
D = \frac{1}{n} \sum_{i=1}^{n} \lvert \Sigma_{ii} \rvert,
\end{equation}
and the mean magnitude of off-diagonal entries
\begin{equation}
O = \frac{1}{n(n-1)} \sum_{i \neq j} \lvert \Sigma_{ij} \rvert.
\end{equation}
We then form the sample-wise ratio
\begin{equation}
R = \frac{D}{O},
\end{equation}
which measures how much larger the diagonal amplitudes are compared to the off-diagonal correlations.
Each sample yields one ratio per channel, and we visualize the empirical distributions of these ratios on FPHA and HDM05.

\mypara{Analysis.} \cref{cornet:fig:fpha-ddr,cornet:fig:hdm-ddr} show that both data sets have ratios well above one. On FPHA, most ratios lie between $1.7$ and $3.0$, indicating that diagonal amplitudes are noticeably larger than off-diagonal correlations. HDM05 exhibits even larger ratios, typically between $2.0$ and $6.0$, with many above $3.0$. These statistics indicate that covariance representations on both data sets are strongly dominated by diagonal entries, with more pronounced dominance on HDM05. When diagonal terms dominate, SPD networks trained on covariance inputs tend to overemphasize variances and underexploit informative pairwise correlations. Correlation matrices normalize variances and highlight off-diagonal interactions, which explains why CorNets outperform SPD baselines on both data sets and why the improvement is substantially larger on HDM05.

\FloatBarrier

\subsubsection{Normalized Covariance vs. Correlation}
\label{cornet:subsubsec:normalized-covariance-vs-correlation}

\mypara{Setup.} We evaluate SPD-based baselines by covariance inputs normalized by their largest eigenvalue. Given a covariance matrix $\Sigma$, we get the normalized SPD input $\widehat{\Sigma} = \Sigma / \lambda_{\max}(\Sigma)$ and feed it into existing SPD networks. This variant is denoted by ``-EigN''. We report results on the Radar, HDM05, and FPHA data sets for representative SPD models: SPDNet, SPDNetBN, SPDResNet, SPDNetLieBN, SPDNetMLR, GyroAI, and GyroSPD++. Here, SPDResNet is implemented under the LEM, while SPDNetLieBN follows the LCM.

\begin{table}[H]
  \centering
  \caption{SPD networks with or without normalized SPD inputs.}
  \label{cornet:tab:spdnets-with-normalized-spd}%
  \resizebox{0.9\linewidth}{!}{
    \begin{tabular}{c|c|ccc}
    \toprule
    \textbf{Manifold} & \textbf{Method} & \textbf{Radar} & \textbf{HDM05} & \textbf{FPHA} \\
    \midrule
    \multirow{14}{*}{$\spd{n}$} & SPDNet & 93.25 ± 1.10 & 64.57 ± 0.61 & 85.59 ± 0.72 \\
           & SPDNet-EigN & 86.91 ± 0.57 & 66.62 ± 0.73 & 84.90 ± 0.62 \\
    \cmidrule{2-5}
           & SPDNetBN & 94.85 ± 0.99 & 71.28 ± 0.79 & 89.33 ± 0.49 \\
           & SPDNetBN-EigN & 89.25 ± 1.19 & 71.59 ± 0.68 & 88.47 ± 0.39 \\
    \cmidrule{2-5}
           & SPDResNet & 95.89 ± 0.86 & 70.12 ± 2.45 & 85.07 ± 0.99 \\
           & SPDResNet-EigN & 92.61 ± 0.96 & 71.02 ± 0.91 & 84.53 ± 0.46 \\
    \cmidrule{2-5}
           & SPDNetLieBN & 94.80 ± 0.71 & 71.78 ± 0.44 & 86.33 ± 0.43 \\
           & SPDNetLieBN-EigN & 88.91 ± 1.21 & 70.61 ± 1.04 & 83.73 ± 0.65 \\
    \cmidrule{2-5}
           & SPDNetMLR & 94.59 ± 0.82 & 65.90 ± 0.93 & 85.60 ± 0.43 \\
           & SPDNetMLR-EigN & 89.41 ± 0.58 & 66.89 ± 0.63 & 83.63 ± 1.09 \\
    \cmidrule{2-5}
           & GyroAI & 96.29 ± 0.48 & 72.34 ± 1.06 & 89.60 ± 0.37 \\
           & GyroAI-EigN & 91.36 ± 0.80 & 72.64 ± 0.70 & 89.90 ± 0.31 \\
    \cmidrule{2-5}
           & GyroSPD++ & 95.20 ± 0.88 & 69.82 ± 1.79 & 89.50 ± 0.37 \\
           & GyroSPD++-EigN & 90.83 ± 1.09 & 66.92 ± 0.28 & 84.29 ± 0.14 \\
    \midrule
    \multirow{5}{*}{$\cor{n}$} & \cellcolor{HilightColor}CorNet-ECM & \cellcolor{HilightColor}97.71 ± 0.61 & \cellcolor{HilightColor}81.35 ± 1.27 & \cellcolor{HilightColor}\firstresults{92.17 ± 0.49} \\
           & \cellcolor{HilightColor}CorNet-LECM & \cellcolor{HilightColor}\firstresults{98.40 ± 0.70} & \cellcolor{HilightColor}78.05 ± 1.14 & \cellcolor{HilightColor}91.17 ± 0.32 \\
           & \cellcolor{HilightColor}CorNet-OLM & \cellcolor{HilightColor}97.57 ± 0.76 & \cellcolor{HilightColor}81.46 ± 0.61 & \cellcolor{HilightColor}91.63 ± 0.12 \\
           & \cellcolor{HilightColor}CorNet-LSM & \cellcolor{HilightColor}96.24 ± 1.48 & \cellcolor{HilightColor}74.89 ± 1.07 & \cellcolor{HilightColor}83.43 ± 0.65 \\
           & \cellcolor{HilightColor}CorNet-PHCM & \cellcolor{HilightColor}96.56 ± 0.86 & \cellcolor{HilightColor}\firstresults{82.26 ± 0.92} & \cellcolor{HilightColor}90.03 ± 0.63 \\
    \bottomrule
    \end{tabular}%
  }
\end{table}%

\mypara{Results.} \cref{cornet:tab:spdnets-with-normalized-spd} summarizes the results. On HDM05, eigenvalue normalization has only a marginal effect and the normalized variants achieve accuracy comparable to their unnormalized counterparts. On FPHA and, in particular, on Radar, normalization usually reduces accuracy. The behavior of GyroSPD++ is especially informative. GyroSPD++ and CorNet share a similar architecture, consisting of one convolution followed by an MLR layer. However, GyroSPD++-EigN performs worse than GyroSPD++ on all three data sets, while CorNet with correlation inputs achieves clear improvements over GyroSPD++. These phenomena can be explained by two factors.
\begin{enumerate}
\item \mypara{Redundancy.} The raw samples on HDM05 and FPHA have already undergone centering, scaling, and normalization before covariance modeling. Dividing by $\lambda_{\max}(\Sigma)$ therefore introduces little additional control over scale, which explains the marginal effect on HDM05.

\item \mypara{Scaled Covariance versus Correlation.} Since EigN is equivalent to uniformly rescaling the raw samples before covariance computation, the normalized covariance matrices remain covariances and do not encode new statistical information. Moreover, forcing the largest eigenvalue to $1$ can remove potentially informative differences in overall energy across samples, which aligns with the degradation observed for EigN variants, especially GyroSPD++-EigN. In contrast, correlation normalization uses a different scaling factor for each pair of variables,
\begin{equation}
\mathrm{Cor}_{ij} = \frac{\Sigma_{ij}}{\sqrt{\Sigma_{ii}\Sigma_{jj}}},
\end{equation}
producing standardized correlation coefficients. Therefore, global eigenvalue scaling is statistically distinct from correlation normalization and fails to capture the benefits of explicit correlation modeling.
\end{enumerate}

\flushbottom
\subsubsection{Ablations on Activations}
\label{cornet:subsubsec:ablations-activations}

\begin{table}[H]
  \centering
  \caption{Comparison of CorNet with or without activations.}
  \label{cornet:tab:activations}%
  \resizebox{0.9\linewidth}{!}{
    \begin{tabular}{c|c|c|cc}
    \toprule
    \textbf{Metric} & \textbf{Activation} & \textbf{Radar} & \textbf{HDM05} & \textbf{FPHA} \\
    \midrule
    \multirow{2}[2]{*}{ECM} & ReLU  & 97.41 ± 0.25 & 81.23 ± 0.46 & 89.80 ± 0.58 \\
          & \cellcolor{HilightColor} None & \cellcolor{HilightColor} \firstresults{97.71 ± 0.61} & \cellcolor{HilightColor} \firstresults{81.35 ± 1.27} & \cellcolor{HilightColor} \firstresults{92.17 ± 0.49} \\
    \midrule
    \multirow{2}[2]{*}{LECM} & ReLU  & 97.23 ± 0.67 & 77.51 ± 1.02 & 91.00 ± 0.15 \\
          & \cellcolor{HilightColor} None & \cellcolor{HilightColor} \firstresults{98.40 ± 0.70} & \cellcolor{HilightColor} \firstresults{78.05 ± 1.14} & \cellcolor{HilightColor} \firstresults{91.17 ± 0.32} \\
    \midrule
    \multirow{2}[2]{*}{OLM} & ReLU  & 97.52 ± 0.47 & \firstresults{81.86 ± 0.65} & 91.47 ± 0.19 \\
          & \cellcolor{HilightColor} None & \cellcolor{HilightColor} \firstresults{97.57 ± 0.76} & \cellcolor{HilightColor} 81.46 ± 0.61 & \cellcolor{HilightColor} \firstresults{91.63 ± 0.12} \\
    \midrule
    \multirow{2}[2]{*}{LSM} & ReLU  & 95.60 ± 0.97 & \na   & \na \\
          & \cellcolor{HilightColor} None & \cellcolor{HilightColor} \firstresults{96.24 ± 1.48} & \cellcolor{HilightColor} \firstresults{74.89 ± 1.07} & \cellcolor{HilightColor} \firstresults{83.43 ± 0.65} \\
    \midrule
    \multirow{2}[2]{*}{PHCM} & ReLU  & 96.40 ± 0.25 & 77.32 ± 1.56 & 88.63 ± 0.22 \\
          & \cellcolor{HilightColor} None & \cellcolor{HilightColor} \firstresults{96.56 ± 0.86} & \cellcolor{HilightColor} \firstresults{82.26 ± 0.92} & \cellcolor{HilightColor} \firstresults{90.03 ± 0.63} \\
    \bottomrule
    \end{tabular}%
  }
\end{table}%

In the main experiments, we follow HNN++~\citep{shimizu2021hyperbolic} and GyroSPD++~\citep{nguyen2024matrix}, and do not use explicit activations, as the manifold itself introduces nonlinearity. We further conduct an ablation on activations. Following \citet[Sec.~3.2]{ganea2018hyperbolic}, we define activations in the tangent space at the identity, \ie $\rieexp_I \circ \delta \circ \rielog_I$ for four Log-Euclidean metrics, and $\rieexp_{\zerovec} \circ \delta \circ \rielog_{\zerovec}$ for PHCM in the $\beta$-concatenated Poincaré vector, where $\delta$ is ReLU~\citep{glorot2011deep}. Specifically, we insert a ReLU after the correlation convolution. As shown in \cref{cornet:tab:activations}, adding activations generally yields no benefits and can even degrade performance. The variant without activation consistently achieves higher or comparable accuracy, except CorNet-OLM for HDM05. Moreover, CorNet-LSM with activation fails to converge on HDM05 and FPHA. These results suggest that CorNet already provides sufficient nonlinearity, rendering additional activations redundant.

\Needspace{24\baselineskip}
\subsubsection{Scalability of Correlation Metrics}
\label{cornet:subsubsec:efficiency-metrics-dim}
\begin{table}[H]
    \centering
    \caption{Average runtime (s) of a single forward pass in CorNet under different metrics and input dimensions. The best results are \firstresults{bold}.}
    \label{cornet:tab:runtime-corr-mlr}
    \begin{tabular}{c|ccccc}
    \toprule
    \textbf{Dim} & \textbf{ECM} & \textbf{LECM} & \textbf{OLM} & \textbf{LSM} & \textbf{PHCM} \\
    \midrule
    30    & \firstresults{0.0004 } & 0.0018  & 0.0012  & 0.0019  & 0.0131  \\
    50    & \firstresults{0.0004 } & 0.0027  & 0.0318  & 0.0334  & 0.0211  \\
    100   & \firstresults{0.0008 } & 0.0054  & 0.0764  & 0.0781  & 0.0413  \\
    150   & \firstresults{0.0015 } & 0.0100  & 0.1247  & 0.1267  & 0.2284  \\
    200   & \firstresults{0.0025 } & 0.0197  & 0.1906  & 0.1938  & 0.3320  \\
    250   & \firstresults{0.0037 } & 0.0345  & 0.2352  & 0.2379  & 0.4414  \\
    300   & \firstresults{0.0053 } & 0.0733  & 0.3434  & 0.3454  & 0.5732  \\
    400   & \firstresults{0.0092} & 0.1796 & 0.5163 & 0.5261 & 0.4807 \\
    500   & \firstresults{0.0143} & 0.3076 & 0.6907 & 0.6961 & 0.5693 \\
    600   & \firstresults{0.0206} & 0.5983 & 0.9331 & 0.9484 & 0.7923 \\
    700   & \firstresults{0.0289} & 1.0961 & 1.2432 & 1.2575 & 1.0417 \\
    800   & \firstresults{0.039} & 1.8689 & 1.6658 & 1.6815 & 1.3387 \\
    900   & \firstresults{0.0535} & 2.9886 & 2.2156 & 2.2303 & 1.7324 \\
    1000  & \firstresults{0.0706} & 3.7259 & 2.539 & 2.5783 & 1.229 \\
    \bottomrule
    \end{tabular}%
\end{table}

We evaluate the computational efficiency of correlation metrics across increasing input dimensions using CorNet with one correlation FC layer followed by one correlation MLR layer. Each input correlation matrix of size $[n,n]$ is mapped to $[20,20]$ by the FC layer and then classified into $10$ classes by the MLR layer. For each listed dimension $30 \leq n \leq 1000$, we randomly generate $30$ correlation matrices and record the average runtime of a single forward pass. As implied by \cref{tab:ch2-correlation-maps}, the runtime is governed by two factors: the codomain computation (Euclidean or hyperbolic) and the complexity of the diffeomorphism. The results are summarized in \cref{cornet:tab:runtime-corr-mlr}. We have the following findings.

\begingroup
\setlength{\abovedisplayskip}{6pt}
\setlength{\belowdisplayskip}{6pt}
\setlength{\abovedisplayshortskip}{3pt}
\setlength{\belowdisplayshortskip}{3pt}
\begin{itemize}[nosep]
    \item
    ECM is consistently the most efficient metric, benefiting from both a Euclidean codomain and the simplest diffeomorphism.

    \item
    At very low dimensions ($n \leq 100$), the relative costs of the non-ECM metrics are not yet stable. At $n=50$ and $n=100$, the ordering is
    \begin{equation}
         \text{ECM} < \text{LECM} < \text{PHCM} < \text{OLM} \approx \text{LSM}.
    \end{equation}
    At the smallest tested dimension $n=30$, all runtimes remain small and their relative ordering differs. At intermediate dimensions ($150 \leq n \leq 300$), the ordering becomes
    \begin{equation}
         \text{ECM} < \text{LECM} < \text{OLM} \approx \text{LSM} < \text{PHCM}.
    \end{equation}
    Here, the cost of PHCM's hyperbolic computations dominates, while the dimension-dependent cost of LECM's matrix functions is not yet pronounced.

    \item
    From $n=400$, PHCM becomes faster than OLM and LSM, and at $n=700$, it also becomes faster than LECM. At high dimensions ($n \geq 800$), the ordering is
    \begin{equation}
        \text{ECM} < \text{PHCM} < \text{OLM} \approx \text{LSM} < \text{LECM}.
    \end{equation}
    Here, diffeomorphisms dominate: ECM and PHCM scale better thanks to relatively lightweight Cholesky decomposition, while OLM and LSM slow down due to matrix logarithm/exponentiation. LECM is the slowest, as its $\log \circ \Theta$ requires two nested matrix functions.
\end{itemize}
\endgroup

\FloatBarrier

\section{Conclusion}
\label{sec:ch5-conclusion}
This chapter developed manifold-specific neural components and architectures by exploiting the additional structures of particular hyperbolic models and correlation manifolds.

The first part addressed representation choice through PVNN. The unconstrained PV model is connected to the Poincar\'e and Lorentz models by Riemannian isometries, but it avoids their explicit constraints. By deriving closed-form core Riemannian operators and relating them to the PV gyrovector structure, we constructed PV MLR, FC, convolutional, activation, and normalization layers. Together, these layers constitute a complete PVNN framework for constructing model-specific hyperbolic architectures. Experiments demonstrate the superior numerical stability and competitive performance of the PV representation.

The second part shifted attention from the representation itself to the geometric principle for building neural layers. Using Busemann functions and horospheres, we developed a common construction for the Poincar\'e and Lorentz models. BMLR admits an exact point-to-horosphere interpretation, avoids an additional manifold-valued point parameter, supports batch-efficient evaluation, and recovers Euclidean MLR in the zero-curvature limit. The same Busemann logits yield explicit BFC layers with practical $\calO(nm)$ complexity and corresponding Euclidean limits. Experiments on image classification, genome sequence learning, node classification, and link prediction showed that these layers generally improve upon existing hyperbolic alternatives while retaining comparable computational cost.

The third part extended manifold-specific network design from hyperbolic vectors to full-rank correlation matrices. The four flat geometries, ECM, LECM, OLM, and LSM, admit Euclidean isometries that yield closed-form correlation MLR, FC, and convolutional layers. For the non-flat PHCM geometry, the Cholesky representation identified correlation matrices with a product of Poincar\'e balls, enabling analogous layers through $\beta$-concatenation and $\beta$-splitting. Analytic gradients for the Riemannian computations enabled accurate end-to-end backpropagation under OLM and LSM. Across the four evaluated data sets, the best CorNet geometry outperformed the considered SPD and Grassmannian baselines.

All three designs nevertheless operate under prescribed Riemannian geometries. The next chapter therefore turns from the design of modules and architectures to that of the underlying metrics themselves.

    \chapter{Fast and Stable Geometries on SPD Manifolds}
\label{chapter:spd-geometries}

\section{Introduction}
\label{sec:ch6-introduction}

The preceding chapters developed Riemannian neural components and architectures under prescribed Riemannian metrics. Despite their different constructions, they share the underlying metric as a common starting point.

As reviewed in \cref{chapter:2}, a Riemannian metric assigns inner products to tangent spaces and determines distance, geodesics, exponential and logarithmic maps, and parallel transport. The induced distance also defines the Fr\'echet statistics used to summarize manifold-valued features. When combined with compatible Lie group or gyrovector structures, a Riemannian metric further supports manifold analogues of addition and scalar multiplication. These geometric primitives offer powerful toolkits for building Riemannian neural networks.

Based on this observation, we shift our attention from designing neural modules and architectures under prescribed geometries to designing the underlying geometry itself. Our goal is to balance geometric flexibility, theoretical convenience, computational efficiency, and numerical stability. Accordingly, we seek geometries whose Riemannian operators and compatible algebraic operations admit closed-form expressions and can be inserted directly into deep learning architectures. We focus on the SPD manifold, which encodes covariance or other second-order statistics and arises naturally in many applications \citep{chakraborty2020manifoldnet,brooks2019riemannian,li2025spdim,dan2024exploring,moakher2006averaging,lopez2021vector,zhao2023modeling,huang2017riemannian}.\footnote{Our focus is distinct from SPD metric-learning methods that learn distance functions induced by an existing Riemannian metric. Instead, we seek to design new Riemannian metrics themselves.}

We pursue two complementary routes. \cref{sec:alem} starts from pullback Euclidean geometry and proposes Adaptive Log-Euclidean Metrics (ALEMs). The resulting adaptive geometry retains closed-form Riemannian operators and a convenient abelian Lie group structure. \cref{sec:pcm-bwcm} instead exposes the product structure of the Cholesky manifold and transfers these geometries to the SPD manifold, yielding simple and stable closed-form Riemannian and gyro operators.

\section{Adaptive Log-Euclidean Metrics}
\label{sec:alem}

\subsection{Introduction}
\label{alem:sec:intro}

As reviewed in \cref{sec:ch2-spd-manifolds}, most popular Riemannian metrics on the SPD manifold are fixed, which can limit the expressive capacity of the associated geometry. A common way to construct SPD metrics is through pullbacks along diffeomorphisms, which transfer Riemannian structures from simpler source manifolds. For instance, \citet{thanwerdas2022theoretically} explained AIM as the pullback metric from a left-invariant metric on the Cholesky manifold. The matrix-power deformations in \cref{subsubsec:spd_param_lie_groups,rmlr:subsec:geom_spd} are also representative examples.

Inspired by the above observations, we leverage pullback techniques to introduce adaptive Riemannian metrics. In particular, we first show that several Riemannian metrics on SPD manifolds, including LEM, LCM, and their generalizations, can be explained as pullback metrics from the standard Euclidean space.
We refer to these metrics as pullback Euclidean metrics.
Then, we propose a general framework for characterizing the properties of pullback Euclidean metrics. Our framework can explain the widely used LEM \citep{arsigny2005fast} and LCM \citep{lin2019riemannian}.
We focus on LEM on SPD manifolds and extend it into \emph{Adaptive Log-Euclidean Metrics (ALEMs)}.
Besides, we present a complete theoretical discussion on the properties of ALEMs. 
Compared with the existing Riemannian metrics, our metrics are learnable, adapting to the characteristics of the data sets.
The effectiveness of our metrics is demonstrated by experiments as well as the applications to recently developed Riemannian building blocks, including the RBN framework developed in \cref{chapter:normalization}, Riemannian residual blocks \citep{katsman2023riemannian}, and the Riemannian classifiers developed in \cref{chapter:rmlr}.
Drawing on this, our \textbf{contributions} are summarized as follows:
\begin{enumerate}
    \item
    We reveal the connection of two popular Riemannian metrics (LEM and LCM) by the pullback technique and propose a general framework for pullback Euclidean metrics.
    \item
    Based on our framework, we propose specific ALEMs on SPD manifolds and conduct comprehensive analyses in terms of the algebraic, analytic, and geometric properties.
    \item
    Extensive experiments on widely used SPD learning benchmarks demonstrate that our metrics exhibit consistent performance gain across data sets.\footnote{The code is available at \url{https://github.com/GitZH-Chen/ALEM}.}
\end{enumerate}

\mypara{Outline.} The necessary background on differential geometry, pullback metrics, and SPD geometry has been established in \cref{chapter:2}. \cref{alem:sec:adaptive-log-euclidean-metrics} develops ALEM and its differentials, and \cref{alem:sec:properties} studies its geometric properties. \cref{alem:sec:param_learning} derives the gradients and parameter-update rules for the general matrix logarithm and exponential. \cref{alem:sec:experiments} instantiates the general matrix logarithm as ALog in SPDNet and applies ALEM to other Riemannian building blocks. Proofs are deferred to \cref{app:alem-proofs}.

\subsection{Adaptive Log-Euclidean Metrics}
\label{alem:sec:adaptive-log-euclidean-metrics}
In this section, we show that both $\biparamLEM$ and LCM are pullback metrics from the Euclidean space.
Inspired by this observation, we present a general framework for characterizing pullback Euclidean metrics.
Then, we focus on generalizing LEM.

\subsubsection{Rethinking \texorpdfstring{$\biparamLEM$}{LEM} and LCM} \label{alem:subsec:thk_lem_lcm}
Among the existing Riemannian metrics on the SPD manifold, LEM is popular in many applications, given its closed form for the Fréchet mean and clear vector-space and Lie-group structures.
In addition, the nascent LCM, which is gaining increasing attention, shares similar properties with LEM.
LEM is derived from Lie group translation \citep{arsigny2005fast}, while LCM is obtained as the pullback of a metric on $\chospace{n}$ \citep{lin2019riemannian}.
Besides, $\biparamLEM$ is obtained as a pullback of LEM \citep{thanwerdas2023n}.
However, the same mathematical logic underlies their derivations.
We denote the Euclidean space of $n \times n$ lower triangular matrices by $\trilspace{n}$.
We define $\clog: \spd{n} \rightarrow \trilspace{n}$ as
\begin{equation} \label{alem:eq:lcm_pullback_map}
    \clog(P)=\lfloor L \rfloor + \dlog(\bbD(L)),
\end{equation}
where $L$ is the Cholesky factor of the SPD matrix $P$, $\lfloor L \rfloor$ is the strictly lower part of $L$, $\bbD(L)$ is a diagonal matrix with diagonal elements of $L$, and $\dlog$ applies the natural logarithm element-wise to the diagonal.
Then, we have the following theorem.
\begin{paristheorem} \label{alem:thm:rethk_lem_lcm}
    \linktoproof{alem:thm:rethk_lem_lcm}
    $\biparamLEM$ is the pullback metric from the Euclidean space of $\sym{n}$ with an $\orth{n}$-invariant inner product $\langle \cdot, \cdot \rangle^{\biparam}$ by matrix logarithm.
    Specifically, the standard LEM is the pullback metric from the Euclidean space of $\sym{n}$ with the standard Frobenius inner product by matrix logarithm.
    LCM is the pullback metric from $\trilspace{n}$ with the Frobenius inner product by $\clog$.
\end{paristheorem}

As Euclidean spaces of the same dimension are naturally isometric, it follows that both $\biparamLEM$ and LCM are pulled back from the standard Euclidean space $\sym{n}$.

\begin{pariscorollary} \label{alem:cor:biparamLEM_pem}
    \linktoproof{alem:cor:biparamLEM_pem}
    $\biparamLEM$ and LCM are pullback metrics from $\sym{n}$ with the standard Frobenius inner product.
\end{pariscorollary}

\subsubsection{Pullback Euclidean Metrics on SPD Manifolds}
\label{alem:subsec:pem_spd}

\cref{alem:subsec:thk_lem_lcm} has shown how LEM is derived from matrix logarithm.
Besides, as shown in \citet{arsigny2005fast}, operations in Lie group and linear space on $\spd{n}$ are also induced from matrix logarithm.
Now, let us explain the underlying mechanism in detail.
A matrix logarithm is a diffeomorphism (a smooth bijection with a smooth inverse).
The property of bijection offers the possibility of transferring algebraic structures from $\sym{n}$ into $\spd{n}$.
The smoothness of the matrix logarithm and its inverse suggests that smooth structures, such as a Lie group structure and a Riemannian metric, can be transferred to $\spd{n}$.
More generally, given an arbitrary diffeomorphism $\phi:\spd{n} \rightarrow \sym{n}$, it suffices to pull various properties from the Euclidean space back to the SPD manifold $\spd{n}$ by $\phi$ as well.
Besides, the computation of the induced operators in $\spd{n}$ by $\phi$ is usually simple.

\begin{parislemma} \label{alem:lem:g_spd}
    \linktoproof{alem:lem:g_spd}
    Let $S_1, S_2 \in \spd{n}$, $V \in T_{S_1}\spd{n}$, and $k \in \bbRscalar$, and let $g^{\rmE}$ be the Frobenius inner product in $\sym{n}$.
    Let $\phi:\spd{n} \rightarrow \sym{n}$ be a diffeomorphism, and denote its differential at $S\in\spd{n}$ by $\phi_{*,S}$.
    We define the following operations:
    \begin{align}
        \label{alem:eq:phi_mul} 
        \text{Element Addition: }& S_1 \phiMul S_2 = \phiinv( \phi(S_1)+\phi(S_2)),\\
        \label{alem:eq:phi_sca_mul}
        \text{Scalar Multiplication: }& k \phiMulScalar S_2 = \phiinv( k\phi(S_2)),\\
        \label{alem:eq:phi_innerpro}
        \text{Inner Product: }& \langle S_1, S_2 \rangle_{\phi} = \langle \phi(S_1), \phi(S_2) \rangle,\\
        \label{alem:eq:phi_g} 
        \text{Riemannian Metric: }& \gphi = \phi^*\geuc,
    \end{align}
    Then, we have the following conclusions:
    \par\leavevmode\vspace{-\baselineskip}
    \begin{enumerate}
        \item \label{alem:itm:spd_hilbet}
        $\{\spd{n}, \phiMul,\phiMulScalar, \langle \cdot, \cdot \rangle_{\phi} \}$ is a Hilbert space over $\bbRscalar$.
        \item 
        $\{\spd{n}, \phiMul \}$ is an abelian Lie group.
        $\{\spd{n}, \gphi \}$ is a Riemannian manifold.
        The associated Riemannian operators are as follows:
        \begin{align}
            \label{alem:eq:dist_phi_spd}
            \dphi (S_1, S_2 ) &= \| \phi(S_1) - \phi(S_2) \|_\rmF,\\
            \label{alem:eq:gene_rie_exp_spd}
            \rieexp_{S_1} V &= \phiinv(\phi(S_1)+\diffphi{S_1}V),\\
            \label{alem:eq:gene_rie_log_spd}
            \rielog_{S_1}S_2 &= \diffphiinv{\phi(S_1)}(\phi(S_2)-\phi(S_1)),\\
            \label{alem:eq:gene_pt_spd} \pt{S_1}{S_2}(V) &= \diffphiinv{\phi(S_2)} \circ \diffphi{S_1}(V),
        \end{align}
        where $\|\cdot\|_\rmF$ is the Frobenius norm, $V \in T_{S_1}\spd{n}$ is a tangent vector, $\rieexp_{S_1}$, $\rielog_{S_1}$, and $\pt{S_1}{S_2}$ are the Riemannian exponential map at $S_1$, logarithmic map at $S_1$, and parallel transport along the geodesic connecting $S_1$ and $S_2$, respectively, and $\phiinv_{*}$ denotes the differential of $\phiinv$.
        Then $\gphi$ is a bi-invariant metric, called a \emph{Pullback Euclidean Metric} induced by $\phi$.
        \item
        $\phi$ is an isomorphism: (a) a linear isomorphism preserving the inner product; (b) a Lie group isomorphism; (c) a Riemannian isometry.
    \end{enumerate}
\end{parislemma}
In fact, $\biparamLEM$ and LCM are special cases of \cref{alem:lem:g_spd}, as are the linear-space and Lie-group structures in \citet{arsigny2005fast} and the Lie-group structure in \citet{lin2019riemannian}.
In addition, neither \citet{arsigny2005fast} nor \citet{lin2019riemannian} reveals the Hilbert space structures in $\spd{n}$.
\subsubsection{Adaptive Log-Euclidean Metrics} \label{alem:subsec:ada_rie_metric}
The key to \cref{alem:lem:g_spd} lies in the diffeomorphism $\phi$.
If we have a proper $\phi$, Riemannian metrics on SPD manifolds can be induced.
In the following, we will present our mappings and then discuss the induced metrics.

As reviewed in \cref{sec:ch2-matrix-functions-differentials}, the matrix logarithm reduces to a scalar logarithm, which is a diffeomorphism between $\bbRplusscalar$ and $\bbRscalar$.
Following this hint, the eigenvalue-based diffeomorphism between $\spd{n}$ and $\sym{n}$ reduces to a scalar diffeomorphism between $\bbRplusscalar$ and $\bbRscalar$.
A very natural idea is to substitute the natural logarithm with logarithms with arbitrary proper bases.
Throughout this part, $\log(\cdot)$ without a subscript denotes the natural scalar or matrix logarithm. Every scalar or matrix logarithm with a general or adaptive base is written explicitly as $\log_\alpha(\cdot)$, without omitting the subscript. The base parameter $\alpha$ is interpreted as either a scalar or a vector according to its argument.
For a scalar base $\alpha \in \bbRplusscalar \setminus \{1\}$ and $x \in \bbRplusscalar$, we define
\begin{equation}
    \log_\alpha(x) = \frac{\log(x)}{\log(\alpha)}.
\end{equation}
When $\alpha=e$, this scalar logarithm reduces to the natural logarithm, which we write without a subscript as $\log(\cdot)$.
For a diagonal matrix $X$, the same notation is extended to a base vector as
\begin{equation} \label{alem:eq:diag_glog}
    \log_\alpha(X) = \diag(\log_{a_1}(x_{11}),\log_{a_2}(x_{22}),\cdots,\log_{a_n}(x_{nn})),
\end{equation}
where $\alpha = (a_1, a_2, \cdots, a_n) \in \left(\bbRplusscalar \setminus \{1\}\right)^{n}$ is the base vector, $\diag(\cdot)$ is the diagonalization operator, and $X$ is an $n \times n$ diagonal matrix. When $\alpha$ is scalar in a matrix expression, it denotes the constant base vector $(\alpha,\ldots,\alpha)$.
Together with eigendecomposition, a general matrix logarithm is defined by
\begin{equation} \label{alem:eq:mlog}
     \log_\alpha(S) = U \log_\alpha(\Sigma) U^\top,
\end{equation}
where $S = U \Sigma U^\top$ is the eigendecomposition.
As a special case,
\begin{equation}
    \alpha=(e,e,\cdots,e) \quad \Longrightarrow \quad \log_\alpha = \log.
\end{equation}
As with the scalar logarithm, we have the following proposition.
\begin{parisproposition}[Diffeomorphism] \label{alem:props:diffeo_mlog}
    \linktoproof{alem:props:diffeo_mlog}
    $\log_\alpha$ is a diffeomorphism, a smooth bijection with a smooth inverse $\log_\alpha^{-1}:\sym{n} \rightarrow \spd{n}$ defined as
    \begin{equation}\label{alem:eq:mgexp}
        \log_\alpha^{-1}(X) = U \diag(a_1^{\Sigma_{11}},a_2^{\Sigma_{22}},\cdots,a_n^{\Sigma_{nn}}) U^\top,
    \end{equation}
    where $X=U\Sigma U^\top$ is the eigendecomposition.
\end{parisproposition}
\begin{parisremark} \label{alem:rmk:proposed_charts}
    The general matrix logarithm $\log_\alpha$ is an arbitrary member of the following family
    \begin{equation}
        \left\{\log_\alpha \mid \alpha = (a_1, \cdots, a_n) \in \left(\bbRplusscalar \setminus \{1\}\right)^{n}\right\}.
    \end{equation}
    Besides, there could be some ambiguity in \cref{alem:eq:mlog} under different arrangements of eigenvalues and eigenvectors.
    In fact, there is a correspondence between scalar $\log_{a_i}$ and eigenvalues and eigenvectors.
    See \cref{alem:app:subsec:well_difined_glog} for more details.
\end{parisremark}
Since $\log_\alpha$ is a diffeomorphism from $\spd{n}$ onto $\sym{n}$, all the results in \cref{alem:lem:g_spd} hold true.
\begin{paristheorem} \label{alem:thm:mlog_spd_properties}
    \linktoproof{alem:thm:mlog_spd_properties}
    Following the notation in \cref{alem:lem:g_spd}, we define $\oplusale$ and $\odotale$ as in \cref{alem:eq:phi_mul,alem:eq:phi_sca_mul}. We define $\langle \cdot, \cdot \rangle_{\log_\alpha}$ and $g^{\log_\alpha}$ as in \cref{alem:eq:phi_innerpro,alem:eq:phi_g}.
    Then, we have the following conclusions:
    \par\leavevmode\vspace{-\baselineskip}
    \begin{enumerate}
        \item \label{alem:enum:hilbert}
        $\{\spd{n}, \oplusale,\odotale, \langle \cdot, \cdot \rangle_{\log_\alpha} \}$ is a Hilbert space over $\bbRscalar$.
        \item \label{alem:enum:mlog_riem_spd}
        $\{\spd{n}, \oplusale \}$ is an abelian Lie group.
        $g^{\log_\alpha}$ is a Riemannian metric on $\spd{n}$.
        We call this metric the \emph{Adaptive Log-Euclidean Metric (ALEM)} and denote $g^{\log_\alpha}$ by $\galem$.
        The associated Riemannian operators are as follows:
        \begin{align} 
            \label{alem:eq:dist_mlog}
            &\dalem (S_1, S_2 ) = \| \log_\alpha(S_1) - \log_\alpha(S_2) \|_\rmF,\\
            \label{alem:eq:rieexp_gmlog} 
            &\rieexp_{S_1} V = \log_\alpha^{-1}\left(\log_\alpha(S_1)+\left(\log_\alpha\right)_{*,S_1}V\right),\\
            \label{alem:eq:rielog_gmlog} 
            &\rielog_{S_1}S_2 = \left(\log_\alpha^{-1}\right)_{*,X_1}\left(\log_\alpha(S_2)-\log_\alpha(S_1)\right),\\
            \label{alem:eq:pt_mlog} 
            &\pt{S_1}{S_2}(V) = \left(\log_\alpha^{-1}\right)_{*,X_2} \circ \left(\log_\alpha\right)_{*,S_1}(V),
        \end{align}
        where $X_i = \log_\alpha(S_i) \in \sym{n}$ for $i=1,2$.
        \item \label{alem:enum:isomorphism}
        $\log_\alpha$ is an isomorphism: (a) a linear isomorphism preserving the inner product; (b) a Lie group isomorphism; (c) a Riemannian isometry.
    \end{enumerate}
\end{paristheorem}
\begin{parisremark}
    Obviously, ALEM varies with different base parameters $\alpha$ in $\log_\alpha$.
    We thus use the plural to describe our metrics.
    Besides, our metrics can be learned.
    This is why we call them adaptive metrics.
\end{parisremark}

Analogously to $\biparamLEM$, we can also define $\biparamALEM$ as the pullback of an $\orth{n}$-invariant inner product:
\begin{equation}
    \gbiparamalem= \log_\alpha^*\gbiparamaE,
\end{equation}
where we denote the $\orth{n}$-invariant inner product $\langle \cdot, \cdot \rangle^{\biparam}$ by $\gbiparamaE$.
$\gbiparamalem$ also shares the properties presented in \cref{alem:thm:mlog_spd_properties}.
Nevertheless, this part focuses on $\biparam=(1,0)$.

\subsubsection{Differentials of General Logarithms}
\label{alem:subsec:differentials}
\cref{alem:eq:rieexp_gmlog,alem:eq:rielog_gmlog,alem:eq:pt_mlog} require the differential maps of $\log_\alpha$ and $\log_\alpha^{-1}$.
This subsection introduces the concrete formulae of the associated differential maps.
\begin{parisproposition}[Differentials] \label{alem:props:diff_mgexp_mlog}
    \linktoproof{alem:props:diff_mgexp_mlog}
    For a tangent vector $V \in T_S\spd{n}$, the differential $\left(\log_\alpha\right)_{*,S} : T_S \spd{n} \rightarrow T_{\log_\alpha(S)} \sym{n}$ of $\log_\alpha$ at $S \in \spd{n}$ is given by
    \begin{equation}
        \left(\log_\alpha\right)_{*,S}(V) = Q+Q^\top + W,
    \end{equation}
    where $Q = D_U\log_\alpha(\Sigma)U^\top$,
    \begin{align*}
        D_U &= (\begin{array}{ccc}
             (\sigma_1 I_n-S)^+ V u_1 & \cdots & (\sigma_n I_n-S)^+ V u_n
        \end{array}),\\
        W &= U \diag\left(\frac{u_1^\top V u_1}{\sigma_1 \log(a_1)},\cdots,\frac{u_n^\top V u_n}{\sigma_n \log(a_n)}\right) U^\top,
    \end{align*}
    $(\cdot)^+$ is the Moore--Penrose inverse, $u_1,\cdots,u_n$ are orthonormal eigenvectors of $S$, and the associated eigenvalues are $\sigma_1,\cdots,\sigma_n$.
    
    Symmetrically, for a tangent vector $\widetilde{V} \in T_X\sym{n}$, the differential $\left(\log_\alpha^{-1}\right)_{*,X} : T_X \sym{n} \rightarrow T_{\log_\alpha^{-1}(X)} \spd{n}$ of $\log_\alpha^{-1}$ at $X \in \sym{n}$ is given by
    \begin{equation} \label{alem:eq:diff_mgexp}
        \left(\log_\alpha^{-1}\right)_{*,X}(\widetilde{V}) = \widetilde{Q}+\widetilde{Q}^\top + \widetilde{W},
    \end{equation}
    where $X = \widetilde{U} \widetilde{\Sigma}\widetilde{U}^\top$ is the eigendecomposition. Here, $D_{\widetilde{U}}$ is defined similarly and $\widetilde{Q} = D_{\widetilde{U}}\diag\left(a_1^{\widetilde{\sigma}_1},\cdots,a_n^{\widetilde{\sigma}_n}\right)\widetilde{U}^\top$. Moreover,
    \begin{equation*}
        \widetilde{W} = \widetilde{U} \diag\left(\log(a_1)a_1^{\widetilde{\sigma}_1}\widetilde{u}_1^\top \widetilde{V} \widetilde{u}_1,\cdots,\log(a_n)a_n^{\widetilde{\sigma}_n}\widetilde{u}_n^\top \widetilde{V} \widetilde{u}_n\right) \widetilde{U}^\top.
    \end{equation*}
\end{parisproposition}

\citet{arsigny2005fast} write the differential of the matrix exponential as an infinite series.
The differential of $\log_\alpha^{-1}$ can also be rewritten in this way.
\begin{parisproposition}[Differential as Infinite Series] \label{alem:props:diff_mgexp_series}
    \linktoproof{alem:props:diff_mgexp_series}
    Following the notation in \cref{alem:props:diff_mgexp_mlog}, the differential of $\log_\alpha^{-1}$ can also be formulated as
    \begin{equation} \label{alem:eq:diff_mgexp_series}
        \begin{aligned}
            &\left(\log_\alpha^{-1}\right)_{*,X}(\widetilde{V}) \\
            &= \sum_{k=1}^{\infty} \frac{1}{k !}(\sum_{l=0}^{k-1} (\widetilde{P}X)^{k-l-1} (D_{\widetilde{P}}X+\widetilde{P}\widetilde{V}) (\widetilde{P}X)^l),
        \end{aligned}
    \end{equation}
    where $\widetilde{P}=\widetilde{U}B\widetilde{U}^\top$, $B=\diag\left(\log(a_1),\cdots,\log(a_n)\right)$, $D_{\widetilde{P}}= D_{\widetilde{U}} B \widetilde{U}^\top + \widetilde{U} B D_{\widetilde{U}}^\top$.
\end{parisproposition}

When $\log_\alpha^{-1}$ is reduced to the matrix exponential, \cref{alem:eq:diff_mgexp_series} coincides with the expression in \citet[Eq.~(8)]{arsigny2005fast}, and our ALEM becomes exactly LEM.

\subsubsection{Properties of ALEM}
\label{alem:sec:properties}
Since our ALEMs are natural generalizations of LEM, they intuitively share many of its properties.
This subsubsection introduces some useful properties of our ALEMs for machine learning.

Fréchet means are important tools for SPD matrix learning \citep{harandi2018dimensionality,chakraborty2018statistical,brooks2019riemannian,chakraborty2020manifoldnorm}.
Like LEM, our ALEM also admits closed-form expressions for Fréchet means.
We present a more general result, the weighted Fréchet mean.
\begin{parisproposition}[Weighted Fréchet Means] \label{alem:props:geo_mean_spd}
    \linktoproof{alem:props:geo_mean_spd}
    For $m$ points $S_1,\ldots,S_m$ on the SPD manifold with associated weights $w_1,\ldots,w_m \in \bbRplus$ satisfying $\sum_{i=1}^m w_i>0$,
    the weighted Fréchet mean $M$ over the metric space $\{\spd{n},\dalem\}$ has a closed form
    \begin{equation} \label{alem:eq:fm_alem}
        M = \log_\alpha^{-1}\left(\sum_{i=1}^{m} \frac{w_i}{\sum_{j=1}^{m} w_j}\log_\alpha(S_i)\right).
    \end{equation}
\end{parisproposition}

Like LEM, although ALEM is not affine-invariant, it enjoys several other invariance properties.
\begin{parisproposition}[Bi-invariance] \label{alem:props:biinvariance}
    \linktoproof{alem:props:biinvariance}
    ALEM is a Lie group bi-invariant metric.
\end{parisproposition}
\begin{parisproposition}[Exponential Invariance] \label{alem:props:exp_invariance}
    \linktoproof{alem:props:exp_invariance}
    The Fréchet means under ALEM are exponential-invariant. 
    In other words, for $S_1,\ldots,S_m \in \spd{n}$ and $\beta \in \bbRscalar$,
    \begin{equation}
        (\mathrm{FM}(S_1,\ldots,S_m))^\beta = \mathrm{FM}(S_1^\beta,\ldots,S_m^\beta),
    \end{equation}
    where $\mathrm{FM}(S_1,\cdots,S_m)$ denotes the Fréchet mean of $S_1,\cdots,S_m$.
\end{parisproposition}

In addition to exponential invariance, the Fréchet mean induced by our ALEM also satisfies various properties presented in \citet{ando2004geometric}.
\begin{parisproposition} \label{alem:props:frechet_means_add_props}
    \linktoproof{alem:props:frechet_means_add_props}
    Let the following be SPD matrices:
    \begin{equation}
        A, B, C, A_0, B_0, C_0 \in \spd{n}.
    \end{equation}
    Let $\fm(A,B,C)$ denote the Fréchet mean of $A,B,C$ under ALEM.
    Then the Fréchet mean satisfies the following properties.
    \par\leavevmode\vspace{-\baselineskip}
    \begin{enumerate}
        \item \label{alem:props:u1}
        U1: permutation invariance.
        For any permutation $\pi$ of $\{A,B,C\}$,
        \begin{equation}
            \fm\bigl(\pi(A,B,C)\bigr)=\fm(A,B,C).
        \end{equation}
        \item \label{alem:props:u2} U2.
        $\fm(A,A,A) = A$.
    \end{enumerate}
    The following properties hold if $A, B, C, A_0, B_0, C_0$ commute.
    \par\leavevmode\vspace{-\baselineskip}
    \begin{enumerate}
        \item \label{alem:props:v1} V1: joint homogeneity.
        \begin{equation}
            \fm(a A, b B, c C)=(a b c )^{1 / 3} \fm(A, B, C), \qquad \forall a,b,c >0.
        \end{equation}
        
        \item V2: monotonicity.
        The map $(A, B, C) \mapsto \fm(A, B, C)$ is monotone. Specifically, if $A \geq A_0$, $B \geq B_0$, and $C \geq C_0$, then $\fm(A,B,C) \geq \fm(A_0,B_0,C_0)$ in the positive semidefinite ordering.
        \item V3: self-duality.
        $\fm(A, B, C)=\fm(A^{-1}, B^{-1}, C^{-1})^{-1}$.
        \item \label{alem:props:v4} V4: determinant identity.
        $\det \fm(A, B, C)=(\det A \cdot \det B \cdot \det C)^{1 / 3}$.
    \end{enumerate}
\end{parisproposition}

In fact, \cref{alem:props:frechet_means_add_props} holds true for any finite number of SPD matrices. 
Besides, the geodesic distance induced by ALEMs has similarity invariance.

\begin{parisproposition}[Similarity Invariance] \label{alem:props:sim_invariance}
    \linktoproof{alem:props:sim_invariance}
    The geodesic distance under ALEM is similarity invariant.
    In other words, let $R \in \so{n}$ be a rotation matrix and let $s \in \bbRplusscalar$ be a scale factor.
    Given any two SPD matrices $S_1$ and $S_2$, we have
    \begin{equation}
        \dalem(S_1,S_2)=\dalem(s^2RS_1 R^\top,s^2RS_2 R^\top).
    \end{equation}
\end{parisproposition}
Let us explain a bit more about the above three kinds of invariance.
First, among metrics on Lie groups, bi-invariant metrics are the most convenient ones \citep[Ch.~V]{sternberg1999lectures}.
Second, exponential invariance offers a fast computation for Fréchet means under exponential scaling.
Finally, similarity invariance is significant for describing frequently encountered covariance matrices \citep{arsigny2005fast}.

The above discussion focuses on the theoretical perspective.
Now, let us reconsider \cref{alem:eq:mlog} in a numerical way.

\begin{parisproposition} \label{alem:prop:rewrit_general_log}
    \linktoproof{alem:prop:rewrit_general_log}
    $\log_\alpha$ can be rewritten as
        \begin{align}
            \label{alem:eq:rw_org_mlog} \log_\alpha(S)
            &= U \log_\alpha(\Sigma) U^\top,\\
            \label{alem:eq:rw_mul_mlog}          
            &= U A \log(\Sigma) U^\top, \\
            \label{alem:eq:rw_div_mlog}          
            &= U \frac{\log(\Sigma)}{B} U^\top,
        \end{align}
    where $\frac{X}{Y}$ is the diagonal division, $B=\diag\left(\log(a_1),\cdots,\log(a_n)\right)$, and $A = \frac{I_n}{B}$.
\end{parisproposition}

Based on the above proposition, more analyses could be carried out from a numerical point of view.
First, $\log_\alpha(\cdot)$ can balance the eigenvalues of an input SPD matrix $S$ by exploiting different bases for different eigenvalues.
In Riemannian algorithms, manifold-valued features usually contain vibrant information.
We expect that by the above adaptation, manifold-valued data could be better fitted and the learning ability of algorithms could be further promoted.
\begin{parisremark}
Note that the discussion in \cref{alem:subsec:ada_rie_metric} and \cref{alem:sec:properties} can also be readily transferred to LCM, generating an adaptive version of LCM.
\end{parisremark}

\subsection{Parameter Learning}
\label{alem:sec:param_learning}
As shown in \cref{alem:thm:mlog_spd_properties}, Riemannian computations under ALEM are built upon the general matrix logarithm $\log_\alpha$ and its inverse. Accordingly, this subsection studies the backpropagation and parameter optimization of these two maps.

\subsubsection{Gradient Computation} \label{alem:subsec:gradients}

For both the general matrix logarithm and exponential, gradients are required with respect to their parameters and inputs.
Since $\log_\alpha$ involves a structured matrix decomposition, the following derivations rely heavily on structured-matrix backpropagation (BP) \citep{ionescu2015matrix}, whose key idea is invariance of the first-order differential form.
The general matrix logarithm is a special case of eigenvalue functions.
Based on the formula given by \citet{bhatia2009positive} and the matrix BP techniques presented by \citet{ionescu2015matrix}, we can obtain all the gradients in the following propositions.
\begin{parisproposition} \label{alem:props:grad_mlog}
    \linktoproof{alem:props:grad_mlog}
    Let us denote $X = \log_\alpha(S)$, where $S \in \spd{d}$ is the input SPD matrix.
    The input gradient $\nabla_S L$ is obtained by specializing the Daleckii--Krein expression in \cref{eq:ch2-symmetric-matrix-function-differential,eq:ch2-loewner-matrix} with $V=\nabla_X L$ and $f(\sigma_i)=A_{ii}\log(\sigma_i)$. The parameter gradient is
    \begin{equation}
        \nabla_{A} L
        = [U^\top (\nabla_{X} L) U] \circledast \log(\Sigma),
    \end{equation}
    where $S = U \Sigma U^\top$ is the eigendecomposition of an SPD matrix and $\sigma_1,\ldots,\sigma_d$ are the diagonal entries of $\Sigma$.
\end{parisproposition}

As the inverse map corresponding to \cref{alem:eq:rw_mul_mlog}, the general matrix exponential $\log_\alpha^{-1}$ can be rewritten as
\begin{equation}
    \label{alem:eq:mexp_rewrotten}
    \begin{aligned}
        \log_\alpha^{-1}(X)
        &= U \diag\left(a_1^{\Sigma_{11}},\cdots,a_n^{\Sigma_{nn}}\right) U^\top\\
        &= U \exp \left(\frac{\Sigma}{A} \right) U^\top,
    \end{aligned}
\end{equation}
where $X=U \Sigma U^\top \in \sym{n}$ is an eigendecomposition of $X$.
Following \cref{alem:props:grad_mlog}, we obtain the backpropagation of $\log_\alpha^{-1}$.

\begin{parisproposition} \label{alem:props:grad_mexp}
    \linktoproof{alem:props:grad_mexp}
    Let us denote $X = \log_\alpha^{-1}(S)$ with $S \in \sym{d}$.
    The input gradient $\nabla_S L$ is obtained by specializing the Daleckii--Krein expression in \cref{eq:ch2-symmetric-matrix-function-differential,eq:ch2-loewner-matrix} with $V=\nabla_X L$ and $f(\sigma_i)=e^{\frac{\sigma_i}{A_{ii}}}$. The parameter gradient is
    \begin{equation}
        \label{alem:eq:gradient_mexp_wrt_A}
        \nabla_{A} L
        = [U^\top (\nabla_{X} L) U] \circledast \left(\diag\left(a_1^{\Sigma_{11}},\cdots,a_d^{\Sigma_{dd}}\right) \frac{-\Sigma}{A^2} \right),
    \end{equation}
    where $S = U \Sigma U^\top$ is the eigendecomposition of a symmetric matrix and $\sigma_1,\ldots,\sigma_d$ are the diagonal entries of $\Sigma$.
\end{parisproposition}

\subsubsection{Parameter Updates}
\label{alem:app:subsec:geom}
\begin{table}[t]
    \centering
    \caption{Parameter learning for the general matrix logarithm and exponential.}
    \label{alem:tb:param_learning}
    \resizebox{\linewidth}{!}{%
    \begin{tabular}{cccc}
    \toprule
    \textbf{Name} & \textbf{Detail} & \textbf{Constraint} & \textbf{Method}\\
    \midrule
    RELU & Optimizing base vector $\alpha$ (\cref{alem:eq:rw_org_mlog}) & Positive & shift-ReLU $\max(\epsilon,\alpha)$\\
    MUL & Optimizing diagonal elements of $A$ (\cref{alem:eq:rw_mul_mlog}) & Unconstrained & Standard BP\\
    DIV & Optimizing diagonal elements of $B$ (\cref{alem:eq:rw_div_mlog}) & Unconstrained & Standard BP\\
    \bottomrule
    \end{tabular}}
\end{table}
The general matrix logarithm and exponential share the same base parameters. Let us focus on the former. Let the input SPD matrix $S$ have dimension $d \times d$.
Recalling \cref{alem:eq:rw_org_mlog,alem:eq:rw_mul_mlog,alem:eq:rw_div_mlog}, there are three ways to implement parameter learning.
We could learn the base vector $\alpha$ in \cref{alem:eq:rw_org_mlog}, diagonal matrix $A$ in \cref{alem:eq:rw_mul_mlog}, or diagonal matrix $B$ in \cref{alem:eq:rw_div_mlog}, respectively.

For learning $A$ in \cref{alem:eq:rw_mul_mlog} or $B$ in \cref{alem:eq:rw_div_mlog}, since the parameters (diagonal elements) lie in a Euclidean space $\bbR{d}$, the optimization can be easily integrated into the BP algorithm.
We call learning $A$ MUL and learning $B$ DIV.

For learning $\alpha$ in \cref{alem:eq:rw_org_mlog}, each element $a$ of $\alpha$ satisfies $a>0$ and $a \neq 1$. Since the equality case can be avoided by setting $a=1+\epsilon$ whenever $a=1$, where $\epsilon \in \bbRplusscalar$, it remains to enforce positivity during optimization. We consider two strategies for doing so. The first strategy applies the shift-ReLU $\max(\epsilon,a)$ to an unconstrained parameter. We call this strategy RELU. Other transformations, such as squaring the parameter, are also feasible, but we focus on RELU. The second strategy takes a geometric approach by viewing $a$ as a point on a one-dimensional SPD manifold and optimizing it using the Riemannian optimization strategy reviewed in \cref{sec:ch2-riemannian-optimization}. We call this strategy GEOM. Its RSGD update is given in the following proposition.
\begin{parisproposition} \label{alem:propos:param_by_geom}
    \linktoproof{alem:propos:param_by_geom}
    Viewing a positive scalar $a$ as a point in a one-dimensional SPD manifold, we have the following RSGD update formula.
    \begin{equation} \label{alem:eq:update_pos_scalar}
        a^{(t+1)} = a^{(t)}e^{-\gamma^{(t)} a^{(t)}  \nabla_{a^{(t)}} L},
    \end{equation}
    where $\nabla_{a^{(t)}} L$ is the Euclidean gradient of $L$ with respect to $a$ at $a^{(t)}$, $\gamma^{(t)}$ is the learning rate, and $e^{(\cdot)}$ is the natural exponential function.
\end{parisproposition}

Moreover, the following proposition shows that GEOM is equivalent to DIV.
\begin{parisproposition} \label{alem:props:geom_equivalent_div}
    \linktoproof{alem:props:geom_equivalent_div}
    For parameter learning in $\log_\alpha$, optimizing the base vector $\alpha$ by RSGD is equivalent to optimizing the divisor matrix $B$ by Euclidean stochastic gradient descent (ESGD).
\end{parisproposition}

Consequently, the three distinct update schemes are RELU, DIV, and MUL, as summarized in \cref{alem:tb:param_learning}.

\subsection{Experiments} 
\label{alem:sec:experiments}
In this section, we validate the efficacy of our approaches on multiple data sets.
Riemannian metrics are foundational to Riemannian neural networks.
Therefore, our ALEM can redesign basic blocks in Riemannian neural networks.
Beyond the main SPDNet experiments, we apply our ALEM to other Riemannian building blocks, including the LieBN framework in \cref{chapter:normalization}, Riemannian residual blocks \citep{katsman2023riemannian}, and Riemannian classifiers \citep{nguyen2023building}.
More details on data sets and experimental settings are provided in \cref{app:datasets,alem:app:experimental-details}.

\subsubsection{Applications in SPDNet}
\label{alem:sec:results_on_MLog}
In existing SPD neural networks, activation and classification layers commonly map SPD features into the logarithmic domain through the matrix logarithm \citep{huang2017riemannian,zhen2019dilated,chakraborty2020manifoldnet,chen2023riemannian}. This mapping is an isomorphism that identifies the SPD manifold under LEM with the Euclidean space $\sym{n}$. Replacing the natural matrix logarithm $\log$ with the learnable general logarithm $\log_\alpha$ allows the resulting layer to adapt the underlying geometry to the learned SPD features.

We focus on the classic SPDNet \citep{huang2017riemannian}, whose BiMap, ReEig, and LogEig layers are reviewed in \cref{app:backbone-spd}. Specifically, replacing the matrix logarithm in its LogEig layer with the learnable general matrix logarithm $\log_\alpha$ yields the \emph{Adaptive Logarithm (ALog)} layer. We compare SPDNet, SPDNetBN, and the ALog-RELU/MUL/DIV variants on HDM05, FPHA, and AFEW.

\begin{table}[t]
    \centering
    \caption{Results of ALog on the HDM05 data set. The best results are \firstresults{bold}.}
    \label{alem:tb:mlog_on_HDM05}
    \resizebox{\linewidth}{!}{%
    \begin{tabular}{c ccc ccc}
    \toprule
    \textbf{Learning rate} & \multicolumn{3}{c}{$1e^{-2}$} & \multicolumn{3}{c}{$5e^{-2}$} \\
    \midrule
    \textbf{Architecture} & \{ 93, 30\} & \{ 93, 70, 30\} & \{ 93, 70, 50, 30\} & \{ 93, 30\} & \{ 93, 70, 30\} & \{ 93, 70, 50, 30\} \\
    \midrule
    SPDNet & 62.92±0.81 & 62.87±0.60 & 63.03±0.67 & 63.89±0.73 & 64.00±0.65 & 63.72±0.61 \\
    SPDNetBN & 63.03±0.75 & 58.27±1.7 & 52.02±2.34 & 63.75±0.69 & 48.78±5.15 & 37.84±6.10\\
    \rowcolor{HilightColor} ALog-MUL & 63.52±0.75 & 63.86±0.58 & \firstresults{63.94±0.44} & 64.4±0.68 & 64.60±0.69 & 64.36±0.49 \\
    \rowcolor{HilightColor} ALog-DIV & \firstresults{63.60±0.79} & 63.93±0.52 & 63.81±0.7 & \firstresults{64.81±0.64} & \firstresults{64.84±0.65} & \firstresults{64.80±0.36} \\
    \rowcolor{HilightColor} ALog-RELU & 63.02±0.79 & \firstresults{63.94±0.64} & 63.14±0.65 & 63.97±0.75 & 64.10±0.63 & 63.78±0.46 \\
    \bottomrule
    \end{tabular}}
\end{table}

On the three data sets, the numbers of training epochs are set to 200, 500, and 100.
We verify our ALog on SPDNet with various architectures.
In addition, we test the robustness of the proposed layer against different learning rates on the HDM05 and FPHA data sets.
Generally speaking, among all three implementations, \textbf{ALog-MUL} shows the most robust performance gain and achieves consistent improvement over the vanilla matrix logarithm. We also observe that ALog-MUL is comparable to or even better than SPDNetBN, which, however, introduces substantially greater complexity than our approach.
The main reason for the superiority of our ALog against the vanilla matrix logarithm is that our ALog can adaptively respect the vibrant geometry of SPD manifolds, depending on the characteristics of data sets, while only LEM can be respected by the matrix logarithm.
The following are detailed observations and analyses.

\mypara{Results on the HDM05 data set.} The 10-fold results are presented in \cref{alem:tb:mlog_on_HDM05}, where the data split and weight initialization are randomized.
Following \citet{huang2017riemannian}, three architectures are implemented on this data set, \ie \{ 93, 30\}, \{ 93, 70, 30\}, and \{ 93, 70, 50, 30\}.
Generally speaking, endowed with ALog, SPDNet achieves consistent improvement.
Among all three implementations, RELU only brings limited improvement.
The reason might be that RELU fails to respect the innate geometry of the positive constraint.
There is another interesting observation worth mentioning.
In \citet{brooks2019riemannian}, only the result of SPDNetBN under the architecture of $\{93,30\}$ is reported on this data set.
Our experiments show that with the network going deeper, SPDNetBN tends to collapse, while our ALog layer performs robustly in all settings.

\begin{figure}[t]
  \centering
  \includegraphics[trim={0cm 0cm 0cm 0cm}, width=0.85\columnwidth]{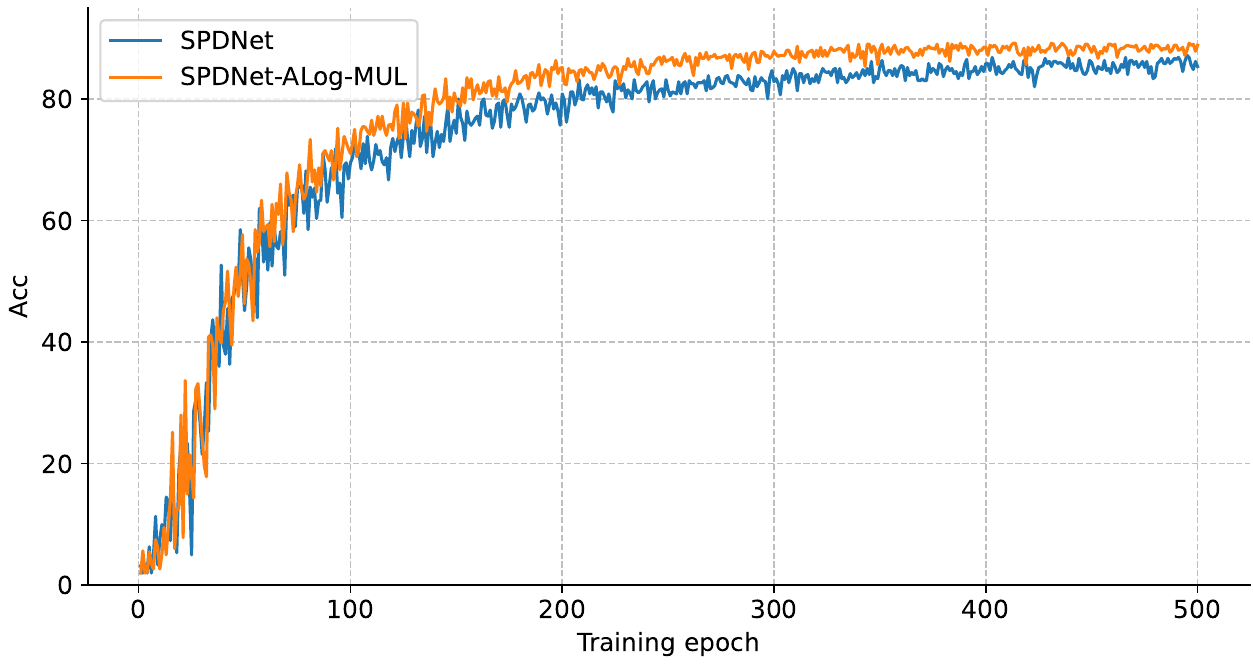}
  \caption{Accuracy curves on the FPHA data set.}
  \label{alem:fig:acc_fpha_mlog}
\end{figure}

\begin{table}[t]
    \centering
    \caption{Results of ALog on the FPHA data set.}
    \label{alem:tb:mlog_on_FPHA}
    \resizebox{0.85\columnwidth}{!}{
    \begin{tabular}{ccccc}
        \toprule
        \multirow{2}[4]{*}{\textbf{SPDNet}} & \multirow{2}[4]{*}{\textbf{SPDNetBN}} & \multicolumn{3}{c}{\textbf{ALog}} \\
    \cmidrule{3-5}          &       & \textbf{MUL} & \textbf{DIV} & \textbf{RELU} \\
        \midrule
        85.73±0.80 & 86.83±0.74 & \cellcolor{HilightColor}87.8±0.71 & \cellcolor{HilightColor}\firstresults{88.07±1.13} & \cellcolor{HilightColor}86.65±0.68 \\
        \bottomrule
    \end{tabular}
    }
\end{table}

\mypara{Results on the FPHA data set.} We validate our approach on this data set, with a learning rate of $1e^{-2}$, over 10-fold cross-validation on random initialization.
Since our experiments indicate that the vanilla SPDNet is already saturated with 1 BiMap layer, we just report the results on the architecture of $\{63,33\}$, which are presented in \cref{alem:tb:mlog_on_FPHA}. 
Although DIV performs best on this data set, it presents the largest variance.
There is an underlying nonlinear scaling mechanism in the update of DIV, which might undermine its robustness.
Without loss of generality, let us focus on a single scalar parameter $b$ in \cref{alem:eq:rw_div_mlog}.
The ultimate factor multiplied by the plain logarithm is $1/b$.
Therefore, the change of the multiplier after the update would be 
\begin{equation} \label{alem:eq:nonlinear_div}
    1/(b-\Delta)-1/b=\Delta/[(b-\Delta)b].
\end{equation}
\cref{alem:eq:nonlinear_div} will scale the original $\Delta$ to some extent.
This scaling mechanism might undermine the robustness of the ALog layer.
However, ALog-MUL achieves robust improvement and even surpasses SPDNetBN.
This again demonstrates the significance of our adaptive mechanism for Riemannian deep networks.
Finally, in terms of convergence analysis, accuracy curves with and without ALog are also reported in \cref{alem:fig:acc_fpha_mlog}.

\Needspace{15\baselineskip}
\begin{wraptable}{r}{0.60\textwidth}
    \centering
    \caption{Results of ALog on the AFEW data set.}
    \resizebox{\linewidth}{!}{
    \begin{tabular}{ccccc}
        \toprule
        \textbf{Depth} & 1 & 2 & 3 & 4 \\
        \midrule
        SPDNet & 48.53 & 46.89 & 48.24 & 47.22 \\
        SPDNetBN & 46.89 & 46.65 & 47.62 & 48.35 \\
        \rowcolor{HilightColor} ALog-MUL & \firstresults{48.57} & \firstresults{48.13} & \firstresults{49.45} & \firstresults{50.62} \\
        \rowcolor{HilightColor} ALog-DIV & 48.42 & 48.02 & 48.13 & 49.89 \\
        \rowcolor{HilightColor} ALog-RELU & 48.06 & 47.25 & 48.86 & 48.1 \\
        \bottomrule
    \end{tabular}
    }
    \label{alem:tb:mlog_on_AFEW}
\end{wraptable}

\mypara{Results on the AFEW data set.} On this data set, the learning rate is $5e^{-2}$, and we validate our method under four network architectures, \ie \{512, 100\}, \{512, 200, 100\}, \{512, 400, 200, 100\}, and \{512, 400, 300, 200, 100\}.
Note that, on this data set, SPDNetBN tends to present relatively large fluctuations in performance, so we compute the median of the last ten epochs.
On various architectures, consistent improvement can be observed when SPDNet is endowed with our ALog.
In addition, MUL performs best among all three implementations.
Another interesting observation is that SPDNetBN seems ineffective on these deep features, while our methods show consistently superior performance, most notably for ALog-MUL. This indicates that our adaptive layer maintains effectiveness when applied to covariance matrices from deep features.

\mypara{Model complexity.} Our ALog manifests the same complexity, no matter how it is optimized.
Without loss of generality, the discussion below focuses on ALog-MUL.
The extra computation and memory costs caused by the ALog layer are minor.
It only depends on the final dimension of the network.
Let us take the deepest one on the AFEW data set as an example.
Our ALog only brings 100 unconstrained scalar parameters, while SPDNetBN needs an SPD matrix parameter for each RBN layer.
The total number of parameters in RBN layers is $400^2+300^2+200^2$, which is much larger than ours.
In addition, SPDNetBN needs to store the running mean of SPD matrices in every RBN layer, while our ALog only needs to store a vector.
In terms of computation, the extra cost of our ALog is secondary as well.
The forward and backward computation of our ALog is generally the same as the plain matrix logarithm, while computation in the RBN layer is much more complex.
All in all, our ALog can consistently improve the performance of SPDNet and achieve comparable or better results than SPDNetBN with much lower computation and memory costs.
\begin{figure}[t]
  \centering
  \begin{subfigure}[t]{0.49\linewidth}
    \centering
    \includegraphics[width=\linewidth]{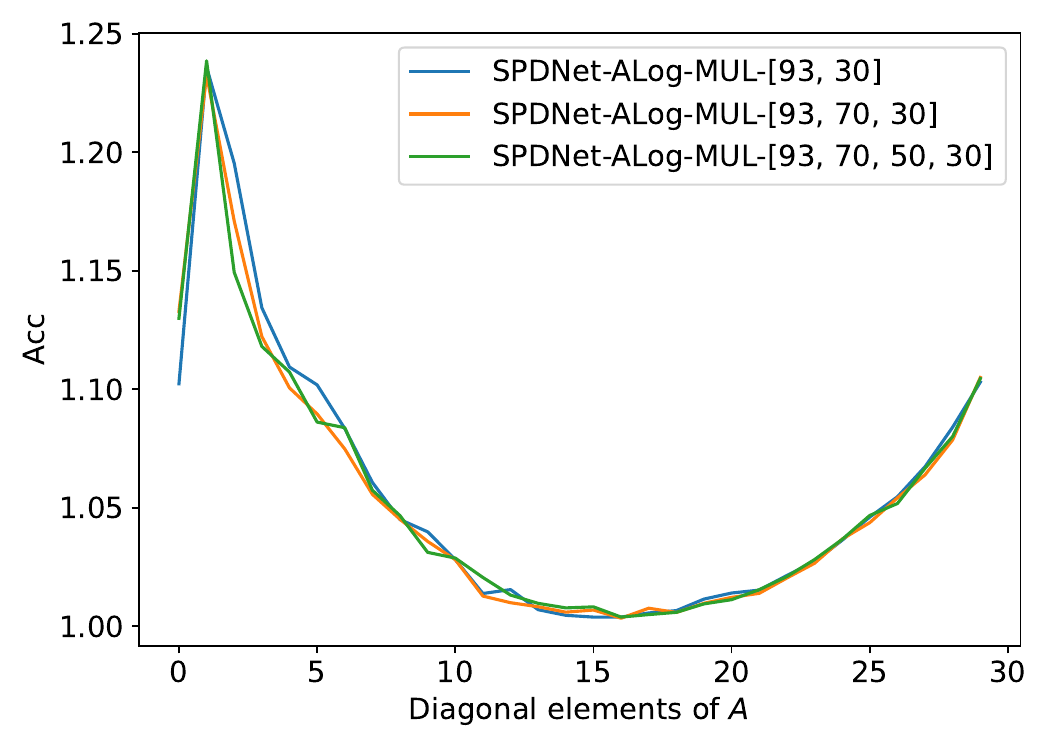}
    \caption{HDM05.}
    \label{alem:fig:vis_params_hdm05}
  \end{subfigure}
  \hfill
  \begin{subfigure}[t]{0.49\linewidth}
    \centering
    \includegraphics[width=\linewidth]{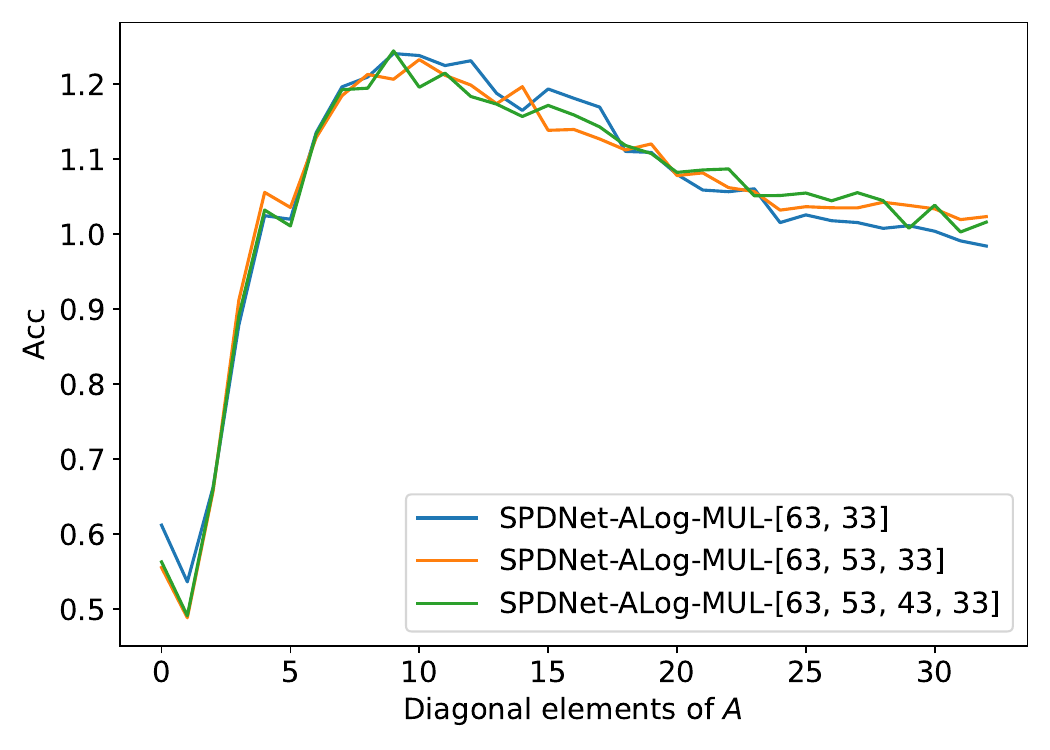}
    \caption{FPHA.}
    \label{alem:fig:vis_params_fpha}
  \end{subfigure}
  \caption{Visualization of parameters in the ALog layer on the HDM05 and FPHA data sets.}
  \label{alem:fig:vis_params}
\end{figure}

\mypara{Visualization.} We visualize the final learned parameters of the ALog layer.
Since ALog-MUL is the most robust strategy, we visualize the parameters of ALog-MUL. 
Specifically, we plot the final values of the diagonal elements of $A$ in \cref{alem:eq:rw_mul_mlog} and visualize the results in \cref{alem:fig:vis_params_hdm05,alem:fig:vis_params_fpha}. 
We observe that the distribution of the parameters is consistent within the same data set but varies between data sets. 
This indicates that our approach can capture vibrant patterns in different data sets, respecting their specific geometry.

\begin{table}[t]
    \centering
    \caption{Results of fixed bases on the HDM05 and FPHA data sets.}
    \label{alem:tab:fix_base_alog}
    \resizebox{0.99\columnwidth}{!}{
    \begin{tabular}{c ccc c}
        \toprule
        \textbf{Data Set} & \multicolumn{3}{c}{\textbf{HDM05}} & \textbf{FPHA} \\
        \textbf{Architecture} & \{93, 30\} & \{93, 70, 30\} & \{93, 70, 50, 30\} & \{63, 33\} \\
        \midrule
        SPDNet-Log2 & 63.93±0.81 & 63.54±0.50 & 63.98±0.63 & 86.65±0.67 \\
        SPDNet & 63.89±0.73 & 64.00±0.65 & 63.72±0.61 & 85.73±0.80 \\
        SPDNet-Log10 & 63.45±0.33 & 63.8±0.71 & 63.64±0.64 & 78.42±0.77 \\
        \rowcolor{HilightColor}SPDNet-ALog-MUL & \firstresults{64.4±0.68} & \firstresults{64.60±0.69} & \firstresults{64.36±0.49} & \firstresults{87.8±0.71} \\
        \bottomrule
    \end{tabular}}
\end{table}

\mypara{Ablation studies.} To further demonstrate the utility of the adaptive mechanisms in our approach, we validate the ALog layer with fixed bases.
As decimal and binary are the two most common systems, we use $\log_{10}$ and $\log_{2}$ as examples of shrinking and expanding the natural logarithm $\log$.
Specifically, we set the scalar base $\alpha$ in $\log_\alpha$ to $10$ and $2$ in \cref{alem:eq:rw_org_mlog}, respectively.
We refer to the network with binary/decimal base as SPDNet-Log2/SPDNet-Log10. 
Note that when $\alpha=e$, $\log_\alpha=\log$, and \cref{alem:eq:rw_org_mlog} reduces to the vanilla matrix logarithm. The network is then our baseline, \ie SPDNet.
We conduct 10-fold experiments on the HDM05 and FPHA data sets and set the learning rate to $5e^{-2}$ and $1e^{-2}$, respectively, while keeping the other settings consistent with previous experiments.
The results are presented in \cref{alem:tab:fix_base_alog}. 
We observe that the fixed logarithms show similar or slightly worse results than the vanilla $\log$, while our ALog shows consistent improvement.
Besides, $\log_{10}$ does not converge on the FPHA data set.
In fact, $\log_{10}$ could shrink the gradient, slowing down convergence, especially under a small learning rate. 
In contrast, our ALog maintains consistent effectiveness.
In summary, our ALog can respect vibrant geometry induced by $\log_\alpha$ and thus benefit SPD network learning.

\subsubsection{Riemannian Batch Normalization}

\begin{table}[t]
  \centering
  \caption{Comparison of RBN methods on the HDM05 data set.}
  \resizebox{0.9\linewidth}{!}{
    \begin{tabular}{ccccc}
    \toprule
    \textbf{Method} & \textbf{Geometry} & [93, 30] & [93, 70, 30] & [93, 70, 50, 30] \\
    \midrule
    None  & \na  & 63.89±0.73 & 64.00±0.65 & 63.72±0.61 \\
    SPDNetBN   & AIM   & 63.75±0.69 & 48.78±5.15 & 37.84±6.10 \\
    SPDBN & AIM   & 64.33±0.89 & 64.31±0.92 & 63.62±1.21 \\
    LieBN-LEM & LEM   & 63.67±0.85 & 65.77±0.89 & 65.34±0.83 \\
    \midrule
    \rowcolor{HilightColor}LieBN-ALEM & ALEM  & \firstresults{65.24±0.71} & \firstresults{70.11±0.96} & \firstresults{68.86±0.72} \\
    \bottomrule
    \end{tabular}
    }
  \label{alem:tab:results_rbn}
\end{table}

The LieBN framework, its normalization guarantee, and its SPD manifestations are developed in \cref{chapter:normalization}.
As shown in \cref{alem:thm:mlog_spd_properties}, $\{\spd{n}, \oplusale \}$ forms a Lie group.
Besides, \cref{alem:props:biinvariance} demonstrates that ALEM is bi-invariant with respect to this group structure.
Therefore, LieBN under ALEM can also normalize Riemannian sample statistics.
Following the LieBN algorithm and its SPD specialization, we implement LieBN under ALEM, denoted LieBN-ALEM. We compare it against AIM-based SPDNetBN \citep{brooks2019riemannian} and SPDBN \citep{kobler2022controlling}, and against LieBN under LEM.

Following previous work \citep{kobler2022controlling,brooks2019riemannian}, we adopt the SPDNet backbone. 
\cref{alem:tab:results_rbn} presents the 10-fold average results on the HDM05 data set under different network architectures.
Our LieBN-ALEM achieves the best performance among the RBN methods.
In particular, the AIM-based SPDNetBN brings worse performance under deeper architectures.
In contrast, our LieBN-ALEM can consistently improve the performance across different architectures.
Besides, compared with LieBN-LEM, our LieBN-ALEM shows better performance, demonstrating the effectiveness of our ALEM.

\subsubsection{Riemannian Residual Blocks}

\begin{wraptable}{r}{0.55\textwidth}
  \centering
  \caption{Experiments on RResNet under different geometries.}
  \resizebox{\linewidth}{!}{
    \begin{tabular}{ccc}
    \toprule
    \textbf{Method} & \textbf{HDM05} & \textbf{NTU60} \\
    \midrule
    SPDNet & 63.89±0.73 & 45.90±1.11 \\
    RResNet-AIM & 63.82±0.58 & 45.22 ± 1.23 \\
    RResNet-LEM & 66.51±0.93 & 48.73±0.60 \\
    \midrule
    \rowcolor{HilightColor}RResNet-ALEM & \firstresults{69.03±1.06} & \firstresults{57.09±0.59} \\
    \bottomrule
    \end{tabular}%
  }
  \label{alem:tab:results_rresnet_hdm05_ntu60}%
\end{wraptable}%

The general RResNet construction and SPD residual block are reviewed in \cref{app:backbone-others}. Since its Riemannian exponential is metric-dependent, the Riemannian residual block under ALEM is obtained by substituting \cref{alem:eq:rieexp_gmlog} into that construction.
The required backpropagation of $\log_\alpha^{-1}$ is derived in \cref{alem:props:grad_mexp}.

Following \citet{katsman2023riemannian}, we compare RResNet under different geometries on the HDM05 and NTU60 data sets.
\cref{alem:tab:results_rresnet_hdm05_ntu60} reports the 10-fold and 5-fold average results on these data sets.
Compared with the vanilla SPDNet, RResNet-AIM brings little improvement, while LEM and ALEM show much better performance.
In particular, the ALEM-based RResNet can bring a clear performance improvement, underscoring the effectiveness of our ALEM.

\subsubsection{Riemannian Classifiers}

\begin{wraptable}{r}{0.55\textwidth}
  \centering
  \caption{Comparison of Gyro MLRs on the NTU60 data set.}
  \resizebox{\linewidth}{!}{
    \begin{tabular}{ccc}
    \toprule
    \textbf{Learning rate} & $1e^{-2}$ & $5e^{-2}$ \\
    \midrule
    GyroMLR-AIM & 54.28±0.47 & 41.41±0.71 \\
    GyroMLR-LCM & 42.68±0.88 & 42.06±0.49 \\
    GyroMLR-LEM & 53.22±0.47 & 39.62±1.30 \\
    \midrule
    \rowcolor{HilightColor}GyroMLR-ALEM & \firstresults{56.21±0.39} & \firstresults{51.65±0.44} \\
    \bottomrule
    \end{tabular}%
  }
  \label{alem:tab:gyro_mlr_ntu60}%
\end{wraptable}%

Euclidean MLR, which consists of FC and softmax, has become a standard classification block in Euclidean neural networks.
Inspired by this, \citet{nguyen2023building} extended Euclidean MLR to the SPD manifold using gyrostructures \citep{nguyen2022gyro} for intrinsic classification, referred to as gyro MLR.
Three gyro MLRs under LCM, AIM, and LEM were introduced by \citet{nguyen2023building}.
Following the logic in \citet[Sec.~2.4.2]{nguyen2023building}, we can obtain the gyro MLR under ALEM.

\begin{paristheorem}[Gyro MLR] \label{alem:thm:gyro_mlr_alem}
    \linktoproof{alem:thm:gyro_mlr_alem}
    Given an SPD feature $S \in \spd{n}$ and $C$ classes, the SPD gyro MLR under ALEM computes the multinomial probability of each class:
    \begin{equation}
    \label{alem:eq:gyro_mlr_alem_start}
    \begin{aligned}
        p(y=k \mid S) \propto
        \exp \left [ \left\langle \log_\alpha(S)-\log_\alpha(P_k), \left(\log_\alpha\right)_{*,P_k}(\tilde{A}_{k}) \right\rangle \right ],
    \end{aligned}
    \end{equation}
    where $k \in\{1, \ldots, C\}$, $P_k \in \spd{n}$, and $\tilde{A}_k \in T_{P_k} \spd{n}$.
\end{paristheorem}

Following \cref{chapter:rmlr}, we set $\tilde{A}_{k}=\pt{I_n}{P_k}(A_k)$ with $A_k \in T_{I_n} \spd{n}$. Therefore, the RHS of \cref{alem:eq:gyro_mlr_alem_start} becomes
\begin{equation}
        \exp \left [ \left\langle \log_\alpha(S)-\log_\alpha(P_k), \left(\log_\alpha\right)_{*,I_n}(A_k) \right\rangle \right ].
\end{equation}
As $\left(\log_\alpha\right)_{*,I_n}(A_k) \in T_\bbzero \sym{n} \cong \sym{n}$, we view $\left(\log_\alpha\right)_{*,I_n}(A_k)$ as the parameter.

We use SPDNet as the backbone. We compare Gyro MLR under our ALEM with those under LEM, LCM, and AIM on the NTU60 data set.
\cref{alem:tab:gyro_mlr_ntu60} presents the 5-fold average results under different learning rates.
Our ALEM outperforms the other metrics within the gyro MLR framework.
When the learning rate is $5e^{-2}$, our GyroMLR-ALEM shows a larger performance advantage, especially compared with GyroMLR-LEM.
These results demonstrate that Riemannian networks can benefit from the adaptivity of our ALEM.

\section{Product Cholesky Metrics}
\label{sec:pcm-bwcm}

\subsection{Introduction}
\label{pcm:sec:introduction}

Whereas \cref{sec:alem} introduces adaptive flexibility through pullback Euclidean geometry, this part pursues a complementary route centered on computational efficiency and numerical stability. LCM provides the natural bridge between these routes. It combines simple closed-form Riemannian operators with the fast and stable computation of the Cholesky decomposition \citep{lin2019riemannian}. LCM is induced by the Cholesky decomposition from a Riemannian metric on the Cholesky manifold, which is the space of lower triangular matrices with positive diagonal entries. We refer to this source metric as the diagonal log metric. Its interpretation as a pullback Euclidean metric was established in \cref{alem:thm:rethk_lem_lcm}. We reveal a simple \emph{product structure} underlying the diagonal log metric: a Euclidean metric on the strictly lower triangular part together with $n$ copies of a Riemannian metric on $\bbRplusscalar$ for the diagonal part. This observation opens up a principled design space, as any metric on $\bbRplusscalar$ can induce a metric on the Cholesky manifold, and further yield a corresponding metric on the SPD manifold via the Cholesky decomposition.

Building on this product structure, we introduce two Cholesky metrics, the diagonal power metric and the diagonal Bures--Wasserstein metric, which induce two SPD metrics via the Cholesky decomposition: the \emph{Power-Cholesky Metric (PCM)} and \emph{Bures--Wasserstein--Cholesky Metric (BWCM)}. Unlike the diagonal logarithm in LCM, our metrics rely on diagonal powers, improving numerical stability by avoiding exponentiation and logarithms. We further define in the Cholesky factors of SPD matrices a diagonal power deformation, which continuously connects existing and new metrics. As power $\theta \to 0$, the deformed metric converges to LCM, while at $\theta=1$ it recovers our proposed metrics, thereby offering a tunable trade-off. All proposed SPD metrics admit closed-form Riemannian operators, including geodesics, logarithmic and exponential maps, parallel transport, as well as gyrovector operators \citep{ungar2022analytic}, which extend vector addition and scalar multiplication into manifolds. These operators make our metrics directly applicable to SPD neural networks. In particular, by substituting these operators into the Riemannian MLR formulation developed in \cref{chapter:rmlr} and residual blocks \citep{katsman2024riemannian}, we directly obtain SPD MLR classifiers and residual blocks under our metrics. We validate our metrics with experiments on SPD neural networks, numerical stability analyses, and tensor interpolation, showing the effectiveness, efficiency, and robustness of our metrics. In summary, our main \textbf{contributions} are:
\begin{enumerate}
    \item 
    Revealing the underlying simple product structure in the Cholesky manifold;
    \item 
    Proposing two Cholesky metrics and their SPD counterparts, PCM and BWCM, which admit fast and stable closed-form operators;
    \item 
    Developing SPD classifiers and residual blocks based on our metrics for SPD neural networks.\footnote{The code is available at \url{https://github.com/GitZH-Chen/PCM_BWCM}.}
\end{enumerate}

\mypara{Outline.} \cref{pcm:sec:preliminaries} recalls the Cholesky geometry used in this part. \cref{pcm:sec:cholesky-product-geometries} develops the Cholesky product geometries, and \cref{pcm:sec:spd-geometries} derives their SPD counterparts. Their applications to SPD neural networks and the experimental evaluation are presented in \cref{pcm:sec:applications,pcm:sec:experiments}. Proofs are deferred to \cref{app:pcm-proofs}.

\subsection{Preliminaries}
\label{pcm:sec:preliminaries}

Pullback metrics are defined in \cref{def:ch2-riemannian-isometry}, while the commonly used SPD geometries are summarized in \cref{tab:ch2-spd-lie-operators,tab:ch2-spd-bwm-pem-operators}. We further recall the Cholesky geometry. The Euclidean space of $n \times n$ lower triangular matrices is denoted $\trilspace{n}$. Its open subset, whose diagonal elements are all positive, is denoted by $\chospace{n}$. The Cholesky space $\chospace{n}$ forms a submanifold of $\trilspace{n}$ \citep{lin2019riemannian}. For a Cholesky matrix $L \in \chospace{n}$ and tangent vectors $X, Y \in T_L\chospace{n}$, the Riemannian metric on the Cholesky manifold, referred to as the diagonal log metric, is
\begin{equation}
    \label{pcm:eq:dlm}
    \gDL _{L}(X,Y)= \langle \trilX, \trilY \rangle + \langle \bbL^{-1} \bbX, \bbL^{-1}\bbY \rangle.
\end{equation}
Here, $\trilX$ and $\trilY$ are the strictly lower triangular parts of $X$ and $Y$, while $\bbL$, $\bbX$, and $\bbY$ are the diagonal matrices formed from their diagonal entries. LCM is the pullback metric of $\gDL$ by the Cholesky decomposition. As shown in \cref{alem:thm:rethk_lem_lcm}, the diagonal log metric is the pullback metric, by the diagonal log map, of the Euclidean metric over $\trilspace{n}$, which rationalizes our nomenclature.

\subsection{Product Geometries on the Cholesky}
\label{pcm:sec:cholesky-product-geometries}
We first unveil the product structure beneath the existing diagonal log metric on the Cholesky manifold. Based on this, we propose two novel Cholesky metrics.

\subsubsection{Disentangling the Cholesky Geometry}
\label{pcm:sec:rethk_chol_m}

We denote the space of $n \times n$ diagonal matrices with positive diagonal elements by $\bbDplus{n}$ and the space of $n \times n$ strictly lower triangular matrices by $\LTzero{n}$. Then, $\LTzero{n}$ is a Euclidean space, and $\bbDplus{n} \cong (\bbRplusscalar)^n$ is an open submanifold of $\bbR{n}$. Recalling \cref{pcm:eq:dlm}, it is defined separately on $\LTzero{n}$ and $\bbDplus{n}$. Besides, $\bbDplus{n}$ can be identified as the product of $n$ copies of $\bbRplusscalar$. The above discussion implies a product structure. We denote the standard Euclidean metric over $\LTzero{n}$ by $\gE$ and define the Riemannian metric on $\bbRplusscalar$ as
\begin{equation}
    \label{pcm:eq:g_rplus}
    \gRplus _{p}(v,w)=p^{-2}vw, \quad \forall p \in \bbRplusscalar \text{ and } v,w \in T_p\bbRplusscalar.
\end{equation}
Then, $\chospace{n}$ is the product manifold of $\LTzero{n}$ and $n$ copies of $\bbRplusscalar$:
\begin{equation}
\label{pcm:eq:prodct_cm} 
    \{ \chospace{n},\gDL \} = \{\LTzero{n},\gE \} \times \overbrace {\{\bbRplusscalar, \gRplus\} \times \cdots \times \{\bbRplusscalar, \gRplus\}}^n.
\end{equation}

\subsubsection{Product Geometries on the Cholesky}
\label{pcm:subsec:prod_metrics_chol}

The following definition characterizes the underlying product structure in \cref{pcm:eq:prodct_cm}.

\begin{parisdefinition}[Product geometries] \label{pcm:def:product_geometries}
    Suppose $g^{\mathrm{LT}^0}$ is a Euclidean inner product on $\LTzero{n}$ and $\{ g^i \}_{i=1}^n$ are Riemannian metrics on $\bbRplusscalar$. Then, the weighted product metric $g$ on $\chospace{n}$ is defined as $g_{L}(X,Y) = g^{\mathrm{LT}^0}(\trilX,\trilY) +  \sum_{i=1}^{n} \alpha_i g^i _{L_{ii}} \left( X_{ii}, Y_{ii}\right)$, with $L \in \chospace{n}$, $X,Y \in T_L\chospace{n}$, and $\alpha_i > 0$. Here, $\trilX$ and $\trilY$ are the strictly lower triangular parts of $X$ and $Y$, while $L_{ii}$, $X_{ii}$, and $Y_{ii}$ are the $i$-th diagonal elements.
\end{parisdefinition}

For simplicity, we focus on the case where $g^{\mathrm{LT}^0}=\gE$ is the standard Euclidean metric, all $\alpha_i$ are equal to 1, and all $g^i$ are identical. Since $\bbRplusscalar$ can be viewed as a one-dimensional SPD manifold $\spd{1}$, the Riemannian metrics reviewed in \cref{sec:ch2-spd-manifolds} can be immediately used to build Riemannian metrics on the Cholesky manifold. We additionally consider the Generalized Bures--Wasserstein Metric (GBWM), which is the pullback of BWM by $S\mapsto M^{-\frac{1}{2}}SM^{-\frac{1}{2}}$ for $S, M \in \spd{n}$ \citep{han2023learning}. For clarity, we denote PEM and GBWM by $\PEM$ and $\GBWM$, respectively. Simple computations show that AIM, LEM, and LCM coincide with \cref{pcm:eq:g_rplus} on $\spd{1}$. Therefore, these metrics reduce to three classes on $\spd{1}$: (1) LEM, LCM, or AIM; (2) $\PEM$; (3) $\GBWM$ or BWM. When the metric on $\bbRplusscalar$ is AIM (LEM or LCM), the resulting product metric on the Cholesky manifold is the diagonal log metric, and the pullback SPD metric via the Cholesky decomposition is exactly LCM.

Inspired by the above analysis, we obtain two new metrics on the Cholesky manifold by setting each $g^i$ in \cref{pcm:def:product_geometries} to $\PEM$ and $\GBWM$, termed the \emph{Diagonal Power Metric} ($\defDPM$) and \emph{Diagonal Bures--Wasserstein Metric} ($\DGBWM$ with $\bbM \in \bbDplus{n}$), respectively. By product geometries \citep{lee2018introduction}, we can obtain closed-form expressions for their Riemannian operators, such as the geodesic, logarithmic map, parallel transport, and weighted Fréchet mean. These operators are particularly important for building concrete learning algorithms \citep{yuan2012local,lezcano2019trivializations,brooks2019riemannian,lopez2021vector}.

\begin{paristheorem}[$\defDPM$]
    \label{pcm:thm:dpem}
    \linktoproof{pcm:thm:dpem}
    Let $L,K \in \chospace{n}$ and $X,Y \in T_L\chospace{n}$, and let $\{L_i \in \chospace{n} \}_{i=1}^N$ have weights $\{w_i\}_{i=1}^N$ satisfying $w_i>0$ for all $i$ and $\sum_{i=1}^N w_i=1$.
    Then, the Riemannian operators under $\defDPM$ with $\theta \neq 0$ are
    \begin{align}
         \gdefDE_{L} (X,Y) &= \langle \trilX, \trilY \rangle + \langle \bbL^{\theta-1} \bbX, \bbL^{\theta-1}\bbY \rangle, \\
         \geodesic{L}{X}(t) &= \trilL + t\trilX + \bbL \left( I_n + t\theta\bbL^{-1} \bbX \right)^{\frac{1}{\theta}}, \\
         \rielog_{L}(K) &= \trilK-\trilL + \frac{1}{\theta} \bbL \left[ \left( \bbL^{-1}\bbK \right)^{\theta}- I_n \right], \\
         \pt{L}{K}(X) &= \trilX + \left( \bbL^{-1} \bbK \right)^{1-\theta}\bbX,\\
         \dist^2(L,K) &= \fnorm{\trilK - \trilL }^2 + \frac{1}{\theta^2} \fnorm{\bbK^\theta-\bbL^\theta}^2, \\
         \wfm(\{w_i\},\{L_i \}) &= \sum \nolimits_{i}w_i \trilLi + \left ( \sum \nolimits_i w_i \bbL_{i}^\theta \right)^\frac{1}{\theta},
    \end{align}
where $\fnorm{\cdot}$ is the Frobenius norm. $\bbX$, $\bbY$, $\bbL$, $\bbK$, and $\bbL_{i}$ are diagonal matrices with diagonal elements from $X$, $Y$, $L$, $K$, and $L_i$. $\geodesic{L}{X}(t)$ denotes the geodesic starting at $L$ with initial velocity $X$. $\pt{L}{K}(\cdot)$ is the parallel transport along the geodesic connecting $L$ and $K$. $\rielog$, $\dist$, and $\wfm$ are the Riemannian logarithm, geodesic distance, and weighted Fréchet mean, respectively. Note that $\geodesic{L}{X}(t)$ is locally defined in $\{t \in \bbRscalar | \bbL + t\theta \bbX \in \bbDplus{n}\}$.
\end{paristheorem}

\begin{paristheorem}[$\DGBWM$]
\label{pcm:thm:dgbwm}
\linktoproof{pcm:thm:dgbwm}
Following the notation in \cref{pcm:thm:dpem}, the Riemannian operators under $\DGBWM$ with $\bbM \in \bbDplus{n}$ are 
\begin{align}
    \gDGBW_{L} (X,Y) &= \langle \trilX, \trilY \rangle +\frac{1}{4} \langle \bbL^{-1} \bbX,\bbM^{-1} \bbY \rangle, \\
    \geodesic{L}{X}(t) &=\trilL + t\trilX + \bbL \left( I_n + t\frac{1}{2} \bbL^{-1} \bbX \right)^2,\\
    \rielog_{L}(K) &= \trilK-\trilL + 2\bbL \left[ \left(\bbL^{-1} \bbK \right)^{\frac{1}{2}}-I_n \right], \\
    \pt{L}{K}(X) &=\trilX + \left( \bbL^{-1} \bbK  \right)^{\frac{1}{2}}\bbX,\\
    \dist^2(L,K) &= \fnorm{\trilK - \trilL}^2 + \fnorm{ \bbM^{-\frac{1}{2}} \left( \bbK^\frac{1}{2}-\bbL^\frac{1}{2} \right)}^2, \\
    \wfm(\{w_i\},\{L_i\}) &= \sum_{i}w_i \trilLi + \left ( \sum_i w_i \bbL_{i}^\frac{1}{2} \right)^2,
\end{align}
where the geodesic $\geodesic{L}{X}(t)$ is locally defined in $\{t \in \bbRscalar \mid \bbL + \frac{t}{2} \bbX \in \bbDplus{n}\}$. When $\bbM=I_n$ in $\DGBWM$, the resulting metric is denoted by DBWM.
\end{paristheorem}
On the SPD manifold, GBWM is locally AIM \citep{han2023learning}.
Similarly, on the Cholesky manifold, our $\DGBWM$ is locally the diagonal log metric at $L \in \chospace{n}$: $g_L^{\bbL\text{-DBW}}(X,Y) = \langle \trilX, \trilY \rangle + \frac{1}{4} \langle \bbL^{-1} \bbX, \bbL^{-1} \bbY \rangle$. GBWM on the SPD manifold generally has no closed-form expression for the Fréchet mean \citep{bhatia2019bures}. Moreover, the closed-form expression of parallel transport under BWM is known only if two SPD matrices commute \citep{thanwerdas2023n}. In contrast, all these operators have closed-form expressions under $\DGBWM$ on $\chospace{n}$.

\subsubsection{Deformed Cholesky Metrics}
\label{pcm:subsec:def_metric_cholesky}

On SPD manifolds, metrics deformed by the matrix power can interpolate between a given metric and an LEM-like metric \citep[Sec.~3.1]{thanwerdas2022geometry}. Inspired by this, we define a diagonal power deformation on the Cholesky manifold. For $\theta\ne0$, we denote the diagonal power by $\dpow_{\theta}: \bbDplus{n} \ni \bbP  \longmapsto  \bbP^\theta \in \bbDplus{n}$. We will show how our proposed metric is connected to the existing diagonal log metric by $\dpow_{\theta}$.

\begin{parisdefinition} \label{pcm:def:dpdm}
    Let $\{\chospace{n}, g\} = \{\LTzero{n}, \gE\} \times \{\bbDplus{n}, \tilde{g}\}$ be a product metric and $\theta\ne0$. We define the diagonal-power-deformed metric of $g$ as $\{\chospace{n}, g^\theta \} = \{\LTzero{n}, \gE\} \times \{\bbDplus{n}, \frac{1}{\theta^2}\dpow_{\theta}^*\tilde{g}\}$.
\end{parisdefinition}

The following lemma shows that $g^\theta$ in \cref{pcm:def:dpdm} converges to a diagonal-log-like metric as $\theta \rightarrow 0$.

\begin{parislemma}
\label{pcm:lem:dpdm_exp_and_limits}
\linktoproof{pcm:lem:dpdm_exp_and_limits}
Given $L \in \chospace{n}$ and $X,Y \in T_L\chospace{n}$, $g^\theta$ in \cref{pcm:def:dpdm} satisfies
\begin{equation}
    \label{pcm:eq:metric_dpdm_expression}
    g^\theta_L(X,Y) 
    = \langle \trilX,\trilY \rangle + \tilde{g}_{\bbL^\theta}\left( \bbL^{\theta-1} \bbX, \bbL^{\theta-1} \bbY \right) \underset{\theta \rightarrow 0}{\longrightarrow} \langle \trilX,\trilY \rangle + \tilde{g}_{I_n}(\bbL^{-1} \bbX, \bbL^{-1} \bbY).
\end{equation}   
\end{parislemma}

Now, we discuss the deformation of the diagonal log metric, $\defDPM$, and $\DGBWM$. First, \cref{pcm:eq:metric_dpdm_expression} indicates that the diagonal-power-deformed metric of the diagonal log metric is itself. Second, $\defDPM$ is the diagonal-power-deformed metric of the Euclidean metric on the Cholesky manifold. Besides, $\defDPM$ interpolates between the diagonal log metric ($\theta \to 0$) and the Euclidean metric ($\theta=1$). Third, the diagonal-power-deformed metric of $\DGBWM$, referred to as $\defDGBWM$, is
\begin{equation}
    \gdefDGBW_L (X,Y) = \langle \trilX, \trilY \rangle +\frac{1}{4} \langle \bbL^{\theta-2} \bbX,\bbM^{-1} \bbY \rangle.
\end{equation}
When $\bbM=I_n$, the deformed metric of DBWM, \ie $\defDBWM$, tends to be a scaled diagonal log metric as $\theta \to 0$:
\begin{equation}
    \lim_{\theta\to0}\gdefDBW_L (X,Y) = \langle \trilX, \trilY \rangle +\frac{1}{4} \langle \bbL^{-1} \bbX,\bbL^{-1} \bbY \rangle.
\end{equation}
As $\defDGBWM$ is the pullback metric by diagonal power and scaled by a constant, the Riemannian operators also have closed-form expressions, which are discussed in \cref{pcm:app:subsubsec:riemannian-operators}.

\subsubsection{Algebraic Structures}

As reviewed in \cref{sec:ch2-algebraic-structures}, \cref{eq:ch2-riem-gyro-addition,eq:ch2-riem-gyro-scalar} define gyroaddition and scalar gyromultiplication on a Riemannian manifold. The gyro operations under the diagonal log metric reduce to vector-space operations, as the metric is induced by the Euclidean metric over the lower triangular matrices. This subsection studies the gyro-structures over $\defDPM$ and $\defDGBWM$. Let the identity matrix be the origin and $\mathcal{C}$ be $\defDPM$ or $\defDGBWM$. We have the following.

\begin{parislemma}[Gyro-structures]
    \label{pcm:lem:gyro_cholesky} 
    \linktoproof{pcm:lem:gyro_cholesky}
    For $L,K \in \chospace{n}$ and $t \in \bbRscalar$, the gyro operations are 
    \begin{align}
        L \oplusChol K &= \trilL + \trilK + \left(\bbL^\beta + \bbK^\beta -I_n \right)^\frac{1}{\beta}, \\
        t \odotChol L &= t\trilL + \left(t\bbL^\beta + (1-t)I_n \right)^\frac{1}{\beta},
    \end{align}
    where $\beta=\theta$ for $\defDPM$, and $\beta=\nicefrac{\theta}{2}$ for $\defDGBWM$. $\oplusChol$ requires $L$ and $K$ to satisfy $\bbL^\beta+\bbK^\beta-I_n \in \bbDplus{n}$, while $\odotChol$ requires $(1-t)I_n+t\bbL^\beta \in \bbDplus{n}$.
\end{parislemma}

These gyro operations are defined only under the assumptions in \cref{pcm:lem:gyro_cholesky}, which arise from the locally defined Riemannian exponential map. Throughout, we impose these assumptions implicitly.

\begin{paristheorem}
    \label{pcm:thm:gyro_spaces_defdem}
    \linktoproof{pcm:thm:gyro_spaces_defdem}
    When the gyro operations are well defined under the conditions in \cref{pcm:lem:gyro_cholesky}, $\{\chospace{n}, \oplusChol \}$ satisfies all the axioms of gyrocommutative gyrogroups (\cref{def:ch2-gyrogroup,def:ch2-gyrocommutative}), and $\{\chospace{n}, \oplusChol, \odotChol \}$ satisfies all the axioms of gyrovector spaces (\cref{def:ch2-gyrovector-space}).
\end{paristheorem}

\begin{pariscorollary}\label{pcm:cor:gyro_inverse}
    The identity element of $\{\chospace{n}, \oplusChol \}$ is the identity matrix, \ie $\forall L \in \chospace{n}, I_n \oplusChol L = L$. The inverses are $\ominusChol L = -1 \odotChol L = -\trilL + \left(2I_n - \bbL^\beta \right)^\frac{1}{\beta}$, for $L \in \{L \in \chospace{n} \mid 2I_n - \bbL^\beta \in \bbDplus{n} \}$.
\end{pariscorollary}

\begin{parisremark}
    The gyrostructures on the Grassmannian have shown success in building Riemannian algorithms \citep{nguyen2022gyro,nguyen2023building}. Like our gyrostructure, the gyrostructures of the Grassmannian also require some assumptions for well-definedness \citep[Sec.~3.2]{nguyen2022gyro}. We therefore examine the well-definedness of the gyrostructures under our metrics. In practice, such positivity constraints can be remedied by numerical techniques. Taking $2I_n - \bbL^\beta \in \bbDplus{n}$ as an example, one can use $d_i \leftarrow \max(d_i, \varepsilon)$ for each diagonal element $d_i$ with a small constant $\varepsilon >0$.
\end{parisremark}

\subsubsection{Numerical Advantages over Diagonal Log Metric}
\label{pcm:subsec:num_adv}

\begin{table}[t]
    \centering
    \caption{Riemannian and gyro operators of different metrics on the Cholesky manifold. For the diagonal log metric, $\log(\cdot)$ and $\exp(\cdot)$ are diagonal logarithm and exponentiation.}
    \label{pcm:tab:riem_gyro_operators_cholesky}
    \resizebox{0.99\linewidth}{!}{
    \begin{tabular}{c ccc}
        \toprule
        \textbf{Operators} & \textbf{Diagonal Log Metric} & $\boldsymbol{\defDPM}$ & $\boldsymbol{\defDGBWM}$\\
        \midrule
        $g_{L}(X,Y)$ 
        & $\langle \trilX, \trilY \rangle + \langle \bbL^{-1} \bbX, \bbL^{-1}\bbY \rangle$
        & $\langle \trilX , \trilY \rangle + \langle \bbL^{\theta-1} \bbX, \bbL^{\theta-1}\bbY \rangle$ 
        & $\langle \trilX, \trilY \rangle +\frac{1}{4} \langle \bbL^{\theta-2} \bbX,\bbM^{-1} \bbY \rangle$
        \\
        \midrule
        $\geodesic{L}{X}(t)$ 
        & $\trilL + t\trilX + \bbL \exp (t \bbL^{-1} \bbX )$ 
        & $\trilL + t\trilX + \bbL \left( I_n + t\theta\bbL^{-1} \bbX \right)^{\frac{1}{\theta}}$
        & $\trilL + t\trilX + \bbL \left( I_n + t\frac{\theta}{2} \bbL^{-1} \bbX \right)^{\frac{2}{\theta}}$ \\
        \midrule
        $\rielog_{L}(K)$ 
        & $\trilK  - \trilL + \bbL \log (\bbL^{-1} \bbK )$  
        & $\trilK-\trilL + \frac{1}{\theta} \bbL \left[ \left( \bbL^{-1}\bbK \right)^{\theta}- I_n \right]$
        & $\trilK-\trilL + \frac{2}{\theta}\bbL \left[ \left(\bbL^{-1} \bbK \right)^{\frac{\theta}{2}}-I_n \right]$\\
        \midrule
        $\pt{L}{K}(X)$ 
        & $\trilX + (\bbL^{-1} \bbK) \bbX $ 
        & $\trilX + \left( \bbL^{-1} \bbK \right)^{1-\theta}\bbX$
        & $\trilX + \left( \bbL^{-1} \bbK  \right)^{1-\frac{\theta}{2}}\bbX$ \\
        \midrule
        $\dist^2(L,K)$ 
        & $\fnorm{\trilK - \trilL }^2 + \fnorm{\log(\bbK) - \log(\bbL)}^2$ 
        & $\fnorm{\trilK - \trilL }^2 + \frac{1}{\theta^2} \fnorm{\bbK^\theta-\bbL^\theta}^2$
        & $\fnorm{\trilK - \trilL}^2 + \frac{1}{\theta^2} \fnorm{ \bbM^{-\frac{1}{2}} \left( \bbK^\frac{\theta}{2}-\bbL^\frac{\theta}{2} \right)}^2$\\
        \midrule
        $\wfm(\{w_i\},\{L_i\})$ 
        &  $\sum_{i}w_i \trilLi + \exp \left ( \sum_i w_i \log(\bbL_{i}) \right)$
        & $\sum_{i}w_i \trilLi + \left ( \sum_i w_i \bbL_{i}^\theta \right)^\frac{1}{\theta}$
        & $\sum_{i}w_i \trilLi + \left ( \sum_i w_i \bbL_{i}^\frac{\theta}{2} \right)^\frac{2}{\theta}$\\
        \midrule
        $L \oplus K$
        & $\trilL + \trilK + \bbL\bbK$
        & $\trilL + \trilK + \left(\bbL^\theta + \bbK^\theta -I_n \right)^\frac{1}{\theta}$
        & $\trilL + \trilK + \left(\bbL^\frac{\theta}{2} + \bbK^\frac{\theta}{2} -I_n \right)^\frac{2}{\theta}$\\
        \midrule
        $t \odot L$
        & $t\trilL + \bbL^t$
        & $t\trilL + \left(t\bbL^\theta + (1-t)I_n \right)^\frac{1}{\theta}$
        & $t\trilL + \left(t\bbL^\frac{\theta}{2} + (1-t)I_n \right)^\frac{2}{\theta}$\\
        \bottomrule
    \end{tabular}
    }
\end{table}

\cref{pcm:tab:riem_gyro_operators_cholesky} summarizes all the Riemannian and gyro operators. The Riemannian operators under $\defDGBWM$ and $\defDPM$ are mostly computed using the diagonal power function, while those under the diagonal log metric are computed using the diagonal logarithm or exponentiation. This indicates that our $\defDPM$ and $\defDGBWM$ may have better numerical stability than the existing diagonal log metric, as logarithm or exponentiation might overly stretch the diagonal elements compared with the power function. The gyro operations under our $\defDPM$ and $\defDGBWM$ also have numerical advantages over those under the diagonal log metric. The former are based on linear operations combined with power and its inverse, causing relatively minor changes to the input magnitude, while the gyro operations under the diagonal log metric are based on products or powers, resulting in more noticeable alterations to the input magnitude.

\subsection{Geometries on the SPD Manifold}
\label{pcm:sec:spd-geometries}

This section discusses the Riemannian metrics on the SPD manifold via the Cholesky decomposition. We first review some basic properties of the Cholesky decomposition, followed by the SPD metrics.

The Cholesky decomposition, denoted by $\chol(\cdot): \spd{n} \rightarrow \chospace{n}$, is a diffeomorphism \citep{lin2019riemannian}.
Therefore, it can pull back the Riemannian and gyro-structures from the Cholesky manifold $\chospace{n}$ to the SPD manifold $\spd{n}$.
We call the pullbacks of $\defDPM$ and $\defDGBWM$ through the Cholesky decomposition the Power-Cholesky Metric ($\defCDEM$) and Bures--Wasserstein--Cholesky Metric ($\defCDGBWM$), respectively. Then, the Riemannian operators under $\defCDEM$ and $\defCDGBWM$ can be obtained from the properties of Riemannian isometries (\cref{def:ch2-riemannian-isometry}).

\mypara{Operators.} Let $\mathcal{C} \in \{\defDPM, \defDGBWM \}$ and $\mathcal{S} \in \{\defCDEM, \defCDGBWM \}$. We denote the Riemannian logarithm, exponential map, geodesic, parallel transport along the geodesic, geodesic distance, weighted Fréchet mean, gyroaddition, and scalar gyromultiplication on $\{\spd{n},g^{\mathcal{S}} \}$ by $\rielog ^{\mathcal{S}}$, $\rieexp ^{\mathcal{S}}$, $\gamma ^{\mathcal{S}}$, $\rmpt ^{\mathcal{S}}$, $\dist ^{\mathcal{S}}(\cdot,\cdot)$, $\wfm ^\mathcal{S}$, $\oplus ^\mathcal{S}$, and $\odot ^\mathcal{S}$, respectively, while $\rielog ^\mathcal{C}$, $\rieexp ^\mathcal{C}$, $\gamma ^\mathcal{C}$, $\rmpt ^\mathcal{C}$, $\dist ^\mathcal{C}(\cdot,\cdot)$, $\wfm ^\mathcal{C}$, $\oplus ^\mathcal{C}$, and $\odot ^\mathcal{C}$ are their counterparts on $\{\chospace{n}, g ^\mathcal{C}\}$.
For $P,Q \in \spd{n}$, $V,W \in T_P\spd{n}$, and $\{P_i \in \spd{n} \}_{i=1}^N$ with weights $\{w_i\}_{i=1}^N$ satisfying $w_i>0$ for all $i$ and $\sum_{i=1}^N w_i=1$, we have the following Riemannian and gyro operators:
\begin{align}
    \gamma ^\mathcal{S} _{(P,V)}(t) &= \chol ^{-1} \left( \gamma ^\mathcal{C} _{(L,\widetilde{V})}(t) \right), \\
    \rielog ^\mathcal{S} _{P}(Q) &= (\chol_{*,P})^{-1} \left( \rielog ^\mathcal{C} _L (K) \right), \\
    \rieexp ^\mathcal{S} _P(V) &= \chol ^{-1} \left( \rieexp ^\mathcal{C}_{L} \left( \widetilde{V} \right)\right), \\
    \pt{P}{Q} ^\mathcal{S} (V) &= (\chol_{*,Q})^{-1} \left( \rmpt ^\mathcal{C}_{L \rightarrow K} \left( \widetilde{V} \right) \right), \\
    \dist ^\mathcal{S} (P,Q) &= \dist ^\mathcal{C} (L,K), \\
    \wfm ^\mathcal{S} (\{ w_i \},\{ P_i \}) &= \chol^{-1} \left( \wfm ^\mathcal{C} \left(\{ w_i \},\{ L_i \} \right) \right), \\
    P \oplus ^\mathcal{S} Q &= \chol ^{-1} (L \oplus ^\mathcal{C} K), \\
    t \odot ^\mathcal{S} P &= \chol ^{-1} (t \odot ^\mathcal{C} L),
\end{align}
where $P=LL^\top$, $Q=KK^\top$, and $P_i=L_i L_i^\top$ are Cholesky decompositions. Here, $\widetilde{V}=\chol_{*,P}(V)$ is the Cholesky differential as defined in \cref{sec:ch2-matrix-functions-differentials}. Besides, when $\mathcal{C}$ is the diagonal log metric, the above recovers the Riemannian and gyro-structures under LCM.

The above gyro operations, when well-defined, also satisfy the gyrovector-space axioms.

\begin{paristheorem}
    \label{pcm:thm:gyro_spaces_spd}
    \linktoproof{pcm:thm:gyro_spaces_spd}
    $\{\spd{n}, \oplus ^\mathcal{S},\odot ^\mathcal{S}\}$ satisfies all the axioms of gyrovector spaces.
\end{paristheorem}

\begin{parisremark}
    \label{pcm:rmk:spd_metric_stable}
    As discussed in \cref{pcm:subsec:num_adv}, the Cholesky metrics $\defDPM$ and $\defDGBWM$ are more numerically stable than the existing diagonal log metric. As pullback metrics through the Cholesky decomposition, our $\defCDEM$ and $\defCDGBWM$ therefore preserve the advantage of numerical stability over the existing LCM. Besides, all the Riemannian operators have closed-form expressions and are easy to use, as the differential maps of the Cholesky decomposition can be easily calculated.
\end{parisremark}

\subsection{Applications to SPD Neural Networks}
\label{pcm:sec:applications}
The closed-form operators derived in \cref{pcm:sec:spd-geometries} make the proposed metrics directly applicable to SPD neural networks. In this section, we apply the proposed SPD metrics $\defCDEM$ and $\defCDGBWM$ to build MLR classifiers and residual blocks on the SPD manifold.

\mypara{MLR.} The Euclidean point-to-hyperplane formulation and its Riemannian extension are developed in \cref{chapter:rmlr}. Substituting the operators of the proposed metrics into that formulation gives the following SPD MLRs.
\begin{paristheorem}
    \label{pcm:thm:spd_mlrs}
    \linktoproof{pcm:thm:spd_mlrs}
    Given an input SPD matrix $S \in \spd{n}$, the $C$-class SPD MLRs under $\defCDEM$ and $\defCDGBWM$ are
    \begin{align}
        \label{pcm:eq:rmlr_pm_lt_v0}
        \defCDEM: p(y=k \mid S \in \spd{n})
        &\propto \exp \left[\begin{aligned}
        &\langle \trilK-\trilLk, \trilAk \rangle \\
        &+\frac{1}{2\theta} \langle \bbK^\theta -\bbL_k^\theta,\bbA_k \rangle
        \end{aligned}\right],\\
        \label{pcm:eq:rmlr_pm_lt}
        \defCDGBWM: p(y=k \mid S \in \spd{n})
        &\propto \exp \left[\begin{aligned}
        &\langle \trilK-\trilLk, \trilAk \rangle \\
        &+\frac{1}{4\theta} \langle \bbK^\frac{\theta}{2} -\bbL_k^\frac{\theta}{2}, \bbM^{-1} \bbA_k \rangle
        \end{aligned}\right],
    \end{align}
	    where $S=KK^\top$ and $P_k = L_k L_k^\top$ are Cholesky decompositions. The parameters are $P_k \in \spd{n}$ and $A_k \in \trilspace{n}$ for each class $k=1,\cdots,C$.
\end{paristheorem}

\mypara{Residual blocks.} The general construction and the specialized SPD residual-block expression are reviewed in \cref{app:backbone-others}. The only component that varies across metrics is the Riemannian exponential map; substituting the operators derived above therefore gives residual blocks under the proposed metrics.

\subsection{Experiments}
\label{pcm:sec:experiments}
We first compare our metrics against the popular AIM, LEM, and LCM when building SPD MLR classifiers and residual blocks. Then, we evaluate the proposed metrics through numerical experiments. More details on data sets and experimental settings are provided in \cref{app:datasets,pcm:app:sec:exp_details}.

\subsubsection{Riemannian Classifiers}

\begin{table}[t]
  \centering
  \caption{SPD MLRs under different metrics on the SPDNet backbone. The best results are \firstresults{bold}.
  } 
  \label{pcm:tab:spd_mlr_results}
  \begin{minipage}[c]{0.24\linewidth}
    \centering
    \resizebox{1\linewidth}{!}{
    \begin{tabular}{c cc}
    \multicolumn{3}{c}{\textbf{(a) Radar}} \\
    \midrule
    \textbf{Metric} & \textbf{Acc}   & \textbf{Time} \\
    \midrule
    AIM   & 94.53 ± 0.95 & 0.80  \\
    LEM   & 93.55 ± 1.21 & 0.76  \\
    LCM   & 93.49 ± 1.25 & 0.72  \\
    \midrule
    \rowcolor{HilightColor}$\defCDEM$   & \firstresults{95.79 ± 0.38} & 0.72  \\
    \rowcolor{HilightColor}$\defCDBWM$  & 93.93 ± 0.79 & 0.71  \\
    \bottomrule
    \end{tabular}%
        }
  \end{minipage}
  \hfill
  \begin{minipage}[c]{0.45\linewidth}
    \centering
    \resizebox{1\linewidth}{!}{
    \begin{tabular}{c cc cc cc}
        \multicolumn{7}{c}{\textbf{(b) HDM05}} \\
        \midrule
        \multirow{2}[4]{*}{\textbf{Metric}} & \multicolumn{2}{c}{\textbf{1-Block}} & \multicolumn{2}{c}{\textbf{2-Block}} & \multicolumn{2}{c}{\textbf{3-Block}} \\
    \cmidrule{2-7}          & \textbf{Acc} & \textbf{Time} & \textbf{Acc} & \textbf{Time} & \textbf{Acc} & \textbf{Time} \\
        \midrule
        AIM   & 58.07 ± 0.64 & 17.32  & 60.72 ± 0.62 & 18.75  & 61.14 ± 0.94 & 19.23  \\
        LEM   & 56.97 ± 0.61 & 2.21  & 60.69 ± 1.02 & 2.92  & 60.28 ± 0.91 & 3.50  \\
        LCM   & 60.69 ± 1.89 & 1.83  & 62.61 ± 1.46  & 2.40  & 62.33 ± 2.15 & 2.90  \\
        \midrule
        \rowcolor{HilightColor}$\defCDEM$   & 62.51 ± 1.65 & 1.58  & 63.66 ± 1.30 & 2.29  & 65.75 ± 2.86 & 2.76  \\
        \rowcolor{HilightColor}$\defCDBWM$  & \firstresults{62.71 ± 0.88} & 1.64  & \firstresults{64.52 ± 0.56} & 2.27  & \firstresults{67.40 ± 0.90} & 2.87  \\
        \bottomrule
    \end{tabular}%
    }
  \end{minipage}
  \hfill
  \begin{minipage}[c]{0.24\linewidth}
    \centering
    \resizebox{1\linewidth}{!}{
    \begin{tabular}{c cc}
        \multicolumn{3}{c}{\textbf{(c) FPHA}} \\
        \midrule
        \textbf{Metric} & \textbf{Acc}   & \textbf{Time} \\
        \midrule
        AIM   & 85.57 ± 0.50 & 7.14  \\
        LEM   & 85.90 ± 0.47 & 0.98  \\
        LCM   & 86.37 ± 0.59 & 0.74  \\
        \midrule
        \rowcolor{HilightColor}$\defCDEM$   & \firstresults{89.40 ± 0.13} & 0.69  \\
        \rowcolor{HilightColor}$\defCDBWM$  & 86.27 ± 0.60 & 0.70  \\
        \bottomrule
    \end{tabular}%
    }
  \end{minipage}
\end{table}

Following the SPD learning setup in \cref{rmlr:sec:experiments} and \citet{huang2017riemannian}, we adopt the Radar data set \citep{brooks2019riemannian} for radar signal classification, and the HDM05 \citep{muller2007documentation} and FPHA \citep{garcia2018first} data sets for human action recognition. We compare the SPD MLRs under our metrics with those under AIM, LEM, and LCM in \cref{rmlr:thm:spdmlrs}. Following \cref{rmlr:sec:experiments} and \citet{nguyen2023building}, we adopt SPDNet \citep{huang2017riemannian} and GyroSPD \citep{nguyen2023building} as two backbones, both mimicking feedforward neural networks. For simplicity, we set $\bbM$ in $\defCDGBWM$ to the identity matrix.

\mypara{SPDNet.} On HDM05, we further evaluate architectures with up to three transformation blocks. \cref{pcm:tab:spd_mlr_results} reports the five-fold accuracy and training time per epoch, from which we draw the following observations.
\begin{itemize}
    \item \mypara{Effectiveness.} Our metrics generally yield higher accuracy than their counterparts. Notably, they outperform LCM, although both originate from the Cholesky product structure. This improvement is attributed to the fact that the diagonal logarithm and exponentiation in LCM tend to overly stretch the diagonal entries, \ie eigenvalues of the Cholesky factors, whereas our diagonal power transformation achieves a more balanced scaling.
    \item \mypara{Efficiency.} Our metrics substantially reduce computational cost compared to AIM, remain faster than LEM, and achieve efficiency comparable to LCM. Together with their superior accuracy, these results highlight the dual advantages of our approach in both effectiveness and efficiency.
\end{itemize}

\begin{wraptable}{r}{0.6\linewidth}
  \centering
  \caption{SPD MLRs on the GyroSPD backbone.}
  \label{pcm:tab:spd_mlr-gyrospd}%
  \resizebox{\linewidth}{!}{
    \begin{tabular}{c cc cc cc}
    \toprule
    \multirow{2}[4]{*}{\textbf{Metric}} & \multicolumn{2}{c}{\textbf{Radar}} & \multicolumn{2}{c}{\textbf{HDM05}} & \multicolumn{2}{c}{\textbf{FPHA}} \\
    \cmidrule{2-7}          & \textbf{Acc} & \textbf{Time} & \textbf{Acc} & \textbf{Time} & \textbf{Acc} & \textbf{Time} \\
    \midrule
    AIM   & 96.80 ± 0.59 & 1.23  & 66.05 ± 1.80 & 21.65 & 85.77 ± 0.52 & 11.48 \\
    LEM   & 96.58 ± 0.27 & 1.18  & 66.42 ± 0.47 & 2.02  & 85.87 ± 0.79 & 1.22 \\
    LCM   & 96.29 ± 0.53 & 1.12  & 68.37 ± 0.66 & 1.66  & 89.83 ± 0.28 & 0.98 \\
    \midrule
    \rowcolor{HilightColor} $\defCDEM$   & \firstresults{97.04 ± 0.64} & 1.18  & 71.93 ± 1.21 & 1.51  & \firstresults{91.17 ± 0.30} & 1.00 \\
    \rowcolor{HilightColor} $\defCDBWM$  & 96.21 ± 0.25 & 1.05  & \firstresults{72.74 ± 0.43} & 1.58  & 91.00 ± 0.11 & 0.96 \\
    \bottomrule
    \end{tabular}%
    }
\end{wraptable}%
\mypara{GyroSPD.} \cref{pcm:tab:spd_mlr-gyrospd} reports the results on the GyroSPD backbone, which consists of a single gyrotranslation layer followed by an SPD MLR. Similar to the SPDNet results, our metrics achieve comparable or superior performance to LCM across all data sets while maintaining comparable efficiency. On HDM05 and FPHA, both proposed metrics deliver higher accuracy with lower runtime than AIM and LEM. On Radar, PCM achieves the highest accuracy, while BWCM has the lowest runtime.

\subsubsection{Riemannian Residual Blocks}

\begin{wraptable}{r}{0.50\textwidth}
  \centering
  \caption{Results on residual blocks.}
  \label{pcm:tab:results_rresnet}
  \resizebox{\linewidth}{!}{
    \begin{tabular}{c cc cc cc}
    \toprule
    \multirow{2}[4]{*}{\textbf{Metric}} & \multicolumn{2}{c}{\textbf{Radar}} & \multicolumn{2}{c}{\textbf{HDM05}} & \multicolumn{2}{c}{\textbf{FPHA}} \\
    \cmidrule{2-7}          & \textbf{Acc} & \textbf{Time} & \textbf{Acc} & \textbf{Time} & \textbf{Acc} & \textbf{Time} \\
    \midrule
    AIM & 96.4  & 1.02  & 57.01 & 1.14  & 87.33 & 0.72 \\
    LEM & 97.07 & 0.81  & 67.52 & 0.52  & 86.17 & 0.32 \\
    LCM & 97.07 & 0.85  & 66.27 & 0.63  & 86.83 & 0.49 \\
    \midrule
    \rowcolor{HilightColor} $\defCDEM$ & \firstresults{97.87} & 0.85  & \firstresults{68.05} & 0.63  & \firstresults{88.33} & 0.48 \\
    \bottomrule
    \end{tabular}%
    }
\end{wraptable}%
Riemannian ResNet (RResNet) was introduced by \citet{katsman2024riemannian}. Its backbone architecture largely follows SPDNet. The key difference lies in the head: while SPDNet directly applies a classification layer, RResNet inserts a residual block before the classification head. Accordingly, we adopt the following classification head under each metric: $\rielog_{I_n} + \mathrm{FC} + \softmax$. Since the Riemannian exponential is similar for $\defCDEM$ and $\defCDGBWM$, we focus on $\defCDEM$ and compare it against AIM, LEM, and LCM in constructing RResNet. \cref{pcm:tab:results_rresnet} reports the best results across three trials, showing that our metric consistently achieves superior accuracy while maintaining comparable efficiency.

\FloatBarrier

\subsubsection{Numerical Stability}
\label{pcm:subsec:numerical-stability}

\begin{table}[t]
    \centering
    \caption{Failure probabilities (\%) of geodesics under different metrics with small eigenvalues in $L \in \chospace{n}$. An output matrix containing any \textsc{Inf} or \textsc{NaN} is considered a failure. Here, DLM denotes the diagonal log metric, while DPM and DBWM denote $\defDPM$ and $\defDBWM$, respectively.}
    \label{pcm:tab:geodesic_stability}
    \resizebox{0.95\linewidth}{!}{
    \begin{tabular}{c c >{\columncolor{HilightColor}}c>{\columncolor{HilightColor}}c >{\columncolor{HilightColor}}c>{\columncolor{HilightColor}}c >{\columncolor{HilightColor}}c>{\columncolor{HilightColor}}c c >{\columncolor{HilightColor}}c>{\columncolor{HilightColor}}c >{\columncolor{HilightColor}}c>{\columncolor{HilightColor}}c >{\columncolor{HilightColor}}c>{\columncolor{HilightColor}}c}
    \toprule
    \multirow{3}[6]{*}{$\boldsymbol{\epsilon}$} & \multicolumn{7}{c}{\textbf{$3 \times 3$ for small matrices}} & \multicolumn{7}{c}{\textbf{$256 \times 256$ for large matrices}} \\
    \cmidrule{2-15}          & \multirow{2}[4]{*}{DLM} & \multicolumn{2}{>{\columncolor{HilightColor}}c}{$\theta=1.5$} & \multicolumn{2}{>{\columncolor{HilightColor}}c}{$\theta=0.5$} & \multicolumn{2}{>{\columncolor{HilightColor}}c}{$\theta=0.15$} & \multirow{2}[4]{*}{DLM} & \multicolumn{2}{>{\columncolor{HilightColor}}c}{$\theta=1.5$} & \multicolumn{2}{>{\columncolor{HilightColor}}c}{$\theta=0.5$} & \multicolumn{2}{>{\columncolor{HilightColor}}c}{$\theta=0.15$} \\
    \cmidrule{3-8}\cmidrule{10-15}          &       & DPM   & DBWM  & DPM   & DBWM  & DPM   & DBWM  &       & DPM   & \multicolumn{1}{>{\columncolor{HilightColor}}c}{DBWM} & DPM   & DBWM & DPM   & DBWM \\
    \midrule
    $1e^{-1}$ & 0.62 & 0 & 0 & 0 & 0 & 0 & 0 & 14.29 & 0 & 0 & 0 & 0 & 0 & 0 \\
    $1e^{-2}$ & 5.70 & 0 & 0 & 0 & 0 & 0 & 0 & 18.48 & 0 & 0 & 0 & 0 & 0 & 0 \\
    $1e^{-3}$ & 51.32 & 0 & 0 & 0 & 0 & 0 & 0 & 58.35 & 0 & 0 & 0 & 0 & 0 & 0 \\
    $1e^{-4}$ & 94.34 & 0 & 0 & 0 & 0 & 0 & 0 & 95.02 & 0 & 0 & 0 & 0 & 0 & 0 \\
    $1e^{-5}$ & 99.39 & 0 & 0 & 0 & 0 & 0 & 0 & 99.47 & 0 & 0 & 0 & 0 & 0 & 0 \\
    $1e^{-10}$ & 100 & 0 & 0 & 0 & 0 & 0 & 0 & 100 & 0 & 0 & 0 & 0 & 0 & 0 \\
    $1e^{-15}$ & 100 & 0 & 0 & 0 & 0 & 0 & 0 & 100 & 0 & 0 & 0 & 0 & 0 & 0 \\
    \midrule
    $1e^{-20}$ & 100 & 0 & 0 & 0 & 0 & 0 & 0.002 & 100 & 0 & 0 & 0 & 0 & 0 & 0.02 \\
    $1e^{-21}$ & 100   & 0     & 0     & 0     & 0     & 0     & 0.03 & 100   & 0     & 0     & 0     & 0     & 0     & 0.01 \\
    $1e^{-22}$ & 100   & 0     & 0     & 0     & 0     & 0     & 0.25 & 100   & 0     & 0     & 0     & 0     & 0     & 0.23 \\
    $1e^{-23}$ & 100   & 0     & 0     & 0     & 0     & 0     & 2.26 & 100   & 0     & 0     & 0     & 0     & 0     & 2.42 \\
    $1e^{-24}$ & 100   & 0     & 0     & 0     & 0     & 0     & 22.98 & 100   & 0     & 0     & 0     & 0     & 0     & 23.13 \\
    $1e^{-25}$ & 100   & 0     & 0     & 0     & 0     & 0     & 86.34 & 100   & 0     & 0     & 0     & 0     & 0     & 86.58 \\
    $1e^{-30}$ & 100   & 0     & 0     & 0     & 0     & 0     & 100   & 100   & 0     & 0     & 0     & 0     & 0     & 100 \\
    \bottomrule
    \end{tabular}%
    }
\end{table}

As discussed by \citet[p.~16]{lin2019riemannian}, LCM is more stable than AIM and LEM owing to the numerical advantage of Cholesky decomposition over SVD. Moreover, as highlighted in \cref{pcm:rmk:spd_metric_stable}, the essential distinction between our metrics and LCM lies in the diagonal operations: ours rely on diagonal power, while LCM employs diagonal exponentiation and logarithm. This structural difference grants our metrics stronger numerical stability and robustness compared with LCM, as well as AIM and LEM. To validate this, we evaluate geodesics on the Cholesky manifold. We generate $100{,}000$ synthetic $n \times n$ Cholesky matrices $L$ and tangent vectors $X \in \trilspace{n}$, where each entry is uniformly sampled from $[0,1]$, and we set the smallest eigenvalue (diagonal entry) of $L$ to $\epsilon$. We test two representative sizes: $3 \times 3$ matrices, commonly used in diffusion tensor imaging \citep{arsigny2007geometric}, and $256 \times 256$ matrices, typical in computer vision \citep{li2018towards,wang2023towards}. The deformation parameter $\theta$ is set to $1.5$, $0.5$, and $0.15$. For $\defDGBWM$, we set $\bbM=I_n$. As shown in \cref{pcm:tab:geodesic_stability}, our $\defDPM$ and $\defDBWM$ remain highly stable across a wide range of $\epsilon$. For $3 \times 3$ matrices, the diagonal log metric already deteriorates at $\epsilon=1e^{-3}$, with failure rates increasing rapidly as $\epsilon$ decreases. For $256 \times 256$ matrices, instability emerges even earlier at $\epsilon=1e^{-1}$. In contrast, our metrics remain stable down to $\epsilon=1e^{-30}$ in most cases. The only exception occurs when $\theta=0.15$, where failures appear around $\epsilon=1e^{-20}$. This behavior is expected since both $\defDPM$ and $\defDBWM$ converge to the diagonal log metric as $\theta \to 0$, thereby inheriting its instability in this limit. Overall, these results demonstrate the superior numerical robustness of our metrics.

\subsubsection{Tensor Interpolation}
As shown by \citet{arsigny2005fast,arsigny2007geometric}, geodesic interpolation of SPD matrices is important in diffusion tensor imaging. This experiment illustrates geodesic interpolation under different SPD metrics. Let $P=LL^\top$ and $Q=KK^\top$ be the Cholesky decompositions of $P,Q \in \spd{n}$. The geodesics connecting $P$ and $Q$ under $\defCDEM$ and $\defCDGBWM$ are
\begin{align}
    \label{pcm:eq:geodesic_defcem}
    \defCDEM \text{: } & \chol^{-1} \left[ \trilL + t(\trilK-\trilL)+ \left(\bbL^\theta + t(\bbK^\theta -\bbL^\theta) \right)^{\frac{1}{\theta}} \right], \\*
    \label{pcm:eq:geodesic_defdbwm}
    \defCDGBWM \text{: } & \chol^{-1} \left[ \trilL + t(\trilK-\trilL)+ \left(\bbL^\frac{\theta}{2} + t(\bbK^\frac{\theta}{2} -\bbL^\frac{\theta}{2}) \right)^{\frac{2}{\theta}} \right].
\end{align}

\begin{figure}[t]
\centering
\includegraphics[width=0.99\linewidth,trim={0cm 0cm 0cm 0cm}]{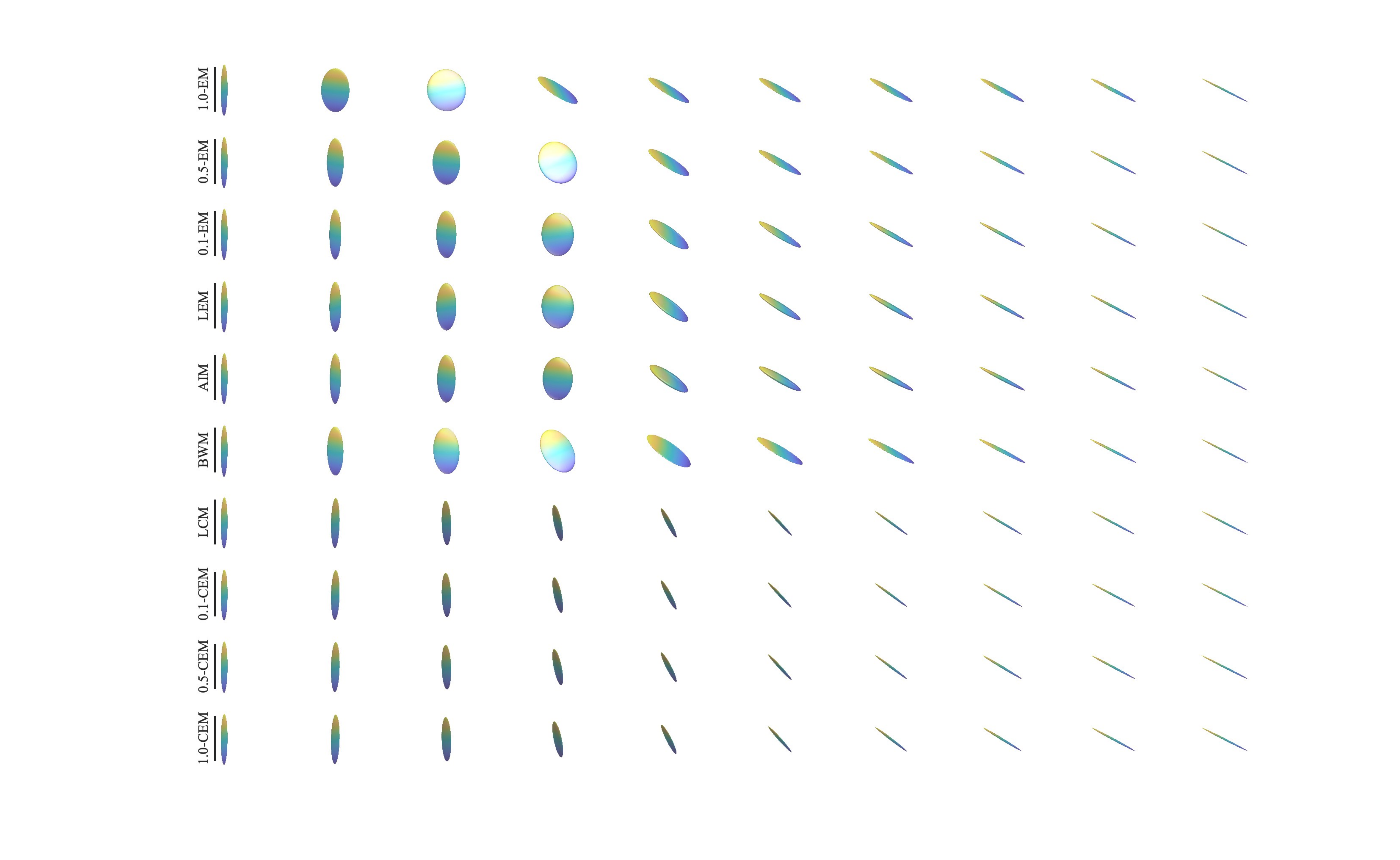}
\caption{
Geodesic interpolation of SPD matrices under different Riemannian metrics. Each $3 \times 3$ SPD matrix can be visualized as an ellipsoid \citep{arsigny2005fast}. The two endpoints are fixed across all metrics.
}
\label{pcm:fig:swelling_effects}
\end{figure}

\begin{table}[t]
  \centering
  \caption{Swelling effects of geodesic SPD interpolations. Deeper greens indicate greater swelling.}
  \label{pcm:tab:interpolation_determinants}%
  \resizebox{0.95\linewidth}{!}{
    \begin{tabular}{c cccccccccc}
    \toprule
    \multirow{2}[4]{*}{\textbf{Metric}} & \multicolumn{10}{c}{\textbf{The determinant of the $i$-th interpolation}} \\
\cmidrule{2-11}          & 0     & 1     & 2     & 3     & 4     & 5     & 6     & 7     & 8     & 9 \\
    \midrule
    \rowcolor[rgb]{ .663,  .816,  .557} 1.0-EM & 3.07  & 104.86 & 182.09 & 234.38 & 261.35 & 262.64 & 237.86 & 186.64 & 108.61 & 3.38 \\
    \rowcolor[rgb]{ .776,  .878,  .706} 0.5-EM & 3.07  & 18.67 & 39.93 & 59.53 & 71.96 & 73.79 & 64.14 & 45.01 & 21.73 & 3.38 \\
    0.1-EM & 3.07  & 4.25  & 5.42  & 6.38  & 6.96  & 7.05  & 6.62  & 5.75  & 4.6   & 3.38 \\
    LEM   & 3.07  & 3.1   & 3.14  & 3.17  & 3.2   & 3.24  & 3.27  & 3.31  & 3.34  & 3.38 \\
    AIM   & 3.07  & 3.1   & 3.14  & 3.17  & 3.2   & 3.24  & 3.27  & 3.31  & 3.34  & 3.38 \\
    \rowcolor[rgb]{ .886,  .937,  .855} BWM   & 3.07  & 15.32 & 32.04 & 48.14 & 59.22 & 62.09 & 55.33 & 39.93 & 19.98 & 3.38 \\
    LCM   & 3.07  & 3.1   & 3.14  & 3.17  & 3.2   & 3.24  & 3.27  & 3.31  & 3.34  & 3.38 \\
    \midrule
    0.1-PCM & 3.07  & 3.15  & 3.23  & 3.29  & 3.34  & 3.37  & 3.39  & 3.4   & 3.4   & 3.38 \\
    0.5-PCM & 3.07  & 3.35  & 3.59  & 3.79  & 3.91  & 3.97  & 3.94  & 3.83  & 3.64  & 3.38 \\
    1.0-PCM & 3.07  & 3.6   & 4.07  & 4.46  & 4.72  & 4.83  & 4.76  & 4.49  & 4.03  & 3.38 \\
    \bottomrule
    \end{tabular}%
    }
\end{table}%

As the geodesics under $\defCDEM$ and $\defCDGBWM$ have similar expressions, we focus on $\defCDEM$. \cref{pcm:fig:swelling_effects} visualizes the geodesic interpolations on $\spd{3}$ under different metrics, including $\PEM$, LEM, AIM, BWM, LCM, and $\defCDEM$. \cref{pcm:tab:interpolation_determinants} presents the associated determinants of the interpolated SPD matrices. We can make the following observations.
\begin{enumerate}
    \item
    The standard Euclidean metric ($1$-EM) exhibits a significant swelling effect, where the maximal determinant of interpolation is much larger than the determinants of the starting and end points. Although matrix power can mitigate the swelling effect, $\PEM$ still suffers from swelling.
    \item
    We find that BWM also demonstrates a clear swelling effect. In contrast, LCM, AIM, and LEM show no swelling effect.
    \item
    The trivial PCM ($\theta=1$) considerably mitigates the swelling effect compared to the Euclidean metric, but it still exhibits some level of swelling. However, by introducing Cholesky power deformation, $\defCDEM$ effectively reduces the swelling effect. Notably, the swelling effect of $\defCDEM$ is significantly weaker than that of $\PEM$ under the same $\theta$.
    \item
    Our PCM shows an interpolation visually similar to that under LCM. This suggests that our metric retains some practical potential of LCM but with better numerical stability (as demonstrated in \cref{pcm:subsec:numerical-stability}).
\end{enumerate}

\subsubsection{Asymptotic Complexity}
We investigate the scalability of our PCM and BWCM in comparison with five existing SPD metrics, namely AIM, LEM, LCM, PEM, and BWM. We use SPD MLR as a representative application. We first analyze the asymptotic complexity of each SPD MLR. We then complement this analysis with synthetic experiments that measure the actual runtime of a single SPD MLR training step across different matrix dimensions.

\begin{table}[t]
\centering
\caption{Number of matrix functions required per sample for a $C$-class SPD MLR. Spectral matrix functions include matrix logarithm, matrix power, and the Lyapunov operator.}
\label{pcm:tab:mlr-decompositions}
\resizebox{\linewidth}{!}{%
\begin{tabular}{ccc}
\toprule
\textbf{Metric} & \textbf{Num. spectral matrix functions} & \textbf{Num. Cholesky decompositions} \\
\midrule
AIM & $1 + 2C$ & $0$ \\
LEM & $1 + C$ & $0$ \\
LCM & $0$ & $1 + C$ \\
PEM & $1 + C$ & $0$ \\
BWM & $1 + 3C$ & $C$ \\
PCM & $0$ & $1 + C$ \\
BWCM & $0$ & $1 + C$ \\
\bottomrule
\end{tabular}}
\end{table}

As shown in \cref{rmlr:thm:spdmlrs,pcm:thm:spd_mlrs}, the $C$-class SPD MLRs under different metrics for the input $S \in \spd{n}$ are
\begin{align*}
\text{AIM:}\quad
p(y=k \mid S)
&\propto \exp\left[\inner{\log\left(P_k^{-\frac{1}{2}} S P_k^{-\frac{1}{2}}\right)}{A_k}\right],\\
\text{LEM:}\quad
p(y=k \mid S)
&\propto \exp\left[\inner{\log(S)-\log(P_k)}{A_k}\right],\\
\text{LCM:}\quad
p(y=k \mid S)
&\propto \exp\left[
\inner{\begin{aligned}
&\trilK - \trilLk + \log(\bbK) \\
&- \log(\bbL_k)
\end{aligned}}{\trilAk + \frac{1}{2} \bbA_k}\right],\\
\text{PEM:}\quad
p(y=k \mid S)
&\propto \exp\left[\frac{1}{\theta}\inner{S^{\theta} - P_k^{\theta}}{A_k}\right],\\
\text{BWM:}\quad
p(y=k \mid S)
&\propto \exp\left[\frac{1}{2}\inner{(P_kS)^{\frac{1}{2}}
 + (SP_k)^{\frac{1}{2}} - 2P_k}{\mathcal{L}_{P_k}(L_kA_kL_k^\top)}\right],\\
\defCDEM: p(y=k \mid S)
&\propto \exp \left[ \inner{\trilK- \trilLk}{\trilAk}
+\frac{1}{2\theta} \inner{\bbK^\theta -\bbL_k^\theta}{\bbA_k} \right],\\
\defCDGBWM: p(y=k \mid S)
&\propto \exp \left[\begin{aligned}
&\inner{\trilK- \trilLk}{\trilAk} \\
&+\frac{1}{4\theta} \inner{\bbK^\frac{\theta}{2} -\bbL_k^\frac{\theta}{2}}{\bbM^{-1} \bbA_k}
\end{aligned}\right],
\end{align*}
where $P_k \in \spd{n}$ and $A_k \in \sym{n}$ are MLR weights, $\log (\cdot)$ is the matrix logarithm, and $\mathcal{L}_P[V]$ is the solution to the matrix linear system $\mathcal{L}_P[V] P+P \mathcal{L}_P[V]=V$, known as the Lyapunov operator.

\mypara{Analysis.} \cref{pcm:tab:mlr-decompositions} summarizes the number of spectral and Cholesky matrix functions required by each SPD MLR. Cholesky decomposition requires $O(\nicefrac{1}{3}n^3)$ flops, while eigendecomposition costs $O(9n^3)$ flops \citep[Algs.~4.2.3 and 8.3.3]{golub2013matrix}. Combining these counts, \cref{pcm:tab:mlr-complexity} reports the resulting asymptotic per-sample complexity for each metric. Cholesky-based metrics (LCM, PCM, BWCM) are asymptotically more efficient than the eigen-based metrics (LEM, PEM, AIM, and BWM), with AIM and BWM being the slowest among the considered methods. In addition, PCM and BWCM can be practically more efficient than LCM, since diagonal powers are cheaper to compute than diagonal logarithms.

\begin{wraptable}{r}{0.4\textwidth}
\centering
\caption{Asymptotic per-sample complexity of a $C$-class SPD MLR for an $n \times n$ input SPD matrix.}
\label{pcm:tab:mlr-complexity}
\resizebox{\linewidth}{!}{%
\begin{tabular}{cc}
\toprule
\textbf{Metric} & \textbf{Asymptotic complexity} \\
\midrule
AIM  & $O\bigl(9(1 + 2C)n^3\bigr)$ \\
LEM  & $O\bigl(9(1 + C)n^3\bigr)$ \\
LCM  & $O\bigl(\tfrac{1 + C}{3}n^3\bigr)$ \\
PEM  & $O\bigl(9(1 + C)n^3\bigr)$ \\
BWM  & $O\bigl((9(1 + 3C) + \tfrac{C}{3})n^3\bigr)$ \\
PCM  & $O\bigl(\tfrac{1 + C}{3}n^3\bigr)$ \\
BWCM & $O\bigl(\tfrac{1 + C}{3}n^3\bigr)$ \\
\bottomrule
\end{tabular}}
\end{wraptable}

\mypara{Setup.} To validate the asymptotic complexity in \cref{pcm:tab:mlr-complexity}, we measure the average wall-clock time of a single forward--backward training step of an SPD MLR classifier as the matrix dimension increases. The model consists of a single SPD MLR layer with 50 output classes followed by a cross-entropy loss. For each dimension $n \in \{32,64,128,256,512\}$, we randomly generate a batch of 30 $n \times n$ SPD matrices. In each run, we perform one forward and one backward pass and record the total runtime of this step. For PEM, we set the matrix power to $0.5$.

\FloatBarrier

\begin{table}[t]
  \centering
  \caption{Average runtime (in seconds) of one SPD MLR training step across different matrix dimensions.}
    \resizebox{\linewidth}{!}{%
    \begin{tabular}{c ccccc>{\columncolor{HilightColor}}c>{\columncolor{HilightColor}}c}
    \toprule
    \textbf{Dim}   & \textbf{AIM}   & \textbf{LEM}   & \textbf{LCM}   & \textbf{PEM}   & \textbf{BWM}   & \textbf{PCM}   & \textbf{BWCM} \\
    \midrule
    32    & 0.2380  & 0.0077  & 0.0046  & 0.0076  & 0.2377  & \firstresults{0.0040}  & \firstresults{0.0040}  \\
    64    & 1.0139  & 0.0395  & 0.0303  & 0.0473  & 1.1205  & 0.0251  & \firstresults{0.0225}  \\
    128   & 3.6256  & 0.1832  & 0.1490  & 0.1844  & 4.0674  & \firstresults{0.1013}  & 0.1019  \\
    256   & 14.5142  & 0.7793  & 0.5833  & 0.7853  & 16.5918  & \firstresults{0.3848}  & 0.4077  \\
    512   & 60.1918  & 3.2948  & 2.5030  & 3.4357  & 70.8647  & 1.7553  & \firstresults{1.7526}  \\
    \bottomrule
    \end{tabular}}%
  \label{pcm:tab:mlr-scalability}%
\end{table}%

\mypara{Results.} As reported in \cref{pcm:tab:mlr-scalability}, our PCM and BWCM are the fastest metrics across all tested dimensions, and the gap becomes particularly pronounced in the high-dimensional case. For small and medium scales (32 and 64), LCM, PCM, and BWCM have very similar runtimes and all are clearly faster than AIM, LEM, PEM, and BWM. When the dimension increases to 512, AIM and BWM require about $60$ and $70$ seconds per training step, whereas PCM and BWCM remain within roughly $1.7$ seconds. In this setting, PCM and BWCM are even faster than LCM.

\section{Conclusion}
\label{sec:ch6-conclusion}

This chapter developed SPD metric design along two complementary routes. The first route proposed a pullback Euclidean framework and used a learnable general matrix logarithm to construct ALEM. Through this pullback construction, ALEM inherits compatible Hilbert-space, abelian Lie-group, and Riemannian structures, together with closed-form Riemannian operators and weighted Fr\'echet means. We further derived the differentials, gradients, and parameter-update schemes required to learn the logarithm bases. Experiments with applications to SPDNet, LieBN, Riemannian residual blocks, and gyro MLR demonstrate the effectiveness of our metrics.

The second route revealed the product structure of the Cholesky manifold. This structure yielded diagonal power and diagonal Bures--Wasserstein geometries, whose deformed versions converge to the diagonal log metric as the power parameter approaches zero. Transferring these geometries through the Cholesky decomposition produced PCM and BWCM on the SPD manifold. The resulting metrics admit closed-form Riemannian and gyro operators, which directly yield SPD MLR classifiers and residual blocks. Experiments on SPD classifiers and residual networks, together with numerical experiments, supported their effectiveness, efficiency, and numerical robustness.

Together, these routes show that the underlying Riemannian geometry can itself be designed to provide the flexibility, tractability, efficiency, and numerical stability required by Riemannian deep learning.

    \chapter{Conclusion and Future Work}
\label{chapter:conclusion}

\section{Conclusion}
\label{sec:ch7-summary}

This dissertation studied Riemannian deep learning from three connected perspectives: unified Riemannian module design across manifolds, manifold-specific Riemannian network design, and the design of the underlying Riemannian geometries. The main contributions are summarized as follows:

\begin{itemize}
    \item \cref{chapter:normalization,chapter:rmlr} developed unified \emph{network modules} from geometric structures shared across different manifolds. The normalization chapter first developed LieBN on Lie groups, then introduced pseudo-reductive gyrogroups, a new algebraic structure that generalizes classical gyrogroups and Lie groups, and finally developed GyroBN on this foundation, with LieBN recovered as a special case. The classification development progressed from SPD MLRs obtained through exact evaluation of the point-to-hyperplane infimum under flat pullback Euclidean metrics to a general RMLR based on a Riemannian-trigonometric formulation requiring only a well-defined Riemannian logarithm.

    \item \cref{chapter:riemannian-neural-networks} developed \emph{manifold-specific Riemannian network designs} by exploiting additional structures. PVNN developed the geometry and core neural layers of the stable PV model, \emph{Hyperbolic Busemann Neural Networks (HBNN)} derived intrinsic and efficient BMLR and BFC layers from Busemann functions and horospheres on the Poincar\'e and Lorentz models, and CorNet developed correlation MLR, FC, and convolutional layers under five geometries together with accurate Riemannian backpropagation under OLM and LSM.

    \item \cref{chapter:spd-geometries} designed the underlying \emph{Riemannian geometries}. ALEM learned pullback Euclidean metrics through general matrix logarithms, whereas PCM and BWCM used the product structure of the Cholesky manifold to obtain efficient and numerically stable SPD geometries.
\end{itemize}

Taken together, these contributions address Riemannian deep learning at the levels of unified network modules, manifold-specific network designs, and underlying Riemannian geometries while balancing intrinsic structure, generality, computational tractability, and numerical stability.

\section{Future Work}
\label{sec:ch7-future-work}

Two directions are particularly promising: representation learning with novel geometries and geometry-aware generative modeling.

\begin{itemize}
    \item \mypara{Novel geometries for complex relational structure.} Hyperbolic spaces provide an effective inductive bias for hierarchical data \citep{nickel2018learning}. Although mixed-curvature products \citep{gu2019learning,skopek2020mixed} and matrix manifolds \citep{cruceru2021computationally} have demonstrated the benefit of matching geometry to heterogeneous structure, real systems may also contain more complex relation types and structures. Future work could develop novel geometric structures to encode such complex relationships while retaining efficient and tractable Riemannian computations.

    \item \mypara{Geometry-aware generative modeling.} Geometry arises both in latent representations and in spaces of probability distributions. At the latent level, Riemannian variational autoencoders \citep{kalatzis2020riemannian}, continuous normalizing flows \citep{mathieu2020riemannian}, score-based models \citep{de2022riemannian}, and flow matching \citep{chen2024flow} have shown how manifold geometry can shape generative dynamics, while other work has analyzed the geometry of learned latent spaces \citep{arvanitidis2017latent,park2023understanding}. At the distribution level, a growing body of work has explored the geometry of probability distributions through information geometry \citep{cheng2024categorical,davis2024fisher,cheng2025alphaflow} and optimal transport \citep{haviv2025wasserstein,choi2024scalable}. A central problem is therefore to balance the geometry of the latent space with the geometry of the distributions evolving on it.
\end{itemize}

\bibliographystyle{plainnat}
\bibliography{source/ref}

\appendix
    \chapter{Experimental Details and Additional Discussions}
\label{app:experimental-details}

\section{Data Sets}
\label{app:datasets}

This section collects descriptions of the data sets used in the experimental chapters.

\subsection{Skeleton-Based Action Recognition and Gesture Data Sets}
\label{app:subsubsec:datasets_preprocessing_grass}

\mypara{HDM05 \citep{muller2007documentation}.} The HDM05 data set\footnote{\url{https://resources.mpi-inf.mpg.de/HDM05/}} consists of 2,273 skeleton-based motion capture sequences executed by different actors. Each frame records the 3D coordinates of 31 joints. Different experiments use the standard task-specific filtered versions adopted by their corresponding backbones, as detailed in the method-specific experimental settings.

\mypara{FPHA \citep{garcia2018first}.} The FPHA data set\footnote{\url{https://github.com/guiggh/hand_pose_action}} includes 1,175 skeleton-based first-person hand gesture videos of 45 different categories, with 600 clips for training and 575 for testing. Each frame contains the 3D coordinates of 21 hand joints.

\mypara{G3D \citep{bloom2012g3d}.} The G3D data set consists of 663 sequences of 20 different gaming actions. Each sequence records the 3D locations of 20 joints, \ie, 19 bones.

\mypara{NTU60 \citep{shahroudy2016ntu}.} The NTU60 data set\footnote{\url{https://github.com/shahroudy/NTURGB-D}} contains 56,880 skeleton sequences classified into 60 classes, where each frame includes 3D coordinates of 25 or 50 body joints.

\mypara{NTU120 \citep{liu2019ntu}.} The NTU120 data set\footnote{\url{https://github.com/shahroudy/NTURGB-D}} contains 114,480 sequences in 120 action classes.

\subsection{Radar and EEG Signal Data Sets}

\mypara{Radar \citep{brooks2019riemannian}.} The Radar data set\footnote{\url{https://www.dropbox.com/s/dfnlx2bnyh3kjwy/data.zip?dl=0}} contains 3,000 synthetic radar signals equally distributed in 3 classes.

\mypara{Hinss2021 \citep{hinss2021eegdata}.} The Hinss2021 data set\footnote{\url{https://zenodo.org/record/5055046}} is a competition data set containing EEG signals for mental workload estimation. The data set is employed for two tasks, inter-session and inter-subject, which are treated as domain adaptation problems. Geometry-aware methods \citep{yair2019parallel,kobler2022spd} have demonstrated promising performance in EEG classification.

\subsection{Image Classification Data Sets}
\label{app:subsec:image-classification-datasets}

\mypara{AFEW \citep{dhall2018emotiw}.} The \emph{Acted Facial Expressions in the Wild (AFEW)} data set contains seven emotion categories, with 773 videos for training and 383 for validation.

\mypara{CIFAR-10 and CIFAR-100 \citep{krizhevsky2009learning}.} The CIFAR-10 and CIFAR-100 data sets each contain 60{,}000 $32\times32$ color images from 10 and 100 classes, respectively. We use the standard PyTorch splits: 50{,}000 training images and 10{,}000 test images.

\mypara{Tiny-ImageNet \citep{le2015tiny}.} Tiny-ImageNet is a subset of ImageNet with 100{,}000 images from 200 classes, resized to $64\times64$. We use the official validation split for evaluation.

\mypara{ImageNet-1k \citep{deng2009imagenet}.} ImageNet-1k contains 1.28M training images, 50K validation images, and 100K test images distributed across 1k classes.

\subsection{Graph Data Sets}
\label{app:subsec:datasets_ccs}

\begin{table}[t]
\centering
\caption{Summary statistics for the graph data sets.}
\label{pvnn:tab:node-dataset-stats}
\begin{tabular}{ccccc}
\toprule
\textbf{Data Set} & \textbf{\#Nodes} & \textbf{\#Edges} & \textbf{\#Classes} & \textbf{\#Features} \\
\midrule
Disease & 1044 & 1043 & 2 & 1000 \\
Airport & 3188 & 18631 & 4 & 4 \\
PubMed & 19717 & 44338 & 3 & 500 \\
Cora & 2708 & 5429 & 7 & 1433 \\
\bottomrule
\end{tabular}
\end{table}
\mypara{Cora \citep{sen2008collective}.} Cora is a citation network where the nodes represent scientific papers in the area of machine learning, the edges are citations between them, and the labels of the nodes are academic subareas.

\mypara{Disease \citep{anderson1991infectious}.} Disease represents a disease propagation tree that simulates the susceptible, infected, and recovered (SIR) disease transmission model, with each node representing either an infection or a non-infection state.

\mypara{Airport \citep{zhang2018link}.} Airport is a transductive data set where nodes represent airports and edges represent airline routes from OpenFlights.org.

\mypara{PubMed \citep{namata2012query}.} PubMed is a standard benchmark that describes citation networks where nodes represent scientific papers in the area of medicine, the edges are citations between them, and the node labels are academic subareas.

\cref{pvnn:tab:node-dataset-stats} summarizes the above data sets.

\subsection{Genomic Sequence Data Sets}
\label{app:subsec:genomic-sequence-datasets}

\mypara{Transposable Elements Benchmark \citep{khan2025hyperbolic}.} The \emph{Transposable Elements Benchmark (TEB)} comprises seven binary classification data sets that investigate transposable elements across plant and human genomes. The seven data sets are LTR Copia, LINEs, and SINEs for plant retrotransposons; CMC-EnSpm and hAT-Ac for plant DNA transposons; and processed and unprocessed pseudogenes for human pseudogenes. For each data set, positive examples are sequences spanning annotated elements of interest, and negatives are randomly sampled, non-overlapping genomic segments outside these regions. We adopt chromosome-level training, validation, and test splits, using chromosomes 8 and 9 for validation and test in plant genomes, and chromosomes 20--22 and 17--19 for validation and test in human genomes, respectively. \cref{app:tab:teb-stats} provides the summary statistics.

\begin{table}[t]
\centering
\caption{Summary statistics for TEB.}
\label{app:tab:teb-stats}
\resizebox{\linewidth}{!}{%
\begin{tabular}{cccccc}
\toprule
\textbf{Task} & \textbf{Species} & \textbf{Data Sets} & \textbf{Num. classes} & \textbf{Max length} & \textbf{Train / Dev / Test} \\
\midrule
\multirow{3}{*}{Retrotransposons} & \multirow{3}{*}{Plant} & LTR Copia & \multirow{3}{*}{2} & 500 & 7666 / 682 / 568 \\
 &  & LINEs &  & 1000 & 22502 / 2030 / 1782 \\
 &  & SINEs &  & 500 & 21152 / 1836 / 1784 \\
\midrule
\multirow{2}{*}{DNA transposons} & \multirow{2}{*}{Plant} & CMC-EnSpm & \multirow{2}{*}{2} & 200 & 19912 / 1872 / 1808 \\
 &  & hAT-Ac &  & 1000 & 17322 / 1822 / 1428 \\
\midrule
\multirow{2}{*}{Pseudogenes} & \multirow{2}{*}{Human} & processed & \multirow{2}{*}{2} & \multirow{2}{*}{1000} & 17956 / 1046 / 1740 \\
 &  & unprocessed &  &  & 12938 / 766 / 884 \\
\bottomrule
\end{tabular}}
\end{table}

\mypara{Genome Understanding Evaluation \citep{zhou2024dnabert}.} The \emph{Genome Understanding Evaluation (GUE)} benchmark contains seven biologically significant genome analysis tasks that span 28 data sets. The data sets contain sequences ranging from 70 to 1000 base pairs in length and originate from yeast, mouse, human, and virus genomes. The HBNN experiments select Core promoter detection and Promoter detection, together with the multi-class tasks Covid variant classification and species classification, totaling nine data sets. \cref{app:tab:gue-stats} provides their summary statistics.

\begin{table}[t]
\centering
\caption{Summary statistics for the adopted GUE data sets.}
\label{app:tab:gue-stats}
\resizebox{\linewidth}{!}{%
\begin{tabular}{cccccc}
\toprule
\textbf{Task} & \textbf{Species} & \textbf{Data Set} & \textbf{Num. classes} & \textbf{Length} & \textbf{Train / Dev / Test} \\
\midrule
\multirow{3}{*}{Core promoter detection} & \multirow{3}{*}{Human} & tata & \multirow{3}{*}{2} & \multirow{3}{*}{70} & 4904 / 613 / 613 \\
 &  & notata &  &  & 42452 / 5307 / 5307 \\
 &  & all &  &  & 47356 / 5920 / 5920 \\
\midrule
\multirow{3}{*}{Promoter detection} & \multirow{3}{*}{Human} & tata & \multirow{3}{*}{2} & \multirow{3}{*}{300} & 4904 / 613 / 613 \\
 &  & notata &  &  & 42452 / 5307 / 5307 \\
 &  & all &  &  & 47356 / 5920 / 5920 \\
\midrule
Covid variant classification & Virus & Covid & 9 & 1000 & 77669 / 7000 / 7000 \\
\midrule
\multirow{2}{*}{Species classification} & Fungi & Fungi & 25 & 5000 & 8000 / 1000 / 1000 \\
 & Virus & Virus & 20 & 10000 & 4000 / 500 / 500 \\
\bottomrule
\end{tabular}}
\end{table}

\section{Backbone Networks}
\label{app:backbone-networks}

This section collects the backbone network architectures used in the dissertation.

\subsection{Backbone Networks on the SPD Manifold}
\label{app:backbone-spd}

\mypara{SPDNet.} SPDNet \citep{huang2017riemannian} is a classical SPD neural network. It mimics conventional densely connected feedforward networks and consists of three basic building blocks:
\begin{align}
    &\text{BiMap layer: }  S^{k} = W^k S^{k-1} W^{k \top}, \text { with } W^k \text { semi-orthogonal,}\\
    &\text{ReEig layer: } S^{k}=U^{k-1} \max (\Sigma^{k-1}, \epsilon I_{n}) U^{k-1 \top},
    \text { with } S^{k-1}=U^{k-1} \Sigma^{k-1} U^{k-1 \top},\\
    &\text{LogEig layer: } S^{k}=\log(S^{k-1}).
\end{align}
where $\max(\cdot)$ is element-wise maximization. BiMap and ReEig mimic transformation and non-linear activation, while LogEig maps SPD matrices into the tangent space at the identity matrix for classification.

\mypara{SPDNetBN.} SPDNetBN \citep{brooks2019riemannian} further proposed RBN based on AIM:
\begin{align}
    & \text{Centering from mean } M \in \spd{n}: \forall i \leq N, \bar{P}_i \gets M^{-\frac{1}{2}} P_i M^{-\frac{1}{2}},\\
    & \text{Biasing towards parameter } B \in \spd{n}: \forall i \leq N, \hat{P}_i \gets B^{\frac{1}{2}} \bar{P}_i B^{\frac{1}{2}}.
\end{align}

\mypara{TSMNet and SPDDSMBN.} TSMNet \citep{kobler2022spd} can be illustrated as $f_{tc} \rightarrow f_{sc} \rightarrow f_{BiMap} \rightarrow f_{ReEig} \rightarrow f_{LogEig}$, where $f_{tc}$ and $f_{sc}$ denote temporal and spatial convolution, respectively. SPDDSMBN \citep{kobler2022spd} is an improved version of SPDNetBN. Apart from controlling the mean, it can also control variance. The key operation in SPDDSMBN for controlling the mean and variance is
\begin{equation}
    \forall i \leq N, \quad \bar{P}_i \gets \spdtrans{I}{B}[(\spdtrans{M}{I}(P_i))^{\frac{s}{v}}],
\end{equation}
where $M$ is the Riemannian mean, $v^2$ is the Fréchet variance, $B \in \spd{n}$ is the biasing parameter, and $s \in \bbRscalar$ is the scaling factor. Inspired by \citet{yong2020momentum}, during training, SPDDSMBN generates running means and running variances for training and testing with distinct momentum parameters. It uses the training running statistics during training and the testing running statistics during testing. SPDDSMBN also applies domain-specific techniques \citep{chang2019domain}, keeping multiple parallel BN layers and distributing observations according to the associated domains. To share cross-domain knowledge, $s$ is uniformly learned across all domains, and $B$ is set to the identity matrix. \citet{kobler2022spd} adopted SPDDSMBN for domain adaptation in EEG classification.

\mypara{SPDGCN \citep{zhao2023modeling}.} SPDGCN is used as a Riemannian graph neural network backbone.

\mypara{GyroSPD \citep{nguyen2023building}.} GyroSPD substitutes the BiMap layer in SPDNet with the AIM-based gyrotranslation $S^k=W^k\oplusGyrAI S^{k-1}=\left(W^k\right)^{\frac{1}{2}}S^{k-1}\left(W^k\right)^{\frac{1}{2}}$, where $W^k$ is an SPD matrix parameter.

\subsection{Backbone Networks on the Grassmannian}
\label{app:backbone-grassmannian}

\mypara{GyroGr.} GyroGr \citep{nguyen2023building} mimics conventional densely connected feedforward networks and is composed of three basic building blocks. Given an ONB Grassmannian matrix $U^{k-1}$, the gyrotranslation and ProjMap layers are defined as
\begin{align}
    \label{app:eq:gyrotranslation}
    &\text{Gyrotranslation: }  U^{k} = W^{k} \oplusGyrONB U^{k-1}, \quad W^{k} \in \grasonb{p,n}, \\
    &\text{ProjMap: }  P^{k} = U^{k-1}(U^{k-1})^\top.
\end{align}

In addition, the pooling is performed via the space of projection matrices \citep{huang2018building}. Specifically, Grassmannian data are first mapped to the space of projection matrices via the ProjMap layer. A standard mean pooling operation is applied to the resulting projection matrices. Finally, the pooled matrices are projected back to the ONB Grassmannian by SVD. The entire procedure can be expressed as
\begin{equation}
    \label{app:eq:grass_pooling}
    \begin{aligned}
        P^k &= f_{\mathrm{p}} \left( {U}^{k-1}({U}^{k-1})^\top \right), \\
        U^{k} &= O^k_{1:p}, \quad \text{where } P^k \stackrel{\mathrm{SVD}}{:=} O^k \Sigma^k (O^k)^\top,
    \end{aligned}
\end{equation}
where $f_{\mathrm{p}}$ is a regular mean pooling.

\subsection{Backbone Networks on Rotation Matrices}
\label{app:backbone-son}

\mypara{LieNet.} LieNet \citep{huang2017deep} is a classical neural network on rotation matrices. Its latent space is the Lie group $\soprod{N}{3}=\so{3} \times \cdots \times \so{3}$, \ie, $R=(R_1,\ldots,R_N) \in \soprod{N}{3}$. The group and manifold structures on $\soprod{N}{3}$ are defined component-wise. For instance, $R^1 \odot R^2 = (R^1_1R^2_1, \ldots, R^1_N R^2_N)$. There are three basic layers in LieNet:
\begin{align}
    &\text{RotMap layer: }  R^{k} = W^k \odot R^{k-1}, \text { with } W^k \in \soprod{N}{3},\\
    &\text{RotPooling layer: } R^{k}_i=
    \begin{cases}
        R_{m_i, n_i}^{k-1}, & \text { if } \Theta\left(R_{m_i, n_i}^{k-1}\right)>\Theta\left(R_{n_i, m_i}^{k-1}\right), \\ R_{n_i, m_i}^{k-1}, & \text { otherwise, }
    \end{cases}, \\
    &\text{LogMap layer: } R^{k}=\log(R^{k-1}),
\end{align}
where $\Theta(\cdot)$ is the Euler angle, and $(n_i,m_i)$ are two indices. The RotMap and RotPooling layers mimic the convolution and pooling layers, while the LogMap layer maps rotation matrices into the tangent space for classification.

In LieNet, each rotation feature has shape $[\text{num},\text{frame},3,3]$, where num and frame denote the spatial and temporal dimensions. The RotPooling layer is applied along either the spatial or temporal dimension, while the RotMap layer is applied along the spatial dimension, with $W^k$ of size $[\text{num},3,3]$.

\subsection{Backbone Networks on Hyperbolic Spaces}
\label{app:backbone-hyperbolic-spaces}

We briefly recap the Lorentz layers, Poincaré MLR, Poincaré FC layer, and Poincaré $\beta$-concatenation used in the experiments. Throughout this subsection, $K<0$.

\mypara{Lorentz Neural Networks.} Let $x \in \lorentz{n}$ be the input vector. The weight parameters are $W \in \bbR{m \times (n+1)}$ and $v \in \bbR{n+1}$. The Lorentz FC layer \citep[Eq.~3]{chen2022fully} and activation layer \citep[Eq.~13]{bdeir2024fully} are defined in a spacetime manner:
\begin{align}
    \label{app:eq:act_lnn}
    \text{Activation: } y 
    &=\begin{bmatrix}
    \sqrt{\left\|\psi\left(x_s\right)\right\|^2-1 / K} \\
    \psi\left(x_s\right)
    \end{bmatrix},  \\  
    \text{FC: } y 
    &=\begin{bmatrix}
    \sqrt{\|\phi(W x, v)\|^2-1 / K} \\
    \phi(W x, v)
    \end{bmatrix}, \\
    & \text{ with } 
    \phi(W x, v) =\lambda \sigma\left(v^\top x+b^{\prime}\right) \frac{W \psi(x)+b}{\norm{W \psi(x)+b}},
\end{align}
where $b \in \bbR{m}$ and $b' \in \bbRscalar$ are biases, $\psi$ is an activation function, $\sigma$ is the sigmoid function, and $\lambda>0$ is a learnable scaling parameter.

\mypara{Poincaré MLR.} \citet{lebanon2004hyperplane} first reformulated the Euclidean MLR $p(y{=}k\mid x) \propto \exp\left(\inner{a_k}{x} - b_k\right)$ via the point-to-hyperplane distance:
\begin{align}
    p(y{=}k\mid x) &\propto \exp \left(\sign(\inner{a_k}{x}{-}b_k) \norm{a_k} d(x,H_{a_k,b_k}) \right), \\
    H_{a, b} &=\left\{x \in \bbR{n} \mid \inner{a}{x}-b=0\right\}, \quad \text{where } a \in \bbR{n}\setminus\{\zerovec\} \text{ and } b \in \bbRscalar.
\end{align}
For $x\in\pball{n}$, \citet[Eqs.~24--25]{ganea2018hyperbolic} generalized this formulation via geometric reinterpretation, and \citet[Sec.~3.1]{shimizu2021hyperbolic} further derived the resulting closed form:
\begin{equation*}
    v_k(x) = \frac{2\|z_k\|}{\sqrt{|K|}}
    \operatorname{asinh} \left(
    \lambda_x^{K} \langle \sqrt{|K|} x, [z_k]\rangle \cosh(2\sqrt{|K|} r_k)
    -\left(\lambda_x^{K} - 1\right)\sinh(2\sqrt{|K|} r_k)\right),
\end{equation*}
where $\lambda_x^{K}=2(1-|K|\|x\|^2)^{-1}$ is the conformal factor, $p(y{=}k\mid x)\propto\exp\left(v_k(x)\right)$, and $[z_k] = \frac{z_k}{\norm{z_k}}$. Here, $z_k \in \bbR{n}\setminus\{\zerovec\}$ and $r_k \in \bbRscalar$ are parameters. Under the identification $a_k=z_k$ and $b_k=\norm{z_k}r_k$, the Euclidean limit is $\lim_{K\to 0}v_k(x)=4(\inner{a_k}{x}-b_k)$.

\mypara{Poincaré FC Layer.} \citet{shimizu2021hyperbolic} extended the Euclidean FC layer to the Poincaré ball via point-to-hyperplane distances. In Euclidean spaces, the FC can be written element-wise as $y_k = \inner{a_k}{x} - b_k$ with $x, a_k\in\bbR{n}$ and $b_k\in\bbRscalar$. Thus, $y_k$ equals $\norm{a_k}$ times the signed distance from $x$ to the hyperplane $H_{a_k,b_k}$. Under this interpretation, the Poincaré FC layer $\calF:\pball{n}\to\pball{m}$ takes the closed form
\begin{equation}
y = \frac{w}{1+\sqrt{1+|K| \|w\|^2}}, 
\qquad
w_k = |K|^{-1/2} \sinh \left(\sqrt{|K|} v_k(x)\right),
\end{equation}
where $|K|$ is the magnitude of the curvature. Here, $Z=\{z_k\}_{k=1}^m$ and $r=\{r_k\}_{k=1}^m$ parameterize the orientations and biases, and $v_k(x)$ is the Poincaré MLR logit.

\mypara{Poincaré $\beta$-Concatenation.} It generalizes the Euclidean concatenation into the hyperbolic Poincaré ball, stabilizing the norm of the Poincaré vector \citep[Sec.~3.3]{shimizu2021hyperbolic}. Given inputs $\{x_i \in \pball{n_i}\}_{i=1}^N$, it is defined as
\begin{equation}
    \rieexp_{\zerovec} \left( \beta_n \left( \beta_{n_1}^{-1} v_1^\top, \cdots, \beta_{n_N}^{-1} v_{N}^\top \right) \right)^\top \in \pball{n},
\end{equation}
where $v_i = \rielog_{\zerovec}(x_i)$, $n=\sum_{i=1}^N n_i$, and $\beta_\alpha = \mathrm{B}\left(\nicefrac{\alpha}{2}, \nicefrac{1}{2}\right)$ is defined using the beta function.

\subsection{Riemannian Residual Network Backbones}
\label{app:backbone-others}

\mypara{RResNet.} Euclidean residual blocks can be written as
\begin{equation}\label{app:eq:euc_res_block}
    x^{(i)} = x^{(i-1)} + n_i \left(x^{(i-1)}\right),
\end{equation}
where $n_i$ is a network. \citet{katsman2024riemannian} generalized this to manifolds by replacing addition with the Riemannian exponential map:
\begin{equation} \label{app:eq:riem_res_block}
x^{(i)} = \rieexp _{x^{(i-1)}}\left( \ell_i(x^{(i-1)}) \right),
\end{equation}
where $\ell_i: \calM \rightarrow T\calM$ outputs a vector field parameterized by the neural network. As $\rieexp_x(v) = x+v$ for the Euclidean space, it can be immediately shown that \cref{app:eq:riem_res_block} naturally extends \cref{app:eq:euc_res_block} to manifolds. On the SPD manifold, the vector field is generated from the eigenvalues. Specifically, the SPD residual block \citep[Eqs.~22--23]{katsman2024riemannian} is
\begin{equation}
    Y=\rieexp_X\left(Q\operatorname{diag}\left(f\left(\operatorname{spec}(X)\right)\right)Q^\top\right),
\end{equation}
where $X\in\spd{n}$, $\operatorname{spec}(X)$ contains all eigenvalues, $f:\bbR{n}\rightarrow\bbR{n}$ is a neural network, and $Q$ is an orthogonal parameter. Here, $\operatorname{diag}(\cdot)$ returns a diagonal matrix from the input vector.

\section{Experimental Details}
\label{app:method-experimental-details}

\subsection{Lie Group Batch Normalization}
\label{app:liebn-experimental-details}

\subsubsection{LieBN on the SPD Manifold}
\label{app:subsec:exp_details_spd}

The SPDNet and TSMNet backbones, including SPDNetBN and SPDDSMBN, are reviewed in \cref{app:backbone-spd}. We next describe the domain-specific momentum LieBN and implementation details.

\mypara{SPD modeling and preprocessing.} The data set descriptions are collected in \cref{app:datasets}. Throughout this dissertation, every sample covariance matrix is made strictly positive definite before subsequent manifold operations by adding a small diagonal perturbation, $\Sigma \leftarrow \Sigma+\epsilon I_n$ with $\epsilon>0$. Unless otherwise stated, this preprocessing convention is applied to all covariance-based SPD representations. Following the protocol of \citet{brooks2019riemannian}, each Radar signal is divided into windows of length 20, whose series yields one $20 \times 20$ SPD covariance matrix. This produces 3,000 covariance matrices equally distributed across 3 classes. For HDM05, each frame consists of 3D coordinates of 31 joints, and each sequence is modeled by a $93 \times 93$ covariance matrix. Following the protocol of \citet{brooks2019riemannian}, we trim the data set down to 2,086 sequences scattered throughout 117 classes by removing some under-represented classes. For the FPHA, we follow \citet{wang2021symnet} to represent each sequence as a $63 \times 63$ covariance matrix. For Hinss2021, we choose the SOTA method, TSMNet \citep{kobler2022spd}, as our baseline model. We follow the Python implementation\footnote{\url{https://github.com/rkobler/TSMNet}} of \citet{kobler2022spd} to carry out preprocessing. In detail, the Python packages MOABB \citep{jayaram2018moabb} and MNE \citep{gramfort2013meg} are used to preprocess the data set. The applied steps include resampling the EEG signals to 250/256 Hz, applying temporal filters to extract oscillatory EEG activity in the 4--36 Hz range, extracting short segments ($\leq 3$s) associated with a class label, and finally obtaining $40 \times 40$ SPD covariance matrices.

\begin{algorithm}[t] \SetKwInOut{Input}{Input}\SetKwInOut{Output}{Output}\SetKwInOut{Parameters}{Parameters}
\caption{Momentum LieBN (MLieBN) Algorithm}
\label{alg:mliebn}
\Input{
A batch of activations $\{P_i\}_{i=1}^N$ over the Lie group $\{\calM, \oplus,g\}$, and a small positive constant $\epsilon$\\
running mean $\bar{M}_r=E$, running variance $\bar{v}^2_r=1$ for training\\
running mean $\tilde{M}_r=E$, running variance $\tilde{v}^2_r=1$ for testing\\
biasing parameter $B \in \calM$, scaling parameter $s \in \bbRscalar \setminus \{0\}$,\\
momentum parameters for training and testing $\eta_{\mathrm{train}}, \eta \in [0,1]$\\
}
\Output{Normalized activations $\{\tilde{P}_i\}_{i=1}^N$}
\BlankLine
\If{training}{
    Compute batch mean $M_b$ and variance $v_b^2$ of $\{P_i\}_{i=1}^N$;\\
    $\bar{M}_r \gets \wfm(\{1-\eta_{\mathrm{train}},\eta_{\mathrm{train}}\},\{\bar{M}_r,M_b\})$;\\
    $\bar{v}^2_r \gets (1-\eta_{\mathrm{train}})\bar{v}^2_r + \eta_{\mathrm{train}} v^2_b$;\\
    $\tilde{M}_r \gets \wfm(\{1-\eta,\eta\},\{\tilde{M}_r,M_b\})$;\\
    $\tilde{v}^2_r \gets (1-\eta)\tilde{v}^2_r + \eta v^2_b$;\\
}
\lIf{training}{$M \gets \bar{M}_r, v^2 \gets \bar{v}^2_r$} \lElse{$M \gets \tilde{M}_r, v^2 \gets \tilde{v}^2_r$}

\For{$i \gets 1$ \KwTo $N$}{
Centering to the neutral element $E$: \\
\Indp \lIf{$g$ is left-invariant}{$\bar{P}_i \gets \ltrans _{\ominus M}(P_i)$} \lElse{$\bar{P}_i \gets \rtrans _{\ominus M}(P_i)$} \Indm
Scaling the dispersion: \\
\hspace{1.5em}  $\hat{P}_i \gets \rieexp_{E} \left [ \frac{s}{\sqrt{v^2+\epsilon}} \rielog_{E}(\bar{P}_i) \right]$ \\
Biasing towards parameter $B$: \\
\Indp\lIf{$g$ is left-invariant}{$\tilde{P}_i \gets \ltrans _{B}(\hat{P}_i)$} \lElse{$\tilde{P}_i \gets \rtrans _{B}(\hat{P}_i)$} \Indm
}
\end{algorithm}

\mypara{Domain-specific momentum LieBN for EEG classification.} \citet{kobler2022spd} proposed SPDDSMBN as a domain adaptation approach for EEG classification.
SPDDSMBN, based on \cref{eq:kobler_rbn}, performed normalization of mean and variance on SPD manifolds under the specific AIM.
Additionally, SPDDSMBN utilized separate momentum parameters for updating training and testing running statistics, inspired by \citet{yong2020momentum}. Following \citet[Alg.~1]{kobler2022spd}, we also present a momentum LieBN (MLieBN) in \cref{alg:mliebn}. Here $\eta$ is fixed and $\eta_{\mathrm{train}}$ is defined as
\begin{equation}
    \eta_{\mathrm{train}}=1-\rho^{\frac{1}{T-1} \max \left(T-t, 0\right)}+\rho,
    \quad \text{where} \quad \rho=\frac{1}{\texttt{domains\_per\_batch}},
\end{equation}
where $T$ and $t$ denote the total number of training epochs and the current epoch, respectively. Furthermore, following \citet{kobler2022spd}, we adopt multi-channel mechanisms for domain-specific MLieBN (DSMLieBN), where each domain has its own MLieBN layer. Similar to \citet{kobler2022spd}, we set the biasing parameter equal to the neutral element, and the scaling factor is shared across all domains.
We denote \cref{alg:mliebn} as $\operatorname{MLieBN}(P_j \mid B, s, \epsilon, \eta,\eta_{\mathrm{train}})$.
Then our DSMLieBN follows
\begin{equation}
    \operatorname{DSMLieBN}(P_j,i)
    =\operatorname{MLieBN}_i(P_j \mid E, s, \epsilon, \eta,\eta_{\mathrm{train}}), \quad \forall P_j \in \{P_k\}_{k=1}^N,
\end{equation}
where $i$ is the index of the domain. We follow the official code of SPDDSMBN\footnote{\url{https://github.com/rkobler/TSMNet}} to implement our DSMLieBN. Thus, the only difference between DSMLieBN and SPDDSMBN is how normalization is performed. Analogous to \cref{thm:liebn_pullback}, computations for DSMLieBN under pullback metrics can also be performed by mapping, calculating, and then remapping.

\mypara{Implementation details.} We use the official code of SPDNetBN\footnote{\url{https://proceedings.neurips.cc/paper_files/paper/2019/file/6e69ebbfad976d4637bb4b39de261bf7-Supplemental.zip}} \citep{brooks2019riemannian} and TSMNet\footnote{\url{https://github.com/rkobler/TSMNet}} \citep{kobler2022spd} to implement our experiments on the SPDNet and TSMNet backbones. For the SPDNet architecture, we compare our LieBN with SPDNetBN \citep{brooks2019riemannian}, which applies the SPDBN (\cref{eq:spdnetbn_centering,eq:spdnetbn_biasing}) to SPDNet. Similar to SPDNetBN, we apply our LieBN after each transformation layer. In the EEG application, one of the state-of-the-art methods is TSMNet+SPDDSMBN \citep{kobler2022spd}, which is a domain adaptation version of \citet{kobler2022controlling}. For a fair comparison, we also implement a domain-specific momentum LieBN, referred to as DSMLieBN. Following \citet{kobler2022spd}, we apply our DSMLieBN before the LogEig layer in TSMNet. We use the standard cross-entropy loss and optimize the parameters with the Riemannian AMSGrad optimizer \citep{becigneul2019riemannian}. The network architectures are represented as $\{d_0, d_1, \ldots, d_L\}$, where the dimension of the parameter in the $i$-th BiMap layer is $d_i \times d_{i-1}$. The experiments are conducted with a learning rate of $5e^{-3}$, a batch size of 30, and 200 training epochs on the Radar, HDM05, and FPHA data sets. For the Hinss2021 data set, following \citet{kobler2022spd}, we use a learning rate of $1e^{-3}$ with a weight decay of $1e^{-4}$, a batch size of 50, and 50 training epochs.

\mypara{Scoring metrics.} In line with the previous work \citep{brooks2019riemannian,kobler2022spd}, we use accuracy as the scoring metric for the Radar, HDM05, and FPHA data sets, and balanced accuracy (\ie the average recall across classes) for the Hinss2021 data set. Ten-fold experiments on the Radar, HDM05, and FPHA data sets are carried out with randomized initialization and split (the split is officially fixed for the FPHA data set), while on the Hinss2021 data set, models are fit and evaluated with randomized leave-5\%-of-sessions-out (inter-session) or leave-5\%-of-subjects-out (inter-subject) cross-validation.

\subsubsection{LieBN on Rotation Matrices}
\label{app:subsec:impl_details_so3}

\mypara{Data sets and preprocessing.} The data set descriptions are collected in \cref{app:datasets}. Following \citet{huang2017deep}, the G3D experiments use the cross-subject setting, with half of the subjects used for training and the other half for testing, while the NTU60 experiments use the cross-view protocol \citep{shahroudy2016ntu}. We use the code\footnote{\url{https://ravitejav.weebly.com/kbac.html}} of \citet{vemulapalli2014human} to represent each skeleton sequence as a point on the Lie group $\soprod{N \times T}{3}$, where $N$ and $T$ denote spatial and temporal dimensions. As preprocessed in \citet{huang2017deep}, we set $T$ to 100, 16, and 64 on the G3D, HDM05, and NTU60 data sets, respectively.

\mypara{LieNet.} The LieNet backbone is reviewed in \cref{app:backbone-son}. Note that the official code of LieNet\footnote{\url{https://github.com/zhiwu-huang/LieNet}} is implemented in MATLAB. We use the open-source PyTorch code\footnote{\url{https://github.com/hjf1997/LieNet}} to implement our experiments. To reproduce LieNet more faithfully, we made the following modifications to this PyTorch code. We reimplemented the LogMap and RotPooling layers to make them consistent with the official MATLAB implementation. In addition, we extended the Riemannian computations of \texttt{Geoopt} \citep{kochurov2020geoopt} to $\so{3}$ to enable the direct Riemannian optimization, which is missing from the current package. We apply our LieBN before the LogMap layer. Note that the dimension of input features in LieNet is $B \times N \times T \times 3 \times 3$. We calculate Lie group statistics along the batch and temporal dimensions ($B \times T$). We denote the LieNet models with our LieBN-Left and LieBN-Right as LieNetLieBN-Left and LieNetLieBN-Right, respectively.

\mypara{Implementation details.} We find that SGD is the most effective optimizer for LieNet, and thus, we adopt it for our experiments. The learning rate is set to $1e^{-2}$. The batch sizes are 30, 30, and 256 for the G3D, HDM05, and NTU60 data sets, respectively. On the NTU60 data set, the learning rate is reduced by a factor of 10 upon model convergence, specifically at the 5th and 25th epochs for LieNetLieBN and LieNet, respectively. For each model, we apply \texttt{torch.nn.utils.clip\_grad\_norm\_} with \texttt{max\_norm=5} to the transformation matrix in the final FC layer.

\subsubsection{LieBN on Correlation Matrices}
\label{app:subsec:impl_details_cor}

We follow the same settings as the experiments on the SPD manifold with respect to the backbone architecture, batch size, number of training epochs, optimizer, and learning rate. The network architecture can be denoted as BiMap-[Power-Cov2Cor-LieBN-Cor]-LogEig, where Power denotes the matrix power and Cov2Cor is $\coropt(\cdot): \spd{n} \to \cor{n}$. The matrix powers used for each data set are presented in \cref{app:tab:liebncor_power}. A single iteration is sufficient to achieve saturated network performance with respect to calculating $\dplus$ and $\dstar$ in OLM and LSM, except for $\dstar$ on the HDM05 data set, which requires convergence with a maximum of 20 iterations.

\begin{table}[t]
  \centering
  \caption{Matrix powers in LieBN-Cor under different metrics on each data set.}
  \label{app:tab:liebncor_power}%
    \begin{tabular}{c|cccc}
    \toprule
    \diagbox{\textbf{Data Set}}{\textbf{Metric}}      & \textbf{ECM}   & \textbf{LECM}  & \textbf{OLM}   & \textbf{LSM} \\
    \midrule
    HDM05 & 0.75  & 0.5   & 0.5   & -0.5 \\
    FPHA  & -0.5  & -0.25 & -0.25 & -0.25 \\
    \bottomrule
    \end{tabular}%
\end{table}%

\subsection{Gyrogroup Batch Normalization}
\label{app:gyrobn-experimental-details}

\subsubsection{GyroBN on the Grassmannian}
\label{app:subsec:ch3-gyrobn-experiments_details_grass}

For HDM05, under-represented clips are removed, yielding 2,086 instances over 117 classes. For NTU60 and NTU120, we focus on mutual actions and adopt the cross-view and cross-setup protocols, respectively \citep{shahroudy2016ntu,liu2019ntu}. Following \citet{nguyen2023building}, each sequence is represented as a Grassmannian matrix of size $93 \times 10$, $150 \times 10$, and $150 \times 10$ for HDM05, NTU60, and NTU120, respectively. Because GyroBN is inserted after the first pooling layer, the corresponding inputs to GyroBN have sizes $47 \times 10$, $75 \times 10$, and $75 \times 10$, as reported in \cref{tab:grass_results}. The GyroGr backbone is reviewed in \cref{app:backbone-grassmannian}.

\mypara{Trivialization.} Following \citet{nguyen2023building}, we apply the trivialization strategy reviewed in \cref{sec:ch2-riemannian-optimization} to the Grassmannian parameters in the gyrotranslation and GyroBN layers. Each Grassmannian parameter $U \in \grasonb{p,n}$ is parameterized by a matrix $\frakU \in \bbR{(n-p) \times p}$ such that
\begin{equation}
    \left[\begin{array}{cc}
    \bbzero & -\frakU^\top \\
    \frakU & \bbzero
    \end{array}\right]=\left[\overline{UU^\top }, \idpp \right],
\end{equation}
where $\overline{(\cdot)}=\rielog_{\idpp}(\cdot)$. The parameter $U$ can be retrieved by
\begin{equation}
    \label{app:eq:trivilization_grass}
    U = \mexp \left(\left[\overline{UU^\top }, \idpp \right]\right) \idonb =\mexp \left(\left[\begin{array}{cc}
    \bbzero & -\frakU^\top \\
    \frakU & \bbzero
    \end{array}\right]\right) \idonb.
\end{equation}
This reparameterization represents the trainable Grassmannian parameters by Euclidean coordinates, thereby allowing the direct use of PyTorch optimizers \citep{paszke2019pytorch} and avoiding direct Riemannian updates of these parameters.

\subsection{Riemannian Multinomial Logistic Regression}
\label{rmlr:app:experimental-details}

\subsubsection{RMLR on the SPD Manifold}
\label{rmlr:app:sec:exp_details}

This subsection offers additional details on the experiments on SPD MLRs.

\mypara{Backbone networks and LogEig MLR.} The SPDNet, TSMNet, SPDNetBN, SPDDSMBN, and SPDGCN backbones are reviewed in \cref{app:backbone-spd}, while RResNet is reviewed in \cref{app:backbone-others}. In the SPD baseline models considered here, the Euclidean MLR in the codomain of matrix logarithm (matrix logarithm + FC + softmax) is used for classification.
Following the terminology introduced in \cref{spdmlr:sec:logeig}, we call this classifier the \textbf{LogEig MLR}.
The LogEig MLR is the Euclidean classifier in the tangent space at the identity, which might distort the innate geometry of the SPD manifold.

\mypara{SPD modeling and preprocessing.} We use the same SPD modeling and preprocessing as the LieBN experiments in \cref{app:subsec:exp_details_spd}.

\mypara{Implementation details.} For SPDNet \citep{huang2017riemannian} and TSMNet \citep{kobler2022spd}, we follow the official PyTorch code of
SPDNetBN\footnote{\url{https://proceedings.neurips.cc/paper_files/paper/2019/file/6e69ebbfad976d4637bb4b39de261bf7-Supplemental.zip}}
and
TSMNet\footnote{\url{https://github.com/rkobler/TSMNet}} to implement our experiments.
To evaluate the performance of our intrinsic classifiers, we substitute the LogEig MLR in SPDNet and TSMNet with our SPD MLRs.
We implement our SPD MLRs induced by five parameterized metrics.
On the Radar and HDM05 data sets, the learning rate is $10^{-2}$, and the batch size is 30.
On the Hinss2021 data set, following \citet{kobler2022spd}, the learning rate is $10^{-3}$ with a $10^{-4}$ weight decay, and the batch size is 50.
The maximum numbers of training epochs are 200, 200, and 50, respectively.
We use the standard cross-entropy loss as the training objective and optimize the parameters with the Riemannian AMSGrad optimizer \citep{becigneul2019riemannian}.

\mypara{RResNet \citep{katsman2023riemannian}.}  We focus on the AIM-based RResNet and use the official code\footnote{\url{https://github.com/CUAI/Riemannian-Residual-Neural-Networks}} and suggested network settings to implement the experiments with RResNet.
We conduct 10-fold and 5-fold experiments on the HDM05 and NTU60 data sets, respectively.
Since RResNet is developed based on SPDNet, we use the same learning settings as SPDNet for the action recognition task and borrow the best $(\theta,\alpha,\beta)$ from \cref{rmlr:tb:results_hdm05} for our SPD MLRs under the RResNet backbone.

\mypara{NTU60 SPD modeling.} For NTU60 \citep{shahroudy2016ntu}, each frame contains the 3D coordinates of 25 body joints and is therefore represented by a $25 \times 3=75$-dimensional coordinate vector. Following \citet{katsman2023riemannian}, each sequence is modeled as a $75 \times 75$ temporal covariance matrix, and evaluation follows the cross-view protocol \citep{shahroudy2016ntu}.

\mypara{SPDGCN \citep{zhao2023modeling}.} We use the official code\footnote{\url{https://github.com/andyweizhao/SPD4GNNs}} and the suggested network settings in \citet{zhao2023modeling}. Note that SPDGCN with SPD MLR retains the same network settings as vanilla SPDGCN.
\cref{rmlr:tab:hyper_spdgcn} presents the hyperparameters $(\theta,\alpha,\beta)$ on different data sets.

\begin{table}[t]
  \centering
  \caption{$(\theta,\alpha,\beta)$ of SPD MLRs on the SPDGCN backbone.}
  \label{rmlr:tab:hyper_spdgcn}%
  \resizebox{\linewidth}{!}{
    \begin{tabular}{cccccc}
    \toprule
    \textbf{Data Sets} & $\triparamAIM$ & $\triparamEM$ & $\biparamLEM$ & $\paramBWM$ & $\paramLCM$ \\
    \midrule
    Disease & (0.25,1,0) & (0.25,1,0) & (1,1) & 0.25  & 0.5 \\
    Cora  & (0.5,1,0) & (0.25,1,$\nicefrac{1}{9}$) & (1,$\nicefrac{1}{9}$) & 0.25  & 0.5 \\
    Pubmed & (0.5,1,0) & (0.5,1,0) & $(1,-\nicefrac{1}{3})$ & 0.25  & 0.5 \\
    \bottomrule
    \end{tabular}%
  }
\end{table}%

\mypara{Network architectures.} We denote the network architecture as $[d_0, d_1,\cdots,d_L]$, where the dimension of the parameter in the $i$-th BiMap layer (\cref{app:backbone-spd}) is $d_{i} \times d_{i-1}$.
For SPDNet, we also validate our SPD MLRs under different network architectures on the Radar and HDM05 data sets.
The network architectures on the Radar data set are $[20,16,8]$ for the 2-block configuration and $[20,16,14,12,10,8]$ for the 5-block configuration, while on the HDM05 data set, the network architectures are $[93,30]$ for 1-block, $[93,70,30]$ for 2-block, and $[93,70,50,30]$ for 3-block.
For TSMNet, the 1-block architecture is $[40,20]$.

\mypara{Scoring metrics and evaluation protocols.} We use the same scoring metrics and evaluation protocols as \cref{app:subsec:exp_details_spd}. For the graph-learning experiments, following \citet{zhao2023modeling}, we report the 10-fold average and maximum node-classification accuracy.

\mypara{Hyperparameters.} We implement the SPD MLRs induced by not only five standard metrics, \ie LEM, AIM, EM, LCM, and BWM, but also five families of parameterized metrics.
Therefore, in our SPD MLRs, we have a maximum of three hyperparameters, \ie $\theta,\alpha,\beta$, where $\alphabeta$ are associated with $\orth{n}$-invariance and $\theta$ controls deformation.
For $\alphabeta$ in $(\theta,\alpha,\beta)$-LEM, $(\theta,\alpha,\beta)$-AIM, and $(\theta,\alpha,\beta)$-EM, recalling \cref{eq:ch2-spd-alpha-beta-inner}, $\alpha$ is a scaling factor, while $\beta$ measures the relative significance of traces.
As scaling is less important \citep{thanwerdas2019affine}, we set $\alpha=1$.
As for the value of $\beta$, we select it from a predefined set: $\{1,\nicefrac{1}{n},\nicefrac{1}{n^2},0,-\nicefrac{1}{n}+\epsilon,-\nicefrac{1}{n^2}\}$, where $n$ is the dimension of the input SPD matrices in SPD MLRs.
The purpose of including $\epsilon \in \bbRplus$ is to ensure the positive definiteness of the inner product, \ie $\alpha+n\beta>0$ and hence $\alphabeta \in \bfst$.
These chosen values for $\beta$ allow for amplifying, neutralizing, or suppressing the trace components, depending on the characteristics of the data sets.
For the deformation factor $\theta$, we roughly select its value around its deformation boundary, \ie $[0.25,1.5]$ for $(\theta,\alpha,\beta)$-AIM, $[0.5,1.5]$ for $\theta$-LCM, $[0.25,1.5]$ for $(\theta,\alpha,\beta)$-EM, and $[0.25,0.75]$ for $2\theta$-BWM.
The detailed values are listed in \cref{rmlr:tab:hyper_param_values}.

\begin{table}[t]
    \centering
    \caption{Candidate values for hyperparameters in SPD MLRs.}
    \label{rmlr:tab:hyper_param_values}%
    \resizebox{\linewidth}{!}{
    \begin{tabular}{ccccc}
    \toprule
    \textbf{Metric} & $(\theta,\alpha,\beta)$-AIM & $(\theta,\alpha,\beta)$-EM & $\theta$-LCM & $2\theta$-BWM \\
    \midrule
    \textbf{Candidate Values} & $\{0.25,0.5,0.75,1,1.25,1.5\}$ & $\{0.25,0.5,1,1.5\}$ & $\{0.5,1,1.5\}$ & $\{0.25,0.5,0.75\}$ \\
    \bottomrule
    \end{tabular}%
    }
\end{table}%

\subsubsection{RMLR on Rotation Matrices}

\mypara{LieNet backbone.} The LieNet backbone is reviewed in \cref{app:backbone-son}. In the official MATLAB implementation, the LogMap layer uses the Euler axis--angle representation.
Classification is performed using the Euler axis--angle representation, followed by an FC layer and a softmax layer.
As the axis--angle representation is equivalent to the matrix logarithm, we call this classifier \textbf{LogEig MLR} as well.
This classifier is, therefore, also non-intrinsic.

\begingroup
\tolerance=9999
\emergencystretch=4pt
\mypara{Preprocessing.} We use the shared rotation-matrix modeling, preprocessing, LieNet implementation, and SGD optimizer choice described in \cref{app:subsec:impl_details_so3}. Following \citet{huang2017deep}, the Lie MLR experiments use the G3D and HDM05 data sets. We trim HDM05 by removing under-represented sequences, resulting in 2,326 sequences across 122 classes.
\par
\endgroup

\begingroup
\tolerance=9999
\emergencystretch=4pt
\mypara{Lie MLR.} We use our Lie MLR to replace the axis--angle classifier in LieNet and call the resulting network LieNet+LieMLR.
To alleviate the computational burden, we set each $P_k$ to have shape $[\mathrm{num},3,3]$, where $\mathrm{num}$ is the spatial dimension of the input of the Lie MLR layer.
In other words, $P_k$ is shared in the temporal dimension.
We adopt PyTorch3D \citep{ravi2020accelerating} to calculate the matrix logarithm.
Due to the instability of \path{pytorch3d.transforms.so3_log_map}, we first use \path{pytorch3d.transforms.matrix_to_axis_angle} to calculate the rotation axis and angle and then convert this representation into the matrix logarithm\footnote{\url{https://github.com/facebookresearch/pytorch3d/issues/188}}.
\par
\endgroup

\mypara{Training details.} Following \citet{huang2017deep}, we focus on the 3-block and 2-block architectures for the G3D and HDM05 data sets, respectively, which are the suggested architectures for these two data sets.
The learning rate is $10^{-2}$ on both data sets, and we further set the weight decay to $10^{-5}$ on the G3D data set.
For LieNet and LieNet+LieMLR, we use \path{torch.nn.utils.clip_grad_norm_} for gradient clipping with a clipping factor of 5.
The clipping is imposed on the dimensionality reduction weight in the final FC linear layer of LieNet or, accordingly, $A=\{A_1,\ldots,A_C\}$ in the Lie MLR layer of LieNet+LieMLR.

\mypara{Scoring metrics.} For the G3D data set, following LieNet \citep{huang2017deep}, we adopt a 10-fold cross-subject test setting, where half the subjects are used for training and the other half are employed for testing.
For the HDM05 data set, following \citet{huang2017deep}, we randomly select half of the sequences for training and the rest for testing.
Due to the instability of LieNet, we conduct 20-fold experiments and select the best 10 folds to evaluate the performance.

\subsection{Proper Velocity Neural Networks}
\label{pvnn:app:experimental-details}

\mypara{Common Implementations.} We use the trivialization strategy reviewed in \cref{sec:ch2-riemannian-optimization} in our MLR, FC, and GyroBN layers. Consequently, all trainable manifold-valued parameters in PVNN are represented by Euclidean coordinates and optimized using standard Euclidean optimizers.

\subsubsection{Image Classification}
\label{pvnn:app:subsec:image-classification-exp-details}

The CIFAR-10 and CIFAR-100 data sets and their PyTorch splits are described in \cref{app:subsec:image-classification-datasets}. Following \citet{bdeir2024fully}, we use data augmentation that includes random cropping with padding of 4 pixels and random horizontal flipping.

\mypara{Implementation Details.} We implement the experiments using the official code\footnote{\url{https://github.com/kschwethelm/HyperbolicCV}} of \citet{bdeir2024fully}. All models share a common backbone, which consists of a ResNet-18 encoder followed by a hyperbolic MLR classifier. Except for PV MLR without $\rieexp_{\zerovec}$, the output embedding of the ResNet-18 backbone is mapped to the target hyperbolic space via the exponential map at the identity $e$, that is, $\rieexp_e\left(x\right)$. Here, $e=\zerovec$ for the Poincar\'e and PV spaces, and $e=\Lzero$ for Lorentz. All models are trained from scratch. Optimization is performed using SGD \citep{robbins1951stochastic} with an initial learning rate of $0.1$, a momentum of $0.9$, and a weight decay of $5\times 10^{-4}$. Training is conducted with a batch size of $128$ for $200$ epochs. The learning rate is decayed by a factor of $\gamma = 0.2$ at epochs $60$, $120$, and $160$. The curvature for the PV space is set as $K = -0.5$.

\subsubsection{Graph Learning}
\label{pvnn:app:subsec:graph-learning-exp-details}

The descriptions and summary statistics of the Disease, Airport, PubMed, and Cora data sets are collected in \cref{app:subsec:datasets_ccs}.

\mypara{Implementation Details.}\label{pvnn:app:impl-details} We adopt the official code of HGCN\footnote{\url{https://github.com/HazyResearch/hgcn}} \citep{chami2019hyperbolic} to conduct experiments. The features of each node are embedded into the hyperbolic space via the exponential map at the identity. The hyperbolic network consists of two FC layers: the first maps the input feature dimension to a $16$-dimensional hidden representation, and the second maps from $16$ to $16$. Each FC layer is followed by an activation function. An MLR layer is then used for classification. All models are trained using the Adam optimizer \citep{kingma2015adam}. We evaluate performance every $10$ epochs and employ early stopping with a patience of $200$ evaluations, restoring the checkpoint with the best test accuracy. \cref{pvnn:tab:pv-hparams-dataset,pvnn:tab:pv-hparams-common} summarize the hyperparameters for PVNN. For KNN \citep{mao2024klein}, HNN \citep{ganea2018hyperbolic}, HNN++ \citep{shimizu2021hyperbolic}, and LNN \citep{bdeir2024fully}, we follow their original papers to implement the experiments. \cref{pvnn:tab:graph-model-architecture} summarizes the hyperbolic layers used in each model.

\begin{table}[t]
\centering
\caption{Hyperparameters for PVNN that vary across graph data sets.}
\label{pvnn:tab:pv-hparams-dataset}
\begin{tabular}{ccccc}
\toprule
\textbf{Hyperparameter} & \textbf{Disease} & \textbf{Airport} & \textbf{PubMed} & \textbf{Cora} \\
\midrule
Learning rate & 0.01 & 0.01 & 0.05 & 0.05 \\
Dropout       & 0.4  & 0.4  & 0.6  & 0.6  \\
Curvature     & -0.3  & -0.3  & -1.0  & -1.0 \\
\bottomrule
\end{tabular}
\end{table}

\begin{table}[t]
\centering
\caption{Hyperparameters for PVNN that are shared across graph data sets.}
\label{pvnn:tab:pv-hparams-common}
\begin{tabular}{lccc}
\toprule
\textbf{Setting} & \textbf{Epochs} & \textbf{Batch size} & \textbf{Weight decay}  \\
\midrule
Value & 2000 & 128 & $5\times10^{-4}$  \\
\bottomrule
\end{tabular}
\end{table}

\begin{table}[t]
\centering
\caption{Summary of the hyperbolic layers used in the graph node classification models.}
\label{pvnn:tab:graph-model-architecture}
\resizebox{\linewidth}{!}{
\begin{tabular}{cccc}
\toprule
\textbf{Model} & \textbf{FC layer} & \textbf{Activation} & \textbf{MLR} \\
\midrule
PVNN & PV FC in \cref{pvnn:thm:pv-fc} & $\rieexp_{\zerovec}\left(\sigma\left(\rielog_{\zerovec}(x)\right)\right)$ & PV MLR in \cref{pvnn:thm:pv-mlr} \\
KNN & $\rielog_{\zerovec}(W\rieexp_{\zerovec}(x))$ & $\rieexp_{\zerovec}\left(\sigma\left(\rielog_{\zerovec}(x)\right)\right)$ & Euclidean MLR after $\rieexp_{\zerovec}$ \\
HNN & $\rielog_{\zerovec}(W\rieexp_{\zerovec}(x))$ & $\rieexp_{\zerovec}\left(\sigma\left(\rielog_{\zerovec}(x)\right)\right)$ & Poincaré MLR \citep{ganea2018hyperbolic} \\
HNN++ & Poincaré FC \citep{shimizu2021hyperbolic} & $\rieexp_{\zerovec}\left(\sigma\left(\rielog_{\zerovec}(x)\right)\right)$ & Poincaré MLR \citep{shimizu2021hyperbolic} \\
LNN & Lorentz FC \citep{chen2022fully} & Lorentz activation \citep{bdeir2024fully} & Lorentz MLR \citep{bdeir2024fully} \\
\bottomrule
\end{tabular}
}
\end{table}

\subsubsection{Genomic Sequence Learning}
\label{pvnn:app:subsec:genomic-sequence-exp-details}

The description and complete summary of TEB are collected in \cref{app:subsec:genomic-sequence-datasets}. We focus on five of its seven data sets.

\mypara{Implementation Details.} For the Euclidean CNN and the hyperbolic CNN baseline (HCNN-S), we directly use the results reported in the original paper \citep[Tab.~2]{khan2025hyperbolic}. Our PVCNN architecture follows their implementation\footnote{\url{https://github.com/rrkhan/HGE}}. Each DNA sequence is represented as a length-$L$ sequence with 4 input channels. We first apply a PV convolution that maps the 4 input channels to 32 channels, followed by PV TBN and a tanh tangent activation. A second PV convolution layer is then applied. The final PV feature is concatenated and passed through an FC layer, and finally classified with a PV MLR head. The curvature is initialized at $K=-0.5$ and learned during training. We train for 100 epochs with a step learning-rate schedule, using milestones at epochs 60 and 85 with a decay factor of 0.1. For the PV FC layer, $\sigma$ in \cref{pvnn:eq:pv-fc-activation} is set to $\tanh$. All other hyperparameters are summarized in \cref{pvnn:tab:hcnn-s-hparams}.

\begin{table}[t]
\centering
\caption{Hyperparameters for TEB.}
\label{pvnn:tab:hcnn-s-hparams}
\begin{tabular}{lc}
\toprule
\textbf{Setting} & \textbf{Value} \\
\midrule
Optimizer        & Adam \\
Learning rate   & $1e^{-4}$ \\
Weight decay     & $2e^{-2}$ \\
Batch size       & $100$ \\
Dropout          & $0.1$ \\
Adam $(\beta_1,\beta_2)$ & $(0.9, 0.999)$ \\
\bottomrule
\end{tabular}
\end{table}

\subsection{Hyperbolic Busemann Neural Networks}
\label{hbnn:app:experimental-details}

\subsubsection{Image Classification}
\label{hbnn:app:subsubsec:image-classification-details}
\begin{table}[t]
	\centering
	\caption{Summary of hyperparameters used in the image classification task.}
	\label{hbnn:app:tab:hyperparameters-cl}
	\begin{tabular}{ccc}
    \toprule
	\textbf{Hyperparameter} & \makecell{\textbf{CIFAR-10/100} \\ \textbf{Tiny-ImageNet}}  & \textbf{ImageNet-1k}\\
    \midrule
    {Epochs} & 200  & 100\\
    {Batch size} & 128 & 256\\
    {Initial learning rate} & $0.1$ & $0.1$\\
    {LR schedule} & $60, 120, 160; \gamma=0.2$ & $30, 60, 90; \gamma=0.1$\\
    {Weight decay} & $5e^{-4}$ & $1e^{-4}$\\
    {Optimizer} & SGD & SGD\\
    {Precision} & 32-bit & 32-bit\\
    {Curvature $K$} & $-1$ & $-1$\\
    \bottomrule
	\end{tabular}
\end{table}

\mypara{Implementation Details.} For CIFAR-10/100 and Tiny-ImageNet, we follow \citet[App.~C.1]{bdeir2024fully}. For ImageNet-1k, we follow \citet[Sec.~4]{guo2022clipped}. \cref{hbnn:app:tab:hyperparameters-cl} summarizes the data-set-specific hyperparameters. For the hyperbolic MLR, before mapping into the hyperbolic space, we clip the feature vector by
\begin{equation}\label{hbnn:app:eq:clip-hyper}
  \operatorname{CLIP}\left(x ; r\right)=\min \left\{1, \frac{r}{\norm{x}}\right\} x
\end{equation}
where $r>0$ is a hyperparameter. The clipped Euclidean embedding is projected via the exponential map to the target hyperbolic space: $\rieexp_e\left(\operatorname{CLIP}(x ; r)\right)$. For the Lorentz model, the clipping parameter is $r=1$ on CIFAR-10/100 and $r=4$ on Tiny-ImageNet and ImageNet-1k. On the Poincaré ball, $r=1$ on all four data sets.

All methods are implemented in PyTorch and trained with cross-entropy loss. The results of MLR, PMLR, and LMLR on CIFAR-10/100 and Tiny-ImageNet are copied from \citet[Tab.~1]{bdeir2024fully}, while those of PBMLR-P on CIFAR-10/100 are copied from \citet[Tab.~2]{nguyen2025neural}. The remaining results are obtained by our careful implementation.

\subsubsection{Genome Sequence Learning}
\label{hbnn:app:subsubsec:genome-sequence-learning-details}
\mypara{Implementation Details.} We mainly follow the official implementations of \citet{khan2025hyperbolic} for data processing, model architecture, and training. We adopt a simple CNN with three convolutional blocks followed by dense, ReLU-activated layers to extract features \citep[Fig.~4]{khan2025hyperbolic}. Before classification, the features are clipped and mapped to the target hyperbolic space as in \cref{hbnn:app:eq:clip-hyper}, then passed to either a prior hyperbolic MLR or our BMLR head. The clipping factor defaults to $r=1$, with the following exceptions on GUE: on the Lorentz model, $r=2.0$ for Covid variant classification and $r=5.0$ for species classification; on the Poincar\'e ball, $r=2.0$ for species classification. Following \citet{khan2025hyperbolic}, we treat the curvature $K$ as a learnable parameter initialized as $-1$. The remaining hyperparameters are listed in \cref{hbnn:app:tab:genome-hyperparams}. All models share these hyperparameters, except LMLR on Covid variant classification, for which we set the weight decay to $1e^{-3}$ to ensure convergence.

All methods are implemented in PyTorch and trained with cross-entropy loss. Results are obtained from our reimplementation.

\begin{table}[t]
\centering
\caption{Hyperparameters for genome sequence learning.}
\label{hbnn:app:tab:genome-hyperparams}
\begin{tabular}{cc}
\toprule
{Batch size} & 100 \\
{Epochs} & 100 \\
{Optimizer} & Adam \\
$\beta_1$, $\beta_2$ & 0.9, 0.999 \\
{Initial learning rate} & $1e^{-4}$\\
{LR schedule} & $60, 85$; $\gamma=0.1$ \\
{Weight decay} & $0.1$ \\
{Initial curvature $K$} & $-1$ \\
\bottomrule
\end{tabular}
\end{table}

\subsubsection{Node Classification}
\label{hbnn:app:subsubsec:node-classification-details}

\mypara{Implementation Details.} We follow the official implementations of HGCN \citep{chami2019hyperbolic} and PBMLR \citep{nguyen2025neural} and conduct experiments on both the Poincaré ball and the Lorentz model. We adhere to their experimental settings. The only changes are weight decay and dropout. We train with cross-entropy loss and the Adam optimizer \citep{kingma2015adam} for 5000 epochs with a learning rate of $1e^{-2}$, curvature set to $-1$, embedding dimension 16, and three GCN layers. We tune weight decay and dropout and report the values in \cref{hbnn:app:tab:node-classification-hyperparams}. All methods are implemented in PyTorch. The results of HGCN-PMLR and HGCN-PBMLR-P on the Poincaré ball are taken from \citet[Tab.~11]{nguyen2025neural}. Results for the remaining baselines are obtained from our reimplementation following the original settings.

\begin{table}[t]
  \centering
  \caption{Hyperparameters for node classification on Disease, Airport, PubMed, and Cora.}
  \label{hbnn:app:tab:node-classification-hyperparams}%
    \begin{tabular}{c|cc}
    \toprule
	    \textbf{Space} & \textbf{Weight decay} & \textbf{Dropout} \\
    \midrule
    $\pball{n}$ & $1e-4, 1e-5, 1e-3,1e-3$ & $0.3, 0, 0,0.2$ \\
    $\lorentz{n}$ & $1e-4, 5e-5, 1e-3,1e-3$ & $0, 0, 0,0.3$ \\
    \bottomrule
    \end{tabular}%
\end{table}%

\subsubsection{Link Prediction}
\label{hbnn:app:subsubsec:link-prediction-details}

\mypara{Implementation Details.} We follow the official implementations of HNN \citep{ganea2018hyperbolic}, HNN++ \citep{shimizu2021hyperbolic}, and HyboNet \citep{chen2022fully}, and adopt the experimental protocol of \citet{chami2019hyperbolic} for link prediction. The encoder consists of two fully connected layers: the first maps the input features to 16, and the second maps 16 to 16. Each FC layer is instantiated as either our BFC or an existing hyperbolic FC layer. After each FC, we apply the activation $\rieexp_e\left(\operatorname{ReLU}(\rielog_e(x))\right)$, where $e$ denotes the model origin. Following \citet{chami2019hyperbolic}, this activation is disabled on Cora. As in the M\"obius layer, we apply a gyro bias after each FC, that is, $x \Hoplus b$. We train with Adam \citep{kingma2015adam} at a learning rate of $1e^{-2}$ and tune weight decay and FC dropout. For BFC, we use $\phi=\tanh$ on Airport and Cora and the identity map on the other two data sets.

\subsection{Full-Rank Correlation Networks}
\label{cornet:app:experimental-details}

The Radar, HDM05, FPHA, and NTU120 data sets are described in \cref{app:datasets}. For HDM05 and FPHA, each sequence is normalized for body-part length, scale, and view using the preprocessing of \citet{vemulapalli2014human}. For NTU120, we follow \citet{chen2021channel} to preprocess the data.

\subsubsection{Input Data}
\label{cornet:app:subsubsec:input-data}
\mypara{Correlation Input in CorNets.} Following \citet{wang2024grassatt,nguyen2024matrix}, we first model each sample as a multi-channel SPD tensor. For HDM05 and FPHA, the skeleton preprocessing and multi-channel covariance construction follow the GyroGr pipeline of \citet{nguyen2023building}; CorNet subsequently converts every covariance matrix into a full-rank correlation matrix by
\begin{equation}
\coropt:\spd{n}\ni\Sigma\longmapsto C=\bbD(\Sigma)^{-\frac{1}{2}}\Sigma\bbD(\Sigma)^{-\frac{1}{2}}\in\cor{n}.
\end{equation}
For Radar, we follow \citet{wang2024grassatt} and use temporal convolution followed by covariance pooling to obtain a multi-channel covariance tensor of shape $[c,20,20]$. After preprocessing, the input correlation tensor shapes are $[7,20,20]$, $[3,28,28]$, $[9,28,28]$, and $[6,28,28]$ on Radar, HDM05, FPHA, and NTU120, respectively.

\subsubsection{Implementation Details}
\label{cornet:app:subsubsec:implementation-details}

\begin{table}[t]
  \centering
  \caption{Hyperparameters in CorNets.}
  \label{cornet:app:tab:hyperparameters}%
  \resizebox{0.85\linewidth}{!}{
    \begin{tabular}{c|c|ccccc}
    \toprule
    \textbf{Data Set} & \textbf{Model} & \textbf{Optimizer} & \textbf{lr} & \textbf{wd} & \textbf{Matrix Power} & \textbf{Converged Epoch}\\
    \midrule
    \multirow{5}[2]{*}{Radar} & CorNet-ECM & Adam  & $1e^{-2}$ & \na & 1.5 & 50\\
          & CorNet-LECM & Adam  & $1e^{-2}$ & \na & -0.25 & 50\\
          & CorNet-OLM & Adam  & $1e^{-2}$ & \na & -0.25 & 50\\
          & CorNet-LSM & Adam  & $1e^{-2}$ & \na & 0.75 & 50\\
          & CorNet-PHCM & Adam  & $1e^{-2}$ & \na & 0.75 & 50\\
    \midrule
    \multirow{5}[2]{*}{HDM05} & CorNet-ECM & Adam  & $1e^{-3}$ & $1e^{-3}$ & 0.125 &100 \\
          & CorNet-LECM & Adam  & $1e^{-4}$ & $1e^{-3}$ & 0.5 & 150 \\
          & CorNet-OLM & SGD   & $5e^{-2}$ & $1e^{-3}$ & 0.25 & 200 \\
          & CorNet-LSM & Adam  & $1e^{-3}$ & \na & -0.75 & 50\\
          & CorNet-PHCM & Adam  & $1e^{-2}$ & \na & -0.25 & 50\\
    \midrule
    \multirow{5}[2]{*}{FPHA} & CorNet-ECM & Adam  & $5e^{-3}$ & \na & -0.25 & 150 \\
          & CorNet-LECM & Adam  & $5e^{-4}$ & $1e^{-4}$ & -0.5 & 150 \\
          & CorNet-OLM & Adam  & $1e^{-4}$ & \na & -1 & 50\\
          & CorNet-LSM & Adam  & $1e^{-3}$ & \na & -1 & 50\\
          & CorNet-PHCM & Adam  & $1e^{-3}$ & $1e^{-4}$ & -0.5 & 150 \\
    \midrule
    \multirow{5}[2]{*}{NTU120} & CorNet-ECM & SGD   & $1e^{-2}$ & \na     & 0.25  & 50 \\
          & CorNet-LECM & SGD   & $1e^{-2}$ & \na     & 0.25  & 50 \\
          & CorNet-OLM & SGD   & $5e^{-3}$ & \na     & 0.25  & 50 \\
          & CorNet-LSM & SGD   & $1e^{-3}$ & \na    & 0.25  & 50 \\
          & CorNet-PHCM & Adam  & $1e^{-3}$ & \na     & 0.25  & 50 \\
    \bottomrule
    \end{tabular}
    }
\end{table}

\mypara{SPD Baselines.} We follow the official PyTorch code of
SPDNetBN\footnote{\url{https://proceedings.neurips.cc/paper\_files/paper/2019/file/6e69ebbfad976d4637bb4b39de261bf7-Supplemental.zip}} to implement SPDNet and SPDNetBN. 
For LieBN\footnote{\url{https://github.com/GitZH-Chen/LieBN}}, we focus on the instantiation under LCM \citep{lin2019riemannian}, while for RResNet\footnote{\url{https://github.com/CUAI/Riemannian-Residual-Neural-Networks}}, we implement the ones induced by AIM \citep{pennec2006riemannian} and LEM \citep{arsigny2005fast}. For SPD MLR\footnote{\url{https://github.com/GitZH-Chen/SPDMLR}}, we implement the one based on LCM. Due to the lack of official code, Gyro-based models are carefully reimplemented from their original papers. Following \citet{nguyen2024matrix}, GyroSPD++ combines an AIM-based convolution with an LEM-based MLR.

\mypara{Grassmannian Baselines.} Since GrNet is officially implemented in MATLAB, we carefully re-implemented it using PyTorch. Additionally, as both GyroGr and GyroGr-Scaling do not release official code, we re-implemented them based on the original paper \citep{nguyen2023building}. For all Grassmannian comparative methods, we use SGD \citep{robbins1951stochastic} with a learning rate of $5e^{-2}$.

\mypara{CorNets.} On all four data sets, we employ a single convolutional kernel for global convolution, \ie applying a global receptive field across the channel dimension. The output dimensions of the correlation convolutional layer are $8 \times 8$, $26 \times 26$, $26 \times 26$, and $11 \times 11$ for the Radar, HDM05, FPHA, and NTU120 data sets, respectively.

We primarily use the Adam \citep{kingma2015adam} and SGD \citep{robbins1951stochastic} optimizers. Inspired by the deformation effect on the latent SPD geometries by the matrix power over the SPD manifold illustrated in \cref{rmlr:fig:illustration_metrics}, we apply the matrix power before correlation modeling ($\coropt(\cdot)$) as activation. In particular, when the data are centered at zero and power is $-1$, $\coropt(\Sigma^{-1})$ corresponds to the partial correlation matrix of the covariance matrix $\Sigma$ \citep[Lem.~1.6]{thanwerdas2024permutation}. The batch size is set to 30, and training is capped at 200 epochs, although most cases converge in fewer than 150 epochs. Due to the different correlation geometries, the hyperparameters vary for CorNets under different geometries. \cref{cornet:app:tab:hyperparameters} summarizes all the hyperparameters.

\mypara{Extra Computational Details for OLM and LSM Layers.} For the MLR, FC, and convolutional layers induced by OLM and LSM, the key computations involve $\offexp$ and $\logscaled$, which depend on the calculations of $\dplus$ and $\dstar$. In our experiments, we empirically observe that iterating until convergence is more effective for $\dplus$, whereas a single step of Newton's method generally performs best for $\dstar$. Accordingly, we set $\dplus$ to iterate until convergence, leveraging \cref{cornet:prop:gradients-d-plus} for accurate backpropagation. For $\dstar$, we adopt a single iteration in Newton's method and use automatic differentiation (autograd) through this single step for backpropagation.

\subsubsection{Additional Details on Visualization}
\label{cornet:app:subsubsec:visualization-details}
We provide additional interpretations of the low-dimensional visualization and clarify how the decision hyperplanes in \cref{cornet:fig:hyperplanes} are obtained. The correlation-in-SPD visualization is discussed in \cref{sec:ch2-full-rank-correlation-manifolds}.

\mypara{Visualization of Low-Dimensional SPD and Correlation Matrices.} Any $2 \times 2$ covariance matrix in $\spd{2}$ can be written as
\begin{equation}
    \Sigma =
    \begin{pmatrix}
        a & b \\
        b & d
    \end{pmatrix},
    \qquad
    a>0,\ d>0,\ ad-b^{2}>0.
\end{equation}
Embedding $\Sigma$ into $\bbR{3}$ via the map $\Sigma \mapsto (a,b,d)$ identifies $\spd{2}$ with the interior of the quadratic cone
\begin{equation}
    \left\{ (a,b,d) \in \bbR{3} \mid a>0,\ d>0,\ ad-b^{2}>0 \right\},
\end{equation}
which is an open cone in $\bbR{3}$.  

For $2 \times 2$ correlation matrices, any $C \in \cor{2}$ has the form
\begin{equation}
    C =
    \begin{pmatrix}
        1 & r \\
        r & 1
    \end{pmatrix},
    \qquad
    r \in (-1,1).
\end{equation}
Thus, $\cor{2}$ is one-dimensional. Embedding $C$ into $\bbR{3}$ as $(1,r,1)$ yields a line segment inside the cone corresponding to $\spd{2}$.

For $3 \times 3$ correlation matrices, any $C \in \cor{3}$ is parameterized by its off-diagonal entries $(r_{12},r_{13},r_{23})$:
\begin{equation}
    C =
    \begin{pmatrix}
        1      & r_{12} & r_{13} \\
        r_{12} & 1      & r_{23} \\
        r_{13} & r_{23} & 1
    \end{pmatrix}.
\end{equation}
Embedding $C$ into $\bbR{3}$ via $C \mapsto (r_{12},r_{13},r_{23})$ produces an open elliptope in $\bbR{3}$. This is the representation of $\cor{3}$ used in \cref{cornet:fig:hyperplanes}, where each point in the elliptope corresponds to one $3 \times 3$ correlation matrix.

\mypara{Construction of \cref{cornet:fig:hyperplanes}.} For ECM, LECM, OLM, and LSM, the decision hyperplane in the correlation MLR is the Riemannian hyperplane in \cref{rmlr:subsec:re_exist_MLR} specialized to $\calM = \cor{n}$:
\begin{equation}
    H_{A,P}
    =
    \left\{ X \in \cor{n} \mid \inner{\rielog_{P}(X)}{A}_{P} = 0 \right\},
    \qquad
    P \in \cor{n},\ A \in T_{P} \cor{n}.
\end{equation}
In \cref{cornet:fig:hyperplanes}, we focus on $\cor{3}$ and visualize it as the open elliptope in $\bbR{3}$ via the embedding
\begin{equation}
    C \in \cor{3}
    \longmapsto
    (C_{21},C_{31},C_{32}) \in \bbR{3}.
\end{equation}
Given a Log-Euclidean metric and parameters $(A,P)$, each correlation matrix $C$ is first mapped to the tangent space at $P$ by $\rielog_{P}(C)$, and we evaluate the linear form $\inner{\rielog_{P}(C)}{A}_{P}$. The set of points in the elliptope where this scalar equals zero corresponds to the decision hyperplane $H_{A,P}$ and is plotted as the separating surface.

For PHCM, the margin hyperplane is defined in the $\beta$-concatenated Poincaré embedding. Let $\Phi$ be the diffeomorphism in \cref{cornet:eq:cor-to-ppb} that maps $C \in \cor{n}$ to the poly-Poincaré space $\bbPPB{n-1}$, and let $\beta$-concatenation be the Poincaré operation in \cref{cornet:subsubsec:revisit-poincare-layers}. We define
\begin{equation}
    \tilde{x}(X)
    =
    \beta\text{-concat}\left(\Phi(X)\right)
    \in \unitpball{N},
    \qquad
    N = \frac{n(n-1)}{2},
\end{equation}
and the PHCM hyperplane
\begin{equation}
    H_{a,p}
    =
    \left\{ X \in \cor{n} \mid \inner{\rielog_{p}(\tilde{x}(X))}{a}_{p} = 0 \right\},
    \qquad
    p \in \unitpball{N},\ a \in T_{p} \unitpball{N}.
\end{equation}
Here $\unitpball{N} = \left\{ x \in \bbR{N} \mid \norm{x}^{2} < 1 \right\}$ is the $N$-dimensional unit Poincaré ball. In \cref{cornet:fig:hyperplanes}, we first map each correlation matrix $C \in \cor{3}$ to $\tilde{x}(C) \in \unitpball{N}$, apply the Poincaré logarithm $\rielog_{p}$ at a reference point $p$, and then visualize the zero level set of the linear form $\inner{\rielog_{p}(\tilde{x}(C))}{a}_{p}$ as the PHCM decision hyperplane.

\subsection{Adaptive Log-Euclidean Metrics}
\label{alem:app:experimental-details}

\subsubsection{Data Sets and Settings}

Although the proposed ALog layers can be plugged into existing SPD networks, we focus on the SPDNet framework \citep{huang2017riemannian}. We follow the PyTorch code provided by SPDNetBN\footnote{\href{https://proceedings.neurips.cc/paper/2019/file/6e69ebbfad976d4637bb4b39de261bf7-Supplemental.zip}{SPDNetBN supplementary code}} to reproduce SPDNet and SPDNetBN and implement our approaches.

Following previous work \citep{huang2017riemannian,brooks2019riemannian}, we evaluate our methods on HDM05 \citep{muller2007documentation}, FPHA \citep{garcia2018first}, and AFEW \citep{dhall2018emotiw}. The data set descriptions are given in \cref{app:datasets}. We use the HDM05 and FPHA temporal covariance representations and the filtered HDM05 setting specified in \cref{app:subsec:exp_details_spd}. For AFEW, we use the released pre-trained FAN\footnote{\url{https://github.com/Open-Debin/Emotion-FAN}} \citep{meng2019frame} to extract deep features and establish a $512 \times 512$ temporal covariance matrix for each video.

We denote the dimensions of the transformation layers in the SPDNet backbone by $\{d_0,d_1,\ldots,d_L\}$. Following the settings in \citet{brooks2019riemannian}, all networks are trained using the default RSGD \citep{becigneul2018riemannian} with a fixed learning rate $\gamma$ and a batch size of 30. To make ALog start from the vanilla matrix logarithm, the parameters in MUL, DIV, and RELU are initialized as $1$, $1$, and $e$, respectively. By abuse of notation, SPDNet-ALog-MUL is abbreviated as ALog-MUL, denoting that we substitute the LogEig layer in SPDNet with the proposed ALog optimized by MUL.

\subsubsection{Implementation Details of Additional Applications}

\mypara{NTU60 data set.} For NTU60, we use the cross-view protocol and $75 \times 75$ covariance representation specified in \cref{rmlr:app:sec:exp_details}.

As reported in \cref{alem:sec:experiments}, MUL shows the best performance. Therefore, we view $A$ in \cref{alem:eq:rw_mul_mlog} as the parameter for all experiments. In the following, we discuss in detail the specific implementation of each method.

\mypara{LieBN.} We follow the official code\footnote{\url{https://github.com/GitZH-Chen/LieBN}} to implement the experiments.
The learning rate is $5e^{-2}$.
Since our LieBN-ALEM shows early convergence, we set the numbers of training epochs to 150, 50, and 30 for the [93, 30], [93, 70, 30], and [93, 70, 50, 30] architectures.
Other settings are the same as \cref{alem:sec:experiments}.

\mypara{RResNet.} We follow the official code\footnote{\url{https://github.com/CUAI/Riemannian-Residual-Neural-Networks}} to implement the experiments.
For the HDM05 data set, we use RSGD \citep{becigneul2018riemannian} with a $5e^{-2}$ learning rate for 200 training epochs.
For the NTU60 data set, we use Riemannian AMSGrad \citep{becigneul2018riemannian} with a $1e^{-2}$ learning rate for 50 training epochs.
We adopt the architectures of [93, 30] and [75, 30] on these two data sets.

\mypara{Gyro MLR.} Since the code of gyro MLR is not publicly available, we carefully re-implement the gyro MLR in \citet{nguyen2023building}.
We adopt an architecture of [75, 30] under an SGD optimizer. The batch size and number of training epochs are 30 and 200, respectively.

\subsection{Product Cholesky Metrics}
\label{pcm:app:sec:exp_details}

\subsubsection{Implementation Details of SPD Neural Networks}
\label{pcm:app:subsec:impl-spdnns}

\mypara{Backbone networks.} SPDNet and GyroSPD are reviewed in \cref{app:backbone-spd}, while RResNet is reviewed in \cref{app:backbone-others}.

\mypara{Data sets and preprocessing.} The Radar, HDM05, and FPHA data sets are described in \cref{app:datasets}. The global covariance representations for SPDNet and RResNet follow \cref{app:subsec:exp_details_spd}; the GyroSPD-specific multichannel construction is given below.

\mypara{SPD MLR.}\label{pcm:app:subsubsec:details_spdmlr} We follow the official PyTorch code\footnote{\url{https://github.com/GitZH-Chen/SPDMLR}} to implement the SPD MLR developed in \cref{spdmlr:sec:spd_mlr}. Due to the lack of official code, the GyroSPD backbone is carefully reimplemented in PyTorch following the original paper \citep{nguyen2023building}. For simplicity, we set $\bbM$ in $\defCDGBWM$ to the identity matrix.

\mypara{SPDNet.} Following \citet{huang2017riemannian}, we replace the vanilla tangent classifier (LogEig + FC + softmax) in SPDNet with SPD MLRs induced by AIM, LEM, LCM, $\defCDEM$, and $\defCDBWM$. We use a Riemannian AMSGrad \citep{becigneul2018riemannian} with a learning rate of $1e^{-2}$, a batch size of $30$, and a maximum of $200$ epochs. We denote the architectures by $[d_0,d_1,\ldots,d_L]$, where $d_i$ is the output dimension of the $i$-th BiMap layer. Following prior work \citep{huang2017riemannian,brooks2019riemannian}, we adopt $[20,16,8]$ on Radar, $[63,33]$ on FPHA, and $[93,30]$, $[93,70,30]$, $[93,70,50,30]$ for 1-, 2-, and 3-block variants on HDM05. As shown in \cref{tab:results_FPHA}, matrix power improves the performance of Cholesky-based metrics on FPHA. We therefore apply a power of $-0.25$ before SPD MLR layers under LCM, $\defCDEM$, and $\defCDBWM$. For $\defCDEM$ on FPHA, we further adopt a weight decay of $1e^{-4}$. The deformation factor $\theta$ is reported in \cref{pcm:app:tab:theta-mlr}.

\mypara{GyroSPD.} Following \citet{nguyen2023building}, the backbone consists of one gyrotranslation layer followed by an SPD MLR. We compare metrics based on LEM, LCM, AIM, and our proposed geometries under the same settings: Riemannian AMSGrad with a learning rate of $1e^{-2}$ and a batch size of $30$. The models are trained for up to $100$, $100$, and $50$ epochs on Radar, HDM05, and FPHA, respectively. Similarly, a matrix power of $0.25$ is applied on FPHA for LCM and our Cholesky-based metrics. The deformation factor $\theta$ is reported in \cref{pcm:app:tab:theta-mlr}.

\begin{table}[t]
  \centering
  \caption{Hyperparameter $\theta$ in $\defCDEM$ and $\defCDBWM$. It is selected from the candidate values used for SPDNet and GyroSPD.}
  \label{pcm:app:tab:theta-mlr}%
    \begin{tabular}{c c ccc}
    \toprule
    \textbf{Backbone} & \textbf{Metric} & \textbf{Radar} & \textbf{HDM05} & \textbf{FPHA} \\
    \midrule
    \multirow{2}[1]{*}{SPDNet} & $\defCDEM$ & -1.5  & -0.5  & 0.75 \\
          & $\defCDBWM$ & -0.75 & -1.5  & 1 \\
    \midrule
    \multirow{2}[1]{*}{GyroSPD} & $\defCDEM$ & -0.75 & -0.75 & -0.75 \\
          & $\defCDBWM$ & -1.5  & -1.5  & -0.5 \\
    \bottomrule
    \end{tabular}%
\end{table}%

\mypara{SPD input of SPDNet.} We use the global covariance representations specified in \cref{app:subsec:exp_details_spd}.

\mypara{SPD input of GyroSPD.} For Radar, the input is the same as SPDNet. For HDM05 and FPHA, we follow \citet{nguyen2023building} to model each sample into a multi-channel covariance tensor $[c,n,n]$. Specifically, we first identify the closest left (right) neighbor of each joint based on its distance to the hip (wrist) joint, and then combine the 3D coordinates of each joint and those of its left (right) neighbor to create a feature vector for the joint. For a given frame $t$, we compute its Gaussian embedding \citep{lovric2000multivariate}:
\begin{equation}
    Y_t=(\det \Sigma_{t})^{-\frac{1}{n+1}}
    \left[\begin{array}{cc} \Sigma_t+\mu_t\left(\mu_t\right)^\top & \mu_t \\
    \left(\mu_t\right)^\top & 1
\end{array}\right],
\end{equation}
where $\mu_t$ and $\Sigma_t$ are the mean vector and covariance matrix computed from the set of feature vectors within the frame. The lower part of the matrix $\log \left(Y_t\right)$ is flattened to obtain a vector $\tilde{v}_t$. All vectors $\tilde{v}_t$ within a time window $[t, t+c-1]$, where $c$ is determined from a temporal pyramid representation of the sequence (the number of temporal pyramids is set to 2 in our experiments), are used to compute a covariance matrix as
\begin{equation}
    \widetilde{\Sigma}_t=\frac{1}{c} \sum_{i=t}^{t+c-1}\left(\tilde{v}_i-\overline{v}_t\right)\left(\tilde{v}_i-\overline{v}_t\right)^\top,
\end{equation}
where $\overline{v}_t=\frac{1}{c} \sum_{i=t}^{t+c-1} \tilde{v}_i$. The resulting $\{\widetilde{\Sigma}_t\}$ are the covariance matrices that we need. On FPHA, we generate the covariances based on three sets of neighbors: left, right, and vertical (bottom) neighbors. After preprocessing, the input covariance matrices are $[3,28,28]$ and $[9,28,28]$ on HDM05 and FPHA, respectively.

\mypara{RResNet implementation.}\label{pcm:app:subsec:details_rresnet} The backbone architectures of RResNet \citep{katsman2024riemannian} are similar to those of SPDNet: $[20,16,8]$ on Radar, $[93,30]$ on HDM05, and $[63,33]$ on FPHA, respectively. The only architectural difference lies in the head: SPDNet directly uses a classification layer, whereas RResNet attaches a residual block before the classification head. Following RResNet, we adopt $\rielog_{I_n}$ + FC + softmax for classification under each metric. We use the official code\footnote{\url{https://github.com/CUAI/Riemannian-Residual-Neural-Networks}} to re-implement the AIM- and LEM-based RResNet. For the RResNets based on LCM and our $\defCDEM$, we adopt the following settings: a learning rate of $1e^{-2}$, a batch size of $30$, and a maximum of $200$ epochs. We use the standard cross-entropy loss as the training objective and optimize the parameters with the Riemannian AMSGrad optimizer \citep{becigneul2018riemannian}. The Cholesky diagonal power is set to $-1$, $0.5$, and $-0.5$ on the Radar, HDM05, and FPHA data sets, respectively. Additionally, on the FPHA data set, a matrix power of $-0.25$ is applied before the residual blocks to activate the latent geometry for both LCM and our metric.

\mypara{SPD input.} The input SPD matrices are the same as those used in SPDNet.

\section{Additional Discussions}
\label{app:additional-discussions}

\subsection{Riemannian Multinomial Logistic Regression}
\label{rmlr:app:additional-discussions}

\subsubsection{RMLR as a Natural Extension of the Euclidean MLR}
\label{rmlr:app:sec:rmlr_gen_emlr}
\begin{parisproposition}
    When $\calM=\bbR{n}$ is the standard Euclidean space, the RMLR defined in \cref{rmlr:thm:rmlr} becomes the Euclidean MLR in \cref{spdmlr:eq:EMLR_reform_start}.
\end{parisproposition}
\begin{proof}
    On the standard Euclidean space $\bbR{n}$, $\rielog_y x=x-y, \forall x,y \in \bbR{n}$.
    Besides, the differential maps of left translation and parallel transport are the identity maps.
    Therefore, given $x,p_k \in \bbR{n}$ and $a_k \in \bbR{n}\backslash\{\zerovec\} \cong T_\zerovec\bbR{n}\backslash\{\zerovec\}$, we have
    \begin{align}
        p(y=k \mid x \in \bbR{n}) 
        &\propto \exp\left( \langle \rielog_{p_k} x,  a_k \rangle_{p_k}\right), \\
        &\propto \exp\left( \langle x - p_k,  a_k \rangle\right),\\
        &\propto \exp\left( \langle x,  a_k \rangle -b_k\right),
    \end{align}
    where $b_k=\langle p_k,a_k\rangle$.
\end{proof}

\subsubsection{Gyro SPSD MLR as a Special Case of Our RMLR}
\label{rmlr:app:sec:gyro_spsd_mlr_as_rmlr}

Gyro SPSD MLR \citep{nguyen2024matrix} is derived from the product of the Grassmannian and SPD gyro spaces.
This section will show that the gyro SPSD MLR is a special case of our RMLR on the product geometry of the SPSD manifold. We first review some necessary results about gyro SPSD MLR and then show the equivalence.

Following the notations in \citet{nguyen2024matrix}, we denote the Grassmannian with canonical metric under the projector and ONB perspective as $\graspp{p,n}$ and $\grasonb{p,n}$, respectively.
The space of $n \times n$ SPSD matrices with a fixed rank $p$, denoted as $\spsd{n,p}$, forms an SPSD manifold \citep{bonnabel2013rank}.
As shown in \citet{bonnabel2013rank,nguyen2024matrix}, the SPSD manifold is a product space, \ie $\spsd{n,p} \cong \grasonb{p,n} \times \spd{p}$.
In other words, every $P \in \spsd{n,p}$ can be decomposed as $P=U_P S_P U_P^\top$ with $U_P \in \grasonb{p,n}$ and $S_P \in \spd{p}$.
We further denote $\spd{p,g}$ as the SPD manifold with metric $g$, where $g$ could be AIM, LEM, or LCM.
As shown in \citet{nguyen2024matrix}, the gyro space in $\spsd{n,p}$ can be defined by the product of gyro spaces of $\grasonb{p,n}$ and $\spd{p,g}$.
By this product structure, \citet{nguyen2024matrix} proposed the SPSD Pseudo-gyrodistance to a hyperplane.

\begin{parisdefinition}[SPSD hypergyroplanes \citep{nguyen2024matrix}]
    \label{rmlr:app:def:spsd_hyperplane}
    Let $P,W \in \grasonb{p,n} \times \spd{p,g}$. Then hypergyroplanes in the structure space $\grasonb{p,n} \times \spd{p,g}$ are defined as
    \begin{equation}
        H_{W,P}^{p s d,g}=\left\{Q \in \grasonb{p,n} \times \spd{p,g}\mid\left\langle\ominus_{p s d,g}P \oplus_{p s d,g}Q,W\right\rangle^{p s d,g}=0\right\},
    \end{equation}
    where $\oplus_{p s d,g}$ and $\langle\cdot,\cdot\rangle^{p s d,g}$ are the gyro addition and gyro inner product, respectively, as defined in \citet{nguyen2024matrix}.
\end{parisdefinition}

\begin{paristheorem}[SPSD pseudo-gyrodistance \citep{nguyen2024matrix}]
    \label{rmlr:app:thm:spsd_gyrodistance}
    Let
    \begin{equation}
        W=\left(U_W,S_W\right),\qquad P=\left(U_P,S_P\right),\qquad X=\left(U_X,S_X\right)\in\grasonb{p,n}\times\spd{p,g},
    \end{equation}
    and let $H_{W,P}^{p s d,g}$ be a hypergyroplane in the structure space $\grasonb{p,n}\times\spd{p,g}$. Then the pseudo-gyrodistance from $X$ to $H_{W,P}^{p s d,g}$ is given by
    \begin{equation}
        \label{rmlr:eq:gyro_spsd_mlr}
        \bar{d}\left(X,H_{W,P}^{psd,g}\right)
        =\frac{
        \left|\begin{aligned}
        &\lambda\left\langle
        \left(\widetilde{\ominus}_{g r}U_P\widetilde{\oplus}_{g r}U_X\right)
        \left(\widetilde{\ominus}_{g r}U_P\widetilde{\oplus}_{g r}U_X\right)^\top,
        U_WU_W^\top\right\rangle^{g r}\\
        &\quad+\left\langle\ominus_g S_P\oplus_g S_X,S_W\right\rangle^g
        \end{aligned}\right|}
        {\sqrt{\lambda\left(\left\|U_WU_W^\top\right\|^{g r}\right)^2+\left(\left\|S_W\right\|^g\right)^2}},
    \end{equation}
    where $\|\cdot\|^{g r}$ and $\|\cdot\|^g$ are the gyro norms on the Grassmann and SPD manifolds \citep{nguyen2024matrix}, and $\langle\cdot,\cdot\rangle^{gr}$ and $\langle\cdot,\cdot\rangle^{g}$ are gyro inner products \citep{nguyen2024matrix}. $\widetilde{\oplus}_{g r}$ and $\oplus_g$ are gyro additions on $\grasonb{p,n}$ and $\spd{p,g}$.
\end{paristheorem}

Denoting $g^{gr}$ as the canonical metric on $\grasonb{p,n}$ and $g$ as AIM, LEM, or LCM, we can prove that \cref{rmlr:app:thm:spsd_gyrodistance} is a special case of \cref{rmlr:thm:rie_margin_dist}.

\begin{paristheorem} \label{rmlr:app:thm:equivalence_rmlr_gyro}
    Under the product metric $g^{psd,g} = \lambda g^{gr} \times g$, 
    the Riemannian hyperplane in \cref{rmlr:eq:r_hyperplane} on the SPSD manifold equals the SPSD hypergyroplane in \cref{rmlr:app:def:spsd_hyperplane}.
    Similarly, the Riemannian margin distance in \cref{rmlr:thm:rie_margin_dist} on the SPSD manifold equals the SPSD Pseudo-gyrodistance in \cref{rmlr:app:thm:spsd_gyrodistance}.
\end{paristheorem}

\begin{proof}
    Following the notations in \cref{rmlr:app:def:spsd_hyperplane,rmlr:app:thm:spsd_gyrodistance}, we further denote $P = U_P S_P U_P^\top$, $Q = U_Q S_Q U_Q^\top$, $W = U_W S_W U_W^\top$, and $X = U_X S_X U_X^\top$ with $U_P, U_Q, U_W, U_X \in \grasonb{p,n}$ and $S_P, S_Q, S_W, S_X \in \spd{p,g}$.
    $I_p$ is the $p \times p$ identity matrix.
    $\idonb=(I_{p},0)^\top$ is the gyro identity on $\grasonb{p,n}$. $\widetilde{\Gamma}^{gr}$, $\widetilde{\rielog}^{gr}$, and $\langle\cdot,\cdot\rangle^{gr}_{U_P}$ denote Riemannian parallel transport along a geodesic, the Riemannian logarithm, and the Riemannian metric at $U_P$ on $\grasonb{p,n}$, respectively.
    $\Gamma^{g}$, $\rielog^{g}$, and $\langle\cdot,\cdot\rangle^{g}_{S_P}$ denote Riemannian parallel transport along a geodesic, the Riemannian logarithm, and the Riemannian metric at $S_P$ on $\spd{p,g}$, respectively.
    $\Gamma^{psd,g}$, $\rielog^{psd,g}$, and $\langle\cdot,\cdot\rangle^{psd,g}_{X}$ denote Riemannian parallel transport along a geodesic, the Riemannian logarithm, and the Riemannian metric at $X$ on $\spsd{n,p}\cong\grasonb{p,n}\times\spd{p}$, respectively.
    
    First, we show that the SPSD hypergyroplane equals our Riemannian hyperplane in \cref{rmlr:eq:r_hyperplane}.
    We have the following:
    \begin{equation}
    \label{rmlr:app:eq:gyro_hyerp_as_riem_hyper}
    \begin{aligned}
        &\left\langle \spsdominus P \spsdoplus Q, W \right\rangle^{p s d, g} \\
        &\stackrel{(1)}{=} \lambda \langle \grassominus U_P \grassoplus U_Q, U_W  \rangle^{gr} + \langle \gominus S_P \goplus S_Q, S_W \rangle^{g} \\
        &\stackrel{(2)}{=} \lambda \langle \rielog^{gr}_{U_P} U_Q, \tilde{A}_{U_W} \rangle^{gr}_{U_P}
        + \langle \rielog^{g}_{S_P} S_Q, \tilde{A}_{S_W} \rangle^{g}_{S_P} \\
        &\stackrel{(3)}{=} \langle \rielog^{psd, g}_P Q, \tilde{A} \rangle^{psd,g}_P
    \end{aligned}
    \end{equation}
    where
    \begin{align}
        \tilde{A}_{U_W} &= \widetilde{\Gamma}^{gr}_{\idonb \rightarrow U_P} \left( \widetilde{\rielog}^{gr}_{\idonb} (U_W) \right), \\
        \tilde{A}_{S_W} &= \Gamma^{g}_{I_p \rightarrow S_P} \left( \rielog^{g}_{I_p} (S_W) \right), \\
        \tilde{A} &= (\tilde{A}_{U_W}, \tilde{A}_{S_W}) \in T_P\spsd{n,p} \cong T_{U_P}\grasonb{p,n} \times T_{S_P}\spd{p,g},
    \end{align}
    where $\times$ denotes the Cartesian product.
    The above derivation comes from the following.
    \begin{enumerate}
        \item 
        The definition of gyro addition, gyro inverse, and gyro inner product on the SPSD manifold \citep[Sec.~3.3]{nguyen2024matrix}.
        \item 
        The proof by \citet[Prop.~3.2]{nguyen2024matrix} indicates that similar results also hold on the Grassmannian. Combining that proposition with its Grassmannian counterparts yields the equation.
        \item 
        The Riemannian product geometry.
    \end{enumerate}
    By the product geometry of the SPSD manifold, we can immediately get
    \begin{equation} \label{rmlr:app:eq:tilde_A}
        \tilde{A}
        = (\tilde{A}_{U_W}, \tilde{A}_{S_W})
        =\Gamma^{psd,g}_{\idpp \rightarrow P } \left( \rielog_{\idpp}^{psd,g} \left( W \right) \right),
    \end{equation}
    where $\idpp=\idonb \idonb^\top$ is the gyro identity on the SPSD manifold.
    
    Next, we show the equivalence between SPSD pseudo-gyrodistance and our Riemannian margin distance:
    \begin{equation}
        \begin{aligned}
            \bar{d}\left(X, H_{W, P}^{psd, g}\right)
            &\stackrel{(1)}{=} \frac{\left|\left\langle \spsdominus P \spsdoplus X, W \right\rangle^{p s d, g} \right|}{ \left \| W \right\|^{psd,g}} \\
            &\stackrel{(2)}{=} \frac{\left| \langle \rielog^{psd, g}_P X, \tilde{A} \rangle^{psd,g}_P \right|}{ \left \| W \right\|^{psd,g}} \\
            &\stackrel{(3)}{=} \frac{\left| \langle \rielog^{psd, g}_P X, \tilde{A} \rangle^{psd,g}_P \right|}{ \left \| \rielog_{\idpp}^{psd,g} \left( W \right) \right\|_{\idpp}^{psd,g}}\\
            &\stackrel{(4)}{=} \frac{\left| \langle \rielog^{psd, g}_P X, \tilde{A} \rangle^{psd,g}_P \right|}{\left \| \Gamma^{psd,g}_{\idpp \rightarrow P } \left(\rielog_{\idpp}^{psd,g} \left( W \right) \right) \right\|_{P}^{psd,g}} \\ 
            &\stackrel{(5)}{=} \frac{\left| \langle \rielog^{psd, g}_P X, \tilde{A} \rangle^{psd,g}_P \right|}{ \left \| \tilde{A} \right\|_{P}^{psd,g}} \\   
            &\stackrel{(6)}{=} d(X,\tilde{H}_{\tilde{A}, P}) \\
        \end{aligned}
    \end{equation}
    \begin{enumerate}
        \item 
        The definition of gyro addition, gyro inverse, gyro inner product, and gyro norm on the SPSD manifold.
        \item 
        \cref{rmlr:app:eq:gyro_hyerp_as_riem_hyper}.
        \item 
        The definition of SPSD gyro norm \citep{nguyen2024matrix}.
        \item 
        Riemannian parallel transport is norm-preserving \citep[Def.~3.1]{do1992riemannian}.
        \item 
        \cref{rmlr:app:eq:tilde_A}.
        \item 
        \cref{rmlr:thm:rie_margin_dist}.
    \end{enumerate}
\end{proof}

\begin{parisremark}
    We make the following remarks regarding the gyro MLR and our MLR on the SPSD manifold.
    \begin{enumerate}
        \item 
        \cref{rmlr:app:eq:tilde_A} indicates that when generating $\tilde{A}$ in our RMLR by parallel transporting a tangent vector $A \in T_{\idpp}\spsd{n,p}$, $\tilde{A}$ is the initial velocity of $W$ in \cref{rmlr:eq:gyro_spsd_mlr}.
        \item
        Putting the pseudo-gyrodistance and Riemannian margin distance into \cref{rmlr:eq:rmlr_v1} yields the gyro MLR and our Riemannian MLR.
        Therefore, \cref{rmlr:app:thm:equivalence_rmlr_gyro} indicates the equivalence of the gyro MLR with our RMLR on the SPSD manifold.
        \item
        As a metric $g$ is required to induce gyro-structures, the metric $g$ in gyro SPSD MLR is confined to AIM, LEM, and LCM.
        However, our SPSD MLR can be defined on the product space of the Grassmannian and SPD manifold under other metrics, such as BWM and PEM, as our framework does not require gyro-structures.
    \end{enumerate}
\end{parisremark}

\subsubsection{Theories on the Deformed Metrics}
\label{rmlr:app:sec:deformed_metrics_theory}

\mypara{Limiting cases of the deformed metrics.} \citet{thanwerdas2019affine} generalized $\biparamAIM$ to a three-parameter family of metrics by power deformation, \ie $\triparamAIM$.
The family of $\triparamAIM$ comprises $\biparamAIM$ for $\theta = 1$ and approaches $\biparamLEM$ with $\theta \rightarrow 0$ \citep{thanwerdas2019affine}.

As reviewed in \cref{subsubsec:spd_param_lie_groups}, LCM and $\biparamLEM$ are extended to power-deformed metrics, denoted as $\triparamLEM$ and $\paramLCM$.
\cref{prop:spd_param_lem_lcm_deformation} shows that $\triparamLEM$ is equal to $\biparamLEM$, and $\paramLCM$ interpolates between $\tilde{g}$-LEM ($\theta \rightarrow 0$) and LCM ($\theta=1$), with $\tilde{g}$-LEM defined as
\begin{equation}
    \langle V_1,V_2 \rangle_P = \frac{1}{2} \langle \widetilde{V_1}, \widetilde{V_2} \rangle -\frac{1}{4} \langle \bbD(\widetilde{V_1}), \bbD(\widetilde{V_2}) \rangle, \forall V_i \in T_P\spd{n},
\end{equation}
where $\widetilde{V_i} = \log_{*,P}(V_i)$ with $\log_{*,P}$ as the differential map of matrix logarithm, and $\bbD(\widetilde{V_i})$ is a diagonal matrix consisting of the diagonal elements of $\widetilde{V_i}$.

\citet{thanwerdas2022geometry} identified the Alpha-Procrustes metric \citep{minh2022alpha} with power-deformed BWM, denoted as $2\theta$-BWM.
Similarly, $2\theta$-BWM becomes BWM with $\theta=0.5$ \citep{thanwerdas2022geometry}. 
We further show the limiting case of $2\theta$-BWM under $\theta \rightarrow 0$.

\begin{parisproposition} \label{rmlr:prop:deformed_bwm_limit}
    $2\theta$-BWM tends to $(\frac{1}{4},0)$-LEM as $\theta \rightarrow 0$.
\end{parisproposition}

Before starting the proof, we first recall a well-known property of deformed metrics \citep{thanwerdas2022geometry}.

\begin{parislemma} \label{rmlr:lem:diformed_metrics_lim}
    Let $\frac{1}{\theta^2}\phi_{\theta}^*g$ be the deformed metric on SPD manifolds pulled back from $g$ by the matrix power $\phi_\theta$ and scaled by $\frac{1}{\theta^2}$.
    Then, as $\theta$ tends to 0, for all $P \in \spd{n}$ and all $V \in T_P\spd{n}$, we have
    \begin{equation} \label{rmlr:eq:deformed_g_lim}
        (\frac{1}{\theta^2}\phi_{\theta}^*g)_P(V,V) \to g_{I}(\log_{*,P}(V),\log_{*,P}(V)).
    \end{equation}
\end{parislemma}

Now, we present our proof for the limiting cases of deformed metrics.
\begin{proof}[Proof of \cref{rmlr:prop:deformed_bwm_limit}]
    First, we have 
    \begin{equation}
        g_I^{\text{BWM}}(V,V) = \frac{1}{4}\langle V, V\rangle.
    \end{equation}

    By \cref{rmlr:lem:diformed_metrics_lim}, we have the following:
    \begin{equation}
        \begin{aligned}
            \gparamBWM_P(V,V) 
            &\xrightarrow{\theta \rightarrow 0} g^{\text{BWM}}_I \left(\log_{*,P}(V),\log_{*,P}(V) \right)\\
            &= \frac{1}{4} \langle \log_{*,P}(V), \log_{*,P}(V) \rangle \\
            &= g_P^{(\frac{1}{4},0)\text{-LEM}}\left(V,V \right).
        \end{aligned}
    \end{equation}
\end{proof}

\mypara{Proof of the properties of the deformed metrics (\cref{rmlr:tab:properties_rie_metrics}).} In this subsection, we prove the properties presented in \cref{rmlr:tab:properties_rie_metrics}.
We first present a useful lemma and then present our detailed proof.
This lemma will be useful in the proof of our SPD MLRs as well.

\begin{parislemma} \label{rmlr:lem:scale_metric_rie_opt}
    Given a Riemannian manifold $(\calM,g)$ and a positive real scalar $a>0$, the scaled metric $ag$ on $\calM$ has the same Riemannian logarithmic maps, exponential maps, and parallel transport as $g$.
\end{parislemma}
\begin{proof}
    Since the Christoffel symbols of $ag$ are identical to those of $g$, the geodesics and parallel transport under both $ag$ and $g$ remain unchanged.
    The equivalence of geodesics implies that the Riemannian exponential maps are the same for $ag$ and $g$.
    As the inverse of the Riemannian exponential maps, the Riemannian logarithmic maps under $ag$ and $g$ are also identical.
\end{proof}

According to \cref{rmlr:lem:scale_metric_rie_opt}, geodesic completeness is independent of the scaling factor $a>0$.
By the definition of $\orth{n}$-, left-, right-, and bi-invariance, these invariant properties are also independent of the scaling factor $a>0$.
Without loss of generality, we will omit the scaling factor in the following proof.

\begin{proof}
    First, we prove $\orth{n}$-invariance of $(\theta,\alpha,\beta)$-LEM, $(\theta,\alpha,\beta)$-EM, $(\theta,\alpha,\beta)$-AIM, and $2\theta$-BWM.
    Since the differential of $\phi_{\theta}$ is $\orth{n}$-equivariant, and $\alphabeta$-LEM, $\alphabeta$-EM, $\alphabeta$-AIM, and BWM are $\orth{n}$-invariant \citep{thanwerdas2023n}, $\orth{n}$-invariance is thus acquired.

    Next, we focus on geodesic completeness.
    It can be easily proven that Riemannian isometries preserve geodesic completeness.
    On the other hand, $\alphabeta$-LEM, $\alphabeta$-AIM, and LCM are geodesically complete \citep{thanwerdas2023n,lin2019riemannian}.
    As a direct corollary, geodesic completeness can be obtained since $\phi_{\theta}$ is a Riemannian isometry.

    Finally, we deal with Lie group invariance.
    Similarly, it can be readily proved that Lie group invariance is preserved under isometries.
    LCM, LEM, and $\alphabeta$-AIM are Lie group bi-invariant \citep{lin2019riemannian}, bi-invariant \citep{arsigny2005fast}, and left-invariant \citep{thanwerdas2022theoretically}.
    As an isometric pullback metric from the standard LEM \citep{thanwerdas2023n}, $\alphabeta$-LEM is, therefore, Lie group bi-invariant.
    As pullback metrics, $(\theta,\alpha,\beta)$-LEM, $(\theta,\alpha,\beta)$-AIM, and $\theta$-LCM are therefore bi-invariant, left-invariant, and bi-invariant, respectively.
\end{proof}

\subsubsection{Computational Details on the SPD MLR under Power-Deformed BWM}
\label{rmlr:app:sec:detail_bwm_spdmlr}

\mypara{Matrix square roots in the SPD MLR under power-deformed BWM.} In the case of MLRs induced by $2\theta$-BWM, computing square roots like $(BA)^{\frac{1}{2}}$ and $(AB)^{\frac{1}{2}}$ with $B,A \in \spd{n}$ poses a challenge.
    Eigendecomposition cannot be directly applied since $BA$ and $AB$ are no longer symmetric, let alone positive definite.
    Instead, we use the following formulas to compute these square roots \citep{minh2022alpha}:
    \begin{equation}
        (BA)^{\frac{1}{2}}=B^{\frac{1}{2}}(B^{\frac{1}{2}}A B^{\frac{1}{2}})^{\frac{1}{2}}B^{-\frac{1}{2}} \text{ and } (AB)^{\frac{1}{2}}=[(BA)^{\frac{1}{2}}]^\top,
    \end{equation}
    where the involved square roots can be computed using eigendecomposition or singular value decomposition (SVD).

\mypara{Numerical stability of the SPD MLR under power-deformed BWM.} Let us first explain why we abandon parallel transport for the SPD MLR derived from $2\theta$-BWM.
Then, we propose our numerically stable methods for computing the SPD MLR based on $2\theta$-BWM.

\mypara{Instability of parallel transport under power-deformed BWM.} As discussed in \cref{rmlr:thm:rmlr}, there are two ways to generate $\tilde{A}$ in SPD MLR: parallel transport and Lie group translation.
However, parallel transport under $2\theta$-BWM can cause numerical problems.
Without loss of generality, we focus on the standard BWM because $2\theta$-BWM is isometric to the BWM.

Although general parallel transport under BWM is obtained by solving an ODE, transport starting from the identity matrix has a closed-form expression \citep{thanwerdas2023n}:
\begin{equation} \label{rmlr:eq:pt_bwm_from_id}
    \Gamma_{I \rightarrow P}(V)= U\left[\sqrt{\frac{\sigma_i+\sigma_j}{2}}\left[U^{\top} V U\right]_{i j}\right] U^{\top},
\end{equation}
where $P=U \Sigma U^\top$ is the eigendecomposition of $P \in \spd{n}$.
The forward computation of \cref{rmlr:eq:pt_bwm_from_id} is numerically stable.
However, its backpropagation requires differentiating through an eigendecomposition, involving the calculation of $\nicefrac{1}{(\sigma_i-\sigma_j)}$ \citep[Prop.~2]{ionescu2015matrix}.
When $\sigma_i$ is close to $\sigma_j$, this backpropagation can be problematic.

\mypara{Numerically stable methods for the SPD MLR under power-deformed BWM.} To bypass the instability of parallel transport under BWM, we use Lie group left translation to generate $\tilde{A}$ in MLRs induced by $2\theta$-BWM.
However, there is another problem that could cause instability.
The computation of the Riemannian metric of $2\theta$-BWM requires solving the Lyapunov operator, \ie $\calL_P[V] P+ P \calL_P[V]=V$.
For symmetric matrices, the Lyapunov operator can be obtained by eigendecomposition:
\begin{equation} \label{rmlr:eq:solution_ly_sym}
    \calL_P[V] = U\left[\frac{V_{i j}^{\prime}}{\sigma_i+\sigma_j}\right]_{i, j} U^{\top},
\end{equation}
where $V \in \sym{n}$, $UV'U^\top=V$, and $P=U\Sigma U^\top$ is the eigendecomposition of $P \in \spd{n}$.
Similar to \cref{rmlr:eq:pt_bwm_from_id}, the backpropagation of \cref{rmlr:eq:solution_ly_sym} requires $\nicefrac{1}{(\sigma_i-\sigma_j)}$, undermining numerical stability.

To remedy this problem, we propose the following formula to stably compute the backpropagation of \cref{rmlr:eq:solution_ly_sym}.

\begin{parisproposition}
    For all $P \in \spd{n}$ and all $V \in \sym{n}$, we denote the Lyapunov equation as
    \begin{equation} \label{rmlr:eq:ly_symbol}
        XP+PX=V,
    \end{equation}
    where $X=\calL_P[V]$.
    Given the gradient $\frac{\partial L}{\partial X}$ of loss $L$ with respect to $X$, the backpropagation of the Lyapunov operator can be computed by
    \begin{align}
        \frac{\partial L}{\partial V} &= \calL_P[\frac{\partial L}{\partial X}],\\
        \frac{\partial L}{\partial P} &= -X\calL_P[\frac{\partial L}{\partial X}]-\calL_P[\frac{\partial L}{\partial X}]X,
    \end{align}
\end{parisproposition}
where $\calL_P[\cdot]$ can be computed by \cref{rmlr:eq:solution_ly_sym}.
\begin{proof}
    Differentiating both sides of \cref{rmlr:eq:ly_symbol}, we obtain
    \begin{align}
        & \diff X P + X \diff P + \diff P X+P \diff X=\diff V, \\
        \implies & \diff X P +P \diff X = \diff V - X \diff P - \diff P X,\\
        \implies & \diff X = \calL_P[\diff V - X \diff P - \diff P X].
    \end{align}
    Besides, easy computations show that
    \begin{equation}
        \calL_P[V] : W = V : \calL_P[W], \forall W,V \in \sym{n},
    \end{equation}
    where $\cdot : \cdot$ denotes the standard Frobenius inner product.

    Then we have the following:
    \begin{align}
        & \frac{\partial L}{\partial X} : \diff X = \frac{\partial L}{\partial X} : \calL_P[\diff V - X \diff P - \diff P X],\\
        \implies & \frac{\partial L}{\partial X} : \diff X = \calL_P[\frac{\partial L}{\partial X}] : \diff V + \left( -X\calL_P[\frac{\partial L}{\partial X}]-\calL_P[\frac{\partial L}{\partial X}]X \right): \diff P.
    \end{align}
\end{proof}
\begin{parisremark}
    \cref{rmlr:eq:solution_ly_sym} needs to be computed in the Lyapunov operator's forward and backward processes.
    Therefore, in the forward process, we can save the intermediate matrices $U$ and $K$ with $K_{i,j}=\left[\frac{1}{\sigma_i+\sigma_j}\right]_{i, j}$, and then use them to compute the backward process efficiently.
\end{parisremark}

\subsection{Hyperbolic Busemann Neural Networks}
\label{hbnn:app:additional-discussions}

\subsubsection{Comparison with Existing Hyperbolic MLRs}
\label{hbnn:app:subsubsec:mlr-comparison}
\begin{table}[H]
    \centering
    \caption{Comparison of hyperplanes. Compact params indicate whether the parameterization requires an additional manifold-valued point.}
    \label{hbnn:app:tab:hyperplane-comparison}
    \resizebox{\linewidth}{!}{%
    \begin{tabular}{ccccc}
        \toprule
        \textbf{Method} & \textbf{Hyperplane} & \textbf{Formulation} & \makecell{\textbf{Applied} \\ \textbf{manifolds}} & \makecell{\textbf{Compact} \\ \textbf{params}} \\
        \midrule
        Euclidean MLR & Euclidean & \makecell{$\left\{ x \in \bbR{n} \mid \inner{a}{x} + b = 0 \right\}$ \\ $a \in \bbR{n},\ b \in \bbRscalar$} & $\bbR{n}$ & \cmark \\
        \midrule
        Poincaré MLR \citep{ganea2018hyperbolic} & \makecell{Geodesic \\ \citep[Def.~3.1]{ganea2018hyperbolic}} & \makecell{$\left\{ x \in \pball{n} \mid \inner{ \rielog_{p}\left(x\right)}{a }_{p} = 0 \right\}$ \\ $ p \in \pball{n},\ a \in T_{p}\pball{n}$} & $\pball{n}$ & \xmark \\
        \midrule
        Pseudo-Busemann MLR \citep{nguyen2025neural} & \makecell{Busemann \& gyro \\ \citep[Def.~4.1]{nguyen2025neural}} & \makecell{$\left\{ x \in \pball{n} \mid B^{v}\left( - p \Moplus x \right) = 0 \right\}$ \\ $v \in \unitsphere{n-1},\ p \in \pball{n}$} & $\pball{n}$ & \xmark\\
        \midrule
        Lorentz MLR \citep{bdeir2024fully} & \makecell{Ambient Minkowski \\ \citep[Eq.~(7)]{bdeir2024fully}} & \makecell{$\left\{ x \in \lorentz{n} \mid \Linner{w}{x} = 0 \right\}$ \\ $p \in \lorentz{n},\ w \in T_{p}\lorentz{n}$} & $\lorentz{n}$ & \xmark\\
        \midrule
        \rowcolor{HilightColor} BMLR & \Gape[0pt][2pt]{Horosphere} & \Gape[0pt][2pt]{\makecell{$\left\{ x \in \hyperspace{n} \mid -\alpha B^{v}\left(x\right) + b = 0 \right\}$ \\ $\alpha > 0,\ v \in \unitsphere{n-1},\ b \in \bbRscalar$}} & $\pball{n}$, $\lorentz{n}$ & \cmark\\  
        \bottomrule
    \end{tabular}
    }
\end{table}

\begin{table}[t]
    \centering
    \caption{Comparison of point-to-hyperplane distances. \greenbf{Real} means the point-to-hyperplane distance is the real distance, obtained by $\inf_{y \in H} \dist(x,y)$ with $H$ as a hyperplane and $\dist$ as the geodesic distance. Instead, \redbf{Pseudo} means the point-to-hyperplane distance is a surrogate, which only equals the real distance in Euclidean geometry.}
    \label{hbnn:app:tab:distance-comparison}
    \resizebox{\linewidth}{!}{%
    \begin{tabular}{cccc}
        \toprule
        \textbf{Method} & \textbf{Point-to-hyperplane distance} & \makecell{\textbf{Applied} \\ \textbf{manifolds}} & \textbf{Dist} \\
        \midrule
        Euclidean MLR & $\displaystyle \frac{|\inner{a}{x} + b|}{\norm{a}}$ & $\bbR{n}$ & \greenbf{Real} \\
        \midrule
        Poincaré MLR \citep{ganea2018hyperbolic} & \makecell{$\displaystyle \frac{1}{\sqrt{-K}} \sinh^{-1}\left(\frac{2\sqrt{-K}  \left|\inner{-p \Moplus x}{a}\right|}{\left(1 + K\norm{-p \Moplus x}^{2}\right)  \norm{a}}\right)$ \\ \citep[Thm.~5]{ganea2018hyperbolic}} & $\pball{n}$ & \greenbf{Real} \\
        \midrule
        Pseudo-Busemann MLR \citep{nguyen2025neural} & \makecell{$\displaystyle \dist(x, p)  \frac{B^{v}\left( -p \Moplus x\right)}{\norm{ -p \Moplus x}}$ \\ \citep[Cor.~4.3]{nguyen2025neural}} & $\pball{n}$ & \redbf{Pseudo} \\
        \midrule
        Lorentz MLR \citep{bdeir2024fully} & \makecell{$\displaystyle \frac{1}{\sqrt{-K}}\left|\sinh^{-1}\left(\sqrt{-K}  \frac{\Linner{v}{x}}{\Lnorm{v}}\right)\right|$ \\ \citep[Eq.~(44)]{bdeir2024fully}} & $\lorentz{n}$ & \greenbf{Real} \\
        \midrule
        \rowcolor{HilightColor} BMLR & $\displaystyle \frac{\left|- \alpha B^v(x) + b\right|}{\alpha}$ & $\pball{n}$, $\lorentz{n}$ & \greenbf{Real} \\
        \bottomrule
    \end{tabular}
    }
\end{table}

All the considered hyperbolic MLRs follow a point-to-hyperplane formulation. Therefore, the key difference lies in hyperplanes and point-to-hyperplane distances across methods. In addition to \cref{hbnn:tab:logit-comparison}, \cref{hbnn:app:tab:hyperplane-comparison,hbnn:app:tab:distance-comparison} further make this comparison. We draw the following three conclusions.
\begin{enumerate}
    \item \mypara{Hyperplanes.} Our BMLR uses Busemann-based horospheres that simultaneously satisfy three desiderata: 
    (i) compact parameterization without a per-class manifold-valued point, whereas other hyperbolic ones\footnote{We note that \citet{shimizu2021hyperbolic,bdeir2024fully} mitigate this issue through re-parameterization: $a=\pt{e}{p}(z)$ and $p=\rieexp_e\left(b \frac{z}{\norm{z}}\right)$, where $e \in \hyperspace{n}$ denotes the origin, $z \in \bbR{n}$, and $b \in \bbRscalar$. Nevertheless, the underlying definitions are over-parameterized.} are over-parameterized;
    (ii) a natural generalization of Euclidean hyperplanes via horospheres, while the Lorentz MLR relies on the ambient Minkowski space, failing to fully respect the intrinsic geometry; 
    and (iii) applicability across hyperbolic models, whereas the Lorentz MLR is tailored to the Lorentz model.
    \item \mypara{Point-to-Hyperplane Distances.} Although pseudo-Busemann MLR also exploits Busemann functions, it relies on a pseudo point-to-hyperplane distance that only coincides with the real distance in Euclidean geometry. In contrast, our BMLR calculates the real point-to-horosphere distance, ensuring geometric fidelity across hyperbolic models.
    \item \mypara{Batch Efficiency.} Recalling \cref{hbnn:tab:logit-comparison}, the Poincaré MLR \citep{ganea2018hyperbolic} computes logits using $\inner{-p_k \Moplus x}{a_k}$ and $\norm{-p_k \Moplus x}^{2}$. For a batch $X \in \bbR{\mathrm{bs} \times n}$ and $C$ classes, evaluating $-p_k \Moplus X$ for every $k$ yields an intermediate tensor of shape $[\mathrm{bs}, C, n]$, and materializing this tensor can cause GPU out of memory (OOM) when $n$ or $C$ is large. The same limitation holds for the pseudo-Busemann MLR. Consequently, their official implementations compute per class in a for-loop, which is batch inefficient. By contrast, BMLR uses logits $-\alpha_k B^{v_k}(x)+b_k$, whose Busemann term reduces to class-wise inner products $\inner{v_k}{x}$ (or $\inner{v_k}{x_s}$). With $X \in \bbR{\mathrm{bs} \times n}$ and $V=[v_1,\ldots,v_C] \in \bbR{n \times C}$, such inner products can be efficiently implemented as a single matrix multiplication $XV$ without any $[\mathrm{bs}, C, n]$ intermediate, yielding high throughput and low memory usage.
\end{enumerate}

\subsubsection{Busemann Fully Connected Layers and Point-to-Horosphere Distances}
\label{hbnn:app:subsec:busemann-fc-p2h-distance}

An apparently natural attempt to define a hyperbolic FC layer is to replace the LHS of \cref{hbnn:eq:hyp-fc-p2h} by the \emph{signed point-to-horosphere} distance. The Euclidean hyperplane passing through the origin and orthogonal to $e_k \in \bbR{m}$ is $H_{e_k,\zerovec}=\left\{ y \in \bbR{m} \mid \inner{e_k}{y} = 0 \right\}$. The corresponding hyperbolic horosphere, following \cref{hbnn:eq:horosphere-param}, is $H_{e_k,1,0}=\{y \in \hyperspace{m} \mid -B^{e_k}(y) = 0 \}$. By \cref{hbnn:eq:point-to-horosphere-dist-param}, the signed point-to-horosphere distance to $H_{e_k,1,0}$ equals $-B^{e_k}(y)$. Accordingly, we can define an alternative FC mapping $\calF: \hyperspace{n} \ni x \mapsto y \in \hyperspace{m}$ via
\begin{equation}\label{hbnn:app:eq:busemann-fc-p2horosphere}
    B^{e_k}(y) = \alpha_k B^{v_k}(x) - b_k, \quad k=1,\ldots,m,
\end{equation}
where $u_k(x) = -\alpha_k B^{v_k}(x) + b_k$, and $\alpha_k>0$, $v_k \in \unitsphere{n-1}$, $b_k \in \bbRscalar$ are learnable. Although this only differs from \cref{hbnn:eq:hyp-fc-p2h} on the LHS, the following discussion shows that such a definition is infeasible in general and fails to deliver a valid hyperbolic FC layer.

\mypara{Poincar\'e Model.} Using \cref{hbnn:eq:poincare-busemann} with $v=e_k$, \cref{hbnn:app:eq:busemann-fc-p2horosphere} becomes
\begin{equation}\label{hbnn:app:eq:pfc-busemann-eq}
    \frac{1}{\sqrt{-K}} \log\left( \frac{\norm{ e_k - \sqrt{-K} y }^{2}}{1 + K\norm{y}^{2}} \right) = - u_k(x).
\end{equation}
Define $t_k = \exp\left( -\sqrt{-K} u_k(x) \right) > 0$. Exponentiating \cref{hbnn:app:eq:pfc-busemann-eq} gives
\begin{equation}\label{hbnn:app:eq:pfc-comp-eq}
    t_k \left( 1 + K\norm{y}^{2} \right) = 1 - 2\sqrt{-K} y_k - K\norm{y}^{2}, \quad k=1,\ldots,m.
\end{equation}
Writing $R=\norm{y}^{2}$, \cref{hbnn:app:eq:pfc-comp-eq} yields an affine expression for each coordinate
\begin{equation}\label{hbnn:app:eq:pfc-affine}
    y_k = c_k + d_k R, \quad c_k = \frac{1 - t_k}{2\sqrt{-K}}, \quad d_k = \frac{\sqrt{-K}}{2}\left(1 + t_k\right).
\end{equation}
Imposing $R = \sum_{k=1}^m y_k^2 = \sum_{k=1}^m\left(c_k + d_k R\right)^2$ gives a quadratic in $R$:
\begin{equation}\label{hbnn:app:eq:pfc-R-quadratic}
    A_2 R^2 + \left(A_1 - 1\right) R + A_0 = 0,
\end{equation}
where, denoting $T = \sum_{k=1}^m t_k$ and $q = \sum_{k=1}^m t_k^2$,
\begin{equation}\label{hbnn:app:eq:pfc-coeffs}
    A_2 = \frac{-K}{4}\left(m + 2T + q\right), \quad A_1 = \frac{m - q}{2}, \quad A_0 = \frac{m - 2T + q}{4(-K)} \geq 0.
\end{equation}
Its discriminant is
\begin{equation}
    \begin{aligned}
    \Delta_{\mathrm{P}}
    &= (A_1-1)^2 - 4 A_2 A_0\\
&= \left( \frac{m-q}{2} - 1 \right)^{2} - 4 \left( \frac{-K}{4}\left(m+2T+q\right) \right) \left( \frac{m-2T+q}{4(-K)} \right) \\
&= \left( \frac{m-2-q}{2} \right)^{2} - \frac{\left(m+2T+q\right)\left(m-2T+q\right)}{4} \\
&= \frac{1}{4} \left[ (m-2-q)^2 - \left( (m+q)^2 - (2T)^2 \right) \right] \\
&= \frac{1}{4} \left[ \left((m-2-q)-(m+q)\right) \left((m-2-q)+(m+q)\right) + 4T^2 \right] \\
&= \frac{1}{4} \left[ (-2-2q)(2m-2) + 4T^2 \right] \\
    &= T^2 - (m-1)\left(1+q\right).
    \end{aligned}
\end{equation}
A real solution $R$ exists only if $\Delta_{\mathrm{P}} \ge 0$. In addition, feasibility requires $0 \le R < -1/K$ so that $y \in \pball{m}$. Such conditions can fail for generic $\{ u_k(x) \}$, in which case no $y \in \pball{m}$ satisfies \cref{hbnn:app:eq:busemann-fc-p2horosphere}.

\mypara{Lorentz Model.} Using \cref{hbnn:eq:lorentz-busemann} with $v=e_k$, \cref{hbnn:app:eq:busemann-fc-p2horosphere} becomes
\begin{equation}\label{hbnn:app:eq:lfc-busemann-eq}
    \frac{1}{\sqrt{-K}} \log\left( \sqrt{-K} \left( y_t - (y_s)_k \right) \right) = - u_k(x).
\end{equation}
Define $t_k = \exp\left( -\sqrt{-K} u_k(x) \right) > 0$ and $t=(t_1,\ldots,t_m)^\top$. Denoting $T = \sum_{k=1}^m t_k$ and $q = \sum_{k=1}^m t_k^2$, we obtain from \cref{hbnn:app:eq:lfc-busemann-eq},
\begin{equation}\label{hbnn:app:eq:lfc-spatial}
    (y_s)_k = y_t - \frac{t_k}{\sqrt{-K}}, \quad k=1,\ldots,m, \quad\Longrightarrow\quad y_s = y_t\vecone - \frac{1}{\sqrt{-K}} t.
\end{equation}
Enforcing the Lorentz constraint $\norm{y_s}^{2} - y_t^{2} = 1/K$ yields a quadratic in $y_t$:
\begin{equation}\label{hbnn:app:eq:lfc-yt-quadratic}
    (m-1) K y_t^{2} + 2\sqrt{-K} T y_t - \left(1 + q\right) = 0.
\end{equation}
The discriminant is
\begin{equation}\label{hbnn:app:eq:lfc-discriminant}
    \begin{aligned}
        \Delta_{\mathrm{L}}
        &= \left(2\sqrt{-K} T\right)^{2} - 4(m-1)K\left( -\left(1 + q\right) \right) \\
        &= 4(-K) T^{2} + 4(m-1)K\left(1 + q\right) \\
        &= 4(-K) \left[ T^{2} - (m-1)\left(1 + q\right) \right].
    \end{aligned}
\end{equation}
Hence, a real $y_t$ exists only if $T^{2} - (m-1)\left(1 + q\right) \ge 0$. In addition, feasibility requires $y_t>0$. Such conditions can fail for generic $\{ u_k(x) \}$, where no $y \in \lorentz{m}$ satisfies \cref{hbnn:app:eq:busemann-fc-p2horosphere}.

\mypara{Summary.} Equating Busemann coordinates as in \cref{hbnn:app:eq:busemann-fc-p2horosphere} requires nontrivial inequalities on the responses $\{ u_k(x) \}$. These constraints are not guaranteed during learning, so the system can become infeasible and the output $y$ undefined. This motivates our choice in \cref{hbnn:eq:hyp-fc-p2h} to use signed point-to-hyperplane distances on the LHS, which admit closed-form solutions that are feasible for all inputs and parameters in both the Poincar\'e and Lorentz models.

\subsection{Full-Rank Correlation Networks}
\label{cornet:app:additional-discussions}

\subsubsection{Connections among FC Layers: Correlation, SPD, Poincaré, and Euclidean}
\label{cornet:app:subsec:connection-fc-layers}
We clarify the correspondence between our FC formulation in \cref{cornet:eq:riem-fc} and previous FC layers.

\mypara{SPD Manifold.} \citet[Props.~3.4--3.6]{nguyen2024matrix} introduced three SPD FC layers based on the gyrovector spaces under LEM, LCM, and AIM, respectively. These gyro SPD FC layers share the same definition as \cref{cornet:eq:riem-fc}, except that their signed distance and $v_{k}$ are defined by gyrovector spaces.

\mypara{Poincaré Ball.} We show that the Poincaré FC layer $\calF(\cdot): \pball{n} \rightarrow \pball{m}$ reviewed in \cref{app:backbone-hyperbolic-spaces} is also defined as our correlation FC layer in \cref{cornet:def:cor-fc}.

We define the zero vector $\zerovec \in \pball{n}$ as the Poincaré origin, as it is the identity element of the Poincaré gyrovector space \citep{ganea2018hyperbolic}. Obviously, $\{e_k\}_{k=1}^m$ is the orthogonal basis over $T_\zerovec \pball{m}$, where $e_k=\left(\delta_{i k}\right)_{i=1}^m$. Corresponding to \cref{cornet:eq:riem-fc}, we have
\begin{equation} \label{cornet:app:eq:sign-dist-v-k}
    \sign( \inner{\rielog _{\zerovec}(y)}{e_k}) \dist(y, H_{e_k,\zerovec}) = v_k(x).
\end{equation}
Compared with \citet[Eq.~(56)]{shimizu2021hyperbolic}, we only need to show the LHS. The sign can be calculated as
\begin{equation}
    \begin{aligned}
        \sign (\inner{\rielog_{\zerovec} (y)}{e_k}_\zerovec)
        &\stackrel{(1)}{=} \sign \left(4 \inner{ \tanh ^{-1} \left(\sqrt{-K}\|y\| \right) \frac{y}{\sqrt{-K}\|y\|} }{e_k} \right)\\
        &\stackrel{(2)}{=} \sign \left( \inner{y}{e_k} \right)\\
        &= \sign \left( y_k \right).
    \end{aligned}
\end{equation}
The above follows from the following.
\begin{enumerate}
    \item 
    $\lambda^K_{\zerovec} = \frac{2}{1+K \norm{\zerovec}^2}=2$ and $\rielog_{\zerovec} (y) = \tanh ^{-1} \left(\sqrt{-K}\|y\| \right) \frac{y}{\sqrt{-K}\|y\|}$.
    \item 
    $\tanh^{-1}(a)>0, \iff a >0 $. 
\end{enumerate}

Therefore, the LHS of \cref{cornet:app:eq:sign-dist-v-k} is simplified as
\begin{equation} \label{cornet:app:eq:sign-dist-last}
    \begin{aligned}
        \sign( \inner{\rielog _{\zerovec}(y)}{e_k}) \dist(y, H_{e_k,\zerovec})
        &= \sign \left( y_k \right) \dist(y, H_{e_k,\zerovec}) \\
        &\stackrel{(1)}{=} \sign \left( y_k \right) \frac{1}{\sqrt{-K}} \sinh ^{-1}\left(\frac{2 \sqrt{-K} |y_k|}{1 + K \| y \|^2}\right) \\
        &\stackrel{(2)}{=} \frac{1}{\sqrt{-K}} \sinh ^{-1}\left(\frac{2 \sqrt{-K} y_k}{1 + K\| y \|^2}\right).
    \end{aligned}
\end{equation}
The above comes from the following.
\begin{enumerate}
    \item 
    \citet[Thm.~5]{ganea2018hyperbolic}.
    \item 
    $1 + K\| y \|^2 > 0$ by definition and $\sign(a) \sinh^{-1}(|a|) = \sinh^{-1}(a), \forall a \in \bbRscalar$.
\end{enumerate}
The last equation in \cref{cornet:app:eq:sign-dist-last} is the LHS of \citet[Eq.~(56)]{shimizu2021hyperbolic}, indicating the equality.

\mypara{Euclidean Space.} We show that the Euclidean FC layer $\calF(\cdot): \bbR{n} \ni x \rightarrow y=Ax+b \in \bbR{m}$ can also be defined as our correlation FC layer in \cref{cornet:def:cor-fc}.

In Euclidean space $\bbR{n}$, the zero vector $\zerovec \in \bbR{n}$ is the origin, and $\{e_k\}_{k=1}^m$ is the orthogonal basis over $T_\zerovec \bbR{m} \cong \bbR{m}$. Then, the RHS of \cref{cornet:eq:riem-fc} becomes
\begin{equation} \label{cornet:app:eq:euc-fc}
    \begin{aligned}
        v_k(x) 
        &= \inner{x-p_k}{a_k}\\
        &\stackrel{(1)}{=} \inner{x}{z_k} - \gamma_k \norm{z_k}.
    \end{aligned}        
\end{equation}
where (1) comes from $\rieexp_{\zerovec} \left(\gamma_k [z_k]\right) = \gamma_k [z_k]$ and $\pt{\zerovec}{p_k}(z_k)=z_k$. The above takes the form of $\inner{x}{a_k}+b_k$.

On the other hand, the LHS of \cref{cornet:eq:riem-fc} becomes
\begin{equation}
    \begin{aligned}
        \sign(\inner{\rielog_{\zerovec}(y)}{e_k})\dist(y, H_{e_k,\zerovec})
        &\stackrel{(1)}{=} \sign(y_k) \dist(y, H_{e_k,\zerovec}) \\
        &\stackrel{(2)}{=} y_k. \\
    \end{aligned}
\end{equation}
The above comes from the following.
\begin{enumerate}
    \item 
    $\rielog_{\zerovec}(y)=y$ and $\inner{y}{e_k}=y_k$.
    \item 
    $\dist(y, H_{e_k,\zerovec}) = \frac{| \inner{y}{e_k}|}{\norm{e_k}}=|y_k|$.
\end{enumerate}

\subsubsection{Log-Euclidean Layers under Product Geometry}
\label{cornet:app:subsec:lem-fc-product}
We first review some basic facts of the product geometry, and then discuss the Log-Euclidean correlation MLR and FC layer under the product geometry. 

\mypara{Product of Correlation Manifolds.} Given a manifold $(\calM,g)$, the $n$-fold product is $(\calM^n,g)=\prod_{i=1}^n (\calM,g)$. Each point and tangent vector over $\calM^n$ are
\begin{align}
    \calM^n \ni P &= (P_1 \in \calM, \cdots, P_n \in \calM), \\
    T_P \calM^n \ni V &= (V_1 \in T_{P_1}\calM, \cdots, V_n \in T_{P_n}\calM).
\end{align}
The product metric is 
\begin{equation}
    \inner{V}{W}_P = \sum_{i=1}^n \inner{V_i}{W_i}_{P_i}, \quad \forall V, W \in T_P \calM^n.
\end{equation}

\mypara{Correlation MLR.}  Following \cref{cornet:thm:flat-mlr}, the logit for the $k$-th class and the input $\boldsymbol{X}=\{X_l \in \cor{n}\}_{l=1}^c \in (\cor{n})^c$ is
\begin{equation} \label{cornet:app:eq:cor-mlr-product}
    \begin{aligned}
    v_k(\boldsymbol{X})
    &= \sum_{l=1}^{c} v_{kl}(X_l;Z_{kl},\gamma_{kl}) \\
    &= \sum_{l=1}^{c}\left(
    \left\langle \phi_n(X_l),(\phi_n)_{*,I_n}(Z_{kl})\right\rangle
    -\gamma_{kl}\norm{(\phi_n)_{*,I_n}(Z_{kl})}
    \right),
    \end{aligned}
\end{equation}
where $d_n=n(n-1)/2$, $\phi_n:\cor{n}\to\bbR{d_n}$ is the diffeomorphism of the selected Log-Euclidean geometry, $\boldsymbol{I}=(I_n,\cdots,I_n)$, $Z_k=(Z_{k1},\cdots,Z_{kc})\in T_{\boldsymbol{I}}(\cor{n})^c$, $Z_{kl}\in T_{I_n}\cor{n}$, and $\gamma_{kl}\in\bbRscalar$. Each $v_{kl}$ is the response of the $l$-th component given by \cref{cornet:thm:flat-mlr}, with the unit direction $[Z_{kl}]=Z_{kl}/\norm{Z_{kl}}_{I_n}$.

\mypara{Correlation FC Layer.} Following \cref{cornet:app:lem:flat-fc,cornet:app:eq:cor-mlr-product}, the FC layer $\calF(\cdot): (\cor{n})^c \to \cor{m}$ for the input $\boldsymbol{X}$ is
\begin{equation} \label{cornet:app:eq:v-k-product}
Y = \phi_m^{-1}\left( \sum_{i=1}^{d_m} \sum_{l=1}^{c} v_{il}(X_l;Z_{il},\gamma_{il}) e_i \right),
\end{equation}
where $d_m=m(m-1)/2$, $\phi_m:\cor{m}\to\bbR{d_m}$ is the output diffeomorphism, $Z_{il} \in T_{I_n}\cor{n}\cong\hol{n}$, and $\gamma_{il} \in \bbRscalar$.

\cref{cornet:app:eq:v-k-product} implies that $\calF(\cdot): (\cor{n})^c \to \cor{m}$ differs from $\calF(\cdot): \cor{n} \to \cor{m}$ only in each scalar response $v_i$, where the former is a summation over the $c$ input components. For example, considering the FC layer $\calF(\cdot): (\cor{n})^c \to \cor{m}$ under ECM, its matrix-coordinate response $v_{ij}$ for the input $\boldsymbol{C} = \{C_l \in \cor{n}\}_{l=1}^c$ is
\begin{equation}
v_{ij}(\boldsymbol{C}) = \sum_{l=1}^c v_{ijl}^{\EC}(C_l;Z_{ijl},\gamma_{ijl}),
\end{equation}
where $Z_{ijl} \in \hol{n}$ and $\gamma_{ijl} \in \bbRscalar$, for $i,j=1,\cdots, m$ with $i > j$, and $1 \leq l \leq c$.

\subsection{Adaptive Log-Euclidean Metrics}
\label{alem:app:additional-discussions}

\subsubsection{Well-definedness of General Matrix Logarithm}
\label{alem:app:subsec:well_difined_glog}

\cref{alem:eq:mlog} does not explicitly specify the correspondence between eigenvalues and diagonal logarithms.
Here, we present a detailed explanation.
In implementations such as PyTorch or MATLAB, eigendecomposition routines return eigenpairs in a prescribed order.

We rewrite the eigendecomposition as $S = \sum \sigma_i E_i$, where $E_i=u_i u_i^\top$ and $u_i$ is the corresponding eigenvector in $U$. Let $S$ be an $n \times n$ SPD matrix, and let $P_n$ be the permutation group on $\{1,\ldots,n\}$. Changing the order of $\{1,\ldots,n\}$ can be viewed as a permutation, so we use $\pi \in P_n$ to represent the corresponding changed order.

Assume that the eigenvalues $\sigma_i$ are sorted in ascending order, \emph{i.e.,} $\sigma_1 \leq \cdots \leq \sigma_n$. We use ``the $i$-th eigenvalue'' to refer to the eigenvalue in the $i$-th pair of the ordered eigenpair sequence $(\sigma_1,u_1),\ldots,(\sigma_n,u_n)$. Since each eigenvector $u_{i}$ is unique, it is safe to say the eigenvalues are ordered, and the $i$-th eigenvalue/eigenvector pair is unique.

Let $\log_\alpha(\Sigma)$ denote applying the scalar logarithm $\log_{a_i}$ to the $i$-th eigenvalue $\sigma_i$. Then $\log_\alpha(S)$ is rewritten as $\log_\alpha(S)=\sum \log_{a_i}(\sigma_i) E_i$, where $S = \sum \sigma_i E_i$. In this way, $\log_\alpha$ is clearly well-defined. By definition, we can observe that the output of $\log_\alpha$ does not depend on the order in eigendecomposition.

Suppose there are two eigendecompositions with different orders, \emph{i.e.,} $S= U\Sigma U^\top = \tilde{U} \tilde{\Sigma}\tilde U^\top$, where $\tilde{U}$ and $\tilde\Sigma$ are rearrangements of $U$ and $\Sigma$. There exists a $\pi \in P_n$ such that, for each $j$, there is a unique $i$ satisfying $\tilde u _j = u_{\pi(i)}$ and $\tilde\sigma_j=\sigma_{(i)}$. We then have $\sum \log_{a_i}(\sigma_i) E_i$ for $S= U\Sigma U^\top$ and $\sum \log_{a_{\pi(i)}}(\sigma_{\pi(i)}) E_{\pi(i)}$ for $S=\tilde U\tilde{\Sigma}\tilde U^\top$, which indicates that the two eigendecompositions are equivalent.

\subsection{Product Cholesky Metrics}
\label{pcm:app:additional-discussions}

\subsubsection{Riemannian Operators under \texorpdfstring{$\defDGBWM$}{(theta, M)-DBWM}}
\label{pcm:app:subsubsec:riemannian-operators}

We first define a map $\phi_{\theta}: \chospace{n} \rightarrow \chospace{n}$ as
\begin{equation} \label{pcm:eq:diag_power_deformation}
    \phi_{\theta}(L) = \trilL + \bbL^\theta, \forall L \in \chospace{n}.
\end{equation}
Its differential at $L \in \chospace{n}$ is given by
\begin{equation} \label{pcm:eq:diff_diag_power}
    \left(\phi_\theta\right)_{*,L}(X) = \trilX + \theta \bbL^{\theta-1} \bbX, \quad \forall X \in T_L\chospace{n}.
\end{equation}

Let $\gDGBW$ and $\gdefDGBW$ be $\DGBWM$ and $\defDGBWM$, respectively.
Since constant scaling of a Riemannian metric preserves the Christoffel symbols, Riemannian operators such as the Riemannian logarithm, exponential map, and parallel transport under $\gdefDGBW$ are the same as those under the pullback metric $\phi_{\theta}^*\gDGBW$.
Following \cref{def:ch2-riemannian-isometry}, these Riemannian operators under $\defDGBWM$ can be obtained by $\phi_{\theta}^*\gDGBW$.
Besides, as constant scaling does not affect the weighted Fréchet mean (WFM), the WFM under $\defDGBWM$ is the same as that under $\phi_{\theta}^* \gDGBW$.
The latter can be calculated by the properties of isometries presented in \cref{def:ch2-riemannian-isometry}.
Therefore, by the properties of isometry in \cref{def:ch2-riemannian-isometry} and \cref{pcm:eq:diag_power_deformation,pcm:eq:diff_diag_power}, we can obtain all the Riemannian operators.

    \chapter{Proofs}
\label{app:proofs}

\section{Mathematical Background}
\label{app:ch2-proofs}

\subsection[Proof of \cref{prop:ch2-isomorphism-preserves-properties}]{Proof of \cref{prop:ch2-isomorphism-preserves-properties}}
\linkofproof{prop:ch2-isomorphism-preserves-properties}
\begin{proof}
The gyration-preservation property is shown by \citet[Eq.~(6.323)]{ungar2022analytic}. The proofs for the gyro identity and gyroinverse follow directly from the isomorphism and the uniqueness of inverse and identity \citep[Thm.~2.10]{ungar2022analytic}.
\end{proof}

\section{Lie Group Batch Normalization}
\label{app:liebn-proofs}

\subsection[Proof of \cref{props:population_gaussian}]{Proof of \cref{props:population_gaussian}}
\label{prf:population_gaussian}
\linkofproof{population_gaussian}

\begin{proof}
    \mypara{\cref{pro:mle_m}.} The MLE of $M$ is
    \begin{equation}
        \begin{aligned}
            M_{\mle}
            &= \argmax \log(k(v)) - \sum_{i=1}^N \frac{\dist(P_i,M)^2}{2v^2}\\
            &= \argmin \sum_{i=1}^N \dist(P_i,M)^2.
        \end{aligned}
    \end{equation}

    \mypara{\cref{pro:hom}.} We denote $Y=\ltrans_B(X)$, and $p_X$ and $p_Y$ as the density of $X$ and $Y$, respectively.
    The density of  $Y$ is
    \begin{equation}
        \begin{aligned}
            p_{Y}(Q)
            &\stackrel{(1)}{=}p_X(\ltrans _{\ominus B}(Q))\\
            &= k(\sigma) \exp \left(-\frac{\dist(\ltrans_{\ominus B}(Q), M)^2}{2 \sigma^2}\right)\\
            &\stackrel{(2)}{=} k(\sigma) \exp \left(-\frac{\dist(Q, \ltrans_{B}(M))^2}{2 \sigma^2}\right).
    \end{aligned}
    \end{equation}
    The above comes from:
    \begin{enumerate}
        \item
        \citet[Thm.~7]{pennec2004probabilities};
        \item
        The isometry of the left translation.
    \end{enumerate}
\end{proof}

\subsection[Proof of \cref{props:samples}]{Proof of \cref{props:samples}}
\label{prf:samples}
\linkofproof{samples}

\begin{proof}
    The isometry of $\ltrans _B$ directly implies the homogeneity of the sample mean.
    Now let us focus on \cref{eq:variance_lie_group}.
    We have the following:
    \begin{equation}
        \begin{aligned}
            \sum\nolimits_{i=1}^N  w_i \dist^2(\phi_{s}(P_i), E)
            &= \sum\nolimits_{i=1}^N w_i \|s\rielog_E P_i\|_E^2\\
            &= s^2\sum\nolimits_{i=1}^N w_i \|\rielog_E P_i\|^2_E\\
            &= s^2\sum\nolimits_{i=1}^N w_i \dist^2(P_i, E),
    \end{aligned}
    \end{equation}
    where $\| \cdot \|_{E}$ is the norm on $T_E\calM$.
\end{proof}

\subsection[Proof of \cref{props:core_operation_right}]{Proof of \cref{props:core_operation_right}}
\label{prf:core_operation_right}
\linkofproof{core_operation_right}

\begin{proof}
    As the right-invariant metric has properties analogous to those of the left-invariant metric, this proof follows the same logic as the two proofs above.

    \mypara{Gaussian homogeneity.} We denote $Y=\rtrans_B(X)$, and $p_X$ and $p_Y$ as the density of $X$ and $Y$, respectively.
    The density of  $Y$ is
    \begin{equation}
        \begin{aligned}
            p_{Y}(Q)
            &\stackrel{(1)}{=}p_X(\rtrans _{\ominus B}(Q))\\
            &= k(\sigma) \exp \left(-\frac{\dist(\rtrans_{\ominus B}(Q), M)^2}{2 \sigma^2}\right)\\
            &\stackrel{(2)}{=} k(\sigma) \exp \left(-\frac{\dist(Q, \rtrans_{B}(M))^2}{2 \sigma^2}\right).
    \end{aligned}
    \end{equation}
    The above comes from:
    \begin{enumerate}
        \item
        \citet[Thm.~13]{pennec2004probabilities};
        \item
        The isometry of the right translation.
    \end{enumerate}

    \mypara{Sample mean homogeneity.} This is a direct corollary of the isometry of right translation.
\end{proof}

\subsection[Proof of \cref{prop:liebn_natural_extension_ebn}]{Proof of \cref{prop:liebn_natural_extension_ebn}}
\label{prf:liebn_natural_extension_ebn}
\linkofproof{liebn_natural_extension_ebn}

\begin{proof}
    As $\bbR{n}$ is an abelian group and the Euclidean inner product is bi-invariant, we focus on left translation in the following.
    The core of this proof lies in the fact that on $\bbR{n}$,
    (1) the Fréchet mean and variance are reduced to the familiar Euclidean statistics.
    (2) the calculation of the running mean becomes the weighted arithmetic mean.
    (3) \cref{eq:liebn_centering,eq:liebn_scaling,eq:liebn_biasing} become \cref{eq:ebn}.
    We prove these three points one by one.

    As stated by \citet[Prop.~G.1 and Cor.~G.2]{lou2020differentiating}, from the view of the product manifold, the element-wise Fréchet mean and variance on $\bbR{n}$ are equivalent to the vector-valued Euclidean mean and variance.

    Besides, by an argument similar to that of \citet[Prop.~G.1]{lou2020differentiating}, the weighted Fréchet mean on $\bbR{n}$ is simplified to the weighted arithmetic average.
    Therefore, on $\bbR{n}$, the calculation of running statistics in our \cref{alg:liebn} becomes the familiar moving average.

    Thirdly, on $\bbR{n}$, we know that $\ltrans _x(y)=x+y$, $\rieexp_x v=x+v$, $\rielog_x y=y-x$, and the neutral element is $0$.
    Since statistics, as well as the Euclidean BN, are calculated element-wise, it suffices to consider a single coordinate, \ie $\bbRscalar$.
    For a batch of activations $\{x_i\}_{i=1}^N \subset \bbRscalar$ with batch mean $\mu_b$ and batch variance $v^2_b$, \cref{eq:liebn_centering,eq:liebn_scaling,eq:liebn_biasing} are rewritten as follows:
    \begin{equation}
        \ltrans_{\beta}\left( \rieexp_0 \left[\frac{\gamma}{\sqrt{v^2_{b}+\epsilon}}\rielog_0(\ltrans_{-\mu_b}(x_i)) \right]\right)
        = \gamma \frac{x_i-\mu_{b}}{\sqrt{v^2_{b}+\epsilon}} + \beta.
    \end{equation}
    The above equation is the exact core computation of the standard Euclidean BN.
\end{proof}

\subsection[Proof of \cref{prop:spd_param_lem_lcm_deformation}]{Proof of \cref{prop:spd_param_lem_lcm_deformation}}
\label{prf:spd_param_lem_lcm_deformation}
\linkofproof{spd_param_lem_lcm_deformation}

\begin{proof}
    We first prove the case of $\triparamLEM$, and then proceed to the case of $\paramLCM$.

    \mypara{$\triparamLEM$.} For clarity, we denote the metric tensor of $\triparamLEM$ as
    \begin{equation}
        \gtriparamLE = \frac{1}{\theta^2}\pow_\theta^*\gbiparamLE,
    \end{equation}
    where $\gbiparamLE$ is the metric tensor of $\biparamLEM$.
    Let $P \in \spd{n}$ and $V,W \in T_P\spd{n}$. Then we have
    \begin{equation}
        \begin{aligned}
            \gtriparamLE_P(V,W)
            &= \frac{1}{\theta^2} \gbiparamLE_{\pow_{\theta}(P)} \left((\pow_\theta)_{*,P}(V),(\pow_\theta)_{*,P}(W) \right)\\
            &= \frac{1}{\theta^2} \langle \left( \mlog \circ \pow_{\theta} \right)_{*,P} (V), \left( \mlog \circ \pow_{\theta} \right)_{*,P} (W) \rangle^{\alphabeta}\\
            &= \langle \mlog_{*,P} (V), \mlog_{*,P} (W) \rangle^{\alphabeta}\\
            &= \gbiparamLE_P(V,W).
        \end{aligned}
    \end{equation}

    \mypara{$\paramLCM$.} Let us first review a well-known fact of deformed metrics \citep{thanwerdas2022geometry}.
    Let $\tilde{g}=\frac{1}{\theta^2}\pow_{\theta}^*g$ be the power-deformed metric on SPD.
    Then when $\theta$ tends to 0, for all $P \in \spd{n}$ and all $V \in T_P\spd{n}$, we have
    \begin{equation} \label{eq:deformed_g_lim}
        \tilde{g}_P(V,V) \to g_{I}(\log_{*,P}(V),\log_{*,P}(V)).
    \end{equation}

    By \cref{eq:deformed_g_lim}, we can readily obtain the results.
\end{proof}

\subsection[Proof of \cref{prop:spd_param_invariance}]{Proof of \cref{prop:spd_param_invariance}}
\label{prf:spd_param_invariance}
\linkofproof{spd_param_invariance}

\begin{proof}
    $\biparamAIM$ is left-invariant \citep{thanwerdas2022theoretically}.
    As the pullback of $\biparamAIM$, $\triparamAIM$ is left-invariant as well.
    Besides, \cref{alem:thm:rethk_lem_lcm} shows that LCM is the pullback metric from the Euclidean space of $\trilspace{n}$.
    Therefore, $\theta$-LCM is bi-invariant.
\end{proof}

\subsection[Proof of \cref{thm:crim}]{Proof of \cref{thm:crim}}
\label{prf:crim}
\linkofproof{crim}

\begin{proof}
    In the following, we denote the Riemannian operators under the left-invariant metric, \ie AIM, by $\dist^{\mrL}$, $\rielog^{\mrL}$, and $\rieexp^{\mrL}$. Note that the Cholesky decomposition pulls back the group operation of matrix product from the Cholesky manifold $\chospace{n}$ \citep{thanwerdas2022theoretically}. For simplicity, we abbreviate $\oplusLieAI$ as $\oplus$.

    The differential maps of the Cholesky decomposition and its inverse are reviewed in \cref{eq:ch2-cholesky-differentials}. Following the notation in this theorem, we further denote $X \in T_L\chospace{n}$.
    Specifically, for the differential map at $I$, we have
    \begin{align}
        \label{eq:diff_chol_at_i}
         \chol_{*,I} (V) &= \left( V \right) _{\frac{1}{2}}, \forall V \in T_I\spd{n}.\\
         \label{eq:diff_chol_inv_at_i}
         (\chol^{-1})_{*,I} (X) &= \symmetrizeSum{X}, \forall X \in T_I\chospace{n}.
    \end{align}
    Denoting $\widetilde{\ltrans}$ and $\widetilde{\rtrans}$ as the group translation on the Cholesky manifold $\chospace{n}$, we have the following w.r.t. the differential maps of left and right translation:
    \begin{align}
        \left((\ltrans_P)_{*,\ominus P} \right)^{-1}
        & \stackrel{(1)}{=} \left( \left( \chol^{-1} \right)_{*, I}
        (\widetilde{\ltrans}_L)_{*,L^{-1}} \circ \chol_{*, \ominus P} \right)^{-1}\\
        \label{eq:diff_ltrans}
        & = \left( \chol_{*, \ominus P} \right)^{-1} \left( (\widetilde{\ltrans}_L)_{*,L^{-1}} \right)^{-1} \circ \chol_{*, I},\\
        \label{eq:diff_rtrans}
        (\rtrans_{\ominus P})_{*,P}
        & \stackrel{(2)}{=} \left(\chol^{-1}\right)_{*, I} \circ (\widetilde{\rtrans}_{L^{-1}})_{*,L} \circ \chol_{*, P}.
    \end{align}
    The above derivation comes from the following:
    \begin{enumerate}
        \item
        $\ltrans _P =  \chol^{-1} \circ \widetilde{\ltrans}_L \circ \chol$;
        \item
        $\rtrans _{\ominus P} =  \chol^{-1} \circ \widetilde{\rtrans}_{L^{-1}} \circ \chol$.
    \end{enumerate}
    \mypara{Riemannian metric.} For the differential of right translation, we have the following:
    \begin{equation}
        \label{eq:diff_rtrans_spd}
        \begin{aligned}
            (\rtrans_{\ominus P})_{*,P} (V)
            & = \left(\chol^{-1}\right)_{*, I} \circ (\widetilde{\rtrans}_{L^{-1}})_{*,L} \circ \chol_{*, P} (V) \\
            & = \symmetrizeSum{ L(L^{-1} V L^{-\top})_{\frac{1}{2}} L^{-1}}. \\
        \end{aligned}
    \end{equation}
    By \cref{eq:diff_rtrans_spd}, one can obtain the expression for the Riemannian metric tensor.

    \mypara{Riemannian geodesic and exponential map.} According to \citet{zacur2014left}, we have the following for the operators between left- and right-invariant metrics:
    \begin{align}
        \label{eq:exp_l2r_start}
        \rieexp _{P} (V) &= \ominus\left\{ \rieexp^{\mrL}_ {\ominus P} \left( -\left((\ltrans_P)_{*,\ominus P} \right)^{-1} \circ (\rtrans_{\ominus P})_{*,P} (V) \right) \right\},\\
        \label{eq:dist_l2r}
        \dist(P,Q) &= \dist^{\mrL} \left( \ominus P, \ominus Q\right).
    \end{align}
    Putting the AIM-based geodesic distance into the RHS of \cref{eq:dist_l2r}, one can obtain the geodesic distance under CRIM.

    Now, we simplify \cref{eq:exp_l2r_start}.
    Putting \cref{eq:diff_ltrans,eq:diff_rtrans} into \cref{eq:exp_l2r_start}, we have the following:
    \begin{equation} \label{eq:exp_crim_derivation}
        \begin{aligned}
        \rieexp_{P}(V)
        & = \ominus\left\{ \rieexp^{\mrL}_{\ominus P} \left(- \left( \chol_{*, \ominus P} \right)^{-1} \circ \left( (\widetilde{\ltrans}_L)_{*,L^{-1}} \right)^{-1} \circ (\widetilde{\rtrans}_{L^{-1}})_{*,L} \circ \chol_{*, P} (V) \right) \right\} \\
        & = \ominus\left\{ \rieexp^{\mrL}_{\ominus P} \left(- \left( \chol_{*, \ominus P} \right)^{-1} \left[ L^{-1}  \chol_{*, P} (V) L^{-1} \right] \right) \right\} \\
        & \stackrel{(1)}{=} \ominus\left\{ \rieexp^{\mrL}_{\ominus P} \left(- \left( \chol_{*, \ominus P} \right)^{-1} \left[ \left(L^{-1} V L^{-\top}\right)_{\frac{1}{2}} L^{-1} \right] \right) \right\} \\
        & \stackrel{(2)}{=} \ominus\left\{ \rieexp^{\mrL}_{\ominus P} \left(- \symmetrizeSum{ \left(L^{-1} V L^{-\top}\right)_{\frac{1}{2}} L^{-1} L^{-\top}} \right) \right\}.\\
        \end{aligned}
    \end{equation}
    The above comes from the following:
    \begin{enumerate}
        \item
        The first identity follows from \cref{eq:ch2-cholesky-differentials};
        \item
        The second identity follows from \cref{eq:ch2-cholesky-differentials}.
    \end{enumerate}

    \mypara{Riemannian logarithm.} From the second equality in \cref{eq:exp_crim_derivation}, we have the following:
    \begin{equation} \label{eq:log_crim_derivation}
        \begin{aligned}
            \rielog _P (Q)
            &= - \chol_{*, P}^{-1} \left\{ L\chol_{*, \ominus P} \left(\rielog ^\mrL _{\ominus P} \left(\ominus Q \right) \right) L \right\}\\
            &\stackrel{(1)}{=} - \chol_{*, P}^{-1} \left\{ \left(L \widetilde{V} L^\top \right)_{\frac{1}{2}} L \right\}\\
            &= -\symmetrizeSum{L L^{\top} \left( L \widetilde{V}L^{\top} \right)_{\frac{1}{2}}^{\top}}
        \end{aligned}
    \end{equation}
    The above comes from the following:
    \begin{enumerate}
        \item
        $\chol_{*,\ominus P} (V) = L^{-1} \left(L V L^\top\right)_{\frac{1}{2}}, \forall V \in T_{\ominus P}\spd{n}$.
    \end{enumerate}
\end{proof}

\subsection[Proof of \cref{cor:crim_geodesic}]{Proof of \cref{cor:crim_geodesic}}
\label{prf:crim_geodesic}
\linkofproof{crim_geodesic}

\begin{proof}
    \mypara{Completeness.} \cref{eq:exp_crim} indicates that $\rieexp _{I}$ is defined over the whole $T _I \spd{n}$. Besides, the SPD manifold is connected \citep{pennec2006riemannian}. By \citet[Cor.~6.20]{lee2018introduction}, CRIM is complete.

    \mypara{Geodesic.} For simplicity, we abbreviate $\oplusLieAI$ as $\oplus$. The geodesic connecting $P$ and $Q$ can be obtained as follows:
    \begin{equation}
        \begin{aligned}
             \gamma{(t; P,Q)}
             &= \rieexp _{P} \left( t\rielog _{P} (Q)\right)\\
             &\stackrel{(1)}{=} \ominus\left\{ \rieexp^{\mrL}_{\ominus P} \left(- \left( \chol_{*, \ominus P} \right)^{-1} \left[ L^{-1}  \chol_{*, P} \left(t \rielog _P (Q) \right) L^{-1} \right] \right) \right\} \\
             &\stackrel{(2)}{=} \ominus\left\{ \rieexp^{\mrL}_{\ominus P} \left( t\rielog ^\mrL _{\ominus P} \left(\ominus Q \right) \right) \right\} \\
             &= \ominus\left\{ \gamma^{\mathrm{AI}}(t; \widetilde{P},\widetilde{Q}) \right\}. \\
        \end{aligned}
    \end{equation}
    The above comes from the following:
    \begin{enumerate}
        \item
        The second equality in \cref{eq:exp_crim_derivation};
        \item
        The first equality in \cref{eq:log_crim_derivation}.
    \end{enumerate}
\end{proof}

\subsection[Proof of \cref{thm:liebn_pullback}]{Proof of \cref{thm:liebn_pullback}}
\label{prf:liebn_pullback}
\linkofproof{liebn_pullback}

\begin{proof}
    Without loss of generality, we focus on the case of the left-invariant metric. The results for the right-invariant metric can be proven similarly.

     We denote \cref{eq:liebn_centering,eq:liebn_scaling,eq:liebn_biasing} on $\calM_i, i=1,2$ as the mapping $\xi^i(\cdot | M,v^2,B,s)$.
     Let $\calB=\{P_i\}_{i=1}^N$ and $f(\calB)=\{f(P_i)\}_{i=1}^N$.

    The core of this proof lies in three points:
    \begin{enumerate}
        \item
        The Fréchet mean and variance of $\calB$ in $\calM_1$ correspond to the counterparts of $f(\calB)$ in $\calM_2$.
        \item
        $\xi^1(P_i|M,v^2,B,s)$ in $\calM_1$ is equal to $f^{-1} (\xi^2(f(P_i)|f(M),v^2,f(B),s))$.
        \item
        The updates of running statistics in $\calM_1$ correspond to the counterparts in $\calM_2$.
    \end{enumerate}

    We denote $M$ as the Fréchet mean of $\calB$, and $v^2$ as the Fréchet variance of $\calB$.
    Then, by the isometry of $f$, the Fréchet mean and variance of $f(\calB)$ are $f(M)$ and $v^2$, respectively.

    On $\calM_i, i=1,2$, we denote $\ltrans^i, \oplus^i, \rieexp^i,\rielog^i$ as the Lie group and Riemannian operators, $E^i$ as the neutral element, and \cref{eq:liebn_scaling} as $\phi^i_s(\cdot)$.
    With the isometry and Lie group isomorphism of $f$, we have the following equations:
    \begin{align}
        \ltrans^1_{\ominus^1 M}
        &=f^{-1} \circ \ltrans^2_{\ominus^2 f(M)} \circ f,\\
        \phi^1_{s} &=\rieexp^1_{E^1}\left[ s \rielog^1_{E^1}(\cdot) \right]\\
        &= f^{-1} \left( \rieexp^{2}_{E^2}\left[ s \rielog^2_{E^2}(f(\cdot))\right]\right)\\
        &= f^{-1} \circ \phi^2_s \circ f,\\
        \ltrans^1_{B}
        &=f^{-1} \circ \ltrans^2_{f(B)} \circ f.
    \end{align}
    Then we have
    \begin{equation} \label{eq:core_ops_pm}
        \xi^1(P_i|M,v^2,B,s)=f^{-1}(\xi^2(f(P_i)|f(M),v^2,f(B),s))
    \end{equation}

    Lastly, we show the correspondence between running statistics.
    Since the Fréchet variance is the same for both $\calB$ and $f(\calB)$, we focus on the running mean.
    Let $M_r$ and $f(M_r)$ denote the initial values of the running means in $\calM_1$ and $\calM_2$, respectively, and let $\wfm^i$ represent the weighted Fréchet mean in $\calM_i$.
    Then the updated running mean in $\calM_1$ is
    \begin{equation}
        \wfm^1(\{1-\eta,\eta\},\{M_r,M\})=
        f^{-1}(\wfm^2(\{1-\eta,\eta\},\{f(M_r),f(M)\}))
    \end{equation}
    We can further simplify the above equation as
    \begin{equation} \label{eq:wfm_pm}
        \wfm^1 = f^{-1} \circ \wfm^2 \circ f
    \end{equation}

    Denoting $\liebn^i$ as the LieBN algorithm on $\calM_i$, \cref{eq:core_ops_pm} and \cref{eq:wfm_pm} imply that
    \begin{equation}
    \liebn^1(P_i|B,s,\epsilon,\eta)=f^{-1} \left[\liebn^2(f(P_i)|f(B),s,\epsilon,\eta) \right].
    \end{equation}
\end{proof}

\section{Gyrogroup Batch Normalization}
\label{app:gyrobn-proofs}

\label{app:sec:gyrobn-proofs-source}

\linkofproof{prop:grassmannian_pseudo_reductive_gyrogroups}
\subsection[Proof of \cref{prop:grassmannian_pseudo_reductive_gyrogroups}]{Proof of \cref{prop:grassmannian_pseudo_reductive_gyrogroups}}
\begin{proof}
    By \citet[Lem.~2.3]{nguyen2023building}, easy computations show that \cref{eq:pseudo_reduction} holds for $\grasonb{p,n}$ if and only if it holds for $\graspp{p,n}$.
    Without loss of generality, we prove the case for the projector perspective.

    Given any $P, Q\in \graspp{p,n}$, \citet[Def.~3.18]{nguyen2022gyro} gives the expression for gyration:
    \begin{equation}
        \gyr[\ominus P,  P] Q=F(\ominus P, P) Q\left(F(\ominus P, P)\right)^{-1},
    \end{equation}
    with $F(\ominus P, P)$ defined as
    \begin{equation}
        F(\ominus P, P)= \mexp \left(-\left[\overline{\ominus P \oplus  P}, \idpp\right]\right) \mexp \left(\left[\overline{\ominus P}, \idpp\right]\right) \mexp \left(\left[\overline{P}, \idpp\right]\right),
    \end{equation}
    where $\overline{(\cdot)} = \rielog_{\idpp}(\cdot)$.
    This equation can be further simplified as
    \begin{equation}
        \begin{aligned}
            F(\ominus P, P)
            &\stackrel{(1)}{=} \mexp (0) \mexp \left(\left[\overline{\ominus P}, \idpp\right]\right) \mexp \left(\left[\overline{P}, \idpp\right]\right) \\
            &\stackrel{(2)}{=} \mexp \left(\left[\overline{\ominus P}, \idpp\right]\right) \mexp \left(\left[\overline{P}, \idpp\right]\right) \\
            &\stackrel{(3)}{=} I_{n}.
        \end{aligned}
    \end{equation}
    The above derivation follows from
    \begin{enumerate}
        \item
        $\overline{\ominus P \oplus P} = \overline{\idpp} = \bbzero_{n\times n} \in \bbR{n \times n}.$
        \item
        $\mexp (0) = I_n$.
        \item
        $\overline{\ominus P}= - \overline{P}$ and $\mexp \left(\left[-\overline{P}, \idpp\right]\right)=\mexp \left(\left[\overline{P}, \idpp\right]\right)^{-1}$.
    \end{enumerate}

    Therefore, $\gyr[\ominus P, P]$ is the identity map.
\end{proof}

\linkofproof{thm:pseudo_reductive_gyrogroups_properties}
\subsection[Proof of \cref{thm:pseudo_reductive_gyrogroups_properties}]{Proof of \cref{thm:pseudo_reductive_gyrogroups_properties}}
\label{app:subsec:prf:pseudo_reductive_gyrogroups_properties}
\begin{proof}
This theorem follows from \citet[Thms.~2.10--2.11]{ungar2022analytic}, which presents some useful properties for gyrogroups.
We argue that all the properties except $\gyr[a, a]=\id$ are independent of the left reduction law (G4), and are therefore satisfied on pseudo-reductive gyrogroups.
All the properties can be proven in the same way as in \citet[Thms.~2.10--2.11]{ungar2022analytic}. We summarize the logic in the following:
\begin{itemize}
    \item
    left gyroassociativity $\Rightarrow$ \cref{enu:prgp_1}
    \item
    left gyroassociativity + \cref{enu:prgp_1} $\Rightarrow$ \cref{enu:prgp_2}
    \item
    definition $\Rightarrow$ \cref{enu:prgp_3}
    \item
    left gyroassociativity + \cref{enu:prgp_1} + \cref{enu:prgp_3}  $\Rightarrow$ \cref{enu:prgp_4}
    \item
    definition $\Rightarrow$ \cref{enu:prgp_5}
    \item
    left gyroassociativity + (G2) + \cref{enu:prgp_1} + \cref{enu:prgp_3} + \cref{enu:prgp_4} + \cref{enu:prgp_5} $\Rightarrow$ \cref{enu:prgp_6}
    \item
    \cref{enu:prgp_1}+\cref{enu:prgp_6} $\Rightarrow$ \cref{enu:prgp_7}
    \item
    left gyroassociativity +\cref{enu:prgp_3} $\Rightarrow$ \cref{enu:prgp_8}
    \item
    left gyroassociativity + left cancellation in \cref{enu:prgp_8} $\Rightarrow$ \cref{enu:prgp_9}
    \item
    gyro identity in \cref{enu:prgp_9} $\Rightarrow$ \cref{enu:prgp_10}
    \item
    \cref{enu:prgp_10} $\Rightarrow$ \cref{enu:prgp_11}
    \item
    left cancellation in \cref{enu:prgp_8} + gyro identity in \cref{enu:prgp_9}   $\Rightarrow$ \cref{enu:prgp_12}
    \item
    left cancellation in \cref{enu:prgp_8} + gyro identity in \cref{enu:prgp_9} $\Rightarrow$ \cref{enu:prgp_13}
\end{itemize}
\end{proof}

\linkofproof{lem:gyro_geodesic_dist}
\subsection[Proof of \cref{lem:gyro_geodesic_dist}]{Proof of \cref{lem:gyro_geodesic_dist}}
\begin{proof}
    \begin{equation}
        \begin{aligned}
          y &\stackrel{(1)}{=}  x \oplus ( \ominus x \oplus y) \\
          &\stackrel{(2)}{=} \rieexp_{x} \left( \pt{e}{x} \left( \rielog_e \left( \ominus x \oplus y   \right)\right)\right)\\
          &\stackrel{(3)}{\Rightarrow} \rielog_{x}(y) = \pt{e}{x} \left( \rielog_e \left( \ominus x \oplus y   \right)\right).
        \end{aligned}
    \end{equation}
        The above comes from the following.
    \begin{enumerate}
        \item
        Left cancellation law.
        \item
        Definition of gyroaddition.
        \item
        Applying $\rielog_x(\cdot)$ to both sides.
    \end{enumerate}
    By the last equation, we have
    \begin{equation}
        \begin{aligned}
            \dist(x,y)
            &= \norm{\rielog_x (y)}_x \\
            &= \norm{\pt{e}{x} \left( \rielog_e \left( \ominus x \oplus y   \right)\right)}_x \\
            &\stackrel{(1)}{=} \norm{ \rielog_e \left( \ominus x \oplus y \right) }_e \\
            &= \gyrdist(x,y),
        \end{aligned}
    \end{equation}
    where (1) comes from
    \begin{itemize}
        \item
        Parallel transport preserving the norm \citep[Sec.~3.1]{do1992riemannian}.
        \item
        $\pt{x}{e} \circ \pt{e}{x}(v)=v, \forall v \in T_e\calM$.
    \end{itemize}
\end{proof}

\linkofproof{lem:isometry_for_gyro}
\subsection[Proof of \cref{lem:isometry_for_gyro}]{Proof of \cref{lem:isometry_for_gyro}}
\begin{proof}
We denote the Riemannian logarithm, gyration, and gyronorm on $\{\calM,g \}$ by $\rielog$, $\gyr$, and $\gyrnorm{\cdot}$, respectively, while $\widetilde{\rielog}$, $\gyrw$, and $\gyrnormw{\cdot}$ are their counterparts on $\{\wcalM, \widetilde{g}\}$. We recall the following from \citet[Lems.~2.1--2.3]{nguyen2023building}:
\begin{align}
    \label{app:eq:iso_gyro_add}
    x \oplus y &= \phi^{-1} \left( \phi(x) \widetilde{\oplus} \phi(y) \right), \\
    \label{app:eq:iso_gyro_scalarprod}
    t \odot x &= \phi^{-1} \left( t \widetilde{\odot} \phi(x) \right), \\
    \label{app:eq:iso_gyro_gyration}
    \gyr[x, y] z &= \phi^{-1} \left( \gyrw[\phi(x), \phi(y)] \phi(z) \right),
\end{align}
where $x,y,z \in \calM$.

\mypara{Gyrodistance.}
\begin{equation}
    \begin{aligned}
        \widetilde{\gyrdist}(\phi(x), \phi(y))
        &= \gyrnormw{ \widetilde{\ominus} \phi(x) \widetilde{\oplus} \phi(y) } \\
        &\stackrel{(1)}{=} \gyrnormw{ \phi(\ominus x \oplus y) } \\
        &= \sqrt{\inner{ \widetilde{\rielog _{\widetilde{e}}} (\phi(\ominus x \oplus y))}{\widetilde{\rielog _{\widetilde{e}}} (\phi(\ominus x \oplus y))}_{\widetilde{e}}} \\
        &\stackrel{(2)}{=} \sqrt{\inner{ \rielog _{e} (\ominus x \oplus y) }{ \rielog _{e} (\ominus x \oplus y) }_{e}} \\
        &= \gyrnorm{\ominus x \oplus y} \\
        &= \gyrdist(x, y).
    \end{aligned}
\end{equation}
The derivation above comes from the following.
\begin{enumerate}
    \item
    By \cref{app:eq:iso_gyro_add,app:eq:iso_gyro_scalarprod}:
    \begin{equation}
        \phi(\ominus x \oplus y) = \phi(\ominus x) \widetilde{\oplus} \phi(y).
    \end{equation}
    \item
    By the isometry:
    \begin{align}
        \rielog_{x} (y)
        &= (\phi_{*, x})^{-1} \left( \widetilde{\rielog}_{ \phi(x) } (\phi(y))  \right), \forall x,y \in \calM, \\
        \inner{v}{w}_{x}
        &= \inner{\phi_{*,x} (v)}{ \phi_{*,x}(w)}_{\phi(x)}, \forall x \in \calM \text{ and } \forall v,w \in T_x\calM,
    \end{align}
    where $\phi_{*,x}$ is the differential map. Here, the RHSs contain the operators over $\widetilde{\calM}$, whereas the LHSs involve those over $\calM$.
\end{enumerate}

\mypara{Gyroisometry.} Given any $x,y,z,a \in \calM$, we have the following by the isometry of $\phi$.

For the gyroinverse:
\begin{equation}
    \begin{aligned}
        \gyrdistw(\widetilde{\ominus} \phi(x), \widetilde{\ominus} \phi(y))
        &= \gyrdistw(\phi(\ominus x), \phi(\ominus y)) \\
        &= \gyrdist( \ominus x, \ominus y) \\
        &= \gyrdist( x, y) \\
        &= \gyrdistw(\phi(x), \phi(y)).
    \end{aligned}
\end{equation}
For the gyration:
\begin{equation}
    \begin{aligned}
        & \gyrdistw(\gyrw[\phi(z), \phi(a)] \phi(x), \gyrw[\phi(z), \phi(a)] \phi(y)) \\
        &= \gyrdistw( \phi(\gyr[z, a] x), \phi(\gyr[z, a] y)) \\
        &= \gyrdist( \gyr[z, a] x, \gyr[z, a] y) \\
        &= \gyrdist( x, y ) \\
        &= \gyrdistw( \phi(x), \phi(y)).
    \end{aligned}
\end{equation}
For the left gyrotranslation:
\begin{equation}
    \begin{aligned}
        & \gyrdistw(\phi(z) \widetilde{\oplus} \phi(x), \phi(z) \widetilde{\oplus} \phi(y) ) \\
        &= \gyrdistw(\phi(z \oplus x), \phi(z \oplus y) ) \\
        &= \gyrdist(z \oplus x, z \oplus y ) \\
        &= \gyrdist( x, y ) \\
        &= \gyrdistw( \phi(x), \phi(y)).
    \end{aligned}
\end{equation}
\end{proof}

\linkofproof{thm:iff_gyroauto_gyroisometries}
\subsection[Proof of \cref{thm:iff_gyroauto_gyroisometries}]{Proof of \cref{thm:iff_gyroauto_gyroisometries}}
We first prove a useful lemma.

 \begin{parislemma}[Left gyrotranslation law]
 \label{app:lem:left_gyrotranslation}
    Every pseudo-reductive gyrogroup $\left\{ G, \oplus \right\}$ verifies the left gyrotranslation law:
    \begin{equation}
        \ominus \left(x \oplus y \right) \oplus \left(x \oplus z \right) = \gyr[x, y]\left(\ominus y \oplus z\right), \quad \forall x, y, z \in G.
    \end{equation}
\end{parislemma}
\begin{proof}
    This lemma generalizes \citet[Lems.~I.1 and~L.1]{nguyen2023building}, which prove the left gyrotranslation law on the specific gyrogroups of the SPD and Grassmannian manifolds.
    Their proof only relies on the left cancellation and the basic axioms (G1--G3).
    Note that the original proof of left gyrotranslation on the Grassmannian \citep[Lem.~I.1]{nguyen2023building} is questionable, as it relies on the left cancellation of gyrogroups, and the Grassmannian is not a gyrogroup but a non-reductive gyrogroup. Fortunately, as we show in \cref{thm:pseudo_reductive_gyrogroups_properties}, the general pseudo-reductive gyrogroups, including the Grassmannian, enjoy left cancellation.
    Therefore, the proof in \citet[Lem.~I.1]{nguyen2023building} can be readily generalized to pseudo-reductive gyrogroups.
\end{proof}

\begin{proof}[Proof of \cref{thm:iff_gyroauto_gyroisometries}]
    $\Rightarrow$:
    For any $z,a \in G$, the gyroautomorphism can be expressed by the gyrator identity in \cref{thm:pseudo_reductive_gyrogroups_properties}:
    \begin{align}
        \gyr[x,y] z &= X \oplus \bar{z},\\
        \gyr[x,y] a &= X \oplus \bar{a},
    \end{align}
    where $X = \ominus (x \oplus y)$, $\bar{z} = x \oplus (y \oplus z)$, and $\bar{a} = x \oplus (y \oplus a)$. Then \citet[Eq.~(30)]{nguyen2023building}, derived for the specific Grassmannian, can be directly extended to pseudo-reductive gyrogroups, as it only relies on left gyrotranslation, invariance of the norm under gyroautomorphisms, and the axioms (G1--G3).

    $\Leftarrow$:
    \begin{equation}
        \begin{aligned}
            \gyrnorm{\gyr[x,y](z)}
            &= \gyrnorm{\gyr[x,y](\ominus e \oplus z)} \quad \text{(\cref{enu:prgp_7} in \cref{thm:pseudo_reductive_gyrogroups_properties} indicates $\ominus e = e$)}\\
            &= \gyrnorm{ \ominus \gyr[x,y](e) \oplus  \gyr[x,y](z)} \quad \text{(automorphism)}\\
            &= \dist \left(\gyr[x,y](e), \gyr[x,y](z) \right) \\
            &= \dist \left(e, z \right) \\
            &= \gyrnorm{\ominus e \oplus z} \\
            &= \gyrnorm{z}.
        \end{aligned}
    \end{equation}
\end{proof}

\linkofproof{thm:gyroisometries}
\subsection[Proof of \cref{thm:gyroisometries}]{Proof of \cref{thm:gyroisometries}}

\begin{proof}
    Given any $x, y, z \in G$, we prove the two claims as follows.

    \mypara{Gyroisometry of the left gyrotranslation.} This property generalizes \citet[Thms.~2.12 and~2.16]{nguyen2023building}, which deal with the gyrotranslation in the SPD and Grassmannian, respectively. We have the following:
    \begin{equation}
        \begin{aligned}
            \dist(L_{x} (y), L_{x} (z))
            &= \dist(x \oplus y, x \oplus z) \\
            &= \gyrnorm{\ominus (x \oplus y) \oplus (x \oplus z)} \\
            &= \gyrnorm{\gyr[x, y]\left(\ominus y \oplus z\right)} \quad \text{ (left gyrotranslation law)} \\
            &= \gyrnorm{\ominus y \oplus z} \quad \text{ (gyroisometry of the automorphism)} \\
            &= \dist(y, z).
        \end{aligned}
    \end{equation}

    \mypara{Gyroisometry of the gyroinverse.}
    \begin{equation}
        \begin{aligned}
            & \dist(\ominus x, \ominus y)\\
            &= \gyrnorm{x \ominus y} \\
            &= \gyrnorm{\ominus y \oplus x} \quad \text{ (gyrocommutativity and gyroisometry of the automorphism)} \\
            &= \dist(y, x) \\
            &= \dist(x, y) \quad \text{ (symmetry of the geodesic distance)}.\\
        \end{aligned}
    \end{equation}
\end{proof}

\linkofproof{thm:gyroinvariance}
\subsection[Proof of \cref{thm:gyroinvariance}]{Proof of \cref{thm:gyroinvariance}}
\label{app:subsec:proof_gyroinvariance}

We give a concise argument based on \cref{lem:gyro_geodesic_dist,thm:iff_gyroauto_gyroisometries,thm:gyroisometries}.

\begin{proof}
    First, all these gyrospaces are characterized by \cref{eq:ch2-riem-gyro-addition,eq:ch2-riem-gyro-scalar}. By \cref{lem:gyro_geodesic_dist}, their gyrodistances agree with the geodesic distances. It thus remains to establish the gyroisometries.

    As shown by \cref{thm:iff_gyroauto_gyroisometries,thm:gyroisometries}, it suffices to show that gyrations in each space preserve the gyronorm. This argument on the SPD and ONB Grassmannian has already been proven \citep[Lems.~L.2 and I.2]{nguyen2023building}. Since the ONB Grassmannian is isometric to the PP Grassmannian via \cref{eq:iso_grass}, \cref{lem:isometry_for_gyro} implies that the same arguments apply to the PP. We therefore only need to treat $\stereo{n}$ with $K \leq 0$. As the Euclidean case $\bbR{n}$ is trivial, we only need to show the Poincaré ball. In the following, $a,b,x,y$ are arbitrary points in $\pball{n}$.

    \mypara{Norm invariance under gyrations.} As the Poincaré ball forms a real inner-product gyrovector space \citep[Def.~6.2 and Thm.~6.85]{ungar2022analytic}, any gyration preserves the Euclidean norm:
    \begin{equation}
        \norm{\gyr[a,b](x)} = \norm{x}, \quad \forall x \in \stereo{n}.
    \end{equation}
    For the gyronorm, we further have
    \begin{equation}
        \begin{aligned}
            \gyrnorm{\gyr[a,b] x}
            &= \| \rielog _{\zerovec}  (\gyr[a,b] x )\|_{\zerovec} \\
            &= \tfrac{2}{\sqrt{|K|}} \tanh^{-1}\left(\sqrt{|K|}\|\gyr[a,b] x \|\right) \\
            &= \tfrac{2}{\sqrt{|K|}} \tanh^{-1}\left(\sqrt{|K|}\|x\|\right) \\
            &= \gyrnorm{x}.
        \end{aligned}
    \end{equation}
\end{proof}

\linkofproof{thm:gyrobn}
\subsection[Proof of \cref{thm:gyrobn}]{Proof of \cref{thm:gyrobn}}
\begin{proof}
        According to \cref{thm:gyroisometries}, any left gyrotranslation is a gyroisometry. For any $y \in \calM$, we have the following:
    \begin{equation}
        \label{eq:gyobn_homogeneity_intermediate}
        \begin{aligned}
             \dist\left(\beta \oplus x_i, y \right)
            &\stackrel{(1)}{=} \dist \left(\ominus \beta \oplus (\beta \oplus x_i), \ominus \beta \oplus  y \right) \\
            &\stackrel{(2)}{=} \dist \left((\ominus \beta \oplus \beta) \oplus \gyr[\ominus \beta,\beta] (x_i) , \ominus \beta \oplus  y \right) \\
            &\stackrel{(3)}{=} \dist \left( x_i, \ominus \beta \oplus  y \right).
        \end{aligned}
    \end{equation}
    The above comes from the following.
    \begin{enumerate}
        \item Any left gyrotranslation is a gyroisometry.
        \item Left gyroassociative law.
        \item $\ominus \beta \oplus \beta = e$ and pseudo-reduction.
    \end{enumerate}

    Denoting the gyromean of $\{x_i\}$ and $\{\beta \oplus x_i \}$ as $\mu$ and $\widetilde{\mu}$, we have the following:
    \begin{equation}
        \begin{aligned}
           \beta \oplus \mu
           &\stackrel{(1)}{=} \beta \oplus (\ominus \beta \oplus  \widetilde{\mu}) \\
           &\stackrel{(2)}{=} \gyr[\beta, \ominus \beta] (\widetilde{\mu}) \\
           &\stackrel{(3)}{=} \widetilde{\mu}. \\
        \end{aligned}
    \end{equation}
    The above comes from the following.
    \begin{enumerate}
        \item \cref{eq:gyobn_homogeneity_intermediate} indicates that $\mu = \ominus \beta \oplus \widetilde{\mu}$.
        \item Left gyroassociative law.
        \item Pseudo-reduction.
    \end{enumerate}

    Now, we proceed to deal with the second property.
    We have the following:
    \begin{equation}
        \label{eq:dispersion_intermediate}
        \begin{aligned}
            \dist(t \odot x_i, e)
            &\stackrel{(1)}{=} \gyrnorm{ \ominus e \oplus (t \odot x_i) } \\
            &\stackrel{(2)}{=} \gyrnorm{ t \odot x_i } \\
            &= \norm{ t \rielog_e (x_i) }_e\\
            &= |t| \norm{\rielog_e (x_i) }_e\\
            &= |t| \gyrnorm{ x_i }\\
            &\stackrel{(3)}{=} |t| \gyrnorm{ \ominus e \oplus x_i }\\
            &= |t| \dist( e, x_i)\\
            &\stackrel{(4)}{=} |t| \dist(x_i, e)\\
    \end{aligned}
    \end{equation}
    The above follows from the following.
    \begin{enumerate}
        \item Symmetry of gyrodistance (as geodesic distance).
        \item $\ominus e = e$.
        \item $x_i = \ominus e \oplus x_i$.
        \item Symmetry of gyrodistance (as geodesic distance).
    \end{enumerate}
    The last equation in \cref{eq:dispersion_intermediate} indicates the homogeneity of dispersion from $e$.
\end{proof}

\linkofproof{prop:fast_bracket_grassmannian}
\subsection[Proof of \cref{prop:fast_bracket_grassmannian}]{Proof of \cref{prop:fast_bracket_grassmannian}}

We first review a fast and stable algorithm for the ONB Grassmannian logarithm \citep[Alg.~5.3]{bendokat2024grassmann}, and the calculation of Grassmannian logarithm under the projector perspective by the ONB Grassmannian logarithm \citep[Prop.~3.12]{nguyen2024matrix}.

\begin{algorithm}[t]
\caption{ONB Grassmannian logarithm \citep[Alg.~5.3]{bendokat2024grassmann}}
\label{alg:modgrasslog}
\KwIn{$U, Y \in \grasonb{p,n}$ are Stiefel representatives under the ONB perspective.}
$QSR^\top \stackrel{\mathrm{SVD}}{:=} Y^\top U$ with $S$ in ascending order, and $Q$ and $R$ flipped column-wise accordingly
$\hat{S} = \sqrt{I_p -S^2}$
$\Delta = (I_n - UU^\top)YQ \frac{\arcsin(\hat{S})}{\hat{S}} R^\top$
\BlankLine
\KwOut{$\rielog_U(Y)=\Delta$}
\end{algorithm}

\cref{alg:modgrasslog} reviews a fast and stable algorithm for the Grassmannian Riemannian logarithm under the ONB perspective $\grasonb{p,n}$.
The vanilla Riemannian logarithm in \cref{tab:ch2-grassmann-riemannian-operators,tab:ch2-grassmann-gyro-operators} requires an $n \times p$ SVD and a $p \times p$ matrix inverse, while \cref{alg:modgrasslog} only requires a $p \times p$ SVD.
Therefore, \cref{alg:modgrasslog} is more efficient than the vanilla logarithm.
Besides, \cref{alg:modgrasslog} can also return a unique tangent vector when $Y$ is in the cut locus of $U$. For more details, please refer to \citet[Sec.~5.2]{bendokat2024grassmann}.

As the projector perspective is isometric to the ONB perspective, the Grassmannian logarithm under the projector perspective can be calculated by the ONB Grassmannian logarithm \citep[Prop.~3.12]{nguyen2024matrix}.

\begin{parisproposition}[\citep{nguyen2024matrix}] \label{app:prop:grass_logarithm_onb_pp}
    Given any $P, Q \in \graspp{p,n}$ with $U=\pi^{-1}(P)$ and $V=\pi^{-1}(Q)$, the Riemannian logarithm $\widetilde{\rielog}_{P}(Q)$ on $\graspp{p,n}$ is given as
    \begin{equation}
        \widetilde{\rielog}_{P}(Q) = \pi_{*,U} \left( \rielog_U V \right),
    \end{equation}
    where $\rielog$ is the Riemannian logarithm under the ONB perspective, $\pi_{*,U}: T_{U} \grasonb{p,n} \rightarrow T_{P} \graspp{p,n}$ is the differential map of $\pi$ at $U$, which is defined as
    \begin{equation}
        \pi_{*,U} (\Delta) = \Delta U^\top + U \Delta^\top, \forall \Delta \in  T_{U} \grasonb{p,n}.
    \end{equation}
\end{parisproposition}

Now, we begin to present the proof.
\begin{proof}[Proof of \cref{prop:fast_bracket_grassmannian}]
    We first show the expression for $\rielog_{\idonb}$ and $\widetilde{\rielog}_{\idpp}$.

    First note the following:
    \begin{equation}
        (I_n - \idonb \idonb^\top) =
        \left( \begin{array}{cc}
            \bbzero & \bbzero \\
            \bbzero & I_{n-p}
        \end{array} \right),
    \end{equation}
    \begin{equation}
        \begin{aligned}
             U^\top \idonb
            &= \left(U_1^\top,U_2^\top \right)
            \left(\begin{array}{c}
                    I_p \\
                    \bbzero
                \end{array} \right) \\
            &= U_1^\top,
        \end{aligned}
    \end{equation}
    By the above two equations, the ONB Grassmannian logarithm at $\idonb$ is
    \begin{equation}
        \label{eq:gras_log_i_onb}
        \begin{aligned}
            \rielog_{\idonb} (U)
            &=  \left( \begin{array}{cc}
                \bbzero & \bbzero \\
                \bbzero & I_{n-p}
            \end{array} \right)
            \left(\begin{array}{c}
                U_1 \\
                U_2
            \end{array} \right)
            Q \frac{\arcsin(\hat{S})}{\hat{S}} R^\top \text{ (\cref{alg:modgrasslog})}\\
            &= \left(\begin{array}{c}
                \bbzero \\
                U_2 Q \frac{\arcsin(\hat{S})}{\hat{S}} R^\top
            \end{array} \right) \\
            &= \left(\begin{array}{c}
                \bbzero \\
                \widetilde{U}_2
            \end{array} \right),
        \end{aligned}
    \end{equation}
    where $QSR^\top \stackrel{\mathrm{SVD}}{:=} U_1^\top$ with $S$ in ascending order, and $Q$ and $R$ flipped column-wise accordingly, and $\hat{S} = \sqrt{I_p -S^2}$.

    For $\widetilde{\rielog}_{\idpp}$, we have
    \begin{equation}
        \begin{aligned}
            \widetilde{\rielog}_{\idpp}(UU^\top)
            &\stackrel{(1)}{=} \pi_{*,\idonb} \left( \rielog_{\idonb} (U) \right) \\
            &\stackrel{(2)}{=} \pi_{*,\idonb} \left( \left(\begin{array}{c}
                \bbzero \\
                \widetilde{U}_2
            \end{array} \right) \right) \\
            &\stackrel{(3)}{=} \left(\begin{array}{cc}
                \bbzero & \widetilde{U}_2^\top \\
                \widetilde{U}_2 & \bbzero
            \end{array} \right).
        \end{aligned}
    \end{equation}
    The above derivation comes from the following.
    \begin{enumerate}
        \item
        \cref{app:prop:grass_logarithm_onb_pp}
        \item
        \cref{eq:gras_log_i_onb}
        \item
        For any $\Delta = (\Delta_1^\top,\Delta_2^\top)^{\top} \in T_{\idonb}\grasonb{p,n}$, where $\Delta_1$ is $p \times p$, we have the following:
        \begin{equation}
            \begin{aligned}
                 \pi_{*,\idonb} \left(
                 \left( \begin{array}{c}
                      \Delta_1 \\
                      \Delta_2
                 \end{array}
                 \right)\right)
                 &=
                 \left( \begin{array}{c}
                      \Delta_1 \\
                      \Delta_2
                 \end{array}
                 \right) \left(I_p,\bbzero \right) +
                 \left( \begin{array}{c}
                      I_p \\
	                      \bbzero
                 \end{array}
                 \right) \left(\Delta_1^\top, \Delta_2^\top \right) \\
                 &= \left( \begin{array}{cc}
	                      \Delta_1 & \bbzero  \\
	                      \Delta_2 & \bbzero
                 \end{array} \right) + \left( \begin{array}{cc}
                      \Delta_1^\top & \Delta_2^\top  \\
	                      \bbzero & \bbzero
                 \end{array} \right) \\
                 &=\left( \begin{array}{cc}
                      \Delta_1 + \Delta_1^\top & \Delta_2^\top  \\
	                      \Delta_2 & \bbzero
                 \end{array} \right).
            \end{aligned}
        \end{equation}
    \end{enumerate}
    Combining all the above results together, we have the following:
    \begin{equation}
        \begin{aligned}
            [\overline{U U^\top},\idpp]
            &= \left[ \widetilde{\rielog}_{\idpp}(UU^\top), \idpp \right]\\
            &= \left[ \left(\begin{array}{cc}
                \bbzero & \widetilde{U}_2^\top \\
                \widetilde{U}_2 & \bbzero
            \end{array} \right), \idpp \right]\\
            &= \left(\begin{array}{cc}
                \bbzero & \widetilde{U}_2^\top \\
                \widetilde{U}_2 & \bbzero
            \end{array} \right)
            \left( \begin{array}{cc}
            I_{p} & \bbzero \\
            \bbzero & \bbzero
        \end{array} \right) -
        \left( \begin{array}{cc}
            I_{p} & \bbzero \\
            \bbzero & \bbzero
        \end{array} \right)
        \left(\begin{array}{cc}
                \bbzero & \widetilde{U}_2^\top \\
                \widetilde{U}_2 & \bbzero
            \end{array} \right) \\
        &=\left(
        \begin{array}{cc}
	             \bbzero & -\widetilde{U}_2^\top \\
	             \widetilde{U}_2 & \bbzero
        \end{array} \right).
        \end{aligned}
    \end{equation}
\end{proof}

\linkofproof{prop:stereographic_gyro_from_riemannian}
\subsection[Proof of \cref{prop:stereographic_gyro_from_riemannian}]{Proof of \cref{prop:stereographic_gyro_from_riemannian}}
\begin{proof}
    Given $v \in T_\zerovec \stereo{n}$, we have the following:
    \begin{align}
        \lambda^K_{\zerovec} &= 2, \\
        \rielog _{\zerovec}(y)
        &= \tank ^{-1} \left( \sqrt{|K|} \norm{y} \right) \frac{y}{ \sqrt{|K|} \norm{y}}, \\
        \pt{\zerovec}{x} (v)
        &= \frac{\lambda_{\zerovec} ^K}{\lambda_{x}^K} \gyr[x,-\zerovec] (v) = \frac{2}{\lambda_{x}^K} v, \\
        \rieexp _{\zerovec}(v)
        &= \tank \left(\sqrt{|K|} \norm{v}\right) \frac{v}{\sqrt{|K|} \norm{v}}.
    \end{align}
    Therefore, we have
    \begin{equation}
        \begin{aligned}
            \rieexp_{x}\left(\pt{\zerovec}{x} \left(\rielog _{\zerovec}(y)\right)\right)
            &= \rieexp_{x}\left( \frac{2}{\sqrt{|K|} \lambda_{x}^K } \tank ^{-1} \left( \sqrt{|K|} \norm{y} \right) \frac{y}{\norm{y}} \right) \\
            &= x \stoplus y,
        \end{aligned}
    \end{equation}
    which implies
    \begin{equation}
        \begin{aligned}
            \rieexp _{\zerovec} \left( t \rielog _{\zerovec} (x) \right) = t \stodot x.
        \end{aligned}
    \end{equation}
\end{proof}

\linkofproof{thm:stereographic_gyro}
\subsection[Proof of \cref{thm:stereographic_gyro}]{Proof of \cref{thm:stereographic_gyro}}
\begin{proof}
We only need to show the case of $K > 0$. Let $x, y, z, w$ be any vectors in $\stereo{n}$, and $s, t \in \bbRscalar$ be real scalars.

We first notice the gyroaddition is a linear combination:
\begin{equation}
    x \stoplus y = \frac{(1 - 2K \inner{x}{y} - K \norm{y}^2)x + (1 + K \norm{x}^2)y}{1 - 2K \inner{x}{y} + K^2 \norm{x}^2 \norm{y}^2}
    = \frac{A x + B y}{D},
\end{equation}
where
\begin{equation}
    \label{app:eq:gyro_addition_setero_rewritten}
    \begin{aligned}
    A &= 1 - 2K \inner{x}{y} - K \norm{y}^2, \\
    B &= 1 + K \norm{x}^2, \\
    D &= 1 - 2K \inner{x}{y} + K^2 \norm{x}^2 \norm{y}^2.
    \end{aligned}
\end{equation}
Recalling \cref{eq:stereographic_gyration}, the gyration $\gyr[x, y](z)$ is also a linear expression:
\begin{equation}
    \gyr[x, y](z) = z + f_1 \cdot x + f_2 \cdot y.
\end{equation}

\mypara{Axioms G1--G3.} Since $e=\zerovec$ and $\stominus x = -x$, G1 and G2 can be immediately verified.
The left gyroassociative law follows from the definition of gyration \citep[Eq.~36]{bachmann2020constant}:
\begin{equation}
    \gyr[x, y] (z)= -\left(x \stoplus y\right) \stoplus\left(x \stoplus\left(y \stoplus z\right)\right).
\end{equation}
To confirm that any gyration is an automorphism of $(\stereo{n}, \stoplus)$, we verify the following identity using symbolic computation:
\begin{equation} \label{app:eq:stereographic_gyration_identity}
    \gyr[x,y](w \stoplus z) = \gyr[x,y](w) \stoplus \gyr[x,y](z).
\end{equation}

Expanding all necessary inner products (e.g., $\inner{x}{w \stoplus z}$, $\inner{y}{w \stoplus z}$) and norms in terms of $\inner{x}{w}$, $\inner{x}{z}$, etc., we express both sides of \cref{app:eq:stereographic_gyration_identity} as linear combinations:
\begin{align}
    \gyr[x,y](w \stoplus z) = f_1 x + f_2 y + f_3 w + f_4 z, \\
    \gyr[x,y](w) \stoplus \gyr[x,y](z) = f_1' x + f_2' y + f_3' w + f_4' z.
\end{align}
We use \texttt{SymPy} to compare the coefficients of $x$, $y$, $w$, and $z$ on both sides. The above is exposed in \verb|stereographic_gyr_automorphism.py|.

\mypara{(G4) Left reduction law.} Similarly, we expand $\gyr[x \stoplus z,y](w)$ and $\gyr[x,y](w)$ and compare the coefficients, as implemented in \verb|stereographic_left_reduction.py|.

\mypara{Gyrocommutative law.} This has been verified by \citet[Lem.~11]{bachmann2020constant}.

\mypara{(V1) Identity scalar multiplication.} This can be directly verified by definition:
\begin{equation}
    t \stodot x= \frac{\tan \left(t \tan ^{-1}(\sqrt{K}\norm{x})\right)}{\sqrt{K}} \frac{x}{\norm{x}}, \forall x \neq 0.
\end{equation}

\mypara{(V2) Scalar distributive law.} We want to verify
\begin{equation} \label{app:eq:stereographic_axiom_v2}
    (s + t) \stodot x = (s \stodot x) \stoplus (t \stodot x).
\end{equation}
We only need to show the case of $x \neq \zerovec$.
Let $\theta = \tan^{-1}(\sqrt{K} \norm{x})$. Then scalar multiplication is simplified as
\begin{equation}
    t \stodot x = \frac{\tan(t \theta)}{\sqrt{K} \norm{x}} x.
\end{equation}
Using symbolic computation (see \verb|stereographic_gyr_v2.py|), we obtain the following w.r.t. \cref{app:eq:stereographic_axiom_v2}:
\begin{align}
    \text{LHS} &= \frac{\tan\left((s + t)\theta\right)}{\sqrt{K} \norm{x}}, \\
    \text{RHS} &= \frac{\tan(s\theta) + \tan(t\theta)}{\sqrt{K} \norm{x} \left( 1 - \tan(s\theta)\tan(t\theta) \right)}.
\end{align}
The identity follows from the tangent addition formula:
\begin{equation}
    \tan((s + t)\theta) = \frac{\tan(s\theta) + \tan(t\theta)}{1 - \tan(s\theta)\tan(t\theta)}.
\end{equation}

\mypara{(V3) Scalar associative law.} This can be directly verified by definition.

\mypara{(V4) Gyroautomorphism.} We now verify
\begin{equation} \label{app:eq:stereographic_axiom_v4}
    \gyr[x,y](t \stodot z) = t \stodot \gyr[x,y](z).
\end{equation}
Let us denote
\begin{equation}
    \alpha_t^{(z)} = \frac{\tan \left(t \tan^{-1}\left(\sqrt{K} \norm{z}\right)\right)}{\sqrt{K} \norm{z}}.
\end{equation}
By the linearity of gyration \citep[Lem.~11]{bachmann2020constant}, the left-hand side of \cref{app:eq:stereographic_axiom_v4} becomes
\begin{equation}
    \text{LHS} = \alpha_t^{(z)} \gyr[x,y](z),
\end{equation}
while the right-hand side reads
\begin{equation}
    \text{RHS} = \alpha_t^{\gyr[x,y](z)} \gyr[x,y](z).
\end{equation}
Since $\alpha_t^{(\cdot)}$ depends only on the norm of its argument, it suffices to show
\begin{equation}
    \norm{z} = \norm{\gyr[x,y](z)},
\end{equation}
which holds as proven by \citet[Lem.~11, iv)]{bachmann2020constant}.

\mypara{(V5) Identity gyroautomorphism.} In each of the three special cases when (i) $x=\zerovec$, or (ii) $y=\zerovec$, or (iii) $x$ and $y$ are parallel in $\bbV$, $x \parallel y$, we have
\begin{align}
    \gyr[\zerovec, x] (z) &\stackrel{(1)}{=} z, \\
    \gyr[x, \zerovec] (z) &\stackrel{(2)}{=} z, \\
    \gyr[x, y] (z) &\stackrel{(3)}{=} z, \quad x \parallel y,
\end{align}
where (1--3) come from $A_{\mathrm{st}}x+B_{\mathrm{st}}y=\zerovec$ in \cref{eq:stereographic_gyration}. Therefore, we have
\begin{equation}
    \gyr[s \stodot x, t \stodot x] = \gyr[\alpha_s^{(x)} x, \alpha_t^{(x)} x] = \id.
\end{equation}
Note that (1--2) are also implied by the first gyrogroup theorem \citep[Thm.~2.10]{ungar2022analytic}.
\end{proof}

\linkofproof{thm:gyroinvariance_stereo}
\subsection[Proof of \cref{thm:gyroinvariance_stereo}]{Proof of \cref{thm:gyroinvariance_stereo}}
This proof largely follows the one for \cref{thm:gyroinvariance}.
\begin{proof}
    \mypara{Norm invariance under gyrations.} By \citet[Lem.~11]{bachmann2020constant}, any gyration preserves the Euclidean norm:
    \begin{equation}
        \| \gyr[a,b](x) \| = \| x \|, \quad \forall x \in \stereo{n}.
    \end{equation}
    For the gyronorm, we further have
    \begin{equation}
        \begin{aligned}
            \gyrnorm{\gyr[a,b] x}
            &= \| \rielog _{\zerovec}  (\gyr[a,b] x )\|_{\zerovec} \\
            &= \tfrac{2}{\sqrt{|K|}} \tank^{-1}\left(\sqrt{|K|}\|\gyr[a,b] x \|\right) \\
            &= \tfrac{2}{\sqrt{|K|}} \tank^{-1}\left(\sqrt{|K|}\|x\|\right) \\
            &= \gyrnorm{x},
        \end{aligned}
    \end{equation}
    with $\tank=\tanh$ for $K<0$ and $\tank=\tan$ for $K>0$.

    \mypara{Isometry of left gyrotranslation and gyroinverse.} As shown in \cref{thm:stereographic_gyro}, $\stereo{n}$ forms a gyrocommutative gyrogroup. By \cref{thm:gyroisometries}, the left gyrotranslation and gyroinverse are gyroisometries.
\end{proof}

\linkofproof{prop:calmk_scalar_prod_inv}
\subsection[Proof of \cref{prop:calmk_scalar_prod_inv}]{Proof of \cref{prop:calmk_scalar_prod_inv}}

\begin{proof}
    \mypara{Non-singular cases.} Denoting $\norm{x}_s = \norm{x_s}, \forall x \in \calMK{n}$, we have
    \begin{align}
    \left( t  \rielog_{\MKzero} x \right)_s
    &= t  \frac{\cosk^{-1}(\sqrt{|K|} x_t)}{\sqrt{|K|} \norm{x_s}} x_s, \\
    \left\| t  \rielog_{\MKzero} x \right\|_s
    &= |t|  \frac{\cosk^{-1}(\sqrt{|K|} x_t)}{\sqrt{|K|}}, \\
    \sqrt{|K|}  \left\| t  \rielog_{\MKzero} x \right\|_s
    &= |t| \cosk^{-1}(\sqrt{|K|} x_t),
    \end{align}
    For $t \neq 0$, the normalized spatial direction satisfies
    \begin{equation}
        \frac{\left(t\rielog_{\MKzero}x\right)_s}{\left\|t\rielog_{\MKzero}x\right\|_s}
        =\operatorname{sgn}(t)\frac{x_s}{\norm{x_s}}.
    \end{equation}
    Since $\cosk$ is even and $\sink$ is odd, substituting the above expressions into $\rieexp_{\MKzero}$ in \cref{tab:ch2-constant-curvature-operators} recovers the signed argument $t\cosk^{-1}(\sqrt{|K|}x_t)$ in \cref{eq:gyro_prod_calmk}. The cases $t=0$ and $x=\MKzero$ are handled by the first branch of \cref{eq:gyro_prod_calmk}.

    For $t=-1$, we have
    \begin{equation}
        \begin{aligned}
        -1 \MKodot x
        &=
        \frac{1}{\sqrt{|K|}}
        \begin{bmatrix}
        \cosk \left( - \cosk^{-1}(\sqrt{|K|} x_t) \right) \\
        \frac{\sink \left( - \cosk^{-1}(\sqrt{|K|} x_t) \right)}{\norm{x_s}}  x_s
        \end{bmatrix} \\
        &\stackrel{(1)}{=}
        \begin{bmatrix}
        x_t \\
        - \frac{\sink \left(\cosk^{-1}(\sqrt{|K|} x_t) \right)}{\sqrt{|K|}\norm{x_s}} x_s
        \end{bmatrix},
        \end{aligned}
    \end{equation}
    where (1) comes from $\cosk(- \theta) = \cosk(\theta)$ and $\sink(-\theta)=-\sink(\theta)$. The rest is to show $\frac{\sink \left(\cosk^{-1}(\sqrt{|K|} x_t) \right)}{\sqrt{|K|}\norm{x_s}} = 1$:
    \begin{equation}
        \frac{\sink \left(\cosk^{-1}(\sqrt{|K|} x_t) \right)}{\sqrt{|K|}\norm{x_s}}
        \stackrel{(1)}{=} \frac{\sqrt{ \sign(K) (1- |K| x_t^2) }}{\sqrt{|K|} \norm{x_s}}
        \stackrel{(2)}{=}1.
    \end{equation}
    The derivation above is based on the following:
    \begin{enumerate}
        \item
        $\cosk ^2 (\theta) + \sign(K) \sink ^2 (\theta) = 1$ implies
        \begin{equation}
            \sink (\theta) = \sqrt{\sign(K) (1 - \cosk ^2 (\theta))},  \forall \theta>0.
        \end{equation}
        Considering $\cosk^{-1}(\theta) \geq 0$ for any $\theta \in \operatorname{dom}(\cosk^{-1}(\cdot))$, we have (1).
        \item
        \begin{equation}
            \begin{aligned}
                &\Kinner{x}{x} = \sign(K) x_t^2 + \norm{x_s}^2 = \frac{1}{K} \\
                & \Rightarrow K \norm{x_s}^2 = 1 - |K| x_t^2 \\
                & \Rightarrow |K| \norm{x_s}^2 = \sign(K) (1 - |K| x_t^2).
            \end{aligned}
        \end{equation}

    \end{enumerate}

    \mypara{Equivalence in the singular cases.} By \cref{eq:iso_calMK_to_stereo},
    \begin{equation}
    \sqrt{K} \norm{u} = \frac{\sqrt{K}  \norm{x_s}}{1+\sqrt{K} x_t}.
    \end{equation}
    Writing $\theta=\cos^{-1}(\sqrt{K} x_t)$ or $\sqrt{K} x_t=\cos\theta$, the sphere constraint gives $\sqrt{K}  \norm{x_s}=\sin\theta$. Hence
    \begin{equation}
    \sqrt{K} \norm{u} = \frac{\sin\theta}{1+\cos\theta} = \tan \frac{\theta}{2},
    \quad \Rightarrow
    \tan^{-1}\left(\sqrt{K} \norm{u}\right)=\frac{\theta}{2}.
    \end{equation}
    Therefore, (i) is equivalent to (ii) since $t \tan^{-1}(\sqrt{K}\norm{u})=\frac{t\theta}{2}$.

    Next, we use the closed form of gyromultiplication \cref{eq:singular-case-radius}. For $K>0$, the gyromultiplication reads
    \begin{equation}
    t \MKodot x = \frac{1}{\sqrt{K}}
    \begin{bmatrix}
    \cos\left(t\theta\right) \\
    \dfrac{\sin(t\theta)}{ \norm{x_s}}x_s
    \end{bmatrix}.
    \end{equation}
    If (ii) holds, then $\sin(t\theta)=0$ and $\cos(t\theta)=-1$, whence $t \MKodot x=[-1/\sqrt{K},0]^\top=-\MKzero$.
\end{proof}

\linkofproof{prop:calmk_gyroaddition}
\subsection[Proof of \cref{prop:calmk_gyroaddition}]{Proof of \cref{prop:calmk_gyroaddition}}
\begin{proof}
We denote $x\MKoplus y=[z_t,z_s^\top]^\top$. As the results are trivial under $x=\MKzero$ or $y=\MKzero$, we assume $x \neq \MKzero$ and $y \neq \MKzero$ in the following.

\mypara{Non-singular cases.} We first consider the non-singular case: i) $K<0$; ii) $K>0$, $x,y \neq \pm \MKzero$, and $u \neq \frac{v}{K \norm{v}^2}$. Since gyroadditions on $\calMK{n}$ and $\stereo{n}$ are both defined by \cref{eq:ch2-riem-gyro-addition}, we have the following under isometries \citep[Lem.~2.2]{nguyen2023building}:
\begin{equation}
x\MKoplus y = \isoSTMK{n} \left( \isoMKST{n}(x)\ \stoplus \isoMKST{n}(y) \right).
\end{equation}

Following \cref{app:eq:gyro_addition_setero_rewritten}, we rewrite the gyroaddition on the stereographic model as
\begin{equation}
u\stoplus v=\frac{(1-2K\langle u,v\rangle-K\|v\|^2)u+(1+K\|u\|^2)v}{1-2K\langle u,v\rangle+K^2\|u\|^2\|v\|^2}= \frac{A u + B v}{\Delta},
\end{equation}
where
\begin{equation}
A=1-2K\langle u,v\rangle-K\|v\|^2,\quad B=1+K\|u\|^2,\quad \Delta=1-2K\langle u,v\rangle+K^2\|u\|^2\|v\|^2.
\end{equation}
Using $\langle u,v\rangle=s_{xy}/(ab)$, $\|u\|^2=n_x/a^2$, $\|v\|^2=n_y/b^2$, one obtains
\begin{equation}\label{app:eq:ABD-simplified}
A=\frac{ab^2-2Kbs_{xy}-Kan_y}{ab^2},\quad B=\frac{a^2+K n_x}{a^2},\quad \Delta=\frac{D}{a^2 b^2}.
\end{equation}
Hence
\begin{equation}\label{app:eq:w}
\stereo{n} \ni w=u\stoplus v=\frac{A u+B v}{\Delta}
=\frac{1}{\Delta} \left(\frac{A}{a}x_s+\frac{B}{b}y_s \right).
\end{equation}

Apply $\isoSTMK{n}$ to $w$:
\begin{equation}\label{app:eq:ztzs-pre}
z_t=\frac{1}{\sqrt{|K|}}\frac{1-K\|w\|^2}{1+K\|w\|^2},\qquad
z_s=\frac{2w}{1+K\|w\|^2}.
\end{equation}
A direct expansion (\texttt{radius\_gyroaddition.py}) yields
\begin{equation}
\|w\|^2=\frac{A^2\|u\|^2+B^2\|v\|^2+2AB\langle u,v\rangle}{\Delta^2}=\frac{N}{D}.
\end{equation}
Substituting the above into \cref{app:eq:ztzs-pre} gives
\begin{equation}\label{app:eq:zt}
z_t=\frac{1}{\sqrt{|K|}}\frac{1-K N/D}{1+K N/D}=\frac{1}{\sqrt{|K|}}\frac{D-KN}{D+KN}.
\end{equation}
For $z_s$, using \cref{app:eq:w}, \cref{app:eq:ABD-simplified} and $1+K\|w\|^2=(D+KN)/D$,
\begin{equation}\label{app:eq:zs}
    \begin{aligned}
        z_s
        &=\frac{2}{1+K N/D}\cdot\frac{1}{\Delta}\left(\frac{A}{a}x_s+\frac{B}{b}y_s\right) \\
        &= \frac{2}{D+KN}\left((Aab^2)x_s+(Ba^2 b)y_s\right) \\
        &= \frac{2 \left( A_s x_s + A_y y_s \right)}{D+KN}.
    \end{aligned}
\end{equation}

\mypara{Gyroaddition in singular cases ($K>0$).} We show that the definition \cref{eq:gyroadd_calmk} indeed returns $-\MKzero$ in this case.

\noindent \emph{Step 1: $\rielog_{\MKzero}(y)$.}
Write $R=1/\sqrt{K}$ and choose the polar angle $\theta\in(0,\pi)$ so that
\begin{equation}\label{app:eq:x_param}
x=\begin{bmatrix}R\cos\theta\\ R\sin\theta\hat s\end{bmatrix},\qquad
y=\begin{bmatrix}-R\cos\theta\\ R\sin\theta\hat s\end{bmatrix},\qquad
\hat s=\frac{x_s}{\|x_s\|}.
\end{equation}
Using $\rielog_{\MKzero}$ in \cref{tab:ch2-constant-curvature-operators},
\begin{equation}
\rielog_{\MKzero}(y)
=\begin{bmatrix}
0\\ \dfrac{\cos^{-1}(\sqrt{K}y_t)}{\sqrt{K}\|y_s\|}y_s
\end{bmatrix}
=
\begin{bmatrix}
0\\ (\pi-\theta)R\hat s
\end{bmatrix}.
\end{equation}

\noindent\emph{Step 2: Parallel transport.}
With $1+\sqrt{K}x_t=1+\cos\theta\neq 0$ (since $x\neq-\MKzero$) and $\pt{\MKzero}{x}$ in \cref{tab:ch2-constant-curvature-operators}, we have
\begin{equation}\label{app:eq:PT_compute}
\pt{\MKzero}{x}\left(\rielog_{\MKzero}(y)\right)
= \begin{bmatrix}
0\\ (\pi-\theta)R\hat s
\end{bmatrix}
-\frac{K\langle x_s,(\pi-\theta)R\hat s\rangle}{1+\sqrt{K}x_t}
\begin{bmatrix}
x_t+\tfrac{1}{\sqrt{K}}\\ x_s
\end{bmatrix}.
\end{equation}
Using $\langle x_s,\hat s\rangle=\|x_s\|=R\sin\theta$, $K R^2=1$, and $x_t=R\cos\theta$, the scalar factor in \cref{app:eq:PT_compute} equals
\begin{equation}
\frac{K(\pi-\theta)R\langle x_s,\hat s\rangle}{1+\sqrt{K}x_t}
=\frac{(\pi-\theta)\sin\theta}{1+\cos\theta}.
\end{equation}
A short simplification then yields the compact form
\begin{equation}\label{app:eq:w_tangent}
w = \pt{\MKzero}{x}\left(\rielog_{\MKzero}(y)\right)
= R(\pi-\theta)
\begin{bmatrix}
-\sin\theta\\
\cos\theta\hat s
\end{bmatrix}.
\end{equation}
Note that $\|w\|_x=R(\pi-\theta)$, so with $\alpha=\sqrt{K}\|w\|_x$ we have $\alpha=\pi-\theta$.

\noindent\emph{Step 3: Exponential at $x$.}
By \cref{tab:ch2-constant-curvature-operators},
\begin{equation}\label{app:eq:Expx_standard}
\rieexp_x(w) = \cos(\alpha)x + \frac{\sin(\alpha)}{\alpha}w,
\qquad \alpha=\sqrt{K}\|w\|_x,
\end{equation}
and here $\cos(\alpha)=\cos(\pi-\theta)=-\cos\theta$, $\sin(\alpha)=\sin(\pi-\theta)=\sin\theta$.
Substituting \cref{app:eq:x_param} and \cref{app:eq:w_tangent} into \cref{app:eq:Expx_standard},
\begin{equation}
\rieexp_x(w)
= R\left[
-\cos\theta\begin{bmatrix}\cos\theta\\ \sin\theta\hat s\end{bmatrix}
+ \sin\theta\begin{bmatrix}-\sin\theta\\ \cos\theta\hat s\end{bmatrix}
\right]
= R\begin{bmatrix}-1\\ 0\end{bmatrix}
= -\MKzero.
\end{equation}
We conclude that $x\MKoplus y=-\MKzero$ in the singular configuration.

\mypara{Equivalence in singular cases ($K>0$).} Recalling the isometry between $\calMK{n}$ and $\stereo{n}$, we write $u=\tfrac{x_s}{a}$ and $v=\tfrac{y_s}{b}$.

\emph{(i) $\Rightarrow$ (ii).}
Write $\alpha=K \norm{v}^2 >0$. From $u=\frac{v}{\alpha}$ and \cref{eq:iso_stereo_to_calMK} we get
\begin{equation}
    \begin{aligned}
        y_t
        &=\frac{1}{\sqrt{K}}\frac{1-\alpha}{1+\alpha}, \\
        x_t
        &=\frac{1}{\sqrt{K}}\frac{1-\frac{1}{\alpha}}{1+\frac{1}{\alpha}}
        =-\frac{1}{\sqrt{K}}\frac{1-\alpha}{1+\alpha}=-y_t, \\
        x_s
        &=\frac{2u}{1+K\|u\|^2}=\frac{2v/\alpha}{1+K\|v\|^2/\alpha^2}=\frac{2v}{1+\alpha}=y_s,
    \end{aligned}
\end{equation}
Hence $x_s=y_s$ and $x_t=-y_t$.

\emph{(ii) $\Rightarrow$ (iii).}
Under $x_s=y_s$ and $x_t=-y_t$, set $n= \norm{x_s}^2 = \norm{y_s}^2$ and note
$s_{xy}=n$, $n_x=n_y=n$. Then,
\begin{equation}
D=a^2b^2-2Kabs_{xy}+K^2n_xn_y=(ab-Kn)^2.
\end{equation}
On the sphere $\sphere{n}$, we have the constraint $x_t^2+ \norm{x_s}^2 = \nicefrac{1}{K}$. Since $y_t=-x_t$ and
$\norm{y_s}^2=\norm{x_s}^2=n$,
\begin{equation}
\begin{aligned}
    ab
    &=(1+\sqrt{K}x_t)(1+\sqrt{K}y_t) \\
    &=(1+\sqrt{K}x_t)(1-\sqrt{K}x_t) \\
    &=1-Kx_t^2 \\
    &=1-K\left(\frac{1}{K}-n\right)=Kn.
\end{aligned}
\end{equation}
Therefore $D=(ab-Kn)^2=0$.

\emph{(iii) $\Rightarrow$ (i).}
This can be obtained using the Cauchy--Schwarz inequality, as in \citet[App.~C.2.1]{bachmann2020constant}.

\mypara{Consequences in singular cases ($K>0$).} Under (ii) we also have $a+b=2$, so
\begin{equation}
N=a^2n+2abn+b^2n=(a+b)^2n=4n>0.
\end{equation}
Using $D=0$ and $N>0$, \cref{app:eq:zt} gives
\begin{equation}
z_t=\frac{1}{\sqrt{K}}\frac{-KN}{KN}=-\frac{1}{\sqrt{K}}.
\end{equation}
Besides, since for $x_s=y_s$ one checks $A_s+A_y=(a+b)(ab-Kn)=0$, \cref{app:eq:zs} yields $z_s=0$. On the other hand, $u\stoplus v = \infty$.
\end{proof}

\linkofproof{cor:isomorphism_calmk_stereo}
\subsection[Proof of \cref{cor:isomorphism_calmk_stereo}]{Proof of \cref{cor:isomorphism_calmk_stereo}}
\begin{proof}
    This is already implied by the proofs of \cref{prop:calmk_scalar_prod_inv,prop:calmk_gyroaddition}.
\end{proof}

\linkofproof{thm:calmk_gyrovector}
\subsection[Proof of \cref{thm:calmk_gyrovector}]{Proof of \cref{thm:calmk_gyrovector}}

\begin{proof}
    The gyrovector space over the stereographic model is also defined by \cref{eq:ch2-riem-gyro-addition,eq:ch2-riem-gyro-scalar}:
    \begin{equation}
        \begin{aligned}
            x \stoplus y &= \rieexp_{x}\left(\pt{\zerovec}{x} \left(\rielog _{\zerovec}(y)\right)\right),\\
            r \stodot x &= \rieexp_{\zerovec} \left( r \rielog _{\zerovec}(x)\right),\\
        \end{aligned}
    \end{equation}
    with $x, y \in \stereo{n}$ and $r \in \bbRscalar$. Since $\pi_{\calMK{n} \to \stereo{n}}(\MKzero)=\zerovec$ and $(\stereo{n},\stoplus,\stodot)$ is a gyrovector space, $(\calMK{n},\MKoplus,\MKodot)$ is also a gyrovector space \citep[Thm.~2.4]{nguyen2023building}.
\end{proof}

\linkofproof{props:klein_poincare_isometry_differentials}
\subsection[Proof of \cref{props:klein_poincare_isometry_differentials}]{Proof of \cref{props:klein_poincare_isometry_differentials}}

\begin{proof}
    \mypara{Isometries.} The Beltrami--Klein model is isometric to the Lorentz model by the following diffeomorphisms \citep[Thm.~3.7]{lee2018introduction}:
    \begin{align}
        \label{app:eq:iso_klein_to_lorentz}
        \pi_{\klein{n} \to \lorentz{n}}
        &: \klein{n} \ni x  \longmapsto\left(\frac{1}{\sqrt{-K} \sqrt{1+K\|x\|^2}}, \frac{x}{\sqrt{1+K \|x\|^2}}\right) \in \lorentz{n},\\
        \label{app:eq:iso_lorentz_to_klein}
        \pi_{\lorentz{n} \to \klein{n}}
        &:
        \lorentz{n} \ni
        \begin{bmatrix}
        x_t \\
        x_s
        \end{bmatrix} \longmapsto \frac{x_s}{\sqrt{-K}x_t} \in \klein{n}.
    \end{align}

    Combining the isometries \cref{app:eq:iso_klein_to_lorentz,app:eq:iso_lorentz_to_klein,eq:iso_calMK_to_stereo,eq:iso_stereo_to_calMK}, one can readily obtain the isometries between Beltrami--Klein and Poincaré ball models. Now, we turn to the differential maps. Given a curve over $c(t) \in \calM$ with $c(0)=x$ and $c'(0)=v$, the differential maps can be calculated by
    \begin{equation}
        \begin{aligned}
        \left. \frac{d \pi_{\klein{n} \to \pball{n}} (c(t))}{d t} \right|_{t=0}
        &=  \left. \frac{d }{d t}  \frac{c(t)}{ \left( 1 + \sqrt{1 + K \|c(t)\|^2} \right)} \right|_{t=0} \\
        &=\frac{\left(1+ \sqrt{1+K\|x\|^2} \right) v - \left(1+K\|x\|^2\right)^{-\frac{1}{2}} K \inner{x}{v} x}{\left(1 + \sqrt{1+K\|x\|^2} \right)^2} \\
        &=\frac{1}{1 + \sqrt{1+K\|x\|^2}} v
        - \frac{K \inner{x}{v}}{\left(1 + \sqrt{1+K\|x\|^2} \right)^2 \sqrt{1+K\|x\|^2}} x, \\
        \left. \frac{d \pi_{\pball{n} \to \klein{n} } (c(t))}{d t} \right|_{t=0}
        &= \left. \frac{d }{d t} \frac{2 c(t)}{1 - K \|c(t)\|^2} \right|_{t=0}  \\
        &= \frac{2 v\left(1- K\|x\|^2\right)+4 K \inner{x}{v} x}{\left(1-K\|x\|^2\right)^2 } \\
        &= \frac{2}{\left(1-K\|x\|^2\right)}v
        + \frac{4 K \inner{x}{v} }{\left(1-K\|x\|^2\right)^2 }x.
        \end{aligned}
    \end{equation}

    \mypara{Homomorphism.} The homomorphism w.r.t. the scalar product can be readily obtained by the Riemannian isometry. Therefore, we first address it before proceeding to the addition. For simplicity, we denote $\phi=\pi_{\pball{n} \to \klein{n}}$.

    \mypara{Scalar product.} As shown by \citet{ungar2022analytic}, the geodesics under the Beltrami--Klein and Poincaré ball models are
    \begin{equation}
    \begin{aligned}
        \gamma ^{\mathbb{K}} _{\phi(x) \rightarrow \phi(y)}(t)
        &=\phi(x) \Eoplus t \Eodot \left(-\phi(x) \Eoplus \phi(y)\right) , \\
        \gamma ^{\mathbb{P}} _{ x \rightarrow y}(t)
        &=x \Moplus t \Modot \left(- x \Moplus y \right).
    \end{aligned}
    \end{equation}
    Here, we use the fact that the gyro inverses in the Möbius and Einstein gyrovector spaces are exactly the familiar vector inverse. The above geodesics satisfy
    \begin{equation}
        \begin{aligned}
            \phi(x \Moplus t \Modot \left(-x \Moplus y\right))=\phi(x) \Eoplus  t \Eodot \left( -\phi(x) \Eoplus \phi(y) \right).
        \end{aligned}
    \end{equation}
    The above comes from $\phi(\gamma ^{\mathbb{P}} _{x \rightarrow y}\left( t \right)) = \gamma ^{\mathbb{K}} _{ \phi(x) \rightarrow \phi(y)} \left( t \right)$. In particular, the geodesic starting from the identity element yields
    \begin{equation}
        \begin{aligned}
            \phi(t \Modot x) = \phi(\gamma ^{\mathbb{P}} _{\zerovec \rightarrow x}\left( t \right)) = \gamma ^{\mathbb{K}} _{\phi(\zerovec) \rightarrow \phi(x)}\left( t \right) \stackrel{(1)}{=} t \Eodot \phi(x), \\
        \end{aligned}
    \end{equation}
    where (1) comes from $\phi(\zerovec)=\zerovec$. The above holds for all $x \in \pball{n}$ and $\forall t \in \bbRscalar$, as every hyperbolic geometry is geodesically complete \citep[p.~139]{lee2018introduction}.

    \mypara{Addition.} We first expand the LHS of \cref{eq:iso_addition}. Inspired by \citet[Eqs.~73--76]{mao2024klein}, we express the Möbius addition as $x \Moplus y = \frac{B x +C y}{A}$ by denoting
    \begin{equation}
        \begin{aligned}
            A &= 1-2 K\langle x, y\rangle+K^2\|x\|^2\|y\|^2 ,\\
            B &= 1-2 K\langle x, y\rangle-K\|y\|^2, \\
            C &= 1+K\|x\|^2.
        \end{aligned}
    \end{equation}
    Then, the Möbius addition is
    \begin{equation} \label{app:eq:prf_moplus_simplify}
    \begin{aligned}
        \phi(x \Moplus y)&=\frac{2 \frac{B x+C y}{A}}{1 - K\left\|\frac{B x+C y}{A}\right\|^2} \\
        & =\frac{2 A B x + 2 A C y}{A^2 - K \left\|B x + C y\right\|^2} \\
        & =\frac{2 A B x+2 A C y}{A^2 - B^2 K \left\|x\right\|^2 - C^2 K \left\|y\right\|^2 - 2 B C K \inner{x}{y} } .
    \end{aligned}
    \end{equation}

    Inspired by \citet[Eq.~77]{mao2024klein}, we denote $\|x\|=a, \|y\|=b$, and $\inner{x}{y}=a b \cos(\theta)$. In this way, we can resort to the symbolic computation package \texttt{SymPy} \citep{meurer2017sympy} for the heavy algebra computation, which brings
    \begin{equation}
    \label{app:prf:eq:iso_addition_lhs_final}
    \begin{aligned}
        \phi(x \Moplus y)
         =&  \frac{2 \left(- 2 K a b \cos{\left(\theta \right)} - K b^{2} + 1\right)}{K^{2} a^{2} b^{2} - K a^{2} - 4 K a b \cos{\left(\theta \right)} - K b^{2} + 1} x \\
         & +   \frac{2 K a^{2} + 2}{K^{2} a^{2} b^{2} - K a^{2} - 4 K a b \cos{\left(\theta \right)} - K b^{2} + 1} y.
    \end{aligned}
    \end{equation}

    Now we turn to the RHS of \cref{eq:iso_addition}. For any $u,v \in \klein{n}$, the Einstein addition can be rewritten as
    \begin{equation}
    \label{app:prf:eq:iso_einstein_addition_rewritten}
        \begin{aligned}
        u \oplus_{\mathrm{E}} v
        &=\frac{1}{1 - K\inner{u}{v}}\left(u+ \frac{1}{\gamma_{u}} v -K \frac{\gamma_{u}}{1+\gamma_{u}} \inner{u}{v} u\right) \\
        &=\frac{1 -K \frac{\gamma_{u}}{1+\gamma_{u}} \inner{u}{v} }{1 - K\inner{u}{v}} u + \frac{1}{\gamma_{u} (1 - K\inner{u}{v}) } v.
        \end{aligned}
    \end{equation}
    The gamma factor and inner product under isometry can be rewritten as
    \begin{equation}
        \begin{aligned}
        \gamma_{\phi(x)}
        &=\frac{1}{\sqrt{1+K\|\phi(x)\|^2}} \\
        &=\frac{1}{ \sqrt{1 + K\left( \frac{2}{1-K\|x\|^2} \right )^2\|x\|^2}} \\
        &=\frac{1-K\|x\|^2}{\sqrt{\left(1-K\|x\|^2\right)^2+4 K\|x\|^2}} \\
        &\stackrel{(1)}{=} \frac{1-K\|x\|^2}{1+K\|x\|^2}, \\
        \inner{\phi(x)}{\phi(y)}
        &= \frac{4 }{(1-K\|x\|^2)(1-K\|y\|^2)} \inner{x}{y}.
        \end{aligned}
    \end{equation}
    Here, (1) comes from $\|x\|^2 < -\frac{1}{K} \Rightarrow K\|x\|^2 + 1 > 0$.
    Putting the above into \cref{app:prf:eq:iso_einstein_addition_rewritten} and following the same notation as \cref{app:prf:eq:iso_addition_lhs_final}, we can obtain the following by \texttt{SymPy}:
    \begin{equation}
        \begin{aligned}
        \phi (x) \Eoplus \phi (y)
        =&  \frac{2 \cdot \left(2 K a b \cos{\left(\theta \right)} + K b^{2} - 1\right)}{4 K a b \cos{\left(\theta \right)} - \left(K a^{2} - 1\right) \left(K b^{2} - 1\right)} x \\
        &+  \frac{- 2 K a^{2} - 2}{4 K a b \cos{\left(\theta \right)} - \left(K a^{2} - 1\right) \left(K b^{2} - 1\right)} y,
        \end{aligned}
    \end{equation}
    which is clearly equal to \cref{app:prf:eq:iso_addition_lhs_final}.
\end{proof}

\linkofproof{thm:einstein_klein}
\subsection[Proof of \cref{thm:einstein_klein}]{Proof of \cref{thm:einstein_klein}}

\begin{proof}
    For simplicity, we denote $\phi=\pi_{\pball{n} \to \klein{n}}$. The homomorphism and bijection of $\phi$ imply the homomorphism of its inverse $\phi^{-1}$. Also note that $\phi({\zerovec})=\phi^{-1}({\zerovec})={\zerovec}$. This yields
    \begin{equation}
        \begin{aligned}
            x \Eoplus y
            &= \phi \left( \phi^{-1} (x) \Moplus \phi^{-1} (y) \right) \\
            &\stackrel{(1)}{=} \phi \left( \rieexp ^\mathbb{P} _ {\phi^{-1}(x)} \left(\pt{{\zerovec}}{\phi^{-1}(x)} ^\mathbb{P} \rielog ^\mathbb{P} _\zerovec (\phi^{-1}(y))\right) \right) \\
            &\stackrel{(2)}{=}  \rieexp ^\mathbb{K} _ x \left(\pt{{\zerovec}}{x} ^\mathbb{K} \rielog ^\mathbb{K} _{\zerovec} (y)\right). \\
        \end{aligned}
    \end{equation}
    The above comes from the following.
    \begin{enumerate}
        \item
        For any Poincaré vectors $u, w \in \pball{n}$, the following holds:
        \begin{equation}
            \pt{{\zerovec}}{u} ^\mathbb{P} (v) = \rielog ^\mathbb{P} _{u} (u \Moplus \rieexp ^\mathbb{P} _{{\zerovec}} (v))
            \Rightarrow
            u \Moplus w
            = \rieexp ^\mathbb{P} _u \left(\pt{{\zerovec}}{u} ^\mathbb{P} (\rielog ^\mathbb{P} _{\zerovec} (w)) \right),
        \end{equation}
        where $v= \rielog ^\mathbb{P} _\zerovec (w)$ and the LHS comes from \citet[Thm.~4]{ganea2018hyperbolic};
        \item
        The second identity follows from the isometry:
        \begin{equation}
            \begin{aligned}
                \rieexp ^\mathbb{P} _ x (v)
                &=  \phi^{-1} \left( \rieexp ^\mathbb{K} _ {\phi(x)} (\phi_{*,x} (v)) \right), \\
                \rielog ^\mathbb{P} _ x (y)
                &=  \phi_{*,x}^{-1} \left( \rielog ^\mathbb{K} _ {\phi(x)} (\phi (y)) \right), \\
                \pt{x}{y} ^\mathbb{P} (v)
                &=  \phi_{*,y}^{-1} \left( \pt{\phi(x)}{\phi(y)} ^\mathbb{K} (\phi_{*,x} (v) ) \right).
            \end{aligned}
        \end{equation}
    \end{enumerate}
    Similar to the gyro addition, the isometry of $\phi$ implies the following with respect to the gyro scalar product:
    \begin{equation}
        \begin{aligned}
            t \Eodot x
            &= \phi \left( t \Modot \phi^{-1} (x) \right)\\
            &\stackrel{(1)}{=} \phi \left( \rieexp ^\mathbb{P} _{\zerovec} \left( t \rielog ^\mathbb{P} _{\zerovec} (\phi^{-1} (x))\right) \right)\\
            &\stackrel{(2)}{=} \rieexp ^\mathbb{K} _{\zerovec} \left( t \rielog ^\mathbb{K} _{\zerovec} (x)\right).
        \end{aligned}
    \end{equation}
    The above comes from the following.
    \begin{enumerate}
        \item
        \citet[Lem.~3]{ganea2018hyperbolic};
        \item The isometry of $\phi$.
    \end{enumerate}
\end{proof}

\linkofproof{thm:riem_klein}
\subsection[Proof of \cref{thm:riem_klein}]{Proof of \cref{thm:riem_klein}}

\begin{proof}
    Following \cref{thm:einstein_klein}, we denote $\phi = \pi_{\pball{n} \to \klein{n}}$ and $\psi =\pi_{\klein{n} \to \pball{n}}$. The results can be obtained by the properties of isometry and isomorphism of $\phi$.

    \mypara{Riemannian exponential and logarithmic maps at the zero vector.} First, we recall that the expressions of $\rieexp ^{\mathbb{P}} _{\zerovec}(v)$ and $\rielog ^{\mathbb{P}} _{\zerovec}(v)$ under the Poincaré ball model are exactly those in \cref{eq:exp_0_klein,eq:log_0_klein} \citep[Eq.~13]{ganea2018hyperbolic}.

    By Riemannian isometry, we have the following:
    \begin{equation}
        \begin{aligned}
            \rieexp ^\mathbb{K} _{\zerovec}(v)
            &= \phi \left( \rieexp ^\mathbb{P} _{\zerovec}(\psi _{*,\zerovec } (v)) \right) \\
            &= \phi\left(\tanh \left(\frac{\sqrt{-K}}{2}\|v\|\right) \frac{v}{\sqrt{-K}\|v\|}\right) \\
            &= \frac{2}{1-K\| v \|^2\left(\frac{\tanh \left(\frac{\sqrt{-K}}{2}\|v\|\right)}{\sqrt{-K}\|v\|}\right)^2} \frac{\tanh \left(\frac{\sqrt{-K}}{2}\|v\|\right)}{\sqrt{-K}\|v\|} v \\
            & =\frac{2 \tanh \left(\frac{\sqrt{-K}}{2}\|v\|\right)}{1+\tanh \left(\frac{\sqrt{-K}}{2}\|v\|\right)^2} \frac{v}{\sqrt{-K}\|v\|} \\
            & \stackrel{(1)}{=}\tanh (\sqrt{-K}\|v\|) \frac{v}{\sqrt{-K}\|v\|}, \\
        \end{aligned}
    \end{equation}
    where (1) comes from $\tanh (2 x)=\frac{2 \tanh x}{1+\tanh ^2 x}$.

    As the inverse of $\rieexp ^\mathbb{K} _{\zerovec}(v)$, $\rielog ^\mathbb{K} _{\zerovec}(x)$, therefore, shares the expression with its counterpart under the Poincaré ball model. Here, we use the properties of isometry to further validate this result:
    \begin{equation}
        \begin{aligned}
            \rielog ^\mathbb{K} _{\zerovec}(x)
            &= \phi _{*,0} \left( \rielog ^\mathbb{P} _{\zerovec}( \psi(x))\right) \\
            & =2 \tanh ^{-1}\left(\frac{\sqrt{-K}\|x\|}{1+\sqrt{1+K\norm{x}^2}}\right) \frac{x}{\sqrt{-K}\|x\|}.
        \end{aligned}
    \end{equation}

    We only need to show
    \begin{equation}
        2 \tanh ^{-1}\left(\frac{\sqrt{-K}\|x\|}{1+\sqrt{1+K\|x\|^2}}\right)=\tanh ^{-1}(\sqrt{-K}\|x\|).
    \end{equation}
    We denote $a = \sqrt{-K}\|x\|<1$. Both sides are
    \begin{equation} \label{app:prf:eq:log-exp-ln-equation1}
        \begin{aligned}
            \text{LHS: }& 2 \tanh ^{-1}\left(\frac{a}{1+\sqrt{1-a^2}}\right) = \ln \left(\frac{1+\frac{a}{1+\sqrt{1-a^2}}}{1-\frac{a}{1+\sqrt{1-a^2}}}\right)=\ln \left(\frac{1+\sqrt{1-a^2}+a}{1+\sqrt{1-a^2}-a}\right), \\
            \text{RHS: }& \tanh ^{-1}(a) = \ln \left( \frac{\sqrt{1+a}}{\sqrt{1-a}} \right) = \ln \left( \frac{\sqrt{1-a^2}}{1-a} \right).
        \end{aligned}
    \end{equation}
    Note that the below equation holds:
    \begin{equation} \label{app:prf:eq:log-exp-ln-equation2}
        \frac{1+\sqrt{1-a^2}+a}{1+\sqrt{1-a^2}-a} = \frac{\sqrt{1-a^2}}{1-a}.
    \end{equation}

    \mypara{Geodesic distances.}
    \begin{equation}
        \begin{aligned}
        \dist ^{\mathbb{K}} (x, y)
        &\stackrel{(1)}{=} \dist ^{\mathbb{P}} (\psi(x), \psi(y)) \\
        &=\frac{2}{\sqrt{-K}} \tanh ^{-1}\left(\sqrt{-K}\left\|-\psi(x) \Moplus \psi(y)\right\|\right) \\
        & \stackrel{(2)}{=} \frac{2}{\sqrt{-K}} \tanh ^{-1}\left(\sqrt{-K}\left\|\psi\left(-x \Eoplus y\right)\right\|\right) \\
        & =\frac{2}{\sqrt{-K}} \tanh ^{-1}\left(\sqrt{-K} \frac{ \norm{-x \Eoplus y}}{1+\sqrt{1+K\left\|-x \Eoplus y\right\|^2}} \right),
        \end{aligned}
    \end{equation}
    where (1) comes from the isometry, while (2) comes from the isomorphism.

    \mypara{Exponential maps.}
    \begin{equation}
        \begin{aligned}
        \rieexp _x ^{\mathbb{K}} (v)
        & \stackrel{(1)}{=} \psi^{-1} \left(\rieexp_{\psi(x)} ^{\mathbb{P}} \left( \psi_{*, x}(v) \right)\right) \\
        & \stackrel{(2)}{=} \psi^{-1} \left( \psi(x) \Moplus \rieexp _{\zerovec}\left(\frac{\lambda_{\psi(x)}^K}{2} \psi_{*, x}(v) \right) \right) \\
        &\stackrel{(3)}{=} x \Eoplus \psi^{-1}\left(\rieexp _{\zerovec}\left(\frac{\lambda_{\psi(x)}^K}{2} \psi_{*, x}(v) \right)\right) \\
        &\stackrel{(4)}{=} x \Eoplus \rieexp _{\zerovec}\left( \lambda_{\psi(x)}^K \psi_{*, x}(v)  \right) . \\
        \end{aligned}
    \end{equation}
    The above comes from the following.
    \begin{enumerate}
        \item
        The isometry of $\psi$;
        \item
        $\rieexp _x ^\mathbb{P} (v) = x \Moplus \rieexp _\zerovec \left(\frac{\lambda_{x}^K}{2} v \right)$;
        \item
        The isomorphism of $\psi$;
        \item
        \begin{equation}
            \begin{aligned}
                \psi^{-1} \circ \rieexp _{\zerovec} \left( v \right)
                &= \psi^{-1} \circ \rieexp ^{\mathbb{P}}_{\zerovec} \left( v \right) \\
                &= \psi^{-1} \circ \psi \circ \rieexp ^{\mathbb{K}}_{\zerovec} \left( 2v \right) \\
                &= \rieexp _{\zerovec} \left( 2v \right).
            \end{aligned}
        \end{equation}
    \end{enumerate}

    It remains to calculate $\lambda_{\psi(x)}^K \psi_{*, x}(v)$:
    \begin{equation}
        \begin{aligned}
            \lambda_{\psi(x)}^K
            &= \frac{2}{1+ K \frac{\norm{x}^2}{\left(1+\sqrt{1+K\|x\|^2}\right)^2}} \\
            &= \frac{2\left(1+\sqrt{1+K\|x\|^2}\right)^2}{\left(1+\sqrt{1+K\|x\|^2}\right)^2+K\|x\|^2} \\
            &= \frac{2\left(1+\sqrt{1+K\|x\|^2}\right)^2}{ 2 +2 \sqrt{1+K\|x\|^2}+2K\|x\|^2 } \\
            &= \frac{\left(1+\sqrt{1+K\|x\|^2}\right)^2}{ 1 + \sqrt{1+K\|x\|^2}+K\|x\|^2} \\
            &= \frac{\left(1+\sqrt{1+K\|x\|^2}\right)}{\sqrt{1+K\|x\|^2}}, \\
        \end{aligned}
    \end{equation}
    \begin{equation}
        \begin{aligned}
            \lambda_{\psi(x)}^K \psi_{*, x}(v)
            &= \lambda_{\psi(x)}^K \left( \frac{1}{1 + \sqrt{1+K\|x\|^2}} v
            - \frac{K \inner{x}{v}}{\left(1 + \sqrt{1+K\|x\|^2}\right)^2 \sqrt{1+K\|x\|^2}} x \right) \\
            &= \frac{1}{\sqrt{1+K\|x\|^2}} v
            - \frac{K \inner{x}{v}}{\left(1 + \sqrt{1+K\|x\|^2}\right) (1+K\|x\|^2)} x .
        \end{aligned}
    \end{equation}

    \mypara{Logarithmic maps.}
    \begin{equation}
        \begin{aligned}
            \rielog _x ^{\mathbb{K}} (y)
            &\stackrel{(1)}{=} \phi _{*, \psi(x)} \left( \rielog _{\psi(x)} ^{\mathbb{P}} \psi(y)\right) \\
            &\stackrel{(2)}{=} \phi _{*, \psi(x)} \left( \frac{2}{ \lambda _{\psi(x)} ^K} \rielog _{\zerovec}\left(-\psi(x) \Moplus \psi(y) \right)\right) \\
            &\stackrel{(3)}{=} \frac{2}{\lambda_{\psi(x)}^K} \phi_{*, \psi(x)}\left(\rielog _{\zerovec}\left(-\psi(x) \Moplus \psi(y)\right)\right) \\
            &\stackrel{(4)}{=} \frac{1}{\lambda_{\psi(x)}^K} \phi_{*, \psi(x)}\left(\rielog _{\zerovec}\left( -x \Eoplus  y\right)\right) .
        \end{aligned}
    \end{equation}
    The above comes from the following.
    \begin{enumerate}
        \item
        The isometry of $\psi$;
        \item
        $\rielog _{x} ^{\mathbb{P}} (y) = \frac{2}{ \lambda _x ^K} \rielog _{\zerovec} \left(  -x \Moplus y \right)$;
        \item
        The linearity of differential maps;
        \item
        \begin{equation}
            \begin{aligned}
                \rielog _{\zerovec}\left(-\psi(x) \Moplus \psi(y)\right)
                &= \rielog ^\mathbb{P} _{\zerovec}\left(-\psi(x) \Moplus \psi(y)\right) \\
                &= \frac{1}{2}\rielog ^\mathbb{K} _{\zerovec}\left( \psi^{-1} \left(-\psi(x) \Moplus \psi(y)\right) \right) \\
                &= \frac{1}{2}\rielog ^\mathbb{K} _{\zerovec}\left( -x \Eoplus  y \right).
            \end{aligned}
        \end{equation}

    \end{enumerate}
\end{proof}

\begin{parisremark}
   \citet[Thm.~9]{mao2024klein} extended the Möbius matrix-vector multiplication \citep[Lem.~6]{ganea2018hyperbolic} to the Einstein space of the Beltrami--Klein model under $K=-1$, namely $\rieexp _{\zerovec} \left(M \rielog _{\zerovec} (x))\right)$. Although their presented formulations are different from the Möbius one, this theorem indicates that matrix-vector multiplications over these two spaces are identical under any negative curvature. The equality can also be readily observed from \cref{app:prf:eq:log-exp-ln-equation1,app:prf:eq:log-exp-ln-equation2}.
\end{parisremark}

\section{SPD Multinomial Logistic Regression}
\label{app:spdmlr-proofs}

\subsection[Proof of \cref{spdmlr:prop:hyperplanes_as_submanifolds}]{Proof of \cref{spdmlr:prop:hyperplanes_as_submanifolds}}
\label{spdmlr:app:subsec:proof_hyperplanes_as_submanifolds}
\linkofproof{spdmlr:prop:hyperplanes_as_submanifolds}

This claim can be proven either by definition \citep[Def.~9.1]{loring2011introduction} or by the constant rank level set theorem \citep[Thm.~11.2]{loring2011introduction}.
We focus on the latter.
\begin{proof}
Consider any $P\in\spd{n}$ and $A\in T_P\spd{n}\backslash\{\bbzero\}$. Define the function
\begin{equation}
    f:\spd{n}\rightarrow\bbRscalar,\qquad S\mapsto\left\langle\rielog_P S,A\right\rangle_P.
\end{equation}
For the SPD hyperplane $\tilde{H}_{A,P}$, we have $\tilde{H}_{A,P}=f^{-1}(0)$.
By assumption, $\rielog_P:\spd{n}\rightarrow T_P\spd{n}$ is a global diffeomorphism, and $f$ is therefore well-defined.
We can rewrite $f$ as a composition, \ie $f=h\circ\rielog_P$, where $h(\cdot)=\langle\cdot,A\rangle_P$ is a linear map.

Since $\rielog_P$ is a diffeomorphism and $h(\cdot)$ is a nonzero linear map, the rank of $f$ is globally constant. Therefore, there exists a neighborhood, \eg the whole SPD manifold, of $f^{-1}(0)$ where the rank of $f$ is constant.
According to the constant rank level set theorem \citep[Thm.~11.2]{loring2011introduction}, we can obtain the claim.
\end{proof}

\subsection[Proof of \cref{spdmlr:lem:dist_to_hyperplane_pems}]{Proof of \cref{spdmlr:lem:dist_to_hyperplane_pems}}
\label{spdmlr:app:subsec:proof_dist_to_hyperplane_pems}
\linkofproof{spdmlr:lem:dist_to_hyperplane_pems}

\begin{proof}
By \cref{alem:lem:g_spd}, we have
\begin{align}
    \left\langle\rielog_P Q,A\right\rangle_P
    &=\left\langle\phi_{*,P}\diffphiinv{\phi(P)}\left(\phi(Q)-\phi(P)\right),\phi_{*,P}A\right\rangle\\
    &=\left\langle\phi(Q)-\phi(P),\phi_{*,P}A\right\rangle.
\end{align}
Therefore, due to the isometry of $\phi$, the SPD hyperplane $\tilde{H}_{A_k,P_k}$ corresponds to the Euclidean hyperplane
\begin{equation}
    H_{\phi_{*,P_k}(A_k),\phi(P_k)}.
\end{equation}
Furthermore, the distance to the margin hyperplane is equivalent to
\begin{gather}
    \inf_{\phi(Q)}\left\|\phi(S)-\phi(Q)\right\|_{\mathrm{F}}\\
    \st \left\langle\phi(Q)-\phi(P_k),\phi_{*,P_k}A_k\right\rangle=0.
\end{gather}
The problem above is the familiar Euclidean distance from a point to a hyperplane.
By simple computation, one can obtain the result.
\end{proof}

\subsection[Proof of \cref{spdmlr:lem:equ_pt_and_lt}]{Proof of \cref{spdmlr:lem:equ_pt_and_lt}}
\label{spdmlr:app:subsec:proof_equ_pt_and_lt}
\linkofproof{spdmlr:lem:equ_pt_and_lt}

\begin{proof}
For simplicity, we abbreviate $\phiMul$ and $\gphi$ as $\odot$ and $g$.
By abuse of notation, we further denote $Q\odot P^{-1}_{\odot}$ as $QP^{-1}$, where $P^{-1}_{\odot}$ is the inverse of $P$ under $\odot$.
According to \cref{alem:lem:g_spd}, $(\spd{n},\odot)$ is an abelian group and $g$ is a bi-invariant Riemannian metric.
By \citet[Lem.~6]{lin2019riemannian}, any parallel transport can be expressed as a differential of left translation,
\begin{equation}
    \pt{P}{Q}=\left(L_{QP^{-1}}\right)_{*,P}, \quad \forall P,Q\in\spd{n}.
\end{equation}
\end{proof}

\subsection[Proof of \cref{spdmlr:lem:pt_anchor_invariance}]{Proof of \cref{spdmlr:lem:pt_anchor_invariance}}
\label{spdmlr:app:subsec:proof_pt_anchor_invariance}
\linkofproof{spdmlr:lem:pt_anchor_invariance}

\begin{proof}
By \cref{alem:eq:gene_pt_spd}, parallel transport under a pullback Euclidean metric is path-independent and satisfies
\begin{equation}
    \pt{Q}{P}(V)=\diffphiinv{\phi(P)}\left(\diffphi{Q}(V)\right).
\end{equation}
For any $\tilde{A}_{1,k}\in T_{Q_1}\spd{n}$, define
\begin{equation}
    \tilde{A}_{2,k}
    =\diffphiinv{\phi(Q_2)}\left(\diffphi{Q_1}\left(\tilde{A}_{1,k}\right)\right)
    \in T_{Q_2}\spd{n}.
\end{equation}
Since $\phi$ is a diffeomorphism, its differential at every point is a linear isomorphism. Hence,
\begin{equation}
    \begin{aligned}
    \pt{Q_2}{P_k}\left(\tilde{A}_{2,k}\right)
    &=\diffphiinv{\phi(P_k)}\left(\diffphi{Q_2}\left(\tilde{A}_{2,k}\right)\right)\\
    &=\diffphiinv{\phi(P_k)}\left(\diffphi{Q_1}\left(\tilde{A}_{1,k}\right)\right)\\
    &=\pt{Q_1}{P_k}\left(\tilde{A}_{1,k}\right).
    \end{aligned}
\end{equation}
Moreover, if $\tilde{B}_{2,k}\in T_{Q_2}\spd{n}$ satisfies the same equality, applying $\diffphi{P_k}$ to both sides gives
\begin{equation}
    \diffphi{Q_2}\left(\tilde{B}_{2,k}\right)
    =\diffphi{Q_1}\left(\tilde{A}_{1,k}\right).
\end{equation}
The invertibility of $\diffphi{Q_2}$ then yields $\tilde{B}_{2,k}=\tilde{A}_{2,k}$, proving uniqueness.
\end{proof}

\subsection[Proof of \cref{spdmlr:thm:general_mlr_pems}]{Proof of \cref{spdmlr:thm:general_mlr_pems}}
\label{spdmlr:app:subsec:proof_general_mlr_pems}
\linkofproof{spdmlr:thm:general_mlr_pems}

\begin{proof}
\begin{align}
    A_k&=\pt{I}{P_k}(\tilde{A}_k)\\
       &=\diffphiinv{\phi(P_k)}\left(\diffphi{I}(\tilde{A}_k)\right).
    \label{spdmlr:eq:pt_i_to_pk}
\end{align}
One can obtain the result by putting \cref{spdmlr:eq:pt_i_to_pk} into \cref{spdmlr:eq:spd_dist_s_to_h}.
\end{proof}

\subsection[Proof of \cref{spdmlr:cor:spd_mlr_param_lem_lcm}]{Proof of \cref{spdmlr:cor:spd_mlr_param_lem_lcm}}
\label{spdmlr:app:subsec:proof_spd_mlr_param_lem_lcm}
\linkofproof{spdmlr:cor:spd_mlr_param_lem_lcm}

\begin{proof}
Denoting the matrix power as $\pow_\theta:\spd{n}\rightarrow\spd{n}$, we have
\begin{align}
    \pow_\theta(I)&=I,\\
    \label{spdmlr:eq:diff_power_diform_map}
    (\pow_\theta)_{*,I}(A)&=\theta A, \quad \forall A\in T_I\spd{n}.
\end{align}
Next, we prove the two cases separately.

\mypara{$\biparamLEM$.} We define the following map:
\begin{equation}
    \label{spdmlr:eq:pb_map_def_lem}
    \psi^{\lem}=f\circ\mlog,
\end{equation}
where $f:\sym{n}\rightarrow\sym{n}$ is the linear isometry between the standard Frobenius inner product and the $\orth{n}$-invariant inner product $\langle\cdot,\cdot\rangle^{\alphabeta}$.
Then $\psi^{\lem}$ pulls back the standard Euclidean metric on $\sym{n}$ to $\biparamLEM$ on $\spd{n}$.
Putting \cref{spdmlr:eq:diff_power_diform_map,spdmlr:eq:pb_map_def_lem} into \cref{spdmlr:eq:spd_dist_s_to_h_final}, we have
\begin{equation}
    \begin{aligned}
    &\exp\left(\left\langle\psi^{\lem}(S)-\psi^{\lem}(P_k),\psi^{\lem}_{*,I}(\tilde{A}_k)\right\rangle\right)\\
    &=\exp\left[\left\langle f\left(\mlog(S)-\mlog(P_k)\right),f(\tilde{A}_k)\right\rangle\right]\\
    &=\exp\left[\left\langle\mlog(S)-\mlog(P_k),\tilde{A}_k\right\rangle^{\alphabeta}\right],
    \end{aligned}
\end{equation}
where the last equality follows from the linearity and isometry property of $f$, with $f=f_{*}$.

\mypara{$\paramLCM$.} We denote
\begin{equation}
    \label{spdmlr:eq:psi_lcm}
    \psi^{\lcm}=\dlog\circ\chol\circ\pow_\theta.
\end{equation}
Then $\psi^{\lcm}$ pulls back the Euclidean metric $\frac{1}{\theta^2}\geuc$ on the Euclidean space $\trilspace{n}$ of lower triangular matrices to $\paramLCM$ on $\spd{n}$.
The differential of Cholesky decomposition is presented by \citet[Prop.~4]{lin2019riemannian}, while the differential of $\dlog$ is obtained as the natural-logarithm specialization of \cref{alem:props:diff_mgexp_mlog}.
Then, simple computations show that
\begin{equation}
    \label{spdmlr:eq:diff_psi_lcm}
    \psi^{\lcm}_{*,I}(A)=\theta\left(\lfloor A\rfloor+\frac{1}{2}\bbD(A)\right), \quad \forall A\in T_I\spd{n}.
\end{equation}
Putting \cref{spdmlr:eq:psi_lcm,spdmlr:eq:diff_psi_lcm} into \cref{spdmlr:eq:spd_dist_s_to_h_final}, we can obtain the result.
\end{proof}

\subsection[Proof of \cref{spdmlr:prop:equivalence_lem_mlr}]{Proof of \cref{spdmlr:prop:equivalence_lem_mlr}}
\label{spdmlr:app:subsec:proof_equivalence_lem_mlr}
\linkofproof{spdmlr:prop:equivalence_lem_mlr}

To prove \cref{spdmlr:prop:equivalence_lem_mlr}, we first present two lemmas about the general cases under pullback Euclidean metrics.

One can observe that \cref{spdmlr:eq:spd_dist_s_to_h_final,spdmlr:eq:spd_hyperplane_final} are very similar to those of a Euclidean MLR.
However, since $\phi$ is normally nonlinear and $P_k$ is an SPD parameter, \cref{spdmlr:eq:spd_dist_s_to_h_final} cannot hastily be identified with a Euclidean MLR.
Under some special circumstances, SPD MLR can be reduced to the familiar Euclidean MLR.
To show this result, we first specialize the RSGD strategy reviewed in \cref{sec:ch2-riemannian-optimization} to pullback Euclidean metrics. In ambient-gradient notation, the required update is
\begin{equation}
    \label{spdmlr:eq:general_rsgd}
    W_{t+1}=\rieexp_{W_t}\left(-\gamma_t\Pi_{W_t}\left(\nabla_{W_t} f\right)\right),
\end{equation}
where $\Pi_{W_t}$ denotes the projection mapping the Euclidean gradient $\nabla_{W_t} f$ to the Riemannian gradient, and $\gamma_t$ denotes the learning rate.
We have already obtained the formula for the Riemannian exponential map in \cref{alem:eq:gene_rie_exp_spd}.
We proceed to formulate $\Pi$.

\begin{parislemma}
\label{spdmlr:lem:rsgd_pems}
For a smooth function $f:\spd{n}\rightarrow\bbRscalar$ on $\spd{n}$ endowed with any kind of pullback Euclidean metric, the projection map $\Pi_P:\sym{n}\rightarrow T_P\spd{n}$ at $P\in\spd{n}$ is
\begin{equation}
    \label{spdmlr:eq:proj_pems}
    \Pi_P(\nabla_P f)
    =\phi^{-1}_{*,\phi(P)}\left(\phi^{-*}_{*,\phi(P)}(\nabla_P f)\right),
\end{equation}
where $\phi^{-*}_{*,\phi(P)}:=\left(\phi^{-1}_{*,\phi(P)}\right)^*:T_P\spd{n}\rightarrow T_{\phi(P)}\sym{n}$ is the Frobenius adjoint of $\phi^{-1}_{*,\phi(P)}:T_{\phi(P)}\sym{n}\rightarrow T_P\spd{n}$. Specifically, for all $U\in T_P\spd{n}$ and $Z\in T_{\phi(P)}\sym{n}$, it satisfies
\begin{equation}
    \left\langle U,\phi^{-1}_{*,\phi(P)}(Z)\right\rangle
    =\left\langle\phi^{-*}_{*,\phi(P)}(U),Z\right\rangle.
\end{equation}
\end{parislemma}
\begin{proof}
Given any smooth function $f:\spd{n}\rightarrow\bbRscalar$, denote its Riemannian gradient at $P$ as $\operatorname{grad}_P f\in T_P\spd{n}$.
\begin{equation}
    \left\langle\operatorname{grad}_P f,V\right\rangle_P=f_{*,P}(V), \quad \forall V\in T_P\spd{n}.
\end{equation}
Let $\nabla_P f$ denote the Euclidean gradient in the canonical chart. For any $Z\in T_{\phi(P)}\sym{n}$, set $V=\phi^{-1}_{*,\phi(P)}(Z)$. By \cref{alem:eq:phi_g} and the definition of the Frobenius adjoint, we have
\begin{equation}
    \begin{aligned}
    \left\langle\phi_{*,P}\left(\operatorname{grad}_P f\right),Z\right\rangle
    &=\left\langle\operatorname{grad}_P f,\phi^{-1}_{*,\phi(P)}(Z)\right\rangle_P\\
    &=f_{*,P}\left(\phi^{-1}_{*,\phi(P)}(Z)\right)\\
    &=\left\langle\nabla_P f,\phi^{-1}_{*,\phi(P)}(Z)\right\rangle\\
    &=\left\langle\phi^{-*}_{*,\phi(P)}\left(\nabla_P f\right),Z\right\rangle.
    \end{aligned}
\end{equation}
Since $Z$ is arbitrary, the nondegeneracy of the Frobenius inner product gives
\begin{equation}
    \phi_{*,P}\left(\operatorname{grad}_P f\right)=\phi^{-*}_{*,\phi(P)}\left(\nabla_P f\right).
\end{equation}
Applying $\phi^{-1}_{*,\phi(P)}$ to both sides yields
\begin{equation}
    \operatorname{grad}_P f
    =\phi^{-1}_{*,\phi(P)}\left(\phi^{-*}_{*,\phi(P)}\left(\nabla_P f\right)\right).
\end{equation}
By definition, $\Pi_P(\nabla_P f)=\operatorname{grad}_P f$, which yields \cref{spdmlr:eq:proj_pems}.
\end{proof}

We can describe the special case mentioned above with this lemma.
\begin{parislemma}
\label{spdmlr:lem:spdmlr_to_emlr}
Suppose the differential map $\phi_{*,I}$ is the identity map, and $P_k$ in \cref{spdmlr:eq:spd_dist_s_to_h_final} is optimized by pullback-Euclidean-metric-based RSGD. Then \cref{spdmlr:eq:spd_dist_s_to_h_final} can be reduced to a Euclidean MLR in the codomain of $\phi$ updated by Euclidean SGD.
\end{parislemma}
\begin{proof}
Define a Euclidean MLR in the codomain of $\phi$ as
\begin{equation}
    p(y=k \mid S)\propto\exp\left(\left\langle\phi(S)-\bar{P}_k,\bar{A}_k\right\rangle\right),
\end{equation}
where $\bar{P}_k,\bar{A}_k\in\sym{n}$. We call this classifier $\phi$-EMLR.

Define the SPD MLR under the pullback Euclidean metric induced by $\phi$ as
\begin{equation}
    p(y=k \mid S)\propto\exp\left(\left\langle\phi(S)-\phi(P_k),\tilde{A}_k\right\rangle\right),
\end{equation}
where $P_k\in\spd{n}$ and $\tilde{A}_k\in\sym{n}$.

Suppose the SPD MLR and $\phi$-EMLR satisfy $\bar{P}_k=\phi(P_k)$.
Other settings of the network are all the same, indicating that the Euclidean gradients satisfy
\begin{equation}
    \frac{\partial L}{\partial\bar{P}_k}=\frac{\partial L}{\partial\phi(P_k)}.
\end{equation}
The update of $\bar{P}_k$ in the $\phi$-EMLR is
\begin{equation}
    \bar{P}'_k=\bar{P}_k-\gamma\frac{\partial L}{\partial\bar{P}_k}.
\end{equation}
The update of $P_k$ in the SPD MLR is
\begin{align}
    P'_k&=\rieexp_{P_k}\left(-\gamma\Pi_{P_k}\left(\frac{\partial L}{\partial P_k}\right)\right)\\
        &=\phi^{-1}\left(\phi(P_k)-\gamma\phi_{*,P_k}^{-*}\frac{\partial L}{\partial P_k}\right).
\end{align}
Therefore, $\phi(P'_k)$ satisfies
\begin{align}
    \phi(P'_k)
    &=\phi(P_k)-\gamma\phi_{*,P_k}^{-*}\frac{\partial L}{\partial P_k}\\
    &=\phi(P_k)-\gamma\phi_{*,P_k}^{-*}\phi_{*,P_k}^{*}\frac{\partial L}{\partial\phi(P_k)}
    \label{spdmlr:eq:phi_p_bp}\\
    &=\phi(P_k)-\gamma\frac{\partial L}{\partial\phi(P_k)}\\
    &=\bar{P}'_k.
\end{align}
\cref{spdmlr:eq:phi_p_bp} comes from the Euclidean chain rule of the differential.
Let $Y=\phi(X)$. Then we have
\begin{equation}
    \frac{\partial L}{\partial Y}:\diff Y
    =\frac{\partial L}{\partial Y}:\phi_{*,X}\diff X
    =\phi_{*,X}^{*}\frac{\partial L}{\partial Y}:\diff X,
\end{equation}
where $:$ denotes the Frobenius inner product.

The equivalence of $\bar{A}_k$ and $\tilde{A}_k$ is obvious.
By mathematical induction, the claim can be proven.
\end{proof}

\begin{proof}
The claim follows directly from \cref{spdmlr:lem:spdmlr_to_emlr}.
\end{proof}

\section{Riemannian Multinomial Logistic Regression}
\label{app:rmlr-proofs}

\subsection[Proof of \cref{rmlr:thm:rie_margin_dist}]{Proof of \cref{rmlr:thm:rie_margin_dist}}
\linkofproof{rmlr:thm:rie_margin_dist}

\begin{proof}
    Let us first solve $Y^*$ in \cref{rmlr:eq:margin_dist_reform_v2}, which is the solution to the following constrained optimization problem:
    \begin{equation} \label{rmlr:eq:q_problem_v1}
        \underset{Y}{\max} \left(\frac{\langle \rielog_P Y, \rielog_P S \rangle_P}{\|\rielog_P Y\|_P\|\rielog_P S\|_P} \right)
        \quad
        \st \langle \rielog_{P} Y, \tilde{A} \rangle_{P}=0.
    \end{equation}
    Note that \cref{rmlr:eq:q_problem_v1} is well-defined due to the existence of the Riemannian logarithm.
    Although \cref{rmlr:eq:q_problem_v1} is normally non-convex, \cref{rmlr:eq:q_problem_v1} and \cref{rmlr:eq:margin_dist_reform_v2} can be reduced to a Euclidean problem:
    \begin{align}
        \label{rmlr:eq:max_cos_tpm}
        \underset{\tilde{Y}}{\max} \frac{\langle \tilde{Y}, \tilde{S}\rangle_P}{\| \tilde{Y} \|_P\| \tilde{S} \|_P} \quad \st \langle \tilde{Y}, \tilde{A} \rangle_P=0,\\
        \label{rmlr:eq:dist_S_to_hp_v2}
        d(S,\tilde{H}_{\tilde{A}, P})=\sin (\angle SPY^*) \| \tilde{S} \|_P,
    \end{align}
    where $\tilde{Y}=\rielog_P Y$ and $\tilde{S}=\rielog_P S$.

    Let us first discuss \cref{rmlr:eq:max_cos_tpm}.
    Denote the solution of \cref{rmlr:eq:max_cos_tpm} as $\tilde{Y}^*$.
    Note that $\tilde{Y}^*$ is not necessarily unique.
    Note that $\rieexp_P$ is only well-defined locally.
    More precisely, $\rieexp_P$ is well-defined in an open ball $\mathrm{B}_{\epsilon}(0)$ centered at $0 \in T_P\calM$.
    Therefore, $\tilde{Y}^*$ might not be in $\mathrm{B}_{\epsilon}(0)$.
    In this case, we can scale $\tilde{Y}^*$ into $\mathrm{B}_{\epsilon}(0)$, and the scaled $\tilde{Y}^*$ is still the maximizer of \cref{rmlr:eq:max_cos_tpm}.
    Therefore, without loss of generality, we assume $\tilde{Y}^* \in \mathrm{B}_{\epsilon}(0)$.

    Putting $\tilde{Y}^*$ into \cref{rmlr:eq:dist_S_to_hp_v2}, \cref{rmlr:eq:dist_S_to_hp_v2} is reduced to the distance to the hyperplane $\langle \tilde{Y}, \tilde{A} \rangle_P=0$ in the Euclidean space $(T_P\calM,\langle\cdot,\cdot\rangle_P)$, which has a closed-form solution:
    \begin{align}
        d(S,\tilde{H}_{\tilde{A}, P}) 
        &= \frac{|\langle \tilde{S}, \tilde{A} \rangle_P|}{\| \tilde{A} \|_P}\\
        \label{rmlr:eq:margin_dist_final}
        &= \frac{|\langle  \rielog_P S, \tilde{A} \rangle _P|}{\| \tilde{A} \|_P}.
    \end{align} 
\end{proof}
\subsection[Proof of \cref{rmlr:thm:rmlr}]{Proof of \cref{rmlr:thm:rmlr}}
\linkofproof{rmlr:thm:rmlr}

\begin{proof}
    Putting the margin distance (\cref{rmlr:eq:rie_margin_dist}) into \cref{rmlr:eq:rmlr_v1}, we have the following:
    \begin{equation}
        \begin{aligned}
            p(y=k \mid S)
            &\propto \exp \left(\operatorname{sign}(\langle \tilde{A}_k, \rielog_{P_k}(S) \rangle_{P_k})\|\tilde{A}_k\|_{P_k} d (S, \tilde{H}_{\tilde{A}_k, P_k}) \right)\\
            &= \exp \left(\operatorname{sign}(\langle \tilde{A}_k, \rielog_{P_k}(S) \rangle_{P_k})\|\tilde{A}_k\|_{P_k} \frac{|\langle  \rielog_{P_k} (S), \tilde{A}_k \rangle _{P_k}|}{\| \tilde{A}_k \|_{P_k}} \right)\\
            &= \exp \left( \langle \rielog_{P_k} S,  \tilde{A}_k \rangle_{P_k} \right).
        \end{aligned}
    \end{equation}
\end{proof}

\subsection[Proof of \cref{rmlr:prop:spd_parametrized_metrics}]{Proof of \cref{rmlr:prop:spd_parametrized_metrics}}
\linkofproof{rmlr:prop:spd_parametrized_metrics}

\begin{proof}
    The Riemannian metric $\biparamEM$ at $I$ is
    \begin{equation}
        g_I^{\alphabeta\text{-EM}}(V,V) = \langle V, V\rangle^{\alphabeta}. 
    \end{equation}
    By \cref{rmlr:lem:diformed_metrics_lim}, we have the following:
    \begin{equation}
        \begin{aligned}
            \gtriparamEM_P(V,V) 
            &\xrightarrow{\theta \rightarrow 0} g_I^{\alphabeta\text{-EM}}\left(\log_{*,P}(V),\log_{*,P}(V) \right)\\
            &= \langle \log_{*,P}(V), \log_{*,P}(V) \rangle^{\alphabeta} \\
            &= g_P^{\alphabeta\text{-LEM}}\left(V,V \right).
        \end{aligned}
    \end{equation}
\end{proof}

\subsection[Proof of \cref{rmlr:thm:spdmlrs}]{Proof of \cref{rmlr:thm:spdmlrs}}
\linkofproof{rmlr:thm:spdmlrs}

As the five families of metrics presented in \cref{rmlr:thm:spdmlrs} are pullback metrics, we first present a general result regarding Riemannian MLRs under pullback metrics.

\begin{parislemma}[Riemannian MLRs under pullback metrics] \label{rmlr:lem:rmlr_pullback}
    Suppose $(\calN,g)$ is a Riemannian manifold and $\phi:\calM \rightarrow \calN$ is a diffeomorphism between manifolds.
    The Riemannian MLR by parallel transport, obtained by combining \cref{rmlr:eq:rmlr_final,rmlr:eq:A_by_pt}, on $\calM$ under $\tilde{g}=\phi^* g$ can be obtained using $g$:
    \begin{align}
        p(y=k \mid S \in \calM)
        &\propto \exp \left[ \langle \tilde{\rielog}_{P_k} S,  \tilde{\Gamma}_{Q \rightarrow P_k} A_k \rangle_{P_k} \right]\\
        \label{rmlr:eq:rmlr_pm_pt}
        &= \exp \left [\langle \rielog_{\phi(P_k)} \phi(S), \tilde{A}_k  \rangle_{\phi(P_k)} \right ],
    \end{align}
    where $\tilde{A}_k = \Gamma_{\phi(Q) \rightarrow \phi(P_k)} \phi_{*,Q} (A_k)$ with $A_k \in T_Q\calM$; $\tilde{\rielog}$ and $\tilde{\Gamma}$ are the Riemannian logarithm and parallel transport under $\tilde{g}$, respectively; and $\rielog$ and $\Gamma$ are their counterparts under $g$.
    
    Furthermore, if $\calN$ has a Lie group operation $\odot$, $\calM$ could be endowed with a Lie group structure $\tilde{\odot}$ by $\phi$.
    The Riemannian MLR by left translation, obtained by combining \cref{rmlr:eq:rmlr_final,rmlr:eq:A_by_lt}, on $\calM$ under $\tilde{g}$ and $\tilde{\odot}$ can be calculated using $g$ and $\odot$:
    \begin{align}
        \label{rmlr:eq:rmlr_pm_lt_v0}
        p(y=k \mid S \in \calM)
        &\propto \exp \left[ \left\langle \tilde{\rielog}_{P_k} S,  \left(\tilde{L}_{\tilde{R}_k}\right)_{*,Q} A_k \right\rangle_{P_k} \right]\\
        \label{rmlr:eq:rmlr_pm_lt}
        &= \exp \left[ \langle \rielog_{\phi(P_k)} \phi(S), \tilde{A}_k \rangle_{\phi(P_k)} \right],
    \end{align}
    where $\tilde{A}_k = \left(L_{R_k}\right)_{*,\phi(Q)}\left(\phi_{*,Q}(A_k)\right)$, $\tilde{R}_k=P_k \tilde{\odot} Q_{\tilde{\odot}}^{-1}$, $R_k=\phi(P_k) \odot \phi(Q)^{-1}_{\odot}$, and $\tilde{L}_{P_k \tilde{\odot} Q_{\tilde{\odot}}^{-1}}$ is the left translation under $\tilde{\odot}$.
\end{parislemma}
\begin{proof}
    Before starting, we should point out that since $\phi$ is a diffeomorphism, $\tilde{\odot}$ and $\tilde{g}$ are indeed well defined, and $(\calM,\tilde{g})$ forms a Riemannian manifold and $(\calM,\tilde{\odot})$ forms a Lie group.
    We first focus on the Riemannian MLR by parallel transport:
    \begin{equation}
        \begin{aligned}
            & p(y=k \mid S \in \calM) \\ 
            &\propto \exp\left( \tilde{g}_{P_k} (\tilde{\rielog}_{P_k} S,  \tilde{\Gamma}_{Q \rightarrow P_k} A_k )\right) \\
            &= \exp \left [g_{\phi(P_k)} \left( \phi_{*,P_k} \circ \phi^{-1}_{*,\phi(P_k)} \rielog_{\phi(P_k)} \phi(S), \phi_{*,P_k} \circ \phi^{-1}_{*,\phi(P_k)} \Gamma_{\phi(Q) \rightarrow \phi(P_k)} \phi_{*,Q} (A_k) \right) \right ] \\
            &= \exp \left [g_{\phi(P_k)}(\rielog_{\phi(P_k)} \phi(S), \Gamma_{\phi(Q) \rightarrow \phi(P_k)} \phi_{*,Q} (A_k)) \right ].
        \end{aligned}
    \end{equation}
    
    In the case of the Riemannian MLR by left translation, we first note that
    \begin{equation}
        \tilde{L}_{\tilde{R}_k}= \phi^{-1} \circ L_{\phi(P_k) \odot \phi(Q)^{-1}_{\odot}} \circ \phi.
    \end{equation}
    Therefore, the associated differential is
    \begin{equation} \label{rmlr:eq:diff_l_pm}
        \left(\tilde{L}_{\tilde{R}_k}\right)_{*,Q}
        = \phi^{-1}_{*,\phi(P_k)} \circ \left(L_{R_k}\right)_{*,\phi(Q)} \circ \phi_{*,Q}.
    \end{equation}
    Putting \cref{rmlr:eq:diff_l_pm} into \cref{rmlr:eq:rmlr_pm_lt_v0}, we can obtain the result.
\end{proof}

Now, we apply \cref{rmlr:lem:rmlr_pullback} to derive the expressions for our SPD MLRs presented in \cref{rmlr:thm:spdmlrs}.
In our cases of SPD MLRs, we set $Q=I$.
For simplicity, we will omit the subscript $k$ for $P_k$ and $A_k$.
We will first derive the expressions for SPD MLRs under $(\theta,\alpha,\beta)$-LEM, $\theta$-LCM, $(\theta,\alpha,\beta)$-EM, and $(\theta,\alpha,\beta)$-AIM from \cref{rmlr:eq:rmlr_pm_pt}.
Then we will derive the expression for MLR under $2\theta$-BWM from \cref{rmlr:eq:rmlr_pm_lt}.
According to \cref{rmlr:lem:scale_metric_rie_opt}, the scaled metric $ag$ shares the same Riemannian operators as $g$.
We will use this fact throughout the following proof.

\begin{proof}
    For simplicity, we abbreviate $\phi_\theta$ as $\phi$ during the proof. Note that for $2\theta$-BWM, $\phi$ should be understood as $\phi_{2\theta}$.
    We first show $\phi(I)$ and the differential map $\phi_{*, I}$, which will be frequently required in the following proof:
    \begin{align} 
        \phi(I) &= I, \\
        \label{rmlr:eq:diff_power_diform_map}
        \phi_{ *, I}(A) &= \theta A, \forall A \in T_I\spd{n}.
    \end{align}
    
    Let $\phi:(\spd{n},\tilde{g})\rightarrow(\spd{n},g)$. Then the SPD MLR under $\tilde{g}$ by parallel transport with $Q=I$ is
    \begin{equation} \label{rmlr:eq:rmlr_pm_pt_simplified}
        p(y=k \mid S \in \spd{n})
        \propto \exp \left [g_{\phi(P)}(\rielog_{\phi(P)} \phi(S), \Gamma_{I \rightarrow \phi(P)} \theta A) \right ].
    \end{equation}
    
    Next, we begin to prove the five SPD MLRs one by one.

    \mypara{$(\alpha,\beta)$-LEM.}     As shown in \cref{alem:thm:rethk_lem_lcm}, the standard LEM is the pullback metric from the Euclidean space $\sym{n}$. Similarly, $(\alpha,\beta)$-LEM is also a pullback metric:
    \begin{equation}
        (\spd{n},g^{\biparamLEM}) \stackrel{\log}{\longrightarrow}
        (\sym{n},g^{\alphabeta}).
    \end{equation}
    By \cref{rmlr:eq:rmlr_pm_pt}, we have
    \begin{align}
        p(y=k \mid S \in \spd{n})
        &\propto \exp \left [ \langle \log(S)-\log(P), \log_{*,I} (A) \rangle^{\alphabeta} \right ]\\
        &= \exp \left [ \langle \log(S)-\log(P), A \rangle^{\alphabeta} \right ].
    \end{align}
    
    \mypara{$\theta$-LCM.} Simple computations show that $\theta$-LCM is the scaled pullback metric of the standard Euclidean metric in the Euclidean space of lower triangular matrices $\trilspace{n}$:
    \begin{equation}
        (\spd{n},\theta^2 g^{\paramLCM}) \stackrel{\phi}{\longrightarrow}
        (\spd{n},g^{\LCM}) \stackrel{\chol}{\longrightarrow}
        (\chospace{n},g^{\text{CM}}) \stackrel{\dlog}{\longrightarrow}
        (\trilspace{n},g^{\text{E}}),
    \end{equation}
    where $g^{\text{E}}$ is the standard Frobenius inner product, and $g^{\text{CM}}$ is the Cholesky metric on the Cholesky space $\chospace{n}$ \citep{lin2019riemannian}.
    Denoting $\zeta=\dlog\circ\chol\circ\phi$, we have
    \begin{equation}
        \zeta_{*,I}(A) = \theta \left( \lfloor A \rfloor + \frac{1}{2}\bbD(A) \right), \forall A \in T_I\spd{n}.
    \end{equation}
    Similar to the case of $\triparamLEM$, we have
    \begin{align}
        p(y=k \mid S \in \spd{n}) &\propto \exp \left [ \frac{1}{\theta^2}\langle \zeta(S)-\zeta(P), \zeta_{*,I} A \rangle \right ]\\
        &=\exp\left[\frac{1}{\theta}\left\langle
        \begin{aligned}
        &\lfloor\tilde{K}\rfloor-\lfloor\tilde{L}\rfloor+\left[\dlog(\bbD(\tilde{K}))-\dlog(\bbD(\tilde{L}))\right],\\
        &\lfloor A\rfloor+\frac{1}{2}\bbD(A)
        \end{aligned}
        \right\rangle\right],
    \end{align}
    where $\tilde{K}=\chol(S^\theta)$, $\tilde{L}=\chol(P^\theta)$, $\bbD(\tilde{K})$ is a diagonal matrix with diagonal elements from $\tilde{K}$, and $\lfloor \tilde{K} \rfloor$ is a strictly lower triangular matrix from $\tilde{K}$.
    
    \mypara{$(\theta,\alpha,\beta)$-EM.} Let $\eta=\frac{1}{|\theta|}\phi$.
    A simple computation shows that $\triparamEM$ is the pullback metric of $\biparamEM$:
    \begin{equation}
        (\spd{n},g^{\triparamEM})
        \stackrel{\eta}{\longrightarrow}
        (\spd{n},g^{\biparamEM}).
    \end{equation}
    Besides, we have the following for $\eta$:
    \begin{equation}
        \eta_{*,I}(A) =\sgn(\theta)A, \forall A \in T_I\spd{n}.
    \end{equation}
    
    According to \cref{rmlr:eq:rmlr_pm_pt}, we have
    \begin{align}
        p(y=k \mid S \in \spd{n})
        &\propto \exp \left[ 
        \langle \eta(S)-\eta(P),\sgn(\theta) A \rangle^{\alphabeta} \right]\\
        &= \exp \left[ \frac{1}{\theta} \langle S^\theta-P^\theta, A \rangle^{\alphabeta} \right].
    \end{align}

    \mypara{$(\theta,\alpha,\beta)$-AIM.}  Putting $g^{\biparamAIM}$ into \cref{rmlr:eq:rmlr_pm_pt_simplified}, we have
    \begin{align}
        p(y=k \mid S \in \spd{n})
        &\propto \exp \left[ \frac{1}{\theta^2}
        g^{\biparamAIM}_{\phi(P)} \left( P^{\frac{\theta}{2}}\log\left(P^{-\frac{\theta}{2}} S^{\theta} P^{-\frac{\theta}{2}}\right)P^{\frac{\theta}{2}},P^{\frac{\theta}{2}} \theta A P^{\frac{\theta}{2}} \right) \right]\\
        &= \exp \left[ \frac{1}{\theta} \left\langle \log\left(P^{-\frac{\theta}{2}} S^\theta P^{-\frac{\theta}{2}}\right), A \right\rangle^{\alphabeta} \right].
    \end{align}    

    \mypara{$2\theta$-BWM.}    We first simplify \cref{rmlr:eq:rmlr_pm_lt} for SPD manifolds and then proceed to focus on the case of $g=g^{\BWM}$.
    Denote $\phi:(\spd{n},\tilde{g},\tilde{\odot})\rightarrow(\spd{n},g,\odot)$, where the Lie group operation $\odot$ \citep{thanwerdas2022theoretically} is defined as
   \begin{equation}
        S_1 \odot S_2 = L_1 S_2 L_1^\top, \forall S_1,S_2\in \spd{n}, \text{ with } L_1=\chol(S_1).
    \end{equation}
    Note that $I$ is the identity element of $(\spd{n},\odot)$, and for any $S \in \spd{n}$, the differential map of the left translation $L_S$ under $\odot$ is
    \begin{equation}
        \left(L_S\right)_{*,Q}(V)=LVL^\top, \forall Q \in \spd{n}, \forall V \in T_Q\spd{n}, \text{ with } L=\chol(S).
    \end{equation}

    For the induced Lie group $(\spd{n},\tilde{\odot})$, the left translation $\tilde{L}_{P \tilde{\odot} I_{\tilde{\odot}}^{-1}}$ under $\tilde{\odot}$ is
    \begin{align}
        \tilde{L}_{P \tilde{\odot} I_{\tilde{\odot}}^{-1}}
        &= \phi^{-1} \circ L_{\phi(P)\odot \phi(I)^{-1}_{\odot}} \circ \phi,\\
        &= \phi^{-1} \circ L_{P^{2\theta}} \circ \phi, \qquad \phi(P)\odot \phi(I)^{-1}_{\odot}=P^{2\theta}.
    \end{align}
    The associated differential at $I$ is
    \begin{align}
        \left(\tilde{L}_{P \tilde{\odot} I_{\tilde{\odot}}^{-1}}\right)_{*,I} (A)
        &= \phi^{-1}_{*,\phi(P)} \circ \left(L_{P^{2\theta}}\right)_{*,\phi(I)} \circ \phi_{*,I} (A)\\
        &= 2\theta \phi^{-1}_{*,\phi(P)} (\bar{L} A \bar{L}^\top),
    \end{align}
    where $\bar{L}=\chol(P^{2\theta})$. Then the SPD MLR under $\tilde{g}$ and $\tilde{\odot}$ by left translation is
    \begin{equation}
        p(y=k \mid S \in \spd{n})
        \propto \exp \left[ 2\theta g_{\phi(P)}\left(\rielog_{\phi(P)} \phi(S), \bar{L} A \bar{L}^{\top} \right) \right].
    \end{equation}

    Setting $g=g^{\BWM}$ (we omit the scaling factor), we obtain the SPD MLR under $2\theta$-BWM:
    \begin{align}
        p(y=k \mid S \in \spd{n})
        &\propto \exp \left [ 2\theta \cdot \frac{1}{4\theta^2}  g_{\phi(P)}^{\BWM} \left(\rielog_{\phi(P)}^{\BWM}\phi(S), \bar{L} A \bar{L}^{\top} \right) \right]\\
        &= \exp \left[ \frac{1}{4\theta}\langle (P^{2\theta}S^{2\theta})^{\frac{1}{2}} + (S^{2\theta}P^{2\theta})^{\frac{1}{2}} -2P^{2\theta}, \calL_{P^{2\theta}}[\bar{L} A \bar{L}^\top] \rangle \right].
    \end{align}
\end{proof}

\subsection[Proof of \cref{rmlr:lem:equi_lie_mlr}]{Proof of \cref{rmlr:lem:equi_lie_mlr}}
\linkofproof{rmlr:lem:equi_lie_mlr}

\begin{proof}
    During this proof, we use the ambient representation of tangent vectors.
    Given rotation matrices $P,Q\in\so{n}$ and a tangent vector $H\in T_Q\so{n}$, let $c(t)$ be a curve on $\so{n}$ satisfying $c(0)=Q$ and $c'(0)=H$.
    The differential of the left translation $L_{PQ^{-1}}$ at $Q$ is
    \begin{equation}
        \left(L_{PQ^{-1}}\right)_{*,Q}(H)
        = \left.\frac{dPQ^{-1}c(t)}{dt}\right|_{t=0}
        = PQ^{-1}H
        = PQ^{\top}H,
    \end{equation}
    which is precisely the vector transport $\vt{Q}{P}(H)$ in \citet[Tab.~1]{boumal2011discrete}.
\end{proof}

\subsection[Proof of \cref{rmlr:thm:lie_mlr}]{Proof of \cref{rmlr:thm:lie_mlr}}
\linkofproof{rmlr:thm:lie_mlr}

\begin{proof}
    Setting $Q=I$ in \cref{rmlr:eq:A_by_lt} and using \cref{rmlr:lem:equi_lie_mlr} give $\tilde{A}_k=P_kA_k$.
    Combining this identity with the logarithmic map and metric in \cref{tab:ch2-so-operators}, we obtain
    \begin{equation}
        \left\langle\rielog_{P_k}R,\tilde{A}_k\right\rangle_{P_k}
        = \left\langle P_k\log\left(P_k^{\top}R\right),P_kA_k\right\rangle
        = \left\langle\log\left(P_k^{\top}R\right),A_k\right\rangle.
    \end{equation}
    Substituting this expression into \cref{rmlr:eq:rmlr_final} yields the result.
\end{proof}

\section{Proper Velocity Neural Networks}
\label{app:pvnn-proofs}

\subsection{Derivation of the Proper Velocity Metric}
\label{pvnn:app:pv-metrics}
The PV line element at $x \in \PVspace{n}$ can be written in terms of the curvature parameter $K<0$ as
\begin{equation}
Q_x(u) = \norm{u}^2 + K \beta_x^2 \inner{x}{u}^2, \quad \forall u\in T_x\PVspace{n}\simeq\bbR{n},
\end{equation}
where $\beta_x = \frac{1}{\sqrt{1-K\norm{x}^2}}$. This is equivalent to the expression in \citet[Eq.~(7.76)]{ungar2022analytic} after substituting $s^2=-1/K$.
Given $u,v \in T_x\PVspace{n}$, the bilinear form $g_x(u,v)$ is obtained by the polarization identity:
\begin{equation}
g_x(u,v) = \tfrac{1}{4}\left(Q_x(u+v) - Q_x(u-v)\right).
\end{equation}

We first expand the two terms in the polarization identity:
\begin{equation}
\begin{aligned}
Q_x(u+v)
&= \norm{u+v}^2 + K \beta_x^2 \inner{x}{u+v}^2 \\
&= \norm{u}^2 + 2\inner{u}{v} + \norm{v}^2
   + K \beta_x^2 \left(\inner{x}{u}^2 + 2\inner{x}{u}\inner{x}{v} + \inner{x}{v}^2\right), \\
Q_x(u-v)
&= \norm{u-v}^2 + K \beta_x^2 \inner{x}{u-v}^2 \\
&= \norm{u}^2 - 2\inner{u}{v} + \norm{v}^2
   + K \beta_x^2 \left(\inner{x}{u}^2 - 2\inner{x}{u}\inner{x}{v} + \inner{x}{v}^2\right).
\end{aligned}
\end{equation}
Taking the difference yields
\begin{equation}
\begin{aligned}
Q_x(u+v) - Q_x(u-v)
&= 4\inner{u}{v}
   + 4K \beta_x^2 \inner{x}{u}\inner{x}{v}.
\end{aligned}
\end{equation}
Substituting this expression into the polarization identity, we obtain
\begin{equation}
\begin{aligned}
g_x(u,v)
&= \tfrac{1}{4}\left(Q_x(u+v) - Q_x(u-v)\right) \\
&= \inner{u}{v} + K \beta_x^2 \inner{x}{u}\inner{x}{v},
\end{aligned}
\end{equation}
which coincides with the expression of the PV metric in \cref{pvnn:eq:pv-metric}.

\subsection[Proof of \cref{pvnn:lem:pv-poincare-diffs}]{Proof of \cref{pvnn:lem:pv-poincare-diffs}}
\linkofproof{pvnn:lem:pv-poincare-diffs}
\begin{proof}

\mypara{Differential of $\PVtoPB$.} Consider the curve $c:(-\varepsilon,\varepsilon)\to\PVspace{n}$ which satisfies $c(0)=x$ and $c'(0)=v$. By definition of the differential,
\begin{equation}
d_x\PVtoPB(v)=\left.\frac{\mathrm{d}}{\mathrm{d}t}\right|_{t=0}\PVtoPB\left(c(t)\right).
\end{equation}
Using $\PVtoPB(x)=\frac{\beta_x}{1+\beta_x}x$ with $\beta_x=\frac{1}{\sqrt{1-K\norm{x}^2}}$, we write
\begin{equation}
\PVtoPB\left(c(t)\right) = h(t)c(t),
\qquad
h(t):=\frac{\beta_{c(t)}}{1+\beta_{c(t)}}.
\end{equation}
Let $r(t)=\norm{c(t)}^2$, so that $\beta_{c(t)}=(1-Kr(t))^{-1/2}$. Then
\begin{equation}
r'(0)=2\inner{x}{v},
\qquad
\beta'_{c(0)}
=\tfrac12(1-Kr(0))^{-3/2}K r'(0)
=K\beta_x^3\inner{x}{v}.
\end{equation}
Differentiating $h(t)$ at $t=0$ gives
\begin{equation}
h'(0)
=\frac{\beta'_{c(0)}}{(1+\beta_x)^2}
=K\frac{\beta_x^3}{(1+\beta_x)^2}\inner{x}{v}.
\end{equation}
Finally, differentiating $h(t)c(t)$ at $t=0$ yields
\begin{equation}
\begin{aligned}
d_x\PVtoPB(v)
&= h'(0)x + h(0)v \\
&= K\frac{\beta_x^{3}}{(1+\beta_x)^2}\inner{x}{v}x
 + \frac{\beta_x}{1+\beta_x}v.
\end{aligned}
\end{equation}

In particular, at $x=\zerovec$ one has $\beta_{\zerovec}=1$ and $\inner{x}{v}=0$. Thus, we have
\begin{equation}
d_{\zerovec}\PVtoPB(v)
 =\frac{\beta_{\zerovec}}{1+\beta_{\zerovec}}v
=\frac{1}{2}v.
\end{equation}

\mypara{Differential of $\PBtoPV$.} Consider the curve $c:(-\varepsilon,\varepsilon)\to\pball{n}$ which satisfies $c(0)=y$ and $c'(0)=w$. By definition of the differential,
\begin{equation}
d_y\PBtoPV(w)
= \left.\frac{\mathrm{d}}{\mathrm{d}t}\right|_{t=0} \PBtoPV(c(t)).
\end{equation}
Using the explicit expression $\PBtoPV(y)=2\gamma_y^2 y$ with $\gamma_y=\frac{1}{\sqrt{1+K\norm{y}^2}}$, we obtain
\begin{equation}
\PBtoPV(c(t))=2\gamma_{c(t)}^2 c(t).
\end{equation}
Let $r(t)=\norm{c(t)}^2$ so that $\gamma_{c(t)}^2=(1+K r(t))^{-1}$. Then
\begin{equation}
r'(0)=2\inner{y}{w},
\qquad
\left.\frac{\mathrm{d}}{\mathrm{d}t}\right|_{t=0}\gamma_{c(t)}^2
= -\frac{K r'(0)}{(1+K r(0))^{2}}
= -2K\gamma_y^4\inner{y}{w}.
\end{equation}
Differentiating $2\gamma_{c(t)}^2 c(t)$ at $t=0$ yields
\begin{equation}
\begin{aligned}
 d_y\PBtoPV(w)
 &= 2\left.\frac{\mathrm{d}}{\mathrm{d}t}\right|_{t=0}\gamma_{c(t)}^2 y
  + 2\gamma_y^2 w \\
 &= -4K\gamma_y^4\inner{y}{w}y + 2\gamma_y^2 w.
\end{aligned}
\end{equation}
In particular, at $y=\zerovec$ we have $\gamma_{\zerovec}=1$ and $\inner{y}{w}=0$. Thus, we have
\begin{equation}
 d_{\zerovec}\PBtoPV(w)=2w.
\end{equation}

\end{proof}

\subsection[Proof of \cref{pvnn:thm:pv-poincare-isometry}]{Proof of \cref{pvnn:thm:pv-poincare-isometry}}
\linkofproof{pvnn:thm:pv-poincare-isometry}

\begin{proof}
It suffices to show that for any $x \in \PVspace{n}$ and $v,w \in T_x\PVspace{n}$,
\begin{equation}
g_{y}^\PB\left(d_x\PVtoPB(v), d_x\PVtoPB(w)\right) = g_x^\PV(v,w),
\end{equation}
where $y = \PVtoPB(x)$.

We first recall the following equations from \cref{pvnn:eq:pv-metric,sec:ch2-constant-curvature-manifolds,pvnn:lem:pv-poincare-diffs}:
\begin{equation}
    \begin{aligned}
        g_x^{\PV}(v,w) &= \inner{v}{w} + K \beta_x^2 \inner{x}{v} \inner{x}{w}, \quad \forall x \in \PVspace{n}, \forall v,w \in T_x\PVspace{n}, \\
        g_y^{\PB}(u,z) &= \left(\lambda_y^{K}\right)^2 \inner{u}{z}, \quad \forall y \in \pball{n}, \forall u,z \in T_y\pball{n}, \\
        \PVtoPB(x) &= \frac{\beta_x}{1+\beta_x}x, \quad \forall x \in \PVspace{n}, \\
        d_x\PVtoPB(v) &= \frac{\beta_x}{1+\beta_x}v  + K \frac{\beta_x^{3}}{(1+\beta_x)^2}\inner{x}{v}x, \quad \forall x \in \PVspace{n}, \forall v \in T_x\PVspace{n}.
    \end{aligned}
\end{equation}

Let
\begin{equation}
a = \frac{\beta_x}{1+\beta_x},
\qquad
b = K \frac{\beta_x^{3}}{(1+\beta_x)^2}.
\end{equation}
Then $d_x\PVtoPB(v) = a v + b \inner{x}{v}x$ and $d_x\PVtoPB(w) = a w + b \inner{x}{w}x$. Thus,
\begin{equation}
\begin{aligned}
& g_y^{\PB}\left(d_x\PVtoPB(v), d_x\PVtoPB(w)\right) \\
&= \left(\lambda_y^{K}\right)^2 \inner{a v + b \inner{x}{v}x}{a w + b \inner{x}{w}x} \\
&= \left(\lambda_y^{K}\right)^2 \left(a^2 \inner{v}{w}
  + ab \inner{x}{w}\inner{v}{x}
  + ab \inner{x}{v}\inner{x}{w}
  + b^2 \inner{x}{v}\inner{x}{w}\inner{x}{x}\right) \\
&= \left(\lambda_y^{K}\right)^2 \left(a^2 \inner{v}{w}
  + \left(2ab + b^2\norm{x}^2\right)\inner{x}{v}\inner{x}{w}\right).
\end{aligned}
\end{equation}

Using $y=\PVtoPB(x)$ and the relation between $\lambda_y^K$, $\beta_x$, and $\norm{x}$ from \cref{sec:ch2-constant-curvature-manifolds}, we simplify the coefficients. First,
\begin{equation}
\begin{aligned}
K\norm{y}^2
& = K\inner{\frac{\beta_x}{1+\beta_x}x}{\frac{\beta_x}{1+\beta_x}x} \\
&= K\frac{\beta_x^2}{(1+\beta_x)^2}\norm{x}^2 \\
&= \frac{\beta_x^2-1}{(1+\beta_x)^2},
\end{aligned}
\end{equation}
where we use $\beta_x^2 = \frac{1}{1-K\norm{x}^2}$. Hence
\begin{equation}
1+K\norm{y}^2
= 1 + \frac{\beta_x^2-1}{(1+\beta_x)^2}
= \frac{2\beta_x}{1+\beta_x},
\end{equation}
which implies $\lambda_y^K = \frac{2}{1+K\norm{y}^2} = \frac{1+\beta_x}{\beta_x}$. Therefore
\begin{equation}
\left(\lambda_y^{K}\right)^2 a^2
= \left(\frac{1+\beta_x}{\beta_x}\right)^2\left(\frac{\beta_x}{1+\beta_x}\right)^2
= 1.
\end{equation}
Next, we compute
\begin{equation}
\begin{aligned}
2ab
&= 2\frac{\beta_x}{1+\beta_x} K \frac{\beta_x^{3}}{(1+\beta_x)^2}
 = \frac{2K\beta_x^{4}}{(1+\beta_x)^3}, \\
b^2\norm{x}^2
&= K^2\frac{\beta_x^{6}}{(1+\beta_x)^4}\norm{x}^2
 = K\frac{\beta_x^{4}(\beta_x^2-1)}{(1+\beta_x)^4},
\end{aligned}
\end{equation}
which brings us to
\begin{equation}
\begin{aligned}
2ab + b^2\norm{x}^2
&= K\frac{\beta_x^{4}}{(1+\beta_x)^4}\left(2(1+\beta_x) + \beta_x^2-1\right) \\
&= K\frac{\beta_x^{4}(\beta_x+1)^2}{(1+\beta_x)^4}
 = K\frac{\beta_x^{4}}{(1+\beta_x)^2}.
\end{aligned}
\end{equation}
Multiplying by $\left(\lambda_y^K\right)^2 = \left(\frac{1+\beta_x}{\beta_x}\right)^2$ yields
\begin{equation}
\left(\lambda_y^{K}\right)^2\left(2ab + b^2\norm{x}^2\right)
 = \left(\frac{1+\beta_x}{\beta_x}\right)^2 K\frac{\beta_x^{4}}{(1+\beta_x)^2}
 = K\beta_x^2.
\end{equation}
Substituting these identities into the expression for $g_\PB$ gives
\begin{equation}
g_y^{\PB}\left(d_x\PVtoPB(v), d_x\PVtoPB(w)\right)
= \inner{v}{w} + K \beta_x^2 \inner{x}{v}\inner{x}{w} = g_x^{\PV}(v,w).
\end{equation}
\end{proof}

\subsection[Proof of \cref{pvnn:thm:pv-exp-log-pt-at-x}]{Proof of \cref{pvnn:thm:pv-exp-log-pt-at-x}}
\label{pvnn:app:pv-exp-log-pt-deriv}
\linkofproof{pvnn:thm:pv-exp-log-pt-at-x}

Following the notation in the main theorem, we further denote:
\begin{equation}
\bar{x} = \pi(x) \in \pball{n},
\quad
\bar{y} = \pi(y) \in \pball{n},
\quad
\bar{v} = d_x\pi(v) \in T_{\bar{x}}\pball{n},
\end{equation}
Recalling \cref{pvnn:eq:pv-poincare-isos,pvnn:lem:pv-poincare-diffs}, we have the following:
\begin{align}
    \pi(x) &= \frac{\beta_x}{1+\beta_x}x, \quad \forall x \in \PVspace{n}, \\
    \quad
    \pi^{-1}(\bar{y}) &= 2 \gamma_{\bar{y}}^2 \bar{y}, \quad \forall \bar{y} \in \pball{n}, \\
    d_x\pi(v) 
    &= K \frac{\beta_x^{3}}{(1+\beta_x)^2}\inner{x}{v}x
    + \frac{\beta_x}{1+\beta_x}v, \quad \forall x \in \PVspace{n}, \forall v \in T_x \PVspace{n}, \\
    d_{\bar{y}}\pi^{-1}(w) 
    &= -4K\gamma_{\bar{y}}^4\inner{\bar{y}}{w}\bar{y} + 2\gamma_{\bar{y}}^2 w, \quad \forall \bar{y} \in \pball{n}, \forall w \in T_{\bar{y}}\pball{n}.
\end{align}

Next, we derive the expressions for each PV operator.

\subsubsection{PV Exponential Map}

We recall from \cref{tab:ch2-constant-curvature-operators} that the Riemannian exponential on the Poincaré ball is
\begin{equation}
\label{pvnn:eq:pb-exp}
\rieexp^{\PB}_{\bar{x}}(\bar{v})
=
\bar{x} \Moplus
\left(
\frac{1}{\sqrt{-K}}
\tanh\left(
\frac{\sqrt{-K}\lambda_{\bar{x}}^{K}\norm{\bar{v}}}{2}
\right)
\frac{\bar{v}}{\norm{\bar{v}}}
\right).
\end{equation}
By the Riemannian isometry and the gyrovector isomorphism of $\pi$, for any $x \in \PVspace{n}$ and $v \in T_x\PVspace{n}$ we have
\begin{equation}
\begin{aligned}
\rieexp_{x}(v) 
&\stackrel{(1)}{=}
\pi^{-1}\left(
\rieexp^{\PB}_{\bar{x}}(\bar{v})
\right) \\
&\stackrel{(2)}{=}
x \PVoplus
\pi^{-1} \left(
\frac{1}{\sqrt{-K}}
\tanh\left(
\frac{\sqrt{-K}\lambda_{\bar{x}}^{K}\norm{\bar{v}}}{2}
\right)
\frac{\bar{v}}{\norm{\bar{v}}}
\right),
\end{aligned}
\end{equation}
The above equalities follow from the following facts.
\begin{enumerate}
    \item 
    Isometry.
    \item 
    Gyrovector isomorphism.
\end{enumerate}

Let
\begin{equation}
\label{pvnn:eq:pb-exp-increment}
u
=
\frac{1}{\sqrt{-K}}
\tanh\left(
\frac{\sqrt{-K}\lambda_{\bar{x}}^{K}\norm{\bar{v}}}{2}
\right)
\frac{\bar{v}}{\norm{\bar{v}}}.
\end{equation}
We have
\begin{equation}
\norm{u} =
\frac{1}{\sqrt{-K}}
\tanh\left(
\frac{\sqrt{-K}\lambda_{\bar{x}}^{K}\norm{\bar{v}}}{2}
\right).
\end{equation}
Let $t = \frac{\sqrt{-K}\lambda_{\bar{x}}^{K}\norm{\bar{v}}}{2}$ so that $\sqrt{-K}\norm{u} = \tanh(t)$. Then
\begin{equation}
\begin{aligned}
\pi^{-1}(u)
&=
\frac{2}{1+K\norm{u}^2}u \\
&=
\frac{2}{1+K\norm{u}^2}
\left(
\frac{1}{\sqrt{-K}}
\tanh(t)
\frac{\bar{v}}{\norm{\bar{v}}}
\right)
\\
&=
\frac{2}{1-\tanh^2(t)}
\frac{1}{\sqrt{-K}}
\tanh(t)
\frac{\bar{v}}{\norm{\bar{v}}}
\\
&=
\frac{2}{\sqrt{-K}}
\frac{\tanh(t)}{1-\tanh^2(t)}
\frac{\bar{v}}{\norm{\bar{v}}}
\\
&\stackrel{(1)}{=}
\frac{2}{\sqrt{-K}}
\tanh(t)\cosh^2(t)
\frac{\bar{v}}{\norm{\bar{v}}}
\\
&=
\frac{2}{\sqrt{-K}}
\frac{\sinh(t)}{\cosh(t)}\cosh^2(t)
\frac{\bar{v}}{\norm{\bar{v}}}
\\
&=
\frac{2}{\sqrt{-K}}
\sinh(t)\cosh(t)
\frac{\bar{v}}{\norm{\bar{v}}}
\\
&\stackrel{(2)}{=}
\frac{1}{\sqrt{-K}}
\left(2\sinh(t)\cosh(t)\right)
\frac{\bar{v}}{\norm{\bar{v}}}
\\
&=
\frac{1}{\sqrt{-K}}
\sinh\left(2t\right)
\frac{\bar{v}}{\norm{\bar{v}}} \\
&=
\frac{1}{\sqrt{-K}}
\sinh\left(
\sqrt{-K}\lambda_{\bar{x}}^{K}\norm{\bar{v}}
\right)
\frac{\bar{v}}{\norm{\bar{v}}} \\
&\stackrel{(3)}{=}
\frac{1}{\sqrt{-K}}
\sinh\left(
\frac{\sqrt{-K} (1+\beta_x)}{\beta_x} \norm{\bar{v}}
\right)
\frac{\bar{v}}{\norm{\bar{v}}}.
\end{aligned}
\end{equation}
The above equalities use:
\begin{enumerate}
    \item 
    $1-\tanh^2(t)=1/\cosh^2(t)$;
    \item 
    $\sinh(2t)=2\sinh(t)\cosh(t)$;
    \item 
    $\lambda_{\bar{x}}^{K} = \frac{1+\beta_x}{\beta_x}$.
\end{enumerate}

\subsubsection{PV Logarithmic Map}

We recall from \cref{tab:ch2-constant-curvature-operators} that the Riemannian logarithm on the Poincaré ball is
\begin{equation}
\label{pvnn:eq:pb-log}
\rielog^{\PB}_{\bar{x}}(\bar{y})
=
\frac{2}{\sqrt{-K}\lambda_{\bar{x}}^{K}}
\frac{\tanh^{-1}\left(\sqrt{-K}\norm{\bar{z}}\right)}{\norm{\bar{z}}}
\bar{z},
\qquad
\bar{z} = (-\bar{x}) \Moplus \bar{y},
\end{equation}
where $\lambda_{\bar{x}}^{K} = \frac{2}{1+K\norm{\bar{x}}^2}$. We define
\begin{equation}
z = (-x)\PVoplus y, \quad
\bar{z} = \pi(z).
\end{equation}
By the Riemannian isometry of $\pi$, we have
\begin{equation}
\begin{aligned}
\rielog_{x}(y)
&=
d_{\bar{x}}\pi^{-1}\left(\rielog^{\PB}_{\bar{x}}(\bar{y})\right)\\
&=
d_{\bar{x}}\pi^{-1}
\left(
\frac{2}{\sqrt{-K}\lambda_{\bar{x}}^{K}}
\frac{\tanh^{-1}\left(\sqrt{-K}\norm{\bar{z}}\right)}{\norm{\bar{z}}}
\bar{z}
\right) \\
&=
\alpha(x,y) d_{\bar{x}}\pi^{-1}\left(\bar{z}\right),
\end{aligned}
\end{equation}
where
\begin{equation}
\label{pvnn:eq:pv-log-alpha-def}
\alpha(x,y)
=
\frac{2}{\sqrt{-K}\lambda_{\bar{x}}^{K}}
\frac{\tanh^{-1}\left(\sqrt{-K}\norm{\bar{z}}\right)}{\norm{\bar{z}}}.
\end{equation}

The differential of $\pi^{-1}$ at $\bar{x}=\pi(x)$ is
\begin{equation}
d_{\bar{x}}\pi^{-1}(h) = -4K\gamma_{\bar{x}}^4 \inner{\bar{x}}{h}\bar{x} + 2\gamma_{\bar{x}}^2 h, \quad \forall h \in T_{\bar{x}}\pball{n},
\end{equation}
where $\gamma_{\bar{x}} = \tfrac{1}{\sqrt{1+K\norm{\bar{x}}^2}}$. Using $\bar{x} = \tfrac{\beta_x}{1+\beta_x}x$ and the relation $1-K\norm{x}^2 = \tfrac{1}{\beta_x^2}$, we have
\begin{equation}
\begin{aligned}
K\norm{\bar{x}}^2
&=
K\frac{\beta_x^2}{(1+\beta_x)^2}\norm{x}^2
=
\frac{\beta_x^2}{(1+\beta_x)^2}\left(\frac{\beta_x^2-1}{\beta_x^2}\right)
=
\frac{\beta_x^2-1}{(1+\beta_x)^2}, \\
1+K\norm{\bar{x}}^2
&=
1+\frac{\beta_x^2-1}{(1+\beta_x)^2}
=
\frac{2\beta_x}{1+\beta_x}.
\end{aligned}
\end{equation}
Hence
\begin{equation}
\gamma_{\bar{x}}^2
=
\frac{1}{1+K\norm{\bar{x}}^2}
=
\frac{1+\beta_x}{2\beta_x},
\qquad
\gamma_{\bar{x}}^4
=
\left(\frac{1+\beta_x}{2\beta_x}\right)^2
=
\frac{(1+\beta_x)^2}{4\beta_x^2}.
\end{equation}
Substituting these into $d_{\bar{x}}\pi^{-1}(h)$ yields
\begin{equation}
\label{pvnn:eq:pv-log-dpi-inv}
\begin{aligned}
d_{\bar{x}}\pi^{-1}(h)
&=
-4K\frac{(1+\beta_x)^2}{4\beta_x^2}\inner{\bar{x}}{h}\bar{x} 
+ 2\frac{1+\beta_x}{2\beta_x}h \\
&=
-K\frac{(1+\beta_x)^2}{\beta_x^2}\inner{\bar{x}}{h}\bar{x}
+\frac{1+\beta_x}{\beta_x}h \\
&=
\frac{1+\beta_x}{\beta_x} h
- K \inner{x}{h}x,
\end{aligned}
\end{equation}
where the last equality uses that $\bar{x} = \tfrac{\beta_x}{1+\beta_x}x$.
Applying \cref{pvnn:eq:pv-log-dpi-inv} with $h = \bar{z}$ and using that $\bar{z}=\pi(z)$ is collinear with $z = (-x)\PVoplus y$, we obtain
\begin{equation}
\label{pvnn:eq:pv-log-sigma-tau-form}
\begin{aligned}
\rielog_{x}(y)
&=
\alpha(x,y) (d_x\pi)^{-1}(\bar{z}) \\
&=
\alpha(x,y)\left(
\frac{1+\beta_x}{\beta_x}\bar{z}
- K \inner{x}{\bar{z}}x
\right).
\end{aligned}
\end{equation}
Since $\bar{z}=\pi(z)$ and $\pi$ is given by \cref{pvnn:eq:pv-poincare-isos}, $z$ and $\bar{z}$ are collinear and
\begin{equation}
\bar{z}
=
\rho z,
\qquad
\rho
=
\frac{\beta_z}{1+\beta_z},
\end{equation}
which also implies $\inner{x}{\bar{z}} = \rho \inner{x}{z}$. Substituting these into \cref{pvnn:eq:pv-log-sigma-tau-form} yields
\begin{equation}
\begin{aligned}
\rielog_{x}(y)
&=
\alpha(x,y)\left(
\frac{1+\beta_x}{\beta_x}\rho z
- K \rho \inner{x}{z}x
\right) \\
&=
\underbrace{\alpha(x,y)\frac{1+\beta_x}{\beta_x}\rho}_{\displaystyle \sigma(x,y)} z
+\underbrace{\left(-K \alpha(x,y)\rho\right)}_{\displaystyle \tau(x,y)} \inner{x}{z}x.
\end{aligned}
\end{equation}
Using the definition of $\alpha(x,y)$ in \cref{pvnn:eq:pv-log-alpha-def} together with $\lambda_{\bar{x}}^{K} = \tfrac{1+\beta_x}{\beta_x}$ and $\rho=\tfrac{\beta_z}{1+\beta_z}$, a straightforward simplification yields
\begin{equation}
\label{pvnn:eq:pv-log-sigma-tau}
\sigma(x,y)
=
\frac{2}{\sqrt{-K}}
\frac{\tanh^{-1}\left(\sqrt{-K}\norm{\bar{z}}\right)}{\norm{z}},
\quad
\tau(x,y)
=
\frac{2\beta_x}{1+\beta_x}
\frac{\sqrt{-K}\tanh^{-1}\left(\sqrt{-K}\norm{\bar{z}}\right)}{\norm{z}}.
\end{equation}
Thus,
\begin{equation}
\rielog_{x}(y)
=
\sigma(x,y) z
+\tau(x,y)\inner{x}{z}x.
\end{equation}

\subsubsection{PV Parallel Transport}

We recall from \cref{tab:ch2-constant-curvature-operators} that the parallel transport on the Poincaré ball is
\begin{equation}
\label{pvnn:eq:pb-pt}
\pt{\bar{x}}{\bar{y}}^{\PB}(w)
=
\frac{\lambda_{\bar{x}}^{K}}{\lambda_{\bar{y}}^{K}}
\Mgyr[\bar{y},-\bar{x}](w), \quad \text{with } w \in T_{\bar{x}}\pball{n}.
\end{equation}
We have
\begin{equation}
\label{pvnn:eq:pv-pt-explicit-step}
\begin{aligned}
&\pt{x}{y}(v) \\
&\stackrel{(1)}{=}
d_{\bar{y}}\pi^{-1}\left(
\pt{\bar{x}}{\bar{y}}^{\PB}\left(d_x\pi(v)\right)
\right)
\\
&=
d_{\bar{y}}\pi^{-1}\left(
\frac{\lambda_{\bar{x}}^{K}}{\lambda_{\bar{y}}^{K}}
\Mgyr[\bar{y},-\bar{x}]\left(d_x\pi(v)\right)
\right)
\\
&\stackrel{(2)}{=}
\frac{\lambda_{\bar{x}}^{K}}{\lambda_{\bar{y}}^{K}}
d_{\bar{y}}\pi^{-1}\left(
\Mgyr[\bar{y},-\bar{x}]\left(d_x\pi(v)\right) \right) \\
&\stackrel{(3)}{=}
\frac{\lambda_{\bar{x}}^{K}}{\lambda_{\bar{y}}^{K}}
\left(
\frac{1+\beta_y}{\beta_y}\Mgyr[\bar{y},-\bar{x}]\left(d_x\pi(v)\right)- K \inner{y}{\Mgyr[\bar{y},-\bar{x}]\left(d_x\pi(v)\right)}y\right) \\
&\stackrel{(4)}{=}
\frac{(1+\beta_x)\beta_y}{(1+\beta_y)\beta_x}
\left(
\frac{1+\beta_y}{\beta_y}\Mgyr[\bar{y},-\bar{x}]\left(d_x\pi(v)\right)- K \inner{y}{\Mgyr[\bar{y},-\bar{x}]\left(d_x\pi(v)\right)}y\right)\\
&=
\frac{1+\beta_x}{\beta_x}\Mgyr[\bar{y}, -\bar{x}]\left(d_x\pi(v)\right)
- K \frac{(1+\beta_x)\beta_y}{(1+\beta_y)\beta_x} \inner{y}{\Mgyr[\bar{y},-\bar{x}]\left(d_x\pi(v)\right)}y.
\end{aligned}
\end{equation}
The above equalities use:
\begin{enumerate}
    \item the isometry property of $\pi$;
    \item linearity of $d_{\bar{y}}\pi^{-1}$;
    \item \cref{pvnn:eq:pv-log-dpi-inv}.
    \item Using the relation between $\lambda_{\bar{x}}^{K}$ and $\beta_x$ in the proof of \cref{pvnn:thm:pv-poincare-isometry},
    \begin{equation}
        \lambda_{\bar{x}}^{K} = \frac{1+\beta_x}{\beta_x}
        \qquad
        \lambda_{\bar{y}}^{K} = \frac{1+\beta_y}{\beta_y}. \\
    \end{equation}
\end{enumerate}

\subsubsection{PV Geodesic Distance}

We recall from \cref{tab:ch2-constant-curvature-operators} that the geodesic distance on the Poincaré ball $\pball{n}$ is
\begin{equation}
\label{pvnn:eq:pb-dist-add}
\dist^{\PB}(y_1,y_2)
=
\frac{2}{\sqrt{-K}}
\tanh^{-1}\left(
\sqrt{-K}\norm{(-y_1)\Moplus y_2}
\right),
\quad
y_1,y_2\in\pball{n}.
\end{equation}

By isometry and isomorphism, the PV geodesic distance is
\begin{equation}
\begin{aligned}
\dist(x,y)
&= \dist^{\PB}\left(\pi(x),\pi(y)\right) \\
&= 
\frac{2}{\sqrt{-K}}
\tanh^{-1}\left(
\sqrt{-K}\norm{(-\pi(x))\Moplus \pi(y)}
\right)\\
&=
\frac{2}{\sqrt{-K}}
\tanh^{-1}\left(
\sqrt{-K}\norm{\pi(-x\PVoplus y)}
\right).
\end{aligned}
\end{equation}

\subsubsection{Special Cases at the Identity}

\mypara{Exponential Map at the Identity.}
\begin{equation}
\begin{aligned}
\rieexp_{\zerovec}(v)
&\stackrel{(1)}{=}
\frac{1}{\sqrt{-K}}
\sinh\left(
\frac{\sqrt{-K} (1+\beta_{\zerovec})}{\beta_{\zerovec}} \norm{\bar{v}}
\right)
\frac{\bar{v}}{\norm{\bar{v}}} \\
&\stackrel{(2)}{=}
\frac{1}{\sqrt{-K}}
\sinh\left(\sqrt{-K}\norm{v}\right)\frac{v}{\norm{v}}.
\end{aligned}
\end{equation}
The above comes from the following.
\begin{enumerate}
    \item $\zerovec$ is the gyro identity;
    \item $\beta_\zerovec=1$ and $d_{\zerovec}\pi(v)=\tfrac{1}{2}v$.
\end{enumerate}

\mypara{Logarithmic Map at the Identity.} As $z=-\zerovec \PVoplus y=y$, we have
\begin{equation}
\begin{aligned}
    \rielog_{\zerovec}(y)
    &=
    \sigma(\zerovec,y) z
    +\tau(\zerovec,y)\inner{\zerovec}{z}\zerovec \\
    &=
    \sigma(\zerovec,y) y \\
    &=
    \frac{2}{\sqrt{-K}}
    \frac{\tanh^{-1}\left(\sqrt{-K}\norm{\pi(y)}\right)}{\norm{y}} y. \\
\end{aligned}
\end{equation}

From $\pi(y)=\tfrac{\beta_y}{1+\beta_y}y$ and $\beta_y=\tfrac{1}{\sqrt{1-K\norm{y}^2}}$, we obtain
\begin{equation}
\sqrt{-K}\norm{\pi(y)}
=
\frac{\beta_y}{1+\beta_y}\sqrt{-K}\norm{y}.
\end{equation}
Let $t=\sqrt{-K}\norm{y}$ and $s=\sqrt{1+t^2}$, so that $\beta_y=\tfrac{1}{\sqrt{1-K\norm{y}^2}}=\tfrac{1}{s}$. Define
\begin{equation}
a
=
\frac{\beta_y}{1+\beta_y}t
=
\frac{t}{s+1}.
\end{equation}
Using the hyperbolic double-angle identity, we have
\begin{equation}
\begin{aligned}
\tanh\left(2\tanh^{-1}(a)\right)
&=
\frac{2a}{1+a^2} \\
&=
\frac{2t/(s+1)}{1+t^2/(s+1)^2} \\
&=
\frac{2t(s+1)}{(s+1)^2+t^2} \\
&=
\frac{2t(s+1)}{s^2+2s+1+t^2} \\
&=
\frac{2t(s+1)}{2(1+t^2+s)} \\
&=
\frac{t(s+1)}{1+t^2+s} \\
&=
\frac{t(s+1)}{s^2+s} \\
&=
\frac{t}{s}
=
\frac{t}{\sqrt{1+t^2}}.
\end{aligned}
\end{equation}
Denoting $u = \sinh^{-1}(t)$, we have
\begin{equation}
\cosh(u)
=
\sqrt{1+\sinh^{2}(u)}
=
\sqrt{1+t^{2}}.
\end{equation}
Therefore,
\begin{equation}
\tanh\left(\sinh^{-1}(t)\right)
=
\tanh(u)
=
\frac{\sinh(u)}{\cosh(u)} 
=
\frac{t}{\sqrt{1+t^{2}}}.
\end{equation}
Since $\tanh$ is strictly increasing on $\bbRscalar$, this implies that
\begin{equation}
2\tanh^{-1}\left(\frac{\beta_y}{1+\beta_y}t\right)
=
\sinh^{-1}(t).
\end{equation}
Substituting this identity back gives
\begin{equation}
\sigma(\zerovec,y)
=
\frac{1}{\sqrt{-K}}
\frac{\sinh^{-1}\left(\sqrt{-K}\norm{y}\right)}{\norm{y}},
\end{equation}
and therefore
\begin{equation}
\rielog_{\zerovec}(y)
=
\frac{1}{\sqrt{-K}}
\sinh^{-1}\left(\sqrt{-K}\norm{y}\right)\frac{y}{\norm{y}}.
\end{equation}

\mypara{Parallel Transport from the Identity.} The gyration satisfies
\begin{equation}
\Mgyr[\zerovec,\bar{y}]=\Mgyr[\bar{y},\zerovec]=\id.
\end{equation}
Substituting this into \cref{pvnn:eq:pv-pt-explicit-step} gives
\begin{equation}
\begin{aligned}
\pt{\zerovec}{y}(v)
&=
\frac{1+\beta_{\zerovec}}{\beta_{\zerovec}} d_{\zerovec}\pi(v)
- K \frac{(1+\beta_{\zerovec})\beta_y}{(1+\beta_y)\beta_{\zerovec}} \inner{y}{d_{\zerovec}\pi(v)}y \\
&=
2\cdot\frac{1}{2}v
- K \frac{2\beta_y}{1+\beta_y}\cdot\frac{1}{2}\inner{y}{v}y \\
&=
v - K \frac{\beta_y}{1+\beta_y}\inner{y}{v}y.
\end{aligned}
\end{equation}
	
\mypara{Parallel Transport to the Identity.} Taking $y=\zerovec$ in \cref{pvnn:eq:pv-pt-explicit-step} and using $\Mgyr[\zerovec,-\bar{x}]=\id$ yields
\begin{equation}
\begin{aligned}
    \pt{x}{\zerovec}(v)
    &=
    \frac{1+\beta_x}{\beta_x}d_x\pi(v) \\
    &=
    \frac{1+\beta_x}{\beta_x}
    \left(
    K\frac{\beta_x^{3}}{(1+\beta_x)^2}\inner{x}{v}x
    + \frac{\beta_x}{1+\beta_x}v
    \right) \\
    &=
    v + K \frac{\beta_x^{2}}{1+\beta_x}\inner{x}{v}x.
\end{aligned}
\end{equation}

\mypara{Distance from the Identity.} This can be directly obtained by the gyro identity.
\begin{equation}
\dist(\zerovec,y)
=
\frac{2}{\sqrt{-K}}
\tanh^{-1}\left(
\sqrt{-K}\norm{\pi(y)}
\right).
\end{equation}
Using the same identity as above with $t=\sqrt{-K}\norm{y}$ yields
\begin{equation}
\dist(\zerovec,y)
=
\frac{1}{\sqrt{-K}}
\sinh^{-1}\left(\sqrt{-K}\norm{y}\right).
\end{equation}

\subsection[Proof of \cref{pvnn:thm:gyro-riem-pv}]{Proof of \cref{pvnn:thm:gyro-riem-pv}}
\linkofproof{pvnn:thm:gyro-riem-pv}
\begin{proof}
    As shown in \cref{prop:stereographic_gyro_from_riemannian}, the Möbius gyroaddition and gyromultiplication can be written by the Riemannian operators. Besides, the isometry $\PVtoPB$ preserves the identity: $\PVtoPB(\zerovec)=\zerovec$. By \citet[Lems.~2.1--2.2]{nguyen2023building}, one can directly obtain the results.
\end{proof}

\subsection[Proof of \cref{pvnn:thm:pv-hyperplane-distance}]{Proof of \cref{pvnn:thm:pv-hyperplane-distance}}
\label{pvnn:app:pv-distance-derivation}
\linkofproof{pvnn:thm:pv-hyperplane-distance}
We first establish the PV hyperplane equivalence and then derive the distance formula.

\subsubsection{Equivalent Characterization of the PV Hyperplane}
We first review a useful lemma from \citet[Lem.~J.1]{chen2025riefc}.
\begin{parislemma}
    \label{pvnn:app:lem:inner_equivalence}
    We assume that the manifold $\calM$ admits a gyrogroup defined by
    \begin{equation}
        x \oplus y = \rieexp_{x} \left( \pt{e}{x} \left( \rielog_e \left( y   \right)\right)\right), \forall x,y \in \calM,
    \end{equation}
    where $e \in \calM$ is the origin of the manifold. Then, we have the following 
    \begin{equation}
        \left \langle \rielog_p (x ), a \right \rangle _{p}
        =\left \langle \rielog_e ( \ominus p \oplus x ), \pt{p}{e}(a) \right \rangle_{e}, \quad \forall x,p \in \calM \text{ and } \forall a \in T_p\calM.
    \end{equation}
\end{parislemma}
Now, we are ready to prove \cref{pvnn:thm:pv-hyperplane-distance}.
\begin{proof}[Proof of PV hyperplane]
    \cref{pvnn:thm:gyro-riem-pv} indicates that the assumption of \cref{pvnn:app:lem:inner_equivalence} holds with $\calM=\PVspace{n}$, $\oplus=\PVoplus$ and $e=\zerovec$. Then, the PV hyperplane
    \begin{equation}
        H_{a,p}
        =
        \left\{x \in \PVspace{n} \mid \inner{\rielog_{p}(x)}{a}_{p}=0\right\}
    \end{equation}
    can be rewritten as
    \begin{equation}
        H_{a,p}
        =
        \left\{x \in \PVspace{n} \mid \inner{\rielog_{\zerovec}(-p \PVoplus x)}{\pt{p}{\zerovec}(a)}_{\zerovec}=0\right\}.
    \end{equation}

    Using the explicit PV operators in \cref{pvnn:thm:pv-exp-log-pt-at-x} and the PV metric in \cref{pvnn:eq:pv-metric}, we have
    \begin{equation}
        \begin{aligned}
            \rielog_{\zerovec}(-p \PVoplus x)&=\alpha (-p \PVoplus x), \quad \text{for some scalar } \alpha \geq 0, \\
            \pt{p}{\zerovec}(a) &=\beta d_p\pi(a), \quad \text{for some scalar } \beta > 0, \\
            g_{\zerovec}(u,v) &=\inner{u}{v}.
        \end{aligned}
    \end{equation}
    As $\alpha=0$ is trivial, we only consider the case $\alpha>0$:
    \begin{equation}
        \inner{\rielog_{\zerovec}(-p \PVoplus x)}{\pt{p}{\zerovec}(a)}_{\zerovec}=0
        \quad \Longleftrightarrow \quad
        \inner{-p \PVoplus x}{d_p\pi(a)}=0.
    \end{equation}
\end{proof}

\subsubsection{PV Point-to-Hyperplane Distance}

We first prove a lemma on the isometry and point-to-hyperplane distance, which will be used to derive the PV point-to-hyperplane distance.

\begin{parislemma}[Isometry and point-to-hyperplane distance]
    \label{pvnn:lem:isometry-hplane-distance}
    Let $(\calM,g)$ and $(\bar{\calM},\bar{g})$ be Riemannian manifolds and let $\phi:\calM \to \bar{\calM}$ be a Riemannian isometry. For $p \in \calM$ and $a \in T_p\calM$, define the hyperplane
    \begin{equation}
        H_{a,p}
        =
        \left\{x \in \calM \mid g_p\left(\rielog_{p}(x),a\right)=0\right\}.
    \end{equation}
    Let $\bar{p}=\phi(p)$ and $\bar{a}=d_p\phi(a) \in T_{\bar{p}}\bar{\calM}$, and define the corresponding hyperplane on $\bar{\calM}$ by
    \begin{equation}
        \bar{H}_{\bar{a},\bar{p}}
        =
        \left\{\bar{x} \in \bar{\calM} \mid \bar{g}_{\bar{p}}\left(\bar{\rielog}_{\bar{p}}(\bar{x}),\bar{a}\right)=0\right\}.
    \end{equation}
    Then $\phi$ maps $H_{a,p}$ onto $\bar{H}_{\bar{a},\bar{p}}$, that is,
    \begin{equation}
        \phi\left(H_{a,p}\right)
        =
        \bar{H}_{\bar{a},\bar{p}}.
    \end{equation}
    Moreover, for every $x \in \calM$ we have
    \begin{equation}
        \dist_{\calM}\left(x, H_{a,p}\right)
        =
        \dist_{\bar{\calM}}\left(\phi(x), \bar{H}_{\bar{a},\bar{p}}\right),
    \end{equation}
    when the point-to-hyperplane distance exists. Here, $\dist_{\calM}$ and $\dist_{\bar{\calM}}$ denote the Riemannian distances on $\calM$ and $\bar{\calM}$, respectively.
\end{parislemma}
\begin{proof}
    Since $\phi$ is a Riemannian isometry, we have
    \begin{equation}
        g_p\left(\rielog_p(x),a\right)
        =
        \bar{g}_{\bar{p}}\left(d_p\phi\left(\rielog_p(x)\right),d_p\phi(a)\right)
        =
        \bar{g}_{\bar{p}}\left(\bar{\rielog}_{\bar{p}}\left(\phi(x)\right),\bar{a}\right).
    \end{equation}
    Therefore,
    \begin{equation}
        g_p\left(\rielog_p(x),a\right)=0
        \quad \Longleftrightarrow \quad
        \bar{g}_{\bar{p}}\left(\bar{\rielog}_{\bar{p}}\left(\phi(x)\right),\bar{a}\right)=0,
    \end{equation}
    which shows that $x \in H_{a,p}$ if and only if $\phi(x) \in \bar{H}_{\bar{a},\bar{p}}$, and hence $\phi\left(H_{a,p}\right)=\bar{H}_{\bar{a},\bar{p}}$.

    For the point-to-hyperplane distance, recall that for a subset $S \subset \calM$ the distance from $x$ to $S$ is
    \begin{equation}
        \dist_{\calM}(x,S)
        =
        \inf_{z \in S} \dist_{\calM}(x,z).
    \end{equation}
    For the point-to-hyperplane distance, we have
    \begin{equation}
        \begin{aligned}
            \dist_{\calM}\left(x, H_{a,p}\right)
            &= \inf_{z \in H_{a,p}} \dist_{\calM}(x,z) \\
            &= \inf_{z \in H_{a,p}} \dist_{\bar{\calM}}\left(\phi(x),\phi(z)\right) \\
            &= \inf_{\bar{z} \in \bar{H}_{\bar{a},\bar{p}}} \dist_{\bar{\calM}}\left(\phi(x),\bar{z}\right) \\
            &= \dist_{\bar{\calM}}\left(\phi(x), \bar{H}_{\bar{a},\bar{p}}\right).
        \end{aligned}
    \end{equation}
\end{proof}

Next, we review the Poincaré hyperplane and point-to-hyperplane distance \citep[Sec.~3.1]{ganea2018hyperbolic}.

\mypara{Poincaré Point-to-Hyperplane Distance.} For a point $p \in \pball{n}$ and a normal vector $a \in T_p\pball{n}$, the Poincaré point-to-hyperplane distance is given by \citet[Thm.~5]{ganea2018hyperbolic}:
\begin{align}
    H^{\PB}_{a,p} 
    &=\left\{x \in \pball{n} \mid \inner{\rielog _p^{\PB}(x)}{a}_{p}=0\right\}
    =\left\{x \in \pball{n} \mid \inner{-p \Moplus x}{a} =0\right\}, \\
    \label{pvnn:eq:poincare-hyperplane}
    d^{\PB}(y, H^{\PB}_{a,p})
    &=
    \frac{1}{\sqrt{-K}}
    \sinh^{-1}\left(
    \frac{2 \sqrt{-K}\left|\inner{-p \Moplus y}{a}\right|}
    {\left(1+K\norm{-p \Moplus y}^2\right)\norm{a}}
    \right).
\end{align}

\begin{proof}[Proof of the PV point-to-hyperplane distance]
    Let
    \begin{equation}
        \bar{p}=\pi(p) \in \pball{n},
        \quad
        \bar{a}=d_p\pi(a) \in T_{\bar{p}}\pball{n},
        \quad
        \bar{y}=\pi(y) \in \pball{n}.
    \end{equation}
    By \cref{pvnn:lem:isometry-hplane-distance}, the point-to-hyperplane distances satisfy
    \begin{equation}
        \label{pvnn:eq:pv-pb-distance}
        \dist^{\PV}\left(y, H_{a,p}\right)
        =
        \dist^{\PB}\left(\bar{y}, \bar{H}_{\bar{a},\bar{p}}\right).
    \end{equation}

    Applying the Poincaré distance formula in \cref{pvnn:eq:poincare-hyperplane} with $p=\bar{p}$, $a=\bar{a}$, and $y=\bar{y}$ gives
    \begin{equation}
        \label{pvnn:eq:pb-hplane-distance}
        \dist^{\PB}\left(\bar{y}, \bar{H}_{\bar{a},\bar{p}}\right)
        =
        \frac{1}{\sqrt{-K}}
        \sinh^{-1}\left(
        \frac{2 \sqrt{-K}\left|\inner{-\bar{p} \Moplus \bar{y}}{\bar{a}}\right|}
        {\left(1+K\norm{-\bar{p} \Moplus \bar{y}}^2\right)\norm{\bar{a}}}
        \right).
    \end{equation}
    The gyrovector isomorphism $\pi$ implies
    \begin{equation}
        -\bar{p} \Moplus \bar{y}
        =
        \pi(-p \PVoplus y).
    \end{equation}
    Denote $z=-p \PVoplus y$. From \cref{pvnn:eq:pv-poincare-isos}, we have the explicit expression
    \begin{equation}
        \pi(z)=\frac{\beta_z}{1+\beta_z} z,
    \end{equation}
    with $\beta_z>0$. Since $\beta_z=\frac{1}{\sqrt{1-K\norm{z}^2}}$, we obtain
    \begin{equation}
        \begin{aligned}
    1+K\norm{\pi(z)}^2
    &= 1+K\left(\frac{\beta_z}{1+\beta_z}\right)^{2}\norm{z}^2 \\
        &= 1+\frac{K\beta_z^2\norm{z}^2}{(1+\beta_z)^2} \\
        &=1+\frac{\beta_z^2 (1-\beta_z^{-2}) }{(1+\beta_z)^2} \\
        &=1+\frac{\beta_z^2 - 1 }{(1+\beta_z)^2} \\
        &=1+\frac{\beta_z - 1 }{1+\beta_z} \\
        &=\frac{2 \beta_z}{1+\beta_z} \\
        \end{aligned}
    \end{equation}
    The above yields
    \begin{equation}
        \frac{2 \sqrt{-K}\left|\inner{\pi(z)}{\bar{a}}\right|}
        {\left(1+K\norm{\pi(z)}^2\right)\norm{\bar{a}}}
        =
        \frac{\sqrt{-K}\left|\inner{z}{\bar{a}}\right|}{\norm{\bar{a}}}.
    \end{equation}
    Therefore,
    \begin{equation}
        d\left(y, H_{a,p}\right)
        =
        \frac{1}{\sqrt{-K}}
        \sinh^{-1}\left(
        \frac{\sqrt{-K}\left|\inner{-p \PVoplus y}{d_p\pi(a)}\right|}{\|d_p\pi(a)\|}
        \right).
    \end{equation}
\end{proof}

\subsection[Proof of \cref{pvnn:thm:pv-mlr}]{Proof of \cref{pvnn:thm:pv-mlr}}
\linkofproof{pvnn:thm:pv-mlr}

\begin{proof}[Proof of PV MLR]
For clarity, we fix a class index $k$ and omit $k$ in the notation whenever possible. We denote $\pi=\PVtoPB$ as in \cref{pvnn:thm:pv-hyperplane-distance}.

\mypara{Step 1: From Hyperplane Distance to a Signed Score.} The PV MLR in \cref{pvnn:eq:pv-mlr-start} associated with parameters $(p,a)$ for $x \in \PVspace{n}$ is
\begin{equation}
\label{pvnn:app:eq:vk-start-form}
\begin{aligned}
&v_k(x) \\
&=
    \sign\left(\inner{-p_k \PVoplus x}{d_{p_k}\pi(a_k)}\right)
    \|a_k\|_{p_k}
    d\left(x,H_{a_k,p_k}\right) \\
&\stackrel{(1)}{=}
    \frac{\|a_k\|_{p_k}}{\sqrt{-K}}
    \sign\left(\inner{-p_k \PVoplus x}{d_{p_k}\pi(a_k)}\right)
    \sinh^{-1}\left(
    \frac{\sqrt{-K}\left|\inner{-p_k \PVoplus x}{d_{p_k}\pi(a_k)}\right|}{\|d_{p_k}\pi(a_k)\|}
    \right) \\
&\stackrel{(2)}{=}
    \frac{\|a_k\|_{p_k}}{\sqrt{-K}}
    \sinh^{-1}\left(
    \frac{\sqrt{-K}\inner{-p_k \PVoplus x}{d_{p_k}\pi(a_k)}}{\|d_{p_k}\pi(a_k)\|}
    \right).
\end{aligned}
\end{equation}
The above comes from the following.
\begin{enumerate}
    \item \cref{pvnn:thm:pv-hyperplane-distance};
    \item $\sinh^{-1}$ is odd and strictly increasing.
\end{enumerate}

\mypara{Step 2: Trivialization and Reduction to a Single Direction.} We adopt the unidirectional parameterization in \cref{pvnn:subsec:pv-mlr}:
\begin{equation}
\label{pvnn:eq:pv-mlr-trivialization}
    p_k
    =
    \rieexp_{\zerovec}\left(r_k [z_k]\right),
    \qquad
    a_k
    =
    \pt{\zerovec}{p_k}\left(z_k\right),
    \qquad
    [z_k]
    =
    \frac{z_k}{\|z_k\|},
\end{equation}
with $z_k \in T_{\zerovec}\PVspace{n} \cong \bbR{n}$ and $r_k \in \bbRscalar$.

As parallel transport is an isometry, we have
\begin{equation}
\label{pvnn:eq:ak-norm-equals-zk}
    \|a_k\|_{p_k}
    =
    \|z_k\|_{\zerovec}
    =
    \|z_k\|.
\end{equation}
Moreover, $p_k$ and $z_k$ are collinear, because $\rieexp_{\zerovec}$ in \cref{pvnn:thm:pv-exp-log-pt-at-x} preserves directions at the origin.
Using the explicit expression of $\pt{\zerovec}{y}$ at the origin in \cref{pvnn:thm:pv-exp-log-pt-at-x}, we see that $\pt{\zerovec}{p_k}$ maps $z_k$ to a linear combination of $z_k$ and $p_k$. Therefore, $a_k$ is also collinear with $z_k$.

The differential $d_{p_k}\pi$ in \cref{pvnn:lem:pv-poincare-diffs} has the form
\begin{equation}
    d_{p_k}\pi(v)
    =
    \alpha_k v
    +
    \beta_k \inner{p_k}{v} p_k,
    \quad
    \alpha_k>0,\ \beta_k \in \bbRscalar,
\end{equation}
so $d_{p_k}\pi$ maps any vector in $\operatorname{span}\{z_k\}$ into the same one-dimensional subspace.
Consequently, there exists a scalar $\lambda_k>0$ such that
\begin{equation}
\label{pvnn:eq:dpi-ak-collinear}
    d_{p_k}\pi(a_k)
    =
    \lambda_k z_k.
\end{equation}
The sign of $\lambda_k$ can be absorbed into $z_k$ by redefining $z_k \leftarrow -z_k$ if necessary. Without loss of generality we may assume $\lambda_k>0$.
Putting \cref{pvnn:eq:pv-mlr-trivialization}, \cref{pvnn:eq:ak-norm-equals-zk}, \cref{pvnn:eq:dpi-ak-collinear} and $\|d_{p_k}\pi(a_k)\| = \lambda_k \|z_k\|$ into \cref{pvnn:app:eq:vk-start-form} yields
\begin{equation}
\label{pvnn:app:eq:vk-step3-final}
    v_k(x)
    =
    \frac{\|z_k\|}{\sqrt{-K}}
    \sinh^{-1}\left(
    \frac{\sqrt{-K}}{\|z_k\|}
    \inner{-p_k \PVoplus x}{z_k}
    \right).
\end{equation}

\mypara{Step 3: Eliminating the Gyroaddition.} The remaining task is to expand the gyro-additive term in \cref{pvnn:app:eq:vk-step3-final}. From \cref{pvnn:sec:preliminaries}, PV gyroaddition is given by
\begin{equation}
    u \PVoplus v
    =
    u
    +
    v
    +
    \left(
    \frac{1-\beta_v}{\beta_v}
    -
    K\frac{\beta_u}{1+\beta_u}\inner{u}{v}
    \right)u,
    \qquad
    \beta_w
    =
    \frac{1}{\sqrt{1-K\|w\|^2}}.
\end{equation}
Setting $u=-p_k$ and $v=x$ yields
\begin{equation}
\begin{aligned}
    -p_k \PVoplus x
    &=
    -p_k
    +
    x
    +
    \left(
    \frac{1-\beta_x}{\beta_x}
    -
    K\frac{\beta_{p_k}}{1+\beta_{p_k}}\inner{-p_k}{x}
    \right)(-p_k).
\end{aligned}
\end{equation}
Taking the inner product with $z_k$ gives
\begin{equation}
\begin{aligned}
    \inner{-p_k \PVoplus x}{z_k}
    &=
    \inner{-p_k}{z_k}
    +
    \inner{x}{z_k}
    +
    \left(
    \frac{1-\beta_x}{\beta_x}
    -
    K\frac{\beta_{p_k}}{1+\beta_{p_k}}\inner{-p_k}{x}
    \right)\inner{-p_k}{z_k} \\
    &=
    \inner{x}{z_k}
    +
    \left(
    1
    +
    \frac{1-\beta_x}{\beta_x}
    -
    K\frac{\beta_{p_k}}{1+\beta_{p_k}}\inner{-p_k}{x}
    \right)\inner{-p_k}{z_k}.
\end{aligned}
\end{equation}

Next, we rewrite the above expression using the unidirectional parameterization of $p_k$.
From \cref{pvnn:eq:pv-mlr-trivialization} and the explicit PV exponential at the origin in \cref{pvnn:thm:pv-exp-log-pt-at-x}, we have
\begin{equation}
    p_k
    =
    \rieexp_{\zerovec}\left(r_k[z_k]\right)
    =
    \frac{1}{\sqrt{-K}}
    \sinh\left(\sqrt{-K}r_k\right)
    \frac{z_k}{\|z_k\|}.
\end{equation}
Thus,
\begin{equation}
    \inner{-p_k}{z_k}
    =
    -\frac{1}{\sqrt{-K}}
    \sinh\left(\sqrt{-K}r_k\right)\|z_k\|.
\end{equation}
Moreover, since $p_k$ and $z_k$ share the same direction, any $x$ admits the decomposition
\begin{equation}
    x
    =
    x_{\parallel}
    +
    x_{\perp},
    \qquad
    x_{\parallel}
    =
    \frac{\inner{x}{z_k}}{\|z_k\|^2}z_k,
    \qquad
    \inner{x_{\perp}}{z_k}
    =
    0,
\end{equation}
which implies
\begin{equation}
    \inner{-p_k}{x}
    =
    \inner{-p_k}{x_{\parallel}}
    =
    \frac{\inner{x}{z_k}}{\|z_k\|^2}
    \inner{-p_k}{z_k}
    =
    -\frac{1}{\sqrt{-K}}
    \sinh\left(\sqrt{-K}r_k\right)
    \frac{\inner{x}{z_k}}{\|z_k\|}.
\end{equation}
The beta factor at $p_k$ is
\begin{equation}
    \beta_{p_k}
    =
    \frac{1}{\sqrt{1-K\|p_k\|^2}}
    =
    \sech\left(\sqrt{-K}r_k\right),
\end{equation}
where we used $\|p_k\|^2 = -\frac{1}{K}\sinh^2\left(\sqrt{-K}r_k\right)$ and the identity $1+\sinh^2(t)=\cosh^2(t)$.

Using $\inner{-p_k}{z_k}$, $\inner{-p_k}{x}$, and $\beta_{p_k}$, we have
\begin{equation}
\begin{aligned}
    & \inner{-p_k \PVoplus x}{z_k}\\
    &=
    \inner{x}{z_k}
    +
    \left(
    1
    +
    \frac{1-\beta_x}{\beta_x}
    -
    K\frac{\beta_{p_k}}{1+\beta_{p_k}}\inner{-p_k}{x}
    \right)\inner{-p_k}{z_k} \\
    &=
    \inner{x}{z_k}
    +
    \left(
    \frac{1}{\beta_x}
    -
    K\frac{\beta_{p_k}}{1+\beta_{p_k}}\inner{-p_k}{x}
    \right)\inner{-p_k}{z_k} \\
    &=
    \inner{x}{z_k}
    +
    \frac{1}{\beta_x}
    \left(
    -\frac{\sinh\left(\sqrt{-K}r_k\right)}{\sqrt{-K}}
    \|z_k\|
    \right) \\
    &\quad -
    K\frac{\beta_{p_k}}{1+\beta_{p_k}}
    \left(
    -\frac{\sinh\left(\sqrt{-K}r_k\right)}{\sqrt{-K}}
    \frac{\inner{x}{z_k}}{\|z_k\|}
    \right)
    \left(
    -\frac{\sinh\left(\sqrt{-K}r_k\right)}{\sqrt{-K}}
    \|z_k\|
    \right) \\
    &=
    \inner{x}{z_k}
    -
    \frac{\sinh\left(\sqrt{-K}r_k\right)}{\sqrt{-K}}
    \frac{\|z_k\|}{\beta_x}
    +
    \frac{\beta_{p_k}\sinh^2\left(\sqrt{-K}r_k\right)}{1+\beta_{p_k}}
    \inner{x}{z_k}\\
    &=
    \left( 1+\frac{\beta_{p_k}\sinh^2\left(\sqrt{-K}r_k\right)}{1+\beta_{p_k}} \right)
    \inner{x}{z_k}
    -
    \frac{\sinh\left(\sqrt{-K}r_k\right)}{\sqrt{-K}}
    \frac{\|z_k\|}{\beta_x}.
\end{aligned}
\end{equation}
Since $\beta_{p_k}=\sech\left(\sqrt{-K}r_k\right)$ and $1+\sinh^2\left(\sqrt{-K}r_k\right)=\cosh^2\left(\sqrt{-K}r_k\right)=1/\beta_{p_k}^2$, we have
\begin{equation}
    \begin{aligned}
    1+\frac{\beta_{p_k}\sinh^2\left(\sqrt{-K}r_k\right)}{1+\beta_{p_k}}
    &=
    \frac{1+\beta_{p_k}+\beta_{p_k}\sinh^2\left(\sqrt{-K}r_k\right)}{1+\beta_{p_k}} \\
    &=
    \frac{1+\beta_{p_k}\cosh^2\left(\sqrt{-K}r_k\right)}{1+\beta_{p_k}} \\
    &=
    \frac{1+1/\beta_{p_k}}{1+\beta_{p_k}}
    =
    \frac{1}{\beta_{p_k}}
    =
    \cosh\left(\sqrt{-K}r_k\right),
    \end{aligned}
\end{equation}
which implies
\begin{equation}
\begin{aligned}
    \inner{-p_k \PVoplus x}{z_k}
    &=
    \cosh\left(\sqrt{-K} r_k\right)
    \inner{x}{z_k}
    -
    \frac{\sinh\left(\sqrt{-K} r_k\right)}{\sqrt{-K}}
    \frac{\|z_k\|}{\beta_x}.
\end{aligned}
\end{equation}
Recalling that $\beta_x = 1/\sqrt{1-K\|x\|^2}$, we obtain
\begin{equation}
\label{pvnn:eq:inner-gyr-pv}
    \inner{-p_k \PVoplus x}{z_k}
    =
    \cosh\left(\sqrt{-K} r_k\right)
    \inner{x}{z_k}
    -
    \frac{\sinh\left(\sqrt{-K} r_k\right)}{\sqrt{-K}}
    \|z_k\|\sqrt{1-K\|x\|^2}.
\end{equation}

Substituting \cref{pvnn:eq:inner-gyr-pv} into \cref{pvnn:app:eq:vk-step3-final}, we arrive at
\begin{equation}
\label{pvnn:eq:vk-final-fully-expanded-app}
    v_k(x)
    =
    \frac{\|z_k\|}{\sqrt{-K}}
    \sinh^{-1}\left(
    \cosh\left(\sqrt{-K} r_k\right)\frac{\sqrt{-K}}{\|z_k\|}\inner{x}{z_k}
    -
    \sinh\left(\sqrt{-K} r_k\right)\sqrt{1-K\|x\|^2}
    \right).
\end{equation}
\end{proof}

\begin{proof}[Proof of PV MLR limits]
By Taylor expansions, we have
\begin{equation}
    \begin{aligned}
    \cosh\left(\sqrt{-K}r_k\right)
    &=
    1 - \frac{K r_k^2}{2} + \mathcal{O}(K^2),\\
    \sinh\left(\sqrt{-K}r_k\right)
    &=
    \sqrt{-K} r_k + \mathcal{O}\left((-K)^{3/2}\right),\\
    \sqrt{1-K\|x\|^2}
    &=
    1 - \frac{K\|x\|^2}{2} + \mathcal{O}(K^2).
    \end{aligned}
\end{equation}

The argument of $\sinh^{-1}(\cdot)$ in \cref{pvnn:eq:vk-final-fully-expanded-app} can be simplified as
\begin{equation}
\begin{aligned}
    &\sinh^{-1} 
    \left\{ 
    \cosh\left(\sqrt{-K} r_k\right)\frac{\sqrt{-K}}{\|z_k\|}\inner{x}{z_k}
    -
    \sinh\left(\sqrt{-K} r_k\right)\sqrt{1-K\|x\|^2} 
    \right\} \\
    &=
    \sinh^{-1} 
    \left\{ 
    \left(
    1 - \frac{K r_k^2}{2} + \mathcal{O}(K^2)
    \right)
    \frac{\sqrt{-K}}{\|z_k\|}\inner{x}{z_k} 
    \right. \\
    &\left. 
    \quad -
    \left(
    \sqrt{-K} r_k + \mathcal{O}\left((-K)^{3/2}\right) 
    \right)
    \left(
    1 - \frac{K\|x\|^2}{2} + \mathcal{O}(K^2)
    \right) 
    \right\}\\
    &=
    \sinh^{-1} 
    \left\{
    \sqrt{-K}
    \left(
    \frac{\inner{x}{z_k}}{\|z_k\|}
    -
    r_k
    \right)
    + \mathcal{O}\left((-K)^{3/2}\right) 
    \right\}\\
    &=
    \sqrt{-K}
    \left(
    \frac{\inner{x}{z_k}}{\|z_k\|}
    -
    r_k
    \right)
    + \mathcal{O}\left((-K)^{3/2}\right).
\end{aligned}
\end{equation}

Substituting this into \cref{pvnn:eq:vk-final-fully-expanded-app} gives
\begin{equation}
\begin{aligned}
    v_k(x)
    &=
    \frac{\|z_k\|}{\sqrt{-K}}
    \left(
    \sqrt{-K}
    \left(
    \frac{\inner{x}{z_k}}{\|z_k\|}
    -
    r_k
    \right)
    + \mathcal{O}\left((-K)^{3/2}\right)
    \right) \\
    &=
    \|z_k\|
    \left(
    \frac{\inner{x}{z_k}}{\|z_k\|}
    -
    r_k
    \right)
    + \mathcal{O}(-K) \\
    &=
    \inner{x}{z_k}
    -
    r_k \|z_k\|
    + \mathcal{O}(-K),\\
    &\xrightarrow{K \to 0^{-}}
    \inner{x}{z_k}
    -
    r_k \|z_k\|.
\end{aligned}
\end{equation}
\end{proof}

\subsection[Proof of \cref{pvnn:thm:pv-fc}]{Proof of \cref{pvnn:thm:pv-fc}}
\linkofproof{pvnn:thm:pv-fc}
\begin{proof}[Proof of PV FC layer]
Specializing \cref{pvnn:thm:pv-hyperplane-distance} to $p=\zerovec$ and $a=e_k$ and using that $-\zerovec \PVoplus y=y$ gives the LHS
\begin{equation}\label{pvnn:eq:dist-coordinate}
    \sign\left(\inner{d_{\zerovec}\pi(e_k)}{-\zerovec \PVoplus y}\right)
    \dist\left(y, H_{e_k,\zerovec}\right)
    =
    \frac{1}{\sqrt{-K}}
    \sinh^{-1}\left(
    \sqrt{-K} y_k
    \right),   
\end{equation}
with $y_k=\inner{y}{e_k}$. Then, we obtain
\begin{equation}
    y_k
    =
    \frac{1}{\sqrt{-K}}
    \sinh\left(\sqrt{-K}v_k(x)\right),
    \quad
    k=1,\dots,m.
\end{equation}
\end{proof}

\begin{proof}[Proof of PV FC limits]
By \cref{pvnn:thm:pv-mlr}, as $K \to 0^{-}$ we have
\begin{equation}
    v_k(x) \to \inner{x}{z_k} + b_k,
    \quad
    b_k = - r_k \|z_k\|.
\end{equation}

For $K<0$ and $v_k(x) \neq 0$, we can rewrite $y_k$ as
\begin{equation}
    y_k
    =
    v_k(x)
    \frac{\sinh\left(\sqrt{-K} v_k(x)\right)}{\sqrt{-K} v_k(x)},
\end{equation}
and we define the fraction to be $1$ when $v_k(x)=0$. Since $\sqrt{-K} \to 0$ and $v_k(x)$ converges to a finite limit, we have $\sqrt{-K} v_k(x) \to 0$. Using the standard limit $\lim_{u \to 0} \sinh(u)/u = 1$, it follows that
\begin{equation}
    \frac{\sinh\left(\sqrt{-K} v_k(x)\right)}{\sqrt{-K} v_k(x)}
    \to 1
    \quad \text{as} \quad
    K \to 0^{-}.
\end{equation}
Combining the above limits yields
\begin{equation}
    \lim_{K \to 0^{-}} y_k
    =
    \lim_{K \to 0^{-}} v_k(x)
    =
    \inner{x}{z_k} + b_k.
\end{equation}
\end{proof}

\subsection[Proof of \cref{pvnn:thm:homogeneity}]{Proof of \cref{pvnn:thm:homogeneity}}
\linkofproof{pvnn:thm:homogeneity}
\begin{proof}
The result is first established in the Poincaré ball model in \cref{thm:gyrobn,thm:gyroinvariance_stereo}.
Since $\pi=\PVtoPB:\PVspace{n}\to\pball{n}$ is a Riemannian isometry, for
$x,y\in\PVspace{n}$, $v\in T_x\PVspace{n}$, and $t\in\bbRscalar$, it intertwines the key
geometric operators used in the proof:
\begin{align}
\pi\left(\rieexp^{\PV}_x(v)\right)
&= \rieexp^{\PB}_{\pi(x)}\left(d_x\pi(v)\right), \\
d_x\pi\left(\rielog^{\PV}_x(y)\right)
&= \rielog^{\PB}_{\pi(x)}\left(\pi(y)\right), \\
\pi\left(x \PVoplus y\right)
&= \pi(x) \Moplus \pi(y),\\
\pi\left(t \PVotimes x\right) 
&= t \Modot \pi(x).
\end{align}
Together with the preservation of geodesic distances and Fréchet means under
the Riemannian isometry $\pi$, these identities imply that both homogeneity
identities are preserved under $\pi$.
Therefore the same theorem holds for the PV model
by the isometry $\pi$.
\end{proof}

\subsection[Proof of \cref{pvnn:prop:pv-lorentz-isometry}]{Proof of \cref{pvnn:prop:pv-lorentz-isometry}}
\linkofproof{pvnn:prop:pv-lorentz-isometry}
\begin{proof}
We first recall the isometries between the Poincaré ball and the Lorentz model \citep[Sec.~2.1]{skopek2020mixed}:
\begin{align}
\LtoPball(x)
&=
\frac{x_s}{1+\sqrt{|K|} x_t},\\
\PballtoL(y)
&=
\begin{bmatrix}
\displaystyle \frac{1}{\sqrt{|K|}} \frac{1-K\norm{y}^2}{1+K\norm{y}^2} \\
\displaystyle \frac{2y}{1+K\norm{y}^2}
\end{bmatrix}.
\end{align}

Hence, the following are Riemannian isometries:
\begin{equation}
\LtoPV
=
\PBtoPV \circ \LtoPball,
\qquad
\PVtoL
=
\PballtoL \circ \PVtoPB.
\end{equation}
It remains to derive the explicit formulas. 

For $x=[x_t,x_s^\top]^\top \in \lorentz{n}$, we first map to the Poincaré ball:
\begin{equation}
y
=
\LtoPball(x)
=
\frac{x_s}{1+\sqrt{|K|} x_t}.
\end{equation}
Applying $\PBtoPV$ from \cref{pvnn:eq:pv-poincare-isos} yields
\begin{equation}
\LtoPV(x)
=
\PBtoPV(y)
=
2 \gamma_y^2 y,
\qquad
\gamma_y
=
\frac{1}{\sqrt{1+K\norm{y}^2}}.
\end{equation}
Using $y = x_s/(1+\sqrt{|K|} x_t)$, we compute
\begin{equation}
1+K\norm{y}^2
=
1
+
K \frac{\norm{x_s}^2}{\left(1+\sqrt{|K|} x_t\right)^2}
=
\frac{\left(1+\sqrt{|K|} x_t\right)^2 + K\norm{x_s}^2}
     {\left(1+\sqrt{|K|} x_t\right)^2}.
\end{equation}
Since $x \in \lorentz{n}$ satisfies $\Linner{x}{x} = 1/K$, we have
\begin{equation}
\Linner{x}{x}
= -x_t^2 + \norm{x_s}^2 = \frac{1}{K}
\quad
\Rightarrow  
\quad
\norm{x_s}^2
=
x_t^2 + \frac{1}{K}.
\end{equation}
Substituting this into the numerator gives
\begin{equation}
\begin{aligned}
\left(1+\sqrt{|K|} x_t\right)^2 + K\norm{x_s}^2
&=
\left(1+\sqrt{|K|} x_t\right)^2 + K\left(x_t^2 + \frac{1}{K}\right) \\
&=
\left(1+\sqrt{|K|} x_t\right)^2 + K x_t^2 + 1 \\
&=
1 + 2\sqrt{|K|} x_t + |K| x_t^2 + K x_t^2 + 1 \\
&=
2\left(1+\sqrt{|K|} x_t\right).
\end{aligned}
\end{equation}
Therefore,
\begin{equation}
1+K\norm{y}^2
=\frac{2\left(1+\sqrt{|K|} x_t\right)}{\left(1+\sqrt{|K|} x_t\right)^2}
=\frac{2}{1+\sqrt{|K|} x_t},
\end{equation}
and hence
\begin{equation}
\gamma_y^2
=
\frac{1}{1+K\norm{y}^2}
=
\frac{1+\sqrt{|K|} x_t}{2}.
\end{equation}
Finally,
\begin{equation}
\LtoPV(x)
=
2 \gamma_y^2 y
=
2 \cdot \frac{1+\sqrt{|K|} x_t}{2} \cdot \frac{x_s}{1+\sqrt{|K|} x_t}
=
x_s.
\end{equation}

For $\PVtoL$, take $x \in \PVspace{n}$ and map to the Poincaré ball by \cref{pvnn:eq:pv-poincare-isos}:
\begin{equation}
y
=
\PVtoPB(x)
=
\frac{\beta_x}{1+\beta_x} x,
\qquad
\beta_x
=
\frac{1}{\sqrt{1-K\norm{x}^2}}.
\end{equation}
Applying $\PballtoL$, we obtain
\begin{equation}
\PVtoL(x)
=
\PballtoL(y)
=
\begin{bmatrix}
\displaystyle \frac{1}{\sqrt{|K|}} \frac{1-K\norm{y}^2}{1+K\norm{y}^2} \\
\displaystyle \frac{2y}{1+K\norm{y}^2}
\end{bmatrix}.
\end{equation}

We now simplify the spatial and temporal components separately. We write
\begin{equation}
y
=
\frac{\beta_x}{1+\beta_x} x
,
\qquad
\norm{y}^2
=
\left(\frac{\beta_x}{1+\beta_x}\right)^{2} \norm{x}^2.
\end{equation}
Using $\beta_x^2 = 1/(1-K \norm{x}^2)$, we obtain
\begin{equation}
\begin{aligned}
&K\norm{y}^2
=
K \norm{x}^2 \frac{\beta_x^2}{(1+\beta_x)^2}
=
\frac{\beta_x^2 - 1}{(1+\beta_x)^2}
=
\frac{\beta_x - 1}{(1+\beta_x)} \\
&\Rightarrow 1+K\norm{y}^2 = \frac{2\beta_x}{1+\beta_x}, 
\quad 
1-K\norm{y}^2 = \frac{2}{1+\beta_x}.
\end{aligned}
\end{equation}
The spatial component of $\PVtoL(x)$ is
\begin{equation}
\frac{2y}{1+K\norm{y}^2}
=
\frac{2 \frac{\beta_x}{1+\beta_x} x}{\frac{2\beta_x}{1+\beta_x}}
=
x,
\end{equation}
and the temporal component is
\begin{equation}
\frac{1}{\sqrt{|K|}} \frac{1-K\norm{y}^2}{1+K\norm{y}^2}
=
\frac{1}{\sqrt{|K|}}
\cdot
\frac{\frac{2}{1+\beta_x}}{\frac{2\beta_x}{1+\beta_x}}
=
\frac{1}{\sqrt{|K|} \beta_x}
=
\frac{\sqrt{1-K \norm{x}^2}}{\sqrt{|K|}}
= \sqrt{\norm{x}^2 - \tfrac{1}{K}}.
\end{equation}
Thus,
\begin{equation}
\PVtoL(x)
=
\begin{bmatrix}
\displaystyle \sqrt{\norm{x}^2 - \tfrac{1}{K}} \\
\displaystyle x
\end{bmatrix}.
\end{equation}
\end{proof}

\section{Hyperbolic Busemann Neural Networks}
\label{app:hbnn-proofs}

\subsection[Proof of \cref{hbnn:thm:limits-bmlr}]{Proof of \cref{hbnn:thm:limits-bmlr}}
\label{hbnn:app:prf:thm:limits-bmlr}
\linkofproof{hbnn:thm:limits-bmlr}
\begin{proof}
Denoting $\kappa^2 = -K$ with $\kappa>0$, we rewrite the Busemann functions in \cref{hbnn:eq:poincare-busemann,hbnn:eq:lorentz-busemann} as
\begin{align}
\text{(Poincaré)}\quad &B^{v}(x) = \frac{1}{\kappa} \log\left( \frac{\left\| v - \kappa x \right\|^2}{1 - \kappa^2\left\|x\right\|^2} \right), \\
\text{(Lorentz)}\quad &B^{v}(x) = \frac{1}{\kappa} \log\left( \kappa x_t - \kappa \inner{x_s}{v} \right).
\end{align}
For the Poincaré case,
\begin{align}
\frac{\left\| v - \kappa x \right\|^2}{1 - \kappa^2\left\|x\right\|^2}
= \frac{1 - 2\kappa\inner{v}{x} + \kappa^2\left\|x\right\|^2}{1 - \kappa^2\left\|x\right\|^2}
= 1 + \frac{-2\kappa\inner{v}{x} + 2\kappa^2\left\|x\right\|^2}{1 - \kappa^2\left\|x\right\|^2}.
\end{align}
Using $\log\left(1+u\right)=u + O\left(u^2\right)$ as $u\to 0$ and $\left(1 - \kappa^2\left\|x\right\|^2\right)^{-1}=1 + O\left(\kappa^2\right)$, we obtain
\begin{equation}
\begin{aligned}
B^{v}(x)
&= \frac{1}{\kappa}\left[\left(-2\kappa\inner{v}{x}+2\kappa^2\norm{x}^2\right)\left(1+O\left(\kappa^2\right)\right) + O\left(\kappa^2\right)\right] \\
&= -2\inner{v}{x} + 2\kappa\norm{x}^2 + O\left(\kappa\right) \\
&= -2\inner{v}{x} + O\left(\kappa\right).
\end{aligned}
\end{equation}
Therefore, $B^{v}(x) \xrightarrow{K\to 0^{-}} -2\inner{v}{x}$.

For the Lorentz case, the Lorentz constraint $-x_t^2 + \left\|x_s\right\|^2 = -\kappa^{-2}$ and $x_t>0$ yield
\begin{equation}
\kappa x_t = \sqrt{1 + \kappa^2 \left\|x_s\right\|^2} = 1 + \frac{1}{2}\kappa^2\left\|x_s\right\|^2 + O\left(\kappa^4\right).
\end{equation}
Set $z = -\kappa\inner{x_s}{v} + \frac{1}{2}\kappa^2\left\|x_s\right\|^2 + O\left(\kappa^4\right)$. Then
\begin{equation}
\log\left( \kappa x_t - \kappa\inner{x_s}{v} \right) = \log\left(1 + z\right) = z + O\left(z^2\right) = -\kappa\inner{x_s}{v} + O\left(\kappa^2\right),
\end{equation}
which implies $B^{v}(x) = -\inner{x_s}{v} + O\left(\kappa\right)$ and therefore $B^{v}(x) \xrightarrow{K\to 0^{-}} -\inner{x_s}{v}$.

Substituting the above two limits into $u_k(x) = -\alpha_k B^{v_k}(x) + b_k$ gives the stated Euclidean limits of the logits.
\end{proof}
\subsection[Proof of \cref{hbnn:thm:distance-horospheres}]{Proof of \cref{hbnn:thm:distance-horospheres}}
\label{hbnn:app:prf:thm:distance-horospheres}
\linkofproof{hbnn:thm:distance-horospheres}

We first review two useful characterizations of Busemann functions in Hadamard spaces. The class $\mathcal{B}$ below is given by \citet[p.~271]{bridson2013metric}. The equivalence follows from \citet[Prop.~II.8.22]{bridson2013metric}, whose formulation in terms of Busemann functions uses the identification of horofunctions with Busemann functions up to additive constants in \citet[Cor.~II.8.20]{bridson2013metric}.
\begin{parisdefinition}
    Let $(\calX,\dist)$ be a Hadamard space, and let $\mathcal{B}$ be the set of functions
    $h:\calX \to \bbRscalar$ on $(\calX,\dist)$ satisfying:
    \par\leavevmode\vspace{-\baselineskip}
    \begin{enumerate}
        \item $h$ is convex;
        \item 1-Lipschitz: $|h(x)-h(y)|\le \dist(x,y)$ for all $x,y\in \calX$;
        \item for any $x_0\in \calX$ and $r>0$, the function $h$ attains its minimum on the sphere 
        $S_r(x_0)$ at a unique point $y$ with $h(y)=h(x_0)-r$.
    \end{enumerate}
\end{parisdefinition}

\begin{parisproposition}\label{hbnn:app:prop:characterization-busemann}
For a function $h:\calX \to\bbRscalar$, the following conditions are equivalent:
\par\leavevmode\vspace{-\baselineskip}
\begin{enumerate}
    \item $h$ is a Busemann function;
    \item $h\in\mathcal{B}$;
    \item $h$ is convex, and for every $t\in\bbRscalar$, the set $h^{-1}(-\infty,t]$ is nonempty; moreover, for each $x\in \calX$, the curve $c_x:[0,\infty)\to \calX$ defined by 
    $t\mapsto \pi_{h^{-1}(-\infty,h(x)-t]}(x)$ is a geodesic ray.
\end{enumerate}
\end{parisproposition}

Now, we are ready to prove \cref{hbnn:thm:distance-horospheres}.
\begin{proof}
As $\tau_2=\tau_1$ is trivial, we only consider $\tau_2 \neq \tau_1$. We assume $\tau_2 > \tau_1$, and discuss the other direction last.

\mypara{Step 1: Symmetry.} By definition,
\begin{equation}
\dist \left(H^\gamma_{\tau_1},H^\gamma_{\tau_2}\right)
=\inf_{x\in H^\gamma_{\tau_1},y\in H^\gamma_{\tau_2}}\dist(x,y)
=\inf_{y\in H^\gamma_{\tau_2},x\in H^\gamma_{\tau_1}}\dist(y,x)
=\dist \left(H^\gamma_{\tau_2},H^\gamma_{\tau_1}\right).
\end{equation}

\mypara{Step 2: Lower Bound.} For any $x \in H^\gamma_{\tau_2}$ and $y\in H^\gamma_{\tau_1}$, the $1$-Lipschitz property of $B^\gamma$ gives
\begin{equation}
\left|B^\gamma(x)-B^\gamma(y)\right| \le \dist(x,y).
\end{equation}
With $B^\gamma(x)=\tau_2$ and $B^\gamma(y)=\tau_1$ this yields
\begin{equation}\label{hbnn:app:eq:lipschitz-lb}
\tau_2-\tau_1 \le \dist(x,y)\qquad\forall y\in H^\gamma_{\tau_1}.
\end{equation}
Taking infimum in \cref{hbnn:app:eq:lipschitz-lb} over $y\in H^\gamma_{\tau_1}$ gives
\begin{equation}\label{hbnn:app:eq:lower-bound}
\tau_2-\tau_1 \le \dist \left(x,H^\gamma_{\tau_1}\right).
\end{equation}

\mypara{Step 3: Upper Bound.} For any $x\in H^\gamma_{\tau_2}$, by property (3) in \cref{hbnn:app:prop:characterization-busemann},
the projection map
\begin{equation}
c_x(t)=\pi_{\{B^\gamma\le B^\gamma(x)-t\}}(x)
=\pi_{HB^\gamma_{\tau_2 - t}}(x), \quad t \in [0,\infty),
\end{equation}
is a unit-speed geodesic ray: $\dist(x,c_x(t))=t$.

Let $t= \tau_2- \tau_1 > 0$ and $z=c_x(t) \in HB^\gamma_{\tau_1}$. If $B^\gamma(z)<\tau_1$, then $z$ lies in the interior of the horoball $HB^\gamma_{\tau_1}$. We can move slightly from $z$ toward $x$ along the geodesic segment $\overline{xz}$ to obtain a point 
$z_\varepsilon$ with $\dist(x,z_\varepsilon)<\dist(x,z)$, contradicting the minimality of the projection.  
Hence, the projected point $z=c_x(t)$ indeed lies on $H^\gamma_{\tau_1}$: $B^\gamma(z)=\tau_1$. Then, we have the following:
\begin{equation}\label{hbnn:app:eq:upper-bound}
\dist(x,H^\gamma_{\tau_1})
\le \dist(x,HB^\gamma_{\tau_1}) = \dist(x,z)
=\dist(x,c_x(t))
=t
=\tau_2-\tau_1.
\end{equation}

\mypara{Step 4: Sandwich Closure.} Combining \cref{hbnn:app:eq:lower-bound,hbnn:app:eq:upper-bound} gives
\begin{equation}
\dist(x,H^\gamma_{\tau_1})=\tau_2-\tau_1.
\end{equation}

Combining \cref{hbnn:app:eq:upper-bound} and \cref{hbnn:app:eq:lower-bound},
\begin{equation}
\dist \left(x,H^\gamma_{\tau_1}\right) = \tau_2-\tau_1,\qquad\text{for every }x\in H^\gamma_{\tau_2}.
\end{equation}
The right-hand side does not depend on $x$, hence
\begin{equation}
\dist \left(H^\gamma_{\tau_2},H^\gamma_{\tau_1}\right)=\tau_2-\tau_1.
\end{equation}

\mypara{Step 5: Opposite Direction.} If $\tau_1>\tau_2$, geodesic completeness of the Hadamard space allows the geodesic ray $c_x$ from Step 3 to extend to a complete geodesic $\widetilde{c}_x:\bbRscalar\to\calX$. For any $x\in H^\gamma_{\tau_2}$, let $s=\tau_1-\tau_2$ and $z=\widetilde{c}_x(-s)$. Since $B^\gamma(c_x(t))=B^\gamma(x)-t$ for $t\geq0$, convexity and the $1$-Lipschitz property of $B^\gamma$ give $B^\gamma(z)=B^\gamma(x)+s=\tau_1$. Thus, $z\in H^\gamma_{\tau_1}$ and $\dist(x,z)=s$. Combining this upper bound with the $1$-Lipschitz lower bound from Step 2 gives
\begin{equation}
\dist\left(x,H^\gamma_{\tau_1}\right)=\tau_1-\tau_2,
\qquad\text{for every }x\in H^\gamma_{\tau_2}.
\end{equation}
Therefore, in both directions,
\begin{equation}
\dist\left(x,H^\gamma_\tau\right)=\left|B^\gamma(x)-\tau\right|,
\qquad\forall x\in\calX,
\end{equation}
and the symmetry of the distance between horospheres gives
\begin{equation}
\dist \left(H^\gamma_{\tau_2},H^\gamma_{\tau_1}\right)=|\tau_2-\tau_1|.
\end{equation}
\end{proof}

\subsection[Proof of \cref{hbnn:thm:bfc-poincare}]{Proof of \cref{hbnn:thm:bfc-poincare}}
\label{hbnn:app:prf:thm:bfc-poincare}
\linkofproof{hbnn:thm:bfc-poincare}

\begin{proof}
    The hyperplane and point-to-hyperplane distance are presented in \cref{hbnn:app:tab:hyperplane-comparison,hbnn:app:tab:distance-comparison}. The origin of the Poincaré ball model is $e = \zerovec \in \pball{m}$. The specific ones w.r.t. the origin \citep[Def.~1 and Eq.~(56)]{shimizu2021hyperbolic} are
    \begin{align}
        H_{e_k, \zerovec}
        &= \{ y \in \pball{m} \mid \inner{e_k}{y} = 0 \}, \\
        \bar{\dist}\left(y, H_{e_k, \zerovec}\right)
        &= \frac{1}{\sqrt{-K}} \sinh^{-1} \left( \frac{2\sqrt{-K} y_k}{1 + K\norm{y}^{2}} \right).
    \end{align}

    Equating $\bar{\dist}\left(y, H_{e_k, \zerovec}\right)$ with $u_k(x)$ from \cref{hbnn:eq:hyp-fc-p2h} gives
    \begin{equation}\label{hbnn:app:eq:poincare-fc-prf-1}
        \sinh^{-1} \left( \frac{2\sqrt{-K} y_k}{1 + K\norm{y}^{2}} \right) = \sqrt{-K} u_k(x), \quad \forall k\in\{1,\ldots,m\}.
    \end{equation}
    Note that \cref{hbnn:app:eq:poincare-fc-prf-1} takes the same form as \citet[Eq.~(56)]{shimizu2021hyperbolic}, except that their responses are different. The proof below is inspired by their derivation.

    Applying $\sinh(\cdot)$ on both sides of \cref{hbnn:app:eq:poincare-fc-prf-1} yields
    \begin{equation}
        \frac{2\sqrt{-K} y_k}{1 + K\norm{y}^{2}} = \sinh \left( \sqrt{-K} u_k(x) \right).
    \end{equation}
    Define $\omega_k := \frac{1}{\sqrt{-K}} \sinh \left( \sqrt{-K} u_k(x) \right)$ and $\omega = [\omega_k]_{k=1}^{m}$. Then, we have
    \begin{equation}
        2 y_k = \left( 1 + K\norm{y}^{2} \right) \omega_k, \quad \forall k,
    \end{equation}
    which is equivalent to the vector identity
    \begin{equation} \label{hbnn:app:eq:poincare-fc-vector-eq}
        2 y = \left( 1 + K\norm{y}^{2} \right) \omega.
    \end{equation}
    Hence, $y$ is collinear with $\omega$. Write $y = \lambda \omega$ with $\lambda \ge 0$. Substituting into \cref{hbnn:app:eq:poincare-fc-vector-eq} and taking norms gives a quadratic in $\lambda$:
    \begin{equation}
        K\norm{\omega}^{2} \lambda^{2} - 2\lambda + 1 = 0.
    \end{equation}
    Solving and selecting the branch that satisfies $y \to 0$ as $\omega \to 0$ yields
    \begin{equation}
        \lambda = \frac{1 - \sqrt{1 - K\norm{\omega}^{2}}}{K\norm{\omega}^{2}} = \frac{1}{1 + \sqrt{1 - K\norm{\omega}^{2}}}.
    \end{equation}
    Therefore,
    \begin{equation}
        y = \frac{\omega}{1 + \sqrt{1 - K\norm{\omega}^{2}}}, \quad \omega_k = \frac{\sinh \left( \sqrt{-K} u_k(x) \right)}{\sqrt{-K}},
    \end{equation}
    which proves the claim. One can check that $y \in \pball{m}$.
\end{proof}

\subsection[Proof of \cref{hbnn:thm:bfc-lorentz}]{Proof of \cref{hbnn:thm:bfc-lorentz}}
\label{hbnn:app:prf:thm:bfc-lorentz}
\linkofproof{hbnn:thm:bfc-lorentz}

\begin{proof}
    Recalling \cref{hbnn:app:tab:hyperplane-comparison}, a Lorentz hyperplane is 
    \begin{equation}
        H_{w,p} = \{ x \in \lorentz{m} \mid \Linner{w}{x} = 0 \}, \text{ with } p \in \lorentz{m},\ w \in T_{p}\lorentz{m}.
    \end{equation}
    The canonical origin is $\Lzero \in \lorentz{m}$. The tangent space at the origin is
    \begin{equation}
    T_{\Lzero}\lorentz{m}=\{ [0,v^\top]^\top \mid v \in \bbR{m} \},
    \end{equation}
    where each tangent vector has a zero time component. Therefore, the coordinate hyperplane through the origin and orthogonal to the $k$-th axis is
    \begin{equation}
        \begin{aligned}
            H_{\bar{e}_k,e} &= \{ y \in \lorentz{m} \mid \Linner{\bar{e}_k}{y} = 0 \} \\
            &= \{ y=(y_t,y_s) \in \lorentz{m} \mid (y_s)_k=0 \},
        \end{aligned}
    \end{equation}
    where $\bar{e}_k=[0,e_k^\top]^\top \in T_{\Lzero}\lorentz{m}$. From \cref{hbnn:app:tab:distance-comparison}, the associated signed point-to-hyperplane distance is
    \begin{equation}
    \begin{aligned}
        \bar{\dist}\left(y, H_{\bar{e}_k,e}\right) 
        &= \sign \left(\Linner{\bar{e}_k}{y}\right) \dist\left(y, H_{\bar{e}_k,e}\right) \\
        &= \frac{1}{\sqrt{-K}} \sinh^{-1} \left( \sqrt{-K} (y_s)_k \right).
    \end{aligned}
    \end{equation}

    Equating $\bar{\dist}\left(y, H_{\bar{e}_k,e}\right)$ with $u_k(x)$ from \cref{hbnn:eq:hyp-fc-p2h} gives
    \begin{equation}\label{hbnn:app:eq:lorentz-fc-prf-1}
        \sinh^{-1} \left( \sqrt{-K} (y_s)_k \right) = \sqrt{-K} u_k(x), \quad 1 \le k \le m.
    \end{equation}
    Applying $\sinh(\cdot)$ to both sides of \cref{hbnn:app:eq:lorentz-fc-prf-1} yields
    \begin{equation}
        (y_s)_k = \frac{1}{\sqrt{-K}} \sinh \left( \sqrt{-K} u_k(x) \right), \quad 1 \le k \le m.
    \end{equation}
    Stacking the coordinates gives
    \begin{equation}
        y_s = \frac{1}{\sqrt{-K}} \sinh \left( \sqrt{-K} u(x) \right), \quad u(x)=\left(u_1(x),\ldots,u_m(x)\right)^{\top}.
    \end{equation}

    Since $y \in \lorentz{m}$, the Lorentz constraint $\Linner{y}{y}=1/K$ implies $-y_t^{2}+\norm{y_s}^{2}=1/K$. Taking the positive time component yields
    \begin{equation}
        y_t = \sqrt{\frac{1}{-K}+\norm{y_s}^{2}}.
    \end{equation}
    Combining the expressions for $y_t$ and $y_s$ proves the claim.
\end{proof}

\subsection[Proof of \cref{hbnn:thm:limits-bfc}]{Proof of \cref{hbnn:thm:limits-bfc}}
\label{hbnn:app:prf:thm:limits-bfc}
\linkofproof{hbnn:thm:limits-bfc}

\begin{proof}
Set $K=-\kappa^{2}$ with $\kappa>0$.

\mypara{Poincaré Case.} Recall that
\begin{equation}
    y=\frac{\omega}{1+\sqrt{1 + \kappa^{2}\norm{\omega}^{2}}}, \quad \omega_k=\frac{\sinh\left(\kappa u_k(x)\right)}{\kappa}.
\end{equation}
For any bounded scalar $z$,
\begin{equation}
    \sinh\left(\kappa z\right)=\kappa z+\frac{\kappa^{3} z^{3}}{3!}+O\left(\kappa^{5}\right)
    \ \Rightarrow\ 
    \frac{\sinh\left(\kappa z\right)}{\kappa}=z+\frac{\kappa^{2} z^{3}}{6}+O\left(\kappa^{4}\right).
\end{equation}
Applying this to $z=u_k(x)$ yields
\begin{equation}
    \omega_k=u_k(x)+\frac{\kappa^{2}}{6} u_k(x)^{3}+O\left(\kappa^{4}\right)=u_k(x)+O\left(\kappa^{2}\right).
\end{equation}
Thus, $\omega=u(x)+O\left(\kappa^{2}\right)$ and $\norm{\omega}^{2}=\norm{u(x)}^{2}+O\left(\kappa^{2}\right)$.

For the denominator,
\begin{equation}
    \sqrt{1+\kappa^{2}\norm{\omega}^{2}}=1+\frac{1}{2}\kappa^{2}\norm{\omega}^{2}+O\left(\kappa^{4}\right),
\end{equation}
which gives
\begin{equation}
    1+\sqrt{1+\kappa^{2}\norm{\omega}^{2}}=2+\frac{1}{2}\kappa^{2}\norm{\omega}^{2}+O\left(\kappa^{4}\right).
\end{equation}
Taking the reciprocal produces
\begin{equation}
    \frac{1}{1+\sqrt{1+\kappa^{2}\norm{\omega}^{2}}}=\frac{1}{2}+O\left(\kappa^{2}\right).
\end{equation}

Multiplying with $\omega=u(x)+O\left(\kappa^{2}\right)$ gives
\begin{equation}
    y=\frac{1}{2} u(x)+O\left(\kappa^{2}\right).
\end{equation}
By \cref{hbnn:thm:limits-bmlr}, $u_k(x)\to 2\alpha_k\inner{v_k}{x}+b_k$, hence
\begin{equation}
    y_k \to \alpha_k\inner{v_k}{x}+\frac{1}{2} b_k.
\end{equation}

\mypara{Lorentz Case.} Recall that
\begin{equation}
    y_s=\frac{1}{\kappa} \sinh\left(\kappa u(x)\right), \quad y_t=\sqrt{\frac{1}{\kappa^{2}}+\norm{y_s}^{2}}.
\end{equation}
Using the same expansion as above,
\begin{equation}
    y_{s,k}=\frac{1}{\kappa} \left(\kappa u_k(x)+\frac{\kappa^{3}}{3!} u_k(x)^{3}+O\left(\kappa^{5}\right)\right)=u_k(x)+O\left(\kappa^{2}\right).
\end{equation}
Therefore, $y_s=u(x)+O\left(\kappa^{2}\right)$ and $\norm{y_s}^{2}=\norm{u(x)}^{2}+O\left(\kappa^{2}\right)$.

Factor out $\kappa^{-1}$ and expand the square root:
\begin{equation}
\begin{aligned}
    y_t&=\frac{1}{\kappa} \sqrt{1+\kappa^{2}\norm{y_s}^{2}} \\
        &=\frac{1}{\kappa} \left(1+\frac{1}{2}\kappa^{2}\norm{y_s}^{2}+O\left(\kappa^{4}\right)\right) \\
        &=\frac{1}{\kappa}+\frac{\kappa}{2}\norm{y_s}^{2}+O\left(\kappa^{3}\right) \\
        &=\frac{1}{\kappa}+O\left(\kappa\right) \to \infty.
\end{aligned}
\end{equation}

By \cref{hbnn:thm:limits-bmlr}, $u_k(x)\to \alpha_k\inner{v_k}{x_s}+b_k$. Using the spatial expansion yields
\begin{equation}
    (y_s)_k \to \alpha_k\inner{v_k}{x_s}+b_k.
\end{equation}
This completes the proof.
\end{proof}

\section{Full-Rank Correlation Networks}
\label{app:cornet-proofs}

\linkofproof{cornet:thm:flat-mlr}
\subsection[Proof of \cref{cornet:thm:flat-mlr}]{Proof of \cref{cornet:thm:flat-mlr}}

We first prove a lemma for MLRs on general isometric manifolds, of which this theorem is a specific case. Notably, the result and proof can be readily extended to the case where $\bbR{m}$ is endowed with an arbitrary inner product.

\begin{parislemma} [Isometric Riemannian MLRs] \label{cornet:app:lem:isometric_MLR}
    Given $m$-dimensional Riemannian manifolds $\left(\widetilde{\calM}, g^{\widetilde{\calM}}\right)$ and $\left(\calM,g^{\calM} \right)$ with a Riemannian isometry $\phi: \widetilde{\calM} \rightarrow \calM$, their origins are $E \in \widetilde{\calM}$ and $\phi(E) \in \calM$. The Riemannian MLR over $\widetilde{\calM}$ for the input $X \in \widetilde{\calM}$ of each class $k=1, \cdots, C$ can be calculated by the one over $\calM$:
    \begin{equation}
        v^{\widetilde{\calM}} _{k}(X; Z_k, \gamma_k)
        = v^{\calM}_{k} ( \phi(X); \phi_{*,E} (Z_k), \gamma_k),
    \end{equation}
    with $\gamma_{k} \in \bbRscalar$, $Z_k \in T_{E} \widetilde{\calM} \cong \bbR{m}$, and $\phi_{*,E}: T_{E} \widetilde{\calM} \rightarrow T_{\phi(E)} \calM$ as the differential map. Here, $v^{\widetilde{\calM}} _{k}$ and $v^{\calM} _{k}$ are the specific realizations of the Riemannian MLR reviewed in \cref{rmlr:subsec:re_exist_MLR} over $\widetilde{\calM}$ and $\calM$, respectively.
\end{parislemma}

\begin{proof}
    We omit the subscript $k$ in $A_k$ and $P_k$ for simplicity. We denote $\widetilde{\Gamma}$, $\widetilde{\rielog}$, $\inner{\cdot}{\cdot}_{P}$, $\norm{\cdot}_{P}$, $\widetilde{\dist}(X, \widetilde{H}_{A,P})$, $\widetilde{H}_{A,P}$ as the parallel transport along the geodesic, Riemannian logarithm, Riemannian metric, the induced norm,  margin distance and hyperplane over $\widetilde{\calM}$, while the counterparts over $\calM$ are denoted as $\Gamma$, $\rielog$, $\inner{\cdot}{\cdot}_{\phi(P)}$, $\norm{\cdot}_{\phi(P)}$, $\dist$, and $H$, respectively.

    From the isometry, we have 
    \begin{align}
        \norm{A}_{P} 
        &= \norm{\phi_{*,P} (A) }_{\phi(P)}, \\
        \label{cornet:app:eq:mlr_inner_product_isometry}
        \inner{\widetilde{\rielog}_{P}(X)}{A}_{P} 
        &= \inner{\rielog_{\phi(P)} (\phi(X))}{\phi_{*,P} (A)}_{\phi (P)}.
    \end{align}
    The above equations imply
    \begin{equation}
        \label{cornet:app:prf:eq:H_widetilde_H}
        \phi \left( \widetilde{H}_{A,P} \right) = H_{\phi_{*,P} (A),\phi(P)}.
    \end{equation}
    Denoting $\mathcal{H} = H_{\phi_{*,P} (A),\phi(P)}$, we have the following for the margin distance
    \begin{equation} \label{cornet:app:prf:margin_dist_isometry}
        \begin{aligned}
            \widetilde{\dist} (X, \widetilde{H}_{A, P}) 
            &= \inf _{Q \in \widetilde{H}_{A, P}} \widetilde{\dist}(X, Q)\\
            &\stackrel{(1)}{=} \inf _{Q \in \widetilde{H}_{A, P}} \dist(\phi(X), \phi(Q))\\
            &\stackrel{(2)}{=} \inf _{R \in \mathcal{H}} \dist( \phi(X), R)\\
            &\stackrel{(3)}{=} \dist ( \phi(X), \mathcal{H}).
        \end{aligned}
    \end{equation}
    The above comes from the following.
    \begin{enumerate}
        \item 
        Isometry.
        \item 
        \cref{cornet:app:prf:eq:H_widetilde_H}.
        \item 
        Definition of margin distance.
    \end{enumerate}

    Combining the above, we have    
    \begin{equation} \label{cornet:app:prf:eq:v_isometry}
        \begin{aligned}
            & v^{\widetilde{\calM}}(X; P,A) \\
            &= \sign(\langle A, \widetilde{\rielog}_{P}(X) \rangle_{P})\|A\|_{P} \widetilde{\dist} (X, \widetilde{H}_{A, P})\\
            &= \sign \left( \inner{\rielog_{\phi(P)} (\phi(X))}{\phi_{*,P} (A)}_{\phi (P)} \right) 
            \norm{\phi_{*,P} (A) }_{\phi(P)}
            \dist (\phi(X), H_{\phi_{*,P} (A),\phi(P)})\\
            &= v^{\calM} \left( \phi(X); \phi(P), \phi_{*,P} (A) \right). 
        \end{aligned}
    \end{equation}

    Finally, let us further consider trivialization. By isometry, we have the following:
    \begin{equation}
        \begin{aligned}
            A 
            &= \widetilde{\Gamma}_{ E \rightarrow P } (Z) \\
            &= \phi_{*,P}^{-1} \left( \pt{\phi(E)}{\phi(P)} (\phi_{*,E} (Z)) \right),
        \end{aligned}
    \end{equation}
    \begin{equation}
        \begin{aligned}
            P 
            &= \widetilde{\rieexp}_{E} \left(\gamma [Z]\right) \\
            &= \phi^{-1} \left( \rieexp_{\phi(E)} ( \gamma [\phi_{*,E} (Z)])  \right).\\
        \end{aligned}
    \end{equation}
    Then, we have 
    \begin{align}
        \phi_{*,P}(A) &= \pt{\phi(E)}{\phi(P)} (\phi_{*,E} (Z)),\\
        \phi(P) &= \rieexp_{\phi(E)} \left( \gamma [\phi_{*,E} (Z)]\right).
    \end{align}
    Putting the above two equations into \cref{cornet:app:prf:eq:v_isometry}, we have
    \begin{equation}
        \begin{aligned}
            v^{\widetilde{\calM}}(X; Z,\gamma)
            &=v^{\widetilde{\calM}}(X; P,A) \\ 
            &= v^{\calM} \left( \phi(X); \phi(P), \phi_{*,P} (A) \right)\\
            &= v^{\calM} \left( \phi(X); \rieexp_{\phi(E)} ( \gamma [\phi_{*,E} (Z)]), \pt{\phi(E)}{\phi(P)} (\phi_{*,E} (Z)) \right)\\
            &= v^{\calM}_{k} ( \phi(X); \phi_{*,E} (Z_k), \gamma_k).
        \end{aligned}
    \end{equation}
\end{proof}

\cref{cornet:thm:flat-mlr} is a special case of \cref{cornet:app:lem:isometric_MLR} and can be readily proven accordingly.

\begin{proof}[Proof of \cref{cornet:thm:flat-mlr}]
    \mypara{MLR.} In Euclidean space $\bbR{m}$, simple computations show that the Riemannian MLR reviewed in \cref{rmlr:subsec:re_exist_MLR} becomes \cref{spdmlr:eq:EMLR_reform_start}, where the latter is equal to $\inner{a_k}{x-p_k}$. Based on \cref{cornet:app:lem:isometric_MLR}, we have
    \begin{equation}
        \label{cornet:app:eq:v_k_flat_mlr}
        \begin{aligned}
            v_{k}(X; Z_k,\gamma_k)
            &= v^{\bbR{m}}_{k} ( \phi(X); \phi_{*,E} (Z_k), \gamma_k),\\
            &= \inner{\phi(X) - \gamma_k [\phi_{*,E} (Z_k)] }{\phi_{*,E} (Z_k)}\\
            &= \left \langle \phi(X), \phi_{*,E}(Z_k) \right \rangle - \gamma_k \norm{\phi_{*,E}(Z_k)},
        \end{aligned}
    \end{equation}    

    \mypara{Margin Hyperplane.} In Euclidean space $\bbR{m}$, the Riemannian margin hyperplane becomes the Euclidean one, which is parameterized by $\inner{a_k}{x-p_k}=0$. Together with \cref{cornet:app:eq:v_k_flat_mlr}, the results can be easily obtained.
\end{proof}

\linkofproof{cornet:prop:diff-at-I}
\subsection[Proof of \cref{cornet:prop:diff-at-I}]{Proof of \cref{cornet:prop:diff-at-I}}

\begin{proof}
    First, we have the following:
    \begin{align}
        \Theta(I) &= I,\\
        \chol(I) &= I,\\
        \log_{*,I} (V) &= V, \quad \forall V \in \hol{n},\\
        \log_{*,I} (V) &= V, \quad \forall V \in \LTzero{n},\\
        \dstar(I) &= I.
    \end{align}
    Putting the above into the differential formulas collected in \cref{sec:ch2-full-rank-correlation-manifolds}, one can directly get the result w.r.t. ECM, LECM, and OLM. For LSM, based on \cref{sec:ch2-full-rank-correlation-manifolds}, we have
    \begin{equation}
        \begin{aligned}
            \logscaled_{*,I}(V) 
            &=\log_{*,\Sigma}  \left(\Delta V \Delta+\frac{1}{2}\left(V^0 \Sigma+\Sigma V^0\right)\right)\\
            &\stackrel{(1)}{=} V + \frac{1}{2}\left(V^0 + V^0\right) \\
            &\stackrel{(2)}{=} V - \diag \left( V \vecone \right).
        \end{aligned}
    \end{equation}
    The above comes from the following.
    \begin{enumerate}
        \item 
        $\Sigma=\Delta=I$
        \item 
        \begin{equation}
            \begin{aligned}
                V^0
                &= -2 \diag \left(\left(I_n+\Sigma\right)^{-1} \Delta V \Delta \vecone \right) \\
                &= - \diag \left( V \vecone \right)
            \end{aligned}
        \end{equation}
    \end{enumerate}
\end{proof}

\linkofproof{cornet:thm:flat-cor-fc}
\subsection[Proof of \cref{cornet:thm:flat-cor-fc}]{Proof of \cref{cornet:thm:flat-cor-fc}}
\label{cornet:app:subsec:prf:flat_cor_fc}

Let $d_n=\frac{n(n-1)}{2}$ and $d_m=\frac{m(m-1)}{2}$ be the manifold dimensions of $\cor{n}$ and $\cor{m}$, respectively. We have the following general results.

\begin{parislemma} \label{cornet:app:lem:flat-fc}
    Let $\left( \cor{n}, g^n \right)$ be isometric to $\bbR{d_n}$ by the diffeomorphism
    \begin{equation}
        \phi_n: \cor{n} \rightarrow \bbR{d_n},
    \end{equation}
    and let $\left( \cor{m}, g^m \right)$ be isometric to $\bbR{d_m}$ by the diffeomorphism
    \begin{equation}
        \phi_m: \cor{m} \rightarrow \bbR{d_m}.
    \end{equation}
    The diffeomorphism satisfies
    \begin{equation}
        I_n = \phi_n^{-1} (\zerovec_{d_n}), \qquad I_m = \phi_m^{-1}(\zerovec_{d_m}).
    \end{equation}
    The correlation FC layer $\calF: \cor{n} \rightarrow \cor{m}$ for the input $X \in \cor{n}$ is
    \begin{equation} \label{cornet:app:eq:flat_v_k_isometry}
        Y= \phi_m^{-1}\left( \sum_{i=1}^{d_m}  v_i(X) e_i \right),
    \end{equation}
    where $\{ e_i \}_{i=1}^{d_m}$ is the canonical orthonormal basis over $\bbR{d_m}$ with $e_i=\left(\delta_{i k}\right)_{k=1}^{d_m}$ for each $i$. Here, $\{v_i(X) \}_{i=1}^{d_m}$ is given by \cref{cornet:thm:flat-mlr}:
    \begin{equation}
        v_i(X) = \left\langle \phi_n(X), (\phi_n)_{*,I_n}(Z_i) \right\rangle -\gamma_{i} \norm{(\phi_n)_{*,I_n}(Z_i)},
    \end{equation}
    with $Z_{i} \in T_{I_n}\cor{n}\cong\bbR{d_n}$ and $\gamma_{i} \in \bbRscalar$ as the FC parameters.
\end{parislemma}
\begin{proof}[Proof of \cref{cornet:app:lem:flat-fc}]
    Let $\{O_k=(\phi_m)_{*,I_m}^{-1}(e_k)\}_{k=1}^{d_m}$. Then $\{O_k\}_{k=1}^{d_m}$ is an orthonormal basis over $T_{I_m}\cor{m}$.
    
    The LHS of \cref{cornet:eq:riem-fc} is
    \begin{equation} \label{cornet:app:eq:prf-lhs-flat-fc}
        \begin{aligned}
            & \sign \left( \inner{\rielog _{I_m}(Y)}{O_k}_{I_m} \right) \dist (Y, H _{O_k, I_m} ) \\
            &\stackrel{(1)}{=} \sign \left( \inner{(\phi_m)_{*,I_m}^{-1} \phi_m(Y)}{O_k}_{I_m} \right)  \dist (Y, H _{O_k, I_m}) \\
            &\stackrel{(2)}{=} \sign \left( \inner{\phi_m(Y)}{e_k} \right) \dist (Y, H _{O_k, I_m})  \\
            &\stackrel{(3)}{=} \sign \left( \inner{\phi_m(Y)}{e_k} \right) \dist (\phi_m(Y), H _{e_k, \zerovec_{d_m}})  \\
            &= \left( \phi_m(Y) \right)_k,  \\
        \end{aligned}
    \end{equation}
    where (1)--(2) come from the isometry, and (3) comes from \cref{cornet:app:prf:margin_dist_isometry}.

    The RHS of \cref{cornet:eq:riem-fc} can be implied by \cref{cornet:thm:flat-mlr}.
\end{proof}

\cref{cornet:app:lem:flat-fc} can be naturally extended to the cases where the inner products of $\bbR{d_n}$ and $\bbR{d_m}$ are not canonical.
\begin{parislemma} \label{cornet:app:lem:flat-fc_general}
    Following all the notation in \cref{cornet:app:lem:flat-fc}, we further assume that the inner products $Q^n(\cdot,\cdot)$ over $\bbR{d_n}$ and $Q^m(\cdot,\cdot)$ over $\bbR{d_m}$ are not necessarily canonical. In addition, $f: (\bbR{d_m}, Q^m(\cdot,\cdot)) \rightarrow (\bbR{d_m}, \inner{\cdot}{\cdot} )$ is a linear isometry to the canonical inner product. Then, we have
    \begin{align}
        Y &= \phi_m^{-1} \circ f^{-1} \left( \sum_{i=1}^{d_m}  v_i(X) e_i \right),\\
        v_i(X) &= Q^n\left( \phi_n(X), (\phi_n)_{*,I_n}(Z_i) \right) - \gamma_{i} \norm{(\phi_n)_{*,I_n}(Z_i)}^{Q^n},
    \end{align}
    where $\norm{\cdot}^{Q^n}$ is the norm induced by $Q^n$.
\end{parislemma}
\begin{proof}[Proof of \cref{cornet:app:lem:flat-fc_general}]
    First, we denote
    \begin{align}
        \psi^m = f \circ \phi_m: \left( \cor{m}, g^m \right) \rightarrow (\bbR{d_m}, \inner{\cdot}{\cdot}).
    \end{align}
    Note that the differential of any linear map between vector spaces is itself. The rest of the proof is identical to that of \cref{cornet:app:lem:flat-fc}.
\end{proof}

Now, we present the proof of \cref{cornet:thm:flat-cor-fc}.

\begin{proof}[Proof of \cref{cornet:thm:flat-cor-fc}]
    As ECM, LECM, OLM, and LSM are pullback metrics from Euclidean spaces, we resort to \cref{cornet:app:lem:flat-fc} and its extension \cref{cornet:app:lem:flat-fc_general}. Denoting the zero matrix as $\bbzero$, we have the following:
    \begin{align}
        \phi^{\EC}(I_n)=\log \circ \Theta(I_n) &=\bbzero \in \LTzero{n}, \\
        \offlog(I_n) &=\bbzero \in \hol{n}, \\
        \logscaled(I_n) &=\bbzero \in \rzero{n}.
    \end{align}
    Therefore, the identity matrix is indeed the origin defined in \cref{cornet:app:lem:flat-fc}.
    
    \begin{figure}[t]
    \centering
    \includegraphics[width=\linewidth,trim={0cm 0cm 0cm 0cm}]{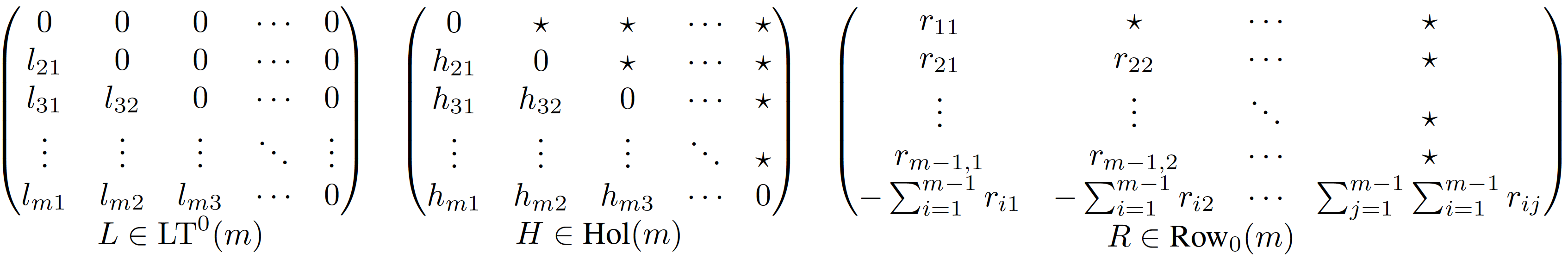}
    \caption{Illustration of the Euclidean spaces $\LTzero{m}, \hol{m}$ and $\rzero{m}$, where $\star$ can be obtained by symmetry.
    }
    \label{cornet:app:fig:euc_spaces}
    \end{figure}
    
    Recalling \cref{cornet:app:lem:flat-fc}, the prototype space is the vector space with the standard vector inner product. Obviously, $\LTzero{m}$, $\hol{m}$, and $\rzero{m}$ are linearly isomorphic to $\bbR{\nicefrac{m(m-1)}{2}}$. As shown in \cref{cornet:app:fig:euc_spaces}, each $L \in \LTzero{m}$ can be identified with a vector of its lower triangular part. Besides, $\LTzero{m}$ with the canonical matrix inner product is identified with $\bbR{\frac{m(m-1)}{2}}$ with standard vector inner product. Therefore, the basis over $\LTzero{m}$ corresponding to the canonical orthonormal basis over $\bbR{\nicefrac{m(m-1)}{2}}$ is
    \begin{equation}
        (\LTzero{m},\inner{\cdot}{\cdot}): U^{\LTzero{m}}_{ij} = E_{ij}, \quad 1 \leq j < i \leq m,
    \end{equation}
    where $E_{ij} \in \bbR{m \times m}$ is the standard basis matrix, with the $(k,l)$-th element defined as 
    \begin{equation}
        \left(E_{i j}\right)_{k l}= 
        \begin{cases}
        1 & \text { if } k=i \text { and } l=j, \\
        0 & \text { otherwise. }
        \end{cases}
    \end{equation}
    Without loss of generality, we identify $( \LTzero{m}, \inner{\cdot}{\cdot} )$ with $( \bbR{\frac{m(m-1)}{2}}, \inner{\cdot}{\cdot} )$, and refer to $\{ E_{ij} \}_{1 \leq j < i \leq m}$ as the canonical orthonormal basis. 
    
    However, $\{ E_{ij} \}$ is neither a canonical orthonormal basis nor even orthonormal for $\hol{m}$ and $\rzero{m}$ under the standard matrix inner product. According to \cref{cornet:app:lem:flat-fc_general}, we only need to find the linear isometry that maps these two spaces into $( \LTzero{m}, \inner{\cdot}{\cdot} )$. By \cref{cornet:app:fig:euc_spaces}, we have the following linear isometries to pull back these two inner products to the standard ones over $\LTzero{m}$:
    \begin{equation}
    \begin{aligned}
        f_{\hol{m} \rightarrow \LTzero{m}} :
        & ( \hol{m}, \inner{\cdot}{\cdot} ) \rightarrow ( \LTzero{m},\inner{\cdot}{\cdot} ), \\
        & \hol{m} \ni H \longmapsto \sqrt{2} \lfloor H\rfloor \in \LTzero{m},\\
        f_{\rzero{m} \rightarrow \LTzero{m}} :
        & ( \rzero{m}, \inner{\cdot}{\cdot} ) \rightarrow ( \LTzero{m},\inner{\cdot}{\cdot} ), \\
        & \rzero{m} \ni R \longmapsto \sqrt{6} \lfloor\widetilde{R}\rfloor + \sqrt{3} \bbD (\widetilde{R}) \in \LTzero{m},
    \end{aligned}
    \end{equation}
    where $\widetilde{R} \in \sym{m-1}$ is the leading principal submatrix of order $m-1$ of $R$. The bases $f_{\hol{m} \rightarrow \LTzero{m}}^{-1} (\{E_{ij}\})$ and $f_{\rzero{m} \rightarrow \LTzero{m}} ^{-1} (\{E_{ij}\})$ are as follows:
    \begin{align}
        ( \hol{m},\inner{\cdot}{\cdot} ) &: U^{\hol{m}}_{ij}= \frac{E_{ij}+ E_{ji}}{\sqrt{2}}, \quad 1 \leq j < i \leq m \\
        ( \rzero{m},\inner{\cdot}{\cdot} ) &: U^{\rzero{m}}_{ij}=
        \begin{cases}
        \frac{E_{ii}-E_{im}-E_{mi}}{\sqrt{3}}, & \text{if } 1 \leq i < m \\
        \frac{E_{ij}+ E_{ji}-E_{mi}-E_{im}-E_{mj}-E_{jm}}{\sqrt{6}}, & \text{if } 1 \leq j < i < m
        \end{cases}
    \end{align}

    Putting the required diffeomorphisms and $v^g_{ij}$ in \cref{cornet:thm:cormlr}  into \cref{cornet:app:lem:flat-fc_general} for ECM, LECM, OLM, and LSM, the corresponding FC layers can be readily obtained. 
    
\end{proof}

\linkofproof{cornet:prop:isometry-hs-pball}
\subsection[Proof of \cref{cornet:prop:isometry-hs-pball}]{Proof of \cref{cornet:prop:isometry-hs-pball}}
\label{cornet:app:subsec:prf:isometry_hs_pball}

\begin{proof}
    First, we review the isometries between the open hemisphere and the Lorentz model \citep[Eqs.~(4.1)--(4.2)]{thanwerdas2022theoretically}, and the one between the Poincaré ball and the Lorentz model \citep[Sec.~2.1]{skopek2020mixed}:
    \begin{align}
    \psi _{\hs{n} \rightarrow \unitlorentz{n}} &: \left(x_1, \ldots, x_{n+1}\right)^\top \in \hs{n} \longmapsto \frac{1}{x_{n+1}}\left(1,x_1, \ldots, x_n\right)^\top \in \unitlorentz{n}, \\
    \psi _{\unitlorentz{n} \rightarrow \hs{n}} &: \left(y_t,y_s^\top\right)^\top \in \unitlorentz{n} \longmapsto \frac{1}{y_t}\left(y_s^\top,1\right)^\top \in \hs{n},\\
    \psi _{\unitlorentz{n} \rightarrow \unitpball{n}} &: \left( x_t,x_s^\top \right)^\top \in \unitlorentz{n} \longmapsto \frac{x_s}{1 + x_t} \in \unitpball{n}, \\
    \psi _{ \unitpball{n} \rightarrow \unitlorentz{n} } &: y \in \unitpball{n} \longmapsto
    \left(\frac{1+ \norm{y}^2}{1 - \norm{y}^2}, \frac{2 y^\top}{1- \norm{y}^2}\right)^\top
    = \frac{1}{1- \norm{y}^2}
    \left(
    \begin{array}{c}
        1+ \norm{y}^2 \\
        2y
    \end{array}\right)
    \in \unitlorentz{n}.
    \end{align}
    For any $(x^\top, x_{n+1})^\top \in \hs{n}$ and $y \in \unitpball{n}$, we have
    \begin{equation}
        \begin{aligned}
            \psi _{\hs{n} \rightarrow \unitpball{n}} \left( \left(
            \begin{array}{c}
                 x  \\
                 x_{n+1} 
            \end{array} \right) \right)
            &=\psi _{\unitlorentz{n} \rightarrow \unitpball{n}} \circ \psi _{\hs{n} \rightarrow \unitlorentz{n}} \left( \left(
            \begin{array}{c}
                 x  \\
                 x_{n+1} 
            \end{array} \right) \right) \\
            &= \psi _{\unitlorentz{n} \rightarrow \unitpball{n}} \left( \frac{1}{x_{n+1}}\left(
            \begin{array}{c}
                 1  \\
                 x
            \end{array} \right) \right)\\
            &= \frac{x}{x_{n+1}} \frac{1}{1+ \frac{1}{x_{n+1}}}\\
            &= \frac{x}{1 + x_{n+1}}
        \end{aligned}
    \end{equation}
    \begin{equation}
        \begin{aligned}
            \psi _{\unitpball{n} \rightarrow \hs{n}} \left( y \right)
            &= \psi _{\unitlorentz{n} \rightarrow \hs{n}} \circ \psi _{\unitpball{n} \rightarrow \unitlorentz{n}} \left( y \right) \\
            &= \psi _{\unitlorentz{n} \rightarrow \hs{n}} \left(
            \frac{1}{1- \norm{y}^2}
            \left(
            \begin{array}{c}
                1+ \norm{y}^2 \\
                2y
            \end{array}\right) \right) \\ 
            &= \frac{1}{1+\norm{y}^2} \left(
                \begin{array}{c}
                     2y \\
                      1-\norm{y}^2
                \end{array} \right)
        \end{aligned}
    \end{equation}
\end{proof}

\linkofproof{cornet:prop:gradients-d-plus}
\subsection[Proof of \cref{cornet:prop:gradients-d-plus}]{Proof of \cref{cornet:prop:gradients-d-plus}}

\begin{proof}
   We denote $D= \dplus(H)$. By \cref{sec:ch2-full-rank-correlation-manifolds}, we have
    \begin{equation}
        \begin{aligned}
            dY &= dD + dH \\
            dD &= -\diag \left( \left(H^0 \right)^{-1} \bbD \left( \exp_{*,Y} (dH) \right) \vecone \right).
        \end{aligned}
    \end{equation}

    Following \citet{ionescu2015training}, we denote the inner product $\inner{\cdot}{\cdot}$ as $\cdot : \cdot$ for simplicity. By the invariance of differential and properties of trace \citep[Eqs.~67--72]{ionescu2015training}, we have the following:
    \begin{equation} \label{cornet:sec:ch2-full-rank-correlation-manifolds_derivation}
        \begin{aligned}
            \frac{\partial l}{\partial Y} : dY 
            &= \frac{\partial l}{\partial Y} : dD + \frac{\partial l}{\partial Y} : dH \\
            &= \frac{\partial l}{\partial Y} : -\diag \left( (H^0)^{-1} \bbD \left( \exp_{*,Y} (dH) \right) \vecone \right) + \frac{\partial l}{\partial Y} : dH \\
            &\stackrel{(1)}{=} \tr \left( -\diagvec \left( \frac{\partial l}{\partial Y} \right)^\top (H^0)^{-1} \bbD \left( \exp_{*,Y} (dH) \right) \vecone \right) + \frac{\partial l}{\partial Y} : dH \\
            &\stackrel{(2)}{=} \tr \left( -\left[ \vecone  \diagvec \left( \frac{\partial l}{\partial Y} \right)^\top (H^0)^{-1} \right] \bbD \left( \exp_{*,Y} (dH) \right) \right) + \frac{\partial l}{\partial Y} : dH \\
            &= -(H^0)^{-1} \diagvec \left( \frac{\partial l}{\partial Y} \right) \vecone^\top : \bbD \left( \exp_{*,Y} (dH) \right) + \frac{\partial l}{\partial Y} : dH \\
            &= -\bbD \left( (H^0)^{-1} \diagvec \left( \frac{\partial l}{\partial Y} \right) \vecone^\top \right) : \exp_{*,Y} (dH) + \frac{\partial l}{\partial Y} : dH \\
            &\stackrel{(3)}{=} \left[ \frac{\partial l}{\partial Y} - \exp_{*,Y} \left( \bbD \left( (H^0)^{-1} \diagvec \left( \frac{\partial l}{\partial Y} \right) \vecone^\top \right) \right) \right] : dH \\
            &\stackrel{(4)}{=} \off \left[ \frac{\partial l}{\partial Y} - \exp_{*,Y} \left( \bbD \left( (H^0)^{-1} \diagvec \left( \frac{\partial l}{\partial Y} \right) \vecone^\top \right) \right) \right] : dH
        \end{aligned}
    \end{equation}
    The above comes from the following.
    \begin{enumerate}
        \item 
        \begin{align}
            A : \diag(b) = \diagvec(A) : b, \quad &\forall A \in \bbR{n \times n}, b \in \bbR{n}, \\
            a : b = a^\top b = \tr(a^\top b), \quad &\forall a,b \in \bbR{n}.
        \end{align}
        \item 
        Cyclic property of the trace for matrices $A, B$, and $C$ of compatible dimensions: $\tr(ABC)=\tr(CAB)$.     
        \item 
        For any $A \in \sym{n}$ and $S \in \sym{n}$, write $Y=U\Delta U^\top$ and let $L=L_{\exp}$ be the Loewner matrix in \cref{eq:ch2-loewner-matrix}. By the Daleckii--Krein formula in \cref{eq:ch2-symmetric-matrix-function-differential} and the properties of trace, we have
        \begin{equation}
            \begin{aligned}
            A: \exp_{*,Y}(S)
            &= A: U\left(L \circledast \left(U^{\top} S U\right)\right) U^{\top} \\
            &= U\left(L \circledast U^{\top} A U\right) U^{\top}: S \\
            &= \exp_{*,Y} (A): S.
            \end{aligned}
        \end{equation}
        \item 
        $H$ has zero diagonal elements.
    \end{enumerate}

    The invariance of the first-order differential gives
    \begin{equation}
        \frac{\partial l}{\partial Y} : d Y = \frac{\partial l}{\partial H} : d H.
    \end{equation}
    By the last equation in \cref{cornet:sec:ch2-full-rank-correlation-manifolds_derivation}, we can obtain $\frac{\partial l}{\partial H}$.
    
\end{proof}

\linkofproof{cornet:prop:gradients-d-star}
\subsection[Proof of \cref{cornet:prop:gradients-d-star}]{Proof of \cref{cornet:prop:gradients-d-star}}

\begin{proof}
    Denoting by $f: \cor{n} \rightarrow \rone{n}$ the map $f(C)=\dstar(C) C \dstar(C)=\Sigma$, we have
    \begin{equation}
        \logscaled_{*, C} = \log_{*, \Sigma} \circ f_{*, C}.
    \end{equation}
    Combining with the differential of $\logscaled$ shown in \cref{sec:ch2-full-rank-correlation-manifolds}, we have the following differential equation:
    \begin{equation} \label{cornet:app:eq:diff_Sigma}
        d \Sigma  = \Delta d C \Delta - \left(V^0 \Sigma+\Sigma V^0 \right),
    \end{equation}
    with $V^0= \diag \left(\left(I_n+\Sigma\right)^{-1} \Delta dC \Delta \vecone \right)$. Similarly to \cref{cornet:prop:gradients-d-plus}, we have the following:
    \begin{equation}
        \label{cornet:app:eq:diff_C_derivation}
        \begin{aligned}
            \frac{\partial l}{\partial \Sigma} : d \Sigma
            &= \frac{\partial l}{\partial \Sigma} : \left( \Delta d C \Delta  - \left(V^0 \Sigma+\Sigma V^0 \right) \right)\\
            &= \left( \Delta \frac{\partial l }{ \partial \Sigma} \Delta \right) : d C   - \frac{\partial l}{\partial \Sigma} : \left(V^0 \Sigma+\Sigma V^0 \right) \\
            &= \left( \Delta \frac{\partial l }{ \partial \Sigma} \Delta \right) : d C - \left( \frac{\partial l}{\partial \Sigma} \Sigma + \Sigma \frac{\partial l}{\partial \Sigma} \right) :  \diag \left(\left(I_n+\Sigma\right)^{-1} \Delta dC \Delta \vecone \right) \\
            &= \left( \Delta \frac{\partial l }{ \partial \Sigma} \Delta \right) : d C - \diagvec \left( \frac{\partial l}{\partial \Sigma} \Sigma + \Sigma \frac{\partial l}{\partial \Sigma} \right) : \left(\left(I_n+\Sigma\right)^{-1} \Delta dC \Delta \vecone \right) \\
            &= \left( \Delta \frac{\partial l }{ \partial \Sigma} \Delta \right) : d C - \tr \left(\widetilde{v}^\top \left(I_n+\Sigma\right)^{-1} \Delta dC \Delta \vecone  \right) \\
            &= \left( \Delta \frac{\partial l }{ \partial \Sigma} \Delta \right) : d C - \tr \left(\Delta \vecone \widetilde{v}^\top \left(I_n+\Sigma\right)^{-1} \Delta dC \right) \\
            &= \left( \Delta \frac{\partial l }{ \partial \Sigma} \Delta \right) : d C - \left( \Delta \left(I_n+\Sigma\right)^{-1} \widetilde{v} \vecone^\top \Delta \right) : dC \\
            &= \left(\Delta \frac{\partial l }{ \partial \Sigma} \Delta - \Delta \left(I_n+\Sigma\right)^{-1} \widetilde{v} \vecone^\top \Delta \right) : dC \\
            &= \left( \Delta \left( \frac{\partial l}{\partial \Sigma}  -(I_n+\Sigma)^{-1} \widetilde{v} \vecone^{\top} \right) \Delta \right) : dC. \\
        \end{aligned}
    \end{equation}
    By imposing symmetrization, we can obtain the results.
\end{proof}

\linkofproof{cornet:thm:invariance-beta-concat}
\subsection[Proof of \cref{cornet:thm:invariance-beta-concat}]{Proof of \cref{cornet:thm:invariance-beta-concat}}
\label{cornet:app:subsec:prf:invariance_beta_concate}

As $\beta$-splitting is the inverse of $\beta$-concatenation \citep{shimizu2021hyperbolic}, we only need to show the case w.r.t. $\beta$-concatenation. Besides, it suffices to prove the 2D case, which is shown in the following lemma.

\begin{parislemma} \label{cornet:app:lem:2d_beta_concat}
    Given ${x_{ij} \in \unitpball{n_j}}$ with $i\in\{1, \dots, N_i\}$ and $j\in\{1, \dots, N_j\}$, applying the $\beta$-concatenation sequentially 2 times in the order $j \rightarrow i$ is equivalent to a single $\beta$-concatenation along all indices simultaneously.
\end{parislemma}
\begin{proof}
    Denoting $d=\sum_{j=1}^{N_j} n_j$ and $v_{ij} = \rielog_{\zerovec} (x_{ij})$, we have the following
    \begin{equation}
        \begin{aligned}
            &\rieexp_{\zerovec} \left( \concat_{i=1}^{N_i} \left( \beta_{N_i \times d} \beta^{-1}_{d} \concat_{j=1}^{N_j} \left( \beta_{d} \beta^{-1}_{n_j}  v_{ij} \right) \right) \right)\\
            &=\rieexp_{\zerovec} \left( \concat_{i=1,j=1}^{i=N_i,j=N_j} \left( \beta_{N_i \times d} \beta^{-1}_{d} \beta_{d} \beta^{-1}_{n_j} v_{ij} \right) \right)\\
            &=\rieexp_{\zerovec} \left( \concat_{i=1,j=1}^{i=N_i,j=N_j} \left( \beta_{N_i \times d} \beta^{-1}_{n_j} v_{ij} \right) \right).
        \end{aligned}
    \end{equation}
    The last line implies the claim.
\end{proof}

A special case of the above lemma is where all $n_j$ are identical.

\begin{pariscorollary} \label{cornet:app:cor:2d_beta_concat_equal}
    Given ${x_{ij} \in \unitpball{n}}$ with $i\in\{1, \dots, N_i\}$ and $j\in\{1, \dots, N_j\}$, applying the $\beta$-concatenation sequentially 2 times in the order $j \rightarrow i$ is equivalent to a single $\beta$-concatenation along all indices simultaneously.
\end{pariscorollary}

\cref{cornet:thm:invariance-beta-concat} can be obtained by \cref{cornet:app:lem:2d_beta_concat,cornet:app:cor:2d_beta_concat_equal}.

\section{Adaptive Log-Euclidean Metrics}
\label{app:alem-proofs}

\subsection[Proof of \cref{alem:thm:rethk_lem_lcm}]{Proof of \cref{alem:thm:rethk_lem_lcm}}
\linkofproof{alem:thm:rethk_lem_lcm}
\begin{proof} [Proof of \cref{alem:thm:rethk_lem_lcm}]
Let us first deal with $\biparamLEM$.
Substituting the differential of the matrix logarithm into \cref{def:ch2-riemannian-isometry} directly yields the result.

Now, let us focus on LCM.
Denote LCM, the standard Euclidean metric, and the metric on the Cholesky manifold \citep{lin2019riemannian} by $\glcm$, $\geuc$, and $\gcm$, respectively.
By \cref{tab:ch2-spd-lie-operators}, $\{\spd{n}, \glcm\}$ is isometric to $\{\chospace{n},\gcm\}$, with the Cholesky decomposition $\chol$ as an isometry.
This is exactly how \citet{lin2019riemannian} derived LCM.
So, the key point lies in the Cholesky metric $\gcm$.
Let us reveal why it is defined in this way.
In fact, $\gcm$ is derived from $\geuc$ by $\clnchart$.
Simple computations show that
\begin{equation} \label{alem:eq:diff_cho_log_chart}
    \left(\clnchart\right)_{*,L}(V) = \lfloor V \rfloor + \bbD(L)^{-1}\bbD(V),
\end{equation}
where $V \in T_L\chospace{n}$.
By \cref{alem:eq:diff_cho_log_chart}, \cref{tab:ch2-spd-lie-operators} can be rewritten as
\begin{equation}
    \gcm_L(X,Y) = \geuc\left(\left(\clnchart\right)_{*,L}(X),\left(\clnchart\right)_{*,L}(Y)\right).
\end{equation}
Therefore, $\clnchart: \chospace{n} \rightarrow \trilspace{n}$ is an isometry.
By transitivity, $\clog: \spd{n} \rightarrow \trilspace{n}$ is also an isometry.
\end{proof}

\subsection[Proof of \cref{alem:cor:biparamLEM_pem}]{Proof of \cref{alem:cor:biparamLEM_pem}}
\linkofproof{alem:cor:biparamLEM_pem}
\begin{proof}[Proof of \cref{alem:cor:biparamLEM_pem}]
    As $\bbR{n(n+1)/2} \cong \trilspace{n} \cong \sym{n}$, LCM is therefore a pullback metric from the standard Euclidean space $\sym{n}$.
    Second, any two Euclidean spaces of the same finite dimension are naturally isometric; hence, $\biparamLEM$ is also a pullback metric from the standard Euclidean space $\sym{n}$.
\end{proof}

\subsection[Proof of \cref{alem:lem:g_spd}]{Proof of \cref{alem:lem:g_spd}}
\linkofproof{alem:lem:g_spd}
\begin{proof} [Proof of \cref{alem:lem:g_spd}]
    By the definitions in \cref{alem:eq:phi_mul,alem:eq:phi_sca_mul,alem:eq:phi_innerpro,alem:eq:phi_g}, the Hilbert-space and isomorphism claims follow directly.
    It remains to establish the geometric claims.
    As every Euclidean space is an abelian Lie group, $\{\spd{n}, \phiMul \}$ is an abelian Lie group. 
    The geodesic distance in \cref{alem:eq:dist_phi_spd} also follows immediately because $\phi$ is a Riemannian isometry.
    
    We only need to prove \cref{alem:eq:gene_rie_exp_spd,alem:eq:gene_rie_log_spd,alem:eq:gene_pt_spd}.
    Note that in the Euclidean space $\sym{n}$, for any $x,y \in \sym{n}$ and tangent vector $v \in T_x\sym{n} \cong \sym{n}$, we have the following:
    \begin{align}
        \rieexp_x v &= x+v,\\
        \rielog_x y &= y-x,\\
        \pt{x}{y} v &= v.
    \end{align}
    By the isometry of $\phi$, we can readily obtain \cref{alem:eq:gene_rie_exp_spd,alem:eq:gene_rie_log_spd,alem:eq:gene_pt_spd}.
\end{proof}
\subsection[Proof of \cref{alem:props:diffeo_mlog}]{Proof of \cref{alem:props:diffeo_mlog}}
\linkofproof{alem:props:diffeo_mlog}
\begin{proof} [Proof of \cref{alem:props:diffeo_mlog}]
    Obviously, $\log_\alpha^{-1}$ is the inverse of $\log_\alpha$.
    What follows is to verify the smoothness of $\log_\alpha$ and its inverse.
    
    According to \citet[Thm.~8.9]{magnus2019matrix}, the map producing an eigenvalue or an eigenvector from a real symmetric matrix is $\cinf$.
    Recalling $\log_\alpha$ and its inverse map $\log_\alpha^{-1}$, it is obvious that they comprise arithmetic calculations or compositions of smooth maps.
    Therefore, $\log_\alpha$ is a diffeomorphism with inverse $\log_\alpha^{-1}$.
\end{proof}

\subsection[Proof of \cref{alem:thm:mlog_spd_properties}]{Proof of \cref{alem:thm:mlog_spd_properties}}
\linkofproof{alem:thm:mlog_spd_properties}
\begin{proof} [Proof of \cref{alem:thm:mlog_spd_properties}]
    This is a direct result of \cref{alem:lem:g_spd}.
\end{proof}

\subsection[Proof of \cref{alem:props:diff_mgexp_mlog}]{Proof of \cref{alem:props:diff_mgexp_mlog}}
\linkofproof{alem:props:diff_mgexp_mlog}
\begin{proof} [Proof of \cref{alem:props:diff_mgexp_mlog}]
    The differentials of $\log_\alpha^{-1}$ and $\log_\alpha$ can be derived similarly.
    In the following, we only present the process of deriving the differential of $\log_\alpha$.
    
    First, let us recall the differentials of eigenvalues and eigenvectors.
    \citet[Thm.~8.9]{magnus2019matrix} offers their Euclidean differentials, which are the exact formulations for differentials under the canonical base on SPD manifolds.
    Thus, we can readily obtain the differentials of eigenvalues and eigenvectors as follows:
    \begin{align}
        \label{alem:eq:diff_eig_value} \sigma_{*,S} (V) &= u^{\top} V u,\\
        \label{alem:eq:diff_eig_vec} u_{*,S} (V) &= (\sigma I_n- S)^{+} V u,
    \end{align}
    where $S u = \sigma u$, $u^\top u = 1$, and $(\cdot)^+$ is the Moore--Penrose inverse.

    By the RHS of \cref{alem:eq:rw_org_mlog}, the differential map of $\log_\alpha$ is
    \begin{equation} \label{alem:eq:diff_mlog}
    \begin{aligned}
        &\left(\log_\alpha\right)_{*,S}(V) \\
        &= U_{*,S}(V) \log_\alpha(\Sigma)U^\top + U \left(\log_\alpha\right)_{*,\Sigma}\left(\Sigma_{*,S}(V)\right) U^\top \\
        & + U \log_\alpha(\Sigma) U^\top_{*,S}(V) \\
        &= Q+Q^\top + U \left(\log_\alpha\right)_{*,\Sigma}\left(\Sigma_{*,S}(V)\right) U^\top,
    \end{aligned}
    \end{equation}
    where $Q= U_{*,S}(V) \log_\alpha(\Sigma)U^\top$.

    For the differential of diagonal logarithm, it is 
    \begin{equation} \label{alem:eq:diff_ln}
        \left(\log_\alpha\right)_{*,\Sigma}\left(\Sigma_{*,S}(V)\right) = A \frac{1}{\Sigma} \Sigma_{*,S}(V),
    \end{equation}
    where $A$ is defined in \cref{alem:eq:rw_mul_mlog}.

    Denote the eigenvectors and eigenvalues of $S=U\Sigma U^\top$ by $U = (u_1,\ldots,u_n)$ and $\Sigma=\diag(\sigma_1,\ldots,\sigma_n)$.
    By \cref{alem:eq:diff_eig_value,alem:eq:diff_eig_vec,alem:eq:diff_mlog,alem:eq:diff_ln}, the differential of $\log_\alpha$ can be obtained.
\end{proof}
\subsection[Proof of \cref{alem:props:diff_mgexp_series}]{Proof of \cref{alem:props:diff_mgexp_series}}
\linkofproof{alem:props:diff_mgexp_series}
\begin{proof} [Proof of \cref{alem:props:diff_mgexp_series}]
    Following the notation in the proposition, we prove the result as follows.
    By abuse of notation, in the following, we omit the wide tilde $\widetilde{~}$.
   
    Now, we proceed to deal with the differential of $\log_\alpha^{-1}$.
    We rewrite the formula of $\log_\alpha^{-1}$ as
    \begin{align}
        & \log_\alpha^{-1}(X) \\
        &= U \diag\left(a_1^{\sigma_1},\cdots,a_n^{\sigma_n}\right) U^\top,\\
        &= U \diag\left(e^{\log(a_1)\sigma_1},\cdots,e^{\log(a_n)\sigma_n}\right)U^\top,\\
        &= U \diag\left(\sum_{k=0}^{\infty}\frac{\left(\log(a_1)\sigma_1\right)^k}{k!},\cdots,\sum_{k=0}^{\infty}\frac{\left(\log(a_n)\sigma_n\right)^k}{k!}\right)U^\top,\\
        \label{alem:eq:rw_mgexp_last2} &= U \left(\sum_{k=0}^{\infty}\frac{(B\Sigma)^k}{k!}\right) U^\top,\\
        \label{alem:eq:rw_mgexp} &= \sum_{k=0}^{\infty}\frac{(PX)^k}{k!}
    \end{align}
     where $P=UBU^\top$, $U$ is obtained from the eigendecomposition $X=U\Sigma U^\top$, and $B=\diag\left(\log(a_1),\cdots,\log(a_n)\right)$ is diagonal.
     By the properties of normed vector algebras \citep[Prop.~15.14]{loring2011introduction}, we can obtain the last equation.
     Then, we can compute the differential of $\log_\alpha^{-1}$ by curves.
     Given a curve $c$ on $\sym{n}$ starting at $X$ with initial velocity $V \in T_X\sym{n}$, write $c(t)=U(t)\Sigma(t)U(t)^\top$ and define $P(t)=U(t)BU(t)^\top$, so that $P(0)=P$. We have
     \begin{equation}
        \begin{aligned}
            \left(\log_\alpha^{-1}\right)_{*,X}(V)
            &= \left. \frac{d}{dt} \right |_{t=0} \log_\alpha^{-1}(c(t))\\
            \label{alem:eq:diff_mgexp_series_proof_stp2}
            &= \left. \frac{d}{dt} \right |_{t=0}\sum_{k=0}^{\infty}\frac{(P(t)c(t))^k}{k!}.
        \end{aligned} 
    \end{equation}
    Term-by-term differentiation gives
    \begin{equation} 
        \begin{aligned} \label{alem:eq:diff_last2step_mgexp}
            &\left(\log_\alpha^{-1}\right)_{*,X}(V)\\
            &= \sum_{k=1}^{\infty} \frac{1}{k !}(\sum_{l=0}^{k-1}(PX)^{k-l-1} \left. \frac{d}{dt} \right|_{t=0}(P(t)c(t)) (PX)^l).
        \end{aligned}
    \end{equation}
    By the chain rule, we have
    \begin{equation} \label{alem:eq:chain_rule_Pc}
        \left. \frac{d}{dt} \right|_{t=0}(P(t)c(t)) = P'(0)X + PV.
    \end{equation}
    $P'(0)$ is obtained by
    \begin{equation}
    \label{alem:eq:diff_P_at0}
        \begin{aligned}
            P'(0) 
            &= \left.\frac{d}{dt}\right|_{t=0}\left(U(t)BU(t)^\top\right)\\
            &= U'(0)BU^\top + UBU'(0)^\top\\
            &= D_U B U^\top + U B D_U^\top,
        \end{aligned}
    \end{equation}

    where $D_U$ is derived from the differential of eigenvectors,
    \begin{equation} \label{alem:eq:D_U}
    D_U = (\begin{array}{ccc}
             (\sigma_1 I_n-X)^+ V u_1 & \cdots & (\sigma_n I_n-X)^+ V u_n
        \end{array}).
    \end{equation}
    Substituting \cref{alem:eq:chain_rule_Pc,alem:eq:diff_P_at0,alem:eq:D_U} into \cref{alem:eq:diff_last2step_mgexp} yields the differential of $\log_\alpha^{-1}$.
\end{proof}

\subsection[Proof of \cref{alem:props:geo_mean_spd}]{Proof of \cref{alem:props:geo_mean_spd}}
\linkofproof{alem:props:geo_mean_spd}
\begin{proof} [Proof of \cref{alem:props:geo_mean_spd}]
    Obviously, the metric space $\{\spd{n},\dalem\}$ is isometric to the space $\sym{n}$ endowed with the standard Euclidean distance.
    Therefore, the weighted Fréchet mean of $\{S_i\}$ in $\spd{n}$ corresponds to the weighted Fréchet mean of associated points $\{\log_\alpha(S_i)\}$ in $\sym{n}$.
    The weighted Fréchet means in Euclidean spaces are clearly the familiar weighted means.
\end{proof}

\subsection[Proof of \cref{alem:props:biinvariance}]{Proof of \cref{alem:props:biinvariance}}
\linkofproof{alem:props:biinvariance}
\begin{proof} [Proof of \cref{alem:props:biinvariance}]
    As $\log_\alpha$ is a Riemannian isometry and $\sym{n}$ is bi-invariant, ALEM is therefore bi-invariant.
\end{proof}

\subsection[Proof of \cref{alem:props:exp_invariance}]{Proof of \cref{alem:props:exp_invariance}}
\linkofproof{alem:props:exp_invariance}
\begin{proof} [Proof of \cref{alem:props:exp_invariance}]
    Following the notation in this proposition, we proceed as follows.
    The right-hand side can be rewritten as
    \begin{equation}
        \begin{aligned}
            (\mathrm{FM}(S_1^\beta,\cdots S_m^\beta)) 
            &= \log_\alpha^{-1}\left(\sum_{i=1}^{m} \frac{1}{m}\beta\log_\alpha(S_i)\right)\\
            &= \log_\alpha^{-1}\left(\beta\sum_{i=1}^{m} \frac{1}{m}\log_\alpha(S_i)\right)\\
            &= \left[\log_\alpha^{-1}\left(\sum_{i=1}^{m} \frac{1}{m}\log_\alpha(S_i)\right)\right]^\beta\\
            &= (\mathrm{FM}(S_1,\cdots S_m))^\beta.
        \end{aligned}
    \end{equation}
\end{proof}

\subsection[Proof of \cref{alem:props:frechet_means_add_props}]{Proof of \cref{alem:props:frechet_means_add_props}}
\linkofproof{alem:props:frechet_means_add_props}
\begin{proof}[Proof of \cref{alem:props:frechet_means_add_props}]
    Recalling \cref{alem:eq:fm_alem}, Properties U1 and U2 obviously hold.
    
    When the SPD matrices $\{A_i\}_{i \leq n}$ commute, we have
    \begin{equation} \label{alem:eq:fm_commute_alem}
        \fm(\{A_i\})= \left(\prod_i A_i\right)^{\frac{1}{n}}.
    \end{equation}
    With \cref{alem:eq:fm_commute_alem}, Properties V1--V4 can be easily proved.
\end{proof}

\subsection[Proof of \cref{alem:props:sim_invariance}]{Proof of \cref{alem:props:sim_invariance}}
\linkofproof{alem:props:sim_invariance}
\begin{proof} [Proof of \cref{alem:props:sim_invariance}]
    Obviously, for a given SPD matrix $S$,
    \begin{align}
        \label{alem:eq:mlog_rotate} \log_\alpha(R S R^\top) &= R \log_\alpha(S) R^\top,\\
        \label{alem:eq:mlog_scale} \log_\alpha(s^2 S ) &= U\left(\log_\alpha(s^2 I_n)+\log_\alpha(\Sigma)\right)U^\top,
    \end{align}
    where $S=U \Sigma U^\top$ is the eigendecomposition.
    These identities yield the result.
\end{proof}

\subsection[Proof of \cref{alem:prop:rewrit_general_log}]{Proof of \cref{alem:prop:rewrit_general_log}}
\linkofproof{alem:prop:rewrit_general_log}
\begin{proof}[Proof of \cref{alem:prop:rewrit_general_log}]
    The three equations can be directly obtained.
\end{proof}

\subsection[Proof of \cref{alem:props:grad_mlog}]{Proof of \cref{alem:props:grad_mlog}}
\linkofproof{alem:props:grad_mlog}
\begin{proof} [Proof of \cref{alem:props:grad_mlog}]
    The input gradient follows from the Daleckii--Krein formula reviewed in \cref{eq:ch2-symmetric-matrix-function-differential,eq:ch2-loewner-matrix}.
    Now, let us focus on the gradient with respect to $A$.
    Differentiating both sides of \cref{alem:eq:rw_mul_mlog} gives
    \begin{equation}
        \diff X = (*) + U\left(\diff A \circledast \log(\Sigma)\right)U^\top,
    \end{equation}
    where $(*)$ denotes other terms involving $\diff U$ and $\diff \Sigma$.
    According to the invariance of the first-order differential form, we have
    \begin{align}
        & \nabla_{X} L : \diff X \\ 
        &= \nabla_{S} L:\diff S + \nabla_{X} L: U\left(\diff A \circledast \log(\Sigma)\right)U^\top \\
        \label{alem:eq:diif_last_row}
        &= \nabla_{S}L :\diff S + \left([U^\top (\nabla_{X} L) U] \circledast \log(\Sigma)\right):\diff A,
    \end{align}
    where $A:B=\tr(A^\top B)$ is the Euclidean Frobenius inner product.
    From the second term on the RHS of \cref{alem:eq:diif_last_row}, we can obtain the gradient with respect to $A$.
\end{proof}

\subsection[Proof of \cref{alem:props:grad_mexp}]{Proof of \cref{alem:props:grad_mexp}}
\linkofproof{alem:props:grad_mexp}
\begin{proof}[Proof of \cref{alem:props:grad_mexp}]
    The derivation follows the same logic as \cref{alem:props:grad_mlog}.
    We only need to show the derivation of \cref{alem:eq:gradient_mexp_wrt_A}.
    Similarly to \cref{alem:props:grad_mlog}, we have the following:
    \begin{equation}
        \diff X = (*) + U\left[\diff A \circledast \left(\diag\left(a_1^{\Sigma_{11}},\cdots,a_n^{\Sigma_{nn}}\right) \frac{-\Sigma}{A^2} \right)\right]U^\top,
    \end{equation}
    \begin{align}
    &\nabla_{X} L : \diff X \\
    &= \nabla_{S}L :\diff S + \left\{[U^\top (\nabla_{X} L) U] \circledast \left(\diag\left(a_1^{\Sigma_{11}},\cdots,a_n^{\Sigma_{nn}}\right) \frac{-\Sigma}{A^2} \right)\right\}:\diff A.
    \end{align}    
\end{proof}

\subsection[Proof of \cref{alem:thm:gyro_mlr_alem}]{Proof of \cref{alem:thm:gyro_mlr_alem}}
\linkofproof{alem:thm:gyro_mlr_alem}
\begin{proof}[Proof of \cref{alem:thm:gyro_mlr_alem}]
    
    Following \citet{nguyen2022gyro,nguyen2023building}, we first define gyrostructures under ALEM:
    \begin{align}
        \label{alem:eq:gyro_addtion}
        P \oplusale Q &= \rieexp_{P}\left(\pt{I_n}{P} \left(\rielog _{I_n}(Q)\right)\right), \\
        \label{alem:eq:gyro_automorphism}
        \gyr[P, Q] R &= (\ominus(P \oplusale Q)) \oplusale(P \oplusale(Q \oplusale R)),\\
        \label{alem:eq:gyro_scalar_product}
        t \odotale P &= \rieexp_{I_n}\left(t \rielog _{I_n}(P)\right), \\
        \label{alem:eq:gyro_inverse}
        \ominus P &= -1 \odotale P = \rieexp_{I_n}\left(- \rielog _{I_n}(P)\right), \\
        \label{alem:eq:gyro_inner_product}
        \gyrinner{P}{Q}&=\left\langle \rielog _{I_n} (P), \rielog _{I_n} (Q)\right\rangle_{I_n},\\
        \label{alem:eq:gyro_norm}
        \gyrnorm{P} &= \sqrt{\gyrinner{P}{P}},\\
        \label{alem:eq:gyro_distance}
        \gyrdist(P, Q) &= \gyrnorm{\ominus P \oplusale Q},
    \end{align}
    where $P,Q,R \in \spd{n}$, and $I_n$ is the identity matrix.
    The above operations are called gyroaddition, gyroautomorphism, scalar gyromultiplication, gyroinverse, gyroinner product, gyronorm, and gyrodistance. Simple computations show that \cref{alem:eq:gyro_addtion} and \cref{alem:eq:gyro_scalar_product} are exactly $\oplusale$ and $\odotale$ in \cref{alem:thm:mlog_spd_properties}.
    As indicated by \cref{alem:thm:mlog_spd_properties}, $\{\spd{n}, \oplusale, \odotale \}$ forms a gyrovector space in the sense of \cref{def:ch2-gyrovector-space}.
    In the following proof, we use $\odotale$ and $\oplusale$.

    The gyro MLR \citep{nguyen2023building} under ALEM is defined as 
    \begin{equation}
        \label{alem:eq:mlr_alem}
        \begin{aligned}
        &p(y=k \mid S) \\
        &\propto \exp \left(\operatorname{sign}(\langle \tilde{A}_k, \rielog_{P_k}(S) \rangle_{P_k})\|\tilde{A}_k\|_{P_k} \bar{d} (S, H_{\tilde{A}_k, P_k}) \right),\\
    \end{aligned}
    \end{equation}
    where $P_k \in \spd{n}$ and $\tilde{A}_k \in T_{P_k} \spd{n}$.
    $\bar{d} (S, H_{\tilde{A}_k, P_k})$ is the margin distance to the SPD hyperplane $H_{\tilde{A}_k, P_k}$, which is defined as
    \begin{align}
        \label{alem:eq:gyro_dist_hyperplane}
        \bar{d} (S, H_{\tilde{A}_k, P_k}) &=\sin (\angle S P_k Q^*) \gyrdist(S,P_k), \\
        Q^* &=\underset{Q \in H_{\tilde{A}_k, P_k} \backslash \{P_k\}}{\argmax}\left(\cos (\angle S P_k Q)\right), \\
        \label{alem:eq:gyro_consie}
        \cos (\angle S P_k Q) &= \frac{\gyrinner{ \ominus P_k \oplusale Q}{\ominus P_k \oplusale S}}{\gyrnorm{ \ominus P_k \oplusale Q} \gyrnorm{ \ominus P_k \oplusale S }},\\
        \label{alem:eq:gyro_hyperplane}
        H_{\tilde{A}_k, P_k} &= \{S \in \spd{n} \mid \langle \rielog_{P_k} S, \tilde{A}_k \rangle_{P_k} =0\}.
    \end{align}
    \cref{alem:eq:gyro_dist_hyperplane,alem:eq:gyro_consie,alem:eq:gyro_hyperplane} are called the SPD pseudo-gyrodistance, SPD gyrocosine, and SPD gyrohyperplane.

    For simplicity, we further omit the subscript $k$ in $P_k$ and $\tilde{A}_k$.
    \cref{alem:eq:gyro_hyperplane} can be simplified:
    \begin{equation}
        \label{alem:eq:alem_gyro_hyperplane_simplified}
        \begin{aligned}
            &\langle \rielog_{P} S, \tilde{A} \rangle_{P}\\
            &\stackrel{(1)}{=} \left\langle \left(\log_\alpha^{-1}\right)_{*,\log_\alpha(P)}\left(\log_\alpha(S)-\log_\alpha(P)\right), \tilde{A} \right\rangle_{P} \\
            &\stackrel{(2)}{=} \left\langle \left(\log_\alpha\right)_{*,P} \circ \left(\log_\alpha^{-1}\right)_{*,\log_\alpha(P)}\left(\log_\alpha(S)-\log_\alpha(P)\right), \left(\log_\alpha\right)_{*,P}(\tilde{A}) \right\rangle \\
            &= \left\langle \log_\alpha(S)-\log_\alpha(P), \left(\log_\alpha\right)_{*,P}(\tilde{A}) \right\rangle.
        \end{aligned}
    \end{equation}
    The above derivation comes from the following.
    \begin{enumerate}
        \item
        \cref{alem:eq:rielog_gmlog}.
        \item 
        The definition of ALEM.
    \end{enumerate}

    Similarly, a simple computation shows that \cref{alem:eq:gyro_consie} can also be simplified as
    \begin{equation} \label{alem:eq:alem_gyro_consine_simplified}
        \frac{\left\langle -\log_\alpha(P)+\log_\alpha(Q), -\log_\alpha(P)+\log_\alpha(S) \right\rangle}{\left\|-\log_\alpha(P)+\log_\alpha(Q)\right\|_\rmF \left\|-\log_\alpha(P)+\log_\alpha(S)\right\|_\rmF}.
    \end{equation}

    Together with \cref{alem:eq:alem_gyro_hyperplane_simplified,alem:eq:alem_gyro_consine_simplified}, \cref{alem:eq:gyro_dist_hyperplane} is equivalent to the distance to the hyperplane in the Euclidean space.
    Therefore, \cref{alem:eq:gyro_dist_hyperplane} has a closed-form solution:
    \begin{equation}
    \label{alem:eq:dist_hyperplane_final}
    \begin{aligned}
        &\bar{d} (S, H_{\tilde{A}, P})\\
        &= \frac{\left|\left\langle \log_\alpha(S)-\log_\alpha(P), \bar{A} \right\rangle\right|}{\left\|\bar{A}\right\|_\rmF}\\
        &= \frac{\left|\left\langle \log_\alpha(S)-\log_\alpha(P), \bar{A} \right\rangle\right|}{\left\|\tilde{A}\right\|_P},
    \end{aligned}
    \end{equation}
    where $\bar{A}=\left(\log_\alpha\right)_{*,P}(\tilde{A})$.
    Substituting \cref{alem:eq:dist_hyperplane_final} into \cref{alem:eq:mlr_alem} yields the claimed result.
\end{proof}

\subsection[Proof of \cref{alem:propos:param_by_geom}]{Proof of \cref{alem:propos:param_by_geom}}
\linkofproof{alem:propos:param_by_geom}
\begin{proof} [Proof of \cref{alem:propos:param_by_geom}]
    To derive the ALEM-specific positive-scalar update, consider the general RSGD update reviewed in \cref{sec:ch2-riemannian-optimization}. For a minimization parameter $w$ on an $n$-dimensional smooth connected Riemannian manifold $\calM$, we have
    \begin{equation} \label{alem:eq:riem_sgd}
        w^{(t+1)}=\rieExp{w^{(t)}} (-\gamma^{(t)} \pi_{w^{(t)}}(\nabla_{w^{(t)}} L )),
    \end{equation} 
    where $\rieExp{w}(\cdot): T_w\calM \rightarrow \calM$ is the Riemannian exponential map, which maps a tangent vector at $w$ back into the manifold $\calM$, and $\pi_{w}(\cdot): \bbR{n} \rightarrow T_w\calM$ is the projection operator, projecting an ambient Euclidean vector into the tangent space at $w$.
    In the case of the SPD manifold, for all $S \in \spd{n}$, $X \in \bbR{n \times n}$, and $V \in \sym{n}$, the exponential map and projection operator are formulated as follows:
    \begin{align}
        \label{alem:eq:spd_proj} \pi_{S}(X) &= S \frac{X+X^\top}{2}S,\\
        \label{alem:eq:spd_exp} \rieExp{S}(V) &= S^{1/2} \exp(S^{-1/2} V S^{-1/2}) S^{1/2},
    \end{align}
    where $\exp(\cdot)$ is the matrix exponential.
    For more details about \cref{alem:eq:spd_proj} and \cref{alem:eq:spd_exp}, see \citet{yger2013review} and \citet{amari2016information}.
    Substituting \cref{alem:eq:spd_proj} and \cref{alem:eq:spd_exp} into \cref{alem:eq:riem_sgd} immediately yields \cref{alem:eq:update_pos_scalar}.
\end{proof}
\subsection[Proof of \cref{alem:props:geom_equivalent_div}]{Proof of \cref{alem:props:geom_equivalent_div}}
\linkofproof{alem:props:geom_equivalent_div}
\begin{proof} [Proof of \cref{alem:props:geom_equivalent_div}]
    Without loss of generality, we focus on the equivalence between $b=B_{11}$ and $a=a_1$.
    Note that $b$ is essentially expressed as $b=\log(a)$.
    Suppose $b^{(t)} = \log\left(a^{(t)}\right)$. Then, we have
    \begin{equation}
        \begin{aligned}
            \nabla_{a^{(t)}} L 
            &= \nabla_{b^{(t)}} L \left.\frac{\partial \log(a)}{\partial a}\right|_{a^{(t)}}\\
            &= \nabla_{b^{(t)}} L \frac{1}{a^{(t)}}.
        \end{aligned}
    \end{equation}

    By \cref{alem:eq:update_pos_scalar}, $\log\left(a^{(t+1)}\right)$ is
    \begin{equation}
        \begin{aligned}
            \log\left(a^{(t+1)}\right)
            &= \log\left(a^{(t)}e^{-\gamma^{(t)} a^{(t)} \nabla_{a^{(t)}} L}\right)\\
            &= \log\left(a^{(t)}\right) -\gamma^{(t)} a^{(t)} \nabla_{a^{(t)}} L\\
            &= \log\left(a^{(t)}\right) - \gamma^{(t)} a^{(t)} \left(\nabla_{b^{(t)}} L/a^{(t)}\right) \\
            &= \log\left(a^{(t)}\right) -\gamma^{(t)}\nabla_{b^{(t)}} L\\
            &= b^{(t)} -\gamma^{(t)}\nabla_{b^{(t)}} L.
        \end{aligned}
    \end{equation}
    The last row is the ESGD update formula for $b$.
    
    Therefore, if $b^{(0)} = \log\left(a^{(0)}\right)$, the two optimization procedures yield equivalent iterates throughout training.
\end{proof}

\section{Product Cholesky Metrics}
\label{app:pcm-proofs}

\linkofproof{pcm:thm:dpem}
\subsection[Proof of \cref{pcm:thm:dpem}]{Proof of \cref{pcm:thm:dpem}}

\begin{proof}
    Since $\defDPM$ is the product metric of $\{\LTzero{n},\gE \}$ and $n$ copies of $\{\bbRplusscalar, \gPE\}$, we first show the Riemannian operators on $\{\bbRplusscalar, \gPE\}$. We can then readily obtain the Riemannian operators on $\{\chospace{n}, \gdefDE \}$ by the principles of product metrics.

    As shown by \citet{thanwerdas2022geometry}, $\gPE$ is the pullback metric of $\gE$ by the power function $\pow_{\theta}(\cdot)$ and scaled by $\frac{1}{\theta^2}$, expressed as $\gPE = \frac{1}{\theta^2} \pow_{\theta}^* \gE$.
    Besides, as constant scaling does not change the Christoffel symbols, the geodesic, Riemannian logarithm and exponential maps, and parallel transport along a geodesic remain the same under $\gPE$ and $\pow_{\theta}^* \gE$.
    These Riemannian operators under $\pow_{\theta}^* \gE$ can be obtained by the properties of Riemannian isometries (\cref{def:ch2-riemannian-isometry}).
    Specifically, given $p,q \in \bbRplusscalar$ and $w,v \in T_p\bbRplusscalar$, we have the following:
    \begin{align}
        \left(\pow_\theta\right)_{*,p}(v) &= \theta p^{\theta-1} v,\\
        \gPE_p(v,w)
        &= \frac{1}{\theta^2} \gE\left(\left(\pow_\theta\right)_{*,p}(v),\left(\pow_\theta\right)_{*,p}(w)\right)
        = \langle p^{\theta-1} v,p^{\theta-1} w \rangle
        = p^{2(\theta-1)} vw,\\
        \gamma_{(p,v)}(t) 
        &= \pow_{\theta}^{-1} \left(\pow_{\theta}(p) + t\left(\pow_\theta\right)_{*,p}(v) \right)\\
        &= (p^{\theta}+ t\theta p^{\theta-1}v )^{\frac{1}{\theta}}\\
        &= p(1 + t\theta p^{-1}v )^{\frac{1}{\theta}}, 
        \text{with } t \in \{t \in \bbRscalar \mid 1 + t\theta p ^{-1} v \in \bbRplusscalar \},\\
        \rielog_{p}(q) 
        &= \left(\pow_\theta\right)_{*,p}^{-1} \left(\pow_{\theta}(q) - \pow_{\theta}(p) \right) \\
        &= \frac{1}{\theta}p^{1-\theta} (q^\theta - p^\theta)
        = \frac{1}{\theta}p \left( \left (\frac{q}{p}\right)^{\theta} -1\right),\\
        \pt{p}{q}(v) 
        &= \left(\pow_\theta\right)_{*,q}^{-1}\left(\left(\pow_\theta\right)_{*,p}(v)\right) = \left ( \frac{q}{p} \right)^{1-\theta}v.
    \end{align}
    The geodesic distance between $p$ and $q$ under $\gPE$ is given by
    \begin{equation}
            \dist^2(p,q) 
            = \gPE_p(\rielog_p(q),\rielog_p(q))
            = \frac{1}{\theta^2} (q^\theta-p^\theta)^2.
    \end{equation}
    The weighted Fréchet mean (WFM) of $\{p_i \in \bbRplusscalar\}_{i=1}^N$ with weights $\{w_i\}_{i=1}^N$ satisfying $w_i>0$ for all $i$ and $\sum_i w_i=1$ under $\gPE$ is defined as
    \begin{equation}
        \wfm(\{w_i\},\{p_i\}) = \underset{p \in \bbRplusscalar}{\argmin} \sum_{i=1}^N w_i \dist^2(p,p_i).
    \end{equation}
    Obviously, the WFM of $\{p_i\}$ under $\gPE$ is the same as the one under $\pow_{\theta}^* \gE$.    
    Due to the isometry of $\pow_{\theta}^* \gE$ to $\gE$, the WFM of $\{p_i\}$ under $\gPE$ can be calculated as
    \begin{equation}
        \label{pcm:eq:defdem_wfm_starting}
        \wfm(\{w_i\},\{p_i\})
        = \pow_{\theta}^{-1} \left ( \wfm^{\rmE}(\{w_i\},\{\pow_{\theta}(p_i)\}) \right)
        = \left ( \sum\nolimits_{i=1}^N w_i p_i^\theta  \right) ^{\frac{1}{\theta}},
    \end{equation}
    where $\wfm^{\rmE}$ in \cref{pcm:eq:defdem_wfm_starting} is the Euclidean WFM, which is the familiar weighted average.

    So far, we have obtained all the necessary Riemannian operators on $\{\bbRplusscalar, \gPE\}$.
    Combining these results with the Euclidean space $\LTzero{n}$ yields the results in the theorem.
\end{proof}

\linkofproof{pcm:thm:dgbwm}
\subsection[Proof of \cref{pcm:thm:dgbwm}]{Proof of \cref{pcm:thm:dgbwm}}
\begin{proof}
    As in the proof of \cref{pcm:thm:dpem}, we only need to show the Riemannian operators on $\{\bbRplusscalar, \gGBWscalar\}$ with $m \in \bbRplusscalar$.
    Expressions for the Riemannian operators under GBWM can be found in \citet{han2023learning}. Here, we further simplify the associated expressions for the one-dimensional case.
    Specifically, given $p,q \in \bbRplusscalar$ and $w,v \in T_p\bbRplusscalar$, we have the following:
    \begin{align}
        \lyp_{p,m}(v) 
        &= \frac{v}{2mp},\\
        \gGBWscalar_{p}(v,w) 
        &= \frac{1}{2}\langle \lyp_{p,m}(v),w \rangle = \frac{vw}{4mp},\\
        \gamma_{(p,v)}(t) 
        &= p + tv + \lyp_{p,m}(tv)^2 m^2 p  
        = p + tv + \frac{(tv)^2}{4p}  
        = p(1+\frac{t}{2}\frac{v}{p})^2,\\
        \rielog_{p}(q) 
        &= 2\left(m(m^{-2}pq)^{\frac{1}{2}}-p \right) 
        = 2\left( \left(pq \right)^{\frac{1}{2}}-p \right) 
        = 2p\left( \left (\frac{q}{p} \right)^{\frac{1}{2}} -1 \right),
    \end{align}
    \begin{equation}
        \begin{aligned}
        \dist^2(p,q) 
        = \gGBWscalar_p(\rielog_p(q),\rielog_p(q))
        &= \frac{1}{4mp} 4\left( \left(pq \right)^{\frac{1}{2}}-p \right)^2\\
        &= \frac{1}{mp} \left( pq - 2 p^\frac{3}{2}q^\frac{1}{2} + p^2 \right)\\
        &= m^{-1} \left( q - 2 p^\frac{1}{2}q^\frac{1}{2} + p \right)\\
        &= \left( m^{-\frac{1}{2}} \left( q^\frac{1}{2}-p^\frac{1}{2} \right) \right)^2.
        \end{aligned}
    \end{equation}
    
    As shown by \citet{han2023learning}, GBWM on $\spd{n}$ is the pullback metric of BWM by $\pi(S)=M^{-\frac{1}{2}} S M^{-\frac{1}{2}}$ for all $S \in \spd{n}$ with $M \in \spd{n}$.
    For the specific $\bbRplusscalar \cong \spd{1}$, the isometry is simplified as
    \begin{equation}
        \pi(p)=m^{-1} p, \forall p \in \bbRplusscalar \text{ with } m \in \bbRplusscalar.
    \end{equation}
    
    The geodesic $\widetilde{\gamma}_{(p,v)}(t)$ under BWM on $\bbRplusscalar$ exists in the interval $\{t \in \bbRscalar \mid 1+ t \lyp_{p}(v) \in \bbRplusscalar \}$ \citep{malago2018wasserstein}, which can be simplified as $\left \{t \in \bbRscalar \mid 1+ t \frac{v}{2p} \in \bbRplusscalar \right\}$.
    Therefore, the geodesic $\gamma_{(p,v)}(t)$ under $\gGBWscalar$ exists in the interval:
    \begin{equation}
        \left \{t \in \bbRscalar \mid 1+ t \frac{\pi_{*,p}(v)}{2\pi(p)} \in \bbRplusscalar \right\} = \left \{t \in \bbRscalar \mid 1+ t \frac{v}{2p} \in \bbRplusscalar \right\}.
    \end{equation}
    
    For BWM, the parallel transport is \citep[Tab.~6]{thanwerdas2023n}
    \begin{equation}
        \tildept{p}{q}(v) = \left( \frac{q}{p} \right)^{\frac{1}{2}} v.
    \end{equation}
    Therefore, the parallel transport on $\{\bbRplusscalar,\gGBWscalar\}$ is
    \begin{equation}
        \begin{aligned}
             \pt{p}{q}(v) 
            = \pi_{*,q}^{-1} \left(\tildept{\pi(p)}{\pi(q)}(\pi_{*,p}(v)) \right)
            &= \pi_{*,q}^{-1} \left( \left(\frac{\pi(q)}{\pi(p)} \right)^{\frac{1}{2}} \pi_{*,p}(v) \right) \\
            &= \left( \frac{q}{p} \right)^{\frac{1}{2}} v.
        \end{aligned}
    \end{equation}
    
    Finally, we show the WFM on $\{\bbRplusscalar,\gGBWscalar\}$.
    Given $\{p_i \in \bbRplusscalar \}_{i=1}^N$ with weights $\{w_i\}_{i=1}^N$ satisfying $w_i>0$ for all $i$ and $\sum_{i=1}^N w_i=1$, the WFM on $\{\bbRplusscalar,\gGBWscalar\}$ is
    \begin{equation}
        \begin{aligned}
            \wfm(\{w_i\},\{p_i\})
            &= \underset{p \in \bbRplusscalar}{\argmin} \sum_{i=1}^N w_i \dist^2(p,p_i)\\
            &=\underset{p \in \bbRplusscalar}{\argmin} \sum_{i=1}^N w_i m^{-1} \left( p^\frac{1}{2}-p_i^\frac{1}{2} \right)^2\\
            &=\underset{p \in \bbRplusscalar}{\argmin} \sum_{i=1}^N w_i \left( p^\frac{1}{2}-p_i^\frac{1}{2} \right)^2\\
            &=\underset{p \in \bbRplusscalar}{\argmin} \sum_{i=1}^N w_i \left( p- 2 p^{\frac{1}{2}} p_i^\frac{1}{2} \right)\\
            &=\underset{p \in \bbRplusscalar}{\argmin} \quad p - 2 p^{\frac{1}{2}}\sum_{i=1}^N w_i  p_i^\frac{1}{2}.
        \end{aligned}
    \end{equation}
    Let $f(p)=p - 2 p^{\frac{1}{2}} \sum_{i=1}^N w_i p_i^\frac{1}{2}$.
    Then, the first- and second-order derivatives are
    \begin{align}
        \label{pcm:eq:1st_derivatives}
        \frac{\diff f}{\diff p} 
        &= 1 - p^{-\frac{1}{2}}\sum\nolimits_i w_i p_i^\frac{1}{2},\\
        \frac{\diff^2 f}{\diff p^2} 
        &= \frac{1}{2} p^{-\frac{3}{2}} \sum\nolimits_i w_i {p_i}^\frac{1}{2}  > 0, \quad \forall p \in \bbRplusscalar.
    \end{align}
    Therefore, the optimal solution can be obtained by setting \cref{pcm:eq:1st_derivatives} equal to 0:
    \begin{equation}
        \frac{\diff f}{\diff p} = 0 \Rightarrow p^*=  \left(\sum\nolimits_i w_i p_i^\frac{1}{2}\right)^2.
    \end{equation}

    Combining the above results with the Euclidean geometry on $\LTzero{n}$, one can readily obtain the results.
\end{proof}

\linkofproof{pcm:lem:dpdm_exp_and_limits}
\subsection[Proof of \cref{pcm:lem:dpdm_exp_and_limits}]{Proof of \cref{pcm:lem:dpdm_exp_and_limits}}

\begin{proof}
    The differential of $\dpow_{\theta}$ at $\bbL \in \bbDplus{n}$ is 
    \begin{equation}\label{pcm:eq:diff_diag_pow_deform}
        \left(\dpow_\theta\right)_{*,\bbL}(\bbV) = \theta \bbL^{\theta-1} \bbV, \quad \forall \bbV \in T_{\bbL}\bbDplus{n}.
    \end{equation}
    Substituting \cref{pcm:eq:diff_diag_pow_deform} into \cref{pcm:def:dpdm} yields the result.
\end{proof}

\linkofproof{pcm:lem:gyro_cholesky}
\subsection[Proof of \cref{pcm:lem:gyro_cholesky}]{Proof of \cref{pcm:lem:gyro_cholesky}}

We first present a useful lemma.

\begin{parislemma}\label{pcm:lem:gyro_dgbwm_sqrt_defdem}
    The Riemannian exponential and logarithmic maps and parallel transport along the geodesic are the same under $\defDGBWM$ and $\nicefrac{\theta}{2}$-DPM.
\end{parislemma}
\begin{proof}
     This follows directly from \cref{pcm:thm:dpem,pcm:thm:dgbwm}.
\end{proof}

Now we begin to prove \cref{pcm:lem:gyro_cholesky}.

\begin{proof}
    We first derive the expressions for DPM with a generic nonzero parameter $\theta$. According to \cref{pcm:lem:gyro_dgbwm_sqrt_defdem}, the expressions for $\defDGBWM$ then follow by replacing $\theta$ with $\nicefrac{\theta}{2}$. We omit the superscript $\mathcal{C}$ for simplicity.
    
    For $X \in T_{I_n}\chospace{n}$, we have the following:
    \begin{align}
        \label{pcm:eq:rielog_i_defdem}
        \rielog_{I_n}(L)
        &= \trilL + \frac{1}{\theta}\left(\bbL^\theta-I_n \right),\\
        \label{pcm:eq:pt_from_i_defdem}
        \pt{I_n}{L}(X)
        &= \trilX + \bbL^{1-\theta}\bbX ,\\
        \label{pcm:eq:rieexp_i_defdem}
        \rieexp_{I_n} X
        &=\trilX + \left(I_n + \theta \bbX \right)^{\frac{1}{\theta}}.
    \end{align}
    For the binary operation, substituting \cref{pcm:eq:rielog_i_defdem,pcm:eq:pt_from_i_defdem} into \cref{eq:ch2-riem-gyro-addition} gives
    \begin{equation}
        \label{pcm:eq:prf_exp_L_condition}
        \begin{aligned}
            L \oplus K 
            &= \rieexp_{L}\left(\pt{I_n}{L} \left(\rielog_{I_n}(K) \right) \right) \\
            &= \rieexp_{L}\left( \pt{I_n}{L} \left( \trilK + \frac{1}{\theta} \left(\bbK^\theta - I_n \right) \right) \right) \\
            &= \rieexp_{L}\left( \trilK + \frac{1}{\theta} \bbL^{1-\theta} \left(\bbK^\theta - I_n \right)  \right) \\
            &= \trilL + \trilK + \bbL \left(I_n + \theta \bbL^{-1} \left( \frac{1}{\theta} \bbL^{1-\theta} \left(\bbK^\theta - I_n \right)   \right) \right)^{\frac{1}{\theta}} \\
            &= \trilL + \trilK + \left( \bbL^\theta + \bbK^\theta -I_n \right)^{\frac{1}{\theta}}.
        \end{aligned}
    \end{equation}
    In the third row of \cref{pcm:eq:prf_exp_L_condition}, the well-definedness of the exponential map requires
    \begin{align}
        \bbL + \theta \left[ \frac{1}{\theta} \bbL^{1-\theta} \left(\bbK^\theta - I_n \right) \right] \in \bbDplus{n} \Leftrightarrow \bbL^\theta + \bbK^\theta -I_n \in \bbDplus{n}.
    \end{align}

    For the gyromultiplication, substituting \cref{pcm:eq:rielog_i_defdem,pcm:eq:rieexp_i_defdem} into \cref{eq:ch2-riem-gyro-scalar} gives
    \begin{equation} \label{pcm:eq:prf_exp_i_condition}
        \begin{aligned}
            t \odot L 
            &= \rieexp_{I_n}(t\rielog_{I_n} (L)) \\
            &= \rieexp_{I_n} \left(t \trilL + \frac{t}{\theta}\left(\bbL^\theta-I_n \right) \right) \\
            &= t \trilL + \left( I_n + t\left(\bbL^\theta-I_n \right) \right)^\frac{1}{\theta}\\
            &= t \trilL + \left(t\bbL^\theta + (1-t)I_n \right)^\frac{1}{\theta}.
        \end{aligned}
    \end{equation}
    In the second row of \cref{pcm:eq:prf_exp_i_condition}, the exponential map requires
    \begin{equation}
        I_n + \theta \left(\frac{t}{\theta}\left(\bbL^\theta-I_n \right) \right) \in \bbDplus{n} \Leftrightarrow t\bbL^\theta + (1-t)I_n \in \bbDplus{n}.
    \end{equation}
\end{proof}

\linkofproof{pcm:thm:gyro_spaces_defdem}
\subsection[Proof of \cref{pcm:thm:gyro_spaces_defdem}]{Proof of \cref{pcm:thm:gyro_spaces_defdem}}

\begin{proof}
    In this proof, we assume $L,K,J \in \chospace{n}$ and $s,t \in \bbRscalar$, with all gyro operations satisfying the conditions in \cref{pcm:lem:gyro_cholesky}. By \cref{pcm:lem:gyro_dgbwm_sqrt_defdem}, we only need to prove the case of $\defDPM$. For simplicity, we omit the superscript $\mathcal{C}$.
    
    \mypara{Axiom (G1).} \cref{eq:ch2-riem-gyro-addition} implies that the identity element is the identity matrix.

    \mypara{Axiom (G2).} We define the inverse element of $L$ as
    \begin{equation} \label{pcm:eq:inv_gyro_defdem}
        \ominus L = -1 \odot L
        = -\trilL + \left(2I_n - \bbL^\theta \right)^\frac{1}{\theta}.
    \end{equation}
    Simple computations show that $\ominus L \oplus L =I_n$.

    \mypara{Axiom (G3).} Gyroaddition in \cref{pcm:lem:gyro_cholesky} indicates that
    \begin{equation}
        L \oplus (K \oplus J) 
        = (L \oplus K) \oplus J
        = \trilL + \trilK + \trilJ + \left( \bbL^\theta + \bbK^\theta + \bbJ^\theta - 2I_n \right)^\frac{1}{\theta}.
    \end{equation}
    Therefore the gyroautomorphism is the identity map, \ie $\gyr[L,K]=\id$.

    \mypara{Axiom (G4).} This is a direct corollary of (G3).

    \mypara{Gyrocommutative law.} Gyroaddition in \cref{pcm:lem:gyro_cholesky} indicates that
    \begin{equation}
        L \oplus K = K \oplus L.
    \end{equation}

    \mypara{Axiom (V1).} This follows from gyromultiplication in \cref{pcm:lem:gyro_cholesky} and \cref{pcm:eq:inv_gyro_defdem}.

    \mypara{Axiom (V2).}
    \begin{equation}
        \begin{aligned}
            (s+t) \odot L 
            &= (s+t)\trilL + \left((s+t)\bbL^\theta + (1-(s+t))I_n \right)^\frac{1}{\theta} \\
            &= s \trilL + t\trilL + \left( s\bbL^\theta + (1-s)I_n + t\bbL^\theta + (1-t)I_n -I_n \right)^\frac{1}{\theta} \\
            &= (s \odot L) \oplus (t \odot L).
        \end{aligned}
    \end{equation}

    \mypara{Axiom (V3).}
    \begin{equation}
        \begin{aligned}
            (st) \odot L 
            &= (st)\trilL + \left(st\bbL^\theta + (1-st)I_n \right)^\frac{1}{\theta} \\
            &= (st)\trilL + \left(s \left(t\bbL^\theta + (1-t)I_n \right) + (1-s)I_n \right)^\frac{1}{\theta} \\
            &= s \odot \left[ t\trilL + \left(t\bbL^\theta + (1-t)I_n \right)^\frac{1}{\theta} \right] \\
            &= s \odot (t \odot L).
        \end{aligned}
    \end{equation}
    
    \mypara{Axioms (V4) and (V5).} These two axioms can be directly obtained, as gyroautomorphisms are all identity maps.
\end{proof}

\linkofproof{pcm:thm:gyro_spaces_spd}
\subsection[Proof of \cref{pcm:thm:gyro_spaces_spd}]{Proof of \cref{pcm:thm:gyro_spaces_spd}}

\begin{proof}
    According to \citet[Thm.~2.4]{nguyen2023building}, gyrovector operations are preserved under Riemannian isometries. Moreover, the Cholesky decomposition is a Riemannian isometry:
    \begin{equation}
        \chol: \{\spd{n}, g ^\mathcal{S}\} \rightarrow \{\chospace{n}, g ^\mathcal{C}\}.
    \end{equation}
\end{proof}

\linkofproof{pcm:thm:spd_mlrs}
\subsection[Proof of \cref{pcm:thm:spd_mlrs}]{Proof of \cref{pcm:thm:spd_mlrs}}
\begin{proof}
    Substituting the associated operators in \cref{pcm:sec:spd-geometries} into \cref{rmlr:lem:rmlr_pullback} yields the results.
\end{proof}

\end{document}